\documentclass[12pt,a4paper]{book}
\usepackage[utf8]{inputenc}
\usepackage[T1]{fontenc}

\usepackage[english]{babel}
\usepackage{csquotes}
\usepackage{hyphenat}

\usepackage{graphicx}

\usepackage[acronym]{glossaries}

\graphicspath{ {./images/} }
\usepackage[table]{xcolor}
\usepackage{array}
\usepackage{multirow}
\usepackage{booktabs}

\usepackage{amsmath}
\usepackage{amssymb}
\DeclareMathOperator*{\argmax}{arg\,max}
\DeclareMathOperator*{\argmin}{arg\,min}

\usepackage{listings,chngcntr,verbatim}
\usepackage[framed,numbered,autolinebreaks,useliterate]{mcode}		

\usepackage{natbib}

\usepackage[acronym]{glossaries}

\makeglossaries
 
\newacronym{gcd}{GCD}{Greatest Common Divisor} 
\newacronym{lcm}{LCM}{Least Common multiple}

\newacronym{olhd}{OLHD}{Optimal Latin Hypercube Design}
\newacronym{svm}{SVM}{Support Vector Machines}
\newacronym{pr}{PR}{Polynomial Regression}
\newacronym{rbf}{RBFN}{Radial Basis Function Network}
\newacronym{nn}{NN}{Neural Networks}
\newacronym{mae}{MAE}{Mean Absolute Error}
\newacronym{lf}{LF}{Low-Fidelity}
\newacronym{hf}{HF}{High-Fidelity}
\newacronym{doe}{DOE}{Design of experiments}
\newacronym{ok}{OK}{Ordinary Kriging}
\newacronym{uk}{UK}{Universal Kriging}
\newacronym{hk}{HK}{Hierarchical Kriging}
\newacronym{pls}{PLS}{Partial Least Squares}
\newacronym{plsok}{PLSOK}{Partial Least Squares Ordinary Kriging}
\newacronym{plshk}{PLSHK}{Partial Least Squares Hierarchical Kriging}
\newacronym{mipt}{MIPT}{MC-intersite-proj-th }
\newacronym{msd}{MSD}{Maximin Scaled Distance}
\newacronym{cv}{CV}{Cross-Validation}
\newacronym{cvv}{CVV}{Cross-Validation Variance}
\newacronym{cvvor}{CVVOR}{Cross-Validation-Voronoi}
\newacronym{cdm}{CDM}{Crowding Distance Metric}
\newacronym{ssa}{SSA}{Smart Sampling Algorithm}
\newacronym{gcv}{GCV}{Generalized Cross-Validation}
\newacronym{mse}{MSE}{Mean-Squared Error}
\newacronym{mmse}{MMSE}{Maximum Mean-Squared Error}
\newacronym{imse}{IMSE}{Integrated Mean-Squared Error}
\newacronym{ame}{AME}{Adaptive Maximum Entropy}
\newacronym{ei}{EI}{Expected Improvement}
\newacronym{cdf}{CDF}{Cumulative Distribution Function}
\newacronym{pdf}{PDF}{Probability Distribution Function}
\newacronym{wei}{WEI}{Weighted Expected Improvement}
\newacronym{awei}{AWEI}{Adaptive Weighted Expected Improvement}
\newacronym{eigf}{EIGF}{Expected Improvement for Global Fit}
\newacronym{haed}{HAED}{Hierarchical Adaptive Experimental Design}
\newacronym{gek}{GEK}{Gradient-Enhanced Kriging}
\newacronym{nurbs}{NURBS}{Non-Uniform Rational B-Spline}
\newacronym{lola}{LOLA}{Local Linear Approximation}
\newacronym{qbc}{QBC}{Query-By-Committee}
\newacronym{masa}{MASA}{Mixed Adaptive Sampling Algorithm}
\newacronym{lhd}{LHD}{Latin Hypercube Design}
\newacronym{tplhd}{TPLHD}{Translational Propagation Latin Hypercube Design}
\newacronym{rmse}{RMSE}{Root Mean-Squared Error}
\newacronym{rmae}{RMAE}{Relative Maximum Absolute Error}
\newacronym{cvd}{CVD}{Cross-Validation Distance}
\newacronym{le}{LE}{Lyapunov Exponents}
\newacronym{lle}{LLE}{Largest Lyapunov Exponent}
\newacronym{epe}{EPE}{Expected Prediction Error}
\newacronym{mepe}{MEPE}{Maximizing Expected Prediction Error}
\newacronym{loocv}{LOCVV}{leave-one-out cross-validation}
\newacronym{gmse}{GMSE}{Generalized Mean Square Cross-Validation Error}
\newacronym{sfcvt}{SFCVT}{Space-Filling Cross Validation Tradeoff}
\newacronym{ace}{ACE}{ACcumulative Error}
\newacronym{doi}{DOI}{Degree-of-Influence}
\newacronym{blup}{BLUP}{Best linear unbiased predictor}
\newacronym{mle}{MLE}{Maximum likelihood estimation}
\newacronym{de}{DE}{Differential evolution}
\newacronym{mivor}{MIVor}{MC-Intersite Voronoi}

\numberwithin{equation}{section}	
\numberwithin{figure}{section}		
\numberwithin{table}{section}       

\usepackage{enumitem}

\usepackage{subcaption}
\usepackage{tikz}
\usetikzlibrary{shapes,arrows}

\usepackage{multirow}
\usepackage{amsthm}
\newtheorem*{mydef}{Definition}

\usepackage{bm}

\usepackage{multicol}
\newcommand{\specialcell}[2][c]{%
  \begin{tabular}[#1]{@{}l@{}}#2\end{tabular}}

\usepackage{mathtools}
\usepackage{pgfplots}   
\usepgfplotslibrary{units}
\usetikzlibrary{spy,backgrounds}
\usepgfplotslibrary{fillbetween}
\usetikzlibrary{patterns}
\usepackage{booktabs}

\usepackage{breqn}
\usepackage{wrapfig,lipsum,booktabs}
\usepackage{pgfplots}
    \usepackage{bbm}

\usepackage{physics}

\renewcommand\paragraph{\@startsection{paragraph}{4}{\z@}%
            {-2.5ex\@plus -1ex \@minus -.25ex}%
            {1.25ex \@plus .25ex}%
            {\normalfont\normalsize\bfseries}}
\usepackage{xcolor}
\usepackage[framemethod=TikZ]{mdframed}
\definecolor{ikmgray}{HTML}{5E5E5E}
\definecolor{ikmgreen}{HTML}{C9DA2B}
\newmdenv[frametitle={},
middlelinecolor=ikmgreen,
middlelinewidth=0pt,
backgroundcolor=ikmgray!20,
roundcorner=2pt,
bottomline=false,
leftline=true,
topline=false,
rightline=false,
skipabove=10pt,
skipbelow=10pt,
leftmargin=10pt,
rightmargin=10pt,
innerleftmargin=10pt,
innerrightmargin=10pt,
innertopmargin=10pt,
innerbottommargin=10pt]{Algorithmus}
\usepackage[labelfont=bf, 
format=plain, 
labelsep=endash, 
justification=raggedright 
]{caption}
\DeclareCaptionType{kasten}[Box]
\usepackage{tabularx}

 \makeatletter
\def\l@figure{\@dottedtocline{1}{1.5em}{3em}}
\makeatother

 \makeatletter
\def\l@table{\@dottedtocline{1}{1.5em}{3em}}
\makeatother

\usepackage{fancyhdr}       
\begin{document}
\counterwithin{lstlisting}{section} 

\newcommand{\authorsName}{Jan N. Fuhg}
\newcommand{\authorsNumber}{} 
\newcommand{\handInDate}{\today} 

\newcommand{\titleInGerman}{Adaptive Ersatzmodelle für parametrische Studien}
\newcommand{\titleInEnglish}{Adaptive surrogate models for parametric studies}
\newcommand{\TypeOfThesisInGerman}{Masterarbeit}
\newcommand{\TypeOfThesisInEnglish}{Master Thesis}

\newcommand{\nameOfSupervisorA}{Dr.-Ing. Amelie Fau }
\newcommand{\nameOfSupervisorB}{}
\newcommand{\nameOfSupervisorC}{ }
\newcommand{\nameOfExaminerA}{Prof. Dr.-Ing. Udo Nackenhorst}
\newcommand{\nameOfExaminerB}{Prof. Dr.-Ing. Michael Beer }
\newcommand{\nameOfExaminerC}{ }

\newcommand{\ThesisLanguage}{1}

\newcommand{\titlepicture}{Bilder/Ackley_Images/SSA.pdf} 


\begin{titlepage}
\enlargethispage{2cm}

\begin{picture}(0,0)(0,35)
\put(-120,0){\includegraphics[height=58mm,angle=0]{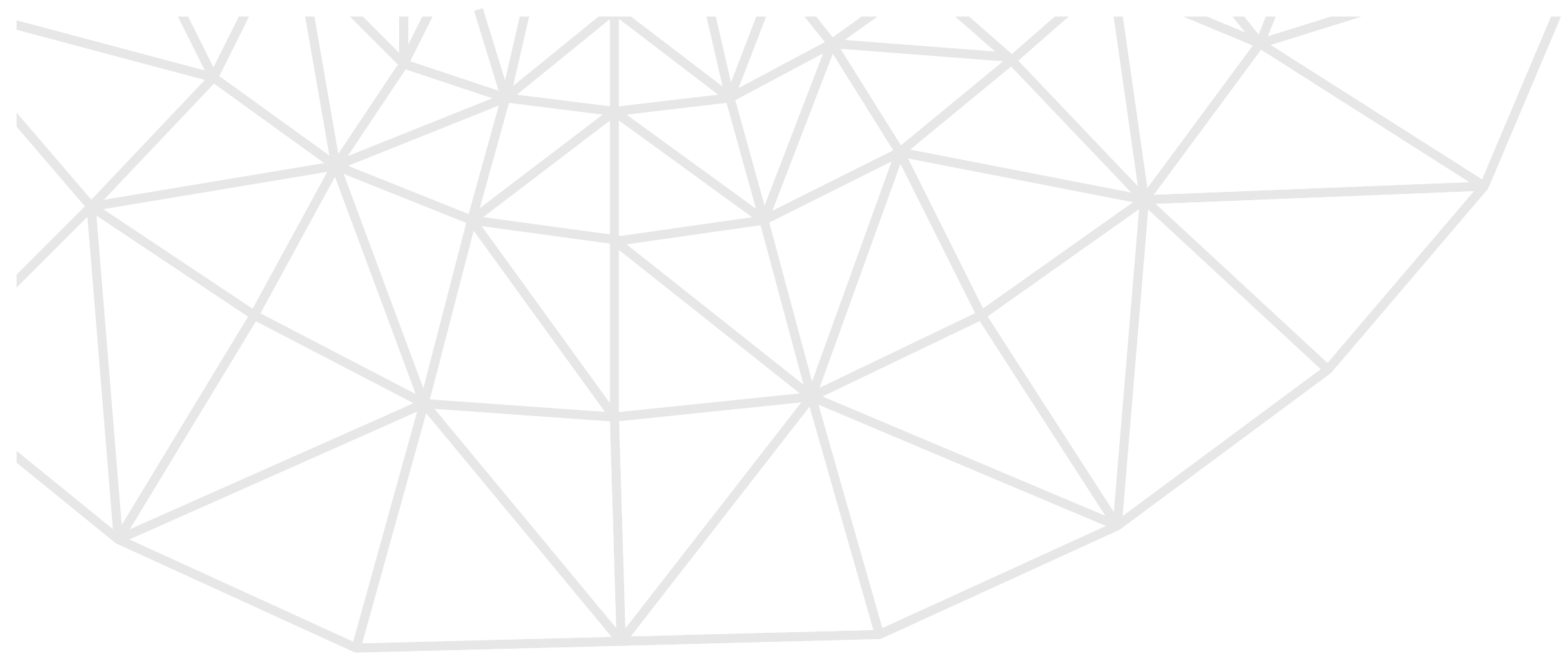}}
\put(-75,112){\includegraphics[height=18mm,angle=0]{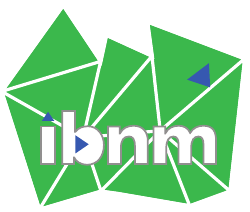}}
\put(0,118){\parbox[b]{\textwidth}{\Large \textbf{Institut für Baumechanik\\ und Numerische Mechanik}}}
\put(300,112){\includegraphics[width=60mm,angle=0]{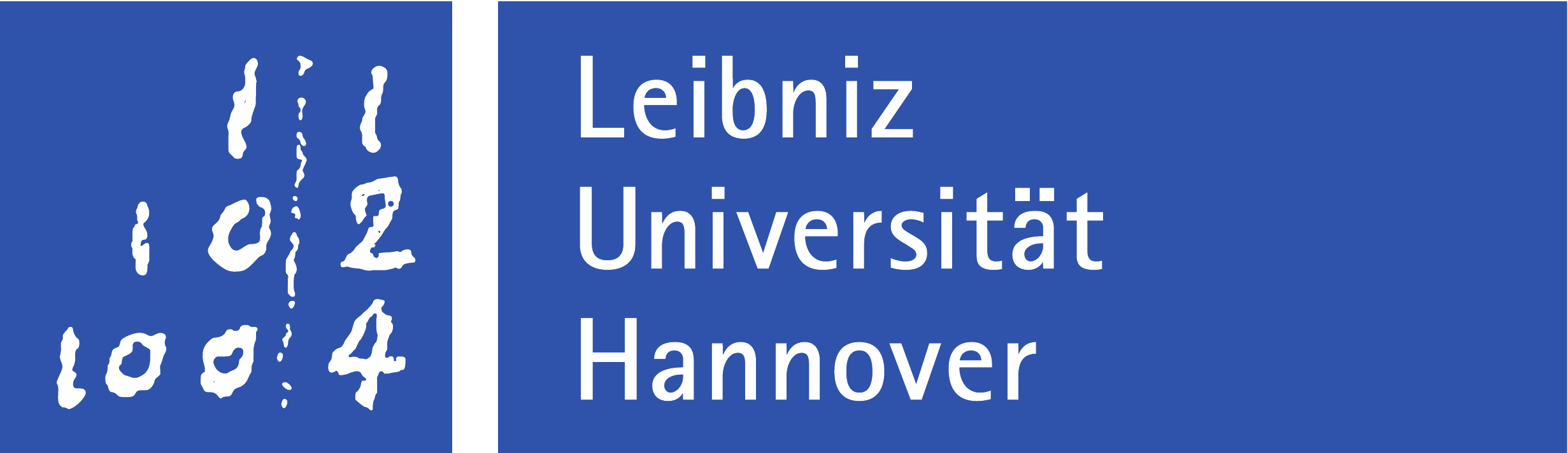}}
\end{picture}
\vspace*{3cm}
   
\begin{center}

\ifnum\ThesisLanguage=1
    {\Huge \bf \titleInEnglish}\\
    \vspace*{0.3cm}
\else
    {\Huge \bf \titleInGerman}\\
    \vspace*{0.3cm}
    {\huge \titleInEnglish}
\fi

\vfill
\includegraphics[width=0.6\textwidth]{\titlepicture}
\vfill
\ifnum\ThesisLanguage=1
    {\Large \bf \TypeOfThesisInEnglish}\\
    {\authorsName} \\
\else
    {\Large \bf \TypeOfThesisInGerman}\\
    {\authorsName\\Matrikelnummer: \authorsNumber} \\
\fi

\vspace{0.5cm}

\textbf{\handInDate}\\

\vspace{0.5cm}

\begin{tabular}{p{0.5\textwidth}p{0.5\textwidth}}
\ifnum\ThesisLanguage=1
  Supervisor: & Examiner: \\
\else
  Betreuer: & Prüfer: \\
\fi
  \nameOfSupervisorA & \nameOfExaminerA \\
  \nameOfSupervisorB & \nameOfExaminerB \\
\end{tabular}

\end{center}
\end{titlepage}

\thispagestyle{empty}
\section*{Acknowledgments}
I would like to thank Dr. Ing. Amelie Fau and Prof. Dr. Ing. Udo Nackenhorst for all the support and helpful comments throughout the process of this thesis. \\
\\
A very special word goes out to my parents and my sister for pointing out my messed-up sleeping schedule and to Fernanda for always being there for me.
\\
\\
\\
The results presented in this thesis were (partially) carried out on the cluster system at the Leibniz University
of Hannover, Germany. 
\newpage
\thispagestyle{empty}
\section*{Abstract}
The computational effort for the evaluation of numerical simulations based on e.g. the finite-element method is high. Metamodels can be utilized to create a low-cost alternative. However the number of required samples for the creation of a sufficient metamodel should be kept low, which can be achieved by using adaptive sampling techniques.
In this thesis adaptive sampling techniques are investigated for their use in creating metamodels with the Kriging technique, which interpolates values by a Gaussian process governed by prior covariances. The Kriging framework with extension to multifidelity problems is presented and utilized to compare adaptive sampling techniques found in the literature for benchmark problems as well as applications for contact mechanics. \\
This thesis offers the first comprehensive comparison of a large spectrum of adaptive techniques for the Kriging framework. Furthermore a multitude of adaptive techniques is introduced to multifidelity Kriging as well as well as to a Kriging model with reduced hyperparameter dimension. \\ In addition, an innovative adaptive scheme for binary classification is presented and tested for identifying chaotic motion of a Duffing's type oscillator.  

\newpage 
\pagenumbering{roman}
\pagestyle{plain}
\pagenumbering{roman}
\tableofcontents

\listoffigures

\listoftables

\printglossary[type=\acronymtype,nonumberlist]
 \glsaddallunused[\acronymtype]
\newpage
\pagestyle{fancy}
\pagenumbering{arabic}
\chapter{Introduction}
\begin{figure}[htpb]
\centering
\includegraphics[scale=0.6]{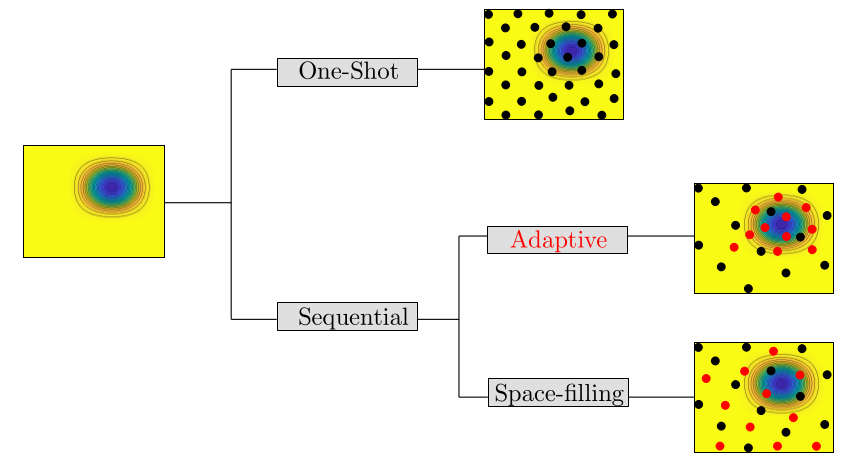}
\caption{Classification of sampling techniques for global metamodeling.}\label{fig::SamplingApproaches}
\end{figure}

Computational models play an increasingly important role in many aspects of science and engineering analysis. Due to the increasing capabilities in computer hardware, modern engineering problems are solved with methods like computational fluid dynamics (CFD) and finite element analysis (FEA) in order to create accurate high-fidelity simulations. 
In general these simulations can assumed to be black box functions, with no prior knowledge about their inner workings, that map parametric input values to an output. This can be used to examine and understand the behavior of the engineering system and to determine regions of interest in the parametric space. 
For real-world physical problems however evaluations of the mapping require high computational resources. \\
Surrogate models, also known as metamodels or response surfaces, are used to replace expensive function evaluations by evaluating the surrogate model itself, which in turn mitigates the computing costs. Herein surrogate models are an approximation of the input/output function. Generally, metamodels can be categorized into \textit{global} and \textit{local} models \citep{crombecq2009space}, where \textit{global} models are used to approximate the simulator on the entire domain. The surrogate aims to become a complete replacement of the original simulator, whereas the goal of \textit{local} models is to accurately approximate the behavior of the simulator locally and therefore e.g. to direct an optimization algorithm towards the global optimum. \\
Surrogate model approaches such as \acrfull{pr}(see e.g. \cite{kleijnen2017regression}), Kriging \citep{kleijnen2009kriging}, \acrfull{svm} \citep{clarke2005analysis} or \acrfull{rbf} models \citep{park1991universal} have been developed and continuously improved in recent years.  The global metamodel approach Kriging as the most intensively investigated metamodel  \citep{jiang2017adaptive}. Originally developed by \cite{krige1951statistical} for the use in geostatistics it has since been extended to deterministic \citep{sacks1989design} and random simulations models \citep{van2003kriging}. The technique has established itself in various fields under different names e.g. Gaussian process regression \citep{williams1996gaussian} in machine learning. This thesis employs the term Kriging. 
In contrast to most of the aforementioned metamodel approaches Kriging is an accurate interpolative Bayesian surrogate modeling technique. Hence, the algebraic difference between the training data and the predicted points equals zero. Furthermore, Kriging exhibits a stochastic property. Therefore, in addition to the predicted values, the predicted variances between the responses of the black box function and the Kriging surrogate model can be obtained. Additionally, Kriging can be used to include multi-fidelity models in the surrogate, whereby numerical models of varying degrees of fidelity and computational expense are combined.  \\
The quality of the approximation of the black box function is dependent on the quality of the metamodel, which is largely influenced by the sample points where the simulator is evaluated. Generally, the design considerations of surrogates are influenced by a conflict between high accuracy and low computational costs \citep{jin2002sequential}. On one hand
more sample points lead to a more accurate metamodel but also more evaluations of the black box function and thus higher computational expenses. On the other hand fewer sample points require less evaluations but result in surrogates that may be inaccurate and even distorted.\\
Sampling approaches, as shown in Figure \ref{fig::SamplingApproaches}, can generally be divided into one-shot and sequential techniques. One-shot sampling generates the sample points in a single step. Common approaches include orthogonal array \citep{owen1992orthogonal},  \acrfull{lhd} \citep{husslage2011space} and \acrfull{olhd} \citep{park1994optimal}. The advantage of one-shot sampling is the ease of implementation and the good coverage of the design space.
However, when the behavior of the input/output function is unknown the determination of an optimal sample size is difficult. Therefore sequential sampling strategies have been introduced, see e.g. \cite{sacks1989design} or \cite{jin2002sequential}, where an iterative sampling process is employed to determine new sample points using the information available from previously executed iterations. The process continues until a threshold number of sample points is exceeded or the desired accuracy is reached. Sequential sampling approaches can be categorized into adaptive and space-filling sequential sampling. Here, space-filling techniques generate samples iteratively from a one-shot sampling method. Adaptive sampling generates points in regions of the parametric space with large prediction errors in order to adapt to the properties of the black box function function. 
Consequently, in comparison to space-filling techniques, surrogates built from adaptively sampled points tend to show more proficient approximations of the simulator while requiring fewer samples.
A comprehensive review and comparison of sequential sampling techniques for Kriging is missing in the literature. Furthermore the existing techniques are seldomly tested for multi-fidelity modeling.\\
The basic principles of proficient sampling techniques are explored in this thesis and common adaptive methods are compared on benchmark problems and mechanical applications with varying degrees of fidelity. \\ 
In this process it has been observed that commonly used sampling techniques found in the literature are unable to sufficiently generate samples when utilizing the Kriging approach to built surrogates for binary classification problems. To overcome this shortcoming, the innovative adaptive sampling technique \acrfull{mivor} is introduced. \\
The thesis is organized as follows.
\begin{itemize}
\item A general overview over the Kriging framework is given in chapter \ref{sec::Surrogate}. Critical modeling consideration including the design choices made in this thesis are introduced. Ordinary Kriging as a proficient compromise between accuracy and computational efficiency is highlighted. Furthermore the method is compared to common surrogate modeling techniques found in the literature.
\item Adaptive sampling strategies are discussed and explained in chapter \ref{sec::adaptiveSampling}. Representative groups of techniques are selected and studied. A review of design considerations for adaptive techniques is given and the choices that have been made in this thesis are presented. 
\item Commonly utilized adaptive sampling strategies are compared for benchmark problems in chapter \ref{ch::AdaptiveSamplinginOK} using Ordinary Kriging. 
\item Adaptive sampling techniques are investigated for their use in creating surrogates for the study of a nonlinear oscillator of Duffing's type with a nonlinear friction force model in chapter \ref{chapt::DynamicApproach}.
\item The novel MIVor technique is introduced in chapter \ref{ch::Novel_techniques} and validated for the generation of a metamodel for binary classification of chaotic motion of a dynamic system.
\item A multifidelity Kriging version called Hierarchical Kriging (HK) is studied in chapter \ref{ch::MFKriging}. Adaptive sampling techniques are utilized for building HK metamodels for benchmark problems as well as a multifidelity Finite-Element application involving nonlinearity with contact.
\item A method to reduce the computational effort present when employing Kriging in high-dimensionality problems by utilizing partial least square dimension reduction is studied and expanded for its use with HK in chapter \ref{ch::PLSOK}. Furthermore adaptive sampling techniques are employed to efficiently create high-dimensional metamodels utilizing these approaches.
\item Concluding remarks and a final evaluation of the tested adaptive sampling methods are presented in chapter \ref{chapter::ConclusionandOutlook}.  
\end{itemize}
\chapter{Kriging surrogate modeling}\label{sec::Surrogate}
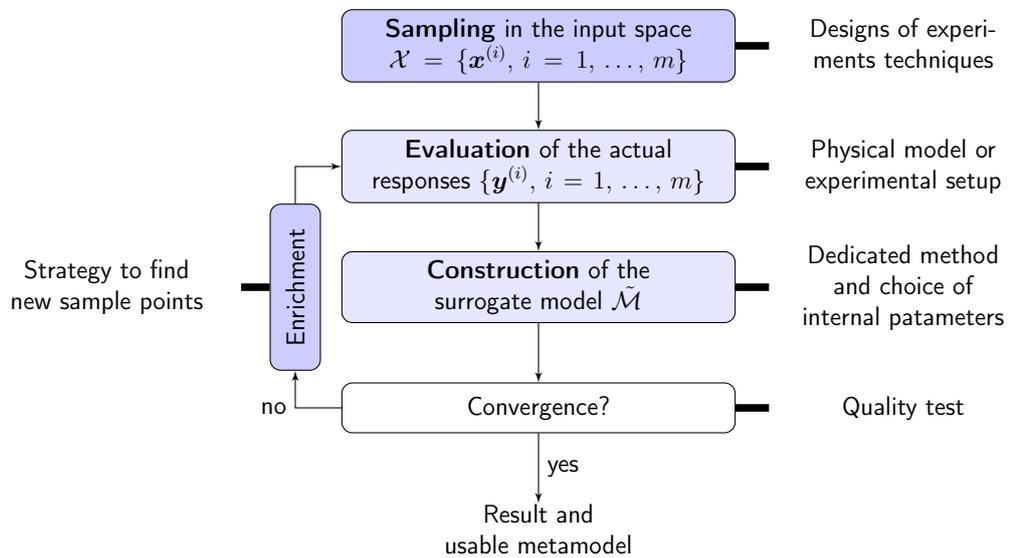
\begin{figure}[h!]
\tikzstyle{decisionA} = [  
    text width=15em, text centered, inner sep=0pt,minimum height=2em]
\tikzstyle{decisionB} = [rectangle, draw,  
    text width=15em, text centered, rounded corners, minimum height=2em]
\tikzstyle{blockB} = [rectangle, draw, fill=blue!20, 
    text width=15em, text centered, rounded corners, minimum height=2em]
\tikzstyle{blockC} = [rectangle, draw, fill=blue!10, 
    text width=15em, text centered, rounded corners, minimum height=2em]
    \tikzstyle{blockD} = [rectangle, draw, fill=blue!20, 
    text width=6em, text centered, rounded corners, minimum height=2em]
\tikzstyle{line} = [draw, -latex']
\tikzstyle{line2} = [draw,line width=1.0mm]
\tikzstyle{cloudA} = [draw, ellipse,fill=red!20,
    minimum height=2em]
    \tikzstyle{blockE} = [rectangle,  
    text width=10em, text centered, rounded corners, minimum height=2em]
 \centering 
\begin{tikzpicture}[node distance = 2.0cm, auto,scale=0.8, transform shape]
    \node [blockB,node distance=8.0cm] (init) {
\textbf{Sampling} in the input space  $\mathcal{X} = \lbrace \bm{x}^{(i)}, \, i=1, \, \ldots  , \, m  \rbrace$
};
    \node [blockC, below of=init] (Evaluate) {\textbf{Evaluation} of the actual\\responses $\lbrace \bm{y}^{(i)}, \, i=1, \, \ldots  , \, m  \rbrace$};
    \node [blockC, below of=Evaluate] (Update) {\textbf{Construction} of the surrogate model $\tilde{\mathcal{M} }$ }; 
          \node [decisionB, below of=Update] (Accurate) {Convergence?};     
     \node [blockD, left of=Update,node distance=4.0cm,rotate=90] (newM) {Enrichment};
          \node [decisionA, below of=Accurate] (Done) {Result and \\usable metamodel};
        
         \node [blockE, right of=Accurate,node distance=6.0cm] (quality) {Quality test};   
         
                  \node [blockE, right of=Update,node distance=6.0cm] (method) {Dedicated method and choice of \\internal patameters};    
                           \node [blockE, right of=Evaluate,node distance=6.0cm] (model) {Physical model or experimental setup};    
         \node [blockE, right of=init,node distance=6.0cm] (techniques) {Designs of experiments techniques};                               
                             
                  \node [blockE, left of=newM,node distance=3.1cm] (newSamples) {Strategy to find new sample points};    

    \path [line2] (Accurate) -- (quality);
\path [line2] (init) -- (techniques);
\path [line2] (Update) -- (method);
\path [line2] (Evaluate) -- (model);

        \path [line2] (newM.north) -- (newSamples.east);
    \path [line] (init) -- (Evaluate);
    \path [line] (Evaluate) -- (Update);
     \path [line] (Update) -- (Accurate);
          \path [line] (Accurate.west) -|  node {no}(newM.west);

          \path [line]
    (newM.east)  |-   (Evaluate.west);
                    \path [line] (Accurate) -- node {yes}(Done);

\end{tikzpicture}
\caption[Surrogate model scheme]{Scheme of the basic steps in building a surrogate model, inspired by \cite{laurent2017overview}.}\label{fig::flowchart_Worklflow}
\end{figure}

From a given set of observations $\mathcal{D} = \lbrace \left( \bm{x}^{(i)}, \,\bm{y}^{(i)} \right), \, i=1, \, \ldots  , \, m  \rbrace$ the surrogate model in global metamodeling aims to reproduce the statistical relationship of a given mapping $\mathcal{M} \, : \mathbb{X} \rightarrow \mathbb{Y}$ between an input $\bm{x} \in \mathbb{X} \subset \mathbb{R}^{n}$ and some output $\bm{y} \in \mathbb{Y} \subset \mathbb{R}^{d}$. The computed surrogate model will be denoted by $\tilde{\mathcal{M} }$. Creating metamodels in this framework is known as supervised learning \citep{rasmussen2006gaussian} or active learning \citep{settles2014active}. For univariate output ($d=1$) the mapping evaluations are gathered in the vector $\bm{y} = \left( y_{i}, \, i=1, \, \ldots  , \, m  \right)$. For the applications of this thesis the input space $\mathbb{X}$ is continuous. The chosen samples in the input space are called \acrfull{doe} and are denoted by the set $\mathcal{X} = \lbrace \bm{x}^{(i)}, \, i=1, \, \ldots  , \, m  \rbrace$. \\
Generally, the construction of the commonly used surrogate models follows a basic work-flow. As illustrated in Figure \ref{fig::flowchart_Worklflow} the basic steps are:
\begin{itemize}
\item \textit{Generation of the initial data $\mathcal{D}$.} Employment of a sampling approach (DOE technique) either one-shot or sequential to generate sample points in the input space $\mathcal{X} = \lbrace \bm{x}^{(i)}, \, i=1, \, \ldots  , \, m  \rbrace$. The responses are calculated by evaluating the mapping $\mathcal{M}$. For real-world engineering problems these evaluations may be computationally expensive.
\item \textit{Construction of the surrogate model $\tilde{\mathcal{M}}$.} With the generated data set $\mathcal{D}$ determine the internal parameters of the surrogate model to approximate $\mathcal{M}$ at best with a low-cost. 
\item \textit{Convergence criteria and enrichment.} Measure the quality of the surrogate model using a chosen, dedicated criterion. 
In a one-shot sampling approach the construction of the surrogate model is finished. In sequential approaches if the criterion is not reached an enrichment strategy is employed to add points to the previous set of sample points in order to update the surrogate model $\tilde{\mathcal{M}}$. Hence,
these strategies post-process the current surrogate, until reaching the criterion and so proposing a satisfactory surrogate model.
\end{itemize}
At the end of this process an approximation of the mapping is found whose evaluation requires less computational expenses than $\mathcal{M}$. Depending on the number of evaluations the construction of the surrogate model can be time consuming. However it should lead to an overall reduction in time in comparison to numerous evaluations needed to evaluate the full solution for the whole parametric space, e.g. required for optimization or reliability analysis problems.

The surrogate modeling approach in this thesis is Kriging, also known as Gaussian process modeling. Originally, Kriging was proposed by a mining engineer called Krige for the use in geostatistics \citep{krige1951statistical}. \cite{matheron1963principles} developed the mathematical foundation for the approach and established the name. For an overview the reader is referred to \cite{kleijnen2009kriging} and \cite{kleijnen2017regression}. The classical types of Kriging are \acrfull{ok}, simple Kriging, \acrfull{uk} and \acrfull{gek} \citep{li2010systematic}. \\
Generally Kriging is the best linear unbiased prediction and can be used as an interpolative surrogate modeling technique utilizing the Bayesian prediction methodology, see e.g. \cite{santner2013design}. In the following the general idea behind UK is introduced and the special case OK is explained. 
\section{Universal Kriging}
Consider the training data set $\lbrace \left( \bm{x}^{(i)}, \,y^{(i)} \right), \, i=1, \, \ldots  , \, m  \rbrace$ of a black-box mapping $\mathcal{M}$. The idea is that the functional relationship between the input and the output is definable by a sample path of a Gaussian process $Y$ and therefore the output will be considered uncertain. A short overview over Gaussian processes is given in \ref{sec::GaussianProcess}. \\ UK is a generalized version of the well known linear regression with a noise term (see Appendix \ref{seq::LeastLinearRegression} for more details about the least squares linear regression model and specifically equation (\ref{eq:LinearRegression})) in which $\bm{Y}$ is approximated by a general linear model where the additional noise term is now represented by a Gaussian process and depends on $\bm{x}$. UK therefore assumes a general polynomial trend in the random function. This concept can be written as
\begin{equation}\label{eq::UK}
Y_{i}(\bm{x}^{(i)})  = \sum_{j=1}^{p} \beta_{j} f_{j}(\bm{x}^{(i)}) + Z(\bm{x}^{(i)}) , \qquad i = 1, \, \ldots \, , m ,
\end{equation}
where $\bm{Y} = \left( Y_{i}, i = 1, \, \ldots , \, m \right)^{T}$ is the vector of observations, \\ $\bm{\beta} = \left( \beta_{j}, j = 1, \, \ldots , \, p \right)^{T}$ is the vector of weights, $\bm{f} = \left( f_{j}, j = 1, \, \ldots , \, p \right)^{T}$ is a collection of regression functions.
Equation (\ref{eq::UK}) describes a two-stage Gaussian process. 
 The first stage contains a linear combination on a given functional basis with $p \leq m$. The second stage consists of a stationary Gaussian process with zero mean and stationary autocovariance $\mathcal{C}$ given by
\begin{equation}
\mathcal{C}(\bm{x}, \bm{x'}) = \sigma^{2} R (\bm{x} - \bm{x'}, \bm{\theta}),
\end{equation}
where $\sigma^{2}$ is the variance and the spatial autocorrelation function $R$ (or kernel) only depends on the euclidean distance between the input combinations $\bm{x}$ and $\bm{x'}$ and some hyperparameters $\bm{\theta}$ which are considered to be known.  
Equation (\ref{eq::UK}) can be reformulated as the matrix equation system 
 \begin{equation}\label{eq::Regression_model}
 \bm{Y}(\bm{x}) = \bm{F}(\bm{x}) \bm{\beta} + Z(\bm{x}).
 \end{equation}
Here, $\bm{F}$ is the so-called regression matrix 
\begin{equation}
F_{i \, j}= f_{j} (\bm{x}^{(i)}), \qquad i=1, \, \ldots \, , m, \qquad j=1, \, \ldots \, , p,
\end{equation}
which for
e.g. $\bm{x} = x$ and $\bm{f} = \lbrace 1, x , x^2, \, \ldots \, , x^{p} \rbrace$ can be written as
\begin{equation}
\bm{F} = \begin{bmatrix}
1 & x^{1} & (x^{1})^{2} & \ldots & (x^{1})^{p} \\
\vdots & \vdots & \vdots & \vdots & \vdots \\
1 & x^{m} & (x^{m})^{2} & \ldots & (x^{m})^{p} 
\end{bmatrix} \, \text{.}
\end{equation} 
Because of the Gaussian assumption in equation (\ref{eq::UK}) the vector consisting of the observations $\bm{Y}$ and a prediction of an unobserved value $Y_{0}$ corresponding to point $\bm{x}^{(0)}$ can be approximated by a random vector with normal distribution. 
Before considering the estimation for the unobserved quantity of interest, kernel formulations will be shortly presented.
\section{Kernel formulations}
\begin{figure}
\centering
\begin{subfigure}[t]{0.5\textwidth}
\begin{tikzpicture}[scale = 0.7, transform shape]
\begin{axis}
[
 axis equal,
xlabel={$x - x'$},
ylabel={$R(x - x')$},
y label style={anchor=north},
x label style={anchor=east},
axis lines=middle,
xtick={0, 0.2,0.4,0.6,0.8,1.0},
ytick={0, 0.2,0.4,0.6,0.8,1.0},
xticklabels={0, 0.2,0.4,0.6,0.8,1.0},
yticklabels={0, 0.2,0.4,0.6,0.8,1.0},
grid=both,
    grid style={line width=.1pt, draw=gray!10},
    major grid style={line width=.2pt,draw=gray!20},
    minor tick num=5,
    ymin=-0.1,ymax=1.1,
    xmin=-0.1,xmax=1.1,
    axis on top=false,
every axis x label/.style={
    at={(ticklabel* cs:1.01)},
    anchor=west,
},
every axis y label/.style={
    at={(ticklabel* cs:1.01)},
    anchor=south,
},
]
\addplot[blue,line width = 0.5mm, domain=0:1.0] {exp(-(x/0.1)^1} ;
\addlegendentry{$\theta = 0.1$}
\addplot[red,line width = 0.5mm, domain=0:1.0] {exp(-(x/0.5)^1} ;
\addlegendentry{$\theta = 0.5$}
\addplot[green,line width = 0.5mm, domain=0:1.0] {exp(-(x/1.0)^1} ;
\addlegendentry{$\theta = 1.0$}

\end{axis}
\end{tikzpicture}
\subcaption{}\label{fig::Exponential_AC}
\end{subfigure}%
\begin{subfigure}[t]{0.5\textwidth}
\begin{tikzpicture}[scale = 0.7, transform shape]
\begin{axis}
[
 axis equal,
xlabel={$x - x'$},
ylabel={$R(x - x')$},
y label style={anchor=north},
x label style={anchor=east},
axis lines=middle,
xtick={0, 0.2,0.4,0.6,0.8,1.0},
ytick={0, 0.2,0.4,0.6,0.8,1.0},
xticklabels={0, 0.2,0.4,0.6,0.8,1.0},
yticklabels={0, 0.2,0.4,0.6,0.8,1.0},
grid=both,
    grid style={line width=.1pt, draw=gray!10},
    major grid style={line width=.2pt,draw=gray!20},
    minor tick num=5,
    ymin=-0.1,ymax=1.1,
    xmin=-0.1,xmax=1.1,
    axis on top=false,
every axis x label/.style={
    at={(ticklabel* cs:1.01)},
    anchor=west,
},
every axis y label/.style={
    at={(ticklabel* cs:1.01)},
    anchor=south,
},
]
\addplot[blue,line width = 0.5mm, domain=0:1.0] {exp(-(x/sqrt(0.1))^2} ;
\addlegendentry{$\theta = 0.1$}
\addplot[red,line width = 0.5mm, domain=0:1.0] {exp(-(x/sqrt(0.5))^2} ;
\addlegendentry{$\theta = 0.5$}
\addplot[green,line width = 0.5mm, domain=0:1.0] {exp(-(x/sqrt(1.0))^2} ;
\addlegendentry{$\theta = 1.0$}

\end{axis}
\end{tikzpicture}
\subcaption{}\label{fig::Squared_Exponential_AC}
\end{subfigure}
\caption[Exponential and squared exponential kernel]{Autocorrelation function for various scale parameters. (a) exponential kernel and (b) squared exponential kernel}\label{fig::Power_Exponential}
\end{figure}
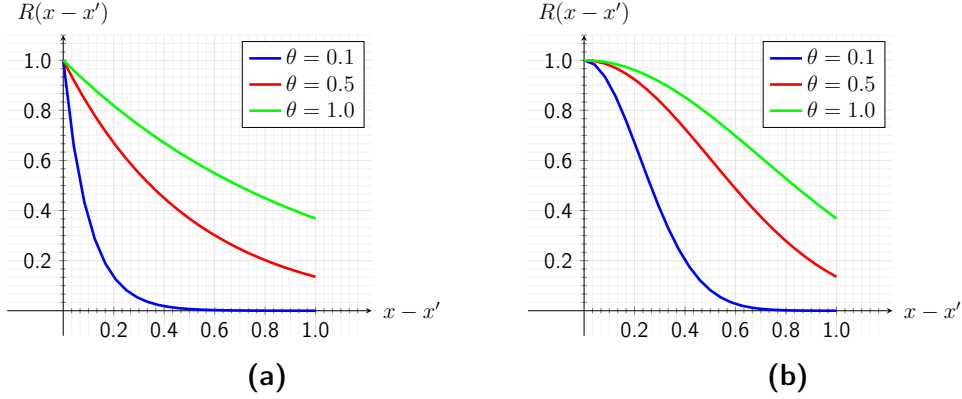

 The details of the autocovariance function need to be chosen before constructing the surrogate model. This entails assuming a functional form of the autocorrelation as well as determining the respective parameters governing the function. The latter are also referred to as hyperparameters. The choice of autocorrelation function should reflect the underlying smoothness and dependence of the dataset. However these properties are rarely a priori known. \\
\cite{han2009improving} mention that to produce better conditioned covariance matrices for multiple dimensions multidimensional kernels should be built by taking the product of unidimensional kernel $h$. This is called the product correlation rule \citep{sasena2002flexibility} and for an $n$-dimensional input space reads
\begin{equation}
R (\bm{x} - \bm{x'}, \bm{\theta}) = \prod_{i=1}^{n} h_{i} (x_{i} - x_{i}', \theta_{i}),
\end{equation}
Different choices for the kernel can be found in the literature. The most common kernels will be reviewed hereafter. \\
The power exponential correlation function reads
\begin{equation}
R (\bm{x} - \bm{x'}, \bm{\theta}) = \exp \left(  -  \sum_{i=1}^{n} \left( \dfrac{\abs{x_{i} - x'_{i}}}{\theta_{i}} \right)^{\nu} \right)
\end{equation}
with the scale parameters $\lbrace \theta_{i} > 0 , i=1, \,  \ldots \, , n \rbrace$ and $0 < \nu \leq 2$. As special cases of this function the exponential ($\nu=1$) and squared exponential or Gaussian ($\nu=2$) forms can be considered (see Figure \ref{fig::Power_Exponential}).

\begin{figure}
\centering
\begin{subfigure}[t]{0.5\textwidth}
\begin{tikzpicture}[scale = 0.7, transform shape]
\begin{axis}
[
 axis equal,
xlabel={$x - x'$},
ylabel={$R(x - x')$},
y label style={anchor=north},
x label style={anchor=east},
axis lines=middle,
xtick={0, 0.2,0.4,0.6,0.8,1.0},
ytick={0, 0.2,0.4,0.6,0.8,1.0},
xticklabels={0, 0.2,0.4,0.6,0.8,1.0},
yticklabels={0, 0.2,0.4,0.6,0.8,1.0},
grid=both,
    grid style={line width=.1pt, draw=gray!10},
    major grid style={line width=.2pt,draw=gray!20},
    minor tick num=5,
    ymin=-0.1,ymax=1.1,
    xmin=-0.1,xmax=1.1,
    axis on top=false,
every axis x label/.style={
    at={(ticklabel* cs:1.01)},
    anchor=west,
},
every axis y label/.style={
    at={(ticklabel* cs:1.01)},
    anchor=south,
},
]
\addplot[blue,samples=50,line width = 0.5mm, domain=0:1.0] {(1+( (sqrt(3)*x)/0.1))*exp(-( (sqrt(3)*x)/0.1) )} ;
\addlegendentry{$\theta = 0.1$}
\addplot[red,samples=50,line width = 0.5mm, domain=0:1.0] {(1+( (sqrt(3)*x)/0.5))*exp(-( (sqrt(3)*x)/0.5) )} ;
\addlegendentry{$\theta = 0.5$}
\addplot[green,samples=50,line width = 0.5mm, domain=0:1.0] {(1+( (sqrt(3)*x)/1.0))*exp(-( (sqrt(3)*x)/1.0) )} ;
\addlegendentry{$\theta = 1.0$}

\end{axis}
\end{tikzpicture}
\subcaption{}\label{fig::Matern3/2}
\end{subfigure}%
\begin{subfigure}[t]{0.5\textwidth}
\begin{tikzpicture}[scale = 0.7, transform shape]
\begin{axis}
[
 axis equal,
xlabel={$x - x'$},
ylabel={$R(x - x')$},
y label style={anchor=north},
x label style={anchor=east},
axis lines=middle,
xtick={0, 0.2,0.4,0.6,0.8,1.0},
ytick={0, 0.2,0.4,0.6,0.8,1.0},
xticklabels={0, 0.2,0.4,0.6,0.8,1.0},
yticklabels={0, 0.2,0.4,0.6,0.8,1.0},
grid=both,
    grid style={line width=.1pt, draw=gray!10},
    major grid style={line width=.2pt,draw=gray!20},
    minor tick num=5,
    ymin=-0.1,ymax=1.1,
    xmin=-0.1,xmax=1.1,
    axis on top=false,
every axis x label/.style={
    at={(ticklabel* cs:1.01)},
    anchor=west,
},
every axis y label/.style={
    at={(ticklabel* cs:1.01)},
    anchor=south,
},
]
\addplot[blue,samples=50,line width = 0.5mm, domain=0:1.0] {(1 + ((sqrt(5)*x)/0.1) + ((5*x^2)/(3*0.1^2)))*exp(- (sqrt(5)*x)/(0.1))} ;
\addlegendentry{$\theta = 0.1$}
\addplot[red,samples=50, line width = 0.5mm, domain=0:1.0] {(1 + ((sqrt(5)*x)/0.5) + ((5*x^2)/(3*0.5^2)))*exp(- (sqrt(5)*x)/(0.5))} ;
\addlegendentry{$\theta = 0.5$}
\addplot[green,samples=50,line width = 0.5mm, domain=0:1.0] {(1 + ((sqrt(5)*x)/1.0) + ((5*x^2)/(3*1.0^2)))*exp(- (sqrt(5)*x)/(1.0))} ;
\addlegendentry{$\theta = 1.0$}

\end{axis}
\end{tikzpicture}
\subcaption{}\label{fig::Matern5/2}
\end{subfigure}
\caption[Matérn kernels]{Matérn autocorrelation functions for various scale parameters. (a) Matérn 3/2 and (b) Matérn 5/2}\label{fig::Matern}
\end{figure}
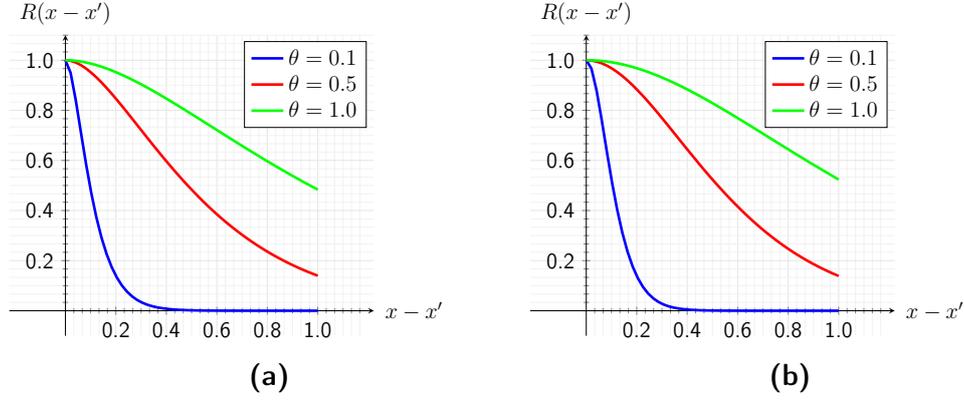

\cite{stein2012interpolation} introduced the Matérn class of autocorrelation function \citep{matern1960spatial} to the Kriging surrogate model framework. They have the convenient property of being highly adjustable. The general Matérn autocorrelation function is given as
\begin{equation}
\begin{aligned}
R (\bm{x} - \bm{x'}, \bm{\theta}, \nu) = \prod_{i=1}^{n} \dfrac{2^{1- \nu}}{\Gamma(\nu)} \left( \frac{\sqrt{2 \nu} \abs{x_{i} - x_{i}'}}{\theta_{i}}\right)^{\nu} \, \mathcal{K}_{\nu} \left( \frac{\sqrt{2 \nu} \abs{x_{i} - x_{i}'}}{\theta_{i}} \right)
\end{aligned}
\end{equation}
with the scale parameters $\lbrace \theta_{i} > 0 , i=1, \,  \ldots \, , n \rbrace$ and $\nu>0$ called shape parameter.\\ Here, $\mathcal{K}_{\nu}$ is the Bessel function of second kind and $\Gamma(\nu)$ is the Gamma function. In practice $\nu=1.5$ and $\nu=2.5$ lead to the most used autocorrelation functions called Matérn 3/2 and Matérn 5/2 respectively.
In closed form Matérn 3/2 yields
\begin{equation}
\begin{aligned}
R (\bm{x} - \bm{x'}, \bm{\theta}, \nu=3/2)  = \prod_{i=1}^{n} \left( 1 + \dfrac{\sqrt{3} \abs{x_{i} - x_{i}'} }{\theta_{i}} \right) \, \exp \left(-\dfrac{\sqrt{3} \abs{x_{i} - x_{i}'} }{\theta_{i}}  \right)
\end{aligned}
\end{equation}
and Matérn 5/2 can be described by
\begin{equation}
\begin{aligned}
R (\bm{x} - \bm{x'}, &\bm{\theta}, \nu=5/2)  = \\&\prod_{i=1}^{n} \left( 1 + \dfrac{\sqrt{5} \abs{x_{i} - x_{i}'} }{l_{i}} + \dfrac{5 (x_{i} - x_{i}')^{2}}{3 \theta_{i}^{2}}   \right) \, \exp \left(-\dfrac{\sqrt{5} \abs{x_{i} - x_{i}'} }{\theta_{i}}   \right) \, \text{.}
\end{aligned}
\end{equation}
Both autocorrelation functions are depicted for different scale parameter values in Figure \ref{fig::Matern}. \\
In practice the squared exponential autocorrelation function is often encountered (see e.g. \cite{aute2013cross}, \cite{bouhlel2017gradient}, \cite{jiang2017adaptive}) in context with Kriging surrogate model construction because of its easy formulation. Such metamodels however inherit an extreme smoothness, which is not necessarily representative of the true mapping and furthermore may lead to ill-conditioning of matrices (see section \ref{sec::limitations}). This is the reason why \cite{laurent2017overview} mention that Matérn autocorrelation functions should generally be the preferred choice. Since computational experiments are the main focus of this thesis and no measurements of input data is available, the Matérn autocorrelation functions will be used in this thesis.
In the following a general form of Kriging surrogate modeling will be summarized.
\subsection*{Best linear unbiased predictor for Kriging}

The \acrfull{blup} offers a prediction for an unobserved quantity $Y_{0}$ of equation (\ref{eq::Regression_model}). For more information about this algorithm see Appendix \ref{sec::BLUP}. It yields that the estimate of the first stage of equation (\ref{eq::Regression_model}) is given by 
\begin{equation}
\hat{\bm{\beta}} = \left( \bm{F}^{T} \bm{R}^{-1} \bm{F}\right)^{-1} \bm{F}^{T} \bm{R}^{-1} \bm{y} \, \text{.}
\end{equation}
The variance estimate of $\bm{\beta}$ yields 
\begin{equation}
\hat{\sigma}^{2} = \frac{1}{m} \left( \bm{y} - \bm{F} \hat{\bm{\beta}} \right)^{T} \bm{R}^{-1} \left( \bm{y} - \bm{F} \hat{\bm{\beta}} \right) \, \text{.}
\end{equation}
These results can be interpreted as the estimates of the generalized least squares linear regression model see (Appendix \ref{seq::LeastLinearRegression}).
Furthermore the best linear unbiased predictor of the metamodel is the Gaussian random variate $\hat{Y}_{0}$ with mean
\begin{equation}
\mu_{\hat{Y}_{0}} = \bm{f}_{0}^{T} \hat{\bm{\beta}} + \bm{r}_{0}^{T} \bm{R}^{-1} (\bm{y}- \bm{F} \hat{\bm{\beta}})
\end{equation}
and minimal variance
\begin{equation}\label{eq::MSE_variance}
\sigma_{\hat{Y}_{0}}^{2}= \sigma^{2} \left( 1 - \bm{r}_{0}^{T} \bm{R}^{-1} \bm{r}_{0} + \bm{u}_{0}^{T} \left( \bm{F}^{T} \bm{R}^{-1} \bm{F}\right)^{-1} \bm{u}_{0}\right),
\end{equation}
where $\bm{u}_{0}$ is given by
\begin{equation}
\bm{u}_{0} = \bm{F}^{T} \bm{R}^{-1} \bm{r}_{0} - \bm{f}_{0} \text{.}
\end{equation}
Furthermore $\bm{f}_{0}$ is the collection of regression functions evaluated at $\bm{x}^{(0)}$.
$\bm{r}_{0}$ describes the cross-correlations between $\bm{x}^{(0)}$ and observations as 
\begin{equation}
r_{0 \, i} = R(\bm{x}^{(0)} - \bm{x}^{(i)}, \bm{\theta}) \, \qquad i=1, \, \ldots \, , m.
\end{equation}
 $\bm{R}$ is the correlation matrix of the observations given by
\begin{equation}
R_{i \, j} = R(\bm{x}^{(i)} - \bm{x}^{(j)}, \bm{\theta}) \, \qquad i, \, j\, =1, \, \ldots \, , m.
\end{equation}
Since the BLUP is a linear combination of Gaussian vectors the estimator has the distribution
\begin{equation}
\hat{Y}_{0} = \bm{a}_{0}^{T} \bm{Y} \sim \mathcal{N} (\mu_{\hat{Y}_{0}}, \sigma^{2}_{\hat{Y}_{0}}) \, \text{.}
\end{equation}
Hence, the predictor inherits convenient properties like a simple formulation of the confidence interval  
\begin{equation}
\hat{Y}_{0} \in \left[ \mu_{\hat{Y}_{0}} - \Phi^{-1} \left( 1 - \frac{\alpha}{2} \right) \sigma_{\hat{Y}_{0}}\,  ;  \, \mu_{\hat{Y}_{0}} + \Phi^{-1} \left( 1 - \frac{\alpha}{2} \right) \sigma_{\hat{Y}_{0}}\right]
\end{equation}
with the probability $1- \alpha$ and where $ \Phi^{-1} (\cdot)$ is the inverse cumulative distribution function of the standard normal distribution. 
\section{Ordinary Kriging}
A simplification of UK is the so-called \acrfull{ok}, see e.g. \cite{kleijnen2009kriging} or \cite{kleijnen2017regression}. Here $\beta_{j} f_{j}(\bm{x}^{(i)})$ of equation (\ref{eq::UK}) is replaced by a constant $\mu$, which can be interpreted as the mean of the term. Therefore instead of a general polynomial trend as in UK, OK sets an unknown mean in the local neighborhood of $\bm{x}^{(0)}$ and hence assumes a stationary constant mean of the underlying random function. The random function is thus defined as
\begin{equation}\label{eq::OK}
\bm{Y}(\bm{x}) =  \mu(\bm{x}) + Z(\bm{x}) \text{.}
\end{equation}
\cite{kleijnen2017regression} points out that OK is used more often than UK in practical applications, which for the author is due to the supplementary needed parameters $\bm{\beta}$ of the latter which often leads to a comparably higher mean squared error. The regression matrix of UK is simplified to the unit vector $\bm{1}$. The generalized least-square estimate of the first stage of equation (\ref{eq::OK}) is then given by
\begin{equation}
\hat{\mu} = (\bm{1}^{T} \bm{R}^{-1} \bm{1})^{-1} \bm{1}^{T} \bm{R}^{-1} \bm{y},
\end{equation}
and the variance estimate of $\bm{\mu}$ yields 
\begin{equation}
\hat{\sigma}^{2} = \frac{1}{m} \left( \bm{y} - \bm{1} \hat{\mu} \right)^{T} \bm{R}^{-1} \left( \bm{y} - \bm{1} \hat{\mu} \right) \, \text{.}
\end{equation}
The best linear unbiased predictor of equation (\ref{eq::OK}) is the Gaussian random variate $\hat{Y}_{0}$, the mean of which reads
\begin{equation}
\mu_{\hat{Y}_{0}}(\bm{x})  =  \hat{\mu} + \bm{r}_{0}^{T}(\bm{x}) \bm{R}^{-1} (\bm{y}- \bm{1} \hat{\mu}),
\end{equation}
and its minimal variance
\begin{equation}
\sigma_{\hat{Y}_{0}}^{2}(\bm{x})  = \sigma^{2} \left( 1 - \bm{r}_{0}^{T}(\bm{x}) \bm{R}^{-1} \bm{r}_{0}(\bm{x}) + \bm{u}_{0}^{T} \left( \bm{1}^{T} \bm{R}^{-1} \bm{1}\right)^{-1} u_{0}(\bm{x})\right)
\end{equation}
with
\begin{equation}
u_{0}(\bm{x}) = \bm{1}^{T} \bm{R}^{-1} \bm{r}_{0}(\bm{x}) - 1 \, \text{.}
\end{equation}
Generally, there is no analytical solution for the estimation of the hyperparameters $\bm{\theta}$. Therefore the correlation matrix consisting of $\sigma^{2} R(\bullet, \bm{\theta})$ is a priori unknown. In order to estimate $\bm{\theta}$ the application-dependent autocorrelation function has to be chosen.  
The most common technique for estimating $\bm{\theta}$ and its variance $\sigma^{2}$ is the so-called \acrfull{mle}. This approach is generally preferred since in contrast to other estimation methods, e.g. Variogram estimation (see \cite{ver1993multivariable} or \cite{chiles1999modelling}) or \acrfull{cv}, it does not depend on the dimension of the input space $\mathbb{X}$. \cite{bachoc2013cross} compared MLE and CV and established that the variance of the surrogate is larger when using CV. Furthermore if the model is correct, i.e. the variance in the observations and the expected output vanished, \cite{bachoc2013cross} points out that MLE always performs better than CV.  MLE will therefore be employed in this thesis. 
\section{Maximum likelihood estimation for the hyperparameters}
MLE tries to maximize the likelihood of the observations $\bm{y}$ defined by the multivariate normal probability density function
\begin{equation}
\begin{aligned}
L(\bm{y} | \bm{\beta}, \sigma, \bm{\theta}) =& 
\dfrac{1}{\left( (2 \pi \sigma^{2})^{m} [\det \bm{R}(\bm{\theta})] \right)^{1/2}} \\ &\exp \left[- \frac{1}{2 \sigma^{2}} (\bm{y} - \bm{F} \bm{\beta})^{T} \bm{R}(\bm{\theta})^{-1} (\bm{y} - \bm{F} \bm{\beta}) \right],
\end{aligned}
\end{equation}
which only depends on $\sigma^{2}, \bm{\beta}$ and $\bm{\theta}$. Maximizing this quantity is equivalent to minimizing its opposite natural logarithm which yields
\begin{equation}\label{eq::oppositeNatu_Log}
\begin{aligned}
- \log L \left( \bm{y} | \bm{\beta}, \sigma^{2}, \bm{\theta} \right) =& \dfrac{1}{2 \sigma^{2}} \left( \bm{y} - \bm{F} \bm{\beta} \right)^{T} \bm{R}(\bm{\theta})^{-1} \left( \bm{y} - \bm{F} \bm{\beta} \right)+ \dfrac{m}{2} \log (2 \pi)\\ 
&+ \dfrac{m}{2} \log (\sigma^{2}) + \dfrac{1}{2} \log \left( [\det (\bm{R}(\bm{\theta}))]\right),
\end{aligned}
\end{equation}
with the maximum likelihood estimates as a function of the hyperparameters given as
\begin{equation}
\begin{aligned}
\hat{\bm{\beta}} (\bm{\theta}) &= \left( \bm{F}^{T} \bm{R}(\bm{\theta}) \bm{F} \right)^{-1} \bm{F}^{T} \bm{R}(\bm{\theta})^{-1} \bm{y} \\
\hat{\sigma}^{2} (\bm{\theta}) &= \dfrac{1}{m} \left( \bm{y} - \bm{F} \hat{\bm{\beta}} \right)^{T} \bm{R}(\bm{\theta})^{-1} \left( \bm{y} - \bm{F} \hat{\bm{\beta}} \right) .
\end{aligned}
\end{equation}
Equation (\ref{eq::oppositeNatu_Log}) can be expressed to be only dependent on the hyperparameters with
\begin{equation}
\begin{aligned}
- \log L \left( \bm{y} | \bm{\beta}, \sigma^{2}, \bm{\theta} \right) &= \dfrac{m}{2} + \dfrac{m}{2} \log (2 \pi) + \dfrac{m}{2} \log \left( \hat{\sigma^{2}}(\bm{\theta}) \right) + \dfrac{1}{2} \log \left([\det R (\bm{\theta})] \right) \\
&= \dfrac{m}{2} \log \left( \psi (\bm{\theta}) \right) + \dfrac{m}{2} \left( \log (2 \pi) + 1 \right) \, \text{.}
\end{aligned}
\end{equation}
Here,
\begin{equation}
\begin{aligned}
\psi (\bm{\theta}) = \hat{\sigma^{2}} (\bm{\theta}) [\det R (\bm{\theta})]^{1/m}
\end{aligned}
\end{equation}
is the so-called reduced likelihood function.
Eventually the maximum likelihood estimate of the hyperparameters $\bm{\theta}$ can be calculated by evaluating an auxiliary optimization problem of the form
\begin{equation}\label{eq::minimization_hyperparameters}
\hat{\bm{\theta}} =   \arg \, \min_{\bm{\theta}^{\star}}  \psi (\bm{\theta}^{\star}) \, \text{.}
\end{equation}
Since there is no analytical solution for the optimization problem, it is necessary to use numerical optimization tools. \cite{bouhlel2017gradient} mention this step as being the most challenging for the construction of surrogate models with Kriging because of the multi-modality of the likelihood function. For this reason, during optimization the hyperparameters are constrained as suggested by \cite{martin2005use}. \\
The software \textit{UQLAB} \citep{lataniotis2015uqlab},
a framework for Uncertainty Quantification in MATLAB of ETH Zurich, which also employs gaussian process modeling has three optimization methods implemented. Firstly, an interior point gradient-based method as described in \cite{byrd1999interior}. Secondly, a genetic algorithm (e.g. \cite{goldberg1989messy}), which is also available as a hybrid version where
the final solution of the genetic algorithm is used as a starting point of the gradient
method that was previously mentioned. And at last a different population based approach known as \acrfull{de} \citep{storn1997differential} as well as multiple hybrid versions of this technique. \\
The DACE toolbox \citep{lophaven2002aspects} uses a modified version of the direct seach algorithm of Hooke and Jeeves \citep{hooke1961direct}. Researchers like \cite{chugh2016surrogate} prefer this solution over population-based techniques due to the
prohibitively high computation time of larger computational models. For a similar reason \cite{bouhlel2016improving} use the derivative-free optimization algorithm COBYLA \citep{powell1994direct}. However, instead of the minimization problem of eq. (\ref{eq::minimization_hyperparameters}) \cite{forrester2008c} solve an equivalent maximization problem for the hyperparameters. The authors point out that a global search method like the genetic algorithm produces the best results for the hyperparameter determination. The comparison of different optimization techniques for the MLE optimization is out of the scope of this thesis. Hence, a hybridized particle swarm optimization similar to the method suggested by \cite{toal2011development} will be employed here. 

\section{Illustrations of Kriging}
This section provides inside into the implications that different autocorrelation functions have on the effective prediction of the stationary Gaussian process and points out differences between UK and OK. Considering the one-dimensional Schwefel function $\mathcal{M}_{Schwfel}^{1d} (x) = x \, \sin (x)$, with $x$ being part of the set $\lbrace x \in \mathbb{R} | 0 \leq x \leq 15 \rbrace $, and $m=10$ equidistant observations.
\subsection{Autocorrelation functions}
\begin{figure}[h!]
\centering
\includegraphics[scale=0.5]{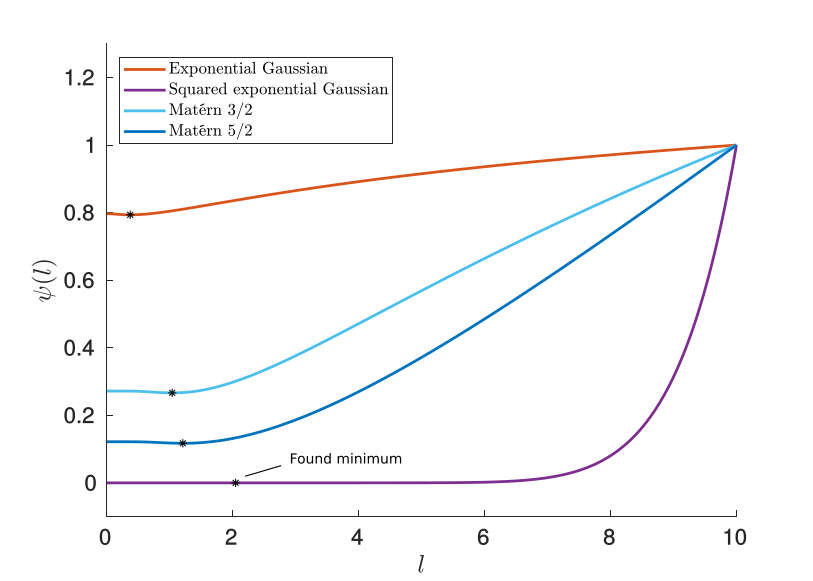}
\caption[Reduced likelihood function for various kernels.]{Normalized form of the reduced likelihood function $\psi$ over hyperparameter $l$ of $\mathcal{M}_{Schwefel}^{1D}$ for different autocorrelation functions. The optimal solution found with differential evolution algorithm is marked with a black dot on each respective line.}\label{fig::OK_ex_psi}
\end{figure}

\begin{figure}[h!]
\centering
\includegraphics[scale=0.5]{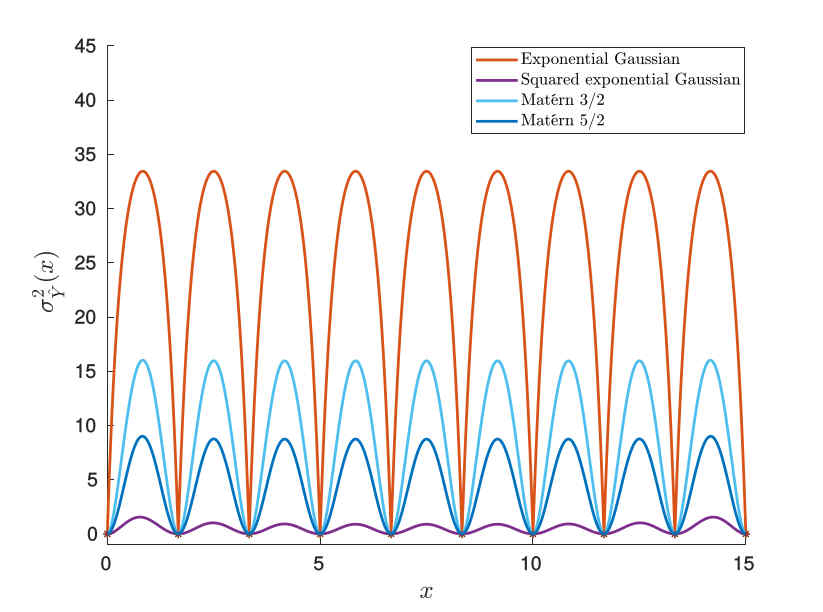}
\caption[Prediction variance of different kernels]{Prediction variance over the input space for the OK approximation of $\mathcal{M}_{Schwefel}^{1D}$ for different kernels. (a) Over the whole space, (b) Close up of (a).}\label{fig::OK_ex_sigma_sq}
\end{figure}

All predictions are done with the OK approach and only the autocorrelation functions are varied in order to investigate their influence. The one-dimensional hyperparameter is fitted using MLE where the optimization was found employing DE. 
\begin{figure}[h!]
\centering
\begin{subfigure}[t]{0.5\textwidth}
\includegraphics[scale=0.35]{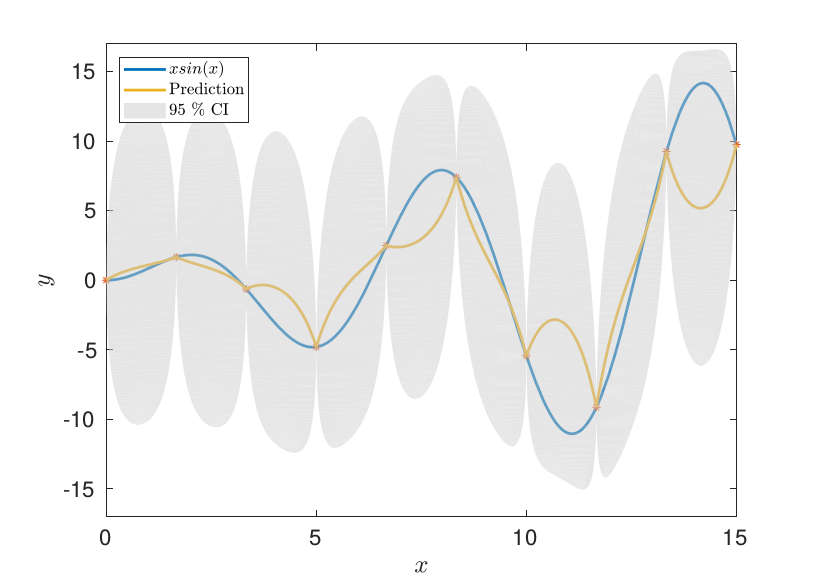}
\subcaption{Gaussian exponential}\label{fig::OK_Rex}
\end{subfigure}%
\begin{subfigure}[t]{0.5\textwidth}
\includegraphics[scale=0.35]{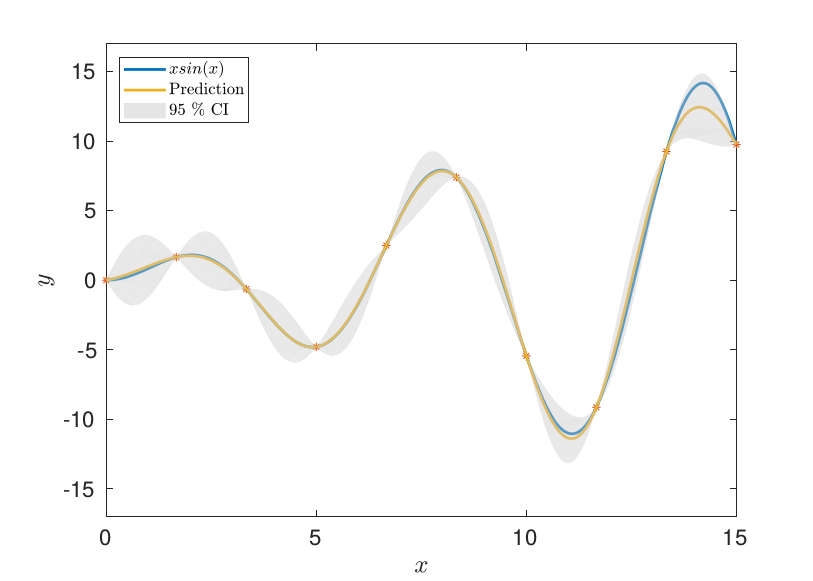}
\subcaption{Gaussian squared exponential}\label{fig::OK_Rsqex}
\end{subfigure}
\begin{subfigure}[t]{0.5\textwidth}
\includegraphics[scale=0.35]{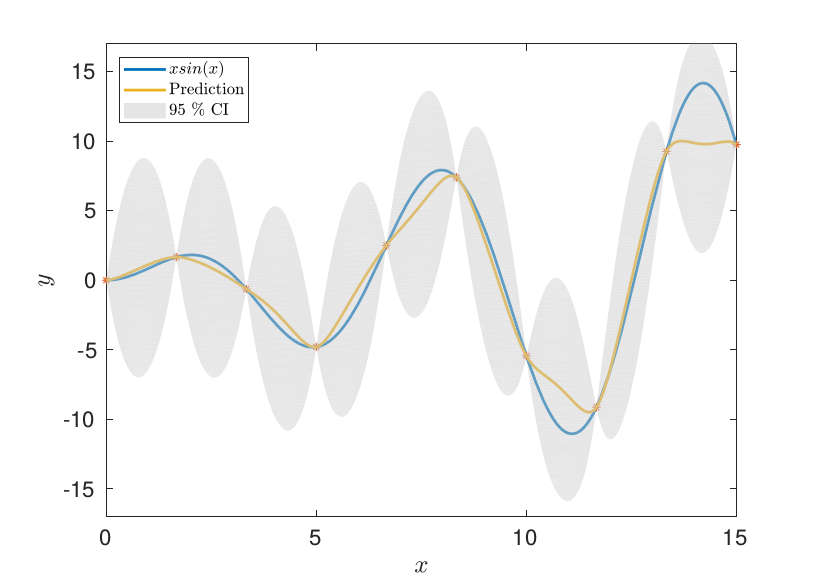}
\subcaption{Matérn 3/2}\label{fig::OK_Matern32}
\end{subfigure}%
\begin{subfigure}[t]{0.5\textwidth}
\includegraphics[scale=0.35]{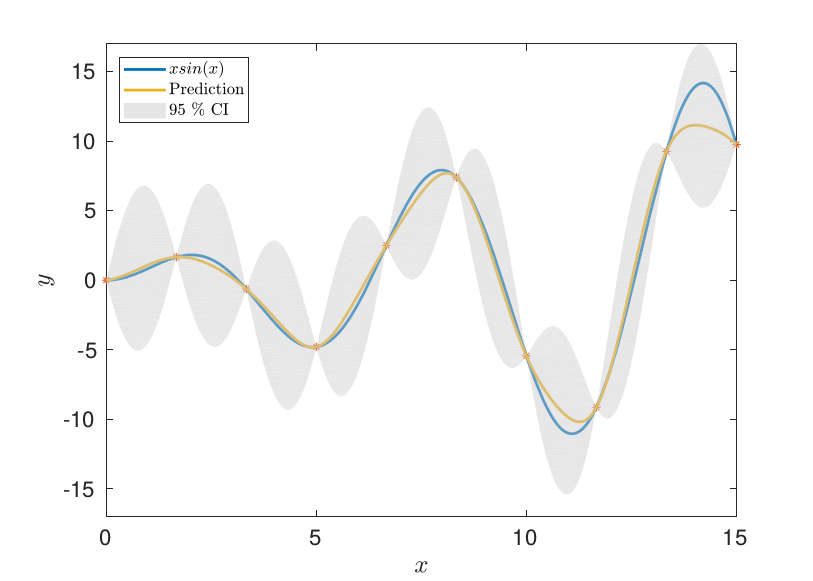}
\subcaption{Matérn 5/2}\label{fig::OK_Matern52}
\end{subfigure}
\caption[Ordinary Kriging models for Schwefel function.]{Ordinary Kriging models for $\mathcal{M}_{Schwefel}^{1D}$ and various autocorrelation functions. The probabilistic predictions (orange curve), the function $\mathcal{M}$ (blue curve) and the $95 \%$ confidence-intervall (grey area) are plotted. (a) Gaussian exponential kernel, (b) Gaussian squared exponential kernel, (c) Matérn 3/2 kernel, (d) Matérn 5/2 kernel}\label{fig::OK_ex}
\end{figure}
The probabilistic predictions as well as the confidence interval for the two introduced power exponential correlation functions as well as the Matérn 3/2 and Matérn 5/2 autocorrelation functions are pictured in Figure \ref{fig::OK_ex}. The $95 \%$ confidence-interval is also depicted as well. The results are highly dependent on the selected correlation kernel. It can be seen that the squared exponential Gaussian holds the least amount of variance, whereas the pure exponential version is the least proficient in terms of prediction quality.
The prediction variance over the input space and depending on the autocorrelation functions is pictured in Figure \ref{fig::OK_ex_sigma_sq}. It can be clearly stated that it depends on the $x$-value positions of the given observations and on the used autocorrelation function.\\
The normalized version of the reduced likelihood function $\psi$ over the hyperparameter $l$ is plotted in Figure \ref{fig::OK_ex_psi}. The normalization was performed with respect to the largest absolute value found. It is noticeable that sharper autocorrelation functions lead to shapes of $\psi$ which are easier to minimize. 
It can be seen that the squared exponential Gaussian autocorrelation function leads to the worst $\psi$ in terms of ease-of-optimization since it features a large plateau which contains the sought optimal solution, which are marked on the respective lines. The Matérn autocorrelation functions offer a trade-off between prediction quality and desirable shape of $\psi$.
\clearpage
\subsection{Difference between UK and OK}
\begin{figure}[h!]
\centering
\begin{subfigure}[t]{0.5\textwidth}
\includegraphics[scale=0.45]{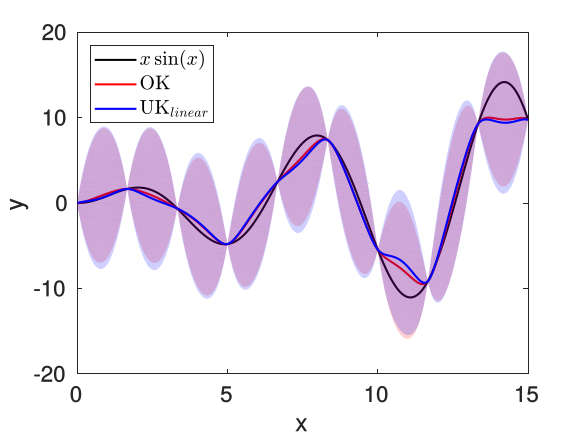} 
\subcaption{Linear regressor}\label{fig::DIFFOKUKLIN}
\end{subfigure}%
\begin{subfigure}[t]{0.5\textwidth}
\includegraphics[scale=0.45]{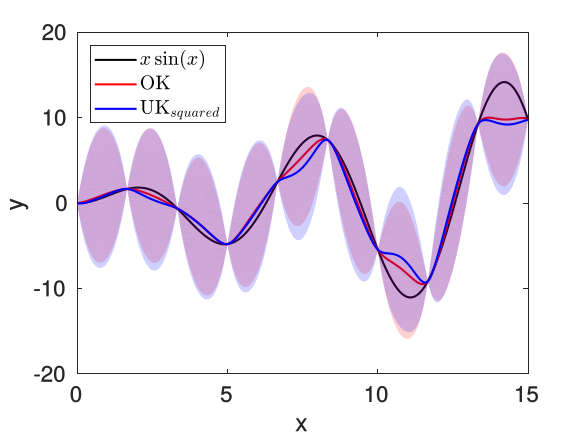} 
\subcaption{Squared regressor}\label{fig::DIFFOKUKSQUARED}
\end{subfigure}
\begin{subfigure}[t]{0.5\textwidth}
\includegraphics[scale=0.45]{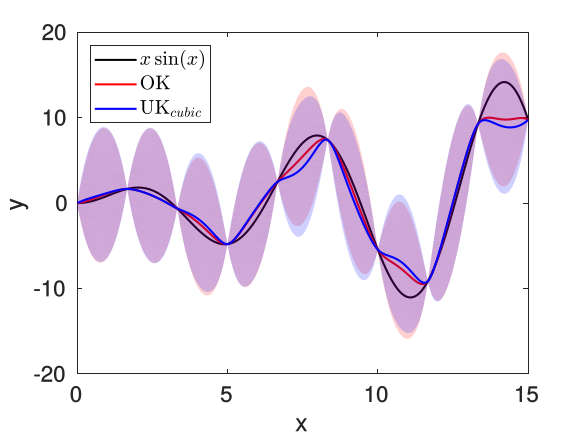} 
\subcaption{Cubic regressor}\label{fig::DIFFOKUKCUBIC}
\end{subfigure}%
\begin{subfigure}[t]{0.5\textwidth}
\includegraphics[scale=0.45]{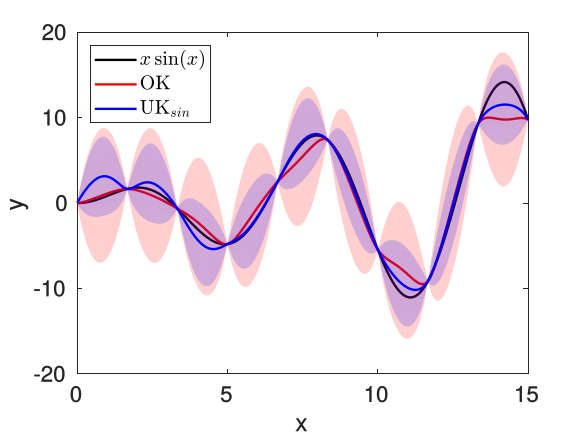} 
\subcaption{Sinus regressor}\label{fig::DIFFOKUKSINUS}
\end{subfigure}
\caption[Function approximation Universal Kriging and Ordinary Kriging]{Differences in function approximation between OK  and different trend functions with UK.
Blue filler indicates the $95 \%$ confidence interval of the respective UK function whereas red filler shows the $95 \%$ confidence interval of UK. (a) Linear Regessor, (b) Squared regressor, (c) Cubic regressor, (d) Sinus regressor.  }\label{fig::DIFFOKUK}
\end{figure}

As explained before the crucial difference between UK and OK is the modeling of the mean of the random function. UK assumes a polynomial trend whereas OK uses a constant mean. The difference in the parameter estimation of an unknown observation lies in the regressor matrix $\bm{F}$ which is the unit vector for the OK case but in UK holds entries that assume the shape of the underlying random function. The matrix can be fully defined by the collection of regression functions $\bm{f} = \left( f_{j}, j = 1, \, \ldots , \, p \right)^{T}$.
Consider four different trends for the problem at hand. A linear model $\bm{f}_{lin} = \left( 1, \abs{x} \right)^{T}$, a quadratic model $\bm{f}_{quadratic} = \left( 1, \abs{x},\abs{x}^{2} \right)^{T}$, a cubic version $\bm{f}_{cubic} = \left( 1, \abs{x},|x|^{2}, |x|^{3}  \right)^{T}$ and a sinusoidal depiction $\bm{f}_{sin} = \left( 1, \sin {x}\right)^{T}$. \\
Both approaches analyzed here utilize Matérn autocorrelation function. MLE has been optimized with DE to find the hyperparameters. 
The function approximations of the metamodels with an OK model are displayed in Figure \ref{fig::DIFFOKUK}. The $95 \%$ confidence intervals are fitted to the color of the respective metamodel. Hence, the  $95 \%$ confidence-interval of OK appears in light red whereas the  $95 \%$ confidence-interval of the UK models appears in light blue. The linear UK approach is compared to OK in Figure \ref{fig::DIFFOKUKLIN}. It can be seen that OK offers a slightly better mean estimation. Furthermore its $95 \%$ confidence-interval also has slightly smaller bounds. The same holds true for the squared UK model (Figure \ref{fig::DIFFOKUKSQUARED}) and the cubic version as seen in Figure \ref{fig::DIFFOKUKCUBIC}. However when the polynomial trend is close to the actual function as it is the case for $\bm{f}_{sin}$ the UK approach yields better approximation results.

\begin{figure}[h!]
\centering
\begin{subfigure}[t]{0.9\textwidth}
\includegraphics[scale=0.7]{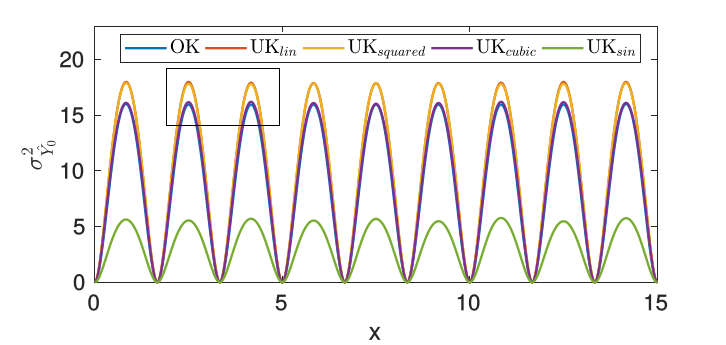} 
\subcaption{Prediction variances of the metamodels.}\label{fig::PredidtVariance3}
\end{subfigure}
\begin{subfigure}[t]{0.9\textwidth}
\includegraphics[scale=0.7]{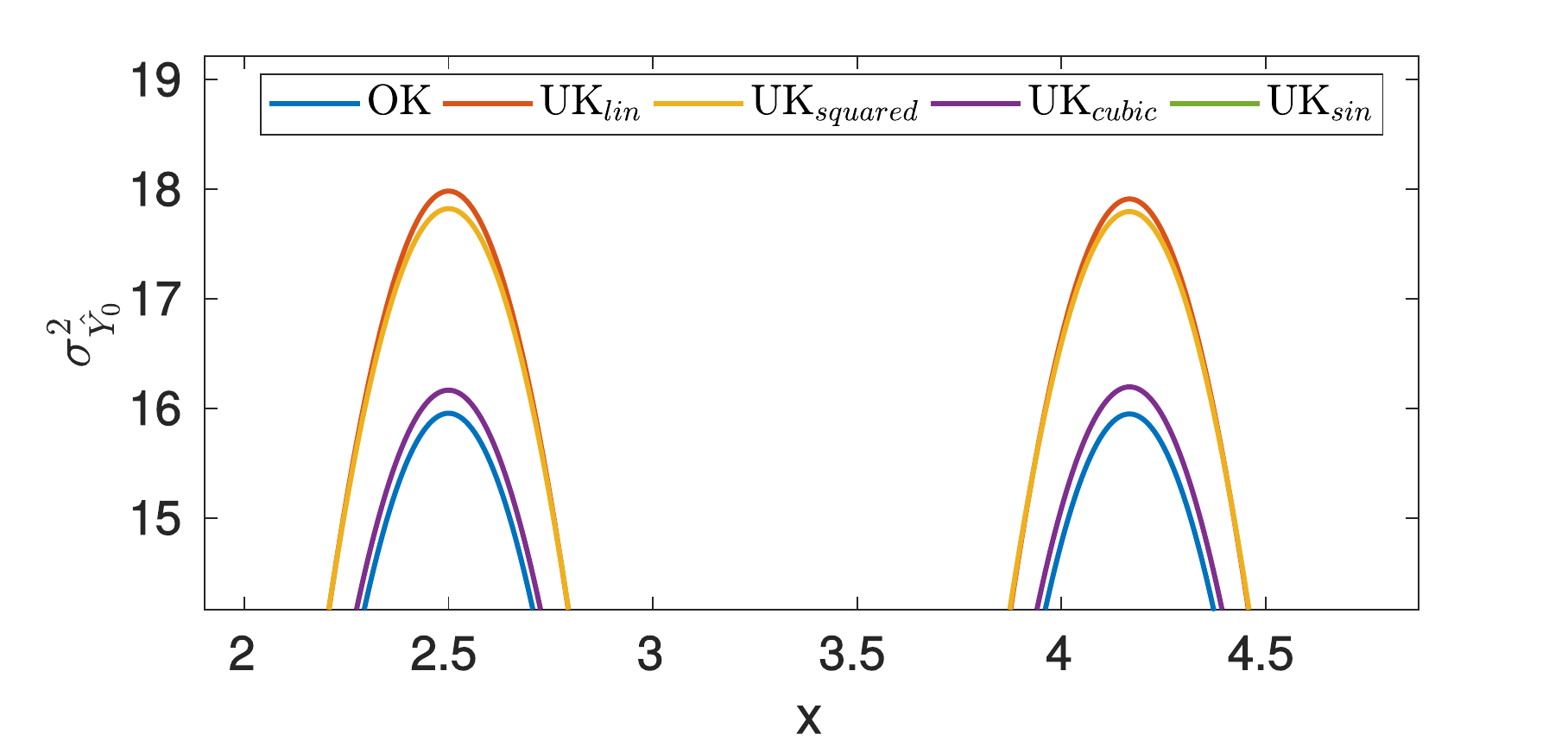} 
\subcaption{Close up of window of (a).}\label{fig::PredidtVariance2}
\end{subfigure}
\caption[Prediction variance Ordinary Kriging and Universal Kriging]{Prediction variance for the OK approximation of $\mathcal{M}_{Schwefel}^{1D}$ for four different UK models defined by their regressor functions. }\label{fig::PredidtVariance1}
\end{figure}

The prediction errors of the UK models and OK over the input space of function of interest $\mathcal{M}_{Schwefel}^{1D}$ are shown in Figure \ref{fig::PredidtVariance1}. A full overview of the error value is shown in the upper figure whereas the focus lies on the area of the rectangle in the lower figure. It can be seen that $\bm{f}_{sin}$ has the lowest prediction error. The OK version however is favored over the other three approaches. \\
This investigation shows that when the underlying polynomial is known and the regressor can be chosen accordingly UK can yield very good results. However this is rarely the case for computer experiments. Therefore OK will be preferred for this work as it offers the best mix between accuracy and time efficiency as no extra weights need to be computed. 
\clearpage
\section{Comparison of metamodeling techniques}\label{ch::ComparisonOKOther}
Ordinary Kriging is compared to commonly found surrogate modeling techniques in this section. The basis of this comparison is the one-dimensional Schwefel function $\mathcal{M}_{Schwefel}^{1D}$ on the domain $[0,35]$. Common metamodel techniques for regression are SVM (e.g. \cite{clarke2005analysis}), RBFN \citep{park1991universal}, \acrfull{nn} in form of multilayer perceptron and PR (see e.g. \cite{edwards2002alternatives}). \\
For the comparison OK is utilized using the Matérn 3/2 autocorrelation function. \\
The SVM is employed using the radial basis kernel as well as an automatic choice of the scale value for the kernel function as defined in the MATLAB software. For the RBFN the hidden layer has 10 neurons. The centers are calculated with K-means clustering algorithm and the widths are set to unity. 
\begin{figure}
\centering
\begin{subfigure}[t]{1.0\textwidth}
\includegraphics[scale=0.7]{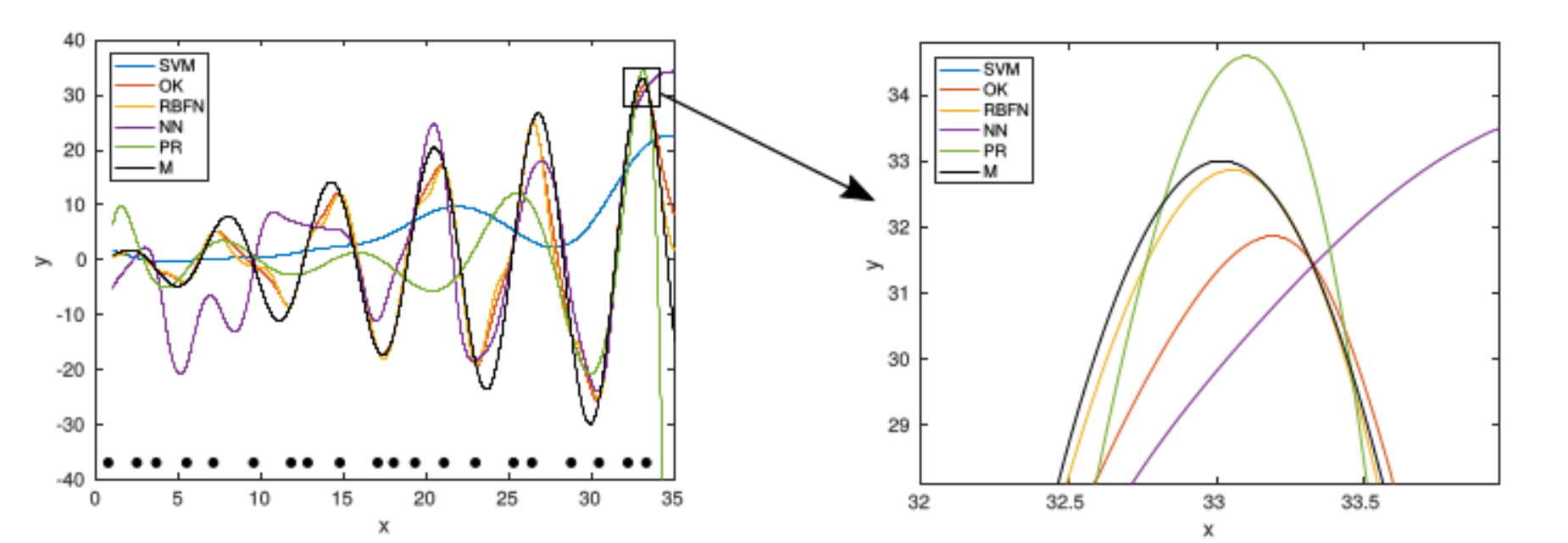} 
\subcaption{20 samples}\label{fig:Compare20}
\end{subfigure}
\begin{subfigure}[t]{1.0\textwidth}
\includegraphics[scale=0.7]{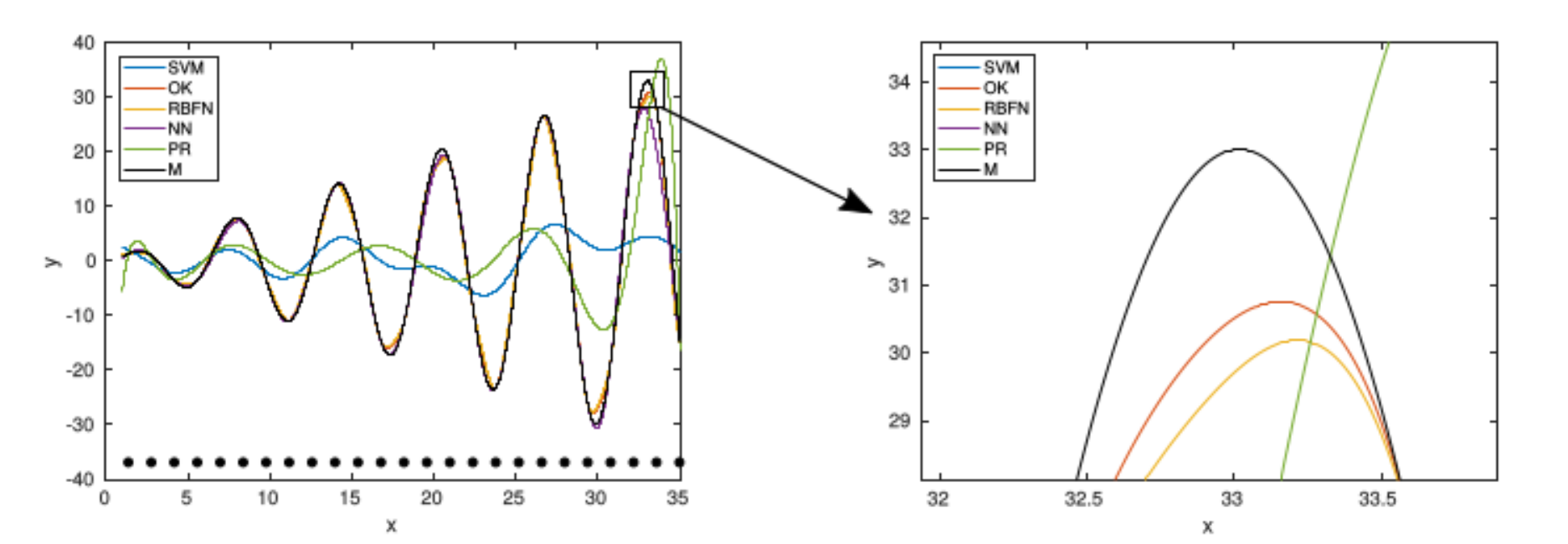} 
\subcaption{25 samples}\label{fig:Compare25}
\end{subfigure}
\begin{subfigure}[t]{1.0\textwidth}
\includegraphics[scale=0.7]{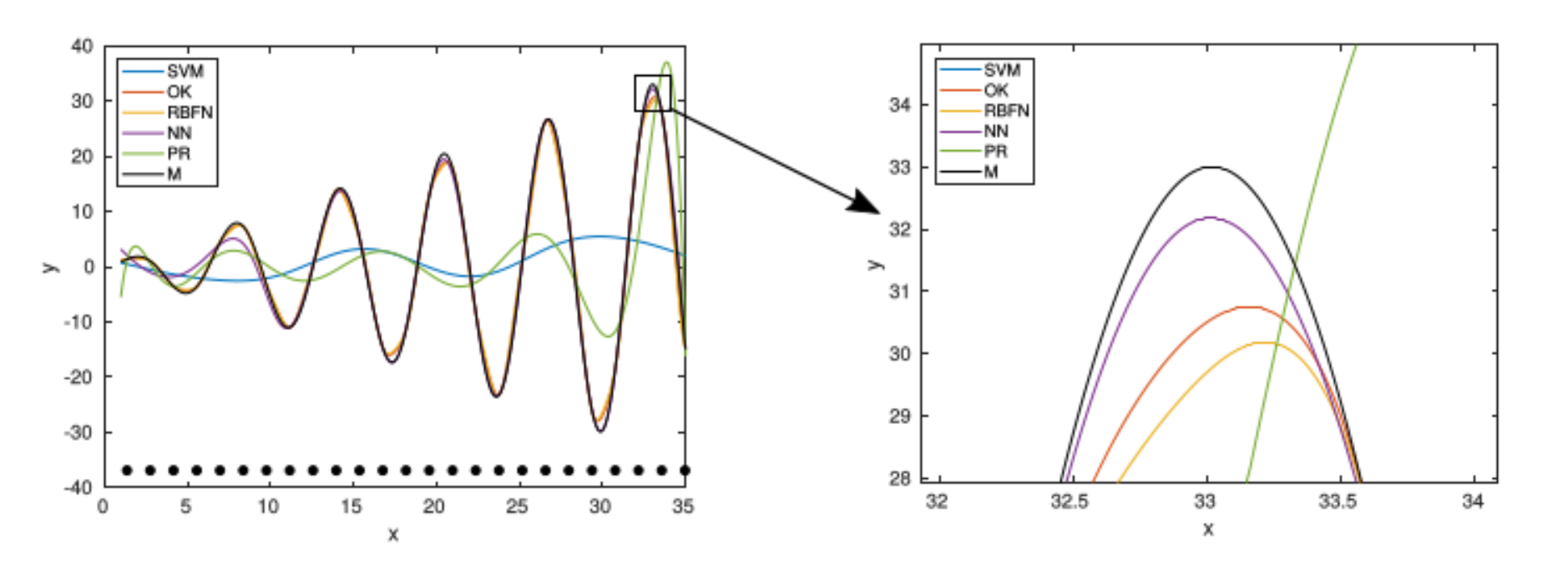} 
\subcaption{30 samples}\label{fig:Compare30}
\end{subfigure}
\caption[Comparison of different metamodel techniques for $\mathcal{M}_{Schwefel}^{1D}$ ]{Comparison of different metamodel techniques for $\mathcal{M}_{Schwefel}^{1D}$ and 20, 25 and 30 samples respectively. The sample locations are indicated by the dots below the curves. Left hand-side corresponds to the whole domain. Right hand-side illustrates the domain in of the highlighted box. }\label{fig:Compare}
\end{figure}

\begin{figure}[hbtp]
\centering
\includegraphics[scale=0.4]{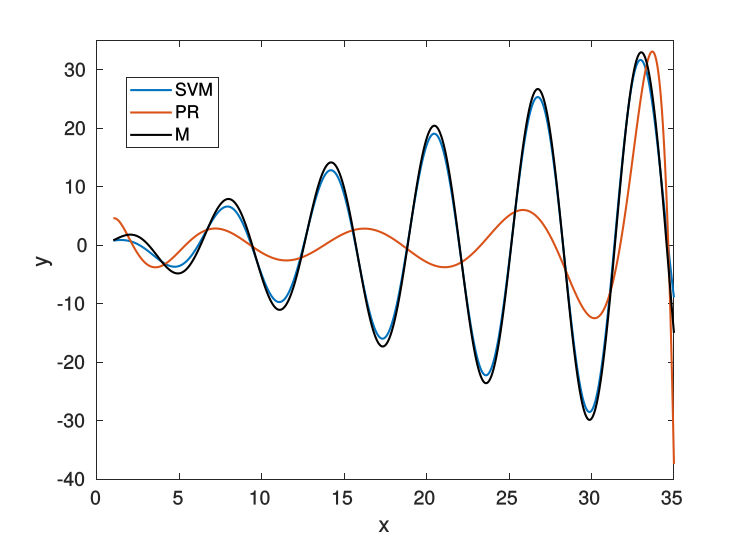}
\caption[Comparison of SVM and PR for $\mathcal{M}_{Schwefel}^{1D}$]{Comparing SVM and PR for 300 sample points in the domain for $\mathcal{M}_{Schwefel}^{1D}$.}\label{fig::SVM300}
\end{figure}

For the polynomial regression a seven degree best fit (in a least-squares sense) for the data is used. The NN is employed with 1 hidden-layer with 20 neurons and $tanh$ activation function where Bayesian regularization is used as the training function.
The approximation results of $\mathcal{M}_{Schwefel}^{1D}$ for the 5 given metamodeling techniques for 20, 25 and 30 sample points respectively are shown in Figure \ref{fig:Compare}. Here, the right-hand sides show the results for the whole input domain. Whereas the left-hand sides focus on the domain as highlighted with the rectangle. It can be seen that SVM and PR are not able to obtain a good approximation with the given data. OK, RBFN and NN show around the same approximation capabilities. With increasing sample size the error gets reduced.
The \acrfull{mae} of the three sample examples is listed in Table \ref{table::CompareMAE}.
The MAE is defined by
\begin{equation}
\text{MAE} = \dfrac{1}{m} \sum_{i=1}^{m} \abs{y_{i} - \hat{Y}_{i}},  
\end{equation}
where $m$ is the number of design points, $y_{i}$ are the exact responses and the estimated response values are given by $\hat{Y}_{i}$.
The results indicate that OK yields the best approximation of the target function for all three sample sizes.
\begin{table}[ht!]
\begin{center}
\begin{tabular}{cr cr cr cr cr} \hline
Metamodel method & 20 Samples & 25 Samples & 30 Samples  \\ \hline\hline \\
SVM  &  11.1664 & 9.9002 & 9.7856  \\
OK &  \textbf{2.6769} & \textbf{0.6724} & \textbf{0.1553} \\
RBFN &  3.098 & 0.8872 &0.2149  \\
NN & 7.6794 & 0.6995 & 0.3571   \\ 
PR &  11.5258 & 9.4992& 9.4619 \\
\end{tabular}
\end{center}
\caption[Comparison of the mean absolute error of different metamodeling techniques for $\mathcal{M}_{Schwefel}^{1D}$]{Comparison of the mean absolute error for $\mathcal{M}_{Schwefel}^{1D}$ of 5 metamodeling techniques for different sample sizes.}\label{table::CompareMAE}
\end{table}
As a side-note as illustrated in Figure \ref{fig::SVM300} SVM needs around 300 sample points to yield equivalent results to OK, RBFN and NN. PR is not able to proficiently fit the data even with 300 sample points.

\clearpage
\section{Limitations of Kriging}\label{sec::limitations}
The limitations and challenges when utilizing Kriging as a surrogate model approach for computer experiments are summarized in the following section.
\subsection{High dimensional problems}
Problematic with the Kriging approach for building surrogate models is the fact that the models are only limitedly suited for large dimensional problems ($n$ > 15). The reason for this is the resulting prohibitively high computational complexity.
Recent works in the field have addressed this problem (see e.g. \cite{bouhlel2016improving} or \cite{damianou2013deep}).
\cite{chugh2016surrogate} and \cite{hensman2013gaussian} mention that the complexity of training in Kriging is $O(m^{3})$, where $m$ is the size of training data. Furthermore when the hyperparameters are determined by using MLE the issue is even more apparent \citep{chugh2016surrogate}. \cite{liu2014gaussian} determined that when constructing a Kriging model for a 50-dimensional benchmark problem for a training set of $m=150$ it takes between 240-400s to obtain the hyperparameters using the MATLAB optimization toolbox on a Xeon 2.66 GHz computer. When the number of training data is increased to 250 and 550 the computation time takes between 1000-1800s and 12000-20000s respectively. These results suggest that using Kriging for
computationally efficient surrogates is no longer viable for
high-dimensional problems. \\
Especially the optimization of the subproblem, MLE, can be an issue because of the matrix inversion needed. Solutions are to reduce the number of hyperparameters, e.g. proposed by \cite{mera2007efficient}, albeit including the assumption that the applicable Kriging models employ the same dynamics in all directions. \cite{bouhlel2016improving} propose to introduce partial least squares dimension reduction to effectively reduce the number of hyperparameters and speed up the surrogate construction.
\subsection{Degeneracy of the covariance}
A second problem is the degeneracy of the covariance and therefore correlation matrix, which makes the inversion of the correlation matrix with numerical tools unreliable because some columns of the matrix are almost identical. This may occur in case of linear dependencies between covariances of subsets of sample points, which is commonly found when observed points are too
close to each other. Generally this occurs when the information of the observations is made redundant by the covariance function.
This is often found in Kriging procedures involving sequential sampling since the sample points tend to pile up around the region's of interest (for more information see section \ref{sec::adaptiveSampling}). \\
Mainly, three different solutions were presented in the literature to circumvent this problem:
\begin{itemize}
\item by controlling the location of generated sample points to guarantee good conditioning (see e.g. \cite{osborne2009gaussian} or \cite{rennen2009subset}),
\item by selecting the correlation function so that respective matrix avoids to become ill-conditioned. Influences on the condition number are researched in \cite{davis1997six},
\item by regularizing the covariance matrix. This can for example be done by introducing a pseudoinverse \citep{mohammadi2016kriging} or by adding a small scalar to the diagonal, which is called a "nugget" (see e.g. \cite{booker1999rigorous} or \cite{santner2013design}).
\end{itemize}
The pseudoinverse and the 'nugget' approach are the literature standard, since in contrast to the other approaches they can be used a posteriori in computer experiments algorithms without major redesign of the methods \citep{mohammadi2016analytic}. 
Approaches involving the pseudoinverse and and a regularization involving "nuggets" are compared in \cite{mohammadi2016analytic}.
\section{Concluding remark}
The mathematical background of Kriging has been presented. The general properties have been explained and differences between Universal and Ordinary Kriging have been highlighted. MLE has been explained as a mean to compute the hyperparameters of the metamodel process. Limitations of the Kriging approach and solutions to these limitations employed in this thesis have been described. As a result of the discussions the following modeling choices will be made:
\begin{itemize}
\item Ordinary Kriging will be utilized.
\item The 3/2 Matérn autocorrelation function will be employed.
\item The MLE will be utilized for the determination of the hyperparameters. The given optimization problem will be solved with a hybridized particle swarm optimization procedure.
\end{itemize}

\chapter{Adaptive Sampling techniques}\label{sec::adaptiveSampling}
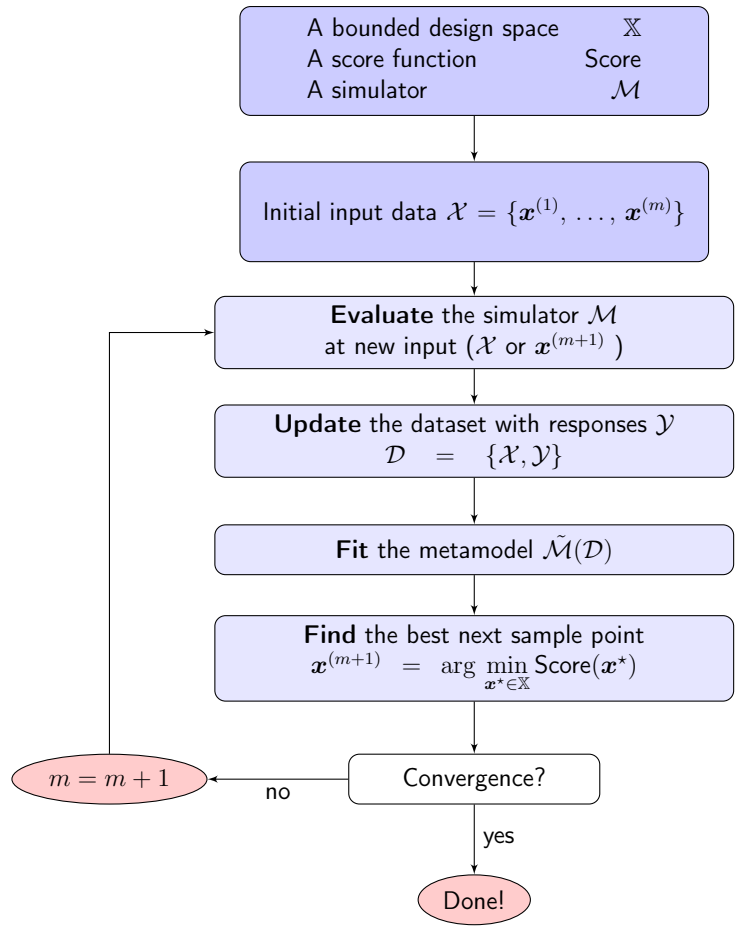
\begin{figure}[h!]
\tikzstyle{decision} = [rectangle, draw,  
    text width=10em, text centered, rounded corners, node distance=2cm, inner sep=0pt,minimum height=2em]
\tikzstyle{block} = [rectangle, draw, fill=blue!20, 
    text width=18em, text centered, rounded corners, minimum height=4em,node distance=4cm,]
\tikzstyle{blockA} = [rectangle, draw, fill=blue!10, 
    text width=20em, text centered, rounded corners, minimum height=2em]
\tikzstyle{line} = [draw, -latex']
\tikzstyle{cloud} = [draw, ellipse,fill=red!20, node distance=2cm,
    minimum height=2em]
 \centering 
\begin{tikzpicture}[node distance = 1.8cm, auto,scale=0.8, transform shape]
    \node [block] (init) {
\begin{tabular}{ l r }
A bounded design space  & $\mathbb{X}$  \\
A score function & $\text{Score}$ \\
A simulator & $\mathcal{M}$ 
\end{tabular}
};
    \node [block, below of=init,node distance=2.5cm] (start) {Initial input data $\mathcal{X} =\lbrace \bm{x}^{(1)}, \, \ldots  , \, \bm{x}^{(m)}  \rbrace$};
    \node [blockA, below of=start,node distance=2.0cm] (Evaluate) {\textbf{Evaluate} the simulator $\mathcal{M}$\\ at new input ($\mathcal{X}$ or $\bm{x}^{(m+1)}$ )};
    \node [blockA, below of=Evaluate] (Update) {\textbf{Update} the dataset with responses $\mathcal{Y} $ \\
    $\mathcal{D} = \lbrace \mathcal{X}, \mathcal{Y}  \rbrace$};
    \node [blockA, below of=Update] (Fit) {\textbf{Fit} the metamodel $\tilde{\mathcal{M}}(\mathcal{D} )$  };    
        \node [blockA, below of=Fit] (Find) {\textbf{Find} the best next sample point
        \\$
        \bm{x}^{(m+1)} = \arg \, \min\limits_{\bm{x}^{\star} \in \mathbb{X}} \text{Score} (\bm{x}^{\star})
        $
         }; 
          \node [decision, below of=Find] (Accurate) {Convergence?};     
     \node [cloud, left of=Accurate,node distance=6.0cm] (newM) {$m = m+1$};
          \node [cloud, below of=Accurate] (Done) {Done!};

    \path [line] (init) -- (start);
     \path [line] (start) -- (Evaluate);
    \path [line] (Evaluate) -- (Update);
     \path [line] (Update) -- (Fit);
     \path [line] (Fit)  -- (Find);
     \path [line] (Find) -- (Accurate);
          \path [line] (Accurate) -- node {no}(newM);

          \path [line]
    (newM.north)  |-  (Evaluate.west);
                    \path [line] (Accurate) -- node {yes}(Done);

\end{tikzpicture}
\caption[Workflow of creating a surrogate model with an adaptive sampling approach]{General workflow of creating a surrogate model with an adaptive sampling approach.}\label{fig::flowchart_adaptiveSampling}
\end{figure}

In this chapter the selection of the best design of experiments\\$\mathcal{X} = \left( \bm{x}^{(i)}, \, i=1, \, \ldots, \, m \right)$ in context of the emulator $\tilde{\mathcal{M}}$ providing an accurate representation of $\mathcal{M}$ over the entire input space $\mathbb{X}$ is investigated.\\
A created emulator should contain as few sample points $m$ as possible. Two general arguments substantiate this thesis. Firstly, a surrogate model is constructed to emulate a computationally expensive simulation model. In order to be viable the building cost of a metamodel together with the resources of performing an analysis of interest (e.g. optimization, reliability analysis) with this model needs to be far smaller than an equivalent computation with the expensive-to-evaluate simulator $\mathcal{M}$. Secondly, when the number of sample points in the dataset $\mathcal{D}$ gets too large a constructed $\tilde{\mathcal{M}}$ becomes computationally inefficient and even insoluble
(see e.g. \cite{dubourg2011adaptive}). \\
Because the mapping $\mathcal{M}$ is assumed to be a black box, the size of the experimental design for the desired accuracy is difficult to predict.  
Sequential design and in particular adaptive sampling offers a solution to this problem since it selects samples in an iterative procedure. 
Adaptive sampling techniques can be distinguished by the number of new sample points that are iteratively added. \cite{liu2017survey} use the term single selection to denote that one sample point is generated per iteration. In contrast batch selection describes the case when more than one sample point is added to the dataset per iteration. This approach is of interest when working with parallelization of surrogate model construction. However, \cite{liu2017survey} mention that single selection adaptive sampling approaches are the main focus of research in the literature because new points are commonly determined with the help of an auxiliary optimization procedure. Hence, single selection approaches will also be employed in this thesis. 
The general workflow of a single selection adaptive sampling technique used for global metamodeling is depicted in Figure \ref{fig::flowchart_adaptiveSampling}.

\newpage
Consider a set of initial data $\mathcal{D}_{ini} = \lbrace \left( \bm{x}^{(i)}, \,\bm{y}^{(i)} \right), \, i=1, \, \ldots  , \, m  \rbrace$. The creation of the surrogate model begins by fitting $\tilde{\mathcal{M}}$ to this data. New points are iteratively added to the dataset $\mathcal{D}$ until a convergence criterion is reached. 
These points are generated based on an adaptive sampling criterion, which solves an optimization problem. 
Assume single selection so that a single new sample point $\bm{x}^{(m+1)}$ is generated in each iteration then this point is determined by minimizing a score function or refinement criterion of the form
\begin{equation}
\bm{x}^{(m+1)} = \arg \, \min_{\bm{x}^{\star} \in \mathbb{X}} \, \text{Score} \left( \bm{x}^{\star} \right) \, \text{.}
\end{equation}
\begin{figure}[h!]
\centering
\begin{subfigure}[t]{0.75\textwidth}
\includegraphics[scale=0.4]{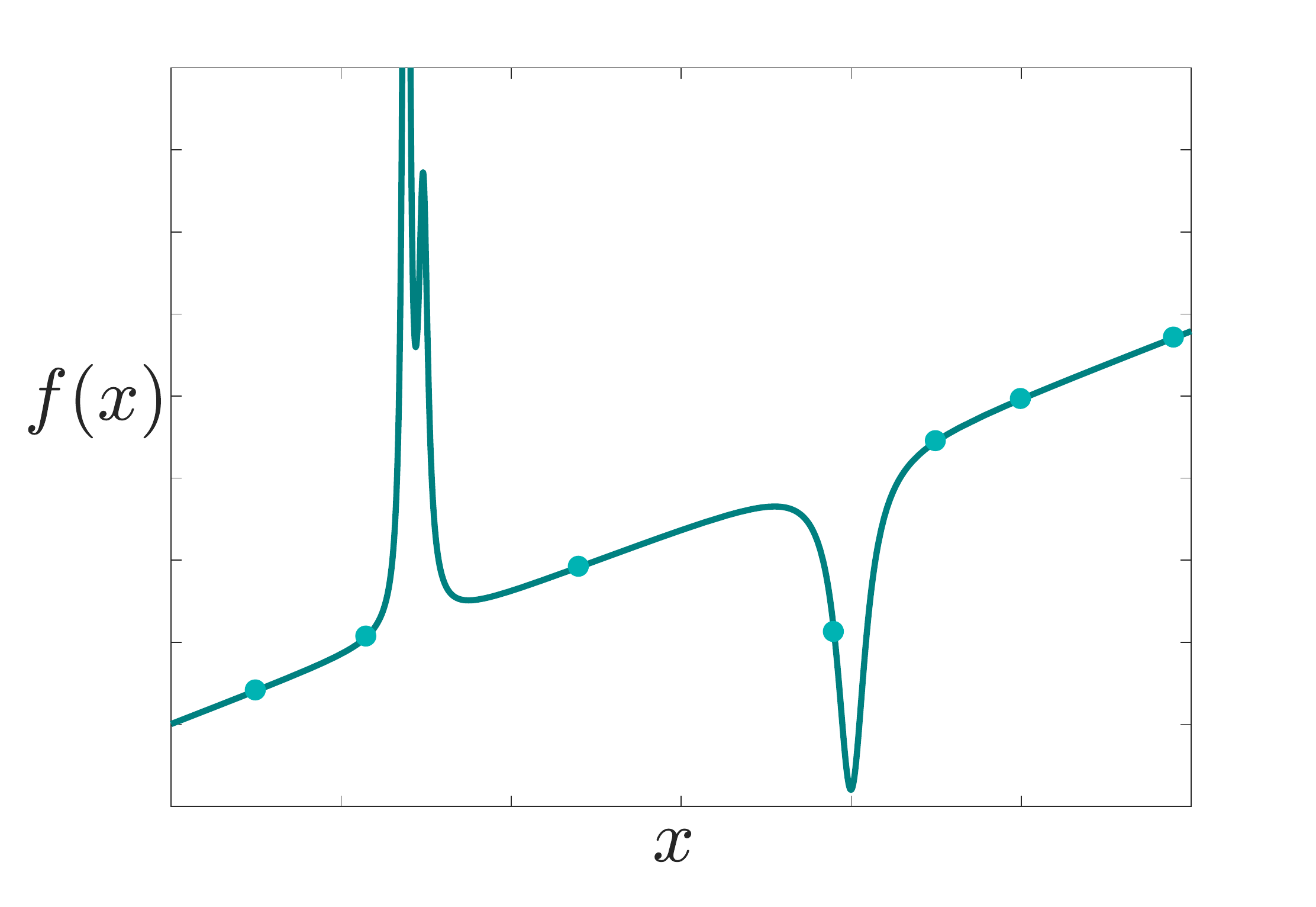} 
\subcaption{Initial dataset}\label{fig:Ex_Ex_initial}
\end{subfigure}
\begin{subfigure}[t]{0.5\textwidth}
\includegraphics[scale=0.28]{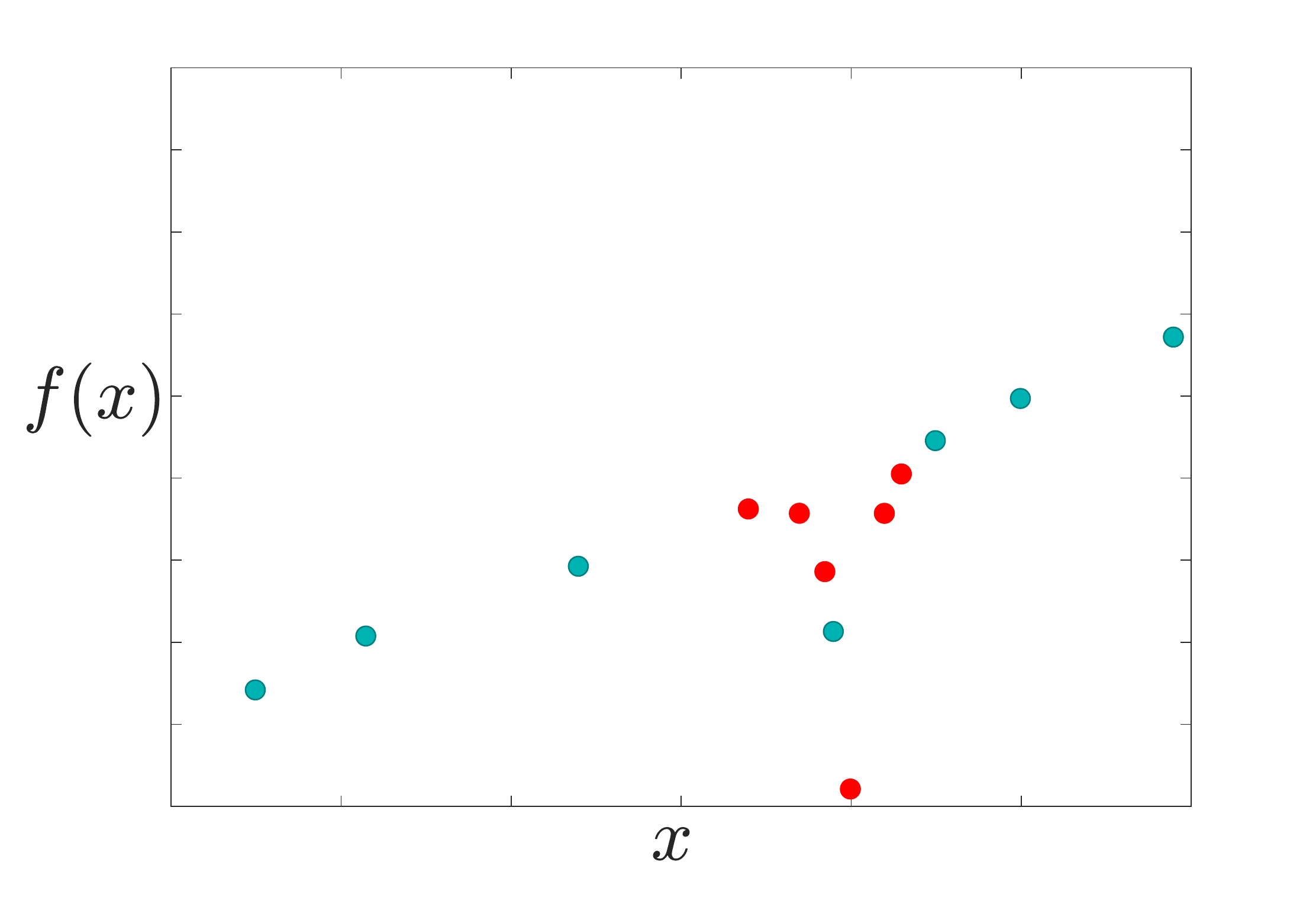}
\subcaption{Local Exploitation}\label{fig:Ex_Ex_Exploitation}
\end{subfigure}%
\begin{subfigure}[t]{0.5\textwidth}
\includegraphics[scale=0.28]{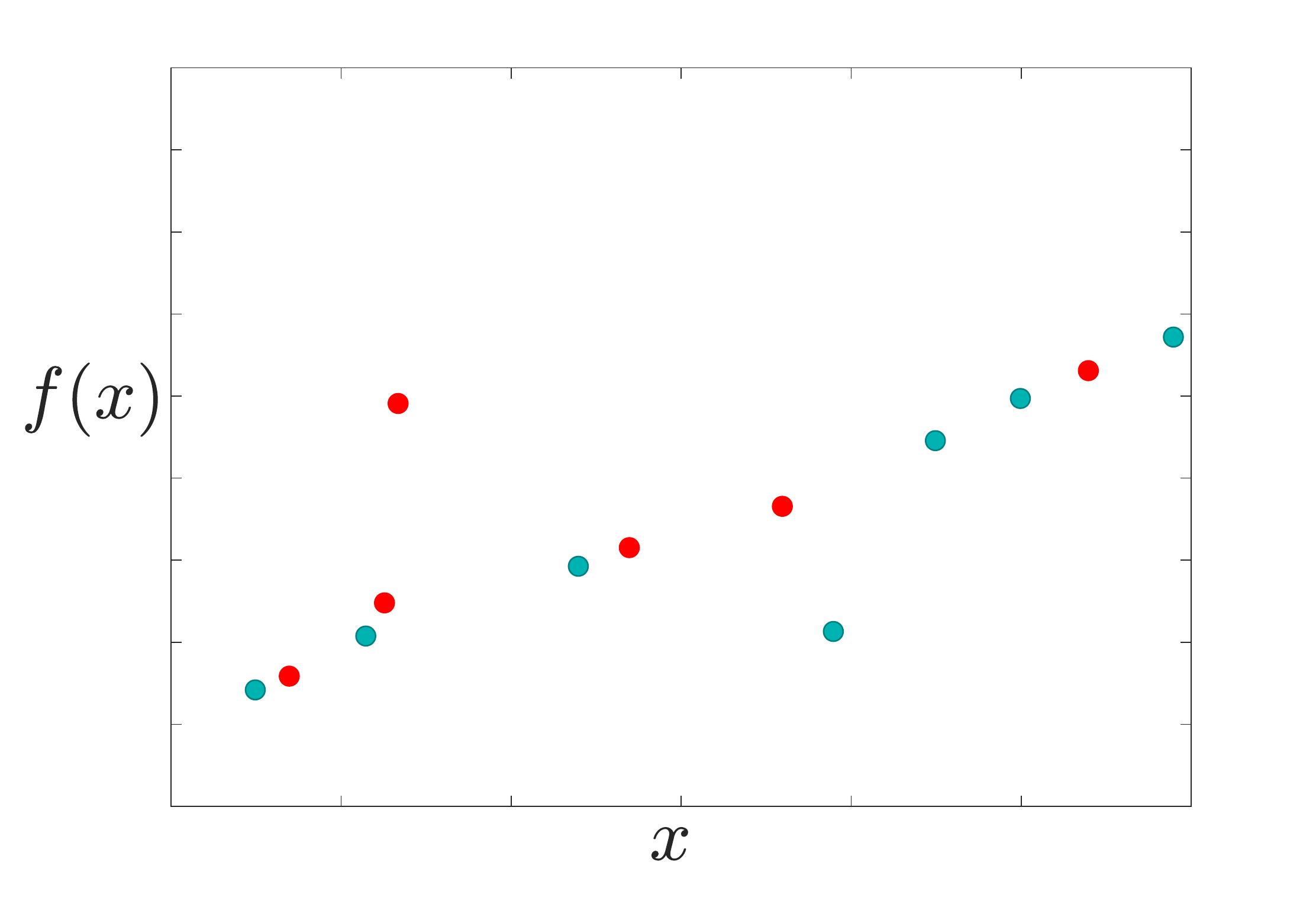}
\subcaption{Global Exploration}\label{fig:Ex_Ex_Exploration}
\end{subfigure}
\caption[Essential trade-off for sampling based approaches]{Essential trade-off for sampling based approaches. Local exploitation and global exploration. (a) Initial dataset with black box function. From a first look the assumption can be made that the function is linear except for one sample in the middle, (b) Local exploitation sets a focus on the nonlinear area and create samples there but missing the second nonlinearity to the left, (c) Global exploration explores the design space evenly and will create a sample in the other nonlinear region, inspired by \cite{crombecq2009space}}%
\label{fig:Ex_Ex}%
\end{figure}
\newpage
Generally, as illustrated in Figure \ref{fig:Ex_Ex}, an adaptive sampling approach needs to consider a trade-off between two conflicting parts, 
namely, local exploitation and global exploration:
\begin{itemize}
\item \textit{Global exploration:} this part aims to explore the domain evenly in order to detect unknown regions of interest, i.e. nonlinear behavior such as discontinuities or optima. This way regions with large prediction error can be identified. Global exploration is independent of the evaluated responses.  
\item \textit{Local exploitation.} This part is key for adaptive sampling approaches. It aims to generate data points in (already identified) interesting regions to reduce the prediction error there. The prediction error however is a priori unknown. As an example consider a global minimum in the black box. In order to accurately represent the minimum with the metamodel the sampling process needs to focus on generating more samples around this domain. 
In contrast to global exploration local exploitation involves the evaluated outputs of earlier iterations in order to guide the sampling process.
\end{itemize} 
Some proposed methods just focus to optimize one of the effects.
For example the global Monte Carlo method \acrfull{mipt} which was developed by \cite{crombecq2011efficient} purely relies on exploration. The idea is to find the best new adaptive sampling point by finding the best possible candidate out of a large number of randomly generated points by just considering the distance to existing sampling points. Another example is the \acrfull{msd} method \citep{jin2002sequential}, which aims to maximize the minimum distance between existing points. \\
However for an optimization of the sampling procedure, exploration and exploitation can be combined to yield a function of the form \citep{liu2017survey}
\begin{equation}
\text{Score}(local(\bm{x}),global(\bm{x})) = w_{\text{local}} \, \cdot \,  local(\bm{x}) + w_{\text{global}} \, \cdot \,  global(\bm{x}) \, \text{.}
\end{equation}
Here $w_{\text{local}}$ is the weight for the local exploitation, whereas $w_{\text{global}}$ represents the corresponding value for global exploration. The summation of the values yields unity. $local(\bm{x})$ and $global(\bm{x})$ are user-defined functions specifying the chosen adaptive sampling technique. This definition leads to a refinement criterion optimization problem for the new sample point of the form
\begin{equation}\label{eq::}
\bm{x}^{(m+1)} = \arg \, \min_{\bm{x}^{\star} \in \mathbb{X}} \, \text{Score} \left( local(\bm{x}^{\star}), global(\bm{x}^{\star}) \right) \, \text{.}
\end{equation}
Inspired by the work of \cite{liu2017survey} three different strategies for flexible balance strategies can be summarized.
\begin{itemize}
\item \textit{Decreasing strategy.} The strategy depicted in Figure \ref{fig:Ex_ex_decreasing_strategy} works as follows. The global weight $w_{\text{global}}$ is near $1$ at the beginning of the metamodel construction. This leads to a reasonable exploration of the input space in order to determine regions of interest. During the computation the $w_{\text{global}}$ decreases while $w_{\text{local}}$ increases until the latter is near zero at the end of the metamodel construction. This implies pure local exploitation of the interesting regions at the end of the computation. Two applications of this strategy are given in \cite{turner2007multidimensional} and \cite{kim2009construction}.
\item \textit{Greedy strategy.} Here, a strategy is introduced which governs a switch between pure exploitation and exploration. (see for example \cite{sasena2002flexibility} and \cite{sasena2002adaptive}). An example procedure is depicted in Figure \ref{fig:Ex_ex_greedy_strategy}. Initially, starting with full global exploration the uncertainty of the entire domain is reduced. Then after the switch the local exploitation starts to efficiently enhance the accuracy of the model locally. If the improvement of accuracy in the local domains is considered sufficient the procedure switches back to global exploration to find new undiscovered regions of interest. Therefore with as few points as possible the strategy iteratively swaps between global exploration and local exploitation.
\item \textit{Switch strategy.} This strategy employs dynamic switching between local and global weights. Herein, the weights are changed by taken information from previous iterations into account, i.e. the differences between prediction errors. This strategy is visualized for an example in Figure \ref{fig:Ex_ex_switch_strategy}. In their paper \cite{singh2013balanced} show that this procedure works more efficiently than the decreasing and greedy strategies. It has also been employed in \cite{liu2017adaptive}.
\end{itemize}
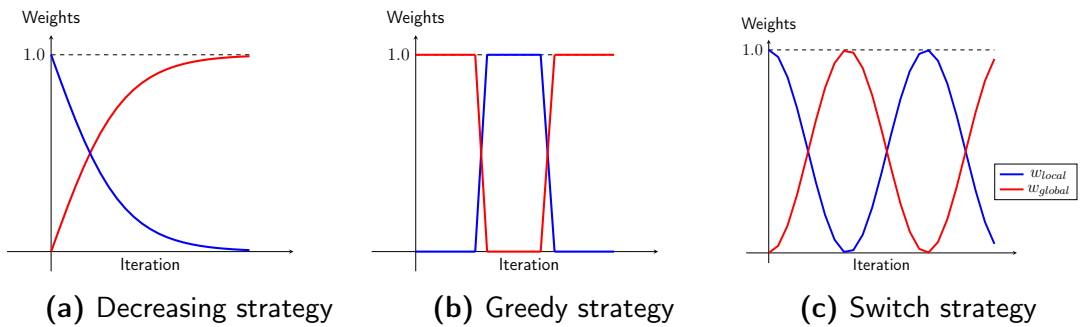
\begin{figure}[ht!]
\centering
\begin{subfigure}[t]{0.35\textwidth}
\begin{tikzpicture}[scale = 0.55, transform shape]
\begin{axis}
[
 axis equal,
xlabel={Iteration},
ylabel={Weights},
y label style={anchor=north},
axis lines=middle,
xtick={0},
xticklabels={},
    ytick={0},
    ymin=-0.1,ymax=1.1,
    xmin=-0.1,xmax=1.1,
    axis on top=false,
every axis x label/.style={
    at={(ticklabel* cs:0.5)},
    anchor=north,
},
every axis y label/.style={
    at={(ticklabel* cs:1.01)},
    anchor=south,
},
legend style={at={(axis cs:1.0,0.5)},anchor=west}
]
\draw[dashed] (axis cs:0.0,1.0) -- (axis cs:1.0,1.0);
\addplot [black, nodes near coords=$1.0$,every node near coord/.style={anchor=0},forget plot] coordinates {( 0, 1)};
\addplot [domain=0.0:1.0,red,line width = 0.5mm]{tanh(2.8*x)};
\addplot [domain=0.0:1.0,blue,line width = 0.5mm]{1-tanh(2.8*x)};
\end{axis}
\end{tikzpicture}
\subcaption{Decreasing strategy}\label{fig:Ex_ex_decreasing_strategy}
\end{subfigure}%
\begin{subfigure}[t]{0.35\textwidth}
\begin{tikzpicture}[scale = 0.55, transform shape]
\begin{axis}
[
 axis equal,
xlabel={Iteration},
ylabel={Weights},
y label style={anchor=north},
axis lines=middle,
xtick={0},
xticklabels={},
    ytick={0},
    ymin=-0.1,ymax=1.1,
    xmin=-0.1,xmax=1.1,
    axis on top=false,
every axis x label/.style={
    at={(ticklabel* cs:0.55)},
    anchor=north,
},
every axis y label/.style={
    at={(ticklabel* cs:1.01)},
    anchor=south,
},
]
\draw[dashed] (axis cs:0.0,1.0) -- (axis cs:1.0,1.0);
\addplot [black, nodes near coords=$1.0$,every node near coord/.style={anchor=0}] coordinates {( 0, 1)};
\draw[blue,line width = 0.5mm] (axis cs:0.0,0.0) -- (axis cs:0.3,0.0);
\draw[blue,line width = 0.5mm] (axis cs:0.3,0.0) -- (axis cs:0.36,1.0);
\draw[blue,line width = 0.5mm] (axis cs:0.36,1.0) -- (axis cs:0.63,1.0);
\draw[blue,line width = 0.5mm] (axis cs:0.63,1.0) -- (axis cs:0.7,0.0);
\draw[blue,line width = 0.5mm] (axis cs:0.7,0.0) -- (axis cs:1.0,0.0);

\draw[red,line width = 0.5mm] (axis cs:0.0,1.0) -- (axis cs:0.3,1.0);
\draw[red,line width = 0.5mm] (axis cs:0.3,1.0) -- (axis cs:0.36,0.0);
\draw[red,line width = 0.5mm] (axis cs:0.36,0.0) -- (axis cs:0.63,0.0);
\draw[red,line width = 0.5mm] (axis cs:0.63,0.0) -- (axis cs:0.7,1.0);
\draw[red,line width = 0.5mm] (axis cs:0.7,1.0) -- (axis cs:1.0,1.0);

\end{axis}
\end{tikzpicture}
\subcaption{Greedy strategy}\label{fig:Ex_ex_greedy_strategy}
\end{subfigure}%
\begin{subfigure}[t]{0.35\textwidth}
\begin{tikzpicture}[scale = 0.52, transform shape]
\begin{axis}
[
xlabel={Iteration},
ylabel={Weights},
y label style={anchor=north},
axis lines=middle,
xtick={0},
xticklabels={},
    ytick={0},
    ymin=-0.1,ymax=2.1,
    xmin=-0.1,xmax=1.1,
    axis on top=false,
every axis x label/.style={
    at={(ticklabel* cs:0.5)},
    anchor=north,
},
every axis y label/.style={
    at={(ticklabel* cs:1.01)},
    anchor=south,
},
legend style={at={(axis cs:1.0,0.7)},anchor=west}
]
\draw[dashed] (axis cs:0.0,2.0) -- (axis cs:1.0,2.0);
\addplot[domain=0:1.0,blue,line width = 0.5mm]{cos(deg(9*x-(pi/0.5)))+1};%
\addplot[domain=0:1.0,red,line width = 0.5mm]{sin(deg(9*x-pi/2))+1};
\addlegendentry{$w_{local}$}
\addlegendentry{$w_{global}$}
\addplot [black, nodes near coords=$1.0$,every node near coord/.style={anchor=0}] coordinates {( 0, 2.0)};
\end{axis}
\end{tikzpicture}
\subcaption{Switch strategy}\label{fig:Ex_ex_switch_strategy}
\end{subfigure}
\caption[Adaptive strategies to balance local exploitation and global exploration]{Adaptive strategies to balance local exploitation and global exploration. (a) Decreasing strategy, (b) Greedy strategy, (c) Switch strategy. }%
\label{fig:Ex_ex_strategy}%
\end{figure} 
Generally, four categories of adaptive sampling approaches can be identified in the literature. Here, it is assumed, that the initial experimental design was built and that a first metamodel $\tilde{\mathcal{M}}$ is available. 
\clearpage
\section{Cross-validation based adaptive sampling}
A generic, i.e. model-independent, approach that is used in the literature is the so-called Cross-validation based adaptive sampling. This group of methods either use the actual CV error or the \acrfull{cvv} to find the location of new sample points by estimating the actual prediction error.  CV has been thoroughly studied by multiple researchers including e.g. \cite{cressie1992statistics} or \cite{meckesheimer2002computationally}. \\
The basic idea of CV is to judge a constructed surrogate model by its performance with respect to unknown data. For the k-fold cross validation (see e.g. \cite{fushiki2011estimation}) for example
a dataset $\mathcal{D}$ and $K$ mutually exclusive and collectively exhaustive subsets of $\mathcal{D}$ with $\lbrace \mathcal{D}_{k}, \, k = 1, \, \ldots \, , K \rbrace$ may be given, then $\mathcal{D}$ can be split with
\begin{equation}
\begin{aligned}
\mathcal{D}_{i} \cap \mathcal{D}_{j} = \emptyset, \qquad \forall (i,j) \in [1 ; K]^{2} \qquad \text{and} \qquad \cup_{k=1}^{K} \mathcal{D}_{k} = \mathcal{D} \, \text{.}
\end{aligned}
\end{equation}
When fitting the model employing all except the k-th subset ($\mathcal{D} \setminus \mathcal{D}_{k}$) the k-th set of cross-validated prediction can be determined if a prediction is made on the specific subset that was left apart. The \acrfull{loocv} is a special case of the general form, whereby $K=m$. This approach is commonly used (see e.g. \cite{liu2016optimal}, \cite{martin2003study} or \cite{laurenceau2008building}) and will be investigated in this thesis. \\ Let the prediction from a surrogate model $\hat{\mathcal{M}}_{LOOCV}$ that was constructed from the reduced dataset $\mathcal{D}_{-1} = \mathcal{D} \setminus \left( \bm{x}^{(i)}, y_{i} \right)$ and evaluated at the point $\bm{x}^{(i)}$ be written as $\hat{Y}_{-i}$ in the following. The general LOOCV error at point $ \bm{x}^{(i)}$ then yields
\begin{equation}
\hat{e}(\bm{x}^{(i)}) = \hat{e}_{i} = \abs{  y^{(i)}(\bm{x}^{(i)})  - \hat{Y}_{-i} (\bm{x}^{(i)})}, \qquad 1\leq i \leq m \, \text{.}
\end{equation}
Small $\hat{e}(\bm{x}^{(i)})$ suggest that the effect of a loss of $\bm{x}^{(i)}$ to the surrogate model is not significant, which implies a well fitted metamodel around $\bm{x}^{(i)}$. On the contrary large values of $\hat{e}(\bm{x}^{(i)})$ give the recommendation that more sample points are needed in the surrounding domain of $\bm{x}^{(i)}$. This conclusion leads to the idea of an adaptive sampling process in areas with larger local LOOCV error. \\
Using the cross-validation errors $\bm{\hat{e}} = \lbrace \hat{e}_{1}, \, \ldots , \, \hat{e}_{m}\rbrace$ a generalized error, the so-called \acrfull{gmse} (see e.g, \cite{liu2016optimal}) can be obtained with
\begin{equation}
\text{GMSE} = \sqrt{\frac{1}{m} \bm{e}^{T} \bm{e}} .
\end{equation}
\newpage
\cite{liu2016optimal} mention that the GMSE generally overestimates the real mean-squared error of the metamodel, however this effect can be diminished by utilizing a sufficient number of observed points as e.g. stated in \cite{viana2009multiple}.\\
\cite{jin2002sequential} proposed a generalized version of the LOOCV defined as 
\begin{equation}
\hat{e}(\bm{x})  = \sqrt{ \frac{1}{m} \sum_{i=1}^{m} \left( y^{(i)}(\bm{x}) - \hat{Y}_{-i} (\bm{x}^{(i)}) \right)^{2} } \text{.}
\end{equation} 
This approach was also employed by \cite{kim2009construction} and extended to include a weighted version of the LOOCV by \cite{jiang2015novel}. \\
Generally, two problems can be observed when using CV for adaptive sampling. Firstly, as pointed out the LOOCV error is not a sufficient measure of the accuracy of the metamodel but in fact a measure of the sensitivity or insensitivity of the model to lost information \citep{cressie1992statistics}. Therefore an error estimation is needed for points which are not part of the current design, i.e. unobserved points. To account for this issue some researchers like \cite{aute2013cross} and \cite{li2010accumulative} propose to develop a metamodel $\mathcal{M}_{\hat{e}}$ for the error at unoberserved points.\\ Secondly, \cite{jin2001comparative} point out that a pure cross-validation based sampling approach can lead to clustered samples. Therefore this approach represents a pure focus on local exploitation.  \\
\cite{aute2013cross} and \cite{li2010accumulative} counter this problem by introducing a distance constraint involving a space-filling metric $S$ in order to only create new sample points within a limit euclidean distance to existing points and therefore to account for global exploration. The optimization problem to generate new sample points then reads
\begin{equation}
\begin{aligned}
&\bm{x}^{(m+1)} = \argmax\limits_{\bm{x}^{\star} \in \mathbb{X}}  \mathcal{M}_{\hat{e}}, \\
& \text{with} \, \, \norm{\bm{x}^{\star} - \bm{x}_{k}} \geq S, \, \forall \, \bm{x}_{k} \in \mathbb{X}.
\end{aligned}
\end{equation}
\cite{aute2013cross} and \cite{li2010accumulative} also suggest formulations to determine the space-filling criterion $S$. \\
\cite{aute2013cross} call their approach \acrfull{sfcvt} and utilize a maximin criterion. In a first step the minimum distance of each point in the design of experiments $\mathcal{X}$ is determined
\begin{equation}
ds(\bm{x}^{(i)}) = \min \left( \norm{\bm{x}^{(i)}- \bm{x}^{(j)}} \right), \qquad \forall \bm{x}^{(i)} \in \mathcal{X} \cap (i \neq j) \, \text{.} 
\end{equation}
In a second step the maximum of these minimum distances is computed and the space-filling criterion is set to be one half of the resulting value
\begin{equation}
S = 0.5 \, \max \left( ds(\bm{x}^{(i)}) \right), \qquad \forall \bm{x}^{(i)} \in \mathcal{X} \, \text{.}
\end{equation}
\cite{li2010accumulative} present a method called \acrfull{ace} for which they define an exponential \acrfull{doi} function 
\begin{equation}
DOI(\bm{x}^{(i)})) = \exp \left( - \alpha \norm{\bm{x}^{(i)} - \bm{x}^{(0)}} \right) \, \text{.}
\end{equation}
The DOI represents a measure for influence that the LOOCV
error has for $\bm{x}^{(i)}$ over $\bm{x}^{(0)}$. The factor $\alpha$ adjusts the rate of decreasing influence. The authors offer a summary of the effect of $\alpha$ on the sampling algorithm and provide a coherent formulation of determining its value. The new sampling point can be generated from the constrained optimization 
\begin{equation}
\begin{aligned}
&\bm{x}^{(m+1)} = \argmax\limits_{\bm{x}^{\star} \in \mathbb{X}} \sum_{i=1}^{n} \hat{e}(\bm{x}^{(i)}) \left(  \exp (- \alpha \norm{\bm{x}^{(i)} - \bm{x}^{\star}})\right), \\
& \text{with} \, \, \norm{\bm{x}^{(i)} - \bm{x}^{\star}} \geq S.
\end{aligned}
\end{equation}
The cluster threshold $S$ is then obtained in a two-step process. Firstly, the minimum distance among all sample points in the DOE is calculated. Secondly the average value of these minimum distances is determined. The cluster threshold is then one-half of this average value. \\
\cite{jiang2015novel} follow a similar approach to account for the global exploration and set the value of $S$ according to a multi-step procedure involving averaging the minimum distance to surrounding points of the design. \\
Other researchers split up the design space and select the partition with the highest CV error and therefore are able to account for global exploration in this way. An early adaption of this technique was given by \cite{devabhaktuni2000neural}, who utilized $2^{n}$ regions for this approach. A similar method known as \acrfull{cvvor} was presented by \cite{xu2014robust} who employs the Voronoi diagram algorithm as described by \cite{aurenhammer1991voronoi} to subdivide the design space into Voronoi cells. The cell with the largest CV error is selected as the one to generate a new sample point in. This approach manages to find a sensible balance between local exploitation and local clustering of sample points. The general scheme of the method is illustrated in Box \ref{alg::CVVOR}. 
\begin{kasten}[h!]
\begin{Algorithmus}
\begin{itemize}
\itemsep1em 
\item Given a design of experiments $\mathcal{X} = \lbrace \bm{x}^{1}, \ldots , \bm{x}^{m} \rbrace$
\item Given a set of observations $\mathcal{D} = \lbrace \left( \bm{x}^{(i)}, \,\bm{y}^{(i)} \right), \, i=1, \, \ldots  , \, m  \rbrace$ .
\item Construct metamodel $\hat{\mathcal{M}}$ with $\mathcal{D}$.
\item[] While stopping criterion is not reached. \\
Do:

\begin{itemize}[topsep=-5px,partopsep=0px]
\item[] Partition the current DOE into a set of Voronoi cells $C = \lbrace C_{1} , \ldots, C_{m}\rbrace$
\item[] Find the sensitive Voronoi cell $C_{sensitive}$ by comparing the prediction error of each cell.
\item[] Determine the new sample point $\bm{x}^{(m+1)}$ by utilizing the point in the cell farthest away from the corresponding cell sample point $\bm{x}_{sensitive}$.
\end{itemize}
end
\end{itemize}
\end{Algorithmus}
\captionof{kasten}{Algorithm for CVVOR}\label{alg::CVVOR}
\end{kasten}

A different approach called the
the \acrfull{ssa} was introduced by \cite{garud2017smart}. Here the authors use the cross-validation technique to create a set of optimization problems to generate a new sample point. \\ 
In order to account for exploration the authors choose the \acrfull{cdm}
\begin{equation}\label{eq::CMD}
CDM(\bm{x}) = \sum_{i=1}^{m} \left( \norm{\bm{x} - \bm{x}^{(i)}}  \right)^{2},
\end{equation}
as presented by \cite{zhang2012adaptive} in order to account for global exploration. A higher metric value indicates a local isolation of a sample point. \\The exploitation is assured by a departure function $\Delta$. It provides a measure for the impact of generating a new sample point close to an existing one. For all points of the observations it is given as 
\begin{equation}
\Delta_{j}(\bm{x}) = \tilde{\mathcal{M}}(\bm{x}) - \tilde{\mathcal{M}}_{j}(\bm{x}), \qquad j=1, \ldots, m \, \text{.}
\end{equation}
Here $\tilde{\mathcal{M}}^{m}_{j}$ is the surrogate model constructed from all points of the DOE except for $\bm{x}^{(j)}$. \\
The crowding metric and the departure function are then combined in order to obtain the new sample point which should maximize both measures. This is achieved by multiplication. For each point of the DOE an optimization problem can be established with
\begin{equation}\label{eq::NLP}
\bm{x}^{(m+1)} = \argmax\limits_{\bm{x}^{\star} \in \mathbb{X}} \left(  \Delta_{j}(\bm{x}^{\star} ) \right)^{2} \, CMD(\bm{x}^{\star} ),  \qquad j=1, \ldots, m \, \text{.}
\end{equation}
The new sample point is the one with the optimal solution of this equation. The algorithm is summarized in Box \ref{alg::SSA}.
\begin{kasten}[h!]
\begin{Algorithmus}
\begin{itemize}
\itemsep1em 
\item Given a set of observations $\mathcal{D} = \lbrace \left( \bm{x}^{(i)}, \,\bm{y}^{(i)} \right), \, i=1, \, \ldots  , \, m  \rbrace$ .
\item Construct metamodel $\hat{\mathcal{M}}$ with $\mathcal{D}$.
\item[] While stopping criterion not reached.\\
Do:
\begin{itemize}[topsep=-5px,partopsep=0px]
\item[] Compute $CMD_{j} = CMD(\bm{x}^{j})$ using equation (\ref{eq::CMD}).
\item[] Arrange $CMD_{j}$ in descending order. Define an index for the order $p= 1, \ldots, m$. Set $p=1$.
\item[] Define subset of observations with $\mathcal{D}_{p} = \lbrace \left( \bm{x}^{(p)}, \,\bm{y}^{(p)} \right)  \rbrace$ .
\item[] Construct metamodel $\hat{\mathcal{M}}_{p}$ with $\mathcal{D}_{-1} = \mathcal{D} \setminus \mathcal{D}_{p}$.
\item[] Solve equation (\ref{eq::NLP}) and gain optimal solution $\bm{x}^{\star}$.
\item[] If $\exists \, i=1,\ldots,m$ such that $\norm{\bm{x}^{\star} - \bm{x}_{(i)}} < \epsilon$, set $p=p+1$. Else set the new sample point
$\bm{x}^{(m+1)} = \bm{x}^{\star}$.
\end{itemize}
end
\end{itemize}
\end{Algorithmus}
\captionof{kasten}{Algorithm for SSA}\label{alg::SSA}
\end{kasten}
\newpage
Generally new sample points for CV based adaptive sampling can be obtained by finding the solution to the optimization problem of the form
\begin{equation}
\bm{x}^{(m+1)} = \argmax\limits_{\bm{x}^{\star} \in \mathbb{X}}   \text{Score} \left(  \hat{e} (\bm{x}^{\star}), d (\bm{x}^{\star}) \right) \, \text{.}
\end{equation}
Here local exploitation is achieved by evaluating some variant of a cross-validation error, and global exploration is controlled by a distance constraint or a partitioned input space. \\
\cite{dubrule1983cross} presented a way to calculate the leave-one-out predictions specifically for universal Kriging. The derivation starts by introducing the matrix $\bm{S}$ as
\begin{equation}
\begin{aligned}
\bm{S} = \begin{bmatrix}
\sigma^{2} \bm{R} & \bm{F} \\
\bm{F}^{T} & \bm{0}
\end{bmatrix} \, \text{.}
\end{aligned}
\end{equation}
Let then the matrix $\bm{B}$ be defined by
\begin{equation}
\begin{aligned}
B_{i\, j} = S_{i \, j}, \qquad i, \, j = 1, \, \ldots \, , m \, \text{.}
\end{aligned}
\end{equation}
\cite{dubrule1983cross} concluded that the mean and the variance of the $m$ leave-one-out predictions come out to 
\begin{equation}
\begin{aligned}
\mu_{\hat{Y}_{-i}} = - \sum_{\substack{j=1 \\ i\neq j}}^{m} \frac{B_{i \, j}}{B_{i \, i}} y^{(j)}, \qquad i = 1, \, \ldots \, , m
\end{aligned}
\end{equation}
and 
\begin{equation}
\begin{aligned}
\sigma_{\hat{Y}_{-i}}^{2} = \frac{1}{B_{i \, i}}, \qquad i=1, \, \ldots \, , m \, \text{.}
\end{aligned}
\end{equation}
Then, the GMSE for Kriging reads
 \begin{equation}
GMSE = \sqrt{\frac{1}{m} \sum_{i=1}^{m} (\mu_{\hat{Y}_{-i}} - y^{(i)})^{2}} .
\end{equation}
Instead of the GMSE the Kriging-related literature introduces the \acrfull{gcv} score as introduced by \cite{golub1979generalized} to measure the cross-validation. The definition however varies in the literature, a good overview is given in \cite{gu2001cross}. In this thesis the GCV score is written as 
\begin{equation}
\begin{aligned}
Q^{2} = 1 - \frac{1}{m} \sum_{i= 1}^{m} \left( \frac{\mu_{\hat{Y}_{-i}} - y^{(i)}}{\sigma_{\hat{Y}_{-i}}}  \right)^{2} \, \text{.}
\end{aligned}
\end{equation}
\section{Variance-based adaptive sampling}
The basic concept of variance-based adaptive sampling relies on the idea that large prediction errors of the surrogate model $\tilde{\mathcal{M}}$ occur in domains with large prediction variances. Since the prediction variance is a byproduct of the Kriging metamodel generation, this strategy evolves natural out of the Kriging procedure. Variance-based adaptive sampling methods rely on the inner properties of the metamodel, i.e. exhibit stochastic properties, and are then model-dependent and hard to extend to other surrogate modeling techniques that do not offer simple ways to obtain their prediction variance.  \\
Intuitively, one can use the \acrfull{mse} (c.f. eq. (\ref{eq::MSE_variance})) to sample a new point. \cite{jin2002sequential} employ this approach, which writes
\begin{equation}
\bm{x}^{(m+1)} = \argmin\limits_{\bm{x}^{\star} \in \mathbb{X}} \left( \argmax\limits_{\bm{x}^{\star} \in \mathbb{X}} \sigma_{\hat{Y}} (\bm{x}^{\star}) \right) \, \text{.}
\end{equation}
Here the new sample is selected at the point which minimizes the largest MSE in the existing Kriging model. The approach is commonly referred to as \acrfull{mmse} \citep{sacks1989design}. In fact, this strategy is a special representation of the entropy approach presented by \cite{shannon1948mathematical} and further developed by \cite{currin1988bayesian} and later \cite{currin1991bayesian} for the case that only one new point is selected at each iteration. 
Another strategy developed by \cite{sacks1989design} called \acrfull{imse} considers a weighted averaged MSE value over the entire input space of the form
\begin{equation}
\bm{x}^{(m+1)} = \argmin\limits_{\bm{x}^{\star} \in \mathbb{X}} \int_{\mathbb{X}} \sigma_{\hat{Y}} (\bm{x}^{\star}) w(\bm{x}^{\star}) d\bm{x}^{\star} \, \text{,}
\end{equation} 
where $w$ is a probability density function which means a unit integral over $\mathbb{X}$. It selects a new point that minimizes the averaged best linear unbiased predictor over the entire domain after adding it in the sample set.
The entropy approach, the MMSE approach and the IMSE approach are in this form not suitable for other metamodels than Kriging. \\
However the representation of the prediction error solely with the prediction variance is inaccurate as can be seen when reintroducing, e.g. the
bias-variance decomposition explained by \cite{geman1992neural} in the field of neuronal networks
\begin{equation}\label{eq::BIASVariancedecomposition}
\begin{aligned}
\underbrace{E \lbrace [ \hat{Y}(\bm{x}) - y(\bm{x}) ]^{2} \rbrace}_{\text{estimation error}} = \underbrace{[ E \lbrace  \hat{Y}(\bm{x}) - y(\bm{x}) \rbrace ]^{2}}_{\text{bias}}  + \underbrace{ E \lbrace [  \hat{Y}(\bm{x}) - E \left( y(\bm{x}) \right) ]^{2}  \rbrace}_{\text{variance}} \, \text{.}
\end{aligned}
\end{equation}
This equation highlights the potential of adjusting the prediction variance to reduce the estimation error. The left-hand side of the equation represents the prediction error. The first term of the right-hand side is the difference between the prediction response $ \hat{Y}$ and the actual response $y$, is the bias \citep{geman1992neural}. The second term is the prediction variance $\sigma_{\hat{Y}}^{2}$ of the surrogate model. Therefore when reducing the estimation error both the bias and the variance need to be decreased. Since the actual response is unknown the bias in itself is non-adjustable. However information about the estimation of the bias can be used to adjust the prediction variance $\sigma_{adj}^{2}$. \\
New points with this strategy can be written in the general form of exploration and exploitation as
\begin{equation}
\bm{x}^{m+1} = \argmin\limits_{\bm{x}^{\star} \in \mathbb{X}} \text{Score} \left( \sigma_{adj}(\bm{x}^{\star}) , \sigma_{\hat{Y}}(\bm{x}^{\star}) \right) \, \text{.}
\end{equation}
The local exploitation is achieved with the prediction variance, whereas the variance of the surrogate model provides the global exploration of the adaptive sampling process. 
In general, there are two ways to make an adjustment on the prediction error to account for the bias, namely internal and external variance-based approaches.
\subsection{Internal variance-based adaptive sampling}
The first adjustment strategy is to incorporate the bias information directly into the prediction variance. \\
As established for stationary Gaussian processes the correlation function only depends on the euclidean distance between two points.
\cite{lin2004sequential} used this to establish an adjusted correlation function of the general form 
\begin{equation}
\hat{\sigma} (\bm{x}^{(i)}, \bm{x}^{(j)}) = \sigma^{2} R_{\text{adj}} (\bm{x}^{(i)}, \bm{x}^{(j)}, \bm{\theta}, \eta_{i}, \eta_{j}) \, \text{.}
\end{equation}
Here the adaption of the prediction variance is achieved by introducing two adjustment factors $\eta_{i}$ and $\eta_{j}$ that try to introduce the prediction bias at the sample points  $\bm{x}_{i}$ and $\bm{x}_{j}$. Since the correlation matrix affects the prediction variance, this method presents a way to include the bias information. However generally the values of the two factors are unknown. Using the cross-validation error to make estimations, \cite{liu2016adaptive} could establish values for these factors. The authors called their approach 
Adaptive maximum entropy (AME). In order to avoid local clustering the adjustments factors are set to be
\begin{equation}
\eta_{i}(\bm{x}^{(i)}) = \left( \frac{\hat{e}_{i}(\bm{x}^{(i)})}{\hat{e}_{max}} \right)^{\gamma} \, ,
\end{equation}
where $\hat{e}_{i}(\bm{x}^{(i)})$ is the LOOCV error at the sample point $\bm{x}^{(i)}$.  $\hat{e}_{max}$ is the maximum LOOCV error.
Given the auxiliary notation 
\begin{equation}
\bm{r}_{0}^{\star} = R_{adj}(\bm{x}^{\star} - \bm{x}^{(i)}, \bm{\theta},  \eta_{i}), \qquad i=1, \ldots, m,
\end{equation}
with the adjusted correlation function, $R_{adj}(\bullet)$ the matrix
\begin{equation}
\bm{R}^{\star} = \begin{bmatrix}
\bm{R}_{adj} & \bm{r}_{0}^{\star} \\
\bm{r}_{0}^{\star} & 1
\end{bmatrix}
\end{equation} 
can be defined with the goal to find a new sample point by maximizing the determinant of the correlation matrix. Therefore the optimization problem is of the form
\begin{equation}
\argmax\limits_{\bm{x}^{\star} \in \mathbb{X}} = \det \left( \sigma^{2} \bm{R}^{\star}(\bm{x}^{\star}) \right) \, \text{.}
\end{equation}
The sampling procedure for AME is summarized in Box \ref{alg::AME}.
\begin{kasten}
\begin{Algorithmus}
\begin{itemize}
\itemsep1em 
\item Given a design of experiments $\mathcal{X} = \lbrace \bm{x}^{(1)}, \ldots , \bm{x}^{(m)} \rbrace$ 
\item Choose a trade-off search pattern $\bm{\gamma} = \lbrace \gamma_{1}, \ldots, \gamma_{N} \rbrace$ and set $\Theta=1$ 
\item[] While the adaptive sampling stopping criterion is not satisfied. \\
Do:
\begin{itemize}[topsep=-5px,partopsep=0px]
\item[] Construct metamodel  $\hat{\mathcal{M}}$ with regular covariance matrix.
\item[] Determine the trade-off coefficient. Set $\gamma = \bm{\gamma}(\Theta)$. If $\Theta > N$ set $\Theta=1$.
\item[] Adjust covariance matrix entries.
\item[] Identify new sampling point with optimization scheme. 
\end{itemize}
end
\end{itemize}
\end{Algorithmus}
\captionof{kasten}{Algorithm for AME}\label{alg::AME}
\end{kasten}
\subsection{External variance-based adaptive sampling}
This strategy adds an external bias term to the prediction variance to adjust it. 
\cite{jones1998efficient} proposed the so-called \acrfull{ei} criterion. 
For this the authors introduce what they call improvement as a random variable of the form
\begin{equation}
\begin{aligned}
I (\bm{x}) = \begin{cases}
y_{min} - \hat{Y}(\bm{x}), &\text{if} \, \hat{Y}(\bm{x}) \leq y_{min} \\
0, &\text{otherwise}
\end{cases}
= \max \lbrace y_{min} - \hat{Y}(\bm{x}), 0 \rbrace \, \text{.}
\end{aligned}
\end{equation}
Here $y_{min} = \lbrace y^{(i)}, i=1, \ldots \, , m  \rbrace$ defines the minimum value of $\mathcal{M}$ in the dataset $\mathcal{D}$. \\
\cite{jones1998efficient} describe this variable with its expectation, yielding the \textit{expected improvement}
\begin{equation}
\begin{aligned}
EI (\bm{x}) \equiv \mathbb{E} [I(\bm{x})] = \int_{-\inf}^{y_{min}} (y_{min} - \hat{y}) \varphi \left( \frac{\hat{y} - \mu_{\hat{y}}(\bm{x})}{\sigma_{\hat{Y}}(\bm{x})} \right) d \hat{y}
\end{aligned}
\end{equation}
which turns into \citep{bichon2010efficient}
\begin{equation}\label{eq::EI}
\begin{aligned}
EI(\bm{x}) = \left( y_{min} - \mu_{\hat{Y}}(\bm{x}) \right) \Phi \left( \frac{y_{min} - \mu_{\hat{Y}}(\bm{x})}{\sigma_{\hat{Y}}(\bm{x})} \right) + \sigma_{\hat{Y}}(\bm{x}) \varphi \left( \frac{y_{min} - \mu_{\hat{Y}}(\bm{x})}{\sigma_{\hat{Y}}(\bm{x})} \right)
\end{aligned}
\end{equation}
where $\varphi$ and $\Phi$ are the standard Gaussian PDF and CDF respectively. \\
The next point that should be added is then defined by
\begin{equation}
\begin{aligned}
\bm{x}^{(m+1)} = \argmax\limits_{\bm{x}^{\star} \in \mathbb{X}} \qquad EI(\bm{x}^{\star}) \, \text{.}
\end{aligned}
\end{equation}
The EI function may be interpreted as a fixed balance between global exploration and local exploitation. The first term of eq. (\ref{eq::EI}) favors exploitation, whereas the second term searches for new regions of interest and is therefore the part responsible for exploration. Various variants and improvements can be found in the literature. \\
\cite{sobester2005design} presented the \acrfull{wei}, which introduces a tuneable parameter $0 \leq w \leq 1$ to the expected improvement in the form of
\begin{equation}
\begin{aligned}
WEI(\bm{x}) =& w \left( y_{min} - \mu_{\hat{Y}}(\bm{x}) \right) \Phi \left( \frac{y_{min} - \mu_{\hat{Y}}(\bm{x})}{\sigma_{\hat{Y}}(\bm{x})} \right) \\ &+ (1-w) \sigma_{\hat{Y}}(\bm{x}) \varphi \left( \frac{y_{min} - \mu_{\hat{Y}}(\bm{x})}{\sigma_{\hat{Y}}(\bm{x})} \right)
\end{aligned}
\end{equation}
in order to be able to adjust the weights on global exploration and local exploitation. Evident values for $w$ are $w=0$ which yields full exploration, $w=1$ which focuses on exploitation. For $w=0.5$ the algorithm becomes equivalent to EI. \\
Another approach is given by \cite{xiao2012exploration} with the \acrfull{awei}, which is built up on WEI. The issue with the latter are the model-dependent weights. AWEI tunes the parameter $w$ in
response to the environment feedback, e.g. initial tests.  \\
\cite{lam2008sequential} introduced the \acrfull{eigf}. It selects new points in regions with significant variation of the response values. It can be written as
\begin{equation}
EIGF(\bm{x}) = \left( \mu_{\hat{Y}}(\bm{x}) - y(\bm{x}^{\star}) \right)^{2} +\sigma_{\hat{Y}}^{2}(\bm{x}),
\end{equation}
where $\bm{x}_{\star}$ is the sample point that is closest in distance to the candidate $\bm{x}$. Hence, $y(\bm{x}_{\star})$ is the response at that point. The first term in the right-hand side represents the role of exploitation. It is large when the estimation of the response $\hat{Y}(\bm{x})$ tends to be different than the response at the nearest point. The global exploration, given by the second term on the right-hand side, is larger when the model has more uncertainty inherent. A different procedure is employed by \citep{busby2009hierarchical} who uses a discontinuous adaptive griding scheme to partition the input space and to account for exploration. The method is termed \acrfull{haed} by the author. The idea is to split a normalized input space in $\mathbb{R}^{n}$ into smaller hypercuboidal subdomains of equal size, where the edge length of such cells in the dimension $i$ corresponds to the correlation length $\theta_{i}$ of the Kriging technique. After that the prediction error is estimated in each cell by calculating the root mean-squared error of all sample points in a given cell. If a cell contains no points the corresponding value is set to $+ \infty$. Therefore when the root mean-squared error is small the metamodel accurately approximates the mapping. In cells with higher local prediction error new sample points need to be added. This can be done until a stopping criterion is reached.  \\
A recently introduced external variance-based technique involving the switch strategy was presented by \cite{liu2017adaptive}. 
The authors employ the decomposition of equation (\ref{eq::BIASVariancedecomposition}) and term the technique \acrfull{mepe}. Here the true bias term $e_{true}^{2}$ incorporating the first term of the right-hand side of equation (\ref{eq::BIASVariancedecomposition}) is estimated by a continuously approximated LOOCV error $\hat{e}(\bm{x})$ the variance is given by the natural Kriging prediction variance. The authors introduce a balance factor $\alpha$ to adjust the exploitative bias term and the exploratory variance term in order to adaptively switch between the two components depending on the estimation quality of the bias term. 
The continuous \acrfull{epe} to be maximized is given by
\begin{equation}\label{eq::MEPE_eq}
EPE(\bm{x}) = \alpha  \hat{e}(\bm{x}) + (1-\alpha) \hat{\sigma}_{\hat{Y}}^{2},
\end{equation}
where $\hat{e}(\bm{x})$ is the value of $\hat{e}(\bm{x}^{(i)})$ when $\bm{x}$ is located in the Voronoi cell of point $\bm{x}^{(i)}$. Otherwise its values are zero. In order to fasten the computation \cite{liu2017adaptive} utilize an approximation of $\hat{e}(\bm{x}^{(i)})$ as given in \cite{sundararajan2000predictive}. \\
The computation of the balance factor for the calculation of the sample point $\bm{x}^{(m+q-1)}$ reads
\begin{equation}\label{eq::alphaFactor}
\begin{aligned}
\alpha = \begin{cases} 0.5, & \text{if} \, q=1 \\
0.99 \min \left[ 0.5 \dfrac{ e_{true}^{2}(\bm{x}^{(m+q-1)}) }{\hat{e}(\bm{x}^{(m+q-1)})}, 1 \right] & \text{if} \, q>1.
\end{cases}
\end{aligned}
\end{equation} 
The algorithm is presented in Box \ref{alg::MEPE}.
\begin{kasten}[h!]
\begin{Algorithmus}
\begin{itemize}
\itemsep1em 
\item Given a design of experiments $\mathcal{X} = \lbrace \bm{x}^{(1)}, \ldots , \bm{x}^{(m)} \rbrace$. 
\item[] While the adaptive sampling stopping criterion is not satisfied. \\
Do: 
\begin{itemize}[topsep=-5px,partopsep=0px]
\item[] Update the balance factor $\alpha$ with equation (\ref{eq::alphaFactor}).
\item[] Obtain the LOOCV error at each sample point $\hat{e}(\bm{x}^{(i)})$.
\item[] Obtain a new point by maximizing the $EPE$-criterion of equation (\ref{eq::MEPE_eq}) over the input domain.
\item[] Update the design of experiments with the new found point and its response.
\end{itemize}
end
\end{itemize}
\end{Algorithmus}
\captionof{kasten}{Algorithm for MEPE}\label{alg::MEPE}
\end{kasten}
\section{Gradient-based adaptive sampling}
The basic idea gradient-based adaptive sampling strategies is the assumption that it is difficult to construct a sufficient metamodel with low prediction errors in regions with large gradients.
Therefore information about the gradients is helpful for the generation of new sample points. 
For some surrogate modeling techniques like RBF, the gradient information can be easily and cheaply derived, however for other models this is not the case. Generally this strategy faces the substantial restriction that gradients need to be obtainable.\\ 
Methods to use gradient information in the Kriging model are under investigation, see e.g. \cite{bouhlel2019gradient} or \cite{ulaganathan2016high}. \newpage The keyword here is GEK \citep{liu2003development} which in some instances is also referred to as Co-Kriging (see e.g. \cite{laurent2013generation}). 
A good overview of general gradient-enhanced metamodels which also includes GEK is given in \cite{laurent2017overview}. However the use of adaptive sampling in combination with gradient-enhanced Kriging models has not been extensively researched.  \\
\cite{rumpfkeil2011dynamic} introduced a gradient and Hessian enhanced Kriging model with a dynamic sampling approach. Under the assumption that the gradient information at the observed points is available the authors use this information to build local metamodels in domains around specific test candidate points by using the so-called Dutch Intrapolation method (see e.g. \cite{kraaijpoel2003seismic}). After that a comparison is employed between the global Kriging surrogate model and
the local Dutch Intrapolation model to find the point in the parametric domain with largest discrepancy between the two models. This point is then added to the DoE. In order to achieve global exploration the authors try to avoid clustering of sample points with the help of a distance criterion. \\
\cite{paul2013sensitivity} employ an adaptive sampling strategy where the sensitivity of the model, which is derived from the GEK gradient information, is coupled with the mean-squared error to present a new error estimation method. This technique is utilized with a distance criterion to sample new points. The results are promising, especially in comparison to CV-based sampling approaches for higher dimensional cases. \\
Gradient-based adaptive sampling techniques have been used more extensively in other types of metamodels.   \cite{yao2009gradient} employ a radial basis function neural network (RBFN) metamodel and use the gradient information of this model to sample in regions of interest. The global exploration was achieved by using OLHD (see e.g. \cite{giunta2003}). Instead of using the gradient information some researchers (see e.g. \cite{mackman2011} or \cite{WEI20129635}) utilize the curvature as a way to sample new points. Gradient-based adaptive sampling has also been used by \cite{pickett2011} in the context of NURBS-based surrogate models.\\
A model-independent technique was proposed by \cite{crombecq2011novel} and denoted LOLA-Voronoi by the authors. It has been revisited by \cite{Deschrijver2011AdaptiveSA} and has been the focus of other research (see e.g. \cite{van2015fuzzy}). Furthermore the method has been implemented in the MATLAB surrogate toolbox (SUMO) \citep{gorissen2010surrogate}. \\
LOLA-Voronoi as presented by \cite{crombecq2011novel} performs a Voronoi Tessellation process to describe the density of existing sample points by calculating the volume of each Voronoi cell. This is done in order to have an accurate measure $V(\bullet)$ for the exploration of the model at hand. 
However, since Voronoi tessellation as for example defined in \cite{aurenhammer1991voronoi} is non-trivial \cite{crombecq2011novel} approximate the cells and the respective volume with a Monte Carlo approach.
In order to account for exploitation the gradient of the mapping $\mathcal{M}$ is linearly approximated with the \acrfull{lola}. Hence, the region's of the input space can be ranked according to their approximated nonlinearity with the help of the nonlinearity measure $E(\bullet)$. 
A new sample point is then obtained by means of optimization of a combination of these two measures
\begin{equation}
\bm{x}^{(m+1)} = \argmax\limits_{\bm{x}^{\star} \in \mathbb{X}}  \, E(\bm{x}^{\star}) + V(\bm{x}^{\star}) \, \text{.}
\end{equation} 
Generally the optimization problem for generating new sample point in gradient-based adaptive sampling reads
\begin{equation}
\bm{x}^{(m+1)} = \argmax\limits_{\bm{x}^{\star} \in \mathbb{X}}   \text{Score} \left(  \hat{g} (\bm{x}^{\star}), d (\bm{x}^{\star}) \right) \, \text{.}
\end{equation}
Local exploitation is performed by using estimated first or second order gradients $\hat{g}$ dependent on the chosen surrogate model. Global exploration is achieved by using a distance metric $d$ to avoid clusters of sample points. 
\section{Query-by-committee based adaptive sampling}
The \acrfull{qbc} approach was presented by \cite{seung1992query} and \cite{freund1993information} as a method for filtering informative queries from a random stream of inputs. The basic idea is to employ a committee of different surrogate models, where each of the models is used to predict the response value at a candidate point $\bm{x}$. The point in the domain with the most "differences" between the committee is chosen as the new sample point. The ambiguous term "differences" is defined by \cite{krogh1995neural} as the variance of the prediction of the committee members.\\ Let a committee $C$ consist of $n_{C}$ members $C = \lbrace \hat{Y}_{i} \rbrace$ with $i=1, \, \ldots, \, n_{C}$, the predicted variance $\hat{\sigma}^{2}_{\text{QBC}}$ is then given as 
\begin{equation}
\hat{\sigma}^{2}_{\text{QBC}} (\bm{x}) = \frac{1}{n_{C}} \sum_{i=1}^{n_{C}} \left(  \hat{Y}_{i} (\bm{x}) - \overline{\hat{Y}} (\bm{x})  \right)^{2}
\end{equation}
where $\overline{\hat{Y}}  = \sum_{i=1}^{t} \hat{Y}_{i} (\bm{x})$ is the average value of the committee evaluated at the candidate point. For Kriging the different surrogate models can for example be built by constructing models employing different autocorrelation functions (e.g. different Matérn and power exponential functions). In comparison to the variance-based adaptive sampling approaches the QBC technique is more generic since it is model-independent. Different types of metamodels can define a committee. Therefore a new sample point for a construction of a Kriging surrogate model can be for example found by different forms of RBF and SVR. Furthermore this strategy supports the use of ensemble metamodels (see e.g. \cite{mendes2012ensemble} or \cite{acar2009ensemble}). \\
\cite{mendes2012ensemble} explain the reasoning why QBC is able to reduce the estimation error of the surrogate models involved in the committee. However,
\cite{melville2004diverse} argue that there must be diversity in the committee members in order for the QBC approach to reduce the generalization error. This is the reason why two general strategies can be distinguished: homogeneous and heterogeneous QBC.
In homogeneous QBC the members of the committee are all part of the same surrogate model type. An example is given in \cite{kleijnen2004application}, where the the idea is to create different committee members by using a statistical technique called jackknifing. Here competing metamodels are generated by leaving out subsets of the DOE. This approach is utilized in the \acrfull{masa} introduced by \cite{eason2014adaptive} developed for neural networks. 
Here in a first step the committee members are constructed. Thereafter a large set of points is randomly generated in the design space. For each of these points a quality parameter based on the variance of the respective point as well as the euclidean distance to existing sample points is calculated. The new sample point is chosen to be the best performing point of the set. The algorithm is summarized in Box \ref{alg::MASA}.  
\begin{kasten}
\begin{Algorithmus}
\begin{itemize}
\itemsep1em 
\item Given a design of experiments $\mathcal{X} = \lbrace \bm{x}^{(1)}, \ldots , \bm{x}^{(m)} \rbrace$
\item Choose number of committee members $K$ and select $K$ subsets $\mathcal{X}_{i} \, \, (i=1, \ldots K)$ of $\mathcal{X}$
\item[] for $i=1:K$
\begin{itemize}[topsep=-5px,partopsep=0px]
\item[] Construct metamodel $\hat{\mathcal{M}}_{i}$ by approximating $\mathcal{M}(\mathcal{X}_{i})$
\end{itemize}
end
\itemsep0em 
\item[] Create randomly generated set of points $P$
\item[] for all $\bm{p}$ in $P$
\begin{itemize}[topsep=-5px,partopsep=0px]
\item[] for $i=1:K$
\begin{itemize}[topsep=-5px,partopsep=0px]
\item[] Predict the metamodel response at point $\bm{p}$. \\
$\hat{Y}_{i} = \hat{\mathcal{M}}_{i}(\bm{p})$
\end{itemize}
end
\item[] Predict the variance at point $\bm{p}$ with $\hat{\sigma}^{2}_{\text{QBC}} (\bm{p})$. Save the maximum value $\hat{\sigma}^{2}_{max, \text{QBC}}$.
\item[] Calculate the euclidean distance $d_{ep}$ between $\bm{x}^{(1)}$ and $\bm{p}$. Save the maximum value $d_{max,ep}$.
\item[] for $j=1:m$
\begin{itemize}[topsep=-5px,partopsep=0px]
\item[] Set $Y_{d} = $ euclidean distance between $\bm{x}^{(j)}$ and $\bm{p}$ 
\item[] if $Y_{d} < d_{ep}$
\begin{itemize}[topsep=-5px,partopsep=0px]
\item[] Set $d_{ep} = Y_{d}$ 
\end{itemize}
end
\end{itemize}
end
\end{itemize}
end
\item[] for all $\bm{p}$ in $P$
\begin{itemize}[topsep=-5px,partopsep=0px]
\item[] Set $\eta_{p} = \frac{d_{ep}}{d_{max,ep}} + \frac{\hat{\sigma}^{2}_{\text{QBC}} (\bm{p})}{\hat{\sigma}^{2}_{max, \text{QBC}}(\bm{p})} $
\end{itemize}
end
\item[] Set the new sampling point as the point $\bm{p}$ with the highest value of $\eta_{p}$
\end{itemize}
\end{Algorithmus}
\captionof{kasten}{Algorithm for MASA}\label{alg::MASA}
\end{kasten}
In contrast heterogeneous means in this context that the committee members are constructed using different metamodeling approaches.
\newpage
For instance, this approach is used by \cite{douak2012active}, where regressors of partial least squares regression, SVM, ridge regression and kernel ridge regression build a committee. According to the authors the results showed good performance in comparison to other adaptive sampling techniques.  \\
The new sample points in QBC adaptive sampling are generally found by
\begin{equation}
\bm{x}^{(m+1)} = \argmax\limits_{\bm{x}^{\star} \in \mathbb{X}}   \text{Score} \left(  \hat{\sigma}_{\text{QBC}} (\bm{x}^{\star}), d (\bm{x}^{\star}) \right) \, \text{.}
\end{equation}
Local exploitation is generated by the QBC-variance, whereas global exploration is achieved by a distance function $d(\bullet)$ that presents local clustering of sample points as e.g. seen in CV-based adaptive sampling approaches. 
\section{Investigated adaptive sampling techniques}
The adaptive sampling techniques investigated in thesis are listed here. To the best of the authors knowledge the presented methods are the most commonly applied techniques throughout the literature. Furthermore it was tried to utilize methods with unique features. The following methods are compared
\begin{itemize}
\item \acrfull{ace} 
\citep{li2010accumulative} 
\item \acrfull{ame}  \citep{liu2016adaptive} 
\item \acrfull{cvd}  \citep{jin2002sequential} 
\item \acrfull{cvvor} \citep{xu2014robust} 
\item \acrfull{ei}   \citep{jones1998efficient} 
\item \acrfull{eigf} \citep{lam2008sequential} 
\item LOLA-Voronoi \citep{crombecq2011novel} 
\item \acrfull{masa}
 \citep{eason2014adaptive} 
\item \acrfull{mepe} \citep{liu2017adaptive}
\item \acrfull{mipt}  \citep{crombecq2011efficient} 
\item \acrfull{msd} \citep{jin2002sequential} 
\item \acrfull{sfcvt} \citep{aute2013cross} 
\item \acrfull{ssa} \citep{garud2017smart} 
\end{itemize}
The general properties and an overview over the features of these methods is given in Table \ref{table::OverViewMethods}.
\begin{table}[h!]
\begin{center}
\resizebox{1.0\textwidth}{!}{%
\begin{tabularx}{1.3\textwidth}{l l l l}
\specialcell{Sampling\\method} & Exploitation & Exploration & Special features \\ \hline\hline \\
ACE &  \specialcell{$\hat{e}^{\text{LOOCV}}$} &  Distance Constraint. & \specialcell{Continuous optimization.\\Approximation of LOOCV\\error at unobserved points.} \\ \\
AME & \specialcell{$\hat{e}^{\text{LOOCV}}$ + adjusted\\prediction variance} & \specialcell{Unsampled point\\error equal to\\closest sample.} & \specialcell{Continuous optimization.\\Adaptive adjustment of E \& E. } \\ \\
CVVor & $\hat{e}^{\text{LOOCV}}$ in each cell & Distance Constraint. & \specialcell{Discontinuous optimization.\\Voronoi tessellation.}  \\ \\
EI &  $y_{min} - \mu_{\hat{Y}}$   & $\sigma_{\hat{Y}}^{2}(\bm{x})$ & \specialcell{Continuous optimization.\\Fixed balance between E \& E.\\Performs well in low\\dimensionality.} \\ \\
EIGF &   $\mu_{\hat{Y}} - y(\bm{x}_{\star})$ & $\sigma_{\hat{Y}}^{2}$ & \specialcell{Continuous optimization.\\Fixed balance between E \& E.} \\ \\
LOLA &  Gradient in cell. & Volume of cell. & \specialcell{Discontinuous optimization.\\Voronoi tessellation.\\Fixed balance between E \& E.}  \\ \\
MASA &  \specialcell{Prediction variance\\between committee\\ members} & $D_{min}$& \specialcell{Discontinuous optimization.\\Fixed balance between E \& E.}  \\ \\
MEPE &  $\hat{e}^{\text{LOOCV}}$ & $\sigma_{\hat{Y}}^{2}$ & \specialcell{Continuous optimization.\\Adaptive adjustment of E \& E. }  \\ \\
MIPT &  - & $D_{min}$  &  \specialcell{Discontinuous optimization.\\Space-filling property.}  \\ \\
MSD & - &$D_{min}$  &  \specialcell{Discontinuous optimization.}    \\ \\
CVD & $\hat{e}^{\text{LOOCV}}$ & Distance considered. & \specialcell{Continuous optimization.}\\ \\
SFCVT & \specialcell{Metamodel for $\hat{e}^{\text{LOOCV}}$} &  Distance Constraint. & \specialcell{Continuous optimization.}    \\ \\
SSA &  Departure function. & $D_{min}$& \specialcell{Continuous optimization.\\Multiple NLP solved.\\Multiplication of E \& E. } 
\end{tabularx}}
\end{center}
\caption[Overview of implemented adaptive sampling algorithms]{Overview of implemented adaptive sampling algorithms ($\hat{e}^{\text{LOOCV}}$: LOOCV error, $y_{min}$: Minimum value of sample points, $\mu_{\hat{Y}}$: Prediction value of Kriging model, $y(\bm{x}_{\star})$: Response value of closest sample to $\bm{x}_{\star}$, $\sigma_{\hat{Y}}^{2}(\bm{x})$: Prediction variance, E \& E: Exploration and Exploitation, NLP: Nonlinear programming with constraint, $D_{min}$: Nearest neighbor distance)}\label{table::OverViewMethods}
\end{table}
In the next section additional design considerations of surrogate models with adaptive sampling are discussed.
\clearpage
\section{Design consideration of surrogate models with adaptive sampling}
The design considerations made for the benchmark problems and applications of this thesis are reviewed in this section. Specifically the size of the initial data and the stopping criterion. Here, the literature is reviewed for these difficulties and a sensible approach is chosen for each case.
\subsection{Initial data}
A sampling procedure, specifically an adaptive one, begins with some initial sample points. For this one-shot and/or sequential space-filling sampling procedures can be employed.  As \cite{kleijnen2009kriging} point out LHD is a commonly used data generation technique. 
Assume that the input space $\mathbb{X}$ is a $\left[0, k-1 \right]^{n}$ hypercube. Then the
$n$-dimensional LHD of k points, is a set of k points of the form $x_{i} = \left( x_{i1}, \, \ldots, \,  x_{in} \right) \in \lbrace 0, \, \ldots , \, k-1 \rbrace^{n}$, such that for each dimension $j$ all $x_{ij}$ are distinct \citep{husslage2011space}.
Two reasons make LHD popular:
\begin{enumerate}
\item LHD is space-filling, see \cite{crombecq2011efficient}. This is especially needed when no details of the mapping is available. Hence, it is important to be able to obtain information from the entire input space $\mathbb{X}$. To further improve the space-filling property, LHD can be combined with the maximin criterion \citep{van2007maximin}.
\item LHD is non-collapsing \citep{husslage2011space}. The collapsing property (see e.g. \cite{janssen2013monte}) describes the phenomenon that when one of the design parameters has almost no influence on the response, then two sample points that are only different in this parameter can be considered as the same point. Hence, they will be evaluated twice to create the surrogate model, which especially for Kriging creates ill-conditioned matrices. The non-collapsing property is enforced by LHD, which means that after removing one or more parameters the spatial design is still useable.
\end{enumerate} 
For these reasons LHD will be used to create the initial data in this thesis in form of the approach proposed by \cite{viana2010algorithm} called \acrfull{tplhd} (a LHD obtained via the translational
propagation algorithm) with a one-point seed. 
TPLHD \label{page::TPLHD} is able to obtain near optimal Latin hypercube
designs without using formal optimization. This leads to less computational effort with results basically provided
in real time. Its aim is to approximately find the the solution to the optimization problem quickly instead of focusing on the best possible solution. The process is found to be able to approximate the optimal solution proficiently in lower dimensions (\cite{liao2010fast} mention this as up to 6). In higher dimensions the sample positioning given by TPLHD diverges from the optimal solution. However for the cases considered in this thesis the use of TPLHD is sufficient even in higher dimensions.\\
However a problem with this creation is the fact that an assumption needs to be made of how many sample points $m_{I}$ need to be created in the initial step in order to start the adaptive sampling procedure. On one hand \cite{kim2009construction} as well as \cite{ghoreyshi2009accelerating} mention that if the initial sample size is too small and hence the metamodel is of poor quality, the adaptive sampling technique may generate points in unwanted regions of interest because global exploration was not sufficiently covered. On the other hand, if the initial sample size is too large then the computational costs are too high and should have better been spent on iterations of the adaptive sampling strategy \citep{crombecq2011novel}. \\
The number of initial sample points is dependent on the application. Specifically on the dimensionality $n$ of the input space, the quality of the initial data (space-filling) and the complexity of the mapping $\mathcal{M}$. \\
However, overall, choosing the sample size for deterministic computer experiment is important but lacks formal guidance. Some researchers presented empirical formulas and rule of thumbs for their own applications. 
A study of the dimensionality and the respective initial data size can be found in \cite{liu2016adaptive}. \\
\cite{jones1998efficient} introduced the rule that $m_{I} = 10 n$. This formula was used and investigated for gaussian processes in  \cite{loeppky2009choosing} who came to the conclusion that it is a reasonable rule of thumb. Furthermore the authors present suggestions for follow up strategies if the rule was a posteriori found to be insufficient. For the applications of this thesis the rule was employed. 
\subsection{Stopping criterion}
When using the adaptive sampling technique in metamodeling a stopping criterion needs to be included to determine at what point of the iteration enough sample points were generated. Generally four stopping criteria can be distinguished. 
\begin{enumerate}
\item The first one is the trivial but not to be underestimated case of external time constraints of building the metamodel, e.g. project deadline. For academic purposes this case will be neglected.  
\item A commonly used stopping criterion is dependent on the available computational budget. In practice this means the procedure will be stopped after the maximal number of mapping evaluations is reached (with the hope that the obtained surrogate model meets the
requirement for accuracy). This criterion is for example employed by \cite{martin2002use}, \cite{van2008customized} or \cite{li2010accumulative}. 
\item The created surrogate model is checked according to a desired accuracy. Possible techniques are the \acrfull{rmse} \citep{coulibaly2000daily} or the 
\acrfull{rmae} \citep{goel2009comparing}.  An overview over some of the more common error formulas is offered in Table \ref{tab::errors}. The choice of validation metric depends on the application. RMSE and the mean absolute error (MAE)
are global performance metrics, while RMAE reflects the presence of poor prediction in local areas. 
The smaller the values of RMSE, MAE and RMAE the more accurate the surrogate. \\
The $R^{2}$ score can be interpreted as giving information about the goodness of fit. A value of 1 indicated a perfect fit, whereas a value of 0 signals bad prediction capability.  
\item The successive relative improvement between iterations is evaluated and when no significant improvements are apparent, the procedure is stopped.
Variations of the cross-validation error are used in this technique, e.g. \cite{dubourg2013metamodel}. \cite{kleijnen2004application} employed cross-validation and computed the jackknifing variance. Another approach was used by \cite{kim2009construction} who utilized the absolute relative error.
\end{enumerate}
Depending on the application a different stopping criterion of the aforementioned will be used.
\setlength{\tabcolsep}{15pt}
\begin{table}[h!]
\begin{center}
\begin{tabular}{cr r } \hline
Validation metric & Formula \\ \hline\hline \\
Mean absolute error &  $\dfrac{1}{m} \sum_{i=1}^{m} \abs{y_{i} - \hat{Y}_{i}}$ \\ \\
Root mean squared error & $\sqrt{\dfrac{1}{m} \sum_{i=1}^{m} (y_{i} - \hat{Y}_{i})^{2}}$ \\ \\
Relative maximum absolute error & $\dfrac{\max \left( \abs{y_{1} - \hat{Y}_{1}}, \, \ldots, \, \abs{y_{m} - \hat{Y}_{m}} \right)}{\sigma_{y_{i}}}$ \\ \\
$\text{R}^{2}$ score & $1 - \dfrac{\sum_{i=1}^{m} (y_{i} - \hat{Y}_{i})^{2}}{\sum_{i=1}^{m} (y_{i} - \overline{y})^{2}} $ 
\end{tabular}
\end{center}
\caption[Error measures used to validate metamodels.]{Error measures used to validate metamodels ( $m$: number of design points, exact responses $y_{i}$ , estimated response values $\hat{Y}_{i}$, mean of the response values $\overline{y}$, standard deviation of the exact responses $\sigma_{y_{i}}$) }\label{tab::errors}
\end{table}
\clearpage
\subsection{Normalization}
In order to avoid numerical problems arising from scaling as recommended by \cite{forrester2008c} a normalization of the input space is conducted before constructing the surrogate model. Consider the given input $\bm{x} \in \mathbb{R}^{n}$, with an upper limit $x_{i}^{u}$ and a lower limit $x_{i}^{l}$ for each dimension $i$. The normalized input value is then given as
\begin{equation}
z_{i} = \frac{x_{i} - x_{i}^{l}}{x_{i}^{u} - x_{i}^{l}} \, \text{.}
\end{equation}
This procedure limits ensures that $0 \leq z_{i} \leq 1$.  
After the construction of the surrogate model the values will be transferred back into their original limits.
\section{Concluding remark}
The framework of adaptive sampling for metamodeling has been presented. Common techniques used in the literature have been explained by a division into four groups: Cross-Validation-based approaches, Variance-based approaches, Query-By-Committee methods and Gradient-based techniques. A summary of the techniques considered in this thesis has been given. Design considerations for adaptive sampling employed in this thesis have been described. The different adaptive sampling techniques will be investigated for benchmark problems in the following chapter.
\chapter{Comparison of sampling technique in Ordinary Kriging}\label{ch::AdaptiveSamplinginOK}
The computational testing consists of two experiments to evaluate the adaptive sampling techniques. In the first experiment the adaptive sampling techniques are compared for benchmark problems. In a second step the techniques are applied to generate metamodels for a dynamic application problem in chapter \ref{chapt::DynamicApproach}. \\
The one-dimensional Schwefel function $\mathcal{M}_{Schwefel}^{1d}$ as proposed by \cite{schwefel1981numerical}, the two-dimensional six-hump camel function $\mathcal{M}_{SHC}^{2d}$  \citep{branin1972widely}, the two-dimensional Ackley function $\mathcal{M}_{Ackley}^{2d}$  (first employed by \cite{ackley1987connectionist}), the three-dimensional Hartmann function $\mathcal{M}_{H3}^{3d}$  \citep{hartman1973some}, the five-dimensional Trid function $\mathcal{M}_{Trid}^{5d}$  \citep{adorio2005mvf} and the seven-dimenional Lévy function $\mathcal{M}_{Levy}^{7d}$  \citep{laguna2005experimental} are chosen in this thesis for the comparison.
They are utilized to represent functions with different shapes and hence different demands for the accurate metamodels.
The interested reader shall be referred to \cite{jamil2013literature} and the project of \cite{WinNT} for more resources on benchmark functions for computer experiments.  
Because an evolutionary algorithm (a hybridized particle swarm optimization) is used to determine the optimal solutions for the hyperparameters the final model may vary after a computation. To avoid unrepresentative results, the models are constructed 10 times and the results are averaged over these computations. To measure the accuracy $5000 \cdot n$ points are randomly selected to calculate the errors. The initial space-filling sample is generated with TPLHD and the Matérn 3/2 autocorrelation function is chosen for all metamodels.
In the following sections these adaptive sampling algorithms are compared for their computational effort and for their ability to create proficient metamodels for OK and HK. 
\section{Computation time in higher dimension}
\begin{figure}[hbtp]
\centering
\includegraphics[scale=0.7]{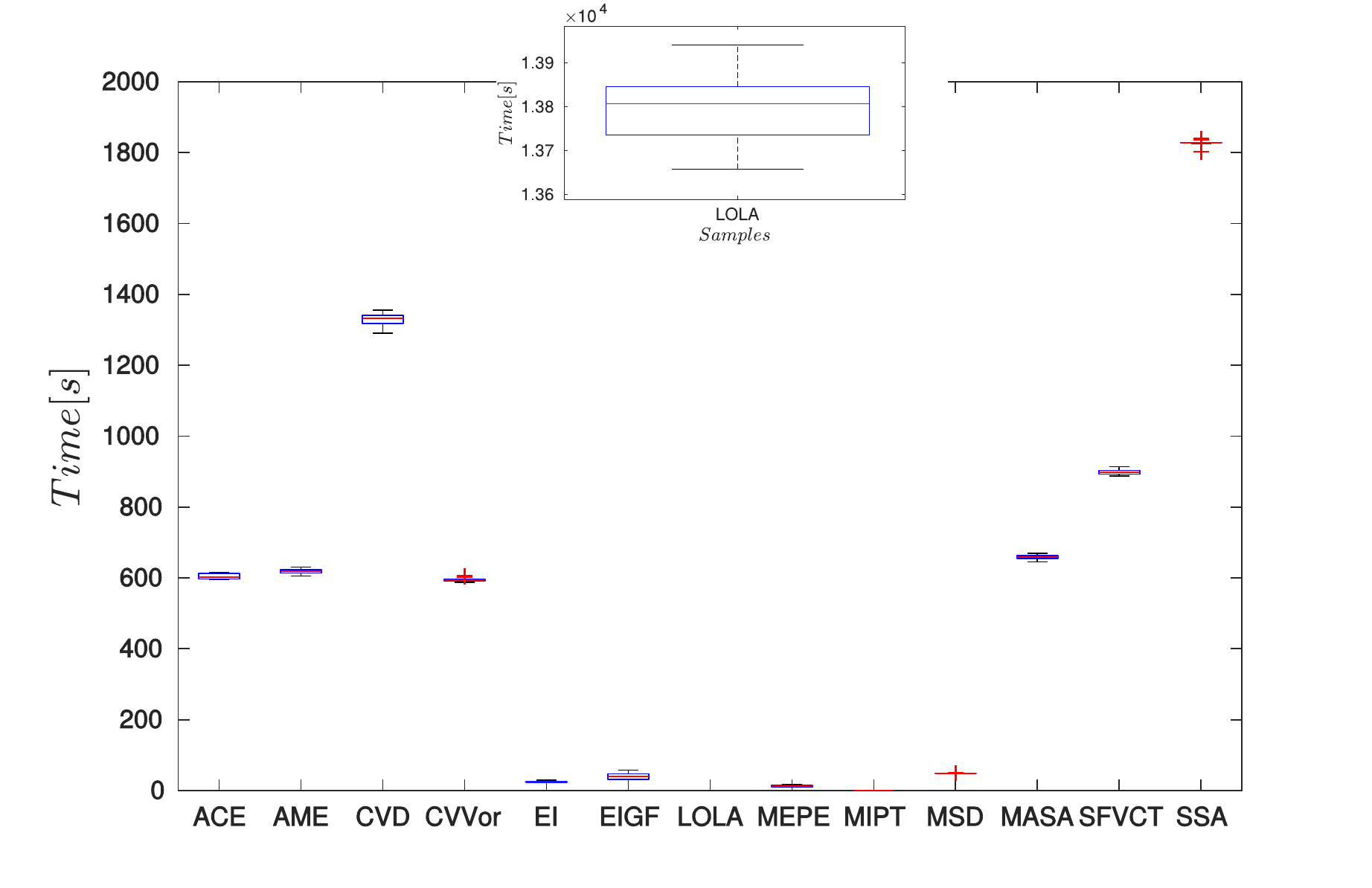}
\caption[Time comparison Ackley function.]{Comparison of time needed to find the 41st sample for $\mathcal{M}_{Ackley}^{2d}$.}\label{fig::timeAckley}
\end{figure}

This section compares the computational time of the adaptive sampling techniques as described the last chapter.
The comparison will be done on the two dimensional Ackley-function $\mathcal{M}_{Ackley}^{2d}$  and the five-dimensional Trid-function $\mathcal{M}_{Trid}^{5d}$. 
For the Ackley function the average time needed to find the 41st sample point after 40 initially created sample points is considered whereas for Trid the time to evaluate the 101st sample starting from 100 initial samples created with TPLHD will be investigated. \\
Since the methods vary between continuous and discontinuous optimizations with or without nonlinear constraints a general comparable approach is difficult to generate. However in order to make the comparisons viable the following restrictions will be set:
\begin{itemize}
\item The code performance is optimized to the best of the authors knowledge and experience.
\item Continuous optimizations as needed with EI,EIGF,SSA,MSE,ACE, MEPE or SFCVT will be done with the default genetic algorithm provided by the Matlab software. The default options will be kept except for the population size which will be set to 1000 times the number of dimensions n.
\item Methods requiring a form of the Monte Carlo method (LOLA, MIPT, MSD, CVVor) are evaluated with a fixed number of 100*$n$*$m$ randomly generated points ($n$:Number of dimensions, $m$:Number of samples points).
\item The MLE optimization for the hyperparameters will also be solved with the genetic algorithm as described above.
\item The average time will be computed over 10 iterations.
\item The computations are carried out on the cluster system at the Leibniz University of Hannover, Germany, to circumvent local computation issues.
\item No parallelization or forms of threading are used. 
\end{itemize} 
For the Ackley function the results can be seen in Figure \ref{fig::timeAckley}. The results are illustrated in boxplot format. The idea behind a boxplot is the following: A central mark is used to show the median of the value of the data. A box is used to show the 25th and 75th percentiles respectively. Whiskers extend until the most extreme data points without considering outliers. The symbol $+$ is used to indicate outliers of data.
The continuous techniques EI, EIGF, MEPE, MIPT and MSD require the least computational effort to find the next sample.\\ LOLA appears as an outlier as 13800 seconds are required. This can be explained by the need to find the best neighbors of sample points for the calculation of the gradient at a point. When trying to find the next point with LOLA starting from 40 then for each sample point a function score needs to be evaluated which for the two-dimensional case involves the computation of all possible neighborhood combinations for each point. This results in $40 \frac{39!}{4! (39-4)!} = 3290040$ evaluations. This is why the LOLA algorithm requires a lot of computational effort in comparison to other techniques. As a result, this thesis will not consider LOLA for the benchmark problems that have more than two dimensions. The rest of the methods need between 600 and 2000 seconds with CVD and SSA needing considerably more time then the other ones. 
\begin{figure}[h!]
\centering
\includegraphics[scale=0.7]{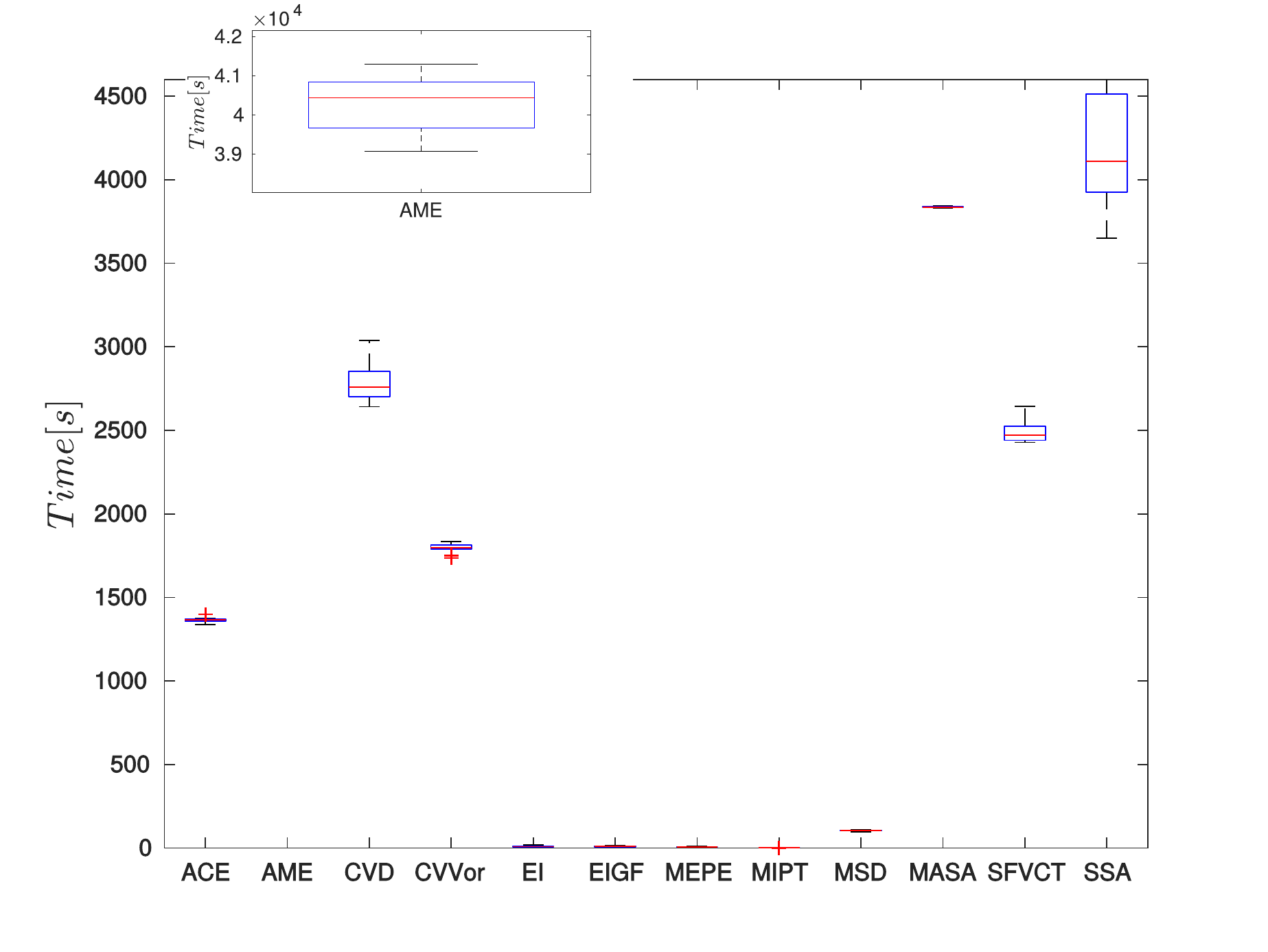} 
\caption[Time comparison Trid function.]{Comparison of time needed to find the 101st sample for the five-dimensional Trid function $\mathcal{M}_{Trid}^{5d}$.}\label{fig::timeTrid}
\end{figure}
For the five-dimensional Trid function the results are illustrated in Figure \ref{fig::timeTrid}. Similarly to the results for the two-dimensional case the five continuous techniques (EI, EIGF, MEPE, MIPT and MSD) need considerably less time than the other methods. It can be seen that all of the methods took longer (between 3 and 4 times). The methods based on LOOCV are usually slower (MEPE utilizes an approximation of this technique and is therefore faster). AME is an outlier. This variance-based method needs notably more computational effort. This is due to the adjustment of the autocorrelation function (see Algorithm \ref{alg::AME}).
\newpage 
It can be observed that in the present implementation of this thesis the most computational effort is put into the assembly of the autocorrelation matrix. Hence, sampling with AME takes longer in higher dimensions with more sample points. Therefore AME will not be regarded for problems with dimensions 5 or higher. 
\section{OK benchmark problems}
The sampling techniques of Table \ref{table::OverViewMethods} are utilized to generate OK metamodels on benchmark problems of varying dimension. The results are compared in the following sections.
\subsection{One-dimensional problem - Schwefel function}\label{sec::Schwefel}
\begin{figure}[hbtp]
 \centering
 \includegraphics[scale=0.5]{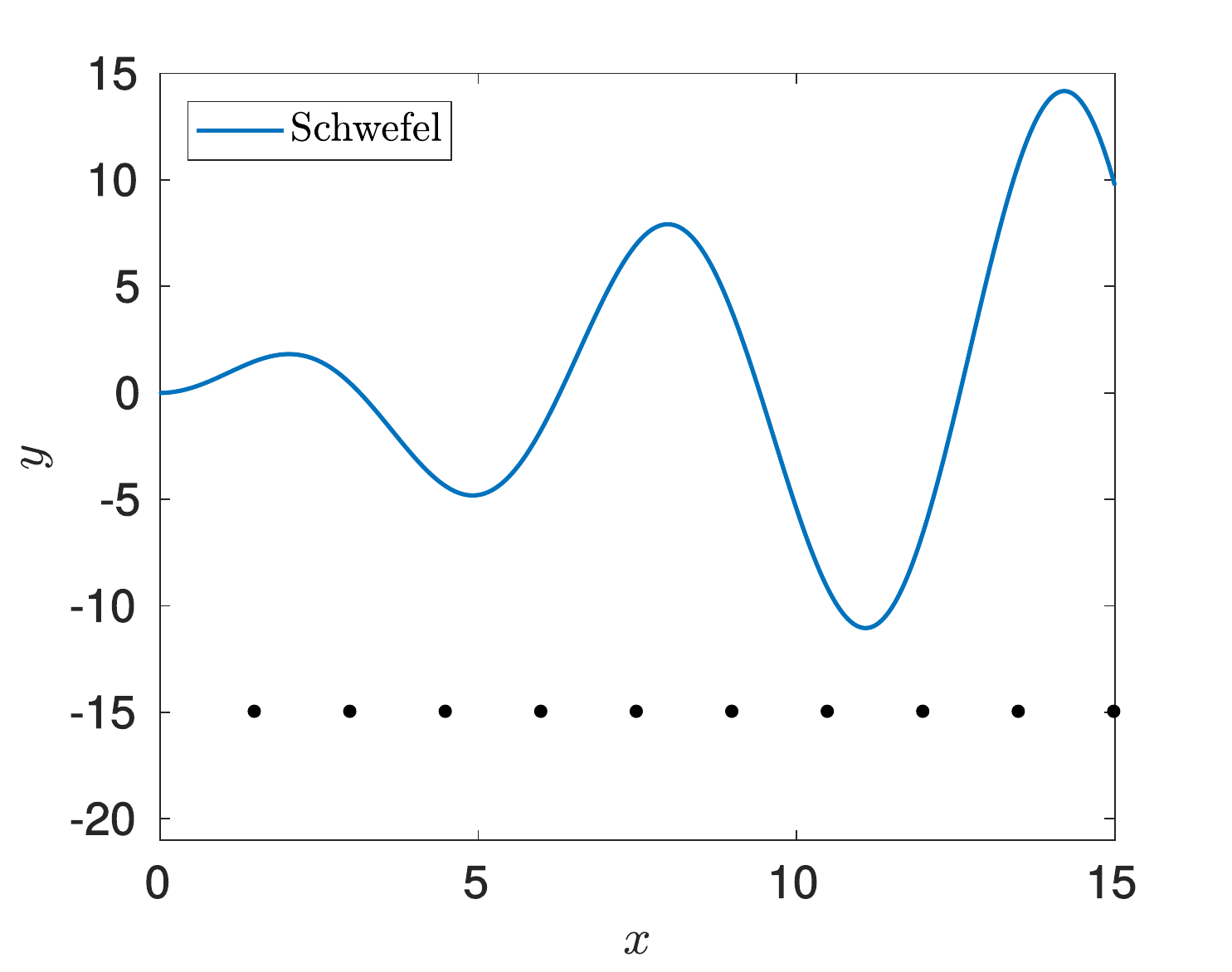}
 \caption[Plot of Schwefel function]{Schwefel function $\mathcal{M}_{Schwefel}^{1d}$ and initial samples in black.}\label{fig:1d_schwefel}
 \end{figure}

At first the results of the previously presented adaptive sampling strategies are compared for their ability to construct a metamodel for the introduced one-dimensional Schwefel function as e.g. used as a benchmark in \cite{laguna2005experimental}.
The function has many local optima and is given by 
\begin{equation}
\mathcal{M}_{Schwefel}^{1d} = x  \sin (x).
\end{equation}
The nonlinear function is depicted in Figure \ref{fig:1d_schwefel} and is constricted to the domain $[0,15]$, in which it can be observed that the function has 2 local minima and 3 local maxima. 

Furthermore the initial samples created by TPLHD are represented by black dots in Figure \ref{fig:1d_schwefel}. For this benchmark an initial $m=10$ observations are randomly chosen with TPLHD as described earlier. It can be seen that initially no sample is found around the lower bound of the parametric space at $x=0$. A good adaptive sampling technique is expected to find at last least one point close to this boundary. This needs to be done with the global exploration of the method. Furthermore it can be seen that there are no samples close to the global minimum and maximum of the function. Here the local exploitation needs to act to sample closer to these required values. \\
In a first step the error of the metamodel created with the initial samples is compared to the error of the Kriging models generated by adding 10 additional samples by utilizing the adaptive sampling techniques. Furthermore a comparison is done with the surrogate model created by 20 randomly selected samples with TPLHD. Here four error measures (MAE, RMAE, RMSE and R$^{2}$) are determined and the results are summarized in Table \ref{table::schwefel}. ACE and CVVor are not able to consistently bring the number of samples to 20 before the process fails because of sample clustering issues. All other strategies are able to achieve a considerable improvement compared to the initially constructed surrogate model. For this benchmark CVD is the method which locates new samples the best.\\ However, TPLHD, the randomly space-filling LHD technique, performs well compared to most other sampling techniques. 
\begin{table}[t!]
\begin{center}
\resizebox{1.0\textwidth}{!}{%
\begin{tabular}{l|l c c c c} \hline
 &Method & MAE & RMAE & RMSE & R$^{2}$ \\ \hline\hline \\
\multirow{1}{*}{\shortstack[l]{Errors after\\10 samples}} & TPLHD &  0.4828 & 0.4094 & 0.7557 & 0.9859 \\\\  \hline \\
\multirow{14}{*}{\shortstack[l]{ Errors after\\20 samples}} &TPLHD &  0.0249 & 0.0769 & 0.0786 & 0.9998 \\
&ACE &  -& -& -&-\\
&AME &  0.0252 & 0.07541 &0.0768 & 0.9998\\
&CVD & \textbf{0.0100} & \textbf{0.0180} & \textbf{0.0179} &\textbf{ 1.0} \\
&CVVor & -&- & -& -\\ 
&EI &  0.2434 & 0.3345 & 0.5017 & 0.9937 \\
&EIGF &  0.06579 & 0.07648 & 0.1171 &0.9996 \\
&LOLA &  0.0806 & 0.1294 & 0.1870 & 0.9991  \\
&MASA &  0.1522 & 0.2101 & 0.3143 & 0.9975 \\
&MEPE &  0.0310 & 0.0370 & 0.0594 & 0.9999 \\
&MIPT &  0.0355 & 0.0771 & 0.0886 & 0.9998  \\ 
&MSD & 0.0813 & 0.0651 & 0.1310 & 0.9995 \\
&SFCVT & 0.0521 & 0.0357 & 0.0803 & 0.9998  \\
&SSA & 0.0173 & 0.0203 & 0.0317 & \textbf{1.0}  
\end{tabular}
}
\end{center}
\caption[Error measures for $\mathcal{M}_{Schwefel}^{1d}$ after 20 samples.]{Error measures for $\mathcal{M}_{Schwefel}^{1d}$ after 20 samples (methods with clustering problems are indicated by empty rows).}\label{table::schwefel}
\end{table}
In a second step it is investigated how many samples with each sampling method are needed to reduce the MAE below $0.01$ until the limit of 50 samples. The results are presented in Table \ref{tab::SchwefelNumberofSamples}. The variance of this number over 10 iterations is symbolized by the $\pm$ value. It can be seen that there is no variation of the number for this one-dimensional case for all sampling technique except for MASA, which could be due to variations in the hyperparameter optimization process. 
\begin{table}[h!]
\begin{center}
\begin{tabular}{c c } \hline
\shortstack[l]{Sampling\\method} & \shortstack[l]{Average number\\of Samples}\\ \hline\hline \\
ACE &  - \\
AME & 23 $\pm$ 0  \\ 
CVD & \textbf{21}   $\pm$ 0 \\
CVVor & -  \\ 
EI &  - \\
EIGF & 46 $\pm$ 0  \\
LOLA  & -  \\
MASA &  43 $\pm$ 1  \\
MEPE & 23 $\pm$ 0 \\
MIPT & 31 $\pm$ 0 \\
MSD & 30 $\pm$ 0  \\
SFCVT &  - \\
SSA &  23  $\pm$ 0  
\end{tabular}
\end{center}
\caption[Average amount of samples to reach before MAE$<0.01$ for $\mathcal{M}_{Schwefel}^{1d}$]{Average amount of samples to each MAE$<0.01$ for $\mathcal{M}_{Schwefel}^{1d}$. The variation of the number of samples over 10 computations is given as an additional value. Methods that do not reach the target value because of clustering or because they need more than 50 samples have been omitted.}\label{tab::SchwefelNumberofSamples}
\end{table}

\begin{figure}[h!]
\centering
\includegraphics[scale=0.45]{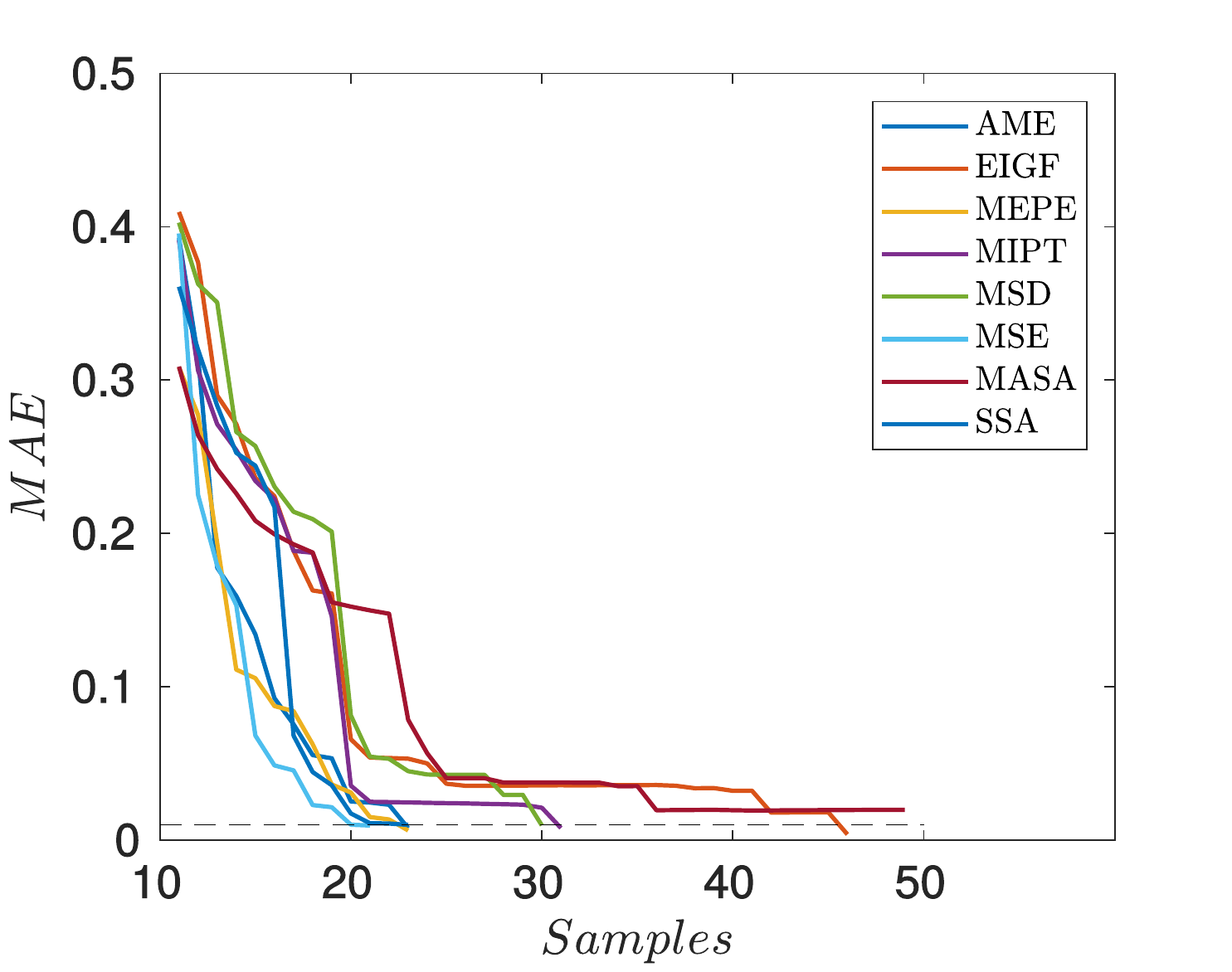}
\caption[Convergence of MAE error for $\mathcal{M}_{Schwefel}^{1d}$]{Convergence of MAE error for the respective sampling methods until threshold value of 0.01 is reached for $\mathcal{M}_{Schwefel}^{1d}$.}%
\label{fig:SchwefelConvergence}%
\end{figure}

Generally 8 out of 13 sampling techniques are able to reduce the MAE error to this threshold within the first 50 samples.
 It can be seen that CVD, SSA and MEPE need the least amount of samples on average with CVD just needing 21. The number of needed samples is illustrated in Figure \ref{fig:SchwefelConvergence} by showing the convergence of the MAE value.  

\begin{figure}[hbtp]
\centering
\begin{subfigure}[t]{0.5\textwidth}
\includegraphics[scale=0.35]{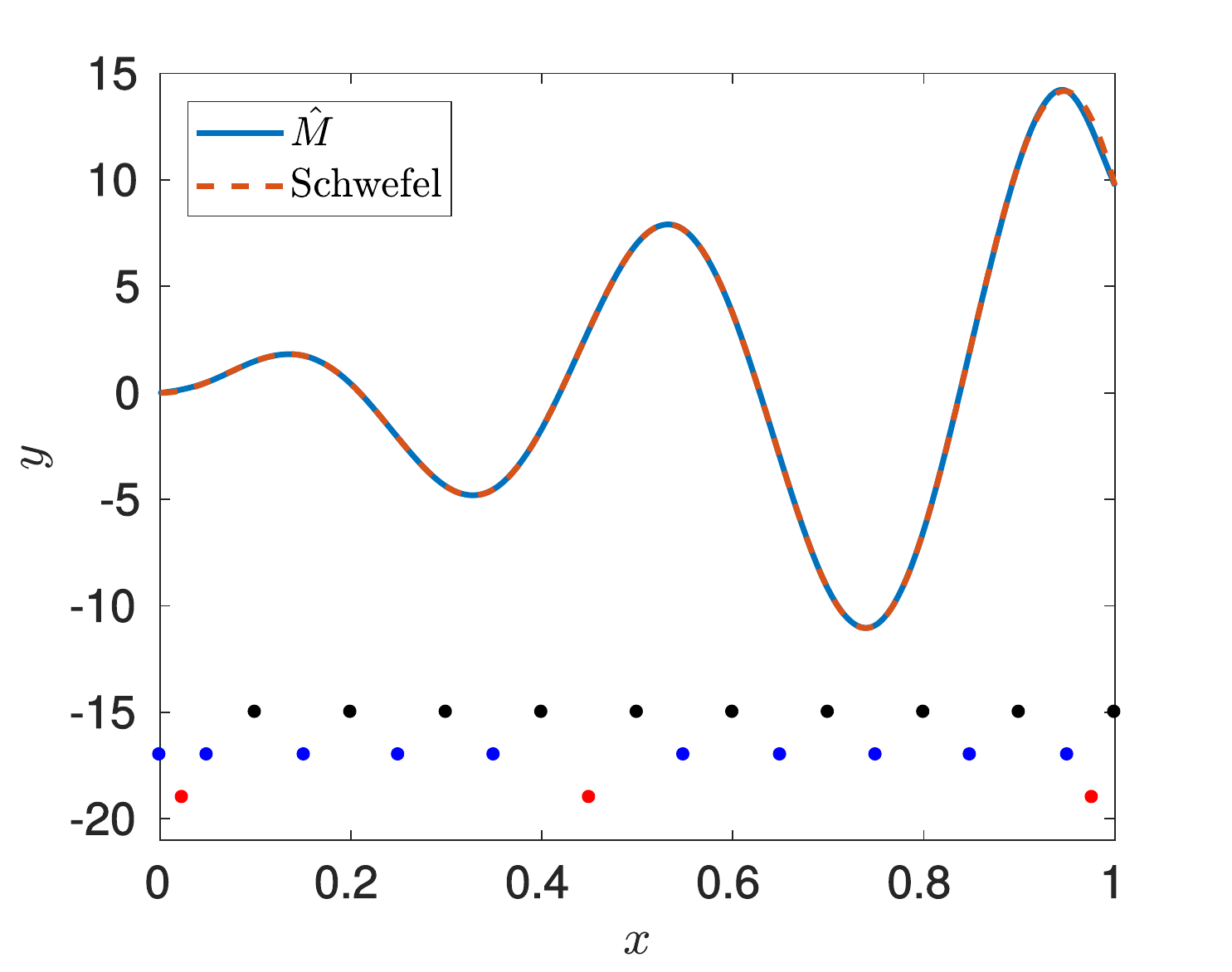}
\subcaption{AME - 23 samples}\label{fig:SchwefelYesAME}
\end{subfigure}%
\begin{subfigure}[t]{0.5\textwidth}
\includegraphics[scale=0.35]{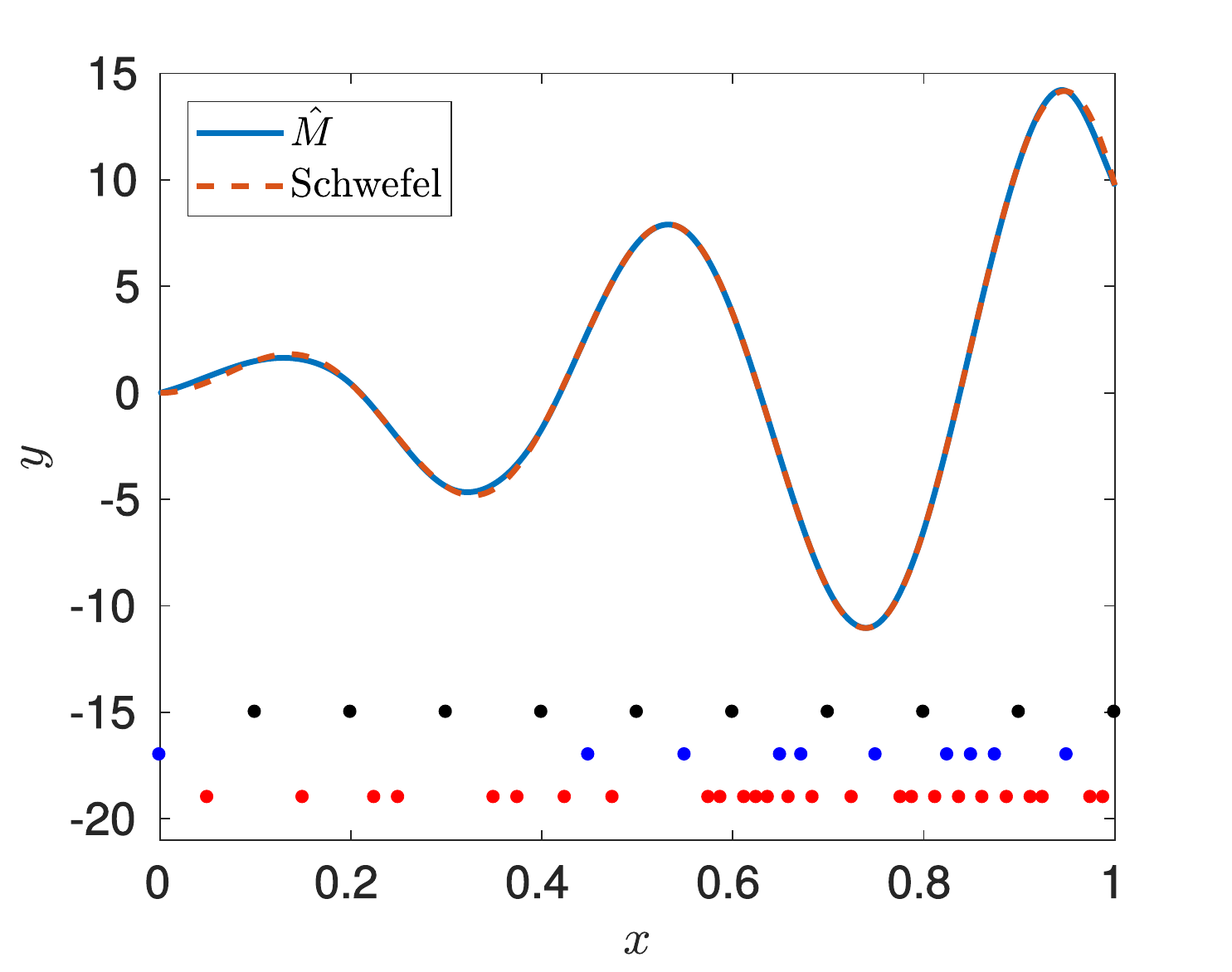}
\subcaption{EIGF - 46 samples}\label{fig:SchwefelYesEIGF}
\end{subfigure}
\begin{subfigure}[t]{0.5\textwidth}
\includegraphics[scale=0.35]{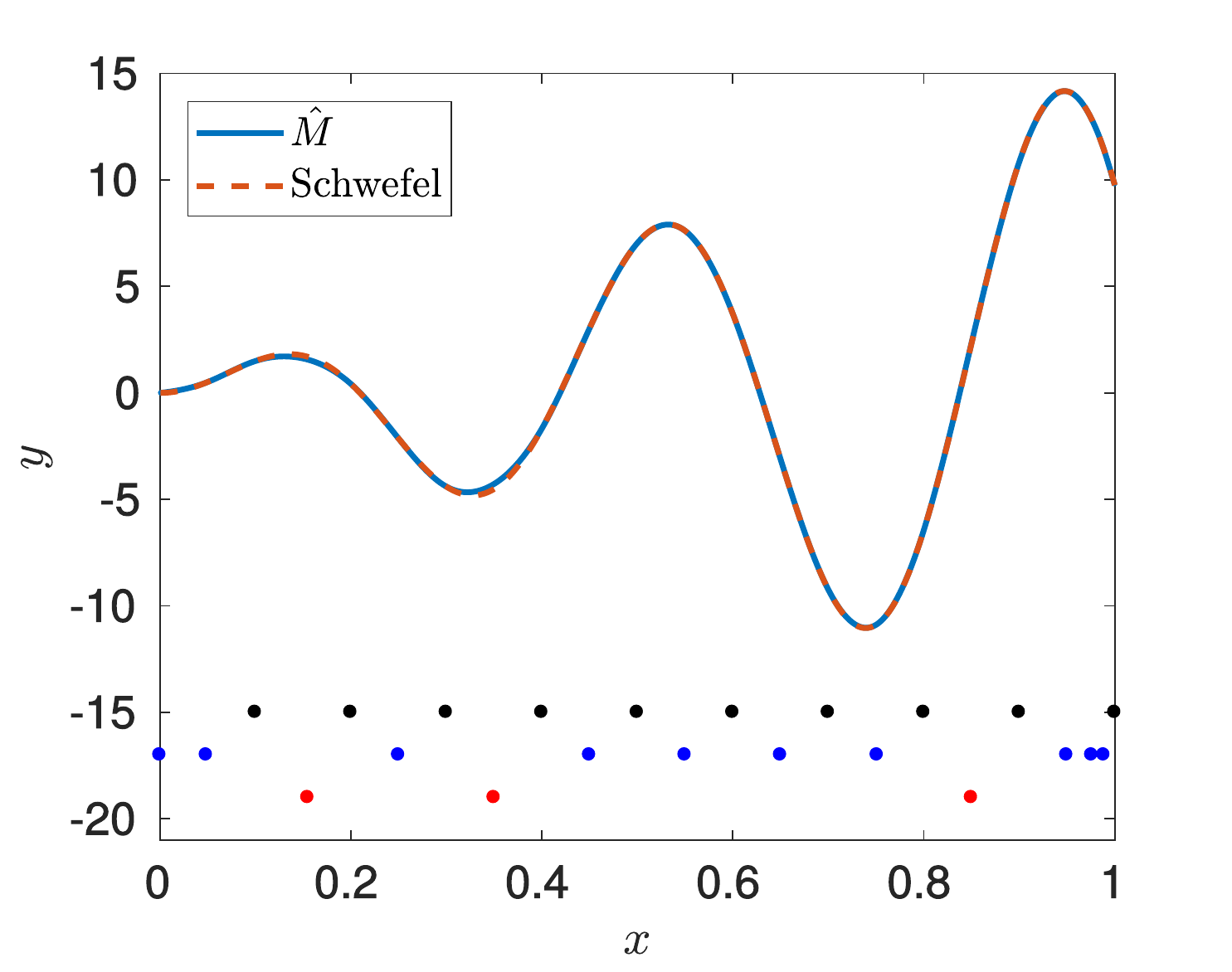}
\subcaption{MEPE - 23 samples}\label{fig:SchwefelYesMEPE}
\end{subfigure}%
\begin{subfigure}[t]{0.5\textwidth}
\includegraphics[scale=0.35]{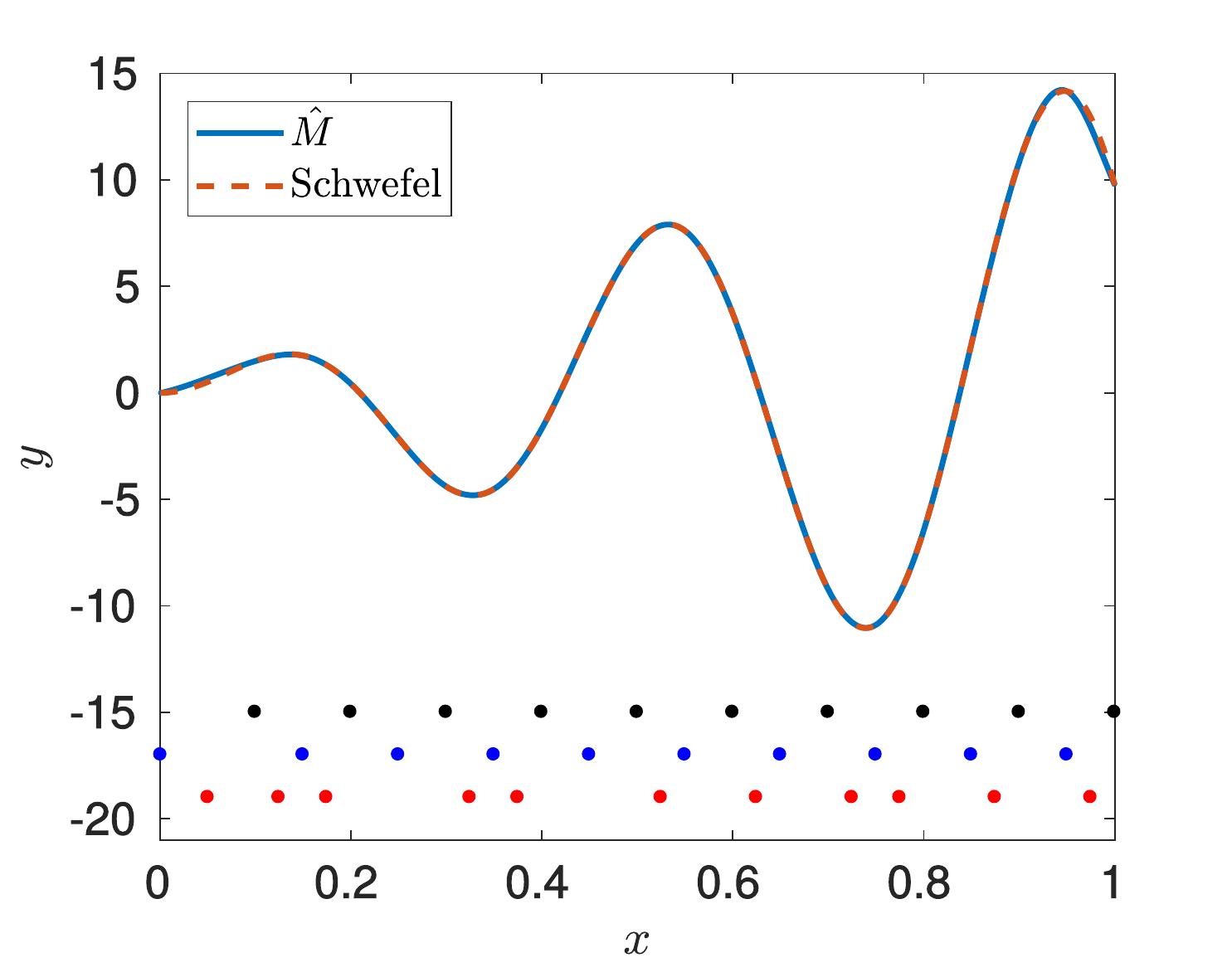}
\subcaption{MIPT - 31 samples}\label{fig:SchwefelYesMIPT}
\end{subfigure}
\begin{subfigure}[t]{0.5\textwidth}
\includegraphics[scale=0.35]{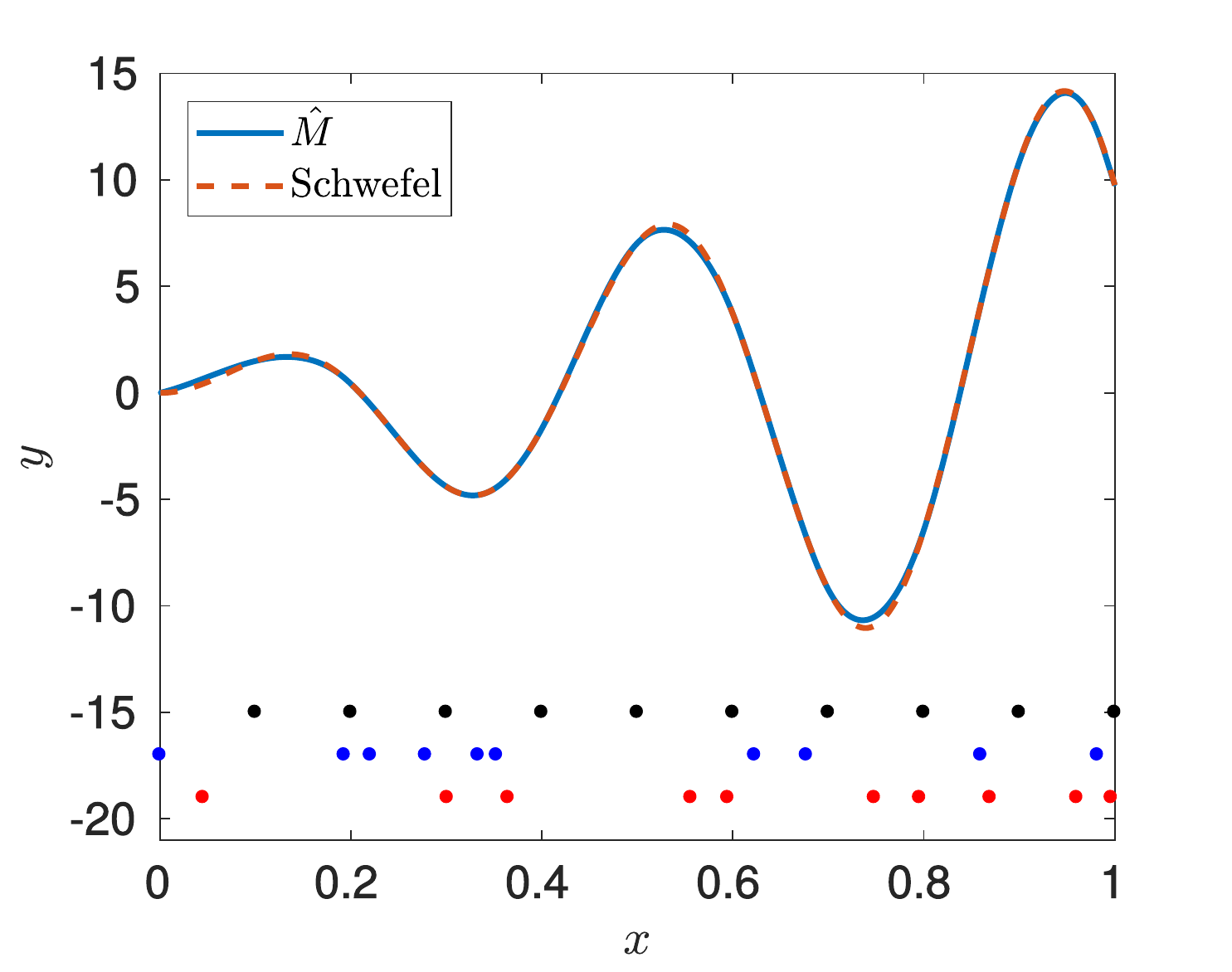}
\subcaption{MSD- 30 samples}\label{fig:SchwefelYesMSD}
\end{subfigure}%
\begin{subfigure}[t]{0.5\textwidth}
\includegraphics[scale=0.35]{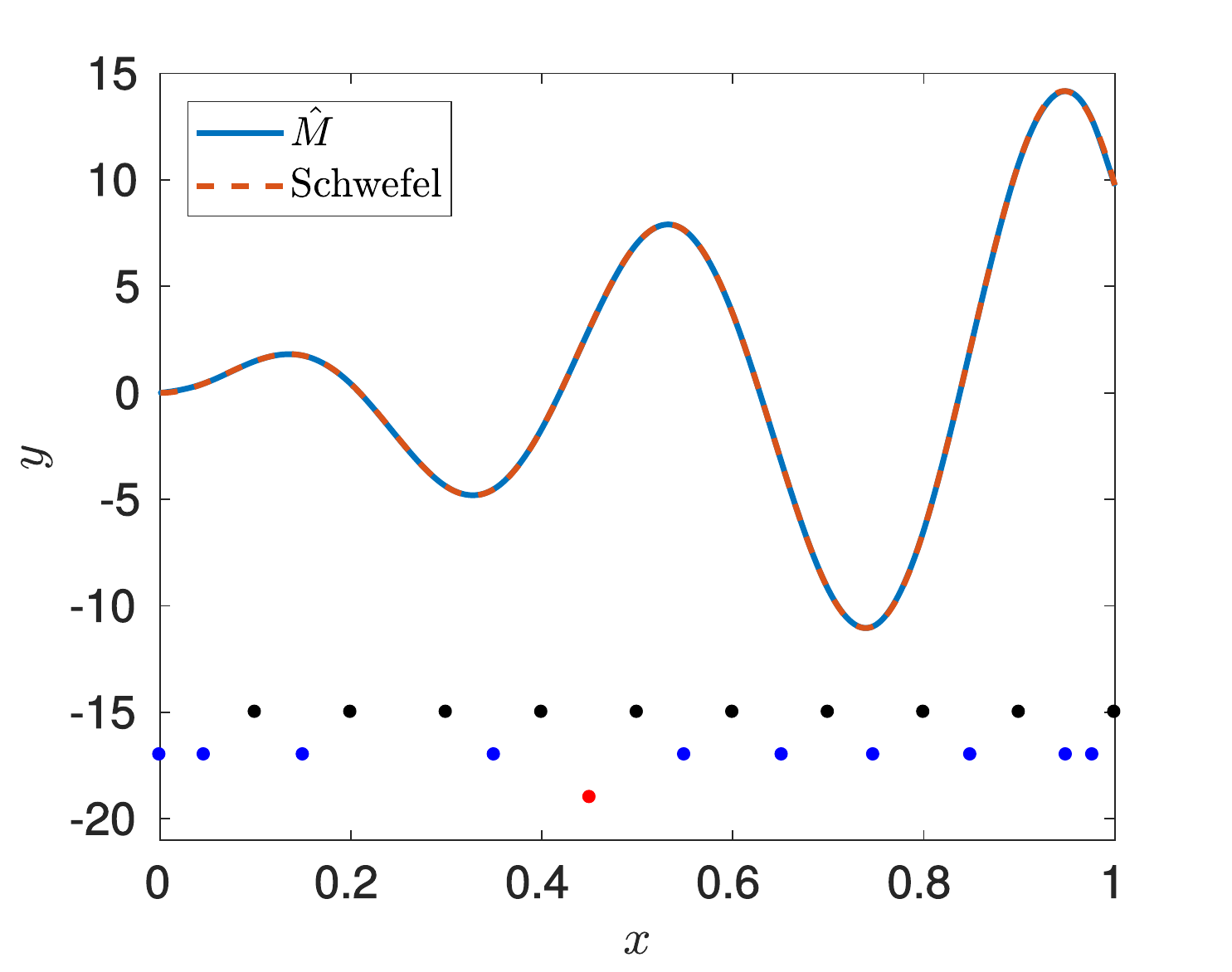}
\subcaption{CVD - 21 samples}\label{fig:SchwefelYesMSE}
\end{subfigure}
\begin{subfigure}[t]{0.5\textwidth}
\includegraphics[scale=0.35]{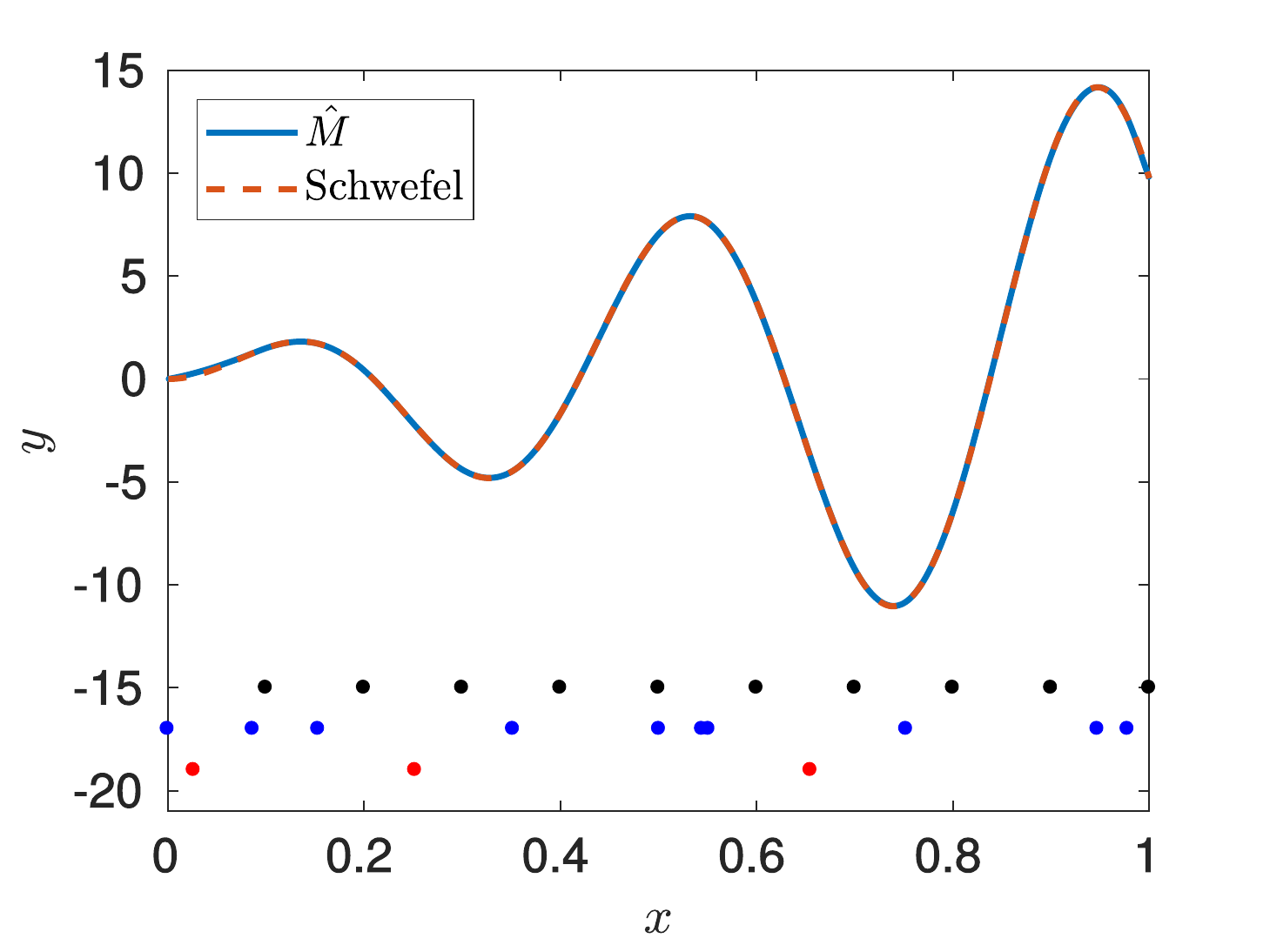}
\subcaption{SSA - 23 samples}\label{fig:SchwefelYesSSA}
\end{subfigure}
\caption[Sample locations for $\mathcal{M}_{Schwefel}^{1d}$ of methods reaching error threshold.]{Sample locations for $\mathcal{M}_{Schwefel}^{1d}$. Target function and metamodel at 20 samples are indicated by lines. Black dots: 10 initial samples. Blue dots: 10th to 20th samples. Red dots: Additional samples until MAE$<0.01$ reached. }\label{fig:SchwefelYes}
\end{figure}

\begin{figure}[h!]
\centering
\begin{subfigure}[t]{0.5\textwidth}
\includegraphics[scale=0.35]{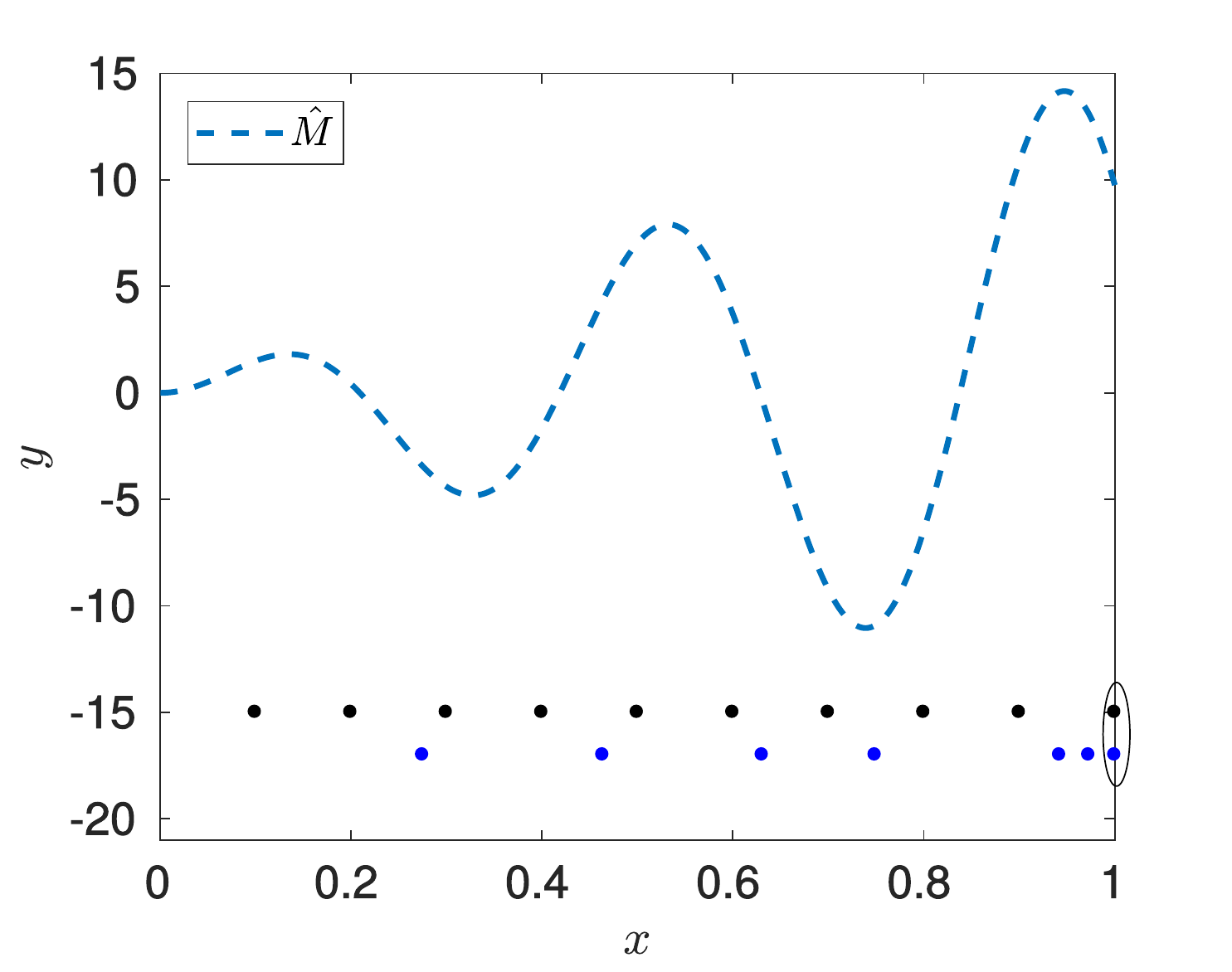}
\subcaption{ACE - 15 samples}\label{fig::SchwefelNotACE}
\end{subfigure}%
\begin{subfigure}[t]{0.5\textwidth}
\includegraphics[scale=0.35]{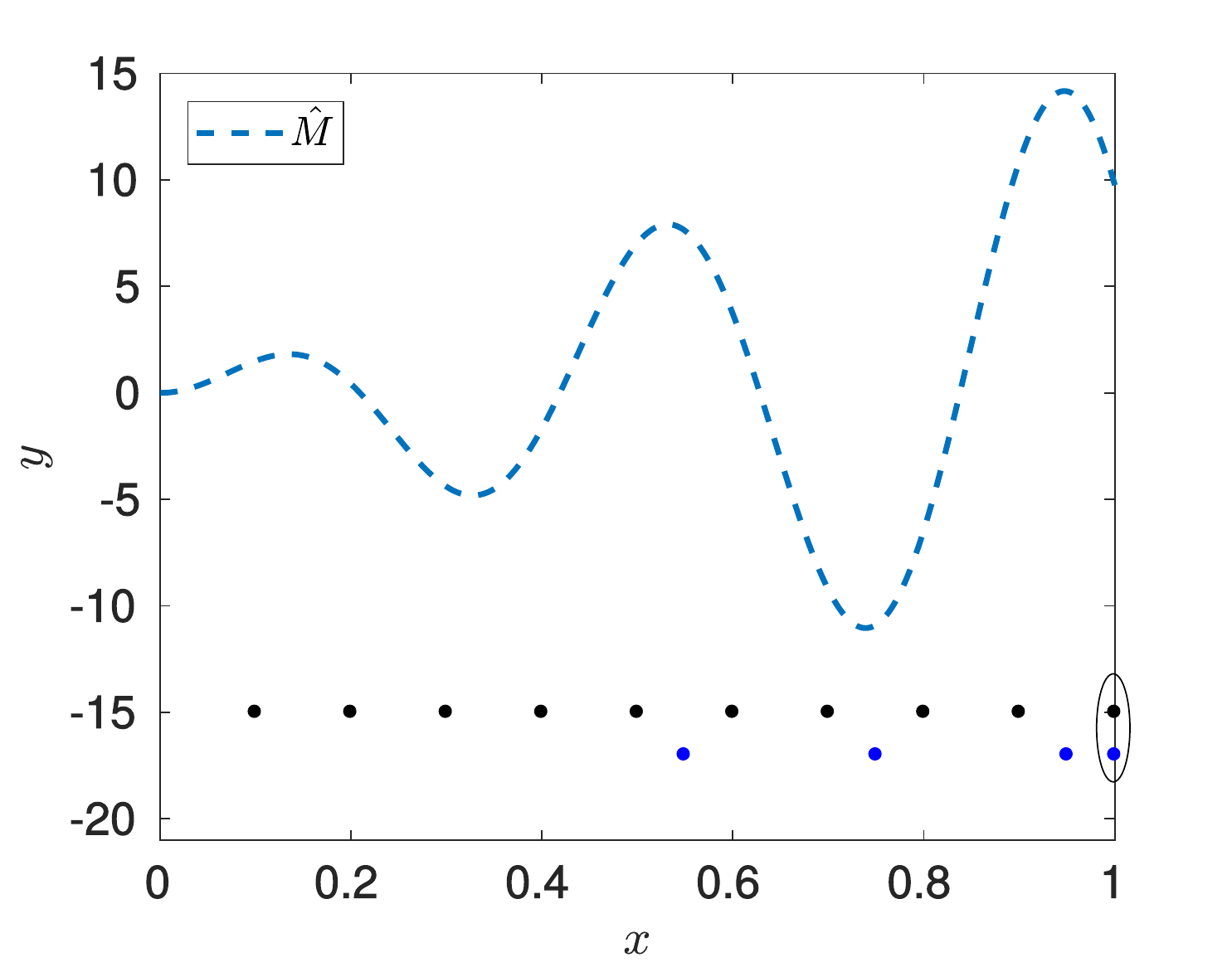}
\subcaption{CVVor - 14 samples}\label{fig::SchwefelNotCVVor}
\end{subfigure}
\begin{subfigure}[t]{0.5\textwidth}
\includegraphics[scale=0.35]{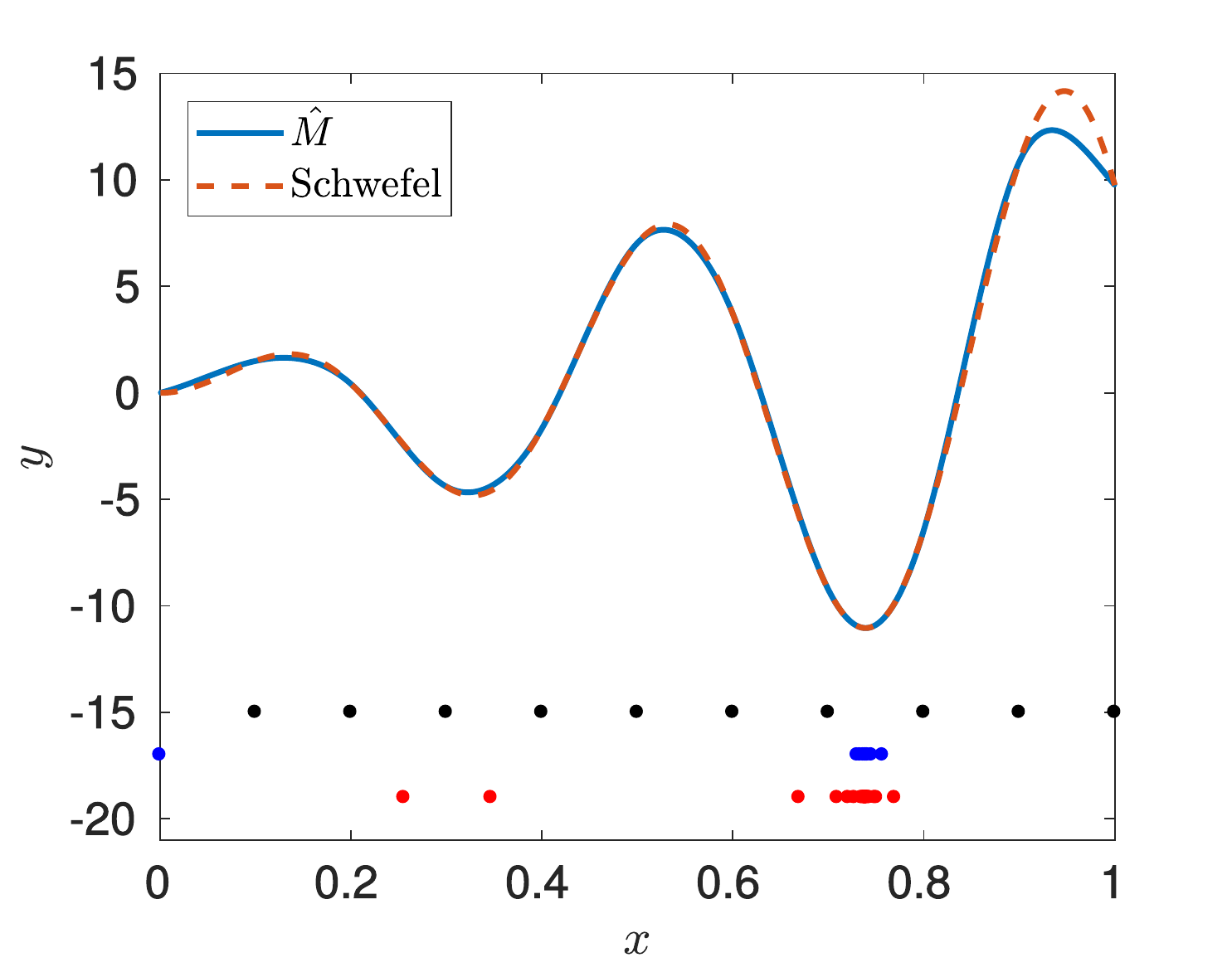}
\subcaption{EI - 50 samples}\label{fig::SchwefelNotEI}
\end{subfigure}%
\begin{subfigure}[t]{0.5\textwidth}
\includegraphics[scale=0.35]{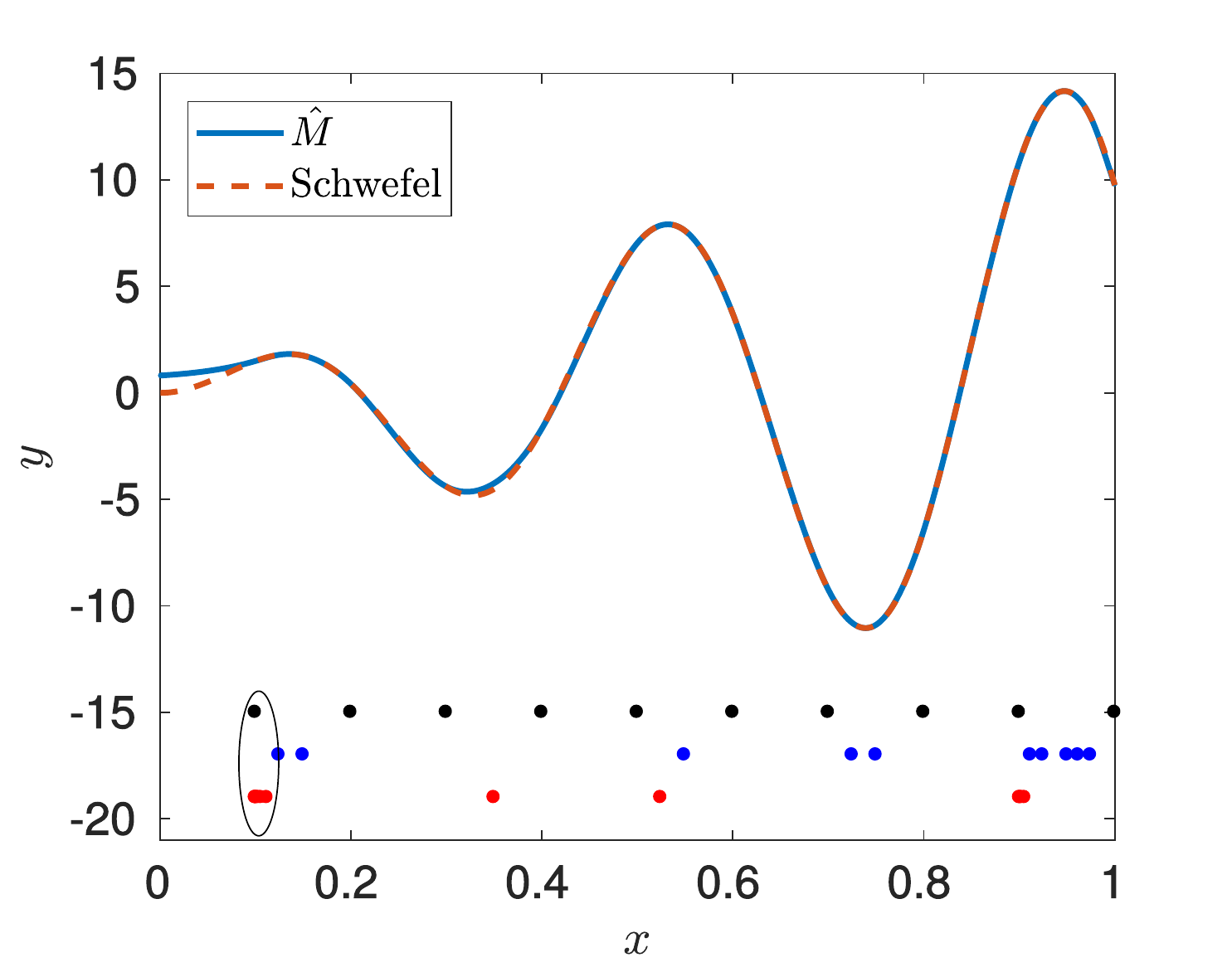}
\subcaption{LOLA - 33 samples}\label{fig::SchwefelNotLOLA}
\end{subfigure}
\begin{subfigure}[t]{0.5\textwidth}
\includegraphics[scale=0.35]{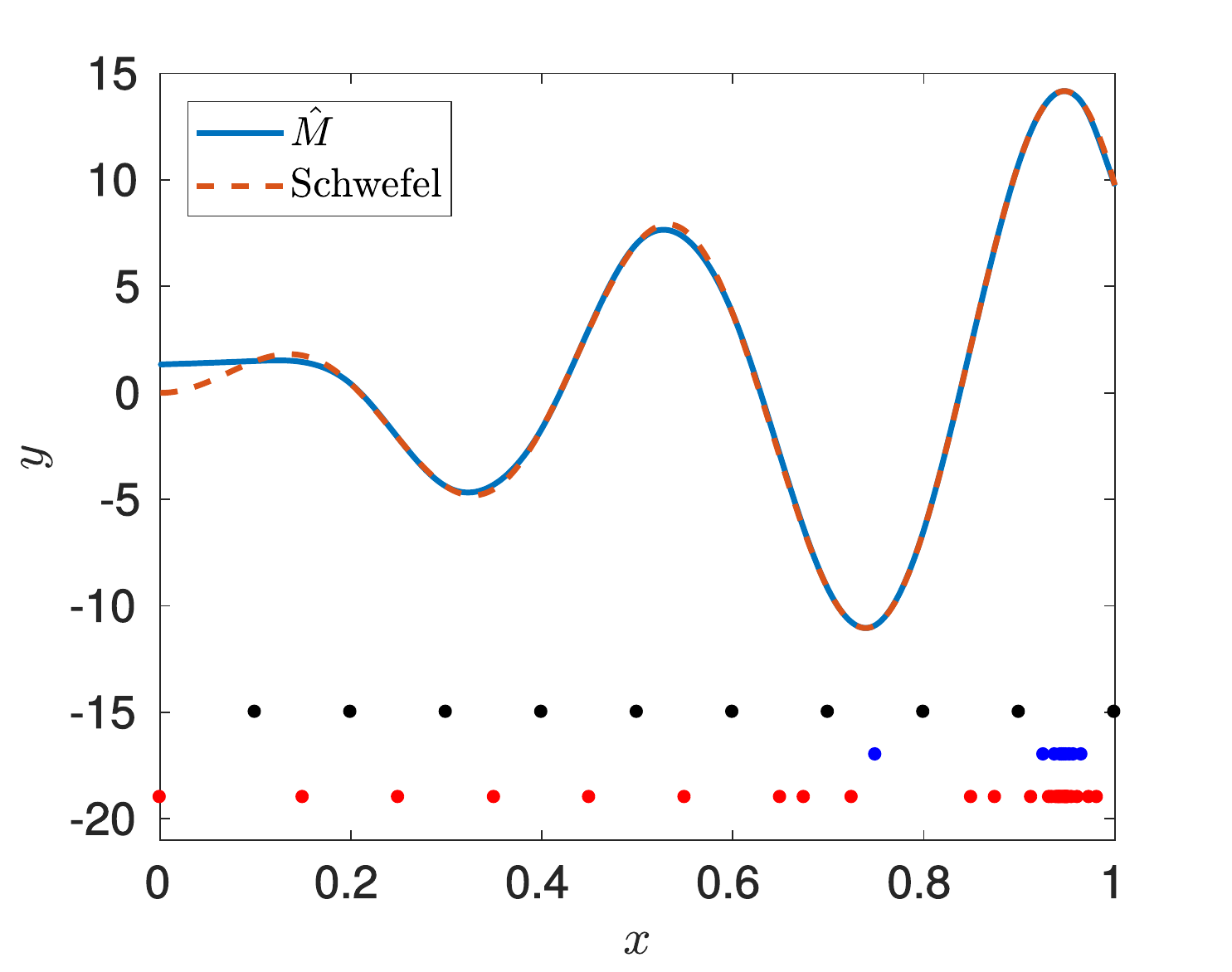}
\subcaption{MASA - 50 samples}\label{fig::SchwefelNotMASA}
\end{subfigure}%
\begin{subfigure}[t]{0.5\textwidth}
\includegraphics[scale=0.35]{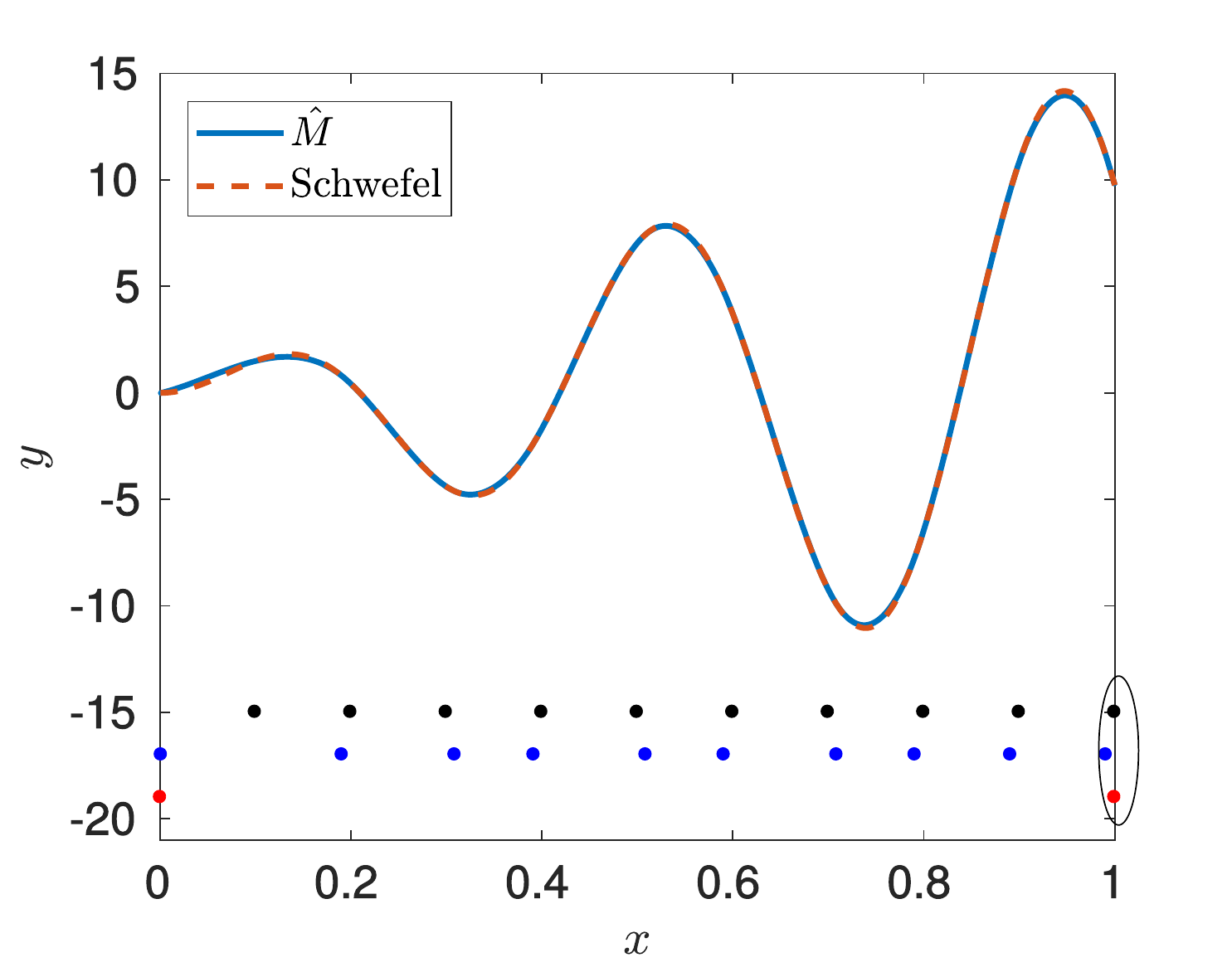}
\subcaption{SFCVT - 22 samples}\label{fig::SchwefelNotSFVCT}
\end{subfigure}
\caption[Sample locations for $\mathcal{M}_{Schwefel}^{1d}$ of methods not reaching error threshold.]{Adaptive sampling techniques unable to reduce MAE below $0.01$ for $\mathcal{M}_{Schwefel}^{1d}$. Potential cluster regions are highlighted with ellipses. EI and MASA need over 50 samples to reach threshold. Black dots: 10 initial samples. Blue dots: 10th to 20th sample. Red dots: Supplementary samples. }\label{fig::SchwefelNot}
\end{figure}

Next, the focus lies on the location of the samples points as given in Figures \ref{fig:SchwefelYes} and \ref{fig::SchwefelNot}. The results of one adaptive sampling technique is illustrated in each subplot of these Figures. The Schwefel function as well its metamodel approximation for each technique after 20 samples is plotted. \newpage Reducing the MAE means that a proficient exploration of the whole domain is needed as well as an exploitation component.\\
The sample positions of the adaptive techniques that were able to reduce the MAE below the threshold are displayed in Figure \ref{fig:SchwefelYes}. The black points are the initial sample locations. The blue points in the middle illustrate the points that are added up to 20 samples. The red points are the remaining points needed to achieve the necessary MAE value. AME is shown in Figure \ref{fig:SchwefelYesAME}.

23 samples on average are needed with AME to reduce the MAE value below the threshold. In the first 20 samples the method already samples in the locations that are needed to achieve a good approximation (around $x=0$ and around the maxima). Therefore it is already difficult to identify any differences between the two plotted lines at that stage. The last 3 points are added in positions in between previous samples. No forms of clustering and space-filling sample positions are shown with AME. 
The samples of EIGF are plotted in Figure \ref{fig:SchwefelYesEIGF}. It can be noticed that a sample is created around $x=0$ in the first 20 samples. However the method seems to focus more on the right-hand side of the plot where the largest absolute values are located. There is one sample at the $x$-value positions of all optima on this side of the parametric domain. Therefore the global minimum and maximum are exactly represented. However the left hand-side of the plot is neglected at first stage. Therefore double the amount of points are needed ($46$) as compared to AME. \\
The next technique is MEPE shown in Figure \ref{fig:SchwefelYesMEPE}. In the first 20 points all optima except the left-most are found with MEPE. Furthermore a sample around $x=0$ is created. The adaptive form of the exploitation and exploration of the method can be seen by means of the three points around the global maximum, where local exploitation of the function is prevalent. Overall MEPE shows good approximation behavior for this function with on average only 23 samples needed to reach the threshold. \\
MIPT as shown in Figure \ref{fig:SchwefelYesMIPT} is a pure exploration-based technique. It is noticeable that for the harmonic behavior of the Schwefel function this procedure is able to reduce the existing error effectively. The method needs 31 points for the reduction. With this procedure it is obvious why TPLHD is a proficient method to create the initial samples as it can be seen as a one-shot counterpart to MIPT. So since TPLHD (in the 1-dimensional case) generates equidistant samples MIPT creates the first ten samples in-between the initial samples and at the previously uncovered lower boundary. \\
Similar to MIPT, MSD (Figure \ref{fig:SchwefelYesMSD}) is also purely exploration-based. When comparing the created samples to MIPT they are not as equidistant. However only 30 samples points are needed to reduce the error measure below the given value. \\ CVD as shown in Figure \ref{fig:SchwefelYesMSE} needs the least amount of samples overall. Due to the distance constrained of the technique a space-filling spread is achieved. The exploitation part (using LOOCV) samples an additional sample close to the global maximum where the highest prediction error can be found. \\
23 samples are needed with SSA in this benchmark problem as shown in Figure \ref{fig:SchwefelYesSSA}. It can however be seen that it has problems to avoid clustering around the local maximum in the middle of the function. But since the origin and all the other optima are sampled the technique achieves an efficient reduction of the error value.
The respective plots for the adaptive sampling techniques that were unable to reduce the MAE error either because of clustering or because more than 50 points are needed, are displayed in Figure \ref{fig::SchwefelNot}. In these subplots the black points also represent the locations of the initial samples. The black points are the samples between the 10th and the 20th sample. All additional samples after this are illustrated by the red dot. \\
The ACE technique is shown in Figure \ref{fig::SchwefelNotACE}. The method runs into problems in the first 10 adaptive samples steps since a clustering issue at the upper boundary emerges (as highlighted by the ellipse around the corresponding points). This leads to an ill-conditioned autocorrelation matrix and numerical problems. \\ The same problem can be found when employing CVVor (Figure \ref{fig::SchwefelNotCVVor}). The exploration of the method is not proficient enough to sample in the lower boundary around zero in this one-dimensional problem. \\
EI is depicted in Figure \ref{fig::SchwefelNotEI}. The method is known to run into problems with clustering especially in lower dimensions since the exploration component is not pronounced enough. Even though exploration of the domain is done (sample at the lower bound) the technique can not avoid to create points around the global minimum. However since there is no superposition of points or clustering the process can continue albeit with a high condition number of the autocorrelation matrix.  \\ LOLA is a discontinuous sampling technique based on a gradient estimation with Voronoi tessellation. The generated samples are depicted in Figure \ref{fig::SchwefelNotLOLA}. The technique stops after 33 samples due to clustering at a value around $0.1$ as indicated by the ellipse in the plot. The exploration character of this technique is not sufficient enough, which can also be seen by the fact that until the 33rd sample no point is created around the lower boundary. \\
The committe-based technique MASA (Figure \ref{fig::SchwefelNotMASA}) also shows problems with clustering. In this case however around the global maximum. This problem is especially obvious in the first 10 adaptively added samples (blue points) since the exploration character of the method is low and no point around the local boundary is generated. However the exploration is proficient enough to let the technique sample at least until 50 points. \\ Lastly SFCVT is discussed (Figure \ref{fig::SchwefelNotSFVCT}). The method is proficient for the first 20 samples, as visible from the good approximation results after 20 samples, which indicates a space-filling property. Here the lower boundary is sampled. However no optima is exactly found with these samples. The 21st and 22nd samples show clustering effects at both boundaries, which leads the technique to run into numerical problems.
Overall this benchmark problem could show the general properties needed of a technique to obtain good approximation results. For this one-dimensional benchmark problem, a proficient space-filling component was necessary and sufficient for good metamodel generation.
\clearpage
\subsection{Two-dimensional benchmark problems}
\begin{figure}[h!]
\centering
\begin{tikzpicture} \begin{axis}[          
          grid=major, 
          grid style={dashed,gray!30},
          xlabel=$x_{1}$, 
          ylabel=$x_{2}$,
          xmin=0.0,xmax=1.0,
          ymin=0.0,ymax=1.0,
        ] 
\addplot+[only marks]%
table { 
x y 
0.2	0.05
0.45	0.1
0.7	0.15
0.9	0.2
0.25	0.25
0.5	0.3
0.75	0.35
0.95	0.4
0.05	0.45
0.3	0.5
0.55	0.55
0.8	0.6
1	0.65
0.1	0.7
0.35	0.75
0.6	0.8
0.85	0.85
0.15	0.9
0.4	0.95
0.65	1
}; 
\end{axis} 
\end{tikzpicture}
\caption[Normalized initial samples for the two-dimensional benchmark tests]{Normalized initial samples for the two-dimensional benchmark tests. }\label{fig::init_twoD}
\end{figure}
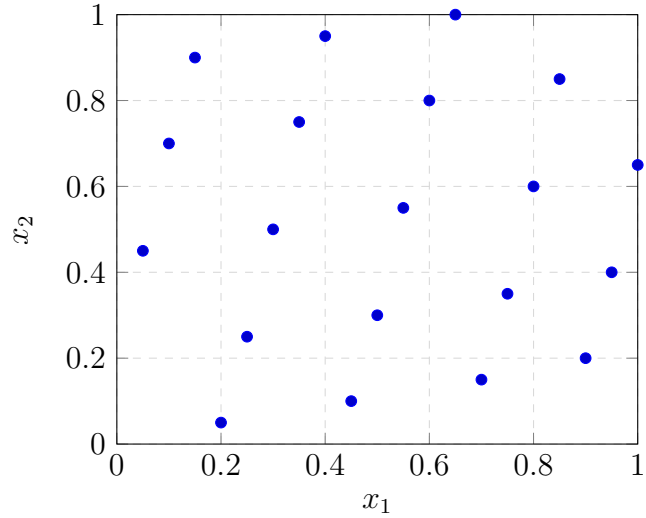

Two two-dimensional benchmark problems with different complexity are compared. $20$ initial samples are computed with TPLHD. The normalized locations are depicted in Figure \ref{fig::init_twoD}. No extra measures were taken for samples to occur at the boundary since the exploration component of the adaptive sampling technique should sample in these areas. 
\subsubsection{Ackley function}\label{sec::Ackley}
\begin{figure}[h!]
\centering
\begin{tikzpicture}
\begin{axis}
[
xlabel={$x_{1}$},
ylabel={$x_{2}$},
zlabel={$\mathcal{M}_{Ackley}^{2d}(x_{1},x_{2})$},
view={75}{25}
]
\addplot3[surf,domain=-2.0:2.0,domain y=-2:2]
{- 20 * exp(-0.2 * sqrt(0.5 * (x^2 + y^2 ) ) ) -  exp(0.5 * (cos(deg(2 * pi * x)) + cos(deg(2 * pi * y) ) )) + exp(1) + 20};
\end{axis}
\end{tikzpicture}
\caption[Two-dimensional Ackley function]{Two-dimensional plot of $\mathcal{M}_{Ackley}^{2d}$.}\label{fig::Ackley}
\end{figure}
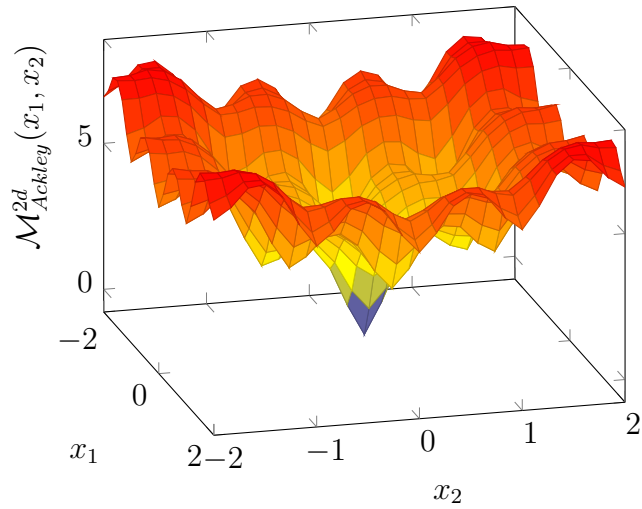

\begin{figure}[h!]
\centering
\begin{subfigure}[t]{0.5\textwidth}
\includegraphics[scale=0.35]{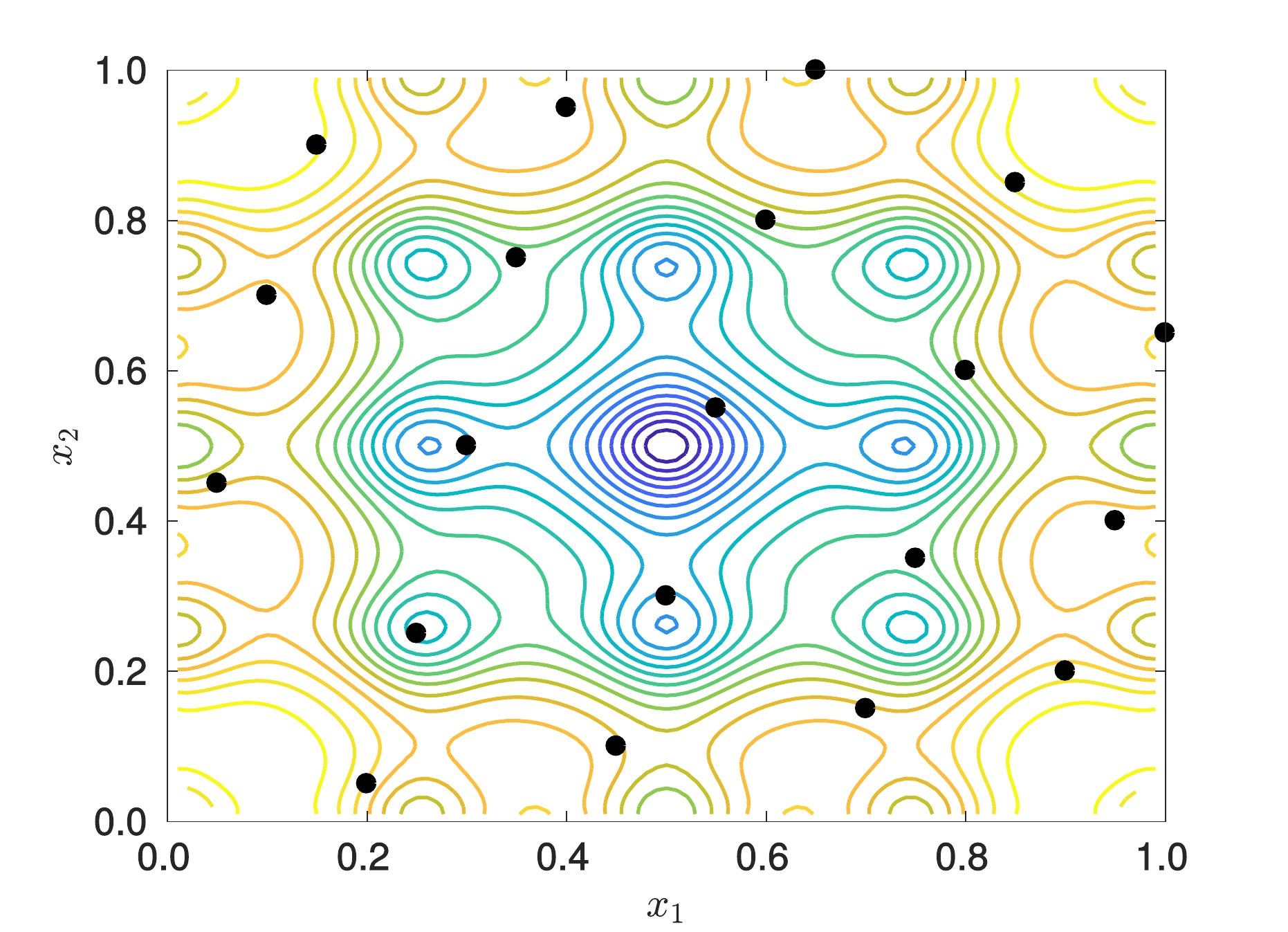} 
\subcaption{Initial samples}\label{fig:AckleyInitial}
\end{subfigure}%
\begin{subfigure}[t]{0.5\textwidth}
\includegraphics[scale=0.35]{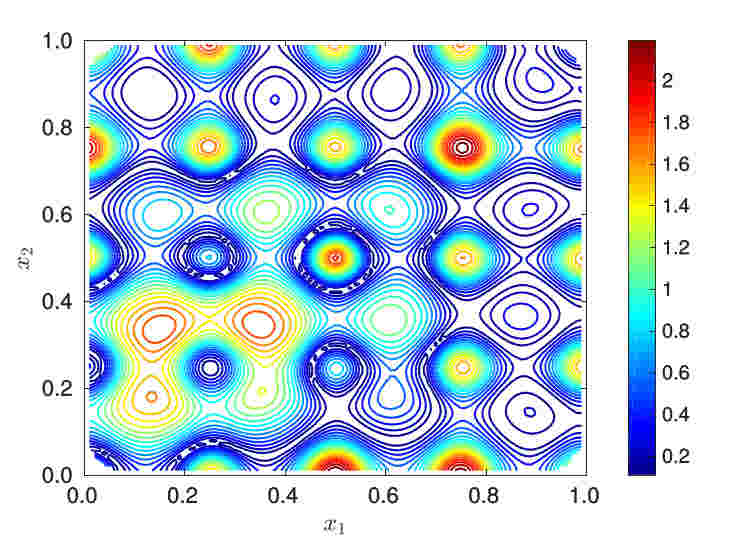} 
\subcaption{Initial absolute error}\label{fig:AckleyInitialError}
\end{subfigure}%
\caption[Initial state Ackley function]{Initial state of $\mathcal{M}_{Ackley}^{2d}$, (a) Location of 20 initial samples created with TPLHD, (b) Initial absolute error over the domain.}\label{fig:AckleyInitialSamples}
\end{figure}

The first two-dimensional test is Ackley's path, which is given by
\begin{equation}
\begin{aligned}
\mathcal{M}_{Ackley}^{2d}(\bm{x}) &= - 20  \exp \left( -0.2  \sqrt{0.5  (x_{1}^2 + x_{2}^2 )} \right) \\ &-  \exp \left( 0.5  \cos(2  \pi  x_{1}) + \cos(2 \pi  x_{2}) \right) + \exp(1) + 20 \, \text{.}
\end{aligned}
\end{equation}
It is defined for $(x_{1}, \,  x_{2}) \in \left[-2,2 \right]^{3}$. The function is plotted in Figure \ref{fig::Ackley}. The shape is multimodal with a global minimum $\mathcal{M}_{Ackley}^{2d}(\bm{x}_{min}) = 0$ with $\bm{x}_{min} = \left[0,0\right]^{T}$. This benchmark was for example employed by \cite{back1993overview}. \\
The location of the initial samples over the function contour is shown in Figure \ref{fig:AckleyInitial}. The resulting absolute error of these initial samples utilizing OK is illustrated in Figure \ref{fig:AckleyInitialError}. 

It can be noted that due to an initial sample close to the global minimum the absolute error around this domain is not as significant as other areas in order to reduce the mean global error. \\
The error measures for the adaptive sampling strategies when starting with 20 samples and iteratively extending the sample size to 40 samples are listed in Table \ref{tab::Ackley40}. 

It can be seen that all of the presented techniques are able to reduce the initial error significantly. 
The best approximation of the shape of the function (when judged by R$^{2}$ and MAE) is given by EIGF. 
\begin{table}[ht!]
\begin{center}
\resizebox{1.0\textwidth}{!}{%
\begin{tabular}{l|l c c c c} \hline
& Method & MAE & RMAE & RMSE & R$^{2}$ \\ \hline\hline \\
\multirow{1}{*}{\shortstack[l]{Errors after\\20 samples}} & TPLHD &  0.6130 &	1.5993 & 0.7819 & 0.6962 \\\\  \hline \\
\multirow{14}{*}{\shortstack[l]{ Errors after\\40 samples}} & TPLHD  &   0.4834 & 1.4352 & 0.6175 & 0.8105  \\
&ACE &   0.4126 & 1.5903 & 0.5622 & 0.8436 \\
&AME &  0.5081 & \textbf{1.2917} & 0.6355 & 0.8002  \\
&CVVor & 0.4035 & 1.5181 & 0.5421 & 0.8546 \\ 
&CVD &  0.4124 & 1.5889 & 0.5526 & 0.8489  \\
&EI & 0.4669 & 1.5788 & 0.6245 & 0.8070  \\
&EIGF &  \textbf{0.3404} & 1.3663 & 0.4653 &\textbf{ 0.8929}\\
&MASA & 0.5291 & 1.4447 & 0.6858 & 0.7673 \\
&MEPE & 0.4729 & 1.4551 & 0.6037 & 0.8197  \\
&MIPT & 0.4752 & 1.3212 & 0.6150 & 0.8129 \\
&MSD & 0.5556 & 1.6283 & 0.7129 & 0.7486 \\
&SFCVT & 0.3869 & 1.5671 & \textbf{0.5342} & 0.8588\\
&SSA &  0.4520 & 1.5941 & 0.6083 & 0.8169
\end{tabular}
}
\end{center}
\caption[Error measures for $\mathcal{M}_{Ackley}^{2d}$ after 40 samples]{Error measures for $\mathcal{M}_{Ackley}^{2d}$ after 40 samples.}\label{tab::Ackley40}
\end{table}

\begin{figure}[h!]
\centering
\includegraphics[scale=0.7]{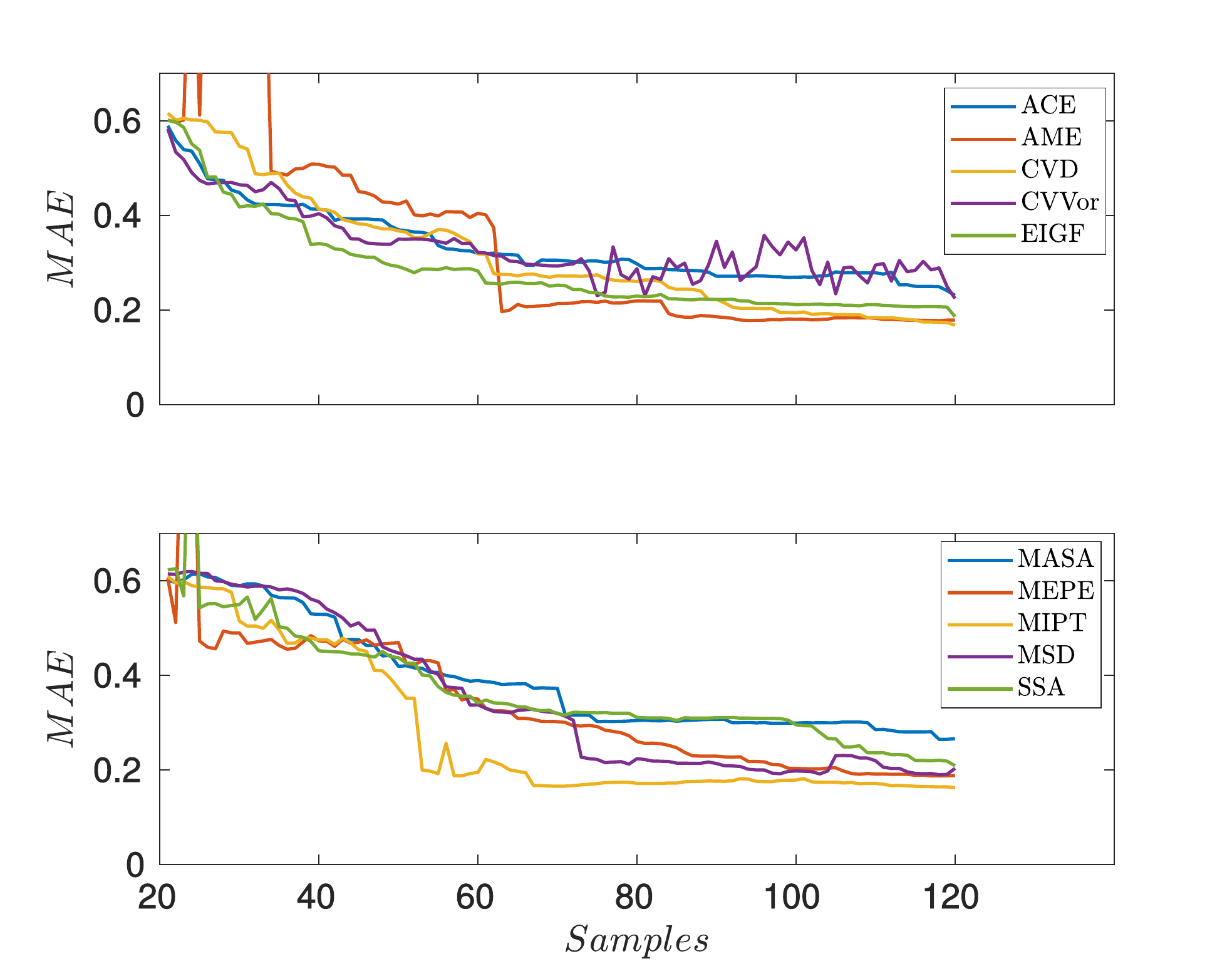} 
\caption[Convergence of MAE error for $\mathcal{M}_{Ackley}^{2d}$]{Average value of MAE error of 10 adaptive sampling techniques for $\mathcal{M}_{Ackley}^{2d}$ over the sample size, until 120 samples.}\label{fig::AckleyConverge}
\end{figure}

\begin{figure}[h!]
\centering
\begin{subfigure}[t]{0.3\textwidth}
\includegraphics[scale=0.28]{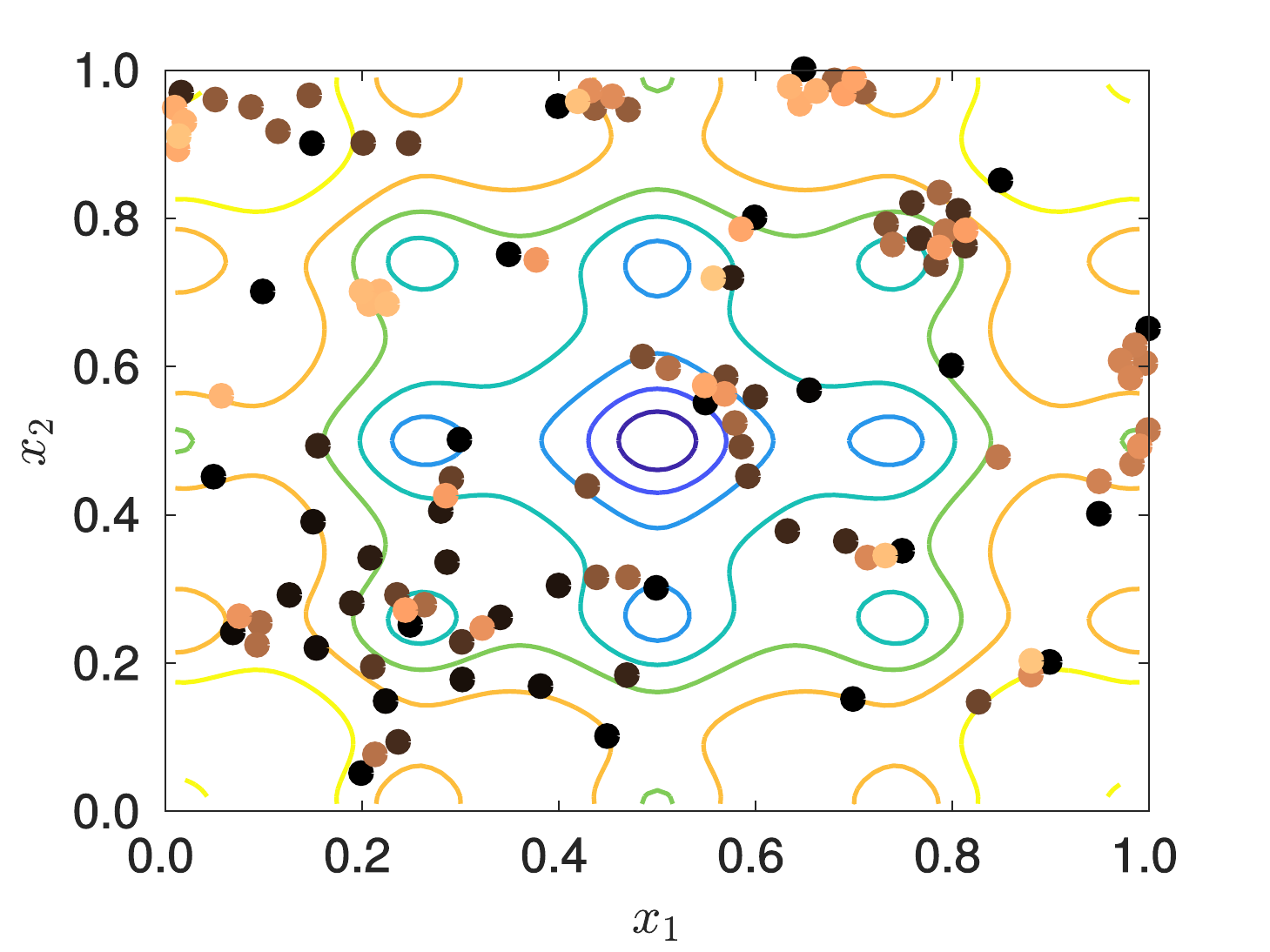}
\subcaption{ACE }\label{fig:AckleyACE}
\end{subfigure}%
\begin{subfigure}[t]{0.3\textwidth}
\includegraphics[scale=0.28]{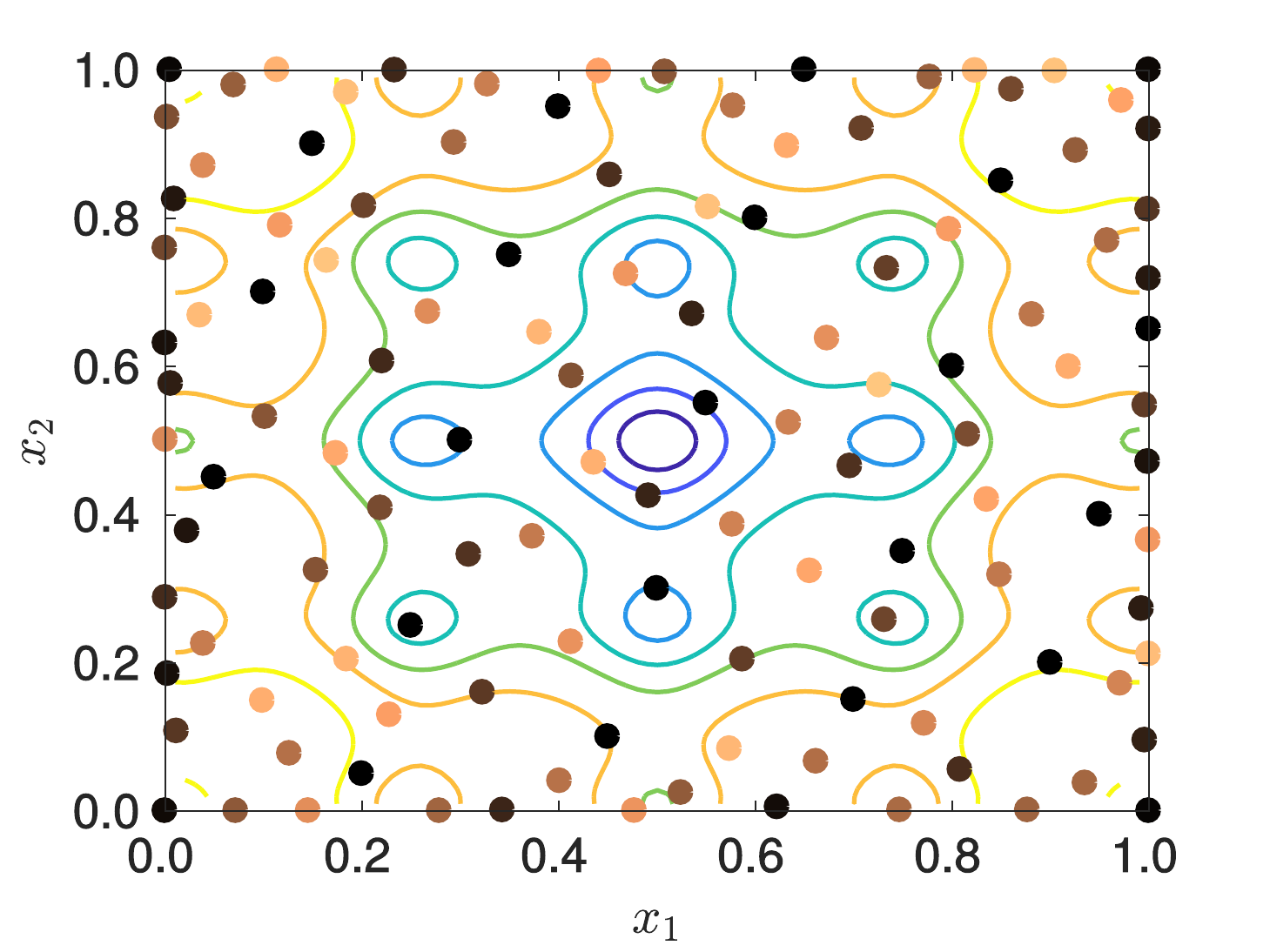}
\subcaption{AME }\label{fig:AckleyAME}
\end{subfigure}
\begin{subfigure}[t]{0.3\textwidth}
\includegraphics[scale=0.28]{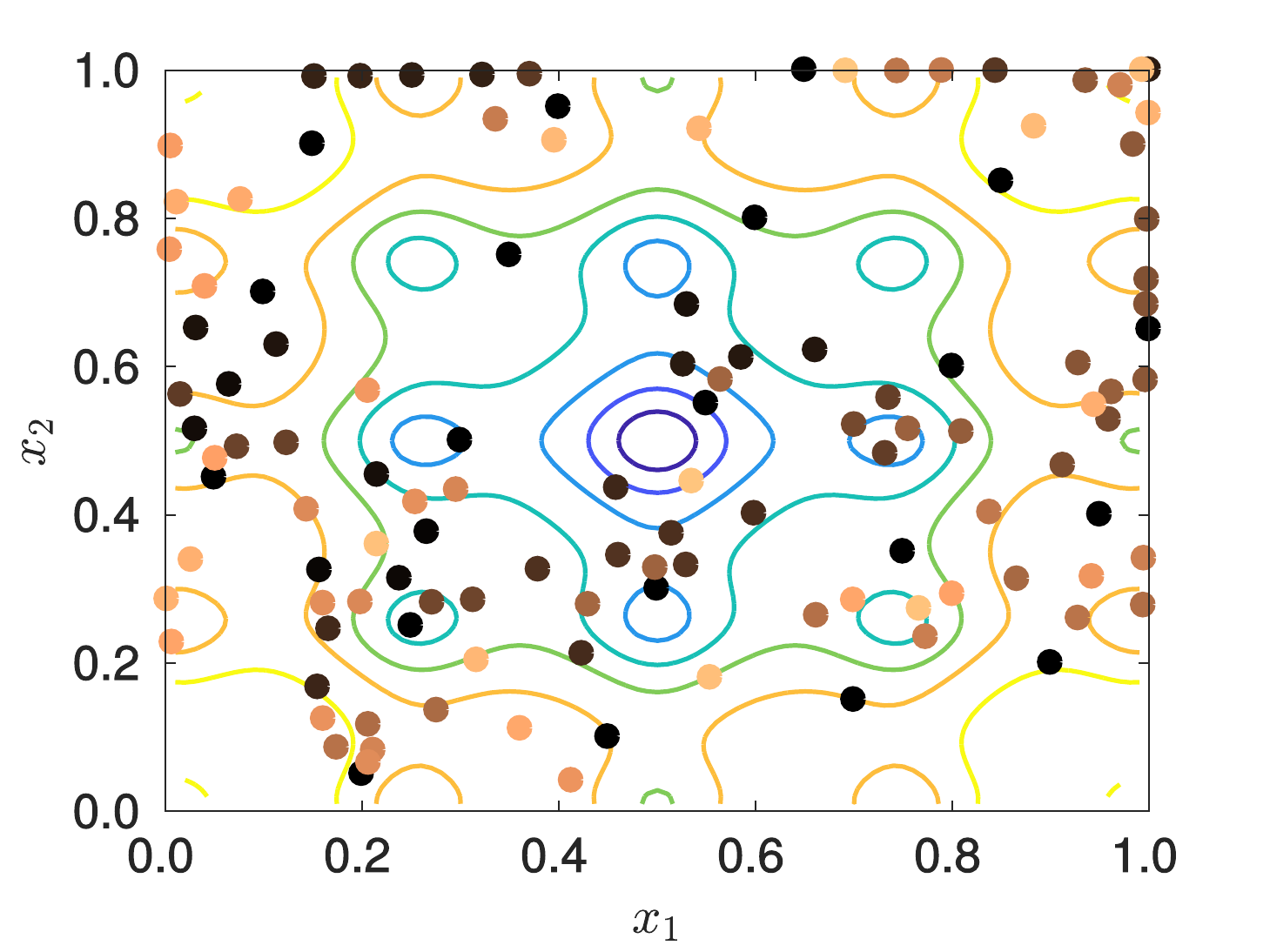}
\subcaption{CVVor }\label{fig:AckleyCVVor}
\end{subfigure}%
\begin{subfigure}[t]{0.3\textwidth}
\includegraphics[scale=0.28]{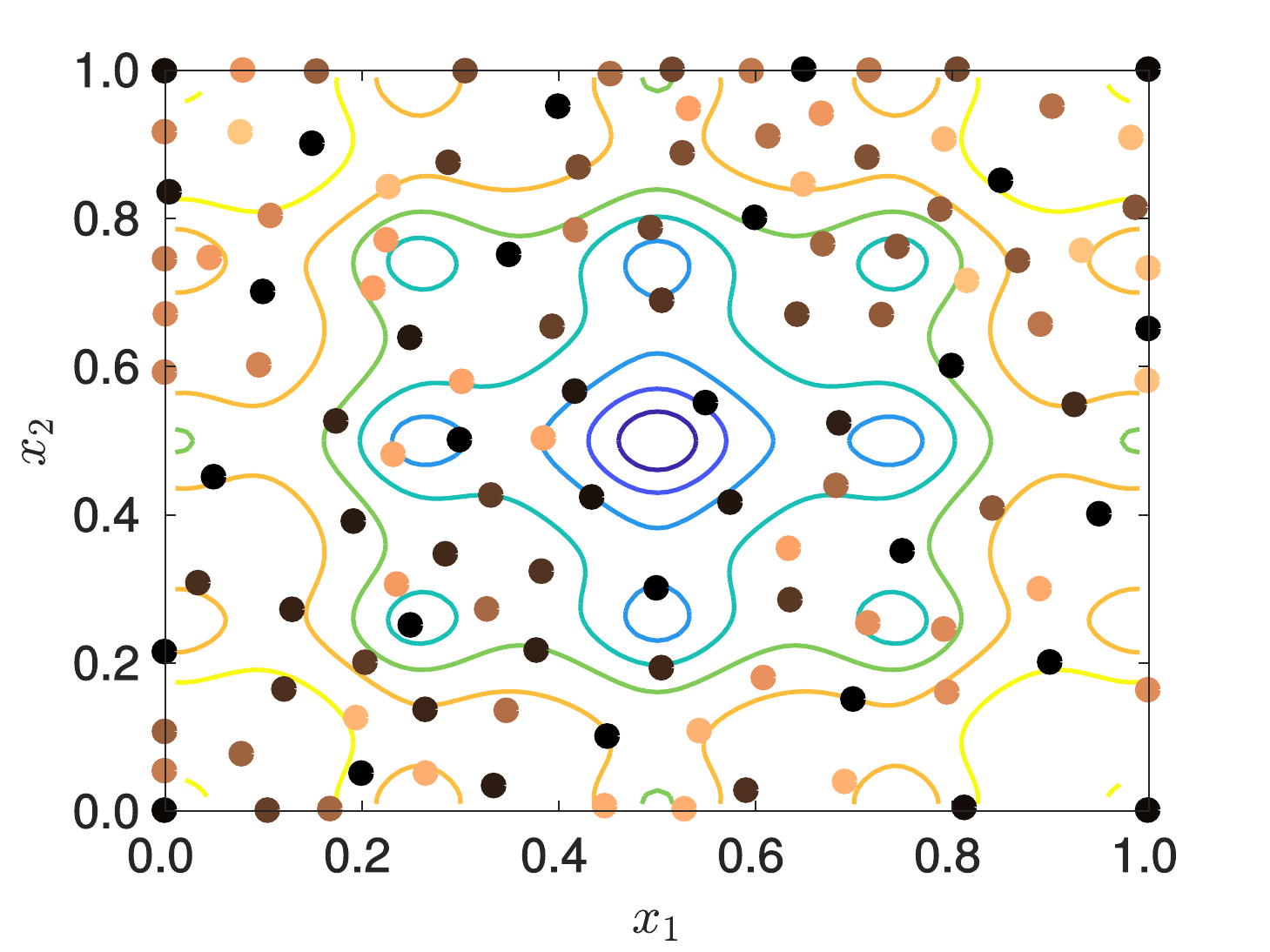}
\subcaption{CVD }\label{fig:AckleyMSE}
\end{subfigure}
\begin{subfigure}[t]{0.3\textwidth}
\includegraphics[scale=0.28]{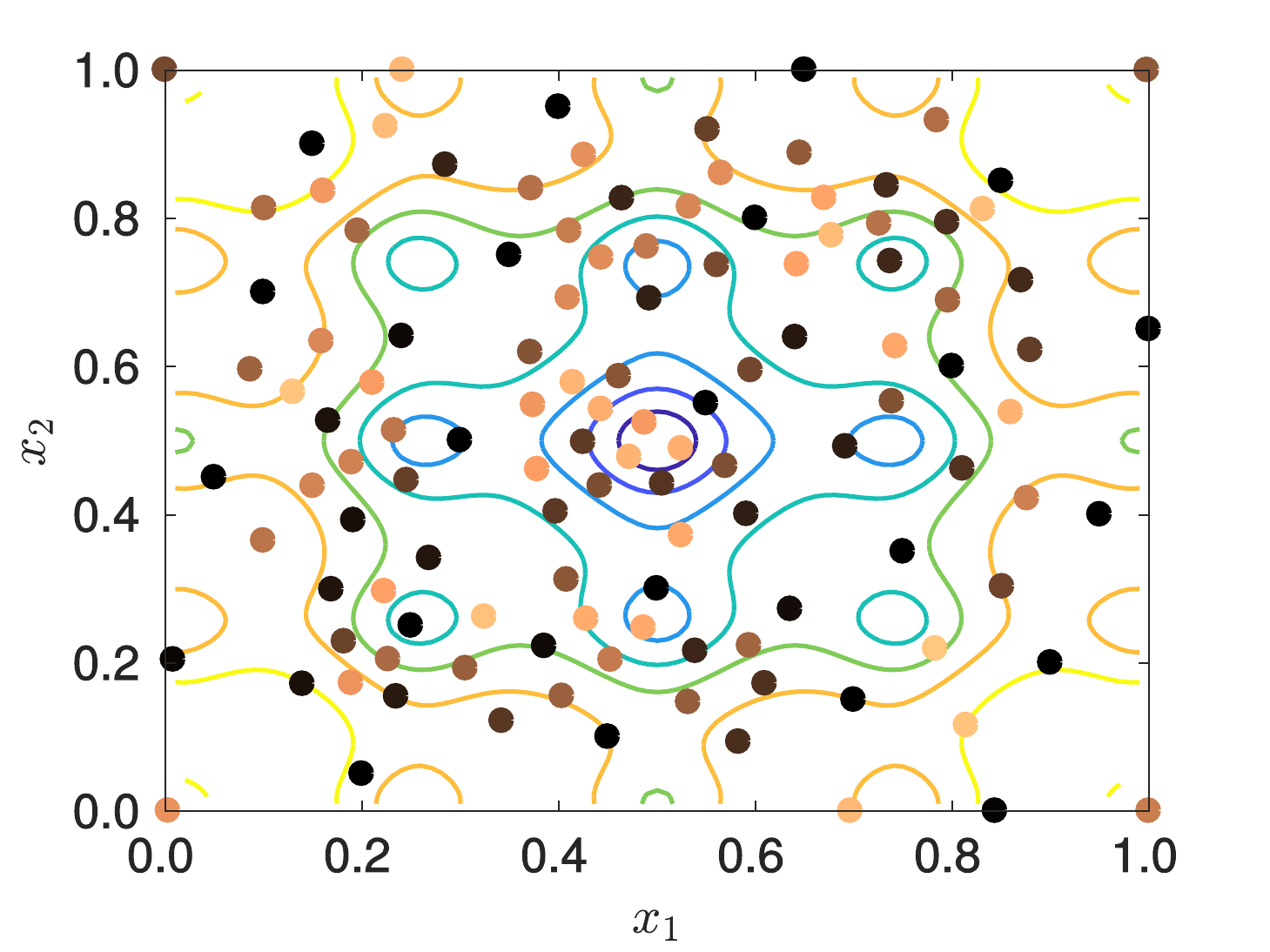}
\subcaption{EIGF}\label{fig:AckleyEIGF}
\end{subfigure}%
\begin{subfigure}[t]{0.3\textwidth}
\includegraphics[scale=0.28]{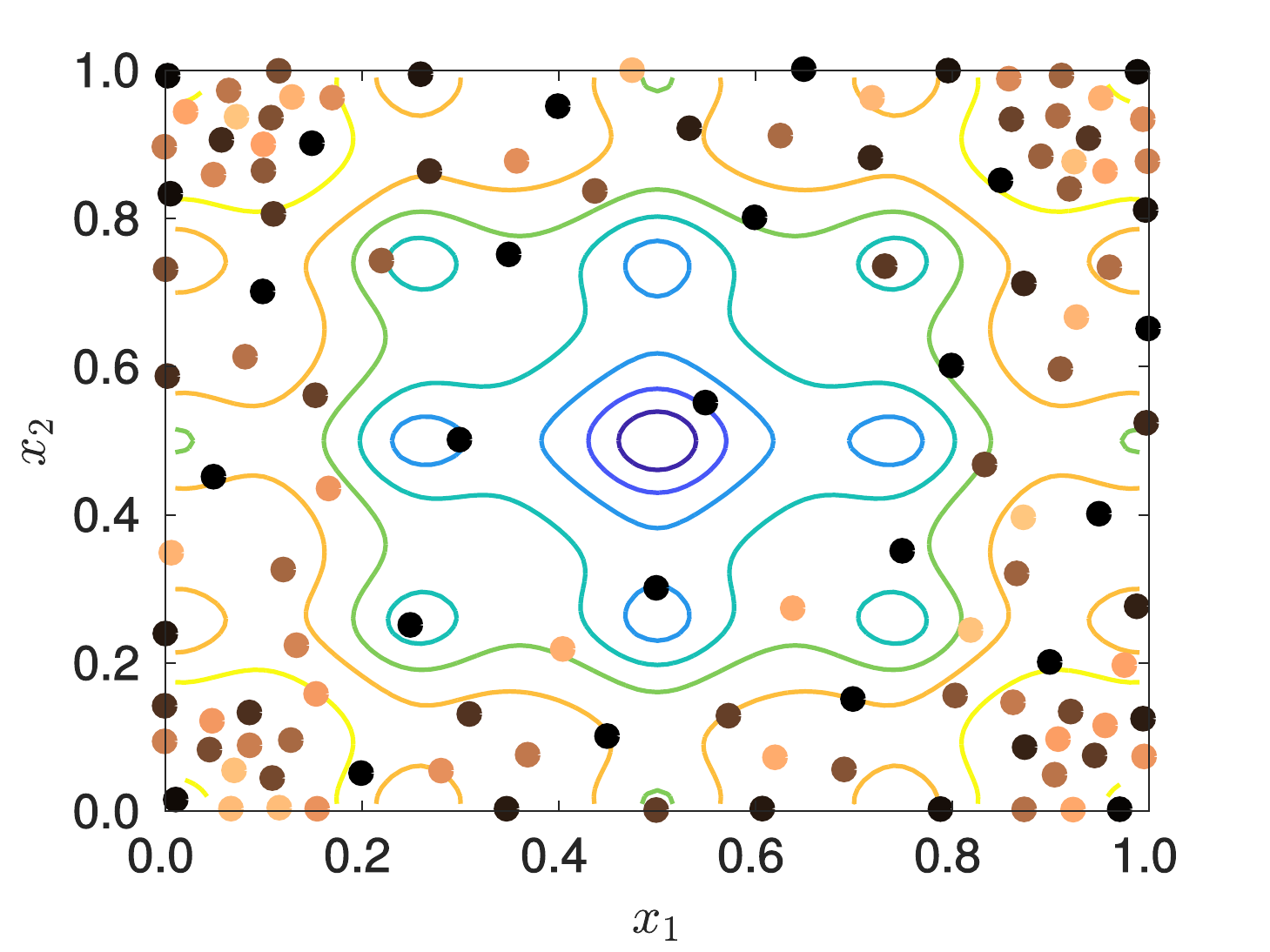}
\subcaption{MASA }\label{fig:AckleyMASA}
\end{subfigure}
\begin{subfigure}[t]{0.3\textwidth}
\includegraphics[scale=0.28]{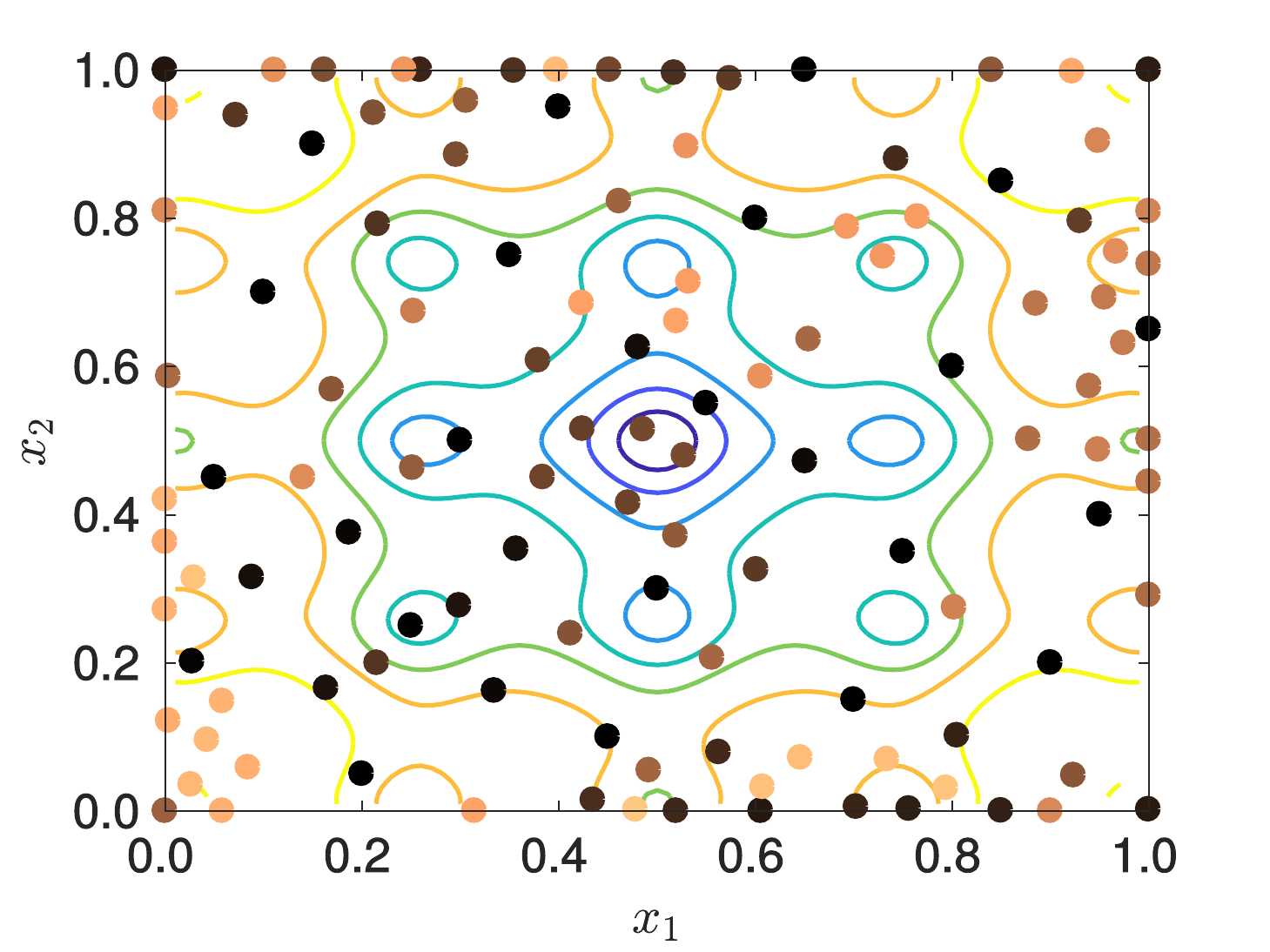}
\subcaption{MEPE }\label{fig:AckleyMEPE}
\end{subfigure}%
\begin{subfigure}[t]{0.3\textwidth}
\includegraphics[scale=0.28]{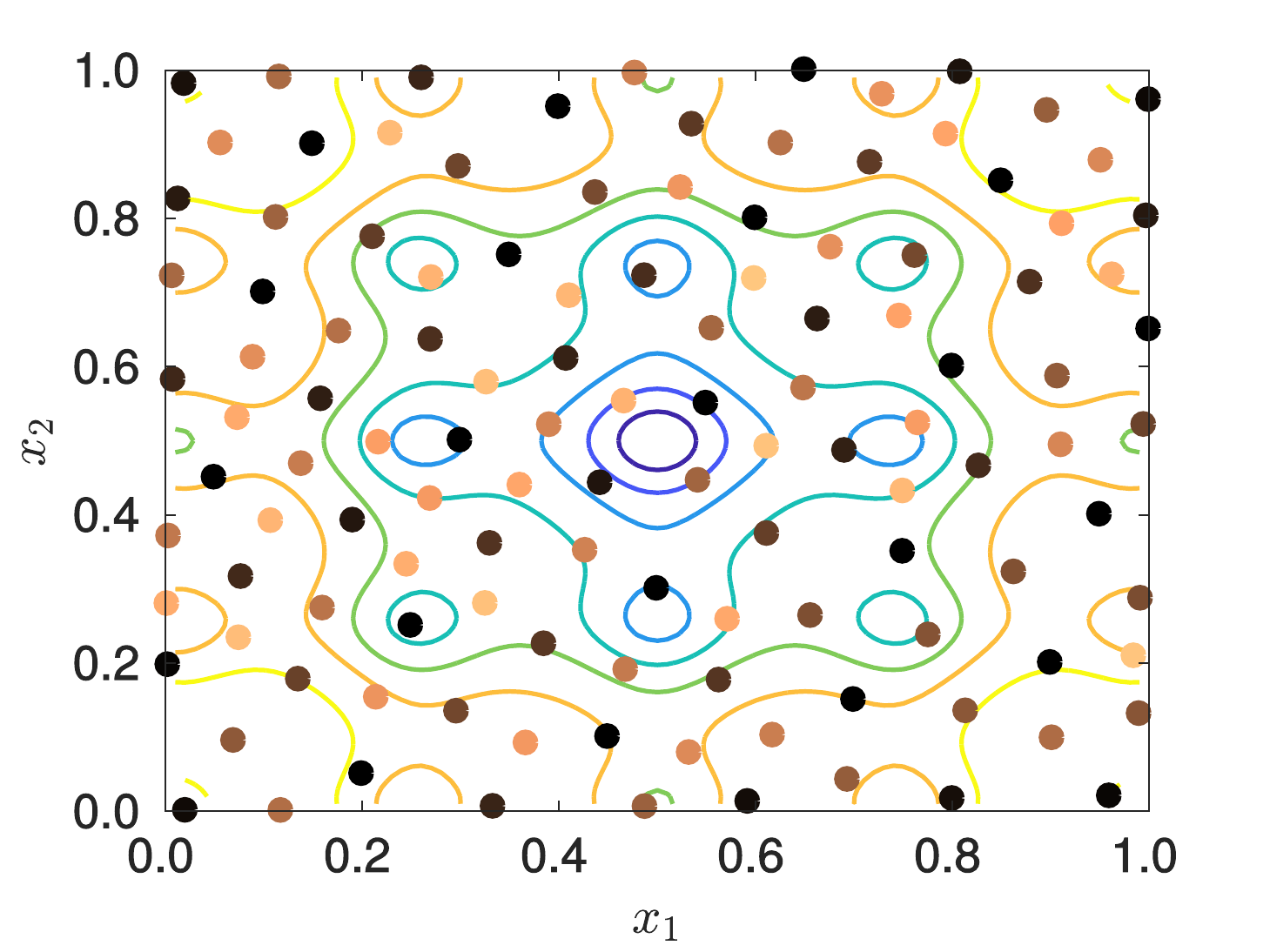}
\subcaption{MIPT }\label{fig:AckleyMIPT}
\end{subfigure}
\begin{subfigure}[t]{0.3\textwidth}
\includegraphics[scale=0.28]{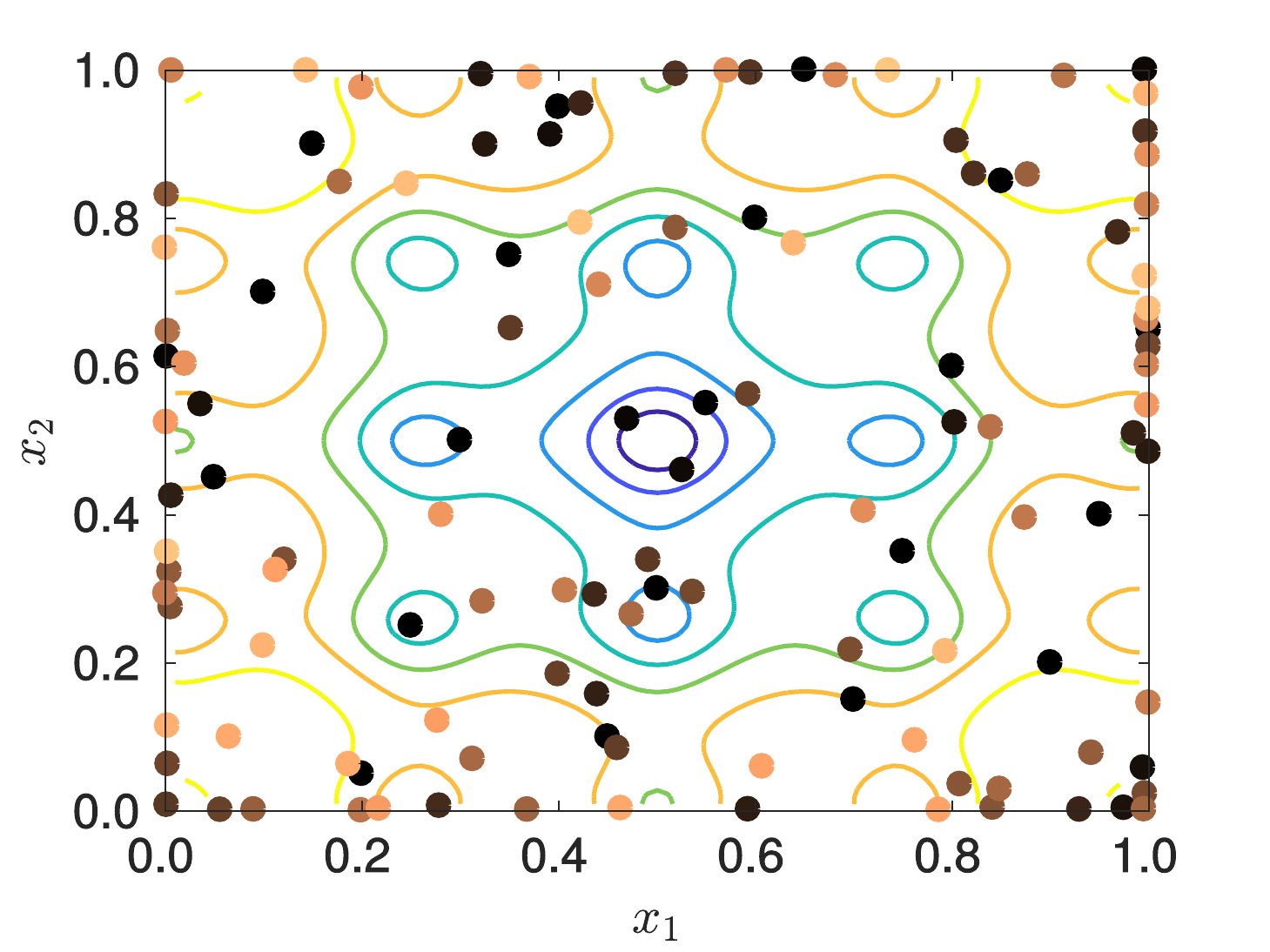}
\subcaption{MSD }\label{fig:AckleyMSD}
\end{subfigure}%
\begin{subfigure}[t]{0.3\textwidth}
\includegraphics[scale=0.28]{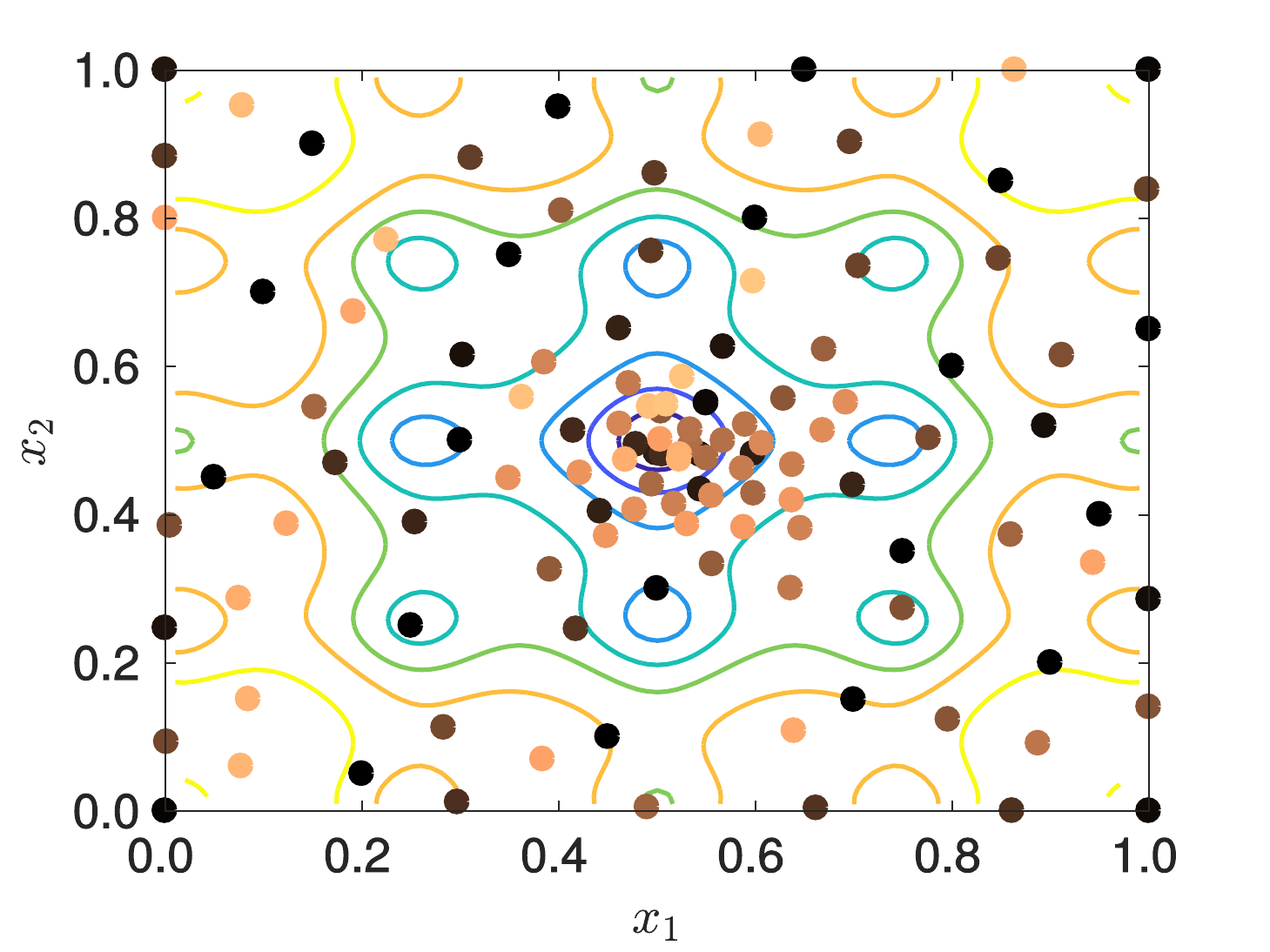}
\subcaption{SSA}\label{fig:AckelySSA}
\end{subfigure}
\caption[Sample positions for Ackley after 120 samples]{Sample positions for the respective adaptive sampling techniques in the normalized input domain after 120 samples for $\mathcal{M}_{Ackley}^{2d}$.}%
\label{fig:AckleyPositions}%
\end{figure}

\newpage
In comparison to a one-shot TPLHD sample generation nine out of the twelve techniques show a more proficient approximation behavior when looking at MAE. Furthermore eight of the methods improve the RMSE error in this respect. 
After 100 samples as shown in Table \ref{tab::100} the results have changed. It should be noted that EI and SFCVT were not able to reach 100 sample points because of clustering issues. The space-filling algorithm MIPT yields on average the best approximation results. Six methods provide a better MAE value then the one-shot TPLHD method and only four a better RMSE measure. 
\begin{table}[h!]
\begin{center}
\resizebox{1.0\textwidth}{!}{%
\begin{tabular}{l|l c c c c} \hline
& Method & MAE & RMAE & RMSE & R$^{2}$ \\ \hline\hline \\
\multirow{1}{*}{\shortstack[l]{Errors after\\20 samples}} & TPLHD &  0.6130 &	1.5993 & 0.7819 & 0.6962 \\\\  \hline \\
\multirow{14}{*}{\shortstack[l]{ Errors after\\40 samples}} & TPLHD  &   0.2319 & 0.8833 & 0.2955 & 0.9608 \\
&ACE &  0.2692 & 0.9398 & 0.3752 & 0.9303\\
&AME &  0.1803 & 0.7206 & 0.2310 & 0.9736 \\
&CVVor &0.2948 & 1.9316 & 0.4286 & 0.9091 \\ 
&CVD &  0.1942 & 1.2137 & 0.2984 & 0.9559  \\
&EI & - & - & - & -  \\
&EIGF &  0.2112 & 1.0552 & 0.3245 & 0.9479\\
&MASA &   0.2991 & 1.3664 & 0.4339 & 0.9068 \\
&MEPE & 0.2033 & 0.8973 & 0.2836 & 0.9602 \\
&MIPT & \textbf{0.1788} & \textbf{0.6755} &\textbf{ 0.2275} & \textbf{0.9743}\\
&MSD &  0.1979 & 0.8698 & 0.2646 & 0.9653 \\
&SFCVT & - & - & - & - \\
&SSA &   0.2955 & 1.3728 & 0.4426 & 0.9030 
\end{tabular}
}
\end{center}
\caption[Error measures for $\mathcal{M}_{Ackley}^{2d}$ after 100 samples]{Error measures for $\mathcal{M}_{Ackley}^{2d}$ after 100 samples (methods with clustering problems are indicated by empty rows).}\label{tab::100}
\end{table}
The convergence of the average MAE error with an increasing sample size starting from 20 and ending at 120 samples is illustrated in Figure \ref{fig::AckleyConverge}. It can be seen that AME shows high variation for smaller sample sizes but is able to finish at a comparatively low MAE value. Most of the methods indicate a similar behavior by decreasing the error measure quickly in the beginning and stalling more at higher sample sizes. MIPT reaches the lowest value.
The positions of the generated samples after 100 added points (with the initial 20) are shown for the respective adaptive sampling techniques in Figure \ref{fig:AckleyPositions}. \\
The sample positions for the ACE method are depicted in Figure \ref{fig:AckleyACE}. It can be noticed that the method tends to generate samples that cluster in certain areas e.g. around ($0.0,1.0$) or ($0.8,0.8$). The exploration component of the method is not pronounced enough. \\This can be noticed by the lack of samples around three of the four corners of the domain. \\
The positions for AME are illustrated in Figure \ref{fig:AckleyAME}. Here the focus lies on exploration. The samples are space-filling and all edges and corners are sufficiently samples. \\
CVVor (Figure \ref{fig:AckleyCVVor}) does not sample in three of the four corners. Furthermore local optima (e.g. around ($0.3,0.8$)) are not sampled. This leads to an overall worse approximation result than other methods like CVD (Figure \ref{fig:AckleyMSE}). CVD creates samples around all local minima. Furthermore the edges and corners are due to the exploration component sufficiently covered. \\
The points for EIGF are displayed in Figure \ref{fig:AckleyEIGF}. The method shows a balance between exploration and exploitation, all corners are sampled but a focus is put onto the center of the domain where the local and global optima are located. \\
MASA (Figure \ref{fig:AckleyMASA}) shows exploitation around the maximum absolute values and does not succeed to sample around the minima in the center. Hence, relatively bad results are obtained.  \\
MEPE has both exploration and exploitation components as seen in Figure \ref{fig:AckleyMEPE}. The edges as well as all local minima are thoroughly sampled. \\
MIPT (Figure \ref{fig:AckleyMIPT}) yields the best MAE error over 120 samples. \\This is due to the space-filling behavior which in this application samples all optima as well as the edges.
The 120 points for MSD are displayed in Figure \ref{fig:AckleyMSD}. It can be seen that the exploitation component is not sufficient enough. The samples are mostly created on the edges. \\
SSA, as depicted in Figure \ref{fig:AckelySSA}, explores the domain edges and corners however it starts to cluster around the global minimum and neglects to catch the overall form of the function while doing so.
\clearpage
\subsubsection{Six-Hump camel}\label{sec::SixHumb}
\begin{figure}[h!]
\centering
\begin{tikzpicture}
\begin{axis}
[
xlabel={$x_{1}$},
ylabel={$x_{2}$},
zlabel={$\mathcal{M}(x_{1},x_{2})$},
view={85}{45}
]
\addplot3[surf,domain=-2.0:2.0,domain y=-2:2]
{(4 - 2.1*x^(2) + (1/3)*x^(4))*x^(2) + x* y + ( - 4 + 4*y^(2) )*y^(2)};
\end{axis}
\end{tikzpicture}
\caption[Six-hump camel function]{Plot of six-hump camel function $\mathcal{M}_{SHC}^{2d}$ }\label{fig::sixHumb}
\end{figure}
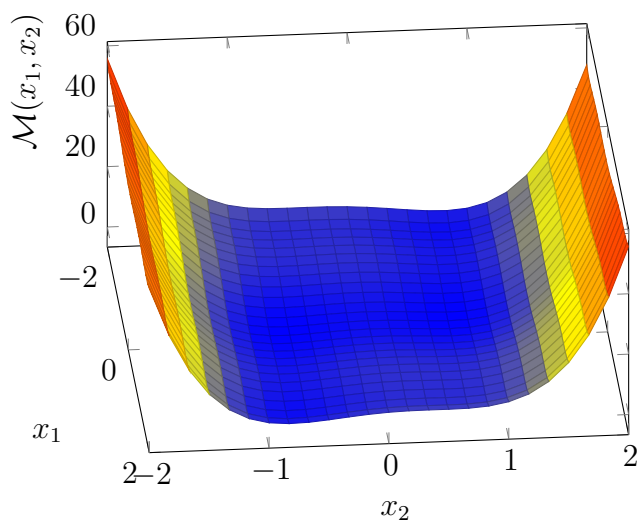

The second two-dimensional benchmark is the six-hump camel function. For usage as a surrogate model benchmark see e.g. \cite{laurent2017overview}. The function reads as
\begin{equation}
\mathcal{M}_{SHC}^{2d}(\bm{x}) = (4 - 2 x_{1}^{2} + \frac{x_{1}^{4}}{3}) x_{1}^{2} + x_{1} x_{2} + \left( - 4 + 4 x_{2}^{2} \right)x_{2}^{2} \, \text{.}
\end{equation}
Let the domain be defined as $(x_{1}, \,  x_{2}) \in \left[-2,2 \right]^{2}$. The function is valley-shaped as shown in Figure \ref{fig::sixHumb}. In the given domain the function has two global minima with $\mathcal{M}_{SHC}^{2d}(\bm{x}_{min}) = -1.0316$ with $\bm{x}_{min} = \left[ 0.0898, -0.7126  \right]^{T}$ and secondly with $\bm{x}_{min} = \left[ -0.0898, 0.7126  \right]^{T}$. The maximum value of the function is around 50 on the edges of the parametric space. Hence the spread of the function is around 51. 

\begin{figure}[h!]
\begin{subfigure}[t]{0.5\textwidth}
\includegraphics[scale=0.35]{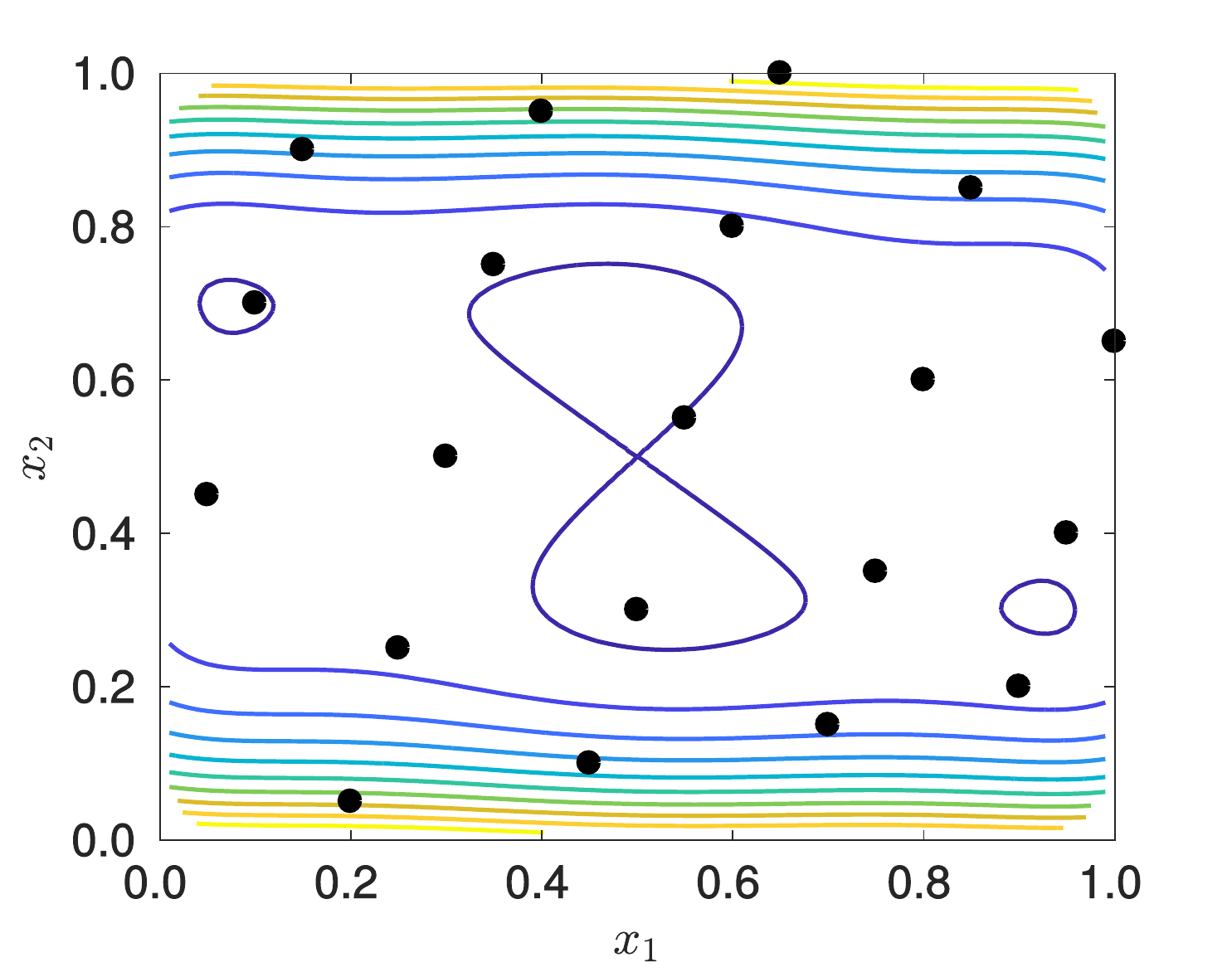}
\subcaption{Initial samples}\label{fig:SixHumbInitialPlot}
\end{subfigure}%
\begin{subfigure}[t]{0.5\textwidth}
\includegraphics[scale=0.35]{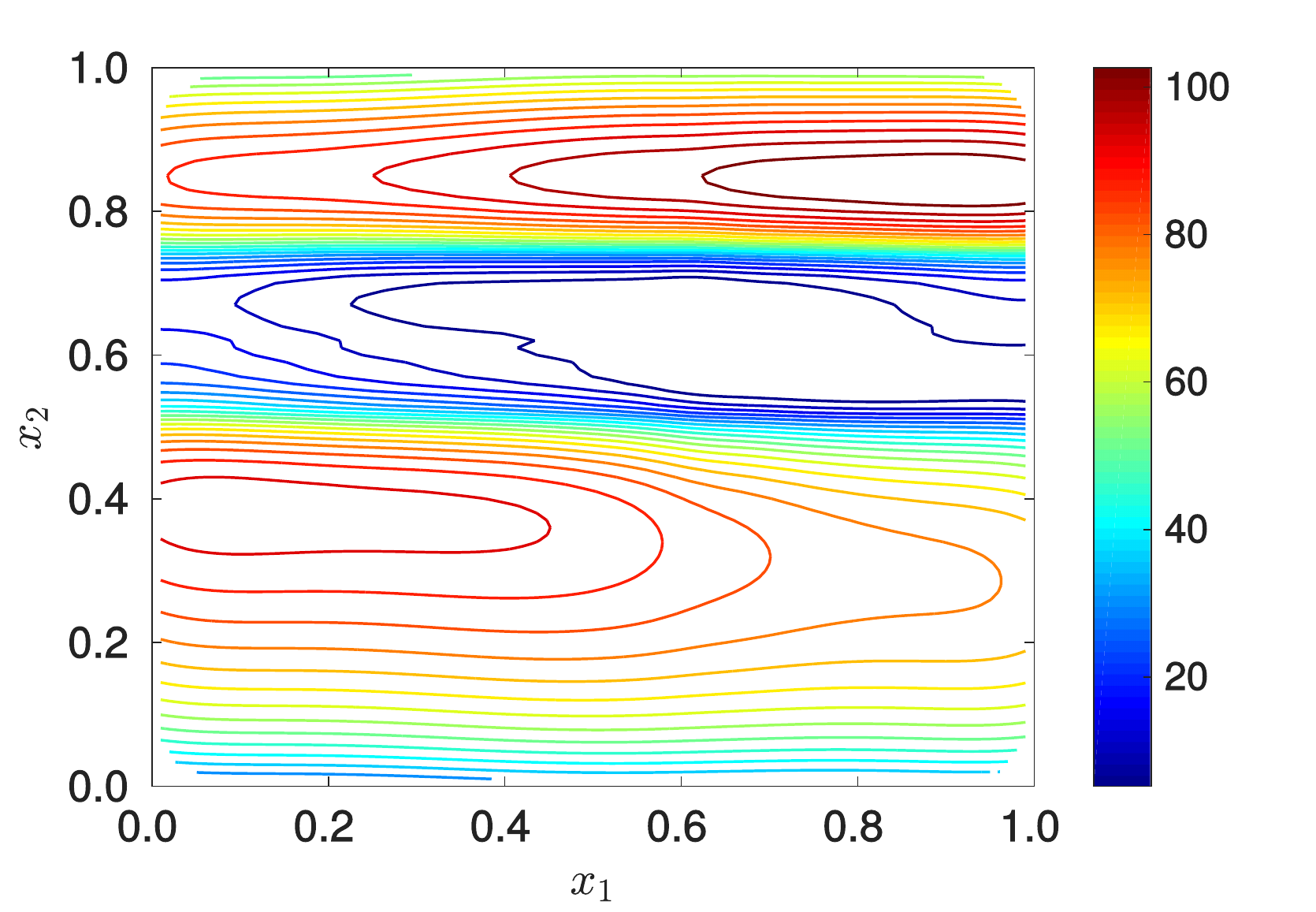} 
\subcaption{Absolute error}\label{fig:SixHumbInitialMAE}
\end{subfigure}
\caption[Initial state six-humb camel function]{Initial state of $\mathcal{M}_{SHC}^{2d}$, (a) Location of 20 initial samples created with TPLHD, (b) Initial absolute error over the domain.}\label{fig::SixHumbInitial}
\end{figure}

The location of the initial sample points as well as a contour plot of the Six-hump camel function are shown in Figure \ref{fig:SixHumbInitialPlot}. It can be seen that the 20 TPLHD samples spread the domain evenly. The quality of the initial metamodel is displayed in Figure \ref{fig:SixHumbInitialMAE} in which the contour level of the absolute error in the domain is plotted.
A proficient adaptive sampling technique needs to generate samples in the following areas:
\begin{itemize}
\item There are no samples around the edges of the domain. Here, the Kriging models needs to extrapolate from the given samples. A task for the adaptive sampling techniques is to sample in these areas.
\item There is already a sample point in the upper left global minimum. The other minimum value is unsampled. 
\item The curvature in the center of domain is not well represented due to lack of samples.
\end{itemize} 
The respective techniques will be judged by these criteria as well as the error measures.

The error measures for the adaptive sampling strategies when starting with 20 samples and iteratively extending this to 40 are listed in table \ref{tab::SixHumb}. 
It can be seen that all adaptive sampling techniques were able to reduce the error significantly compared to the initial metamodel. Judging by the RMSE error, which can give hints to the shape approximation, most of the adaptive methods do worse then general TPLHD at 40 samples. In this category only 4 algorithms perform well. The relative maximum absolute error however was better for 10 out of the 13 compared techniques, than the TPLHD counterpart. MEPE is the best technique for this application. It is able to reduce the error far more proficient and has the best values in all error measurements.
\begin{table}[h!]
\begin{center}
\resizebox{1.0\textwidth}{!}{%
\begin{tabular}{l|l c c c c} \hline
& Method & MAE & RMAE & RMSE & R$^{2}$ \\ \hline\hline \\
\multirow{1}{*}{\shortstack[l]{Errors after\\20 samples}} & TPLHD  &  0.7861 & 0.6409 & 1.3227 & 0.9890  \\ \\  \hline \\
\multirow{14}{*}{\shortstack[l]{ Errors after\\40 samples}} & TPLHD  &  0.2264 & 0.1553 & 0.3437 & 0.9992 \\
&ACE &  0.2165 & 0.1182 & 0.2958 & 0.9994 \\
&AME &  0.4472 & 0.1366 & 0.5630 & 0.9980 \\ 
&CVVor &  0.2796 & 0.4348 & 0.5518 & 0.9981 \\
&EI &  0.4744 & 0.5376 & 0.9395 & 0.9945  \\
&EIGF &  0.2922 & 0.1137 & 0.3823 & 0.9991 \\ 
&LOLA &  0.2833 & 0.5586 & 0.7026 & 0.9969 \\
&MASA & 0.3071 & 0.1116 & 0.3918 & 0.9990  \\
&MEPE & \textbf{0.1284} & \textbf{0.0464} & \textbf{0.1636}& \textbf{0.9998 } \\
&MIPT &  0.2949 & 0.1308 & 0.3849 & 0.9990   \\
&MSD & 0.2309 & 0.1013 & 0.3200 & 0.9993  \\
&MSE &  0.3458 & 0.1464 & 0.4528 & 0.9987 \\
&SFCVT &  0.2533 & 0.1205 & 0.3459 & 0.9992   \\
&SSA & 0.2655 & 0.1113 & 0.3421 & 0.9992
\end{tabular}
}
\end{center}
\caption[Error measures for $\mathcal{M}_{SHC}^{2d}$ after 20 samples.]{Error measures for $\mathcal{M}_{SHC}^{2d}$ after 20 samples. }\label{tab::SixHumb}
\end{table}

\begin{figure}[htbp]
\centering
\includegraphics[scale=0.5]{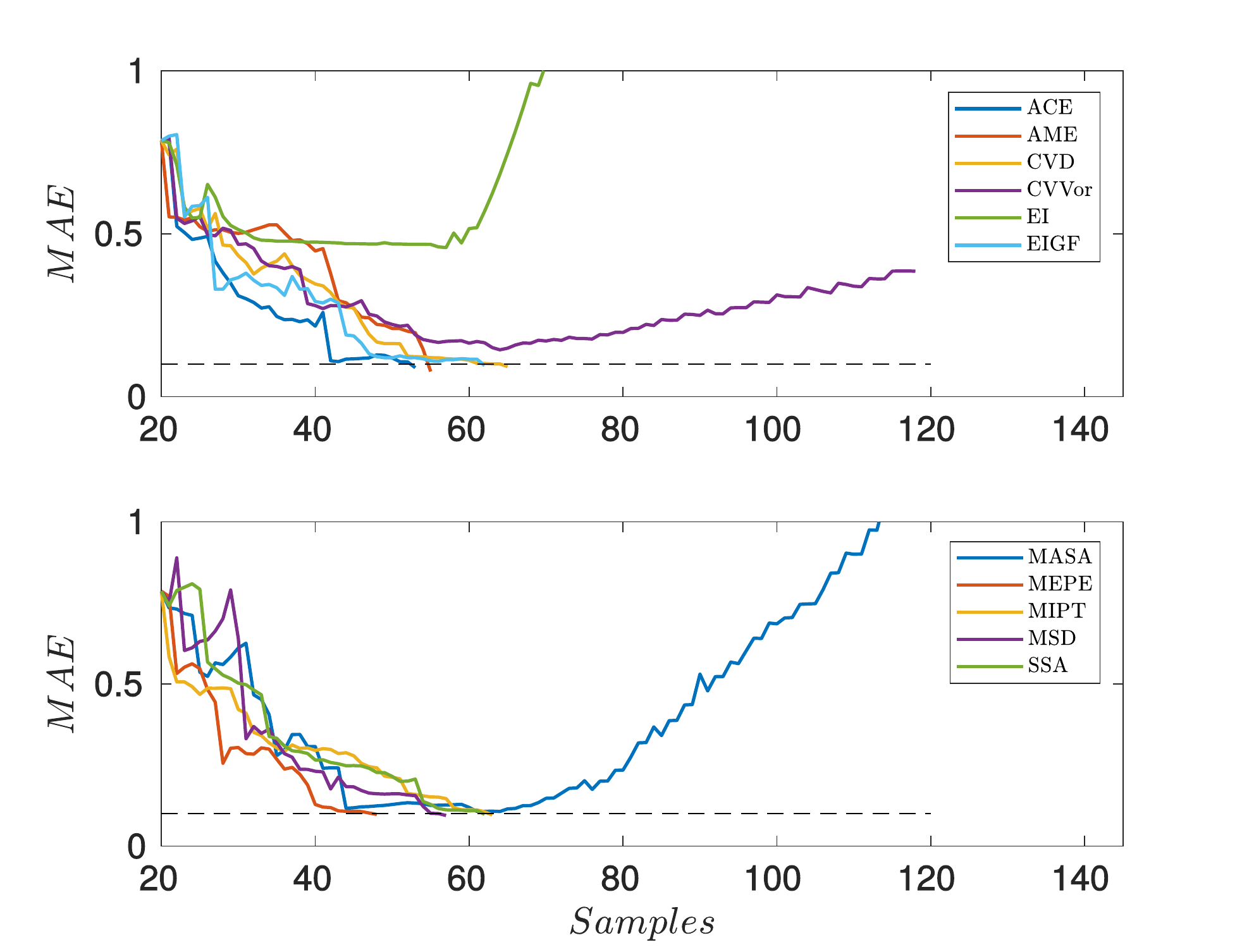}
\caption[Convergence of MAE error for $\mathcal{M}_{SHC}^{2d}$]{Convergence of the MAE value for $\mathcal{M}_{SHC}^{2d}$ for different adaptive sampling techniques.}\label{fig::SixHumbAdapConv}
\end{figure}

\begin{figure}[htbp]
\centering
\includegraphics[scale=0.6]{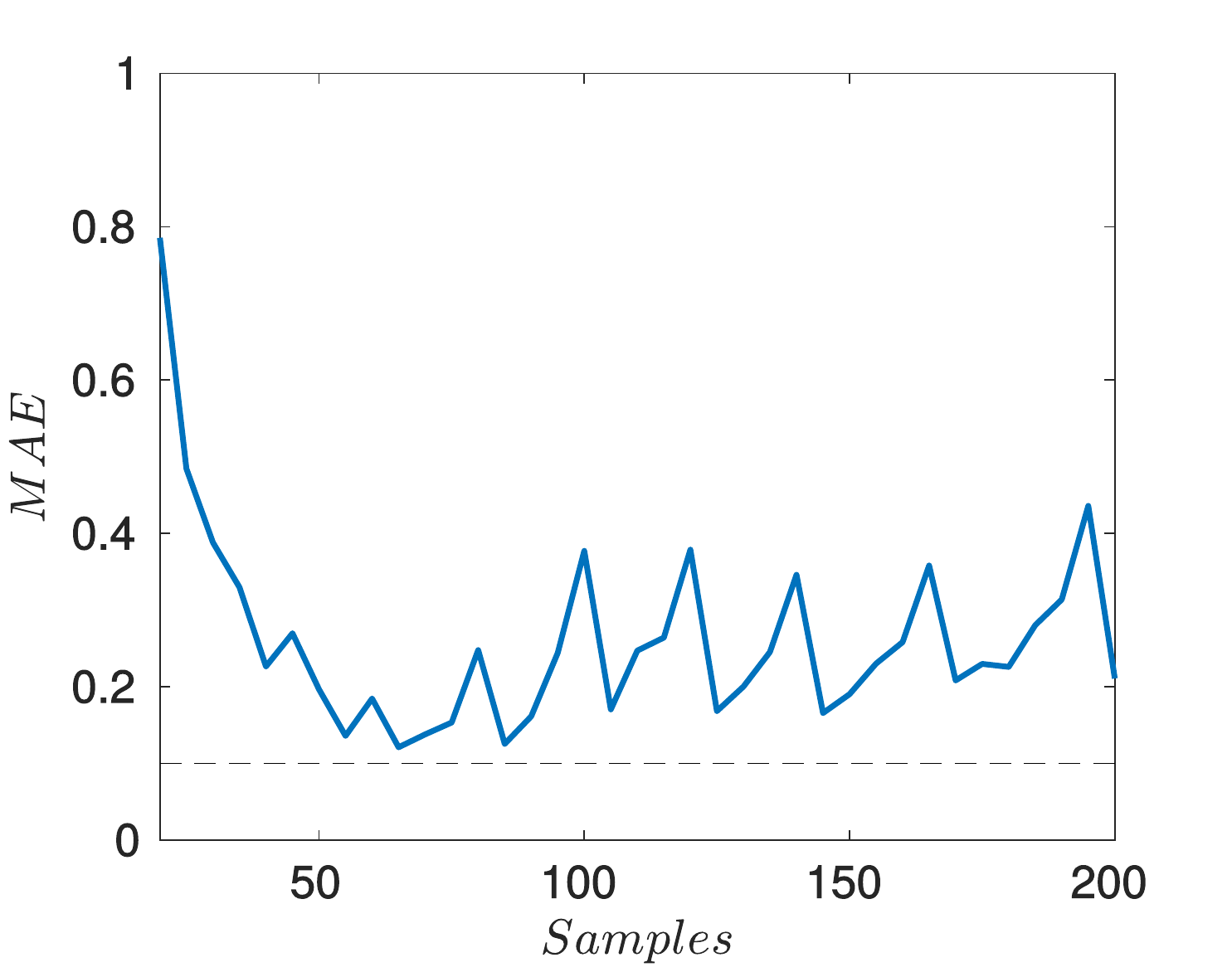}
\caption[Convergence of MAE error for $\mathcal{M}_{SHC}^{2d}$ with TPLHD samples.]{Convergence of MAE for $\mathcal{M}_{SHC}^{2d}$ with samples created by TPLHD.}\label{fig::SixHumbTPLHDConv}
\end{figure}

\begin{table}[t!]
\begin{center}
\resizebox{0.5\linewidth}{!}{%
\begin{tabular}{c c } \hline
\shortstack[l]{Sampling\\method} & \shortstack[l]{Average number\\of Samples}\\ \hline\hline \\
ACE &  53  $\pm$ 1 \\
AME & 55 $\pm$ 0  \\ 
CVD & 57   $\pm$ 1 \\
CVVor &  - \\ 
EI  &  - \\
EIGF & 62 $\pm$ 0  \\
MASA &  -  \\
MEPE & 48  $\pm$ 0 \\
MIPT &  63 $\pm$ 0 \\
MSD & 57 $\pm$ 0  \\
SFCVT &  - \\
SSA &  62  $\pm$ 1  \\ 
\end{tabular}
}
\end{center}
\caption[Average amount of samples required before MAE$<0.1$ for $\mathcal{M}_{SHC}^{2d}$]{Average amount of samples required before MAE$<0.1$ for $\mathcal{M}_{SHC}^{2d}$. The variation of the number of samples over 10 computations is given as an additional value. Methods that do not reach the target value because of clustering or because they need more than 120 samples were omitted.}\label{tab::SixHumbConversionValues}
\end{table}
In a next step the convergence of the techniques is studied. The aim is to reduce the MAE error below 0.1, which is considered to be a satisfactory approximation of the target function. 
To avoid the randomness in the optimization methods, 10 iterations were computed and their performance averaged. 
The number of samples needed to achieve this threshold for the respective techniques are listed in Table \ref{tab::SixHumbConversionValues}. The cutoff was set to 120 samples. The number behind the $\pm$-symbol indicate the variation of the average number. It can be seen that 8 out of the 13 methods were able to reach the target. MEPE is again the best method in this study with an average of only 48 samples needed. \\
 The purely exploration-based approach MIPT requires the most samples with 63. 
SFCVT, EI, CVVor and MASA have problems with clustering. 
The variation of the numbers are low for all studied methods which validates the implementation. 
The convergence towards the threshold is displayed in Figure \ref{fig::SixHumbAdapConv}. The methods with clustering problems run into numerical issues and therefore the error measure increases. The other methods show a constant decrease.
As a comparison consider the converged obtained using TPLHD (see Figure \ref{fig::SixHumbTPLHDConv}). Here, every 5 samples steps a TPLHD metamodel is created and the resulting MAE error is plotted here over the respective number of samples. The target threshold of $0.1$ is highlighted with a dotted line. It can be seen that the plot is jittery for a sample size over 50. This is due to the TPLHD sampling which shifts the samples in the parametric domain. However the MAE target value is not reached.
This result shows the difference between a space-filling method and the adaptive sampling methods. Space-filling methods require more sample points when trying to reduce the approximation error to a very low target.
Here, the adaptive schemes perform better when they are not accompanied by numerical problems. To highlight this consider the data listed in Table \ref{tab::SixHumbTable60}. Recall that after 40 samples TPLHD was performing better or equal to basically all adaptive sampling methods except SSA. \\ After 60 samples this notion is completely changed. Here, TPLHD is by far the worst performing method. 
\begin{table}[htbp]
\begin{center}
\resizebox{1.0\textwidth}{!}{%
\begin{tabular}{l|l c c c c} \hline
& Method & MAE & RMAE & RMSE & R$^{2}$ \\ \hline\hline \\
\multirow{1}{*}{\shortstack[l]{Errors after\\20 samples}} & TPLHD  &  0.7861 & 0.6409 & 1.3227 & 0.9890  \\ \\  \hline \\
\multirow{14}{*}{\shortstack[l]{ Errors after\\60 samples}} &TPLHD & 0.1846 & 0.1698	& 0.3555 & 0.9992 \\
&ACE &  \textbf{0.0695} & \textbf{0.0560} & 0.0999 & 0.9999 \\
&AME & 0.0669 & 0.0341 & \textbf{0.0911} & 0.9999 \\ 
&CVD &  0.1117 & 0.0670 & 0.1583 & 0.9998 \\
&EIGF &  0.1153 & 0.0444 & 0.1581 & 0.9998 \\ 
&MEPE & 0.0772 & 0.0285 & 0.0997 & 0.9999 \\
&MIPT &  0.1123 & 0.0621 & 0.1580 & 0.9998   \\
&MSD & 0.0900 & 0.0503 & 0.1404 & 0.9998  \\
&SSA & 0.1087 & 0.0602 & 0.1494 & 0.9998
\end{tabular}
}
\end{center}
\caption[Error measures for $\mathcal{M}_{SHC}^{2d}$ after 60 samples.]{Error measures for $\mathcal{M}_{SHC}^{2d}$ after 60 samples. }\label{tab::SixHumbTable60}
\end{table}

This proves once again the effectiveness of adaptive schemes.
The next step studies the positions of the generated samples for the respective techniques to reach an $MAE$ value of below $0.1$.
\begin{figure}[h!]
\centering
\begin{subfigure}[t]{0.5\textwidth}
\includegraphics[scale=0.37]{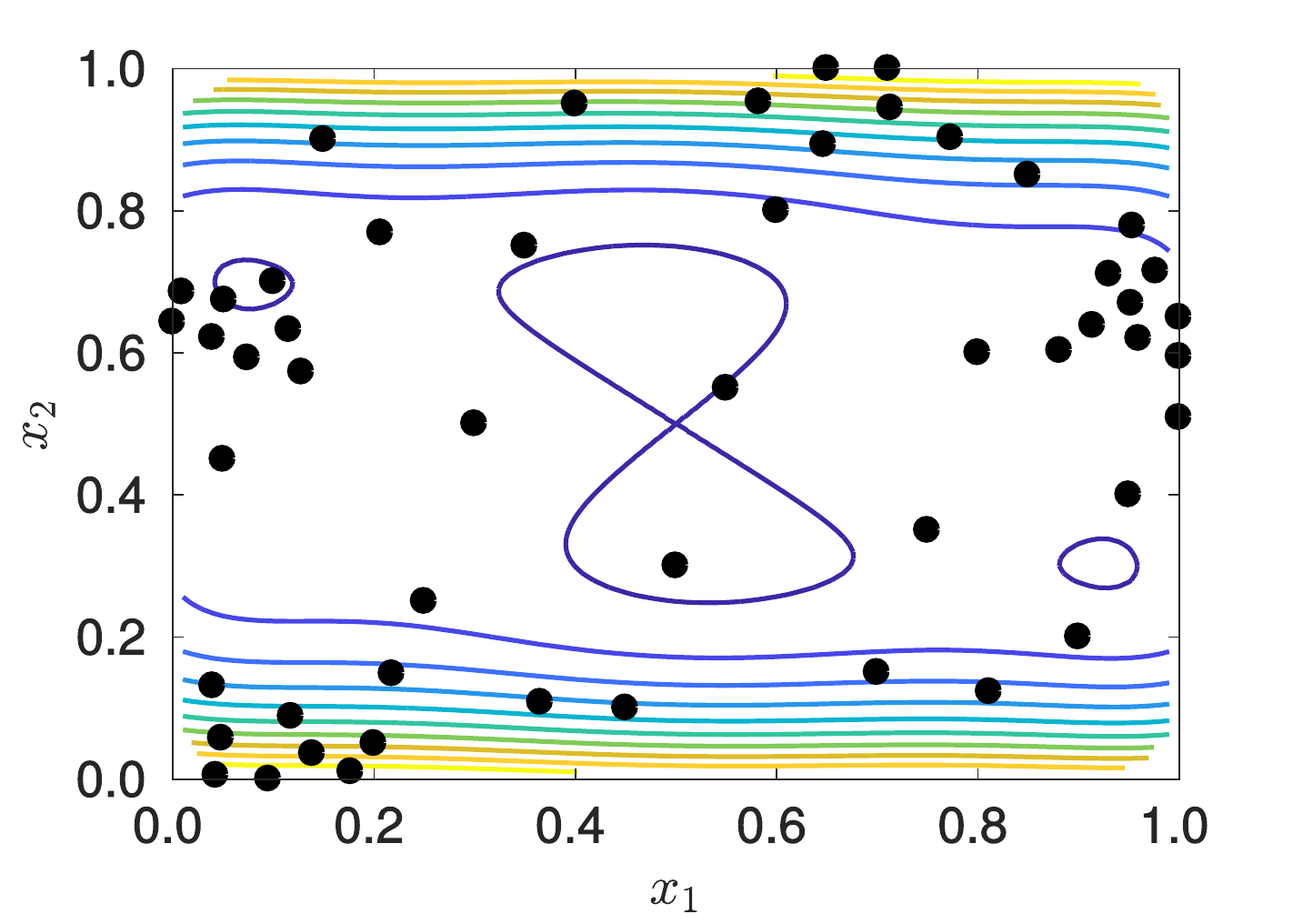}
\subcaption{ACE - 53 samples}\label{fig:SixHumbYesACE}
\end{subfigure}%
\begin{subfigure}[t]{0.5\textwidth}
\includegraphics[scale=0.35]{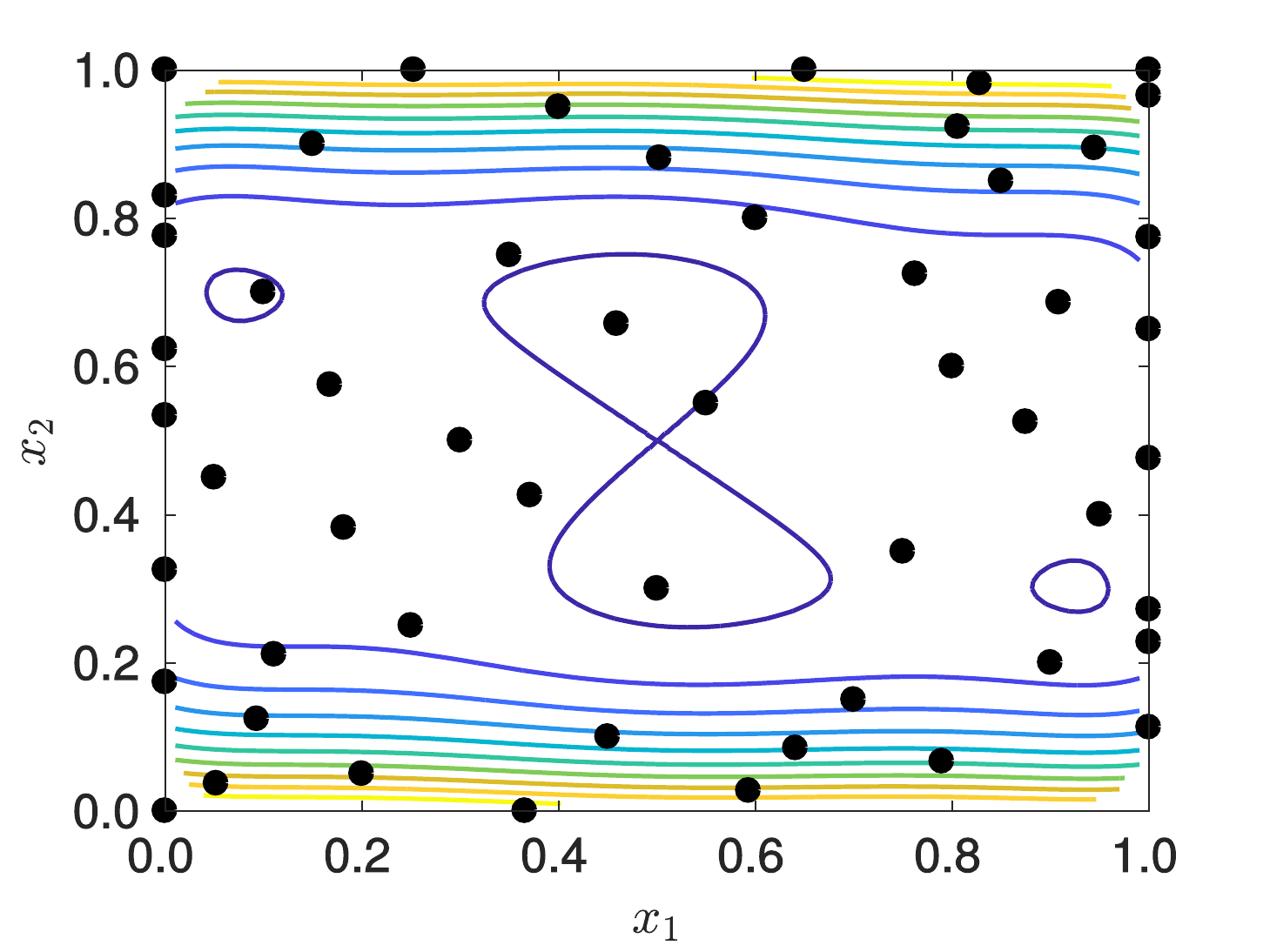}
\subcaption{AME - 55 samples}\label{fig:SixHumbYesAME}
\end{subfigure}
\begin{subfigure}[t]{0.5\textwidth}
\includegraphics[scale=0.35]{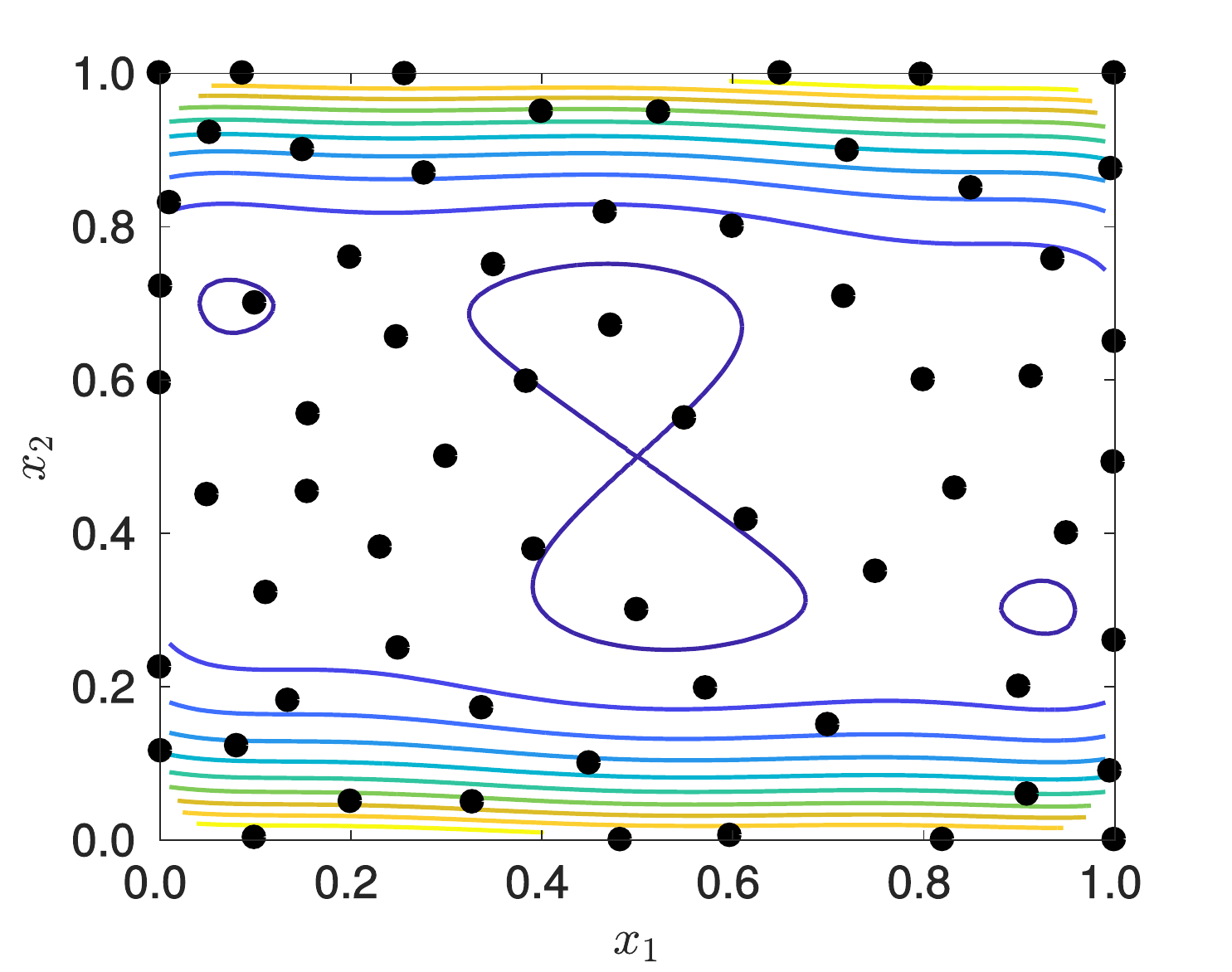}
\subcaption{CVD - 65 samples}\label{fig:SixHumbYesMSE}
\end{subfigure}%
\begin{subfigure}[t]{0.5\textwidth}
\includegraphics[scale=0.35]{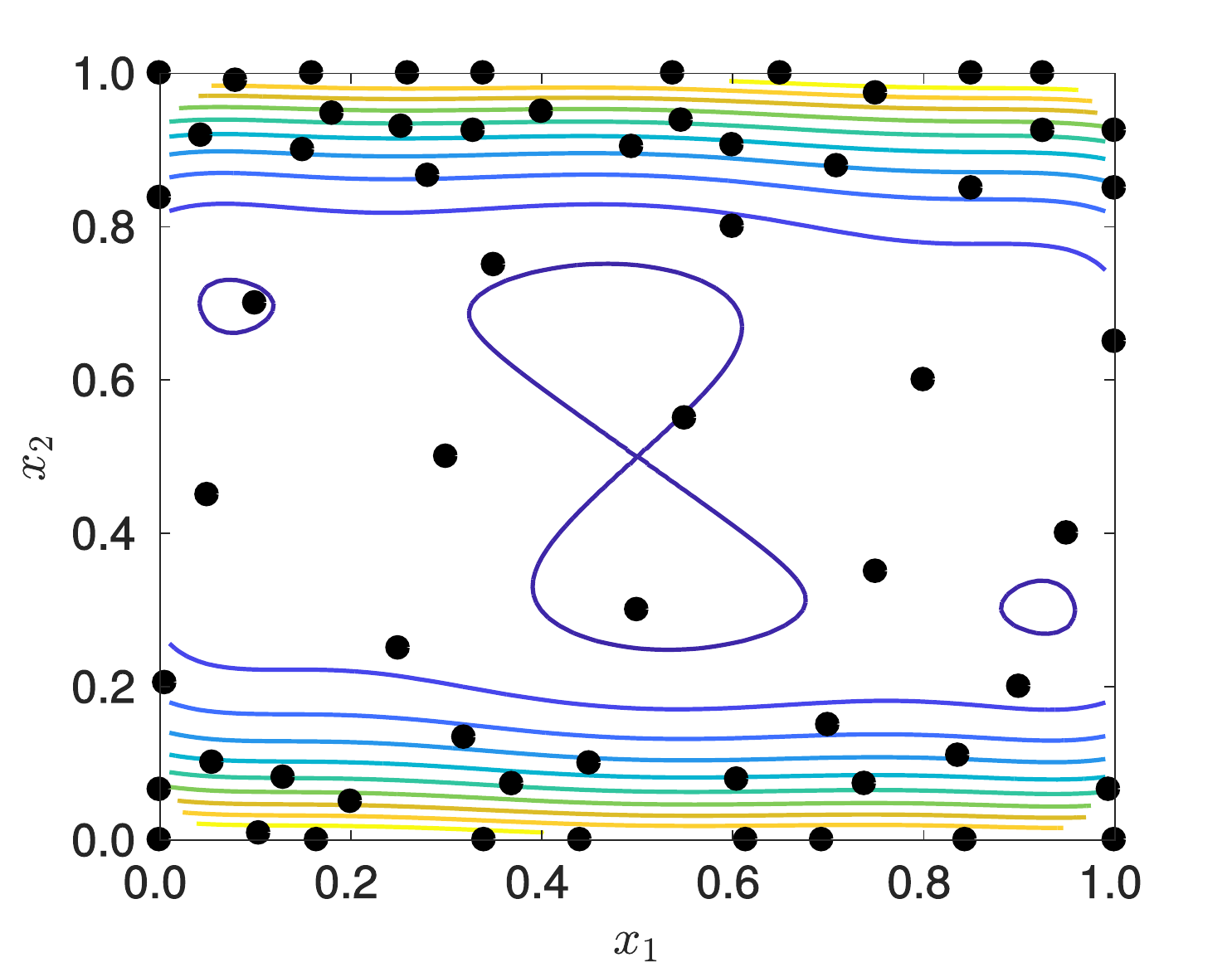}
\subcaption{EIGF - 62 samples}\label{fig:SixHumbYesEIGF}
\end{subfigure}
\begin{subfigure}[t]{0.5\textwidth}
\includegraphics[scale=0.35]{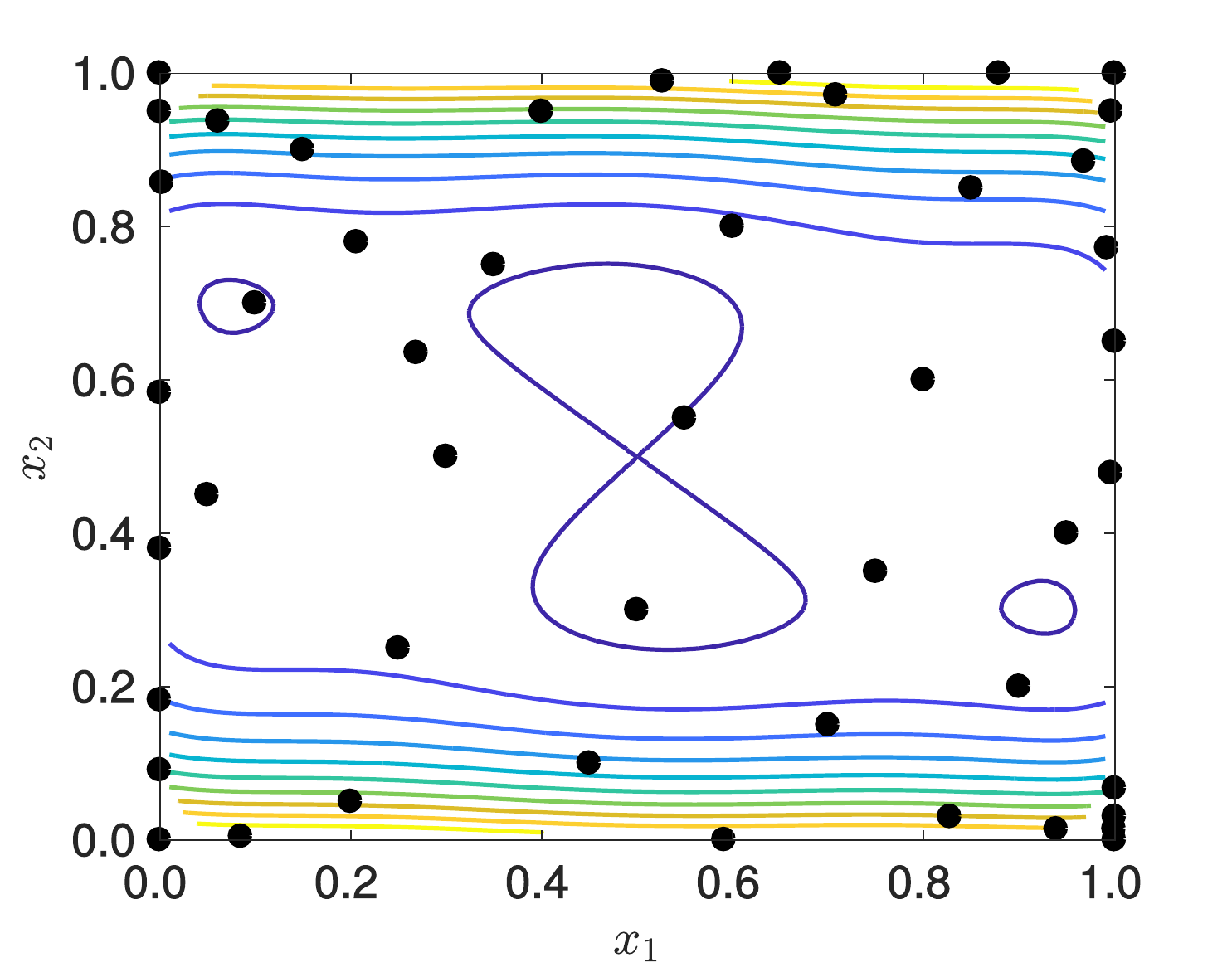}
\subcaption{MEPE- 48 samples}\label{fig:SixHumbYesMEPE}
\end{subfigure}%
\begin{subfigure}[t]{0.5\textwidth}
\includegraphics[scale=0.35]{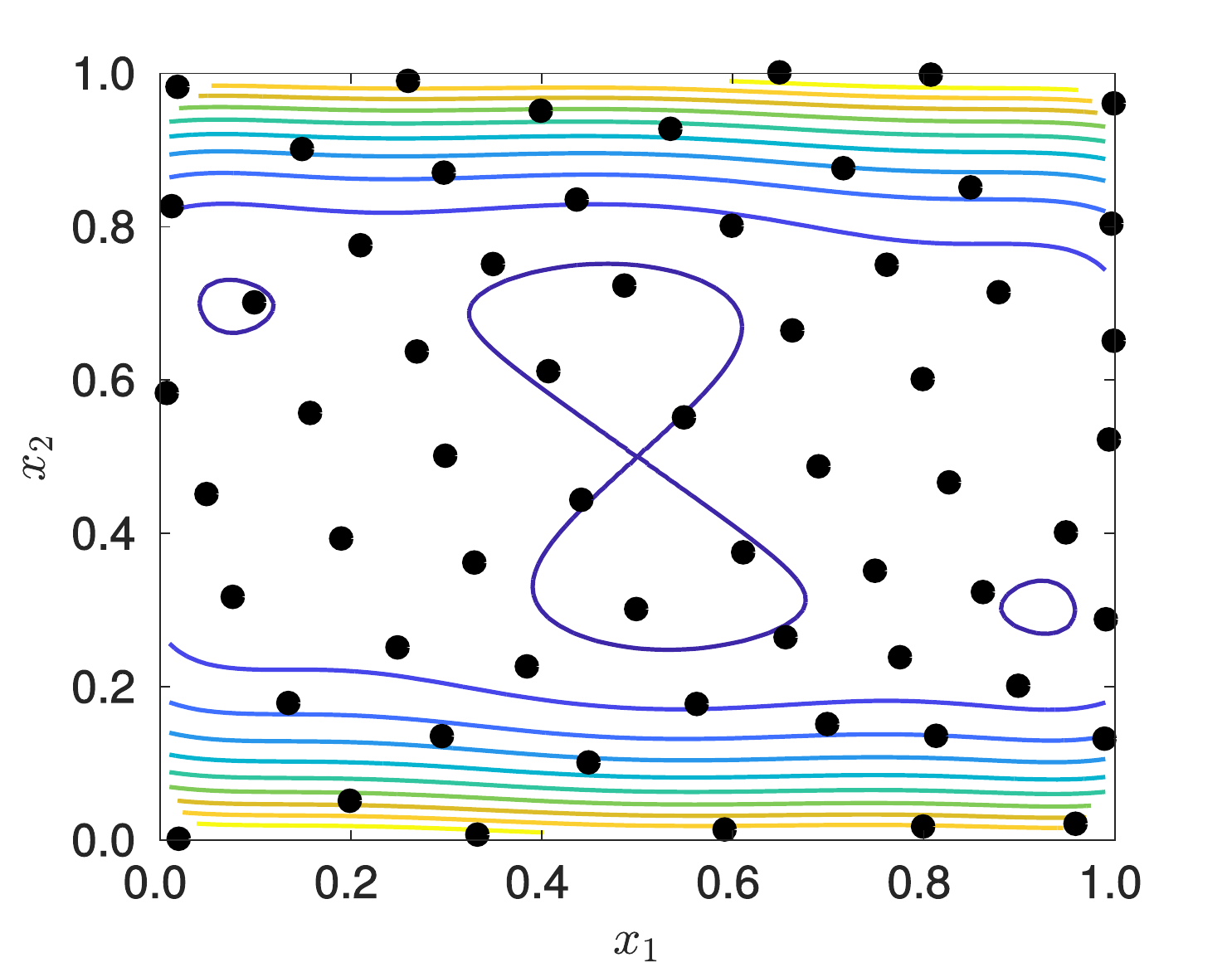}
\subcaption{MIPT - 63 samples}\label{fig:SixHumbYesMIPT}
\end{subfigure}
\begin{subfigure}[t]{0.5\textwidth}
\includegraphics[scale=0.35]{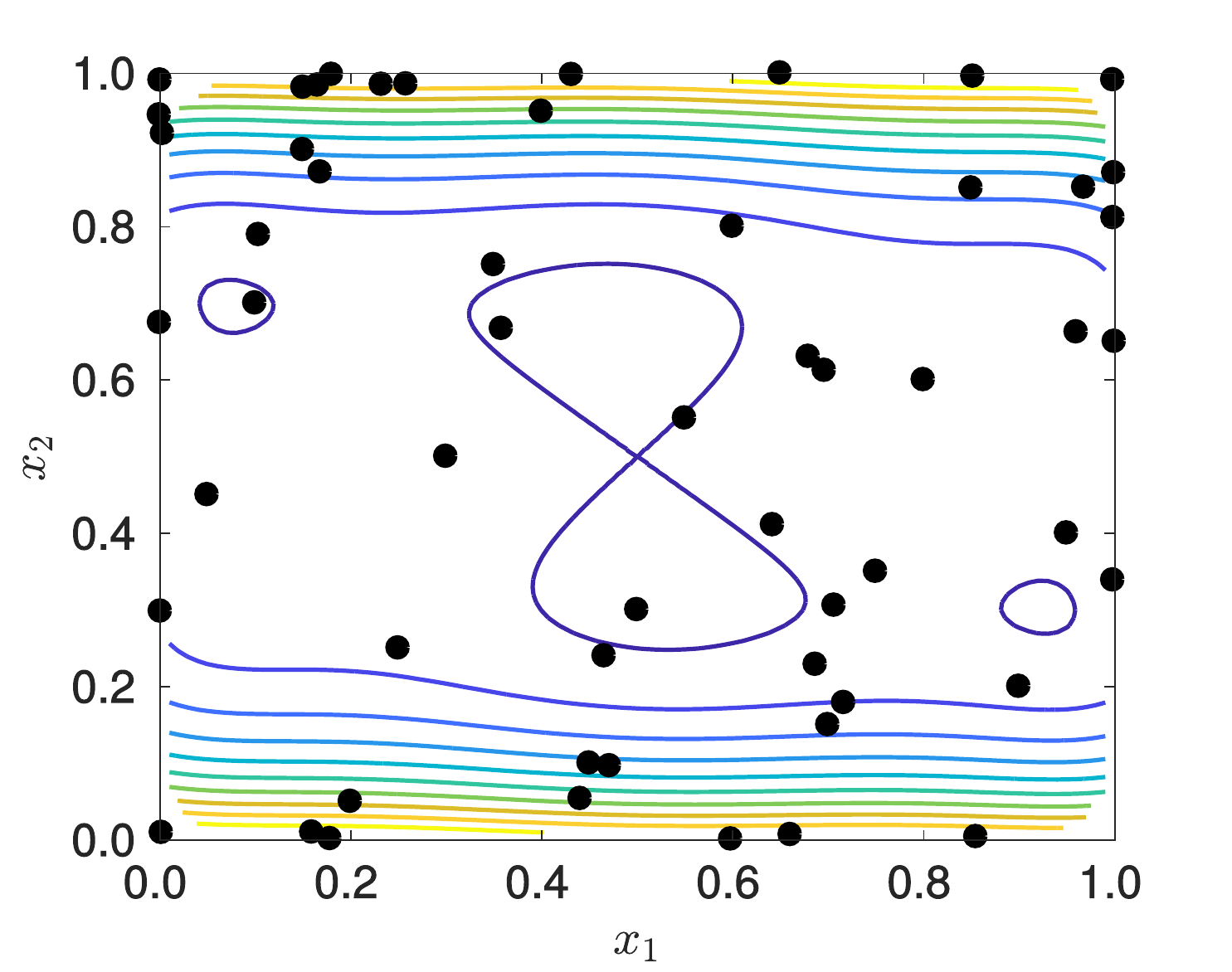}
\subcaption{MSD - 57 samples}\label{fig:SixHumbYesMSD}
\end{subfigure}%
\begin{subfigure}[t]{0.5\textwidth}
\includegraphics[scale=0.37]{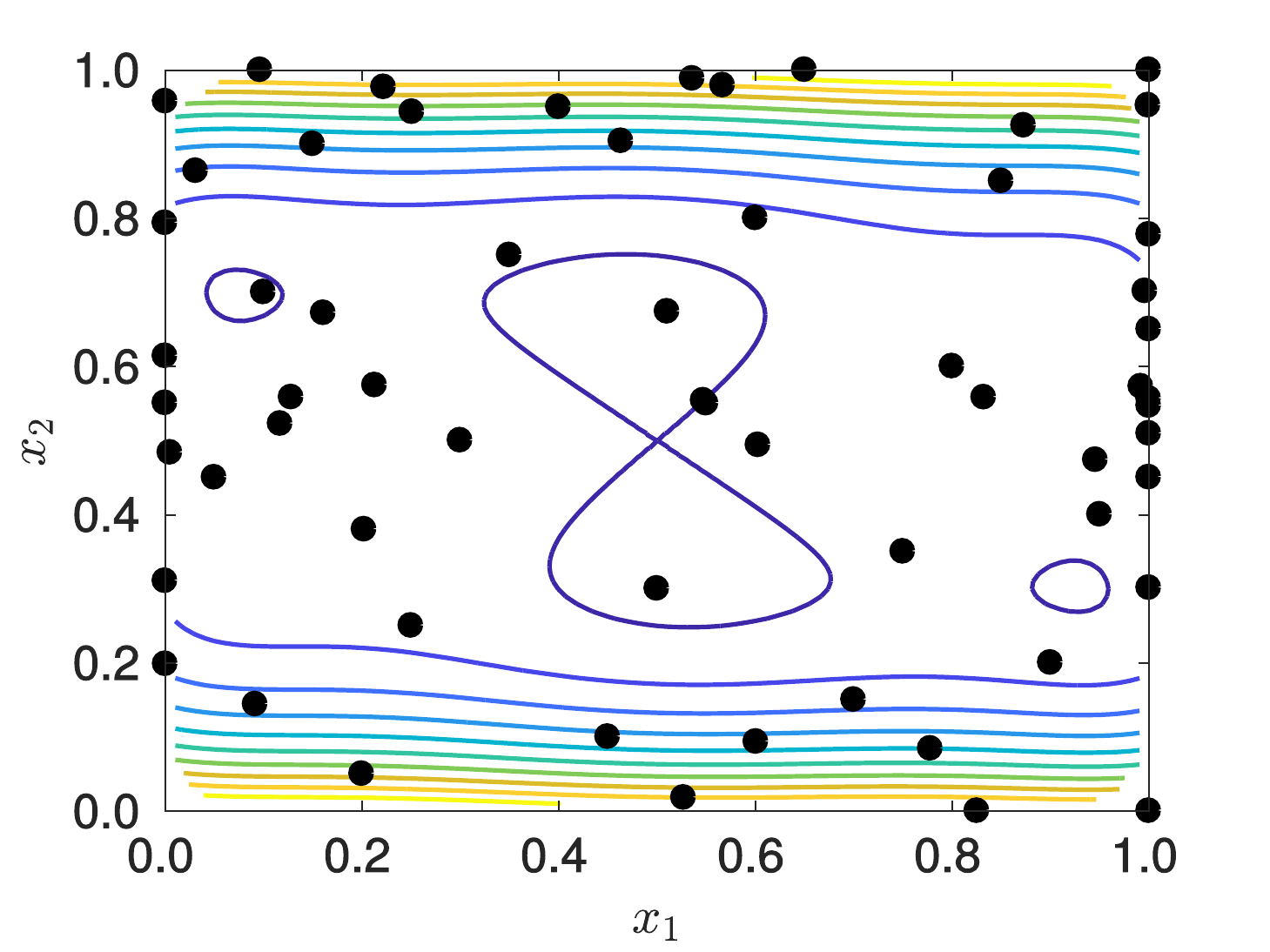}
\subcaption{SSA - 62 samples}\label{fig:SixHumbYesSSA}
\end{subfigure}%
\caption[Sample locations for $\mathcal{M}_{SHC}^{2d}$ of methods reaching error threshold.]{Sample locations for $\mathcal{M}_{SHC}^{2d}$ until threshold MAE$<0.1$ reached for different methods. }\label{fig:SixHumbYes}
\end{figure}
The plots are summarized in Figure \ref{fig:SixHumbYes} with the contour of the target function given in the background and recall the requirements for the sample position of a proficient metamodel stated earlier in this section.
The sample points of ACE which needs 53 samples until the threshold are shown in Figure \ref{fig:SixHumbYesACE}. A lot of points are used to sample three distinct locations in the domain: The area around $(0,0)$, the already found global minimum as well as an area on the right where the function value suddenly increases and the biggest absolute error was recorded in the initial metamodel (c.f. Figure \ref{fig:SixHumbInitialMAE}). It can be seen that the exploitation factor of the function is more pronounced than the exploration. Nevertheless the function is able to reduce the MAE error.   The AME (Figure \ref{fig:SixHumbYesAME}) samples show a distinct exploration-based approach. The edges of the domain are evenly samples. Even three out of fours corner have obtained a point.  
CVD, as depicted in Figure \ref{fig:SixHumbYesMSE}, balances exploration and exploitation with a constant factor. \\It seems in this application the exploration was more pronounced. The domain is evenly covered and not much exploitation of certain areas are visible. Therefore CVD needs 65 sample points to reach the threshold.
EIGF (Figure \ref{fig:SixHumbYesEIGF}) has its focus on the valleys on the upper and lower sides of the domain. As it was noticed before the exploitation factor of EIGF draws points towards high function values. It can be seen that EIGF does not sample in the center and therefore needs 62 samples to reduce the error. 
\begin{figure}[htpb]
\centering
\begin{subfigure}[t]{0.5\textwidth}
\includegraphics[scale=0.35]{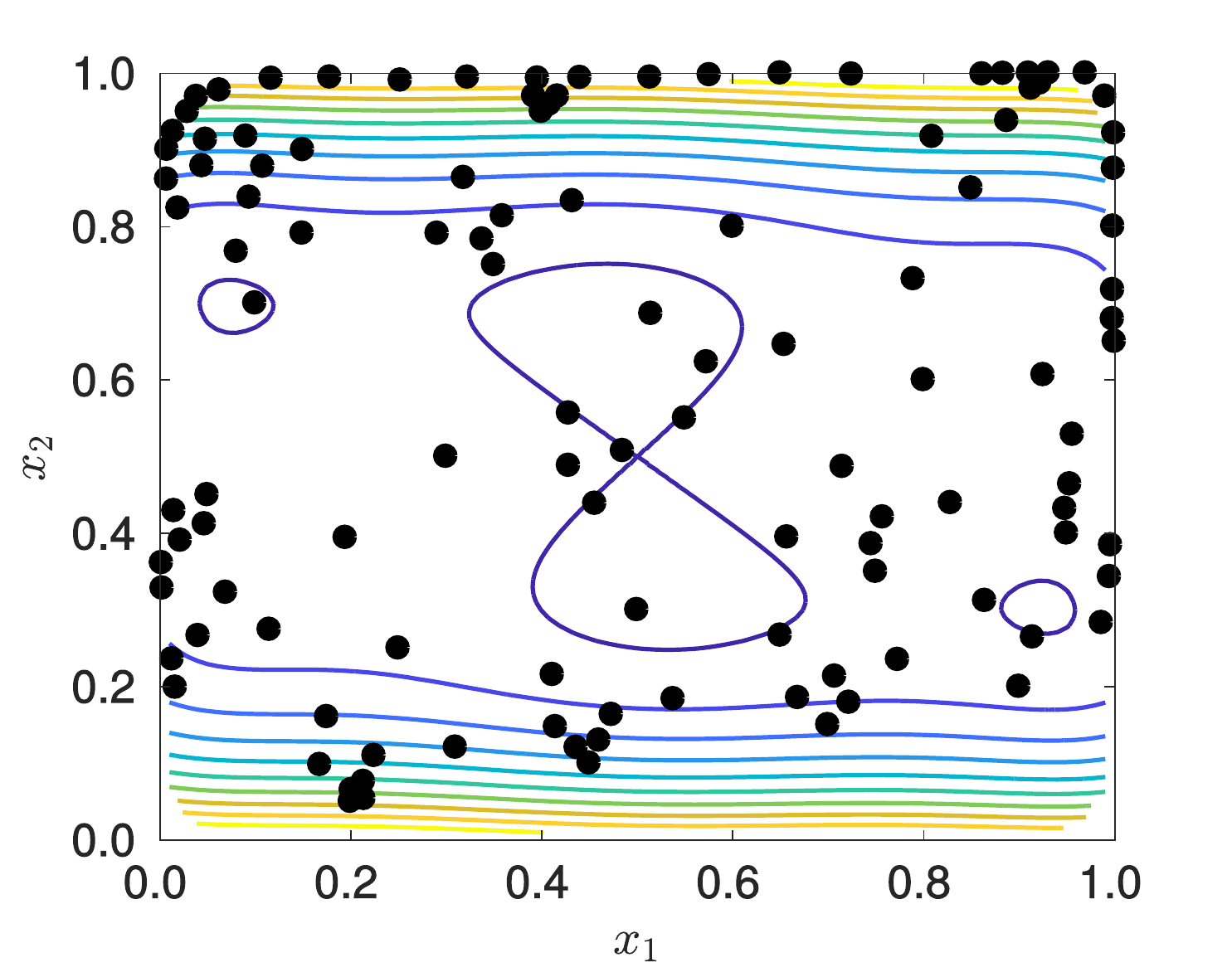}
\subcaption{CVVor - 120 samples}\label{fig:SixHumbNOCVVor}
\end{subfigure}%
\begin{subfigure}[t]{0.5\textwidth}
\includegraphics[scale=0.35]{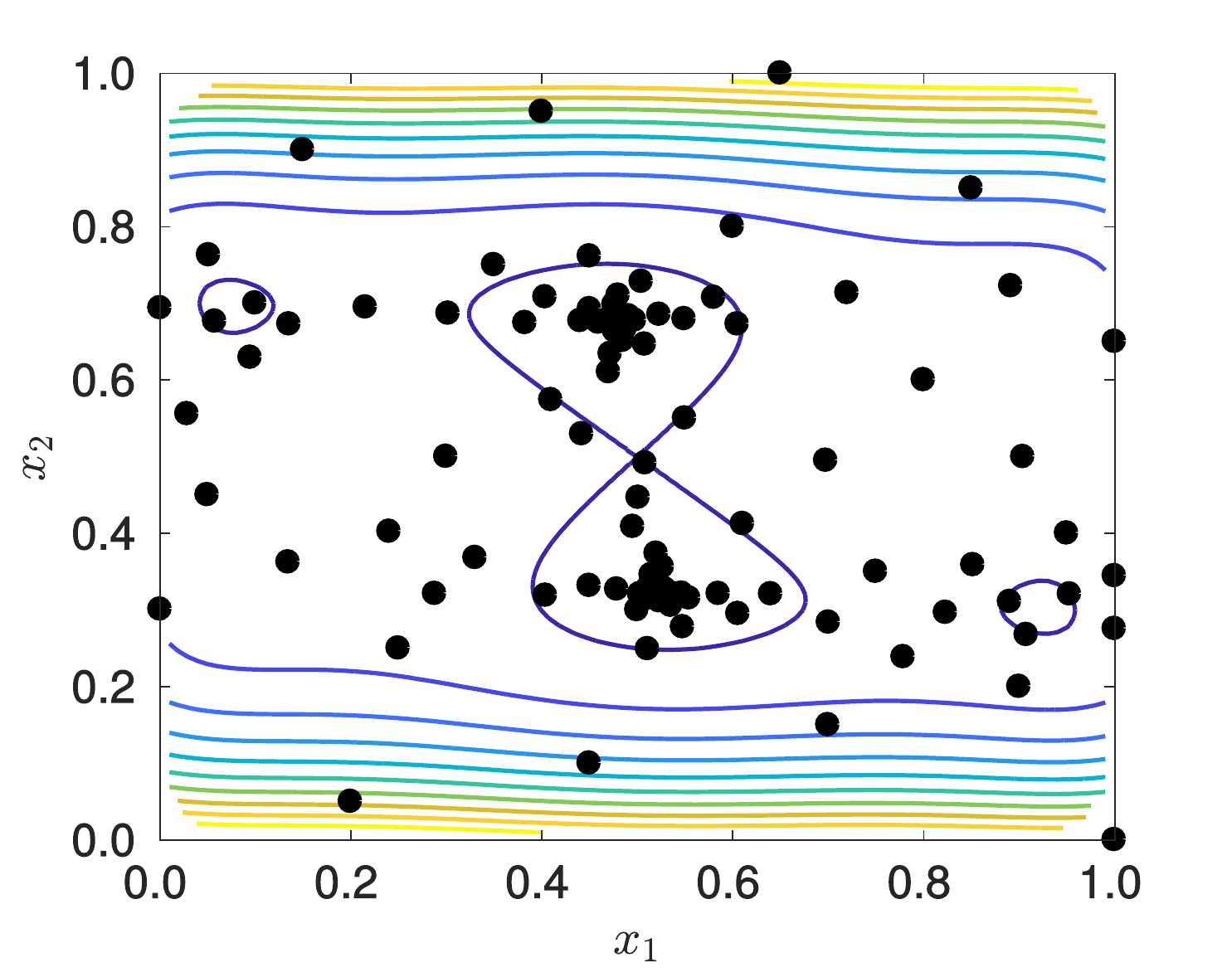}
\subcaption{EI - 120 samples}\label{fig:SixHumbNOEI}
\end{subfigure}
\begin{subfigure}[t]{0.5\textwidth}
\includegraphics[scale=0.35]{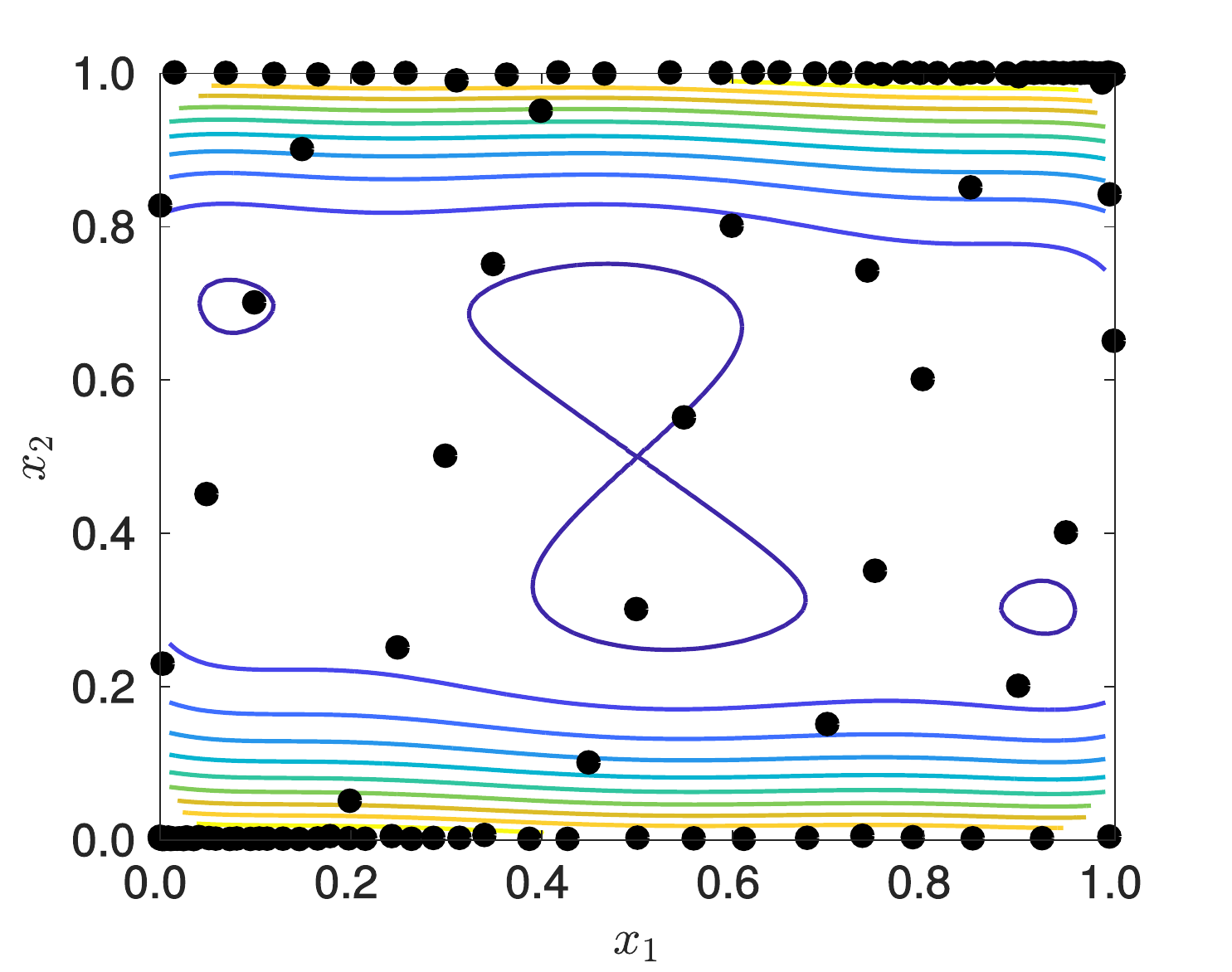}
\subcaption{MASA -120 samples}\label{fig:SixHumbNOMASA}
\end{subfigure}%
\begin{subfigure}[t]{0.5\textwidth}
\includegraphics[scale=0.35]{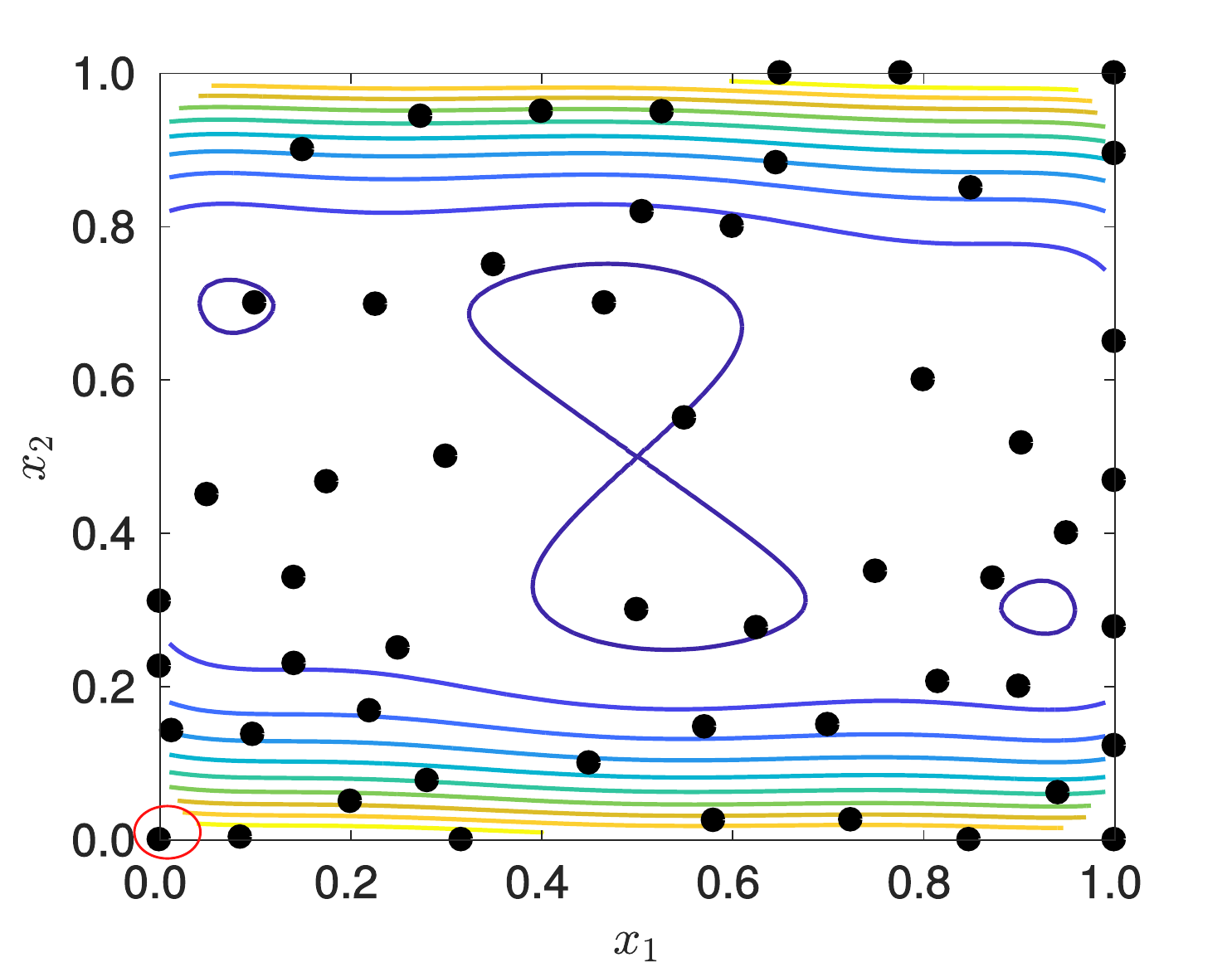}
\subcaption{SFCVT - 55 samples}\label{fig:SixHumbNOSFVCT}
\end{subfigure}
\caption[Sample locations for $\mathcal{M}_{SHC}^{2d}$ of methods failing to reach error threshold.]{Sample locations for $\mathcal{M}_{SHC}^{2d}$ of methods failing to reach threshold of MAE$<0.1$ until 120 samples or numerical problems.}\label{fig:SixHumbNO}
\end{figure}
MEPE as shown in Figure \ref{fig:SixHumbYesMEPE} is the best method in this application. The adaptive balance between exploration and exploitation can be seen as the method is space-filling but also focuses unnecessarily on an area around $(1.0,0.0)$.  
MIPT is purely exploration-based as seen in 
Figure \ref{fig:SixHumbYesMIPT}. Therefore it takes 63 samples, i.e. 15 samples more than MEPE to reduce MAE to the target. 
The 57 samples of MSD are displayed in Figure \ref{fig:SixHumbYesMSD}. The exploration of the algorithm is visible since the edges are evenly covered. However in comparison to the other exploration-based methods MSD also seems to create clusters of samples. 
Finally SSA is covered in Figure \ref{fig:SixHumbYesSSA}. The algorithm needs 62 sample points and shows both exploration and exploitation in its sample generation. 
The adaptive techniques that were unable to reduce the MAE error are presented in Figure \ref{fig:SixHumbNO}.
CVVor, as depicted in Figure \ref{fig:SixHumbNOCVVor}, needs more than 120 samples. It can be seen that some parts of the domain are not covered at all, and that CVVor tends to cluster around certain points. 
As remarked earlier the exploitation of EI (Figure \ref{fig:SixHumbNOEI}) focuses on minimal values. The exploration character underperforms here. Therefore the method exceeds the sample size cap. 
The samples of the MASA algorithm, which focuses only on the maximum values at the edges of the domain and therefore exceeds 120 samples, are shown in Figure \ref{fig:SixHumbNOMASA}. 
SFCVT (Figure \ref{fig:SixHumbNOSFVCT}) is the only algorithm which runs into numerical problems since two points were created at the same spot, here at $(0,0)$. Hence, SFCVT is not able to pursue the adaptive scheme after reaching 55 sample points.
\clearpage
\subsection{Three-dimensional Hartmann function}
\begin{figure}[b!]
\centering
\includegraphics[scale=0.4]{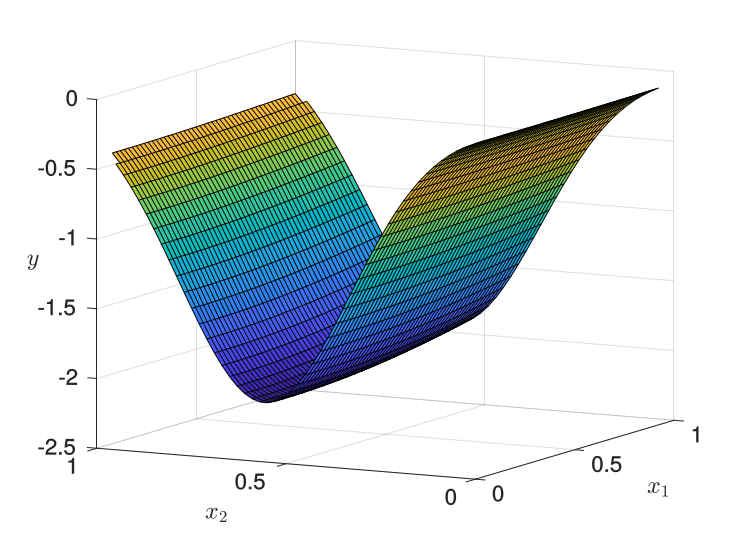}
\caption[$x_{1}$-$x_{2}$-plane of $\mathcal{M}_{H3}^{3d}$]{$x_{1}$-$x_{2}$-plane of $\mathcal{M}_{H3}^{3d}$ for $x_{3}=1$.}\label{fig::Hartman2d}
\end{figure}

The three-dimensional Hartmann function, denoted by $\mathcal{M}_{H3}^{3d}$, has 4 local minima.The function is evaluated on the hypercube $x_{i} \in \left[ 0, 1 \right]$, for all $i= 1,2,3$ and has a global minimum at $\mathcal{M}_{H3}^{3d}(0.1146, 0.5556, 0.8525) = -3.8627$.
The function is given as
\begin{equation}
\mathcal{M}_{H3}^{3d}(\bm{x}) = - \sum_{1}^{4} \alpha_{i} \exp \left( - \sum_{j=1}^{3} A_{ij} (x_{j}- P_{ij})^{2} \right)
\end{equation}
with
\begin{equation}
\begin{aligned}
\bm{\alpha} &= \begin{bmatrix}
1.0 & 1.2 & 3.0 & 3.2
\end{bmatrix}^{T}, \\
\bm{A} &= \begin{bmatrix}
3.0 & 10.0 & 30.0 \\
0.1 & 10.0 & 35.0 \\
3.0 & 10.0 & 30.0 \\
0.1 & 10.0 & 35.0
\end{bmatrix}, \\
\bm{P} &= 10^{-4} \begin{bmatrix}
3689.0 & 1170.0 & 2673.0 \\
4699.0 & 4387.0 & 7470.0 \\
1091.0 & 8732.0 & 5547.0 \\
381.0  & 5743.0 & 8828.0
\end{bmatrix} \, \text{.}
\end{aligned}
\end{equation}
This function has for example been used as a benchmark for metamodel construction in \cite{jiang2018two}. The Hartman function over the $x_{1}-x_{2}$-plane with the third value fixed to unity is displayed in Figure \ref{fig::Hartman2d}. It shows that there is high nonlinearity involved in the function. Furthermore the spread of the function is around $3.7$. This gives an idea for the non-relative error measures that will follow later. 

\begin{figure}[t!]
\centering
\includegraphics[scale=0.6]{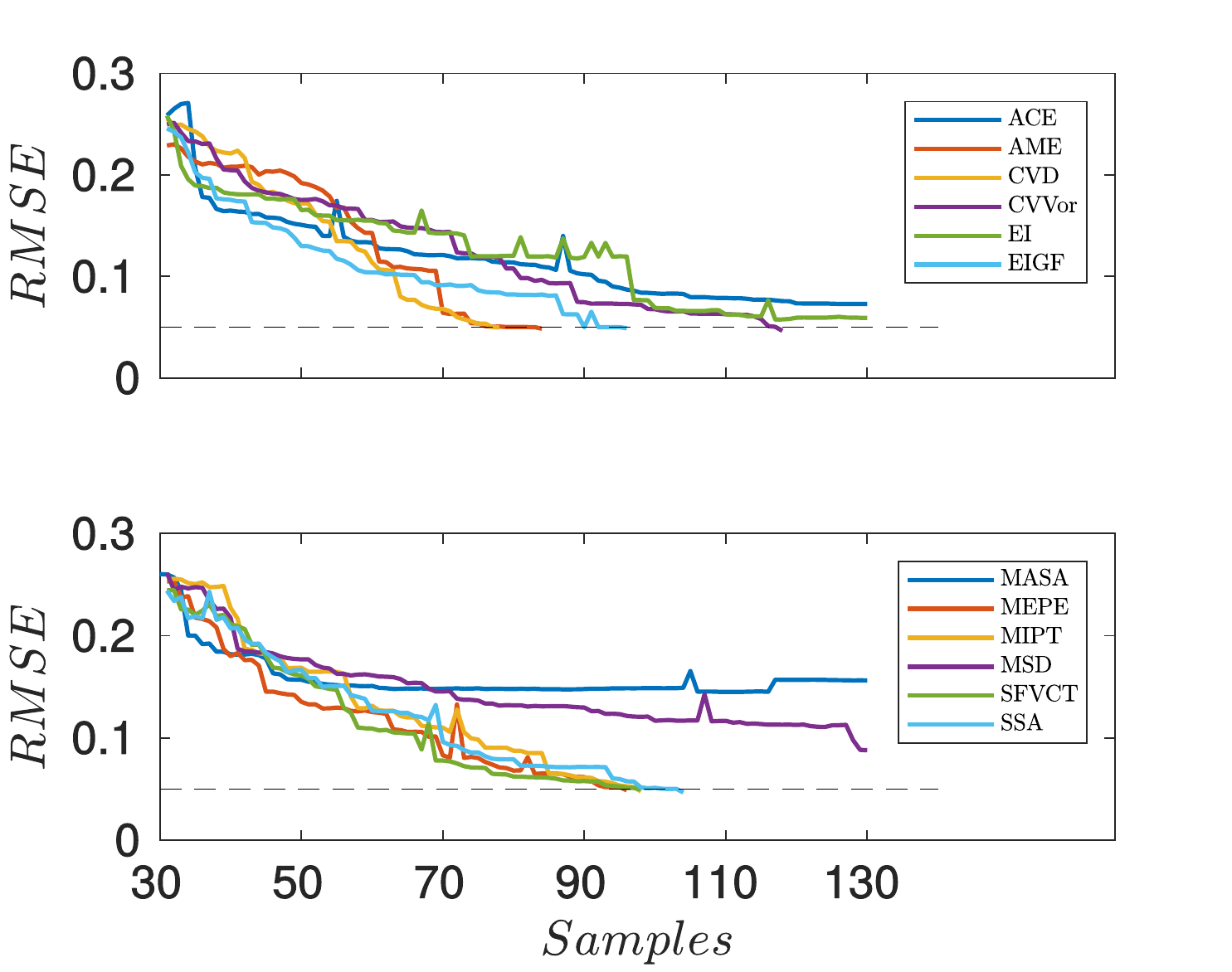}
\caption[Convergence of MAE value for $\mathcal{M}_{H3}^{3d}$ ]{Convergence of MAE value for $\mathcal{M}_{H3}^{3d}$ until MAE$<0.05$ (capped at 130 samples).}\label{fig::HartmanMAEConv}
\end{figure}

\begin{figure}[b!]
\centering
\includegraphics[scale=0.5]{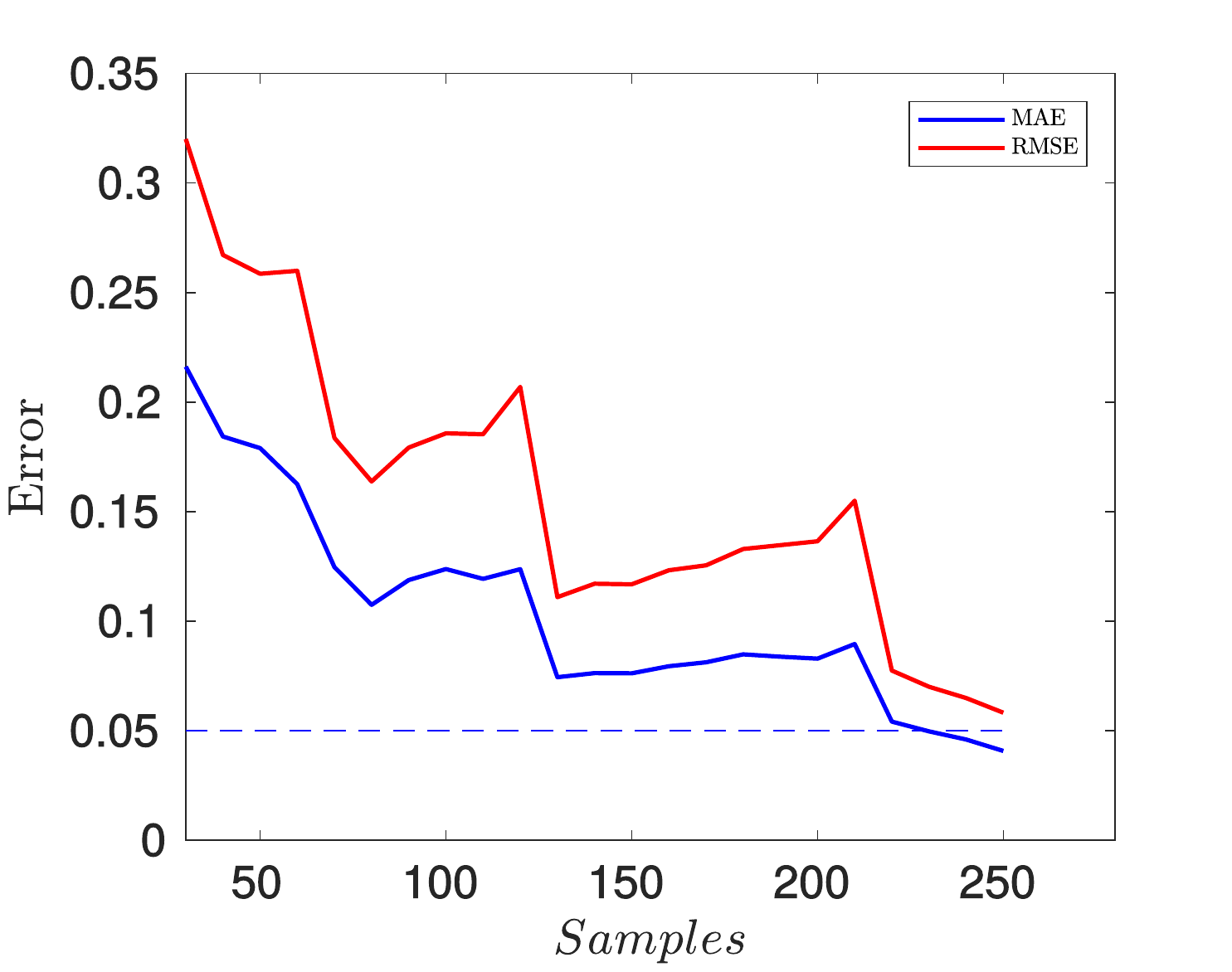}
\caption[MAE and RMSE values for $\mathcal{M}_{H3}^{3d}$ with TPLHD samples]{MAE and RMSE values for $\mathcal{M}_{H3}^{3d}$ in with samples created with TPLHD with step size 10 (blue dotted line illustrates MAE threshold).}\label{fig::TPLHD_Hartman}
\end{figure}

\begin{figure}[t!]
\centering
\includegraphics[scale=0.6]{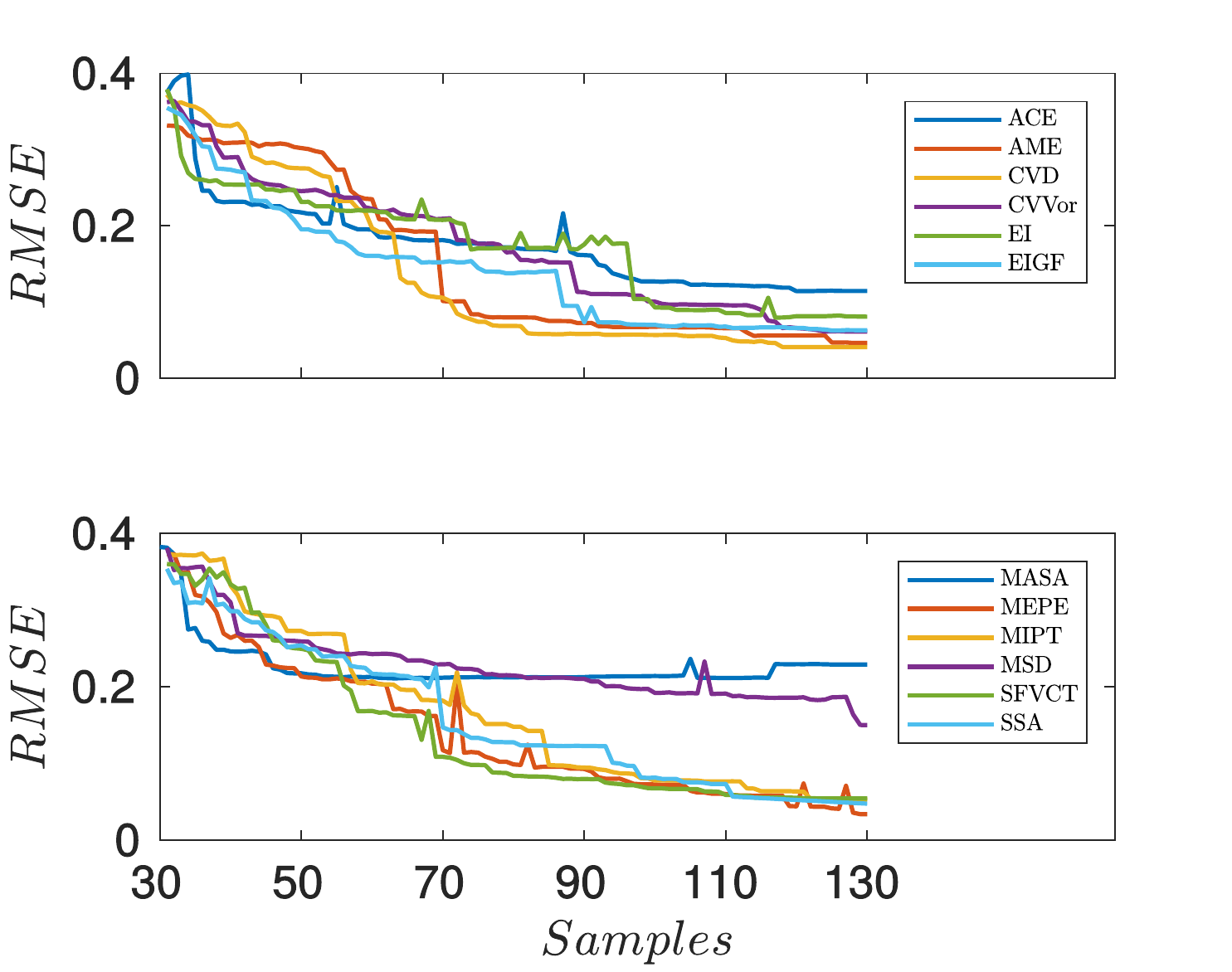}
\caption[Convergence of RMSE value for $\mathcal{M}_{H3}^{3d}$ ]{Convergence of MAE value for $\mathcal{M}_{H3}^{3d}$ until 130 samples for different adaptive sampling technqiues.}\label{fig::HartmanRMSEConv}
\end{figure}

Since the computational effort for LOLA would be too high, the technique is omitted here. The error measures for the adaptive sampling strategies when starting with 30 samples and iteratively extending to 60 are listed in table \ref{tab::H360}. 
It can be pointed out that in three dimensions no sampling technique stopped due to numerical problems e.g. clustering.
\begin{table}[t!]
\begin{center}
\resizebox{1.0\textwidth}{!}{%
\begin{tabular}{l|l c c c c} \hline
Sampling method & MAE & RMAE & RMSE & R$^{2}$ \\ \hline\hline \\
\multirow{1}{*}{\shortstack[l]{Errors after\\30 samples}} & TPLHD  & 0.2161 & 1.3858 & 0.3201 & 0.8883 \\ \\ \hline \\ 
\multirow{13}{*}{\shortstack[l]{ Errors after\\60 samples}} &TPLHD &   0.1624 & 1.0775 & 0.2600 & 0.9263 \\
&ACE &  0.1333 & \textbf{0.9687 }& 0.1949 & 	0.9583 \\
&AME &  0.1147 & 1.1641 & 0.2085 & 0.9523 \\ 
&CVD &   0.1136 & 1.1544 & 0.1966 & 	0.9576  \\ 
&CVVor &   0.1556 & 1.0840 & 0.2221 & 0.9459  \\ 
&EI &   0.1551 & 	1.1542 & 0.2198 & 0.9470 \\
&EIGF &  \textbf{0.1039} & 1.1272 & \textbf{0.1602}   & \textbf{0.9718}\\
&MASA &  0.1508 & 1.1572 & 0.2131  & 0.9502 \\
&MEPE &  0.1253 & 1.1726 & 0.2041 & 0.9543 \\
&MIPT &  0.1315 & 1.0121 & 0.2071  & 0.9529\\
&MSD & 0.16124 & 1.1694 & 0.2426 & 0.9354  \\
&SFCVT &   0.1073 & 1.0228 & 0.1662  & 0.9696 \\
&SSA &   0.1266 & 1.2507 & 0.2168 & 0.9484
\end{tabular}
}
\end{center}
\caption[Error measures for $\mathcal{M}_{H3}^{3d}$ after 60 samples]{Error measures for $\mathcal{M}_{H3}^{3d}$ after 60 samples.}\label{tab::H360}
\end{table}
It can be seen that up to 60 samples EIGF yields the best results for 3 out of the 4 error measures. RMAE is the lowest with ACE after 60 samples. The results show that all adaptive sampling methods were able to reduce the initial error considerably. Furthermore for MAE, RMSE and R$^{2}$ sampling points individually offers the better results than just using a space-filling TPLHD approach. Since all of these measures describe the general goodness of shape approximation it can be said that the form of the Hartmann function is fitted more proficiently this way. RMAE for 60 TPLHD samples performs better than for 9 out of the 12 methods, which can be explained by the fortunate point placement of the space-filling method and the accurate prediction of OK at sample points.
Generally all tested methods provide good results and the difference between them is not drastic. 
In a second step the convergence of the described techniques is studied. The number of samples needed for each method to reduce the MAE error below $0.05$ are listed in Table \ref{tab::HartmanLimit}. A number of 130 samples is taken as the cut-off. It can be seen that 8 of the 12 methods are able to reduce the measure accordingly.
\newpage
The variation of the needed points is small which speaks for the hybrid particle swarm optimization as a proficient optimization scheme for the hyperparameters and furthermore that the implementations of these methods are sufficient in reducing the randomness of the optimizations involved. 
Amongst the methods listed in Table \ref{tab::HartmanLimit} CVD is the most efficient method with a need of only 78 samples. CVVor needs the most with 118. 
\begin{table}[t!]
\begin{center}
\begin{tabular}{c c } \hline
\shortstack[l]{Sampling\\method} & \shortstack[l]{Average number\\of Samples}\\ \hline\hline \\
ACE &  -  \\
AME & 84 $\pm$ 1  \\ 
CVD & 78 $\pm$ 0 \\
CVVor &  118 $\pm$ 1 \\ 
EI & - \\
EIGF & 96 $\pm$ 0 \\
MASA & -  \\
MEPE & 96 $\pm$ 0 \\
MIPT & 98 $\pm$ 1 \\
MSD & -  \\
SFCVT &  98 $\pm$ 1 \\
SSA &  104 $\pm$ 2  \\ 
\end{tabular}
\end{center}
\caption[Average amount of samples required before MAE$<0.05$ for $\mathcal{M}_{H3}^{3d}$]{Average amount of samples required before MAE$<0.05$ for $\mathcal{M}_{H3}^{3d}$. The variation of the number of samples over 10 computations is given as an additional value. Methods that do not reach the target value because of clustering or because they need more than 130 samples are omitted.}\label{tab::HartmanLimit}
\end{table}
The rate of convergence of MAE of all methods until 130 samples is shown in Figure \ref{fig::HartmanMAEConv}. The values given are the average over ten computations. The target MAE value of $0.05$ is illustrated as a horizontal line. It can be seen that after around 70 samples the accuracy of MASA and MSD is stagnating with MSD only dipping again in the last samples. EI and ACE are close to the threshold and might be able to reduce the measure further with more points. The rest of the technique show a more or less constant decline with occasional jumps involved. 
The errors of Kriging models based on                       samples generated with TPLHD starting with 30 samples, up to 250 and incremented with 10 each step, is displayed in Figure \ref{fig::TPLHD_Hartman}. 
Plotted are the MAE and RMSE measures.
It can be noticed that without an adaptive approach the MAE error threshold (blue-dotted line) is only reached after 230 sample points of TPLHD.\newpage  As a comparison the best adaptive sampling technique (CVD) needs around a third of the points.
The evolution of the error measure RMSE is plotted for all the methods between 30 to 130 samples in Figure \ref{fig::HartmanRMSEConv}. With this measure it is more obvious that some methods start stagnating after a certain amount of points are reached. Besides, the techniques that have been successfully able to reduce the MAE error show a more proficient converging behavior with RMSE as well. 

\clearpage
\subsection{Five-dimensional Trid function}\label{sec::Trid}
\begin{figure}[hbtp]
\centering
\includegraphics[scale=0.4]{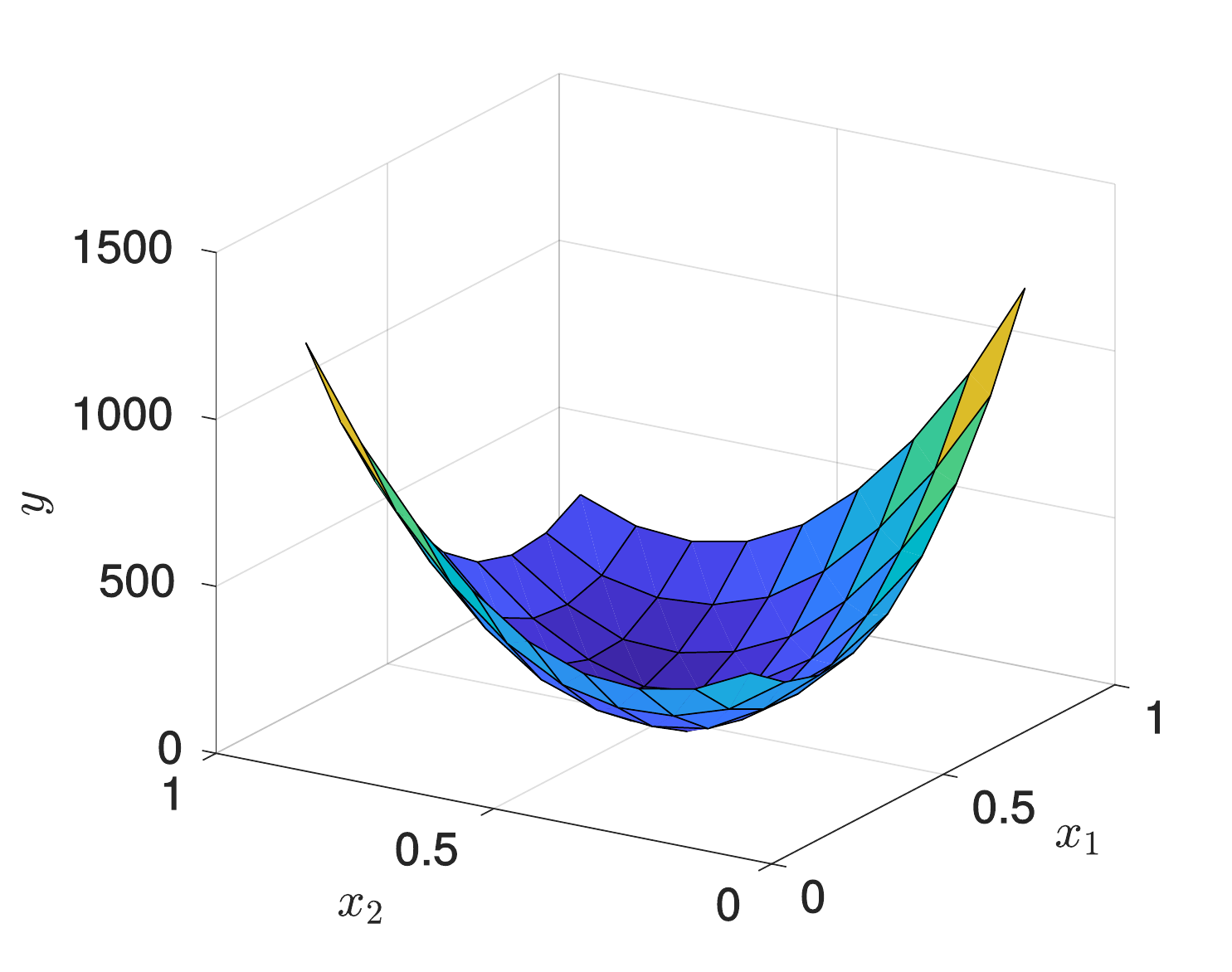}
\caption[Trid function $\mathcal{M}_{Trid}^{5d}$ projected onto to the normalized $x_{1}$-$x_{2}$-plane]{Trid function $\mathcal{M}_{Trid}^{5d}$ projected onto to the normalized $x_{1}$-$x_{2}$-plane with all other parameter values equaling zero.}\label{fig::TridProjected}
\end{figure}

The five-dimensional Trid function as e.g. found in \cite{adorio2005mvf} has the form
\begin{equation}
\mathcal{M}_{Trid}^{5d}(\bm{x}) = \sum_{i=1}^{5} (x_{i}-1)^{2} - \sum_{i=1}^{5} x_{i} x_{i-1}.
\end{equation}
The function is here evaluated in the input domain $[-25,25]$ for each dimension. The function has one global minimum $\mathcal{M}_{Trid}^{5d}(\bm{x}_{min}) = -30$ at $\bm{x}_{min} = (5,8,9,8,5)^{T}$.  The underlying nonlinearity of the function is highlighted in Figure \ref{fig::TridProjected}. The function response over the normalized $x_{1}$-$x_{2}$-plane with the rest of the 5 parameters being set to zero is depicted. 

\begin{figure}[t!]
\centering
\includegraphics[scale=0.5]{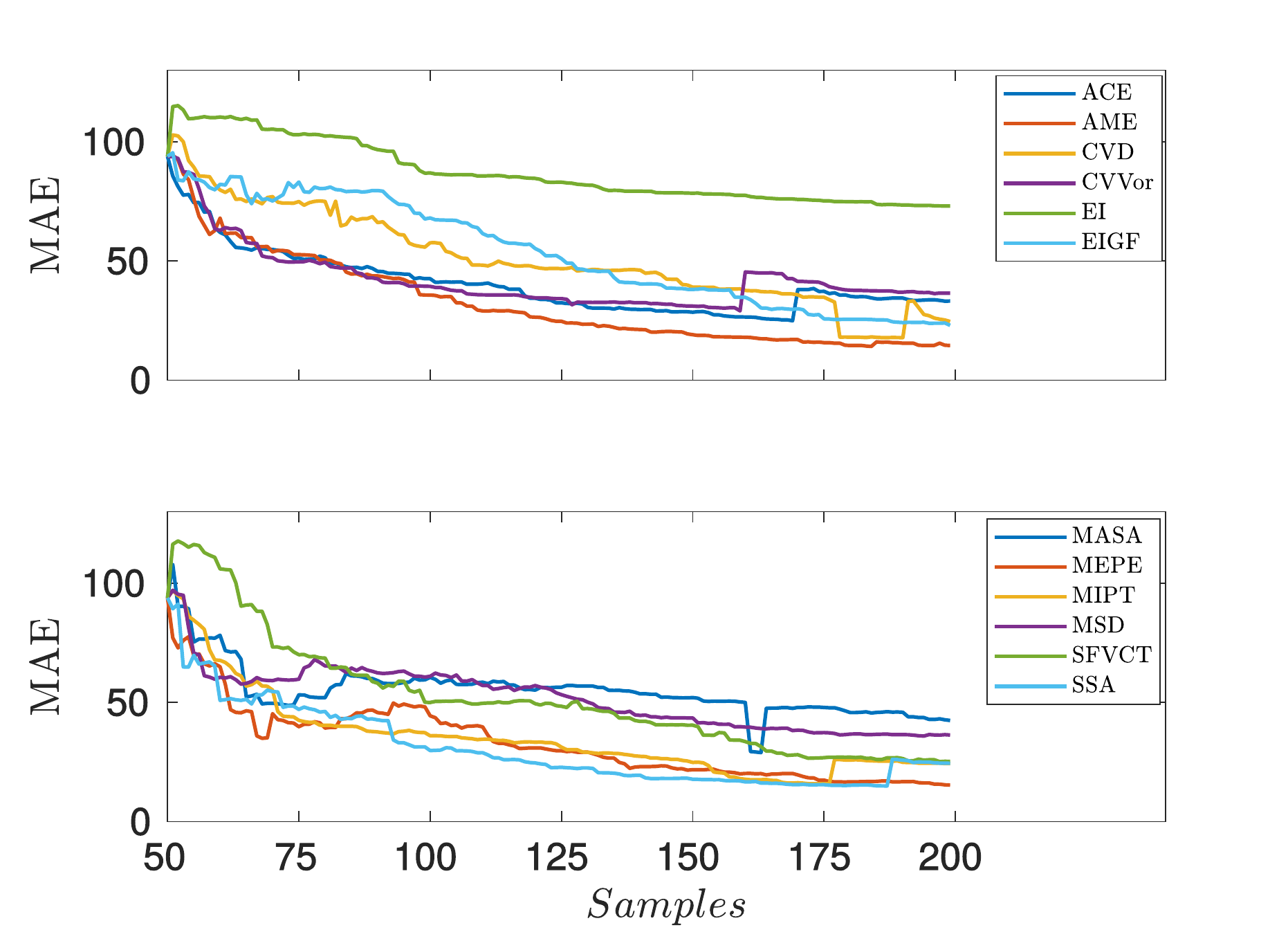}
\caption[Convergence of the MAE error for $\mathcal{M}_{Trid}^{5d}$ until 200 samples]{Connvergence of the MAE error for $\mathcal{M}_{Trid}^{5d}$ until 200 samples.}\label{fig::TridMAEBehav}
\end{figure}

The average error measures of the adaptive sampling techniques when starting from 50 samples and extending the sample size to 150, is listed in Table \ref{table::Tried150}. It can be seen that all the methods are able to reduce the error with respect to the initial metamodel with 50 TPLHD samples. However EI shows clustering behavior and therefore does not show a substantial improvement. Except EI, the other presented methods show more proficient results than a metamodel created with 150 TPLHD samples. SSA, AME, MEPE and MIPT are able to reduce the MAE error by 3 with respect to the one-shot model. As highlighted, SSA shows the best results in all four error categories.  
\begin{table}[t!]
\begin{center}
\resizebox{1.0\textwidth}{!}{%
\begin{tabular}{l|l c c c c} \hline
& Method & MAE & RMAE & RMSE & R$^{2}$  \\ \hline\hline \\
\multirow{1}{*}{\shortstack[l]{Errors after\\50 samples}} & TPLHD & 93.8611 & 1.1550 & 118.1866 & 0.9455 \\ \\  \hline \\
\multirow{13}{*}{\shortstack[l]{ Errors after\\150 samples}} &TPLHD &62.6720 & 0.9387 & 87.9973 & 0.9698 \\
&ACE & 46.2450 & 0.7657 & 68.7834 & 0.9826 \\
&AME & 19.0769 & 0.2626 & 25.9363 & 0.9975 \\
&CVD & 38.9972 & 0.4285 & 49.3561 & 0.9910  \\
&CVVor & 31.0931 & 0.5944 & 46.9048 & 0.9919  \\ 
&EI &  78.4929 & 1.2724 & 116.2119 & 0.9505 \\
&EIGF & 38.0913 & 0.5281 & 51.0940 & 0.9904\\
&MASA &  51.9301 & 0.5639 & 67.2773 & 0.9834 \\
&MEPE &  21.7821 & 0.2394 & 27.5633 & 0.9972\\
&MIPT &  24.826 & 0.4689 & 33.1095 & 0.9959 \\ 
&MSD & 43.4871 & 0.3884 & 53.7404 & 0.9894\\
&SFCVT &40.4675 & 0.8251 & 58.0677 & 0.9876\\ 
&SSA &  \textbf{17.7730} & \textbf{0.2524} &\textbf{ 23.6932} & \textbf{0.9979}
\end{tabular}
}
\end{center}
\caption[Error measures for $\mathcal{M}_{Trid}^{5d}$ after 150 samples.]{Error measures for $\mathcal{M}_{Trid}^{5d}$ after 150 samples. }\label{table::Tried150}
\end{table}
The average scores after adding additional 50 samples are listed in Table \ref{table::Trid200}. It can be seen that with respect to the error after 150 samples of Table \ref{table::Tried150} the TPLHD created metamodel after 200 samples shows an improved approximation. When looking at the MAE error 10 out of the 12 adaptive techniques achieve a more sufficient solution than the one-shot method. In addition to EI, MASA is the second method that shows clustering. \newpage
It can be seen that MEPE overall is able to add the samples most effectively after 200 points. 
\begin{table}[t!]
\begin{center}
\resizebox{1.0\textwidth}{!}{%
\begin{tabular}{l|l c c c c} \hline
&Method & MAE & RMAE & RMSE & R$^{2}$  \\ \hline\hline \\
\multirow{1}{*}{\shortstack[l]{Errors after\\50 samples}} & TPLHD  & 93.8611 & 1.1550 & 118.1866 & 0.9455\\\\  \hline \\
\multirow{13}{*}{\shortstack[l]{ Errors after\\200 samples}} & TPLHD  & 40.2345 & 0.6537 & 60.8234 & 0.9960 \\
&ACE & 28.5187 & 0.7657 & 40.4943 & 0.9947 \\
&AME & \textbf{14.6074} & 0.2491 & 23.0186 & 0.9983\\
&CVD & 24.4153 & 0.2436 & 30.2288 & 0.9966  \\
&CVVor & 36.4604 & 0.8105 & 57.7857 & 0.9877   \\ 
&EI &  73.0383 & 1.2359 & 110.1668 & 0.9582 \\
&EIGF &  22.6122 & 0.4153 & 30.1483 & 0.9966\\
&MASA &  42.4401 & 0.5633 & 55.6606 & 0.9886 \\
&MEPE &  14.7321 & \textbf{0.1427} & \textbf{18.7683} & \textbf{0.9989}\\
&MIPT &  24.3554 & 0.3861 & 31.7237 & 0.9963\\ 
&MSD &36.2998 & 0.3130 & 44.5949 & 0.9927\\
&SFCVT &24.9267 & 0.3513 & 33.4248 & 0.9959 \\ 
&SSA &  23.8567 & 0.4013 & 32.1571 & 0.9962
\end{tabular}
}
\end{center}
\caption[Error measures for $\mathcal{M}_{Trid}^{5d}$ after 200 samples.]{Error measures for $\mathcal{M}_{Trid}^{5d}$ after 200 samples. }\label{table::Trid200}
\end{table}
The MAE error over the added samples (from 50 to 200) is displayed in Figure \ref{fig::TridMAEBehav}. All methods show a trend towards a constant decrease of the measure.
However after reaching 150 samples, the methods appear to be stalling or decreasing at a slower rate.

\clearpage
\subsection{Seven-dimensional Lévy function}  
\begin{figure}[h!]
\centering
\includegraphics[scale=0.5]{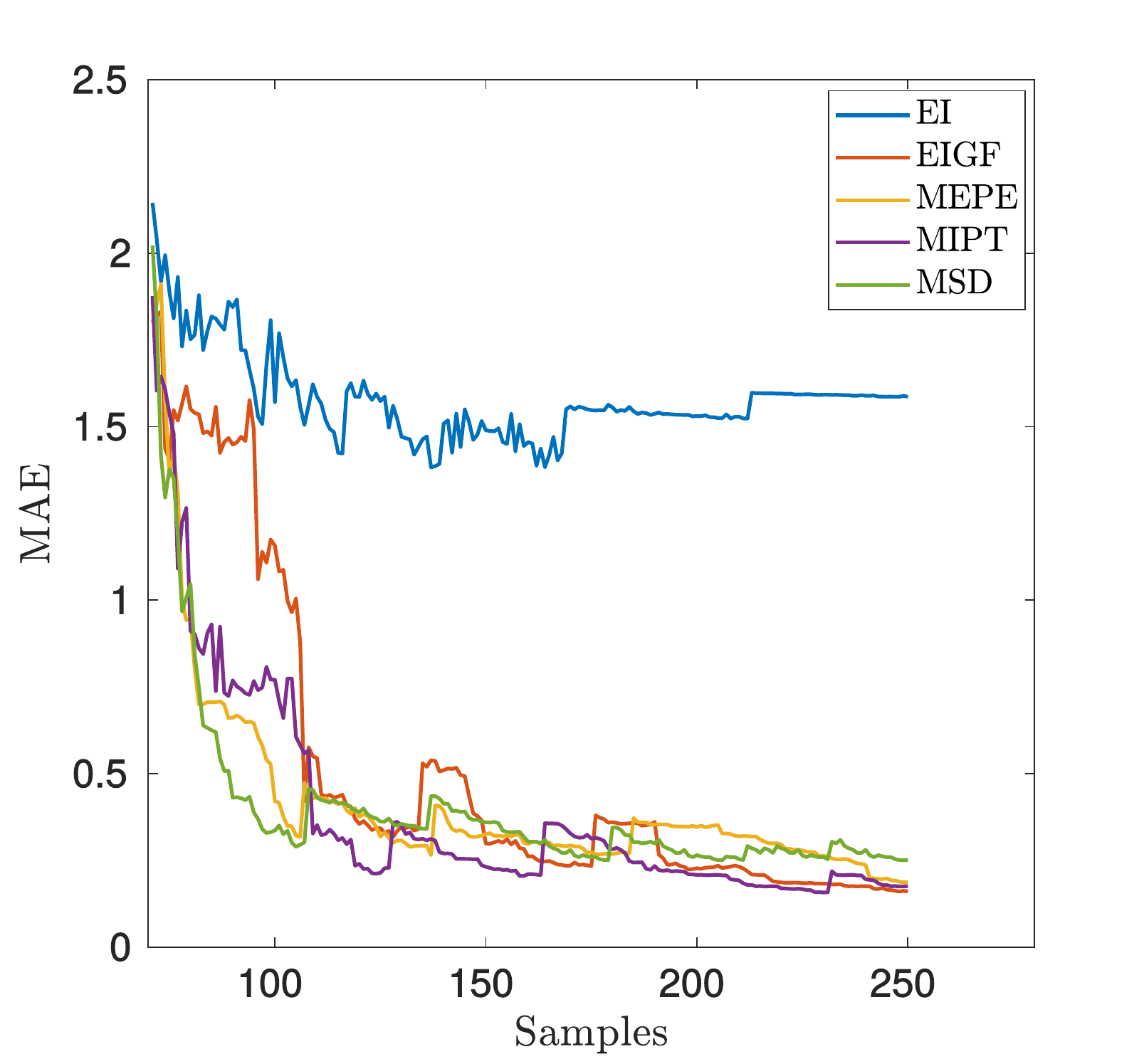}
\caption[Convergence of the MAE error for $\mathcal{M}_{Levy}^{7d}$ until 250 samples]{Convergence of the MAE error for $\mathcal{M}_{Levy}^{5d}$ until 250 samples.}\label{fig::Levy_Conv}
\end{figure}

The next benchmark function is the 7-dimensional form of the Lévy function and was for example utilized in \cite{laguna2005experimental}.
The function is of the form
\begin{equation}
\begin{aligned}
\mathcal{M}_{Levy}^{7d}(\bm{x}) &= \sin^{2} (\pi w_{1}) + \sum_{i=1}^{7} (w_{i}-1)^{2} \left[ 1 + 10 \sin^{2}(\pi w_{i} + 1) \right] \\&+ (w_{7} - 1)^{2} \left[1 + \sin^{2} (2 \pi w_{7}) \right],
\end{aligned}
\end{equation}
where $w_{i} = 1 + \frac{x_{i} - 1}{4}$, for all $i=1, \ldots, 7$.
The input domain for all $x_{i}$ is [-2,2].The function has a global minimum $\mathcal{M}_{Levy}^{7d}(\bm{x}_{min}) = 0$ at $\bm{x}_{min} = (1, \ldots, 1)^{T}$. To put the following error measures into perspective the spread of the function in the specified domain is around $17.5$.\\
The average results of the error measures after 150 samples when starting from 70 samples for 10 adaptive sampling techniques are listed in Table \ref{table::Levy150}. It can be noticed that all sampling techniques except for EI reduce the error significantly with respect to the one-shot TPLHD metamodel and the initial model. 
The space-filling algorithm MIPT yields the best results followed by SSA.
\begin{table}[b!]
\begin{center}
\resizebox{1.0\textwidth}{!}{%
\begin{tabular}{l|l c c c c} \hline
& Method & MAE & RMAE & RMSE & R$^{2}$  \\ \hline\hline \\
\multirow{1}{*}{\shortstack[l]{Errors after\\100 samples}} & TPLHD & 1.9768 & 2.9701 & 2.52960 & 0.44031 \\\\  \hline \\
\multirow{12}{*}{\shortstack[l]{ Errors after\\150 samples}} &TPLHD &1.0392 & 1.3564 & 1.2974 & 0.8527\\
& ACE & 0.4014 & 0.6603 & 0.5157 & 0.9767  \\
&CVVor & 0.3259 & 0.4516 & 0.4259 & 0.9892    \\ 
&EI &  1.4894 & 2.5317 & 1.9733 & 0.6594 \\
&EIGF &  0.2984 & 0.4645 & 0.3827 & 0.9871 \\
&MASA & 0.4466 & 0.7158 & 0.5613 & 0.9724  \\
&MEPE & 0.3256 & 0.4425 & 0.4067 & 0.9855 \\
&MIPT &  \textbf{0.2314} & \textbf{0.3040} & \textbf{0.2888} & \textbf{0.9927} \\ 
&MSD & 0.4487 & 0.7740 & 0.5614 & 0.9724 \\
&MSE &0.3268 & 0.4979 & 0.4064 & 0.9855   \\
&SFCVT &0.3003 & 0.5501 & 0.3915 & 0.9865 \\ 
&SSA &  0.2915 & 0.4137 & 0.3616 & 0.9885
\end{tabular}
}
\end{center}
\caption[Error measures for $\mathcal{M}_{Levy}^{7d}$ after 150 samples]{Error measures for $\mathcal{M}_{Levy}^{7d}$ after 150 samples. }\label{table::Levy150}
\end{table}
Since the the problem is seven-dimensional, additional samples with regard to the results of Table \ref{table::Levy150} are only found with the 5 most time-efficient methods (EI, EIGF, MEPE,MIPT and MSD). \\ The error values for these five methods after 250 samples are shown in Table \ref{table::Levy250}. 
EI is clearly the worst and yields less proficient results than the one-shot method. However, all other methods reduce the error significantly with EIGF approximating the shape of the function best (judged by MAE and RMSE) whereas MEPE reduces the absolute error most noticeable.
\begin{table}[t!]
\begin{center}
\resizebox{1.0\textwidth}{!}{%
\begin{tabular}{l|l c c c c} \hline
& Method & MAE & RMAE & RMSE & R$^{2}$ \\ \hline\hline \\
\multirow{1}{*}{\shortstack[l]{Errors after\\100 samples}} & TPLHD & 1.9768 & 2.9701 & 2.52960 & 0.44031 \\\\  \hline \\
\multirow{6}{*}{\shortstack[l]{ Errors after\\250 samples}} &TPLHD &0.9310&1.2336 & 1.1615 & 0.8820 \\
&EI & 1.5860 & 3.1906 & 2.1284 & 0.6037  \\
&EIGF &  \textbf{0.1602} & 0.2934 & \textbf{0.2116} & 0.9944  \\
&MEPE &  0.1872 & \textbf{0.2239} & 0.2380 & \textbf{0.9981}\\
&MIPT & 0.1754 & 0.2806 & 0.2226 & 0.9956  \\ 
&MSD & 0.2510 & 0.2830 & 0.2899 & 0.9957   
\end{tabular}
}
\end{center}
\caption[Error measures for $\mathcal{M}_{Levy}^{7d}$ after 250 samples]{Error measures for $\mathcal{M}_{Levy}^{7d}$ after 250 samples. }\label{table::Levy250}
\end{table}
The convergence behavior of the MAE error of the 5 techniques is illustrated in Figure \ref{fig::Levy_Conv}. It can be seen that all methods except EI show a sufficient decreasing rate. However after 150 samples the error is not reducing with the same rate and is more or less stagnating.
\clearpage
\section{Concluding remark}
The introduced adaptive sampling techniques have been tested on different benchmark problems in the OK framework. An analysis of the computation time has been employed. It was found that five techniques (EI, EIGF, MEPE, MIPT and MSD) can be used in higher dimensionality $\geq 7$ in order to efficiently add new sample points to the DOE in an efficient manner. CVD yields the best results for the one-dimensional case. In the higher dimensions MEPE shows the most proficient balance between exploration and exploitation and hence can be used as a versatile adaptive sampling scheme for black-box functions.
\chapter{Investigation of a dynamic problem utilizing a Kriging approach}\label{chapt::DynamicApproach}
Parametric studies for dynamic systems are of high interest to detect the risks of instabilities. The ability of Kriging with adaptive sampling is for such application is herein investigated, 
An indicator for the system chaotic motion is presented. It is found that for the given application the presented adaptive sampling techniques are not sufficient. Therefore an adaptive method for classification with Kriging is suggested, which to the best of the authors knowledge is a novelty. 
\section{Mathematical modeling of the dynamic system}\label{sec::MathematicalModelOfSystem}
\begin{figure}[ht!]
\centering
  \includegraphics[scale=0.45]{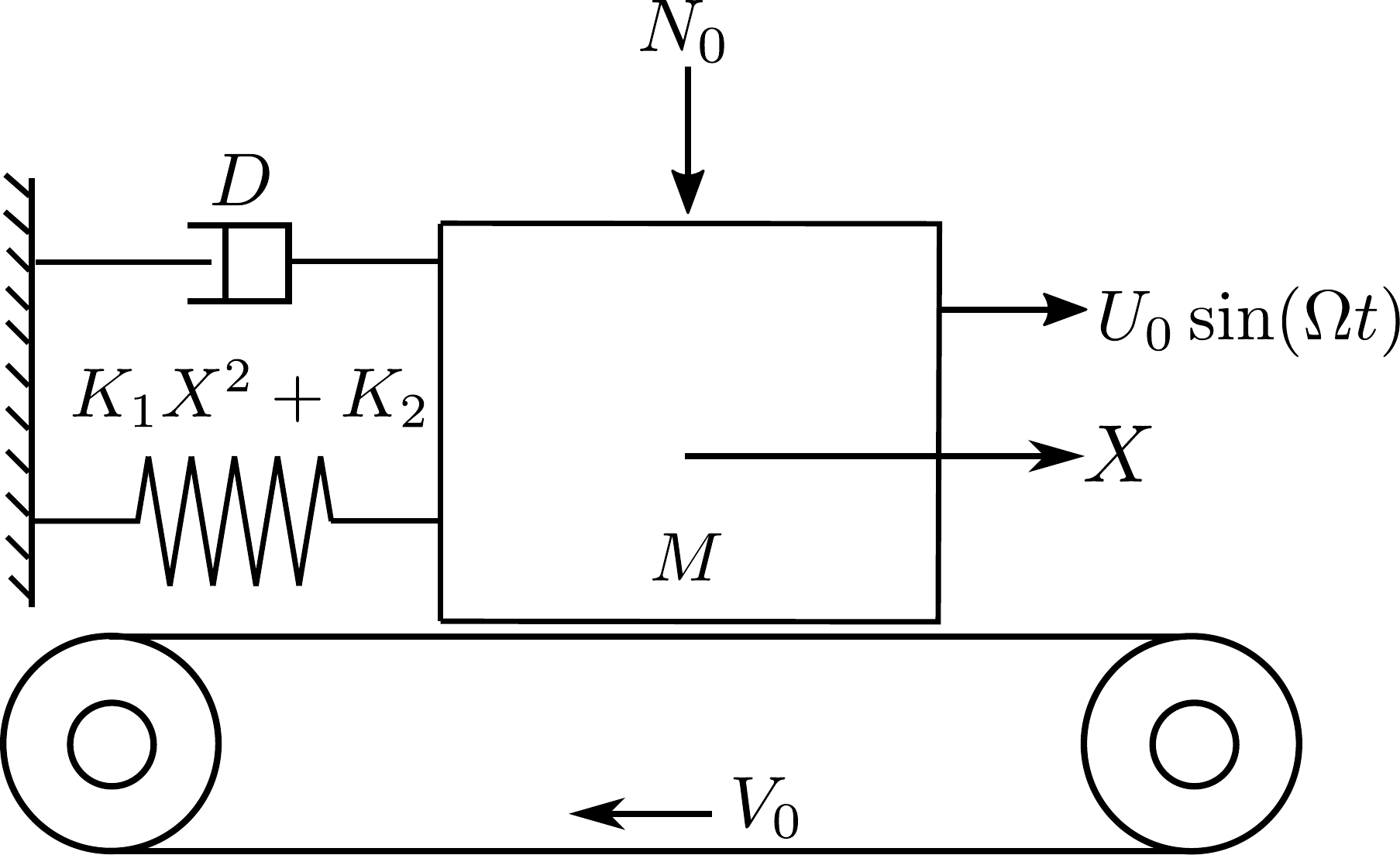}
\caption[Scheme of the analyzed nonlinear mass-on-belt system]{Scheme of the analyzed nonlinear mass-on-belt system.}
\label{fig:Application}       
\end{figure}

Friction can be the cause of dynamic instabilities leading to self-excited vibrations which can result in a performance decrease of mechanical systems. Dry friction refers to two solid bodies in contact. Oscillating systems excited by dry friction are frequently encountered in many practical applications, including brake squeal systems \citep{barton2004braking} and hydraulic cylinders \citep{owen2003reduction}.  \\
Linear spring mass systems placed on a moving belt, commonly referred to as mass-on-belt systems, are generally utilized as a mathematically simplified version of a stick-slip system, see e.g. \cite{stelter1990stick}, \cite{hinrichs1997dynamics}, \cite{galvanetto1999dynamics} or \cite{jimenez2007two}. Nonlinear systems of this type have yet to be analyzed to the same extent \citep{devarajan2017analytical}. However, the practical applications e.g. brake systems contain inherent nonlinearities. Therefore studies of these systems are of importance. \\
The mass-on-belt system with additional nonlinearities for its bifurcation behavior is investigated in \cite{santhosh2016discontinuity,awrejcewicz2005stick}.
\cite{devarajan2017analytical} studied a Duffing's type oscillator analytically in order to obtain expressions for stick–slip and pure-slip vibration amplitudes and frequencies.
An estimation method for the spectrum of Lyapunov exponents proposed by \citep{balcerzak2018spectrum} is employed in \cite{pikunov2019numerical} to analyze the stability of a nonlinear mass-on-belt system.
The mechanical problem investigated in this paper is the oscillator of  Duffing's type as schematized in Figure \ref{fig:Application} with a damping term. A mass $M $ is placed upon a moving belt with constant velocity $V_{0}$. The displacement of the mass over the time $t$ is represented by $X(t)$. The movement of the mass is restricted by a nonlinear spring of stiffness $K_{1} X^{2} + K_{2}$ and a dashpot with the damping coefficient $D$ parallel to
the spring.
The relative velocity between the belt and the body is denoted by $V_{R}(t) = V_{0} - \dot{X}(t) $. 
A time-dependent harmonic force $U(t) = U_{0} \sin (\Omega t)$ with the amplitude $U_{0} $ and the angular frequency $\Omega $ as well as a normal load $N_{0}$ are applied on the mass. 

\begin{figure}[hbtp]
\centering
\includegraphics[scale=0.5]{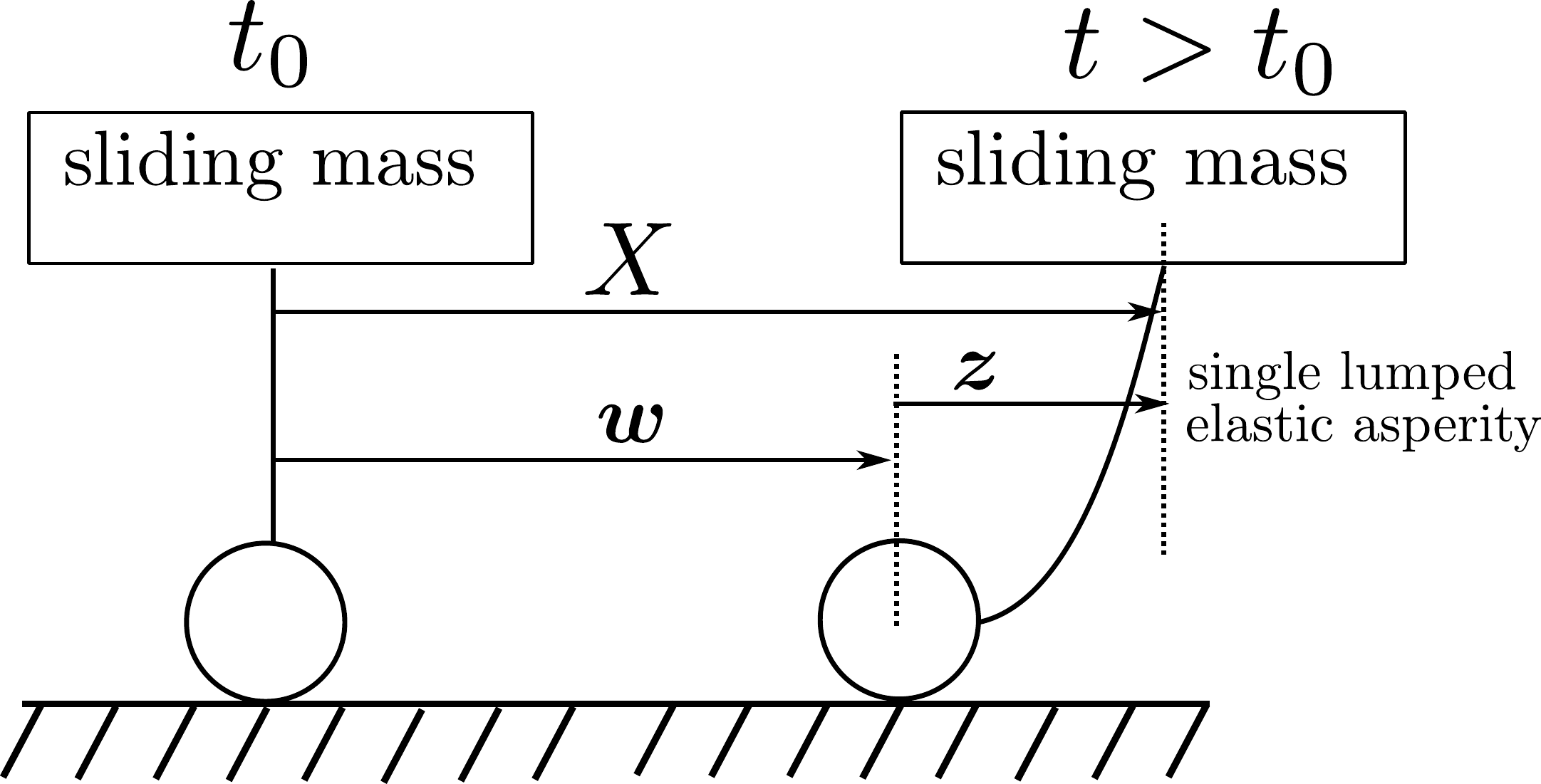}
\caption[Physical analogy of elasto-plastic model]{Physical analogy of elasto-plastic model. Block moves over surface. Displacement $X$ can be broken down into elastic component $z$ and plastic part $w$ (modified from \cite{dupont2000elasto})}\label{fig::elasto-plastic}
\end{figure}

The equation of motion of the system reads 
\begin{equation}
\begin{aligned}
M \ddot{X}(t) = - D \dot{X}(t)  - K_{1} X^{3}(t) -K_{2} X(t)   + F_{R}(V_{R}) + U_{0} \sin(\Omega t) \, \text{,}
\end{aligned}
\end{equation}
where $F_{R}(V_{R}) = N_{0} f_{R}(V_{R})$ is the friction force. $f_{R}(V_{R})$ is called the friction function which denotes the friction force per unit of normal load and is a user-chosen function. An overview of common choices is presented in \cite{pennestri2016review}. Generally, static and dynamic friction force models can be distinguished \citep{olsson1998friction}, where in a dynamic model the friction force does not only depend on the relative velocity between two bodies but also on other state variables.
Herein, a dynamic model called the
elasto-plastic model is employed as found in \cite{dupont2000elasto} and \cite{dupont2002single}. Elasto-plastic models were first introduced by 
\cite{prandtl1924spannungsverteilung} to model the behavior of solids subjected to stress. Here, the displacement $X$ of the body is the sum of an elastic (memoryless) contribution denoted $z$ and a plastic (history-dependent) part $w$ such that
\begin{equation}
X = z + w \, \text{.}
\end{equation} 
During the sticking phase, the plastic displacement is constant. In contrast, while slipping the elastic component remains fixed. A physical analogy of this model is depicted in Figure \ref{fig::elasto-plastic}.

\begin{figure}[hbtp]
\centering
\includegraphics[scale=0.5]{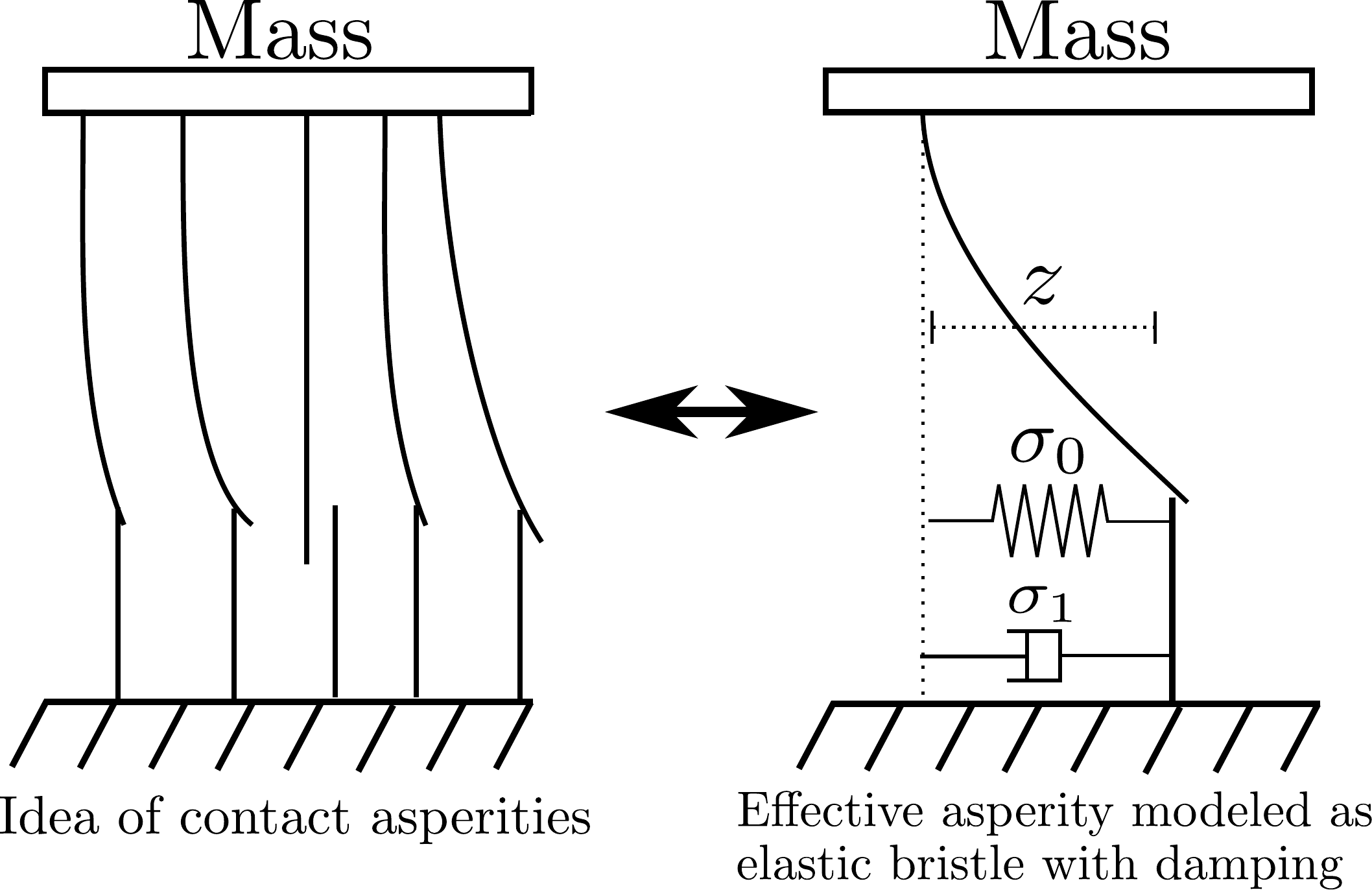}
\caption[Basic concept of friction force formulation]{Basic concept of friction force formulation. Asperities in the contact surface are modeled as elastic bristle with damping. For simplicity the LuGre-model takes an average bristle deflection $z$ for the determination of the friction force.}\label{fig::bristle}
\end{figure}

Here, due to the deformation of a single lumped elastic asperity or bristle the mass is subject to the friction force $F_{R}$. 
The state history of this model is only characterized by the state variable $z$, which in the physical analogy is the deflection of the bristle. 
The friction function is defined as
\begin{equation}
\begin{aligned}
f_{R}(V_{R},z)= \sigma_{0} z + \sigma_{1} \dot{z} + \sigma_{2} V_{R},
\end{aligned}
\end{equation}
which can be compared to the force model of the classical LuGre model of \cite{de1995new}. $\sigma_{0}$ is the bristle stiffness, $\sigma_{1}$ is the average bristle damping coefficient and $\sigma_{2}$ is a viscous component of the friction force. The concept is illustrated in Figure \ref{fig::bristle}. 

\begin{figure}[htp]
\centering
\begin{subfigure}[t]{0.5\textwidth}
\includegraphics[scale=0.35]{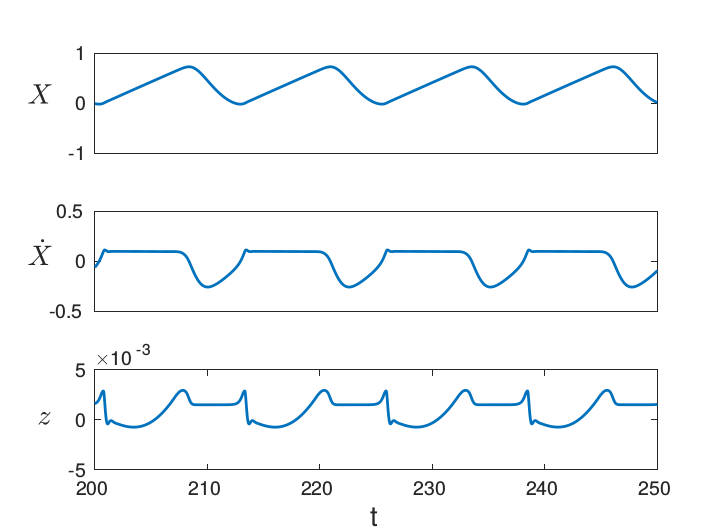}
\subcaption{}\label{fig:TimeResponse}
\end{subfigure}%
\begin{subfigure}[t]{0.5\textwidth}
 \includegraphics[scale=0.35]{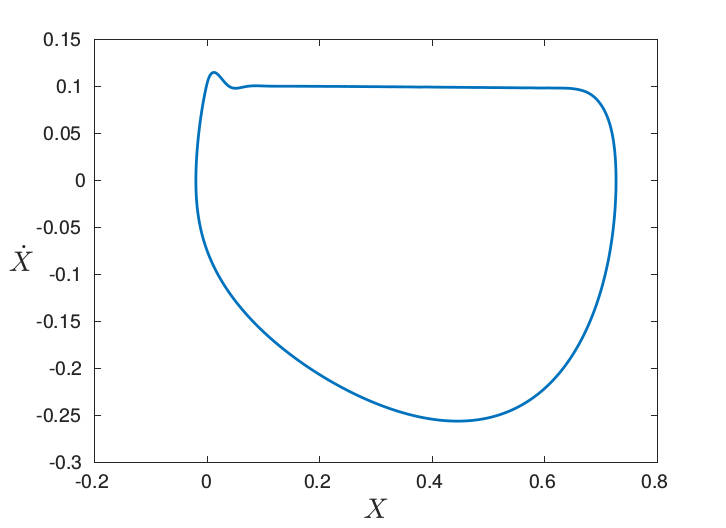}
\subcaption{}\label{fig:PhaseResponse}
\end{subfigure}%
   \caption[Time response and phase-space for application]{Time response and phase space for the nonlinear mass-on-belt system with $\Omega = 0.5 \, \text{rad}/\text{s}$,
$M = 1 \, \text{kg}$,
$V_{0} = 0.1 \, \text{m}/\text{s}$,
$D = 0.0 \, \text{Ns}/\text{m}$, 
$K_{1} = 1\, \text{N}/\text{m}^{3}$ ,
$K_{2} = 0.0 \, \text{N}/\text{m}$, 
$\mu_{s} = 0.3$,
$\mu_{k} = 0.15$,
$V_{s} = 0.1 \, \text{m}/\text{s}$,
$U_{0} = 0.1 \,\text{N}$,
$N_{0} = 1.0 \, \text{N}$,
$\sigma_{0} = 100.0 \, \text{N}/\text{m} $,
$\sigma_{1} = 10.0 \,  \text{Ns}/\text{m}$ and
$\sigma_{2} = 0.1 \, \text{Ns}/\text{m}$.  $X$ given in $m$. $\dot{X}$ in $m/s$ and the time $t$ in seconds.}\label{fig:TimeAndPhase}
\end{figure}

The velocity of bristle deflection has been defined by  \cite{dupont2000elasto} as
\begin{equation}
\begin{aligned}
\dot{z} = \left(1 - \alpha(z, V_{R}) \dfrac{\sigma_{0}}{g(V_{R})} z \cdot \text{sgn} \left( V_{R} \right) \right) V_{R} \, \text{,}
\end{aligned}
\end{equation}
where $\alpha(\bm{z}, \bm{\dot{g}}_{T})$ is a state variable, which depends on the relative velocity and is utilized to capture the stiction effect see e.g. \cite{townsend1987effect}. The function $\alpha$ defines the elastic deformation until the breakaway force of the system is exceeded, i.e.
\begin{equation}
\begin{aligned}
\alpha(\bm{z}, \bm{\dot{g}}_{T}) = \begin{cases}
\alpha(\bm{z}) , & \text{if} \,  \bm{\dot{g}}_{T} \cdot \bm{z} \geq 0 \\
0 , & \text{if} \,  \bm{\dot{g}}_{T} \cdot \bm{z} < 0 \, \text{,}
\end{cases}
\end{aligned}
\end{equation}
where
\begin{equation}
\begin{aligned}
\alpha(z)= \begin{cases}
0 , & \text{if} \, 	 z \leq z_{ba}\\
\frac{1}{2} \left( \sin \left( \pi \dfrac{\norm{z}  - \frac{z_{max} + z_{ba}}{2} }{z_{max} - z_{ba}} \right) + 1 \right) , & \text{if} \,  z_{ba} < \norm{ z} < z_{max}  \\
1 , & \text{if} \,  z_{max} \leq \norm{ z } \, \text{,}
\end{cases}
\end{aligned}
\end{equation}
with $z_{max}$ representing the maximum bristle deflection and $z_{ba}$ the bristle deflection for the breakaway condition.
The function $g(V_{R})$ models the Stribeck-effect, which describes an increase of the friction coefficient with increasing relative contact velocity \cite{stribeck1902wesentlichen}.
It is given by
\begin{equation}
g(V_{R}) = N_{0} \left( \mu_{k} + (\mu_{s}-\mu_{k}) \exp \left( - \frac{V_{R}^{2}}{V_{S}^{2}} \right)   \right),
\end{equation}
where $V_{S}$ is a parameter called the characteristic Stribeck velocity, and $\mu_{s}$ and  $\mu_{k}$ denote the static and kinetic friction coefficients respectively.
Therefore, the complete equation of motion of the system in state-space form with the modified elasto-plastic friction model reads 
\begin{equation}\label{eq::dynamic_system}
\begin{cases}
\frac{dX}{dt} &= \dot{X} \\
\frac{d^{2} X}{d^{2} t} &= - \frac{D}{M} \dot{X}  - \frac{K_{1}}{M} X^{3} - \frac{K_{2}}{M} X   +  \frac{N_{0}}{M} f_{R}(V_{R}) + \frac{U_{0}}{M} \sin(\Omega t) \\
\frac{d z}{d t} &= \left(1 - \alpha(z, V_{R}) \dfrac{\sigma_{0}}{g(V_{R})} z \cdot \text{sgn} \left( V_{R} \right) \right) V_{R} \, \text{,}
\end{cases}
\end{equation}
which can be solved e.g. via numerical integration. Knowing the equations of motion, the next step is to analyze the dynamic behavior of the system. An example of the time-response and a phase-plot are displayed in Figure \ref{fig:TimeAndPhase}. 

This dynamic system is studied with the help of a surrogate model to evaluate at low-cost the critical values. Hereby two different values are investigated:
\begin{enumerate}
\item the sticking time of the system,
\item the \acrfull{lle} in order to characterize instability.
\end{enumerate} 
The LLE is investigated in section \ref{sec::SurModelForLLE}.
The importance of the sticking time is explained in the next section.
\clearpage
\section{Surrogate model for the Sticking time}
Qualitatively, the sticking and slipping mode define the oscillation of a dry-friction oscillator. 
\cite{lima2015stick} point out that in order to reduce the sticking time and therefore improve properties e.g. endurance of mechanical systems the role of the acting parameters needs to be determined. These can then be used to find countermeasures for systems where undesirable stick occurs. This can be used for example to the application described by \cite{navarro2004practical} and later \cite{navarro2007sliding}, where the authors model a generic oilwell drillstring with multiple degrees of freedom as commonly found in the petroleum extraction industry. Sticking of the drillstring is the origin of problems such as drill pipe fatigue or components failure. Therefore a reduction of the harmful sticking time has an economic interest. However as \cite{lima2015stick} point out the duration of the sticking mode and the cumulative sticking time has not got the needed attention in the literature. The adaptive sampling techniques are herin studied for their use in generating a surrogate model for the sticking time of the nonlinear mass-on-belt problem.
\begin{figure}[h!]
\centering
\begin{subfigure}[t]{0.5\textwidth}
\includegraphics[scale=0.35]{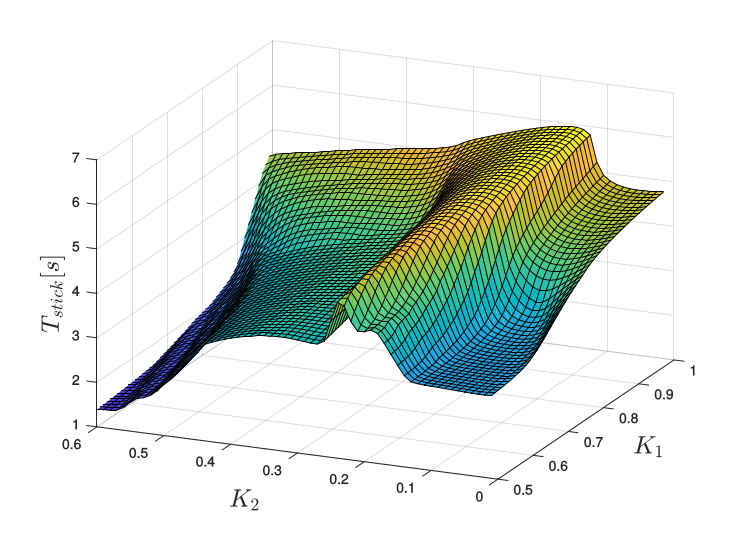} 
\subcaption{Sticking time}\label{fig:STickingKSeta06Plot}
\end{subfigure}%
\begin{subfigure}[t]{0.5\textwidth}
\includegraphics[scale=0.35]{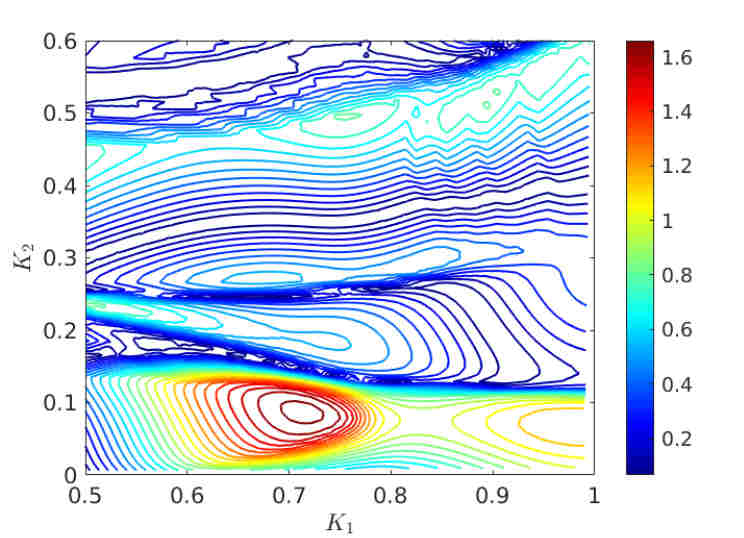} 
\subcaption{Contour of absolute error}\label{fig:STickingKSeta06Error}
\end{subfigure}
\caption[Sticking time and absolute error of the evaluation of the sticking time]{Sticking time and absolute error of the evaluation of the sticking time. $K_{1}$ is given in $\text{N}/\text{m}^{3}$, $K_{2}$ in $\text{N}/\text{m}$ and the absolute error in seconds.}\label{fig:STickingKSeta06}
\end{figure}

\begin{figure}[t!]
\centering
\includegraphics[scale=0.5]{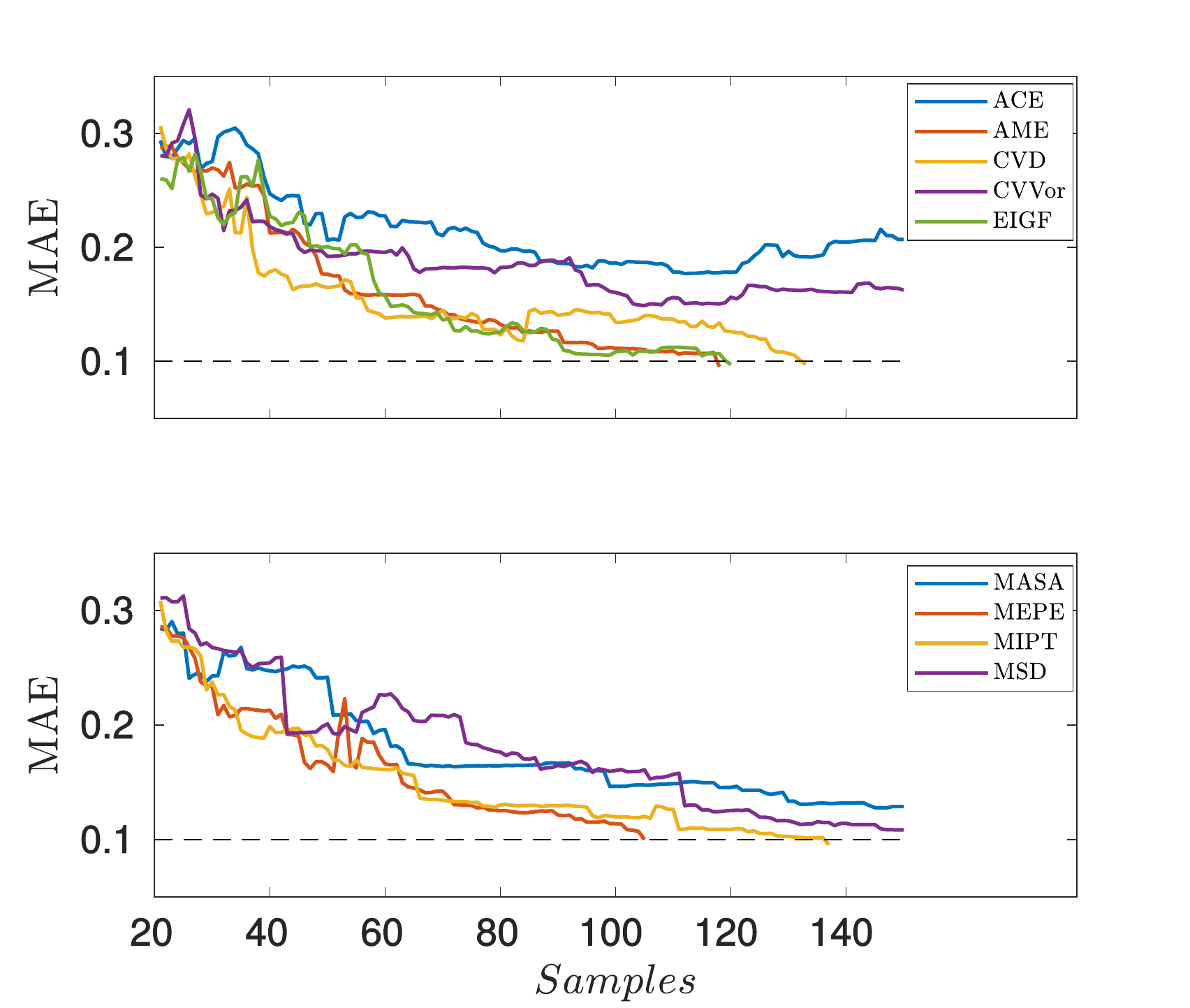}
\caption[MAE error over the sample size for the sticking time problem]{MAE error over the sample size for selected adaptive sampling techniques for the sticking time until 150 samples.}\label{fig::KSetaConvergence}
\end{figure}

\begin{figure}[h!]
\centering
\begin{subfigure}[t]{0.5\textwidth}
\includegraphics[scale=0.4]{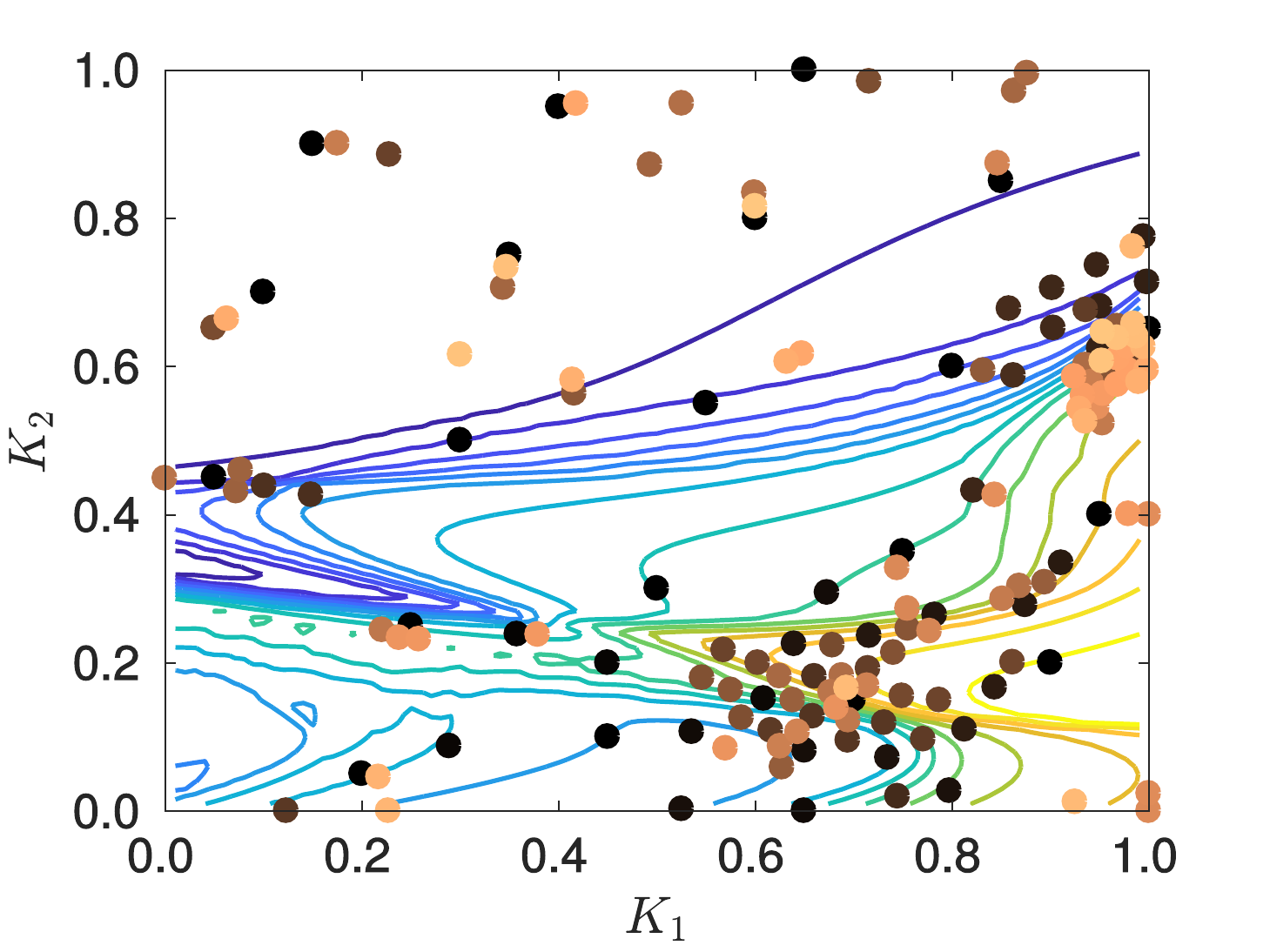} 
\subcaption{ACE - 150 samples}\label{fig:TStickKSeta06NoACE}
\end{subfigure}%
\begin{subfigure}[t]{0.5\textwidth}
\includegraphics[scale=0.4]{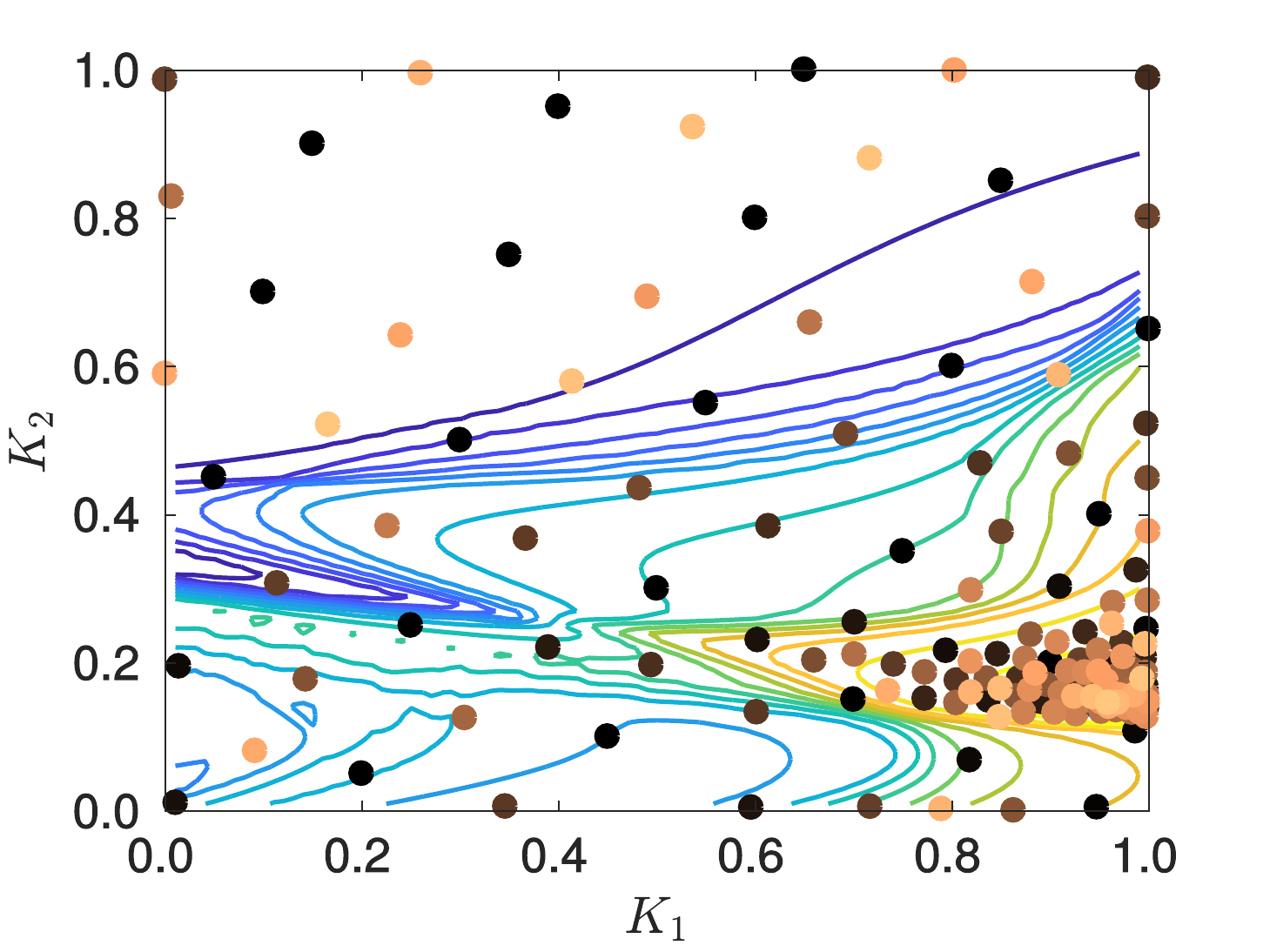} 
\subcaption{MASA - 150 samples}\label{fig:TStickKSeta06NoMASA}
\end{subfigure}
\caption[Sample positions and average number of samples for sticking sticking time problem unable to reach error threshold]{Sample positions and average number of samples for methods unable to reach the error threshold for the sticking time problem.}\label{fig:TStickKSeta06No}
\end{figure}

In the following sticking of the mass on the belt is defined when the relative velocity $V_{R}(t) = V_{0} - \dot{X}(t)$ is below the threshold of $10^{-4}m/s$. The sticking time is evaluated by numerically integrating equation (\ref{eq::dynamic_system}) with a six-stage, fifth-order, Runge-Kutta method with variable step-size until 250 seconds. In order to avoid transient behavior the sticking time is the cumulative time from $150$ to $250$ seconds with $V_{R} < 10^{-4}m/s$. \\
The parametric domain for the sticking time 
is given by the two spring stiffnesses with $K_{1} \in  [0.5 , 1.0]\, \text{N}/\text{m}^{3}$
and $K_{2} \in [0.0,0.6] \, \text{N}/\text{m}$. 
The rest of the parameter values are given by
$M = 1 \, \text{kg}$,
$V_{0} = 0.1 \, \text{m}/\text{s}$,
$D = 0.0 \, \text{Ns}/\text{m}$, 
$\mu_{s} = 0.3$,
$V_{s} = 0.1 \, \text{m}/\text{s}$,
$U_{0} = 0.1 \,\text{N}$,
$N_{0} = 1.0 \, \text{N}$,
$\sigma_{0} = 100.0 \, \text{N}/\text{m} $,
$\sigma_{1} = 10.0 \,  \text{Ns}/\text{m}$,
$\sigma_{2} = 0.1 \, \text{Ns}/\text{m}$, 
$\Omega = 0.6 \, \text{rad}/\text{s}$ and
$\mu_{k} = 0.15$. The sticking time over the input domain is plotted in Figure \ref{fig:STickingKSeta06Plot}. Initially 20 samples are spread across the domain with TPLHD leading to a metamodel with an absolute error as distributed in Figure \ref{fig:STickingKSeta06Error}.
It can be noticed that the spread of the sticking time in the parametric domain $\lbrace K_{1}, K_{2}\rbrace$ is around 5 seconds and the highest initial absolute error around 1.6 seconds. 
From an initial TPLHD model using 20 samples surrogate models utilizing 60 samples are built using 12 alternative sampling techniques. Furthermore a direct TPLHD model with 60 samples is generated.
The error measures for the different surrogate models are listed in Table \ref{table::StickingTime}. 
In the majority of the ten iterations EI, SFCVT and SSA provoke numerical issues because of clustering. 
It can be noticed that CVD yields the best MAE error value, whereas EIGF excels in the other criteria. All other techniques perform more proficient then the initial and final one-shot metamodels.
\begin{table}[b!]
\begin{center}
\resizebox{1.0\textwidth}{!}{%
\begin{tabular}{l|l c c c c} \hline
& Method & MAE[s] & RMAE & RMSE[s] & R$^{2}$  \\ \hline\hline \\
\multirow{1}{*}{\shortstack[l]{Errors after\\20 samples}} & TPLHD & 0.3070 & 1.2916 & 0.4569 & 0.8830\\ \\  \hline \\
\multirow{13}{*}{\shortstack[l]{ Errors after\\60 samples}} &TPLHD &0.2847 & 1.2684 & 0.3822 & 0.9251
\\
&ACE &0.2277 & 1.4008 & 0.3370 & 0.9364\\
&AME & 0.1586 & 1.6147 & 0.3142 & 0.9447\\
&CVVor & 0.1955 & 1.2527 & 0.3205 & 0.9424 \\
&CVD & \textbf{0.1380} & 1.2326 & 0.2412 & 0.9674\\
&EI& -  & - & - & -\\
&EIGF &  0.1576 & \textbf{0.9986} & \textbf{0.2205} & \textbf{0.9727} \\
&MASA &0.1959 & 1.6615 & 0.3152 & 0.9443 \\
&MEPE &0.1660 & 1.0896 & 0.2687 & 0.9595 \\
&MIPT &0.1610 & 1.680 & 0.3091 & 0.9464\\ 
&MSD & 0.2259 & 1.6384 & 0.3992 & 0.9107  \\
&SFCVT &  - & - & -&  -\\
&SSA &-   &-  &- & -
\end{tabular}
}
\end{center}
\caption[Error measures for the sticking time metamodel 60 samples]{Error measures for the sticking time metamodel 60 samples (methods with clustering problems are indicated by empty rows).}\label{table::StickingTime}
\end{table}
As a next step the average amount of samples needed to reduce the MAE error below $0.1s$ is studied. Here the limit is set to 150 points. \newpage Amongst the nine methods that do not provoke numerical issues five are able to reach this threshold within this limit. 
\begin{figure}[t!]
\centering
\begin{subfigure}[t]{0.5\textwidth}
\includegraphics[scale=0.4]{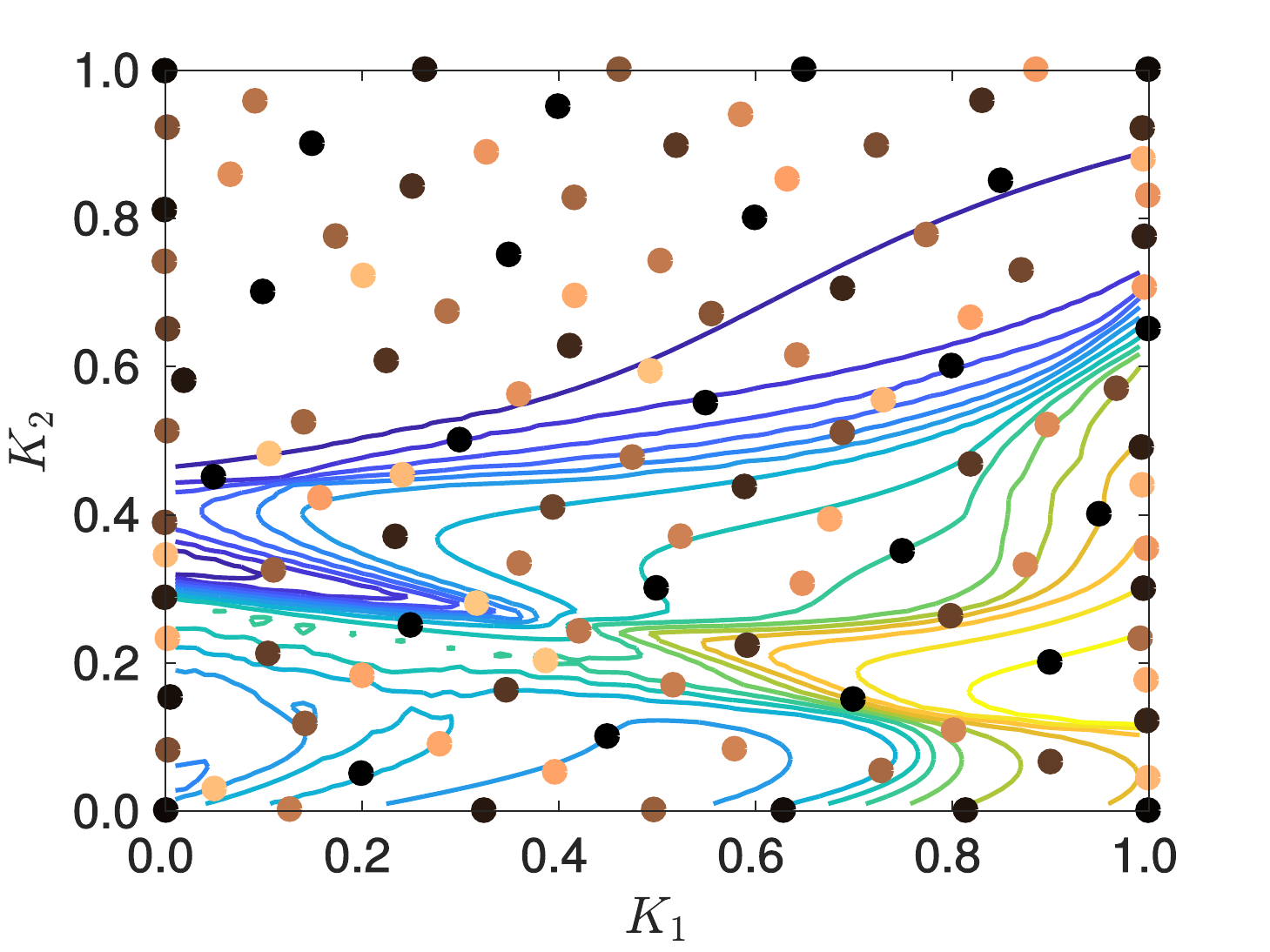} 
\subcaption{AME - 118 samples}\label{fig:TStickKSeta06YesAME}
\end{subfigure}%
\begin{subfigure}[t]{0.5\textwidth}
\includegraphics[scale=0.4]{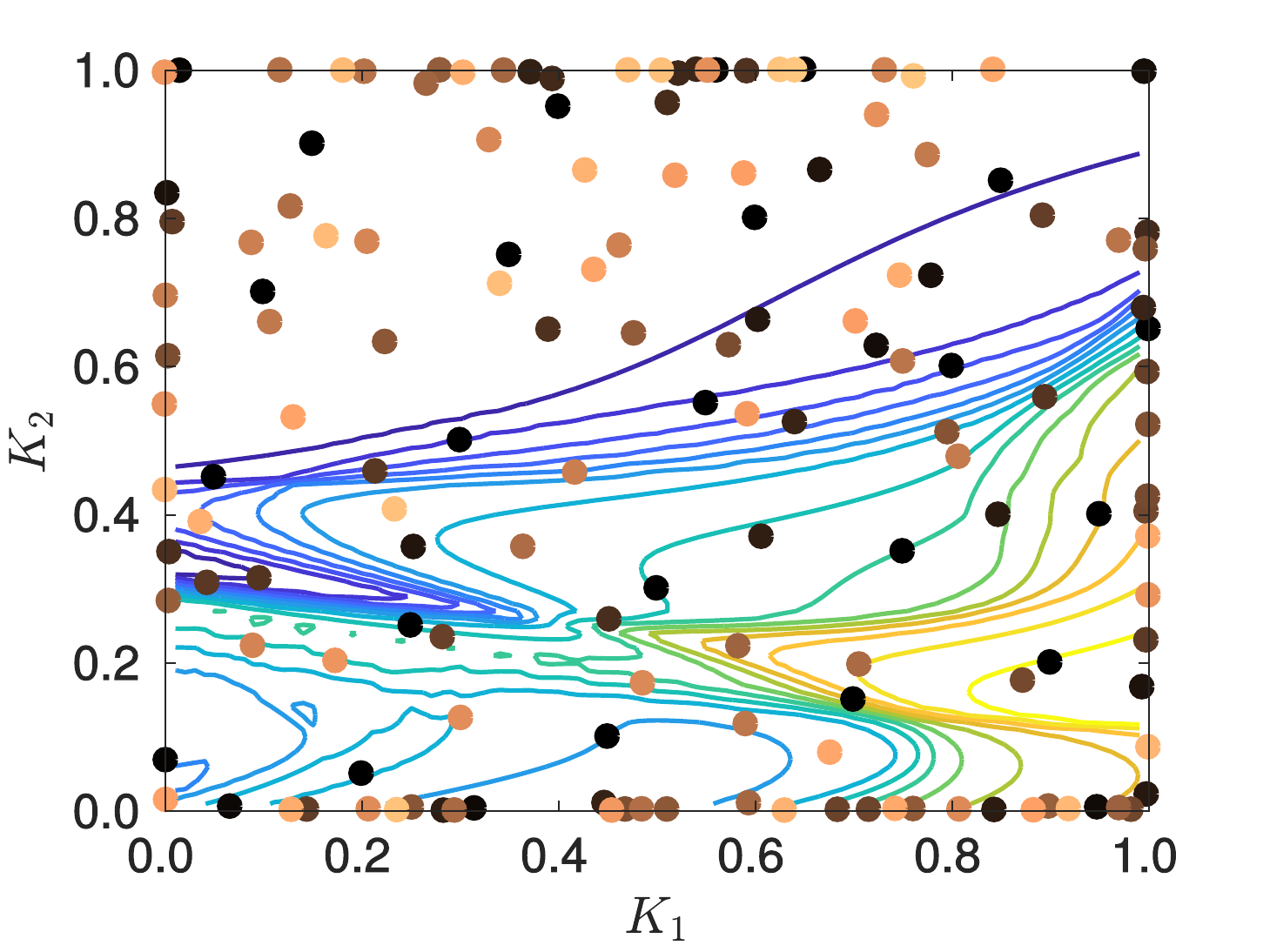}
\subcaption{CVD - 133 samples}\label{fig:TStickKSeta06YesMSE}
\end{subfigure}
\begin{subfigure}[t]{0.5\textwidth}
\includegraphics[scale=0.4]{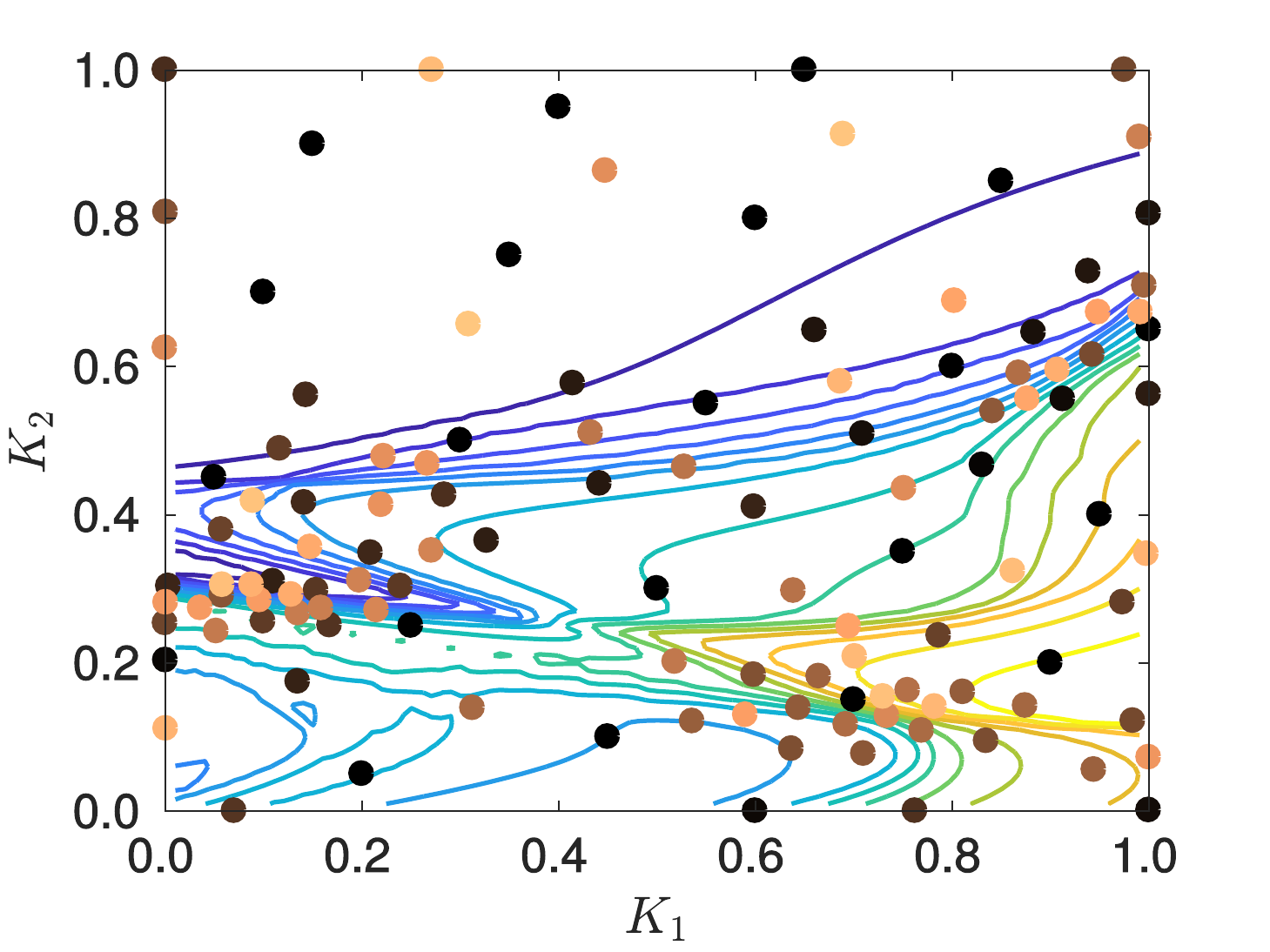}
\subcaption{EIGF - 120 samples}\label{fig:TStickKSeta06YesEIGF}
\end{subfigure}%
\begin{subfigure}[t]{0.5\textwidth}
\includegraphics[scale=0.4]{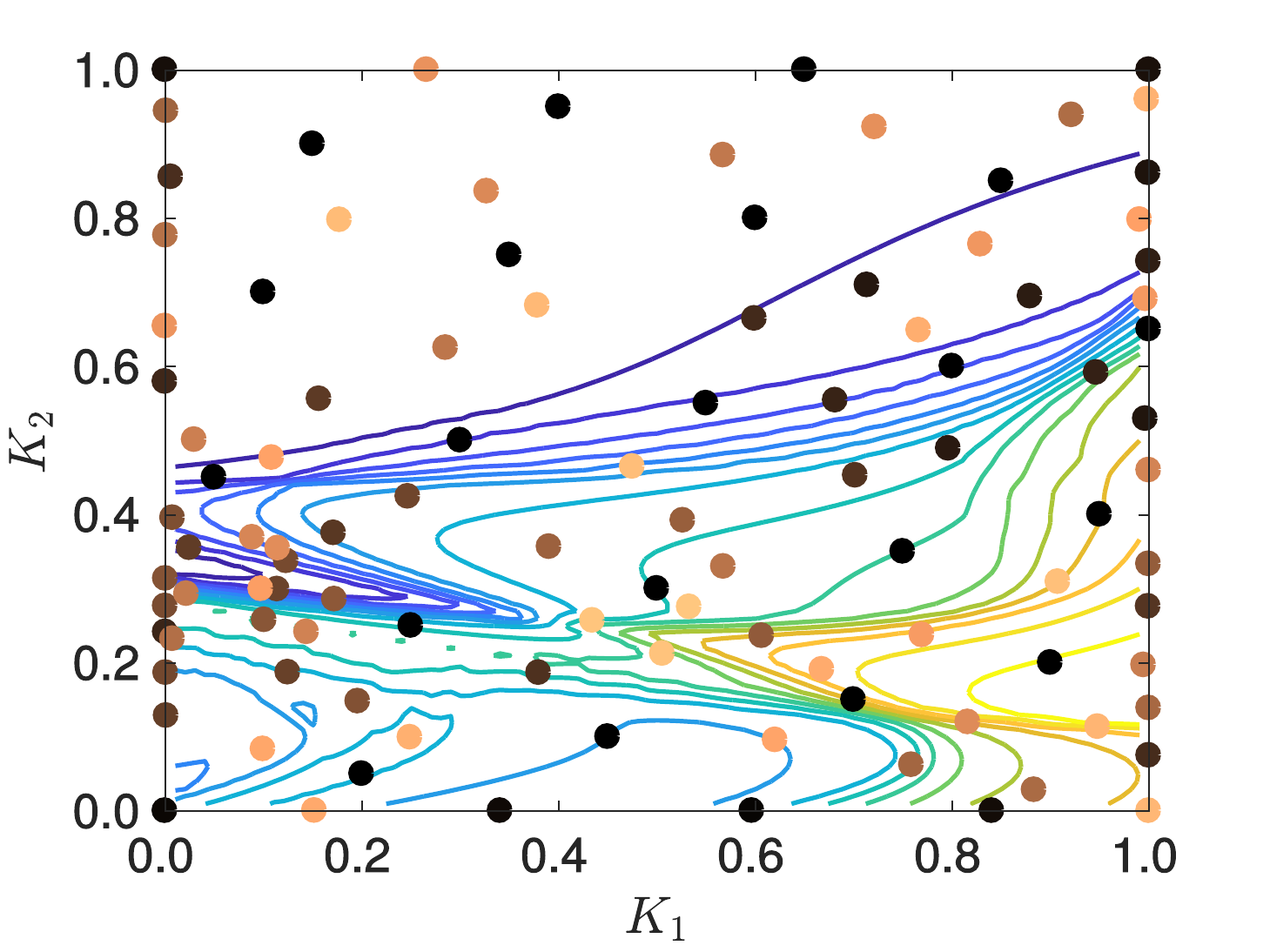}
\subcaption{MEPE - 105 samples}\label{fig:TStickKSeta06YesMEPE}
\end{subfigure}
\begin{subfigure}[t]{0.5\textwidth}
\includegraphics[scale=0.4]{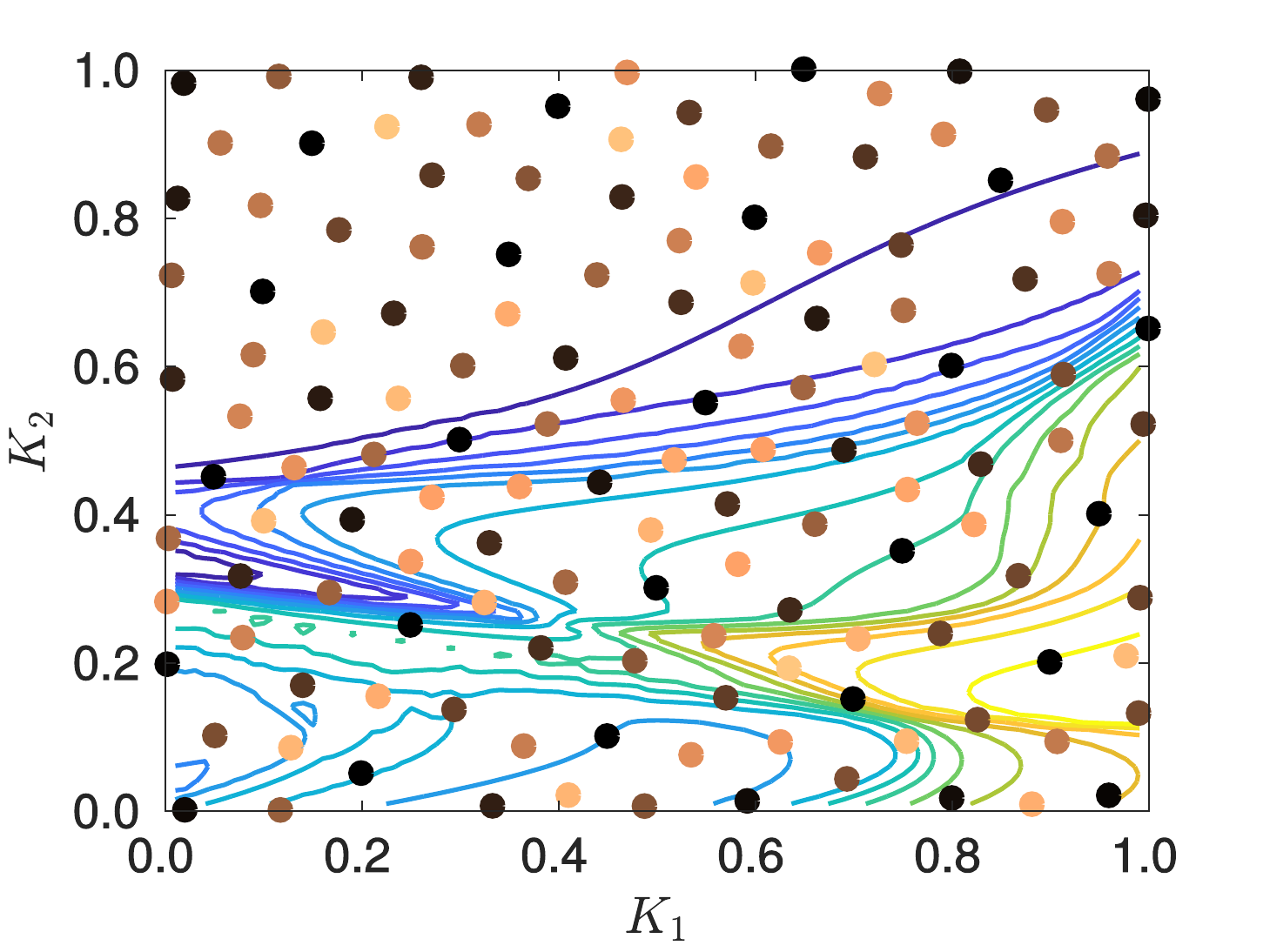}
\subcaption{MIPT- 117 samples}\label{fig:TStickKSeta06YesMIPT}
\end{subfigure}%
\caption[Sample positions and average number of samples for sticking sticking time problem]{Sample positions and average number of samples needed to reduce the reach MAE$<0.1$ for the sticking time problem.}\label{fig:TStickKSeta06Yes}
\end{figure}
This can be seen in the convergence of the MAE error over the sample size as given in Figure \ref{fig::KSetaConvergence}. 
The lowest average number of points is needed with MEPE ($105$ samples on average). The sample positions of the added points of the 5 successful methods as well as the number of points needed are shown in Figure \ref{fig:TStickKSeta06Yes}.\\ The later a point is added the more its color tends towards light red. MEPE as displayed in Figure \ref{fig:TStickKSeta06YesMEPE} over the contour of the target function balances exploration and exploitation most proficiently.
The sample locations of two of the methods (ACE and MASA) that need more than 150 samples to reach the threshold are shown in Figure \ref{fig:TStickKSeta06No}. It can be seen that these methods are clustering around regions of the input domain and therefore do not gain more information while increasing the number of samples.
\clearpage
\section{Surrogate model for classification of chaotic motion}\label{sec::SurModelForLLE}
\acrfull{le} introduced by \cite{oseledec1968multiplicative} are defined as the average divergence or convergence rate of nearby orbits in the state space. They are one of the most useful tools to characterize the stability of a dynamic system \cite{kocarev2006discrete}. 
A brief introduction to LE as well as suitable estimation methods for the problem at hand are given in the next section. 
Afterwards the ability of the presented adaptive sampling techniques to create a metamodel for this indicator is discussed and a new and more suitable sampling method is introduced and tested on different examples of data.
\subsection{Estimation method for Lyapunov exponent}\label{sec::EstimationLLE}
\begin{figure}[h!]
\centering
\includegraphics[scale=0.5]{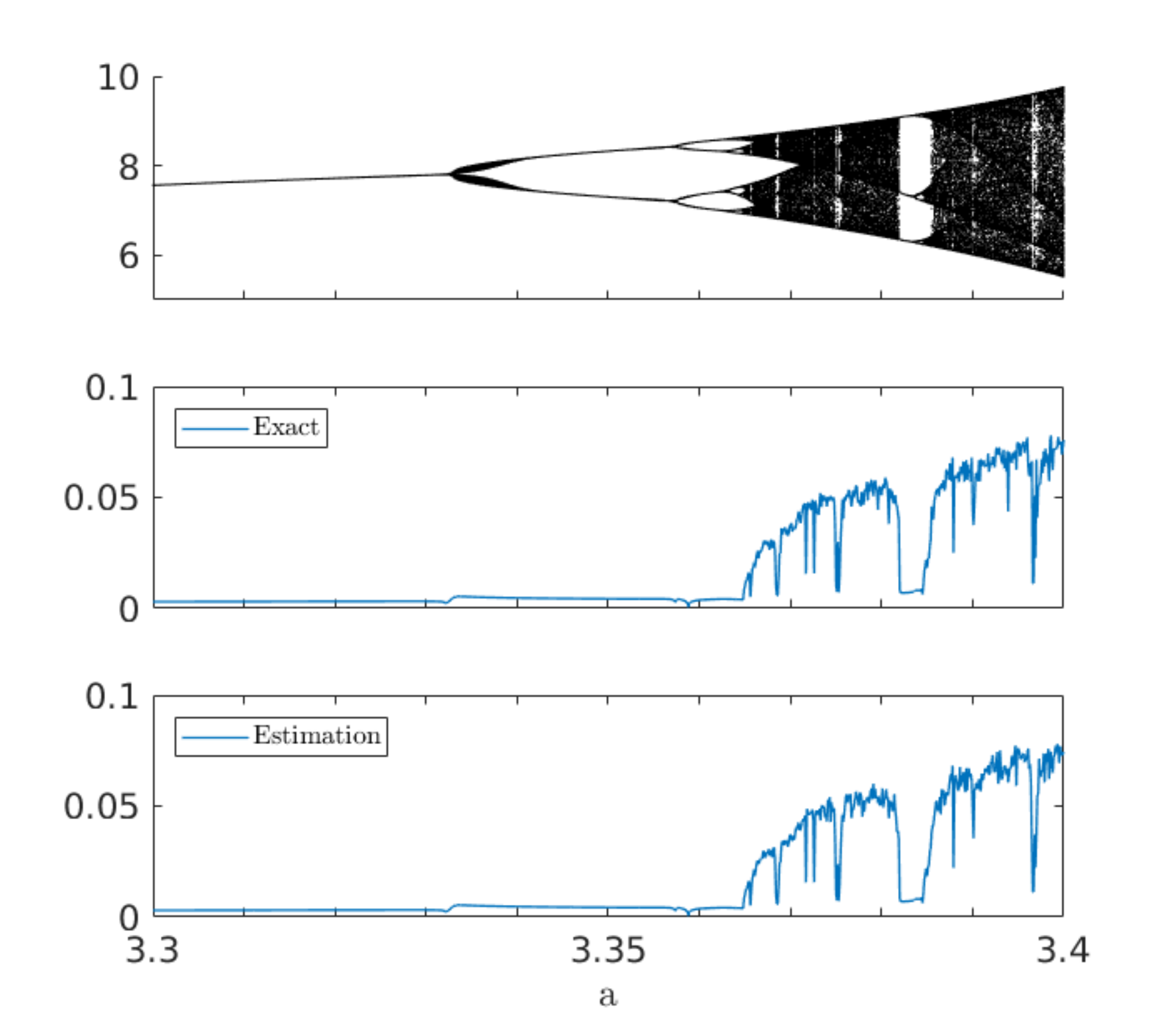}
\caption[LLE estimation technique for continuous problem]{Upper image bifurcation diagram of the problem of equation (\ref{eq::ProblemFromPaper}) over the parameter $a$. Middle image: LLE with exact Jacobi matrix for the equation. Lower image: LLE with utilized estimation method of the Jacobi matrix.}\label{fig:LLEProblemFromPaper}
\end{figure}

Consider a general equation of motion of an $N$-dimensional time-continuous system
can be recast into first order differential equation system of the form
\begin{equation}\label{eq::n_dim_system}
\dot{\bm{x}} = \bm{f} (\bm{x}) \, \text{,}
\end{equation}
with the state-space vector $\bm{x} \in \mathbb{R}^{N}$ and $\bm{f}$ being a set of $N$ functions i.e. $\bm{f}= \left[f_{1}(\bm{x}), \ldots f_{N}(\bm{x}) \right]$. \\
Consider an N-dimensional sphere of initial conditions. With progressing time the sphere evolves into an ellipsoid whose principal axes are contracting or expanding with rates governed by the spectrum of LEs, here denoted by $\left\lbrace \lambda_{i} \right\rbrace_{i \in \left[ 1,N\right] }$. A positive exponent indicates local instability in its particular direction and hence characterizes chaotic motion, see \cite{rosenstein1993practical}.
Therefore in practice and for the purpose of this thesis the computation of the largest Lyapunov exponent is sufficient.
The presence of a positive LLE indicates chaos, whereas negative values are characteristic for regular motion. Since the analytic calculation is mostly restricted to simple linear systems, LEs are usually determined with numerical methods see e.g. \cite{shimada1979numerical},\cite{benettin1980lyapunov}, \cite{kantz1994robust} or \cite{benettin1980lyapunov}. 
\\
When the equations of the analyzed system are available and the system is smooth with an analytically obtainable Jacobian matrix, the LLE can be calculated with e.g. the algorithm as given in \cite{wolf1986quantifying}, where the system of equations is solved for $N$ nearby initial conditions and correct state space orientation is maintained by repeatedly orthonormalizing the corresponding set of vectors using the Gram-Schmidt procedure. However, the dry friction problem at hand is a non-smooth system. The Jacobian matrix cannot be determined or is strongly ill-conditioned, which leads to significant numerical problems when employing the algorithm proposed by \cite{wolf1986quantifying}. 
To overcome this obstacle, a novel method has recently been presented in \cite{balcerzak2018spectrum}, which estimates the Jacobian matrix by a truncated Taylor series expansion with small orthogonal perturbations. Consider a discretization in time given by the set $\lbrace t_{i} | i=1, \ldots, n \rbrace$. Let $x(t_{i})$ be denoted by $x_{i}$.
The discrete dynamical system of the mapping $\bm{f}$ introduced in equation (\ref{eq::n_dim_system})
at time $t_{i}$ reads
\begin{equation}\label{eq::gen_form}
\bm{x}_{i+1} = \bm{G}(\bm{x}_{i}),
\end{equation}
where $\bm{G}(\bm{x}_{i}) = \left[G_{1}(\bm{x}_{i}), \ldots,G_{N}(\bm{x}_{i})  \right]$.
Because the operator $\bm{G}$ in equation (\ref{eq::gen_form}) may be discontinuous, an analytic expression for the Jacobian can not be given. However,
by introducing a perturbation vector
$\Delta_{i} = \left[ \delta_{1} , \ldots \delta_{N}  \right]^{T}$, where each element is a value of small magnitude, equation (\ref{eq::gen_form}) can be rewritten in two ways
\begin{equation}
\begin{aligned}
\bm{x}_{i+1} + \Delta_{i+1} &= \bm{G}(\bm{x}_{i}+ \Delta_{i}) \approx   \bm{G}(\bm{x}_{i})+ J\bm{G}(\bm{x}_{i}) \Delta_{i}\\
\bm{x}_{i+1} - \Delta_{i+1} &= \bm{G}(\bm{x}_{i}- \Delta_{i}) \approx   \bm{G}(\bm{x}_{i})- J\bm{G}(\bm{x}_{i}) \Delta_{i}\\
\end{aligned}
\end{equation}
Here $J\bm{G}(\bm{x}_{i})$ is the Jacobian matrix of $\bm{G}(\bm{x}_{i})$,
which can be approximated as
\begin{equation}
\begin{aligned}
J\bm{G}(\bm{x}_{i}) \Delta_{i} &\approx \bm{G}(\bm{x}_{i} + \Delta_{i}) -   \bm{G}(\bm{x}_{i}), \\
J\bm{G}(\bm{x}_{i}) \Delta_{i} &\approx -\bm{G}(\bm{x}_{i} - \Delta_{i}) +   \bm{G}(\bm{x}_{i}) \, \text{.}
\end{aligned}
\end{equation}
The equations correspond to forward difference and backward difference schemes respectively. Adding these two expressions yields an estimation of 
each column vector of the Jacobian as
\begin{equation}
J\bm{G}_{j}(\bm{x}_{i}) \approx \frac{\bm{G}(\bm{x}_{i} + \Delta_{i}^{j}) - \bm{G}(\bm{x}_{i} - \Delta_{i}^{j})}{2 \delta}
\end{equation}
where $\Delta_{i}^{j} = \delta \bm{e}^{T}_{j}$ and $\bm{e}_{j}$ is the unit vector with unit in the j-th element.
From this robust numerical estimation of the Jacobian matrix, any algorithm for calculating the LLE can be easily employed even for non-smooth applications. Here, the algorithm mentioned in \cite{wolf1986quantifying} has been implemented.
The presented method has been used for a non-smooth application by \cite{pikunov2019numerical}. \\
In the following a short study of the validity of the presented LLE estimation method is given.
This will be done by first comparing the results of the estimation method on a continuous three-dimensional problem with the exact Jacobi matrix and afterwards validating the method on the discontinuous ODE-system of equation (\ref{eq::dynamic_system}). \\
The continuous example is taken out of \cite{molaie2013simple}. It reads
\begin{equation}\label{eq::ProblemFromPaper}
\begin{aligned}
\dot{x} &= y \\
\dot{y} &= z \\
\dot{z} &= -ax - y - 4z + y^{2} + xy 
\end{aligned}
\end{equation}
Here $a$ is a given parameter. The bifurcation diagram of the problem for the domain $a \in [3.3,3.4]$ is plotted in the upper image of Figure \ref{fig:LLEProblemFromPaper}. 

\begin{figure}[h!]
\centering
\includegraphics[scale=0.5]{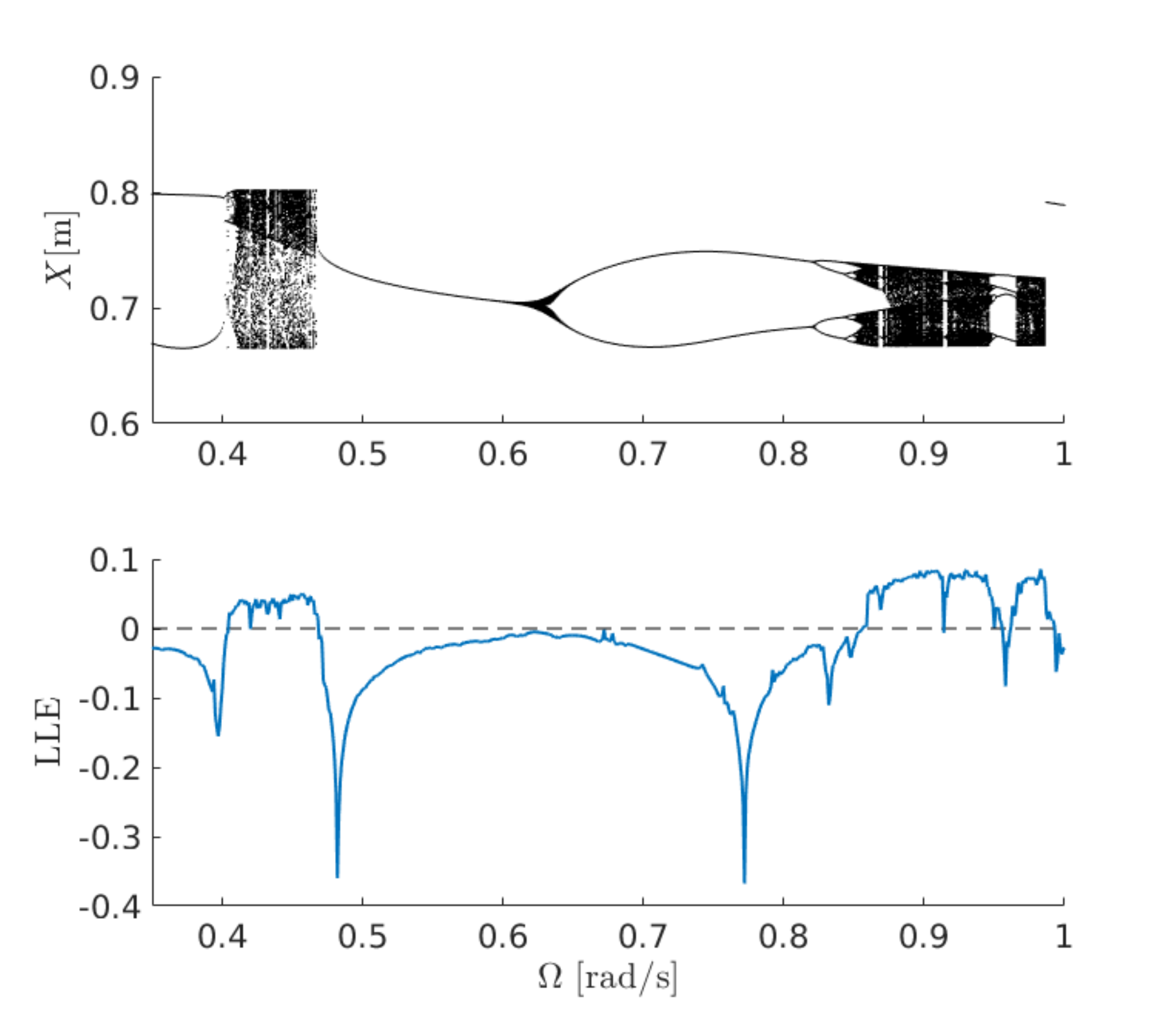} 
\caption[LLE estimation technique for discontinuous problem]{Upper image: Bifurcation diagram over the angular frequency. Below: LLE over $\Omega$. }
\label{fig:LLEOverBifurcationMyProblem}       
\end{figure}
It can be seen that with an increasing value of the parameter the response gets more chaotic.
The LLE estimated with the algorithm as described in \cite{wolf1986quantifying}, utilizing the exact Jacobi matrix of the system, is illustrated in the middle picture. It can be seen that the LLE mirrors the chaotic behavior of the bifurcation diagram. The same algorithm but with the estimated Jacobi matrix of section \ref{sec::EstimationLLE} yields the LLE values of the lowest image in Figure \ref{fig:LLEProblemFromPaper}. Here a perturbation vector given by $\Delta = 10^{-4} \left[ 1 , 1 , 1 \right]^{T}$ was employed. It can be seen that the presented Jacobi estimation method is able to achieve similar results to the exact Jacobi matrix representation. \\
The bifurcation diagram and the respective LLE values for the dynamic problem defined by equation (\ref{eq::dynamic_system}) are plotted over the angular frequency in Figure \ref{fig:LLEOverBifurcationMyProblem}. Here the same perturbation vector as in the previous example is employed.
It can be seen that the estimated LLE value with the given method is also consistent with the bifurcation graph.
\subsection{Metamodel for the LLE indicator}\label{sec::MetaModelForLLEIndi}
\begin{figure}[h!]
\centering
\begin{subfigure}[t]{0.5\textwidth}
\includegraphics[scale=0.4]{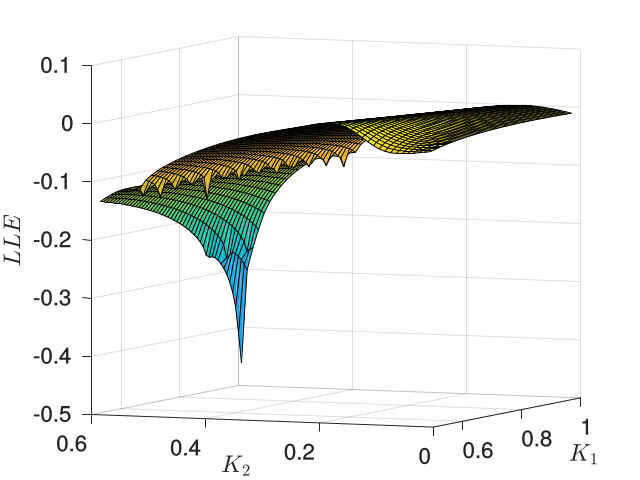} 
\subcaption{}\label{fig:LLEExample1_plot}
\end{subfigure}%
\begin{subfigure}[t]{0.5\textwidth}
\includegraphics[scale=0.4]{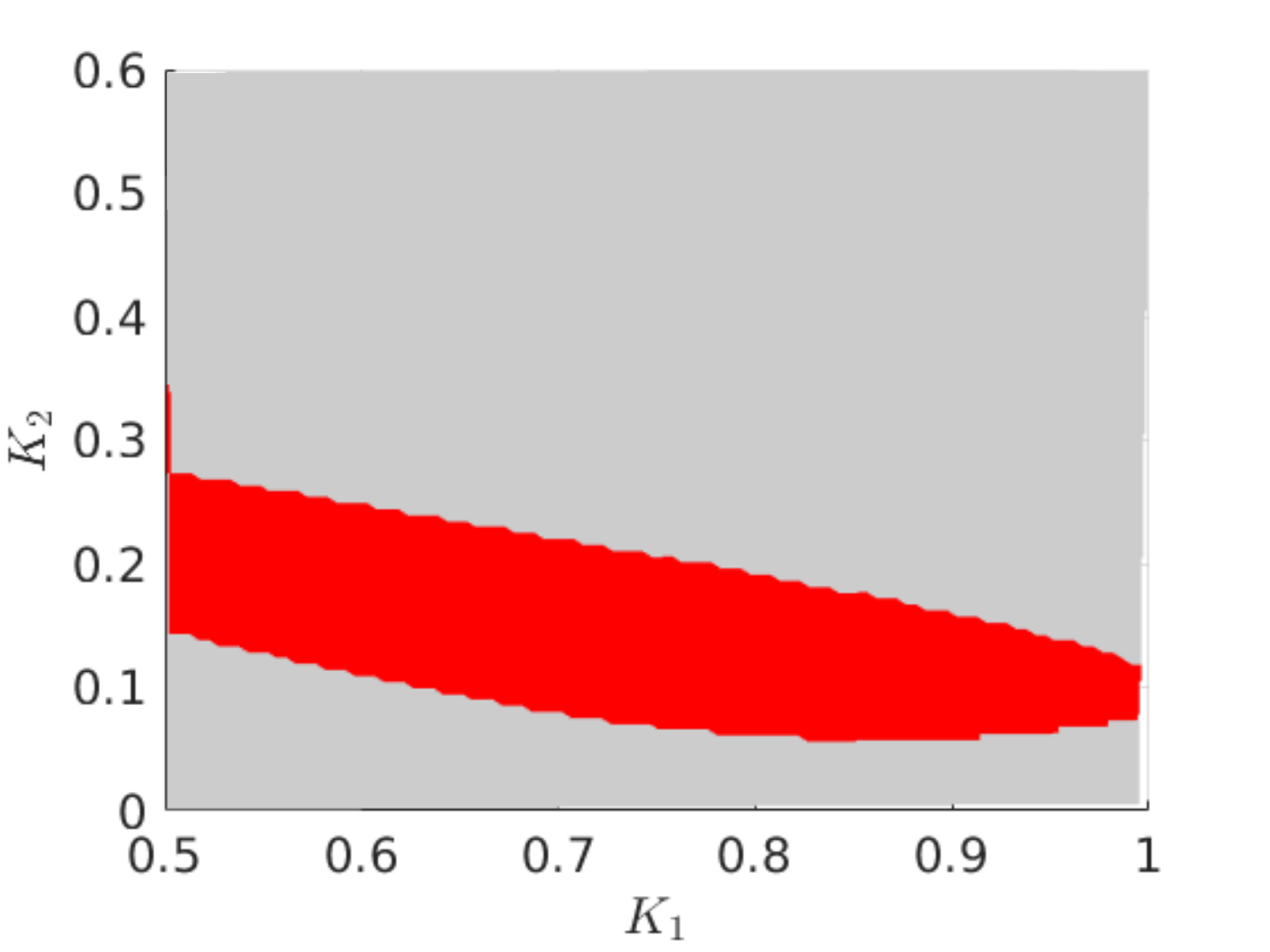} 
\subcaption{}\label{fig:LLEExample1_redgray}
\end{subfigure}
\caption[Example of LLE values over the spring-stifnesses]{Example of LLE values over the spring-stifnesses, (a) LLE values over parametic space, (b) Outcome of $\mathcal{C}_{MoB}$ (red: $\mathcal{C}_{MoB}=1$, gray = $\mathcal{C}_{MoB}=0$) .With $K_{1}$ given in $\text{N}/\text{m}^{3}$, $K_{2}$ in $\text{N}/\text{m}$ and LLE is unitless.  }\label{fig:LLEExample1}
\end{figure}

Issues arise when trying to utilize the presented adaptive sampling techniques for a metamodel of the LLE. The LLE value over the two spring stiffnesses that define the spring of the nonlinear mass-on-belt problem (of Figure \ref{fig:Application}) is depicted in Figure \ref{fig:LLEExample1_plot}. 
A given LLE value classifies the behavior of the given dynamic system into chaotic or regular motion. \\
Consider a user defined output for the mass-on-belt problem (MoB) (e.g. output after 150 seconds, output after one integration step, ...) of the systems of equations of (\ref{eq::dynamic_system}) given as a blackbox function denoted by $\mathcal{M}_{MoB}(\bm{x})$ where $\bm{x}$ defines the parametric input. Let $\mathcal{C}_{MoB}(\bm{x})$ be a blackbox function classifying the motion of the system with
\begin{equation}\label{eq::ClassifierMotion}
\mathcal{C}_{MoB}(x) = \begin{cases}
1, & \text{if} \, \, \mathcal{M}_{LLE,MoB}(\bm{x}) \geq 0 \\
0, & \text{if} \, \, \mathcal{M}_{LLE,MoB}(\bm{x}) < 0 
\end{cases}
\end{equation} 
where $\mathcal{M}_{LLE,MoB}(\bm{x})$ provides the LLE value of the function $\mathcal{M}_{MoB}$ with input $\bm{x}$. \\
The output of the classification function $\mathcal{C}_{MoB}$ of the data illustrated in Figure \ref{fig:LLEExample1_plot} is depicted in Figure \ref{fig:LLEExample1_redgray}.
On one hand values yielding $\mathcal{C}_{MoB}=1$, characterizing chaotic behavior, are highlighted in red in Figure \ref{fig:LLEExample1_redgray}. On the other hand values generating the output $\mathcal{C}_{MoB}=0$, which characterize regular motion, are indicated in gray.

The main focus of a metamodel for this application should be to identify if LLE$= \mathcal{M}_{LLE,MoB}(\bm{x})$ for a point $\bm{x}$ of the parametric domain is above or below 0, whereas the precise value of the LLE, as seen in Figure \ref{fig:LLEExample1_plot}, is not of particular interest.

\subsubsection{Comparison of Kriging metamodel for classifier or data}
\begin{figure}
\centering
\includegraphics[scale=0.5]{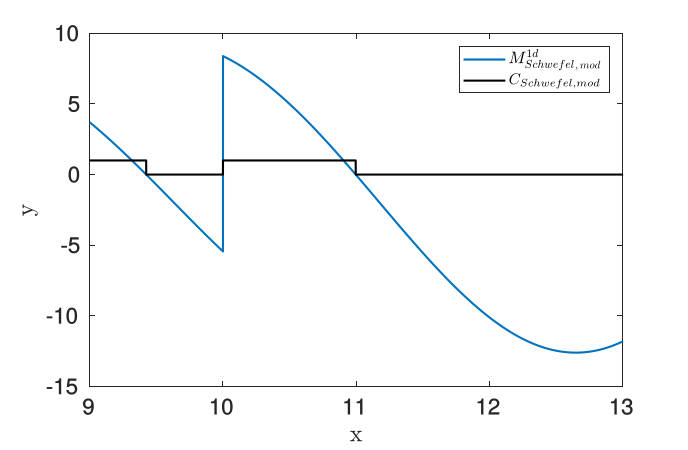} 
\caption[$\mathcal{M}_{Schwefel,mod}$ and its classifying function]{$\mathcal{M}_{Schwefel,mod}$ and its classifying function $C_{Schwefel,mod}$}\label{fig::FunctionandClassifier}
\end{figure}

There are to be two different ways of creating a metamodel for this application:
\begin{enumerate}
\item Generating a metamodel $\hat{\mathcal{M}}$ for the smooth data function $\mathcal{M}_{LLE,MoB}(\bm{x})$ after which it can be evaluated depending on the algebraic sign of the output, e.g. with 
\begin{equation}
\begin{aligned}
f_{\hat{\mathcal{M}}}(\bm{x}) = \begin{cases}
1,  &\text{if} \, \, \hat{\mathcal{M}}(\bm{x}) \geq 0 \\
0, &\text{if} \, \, \hat{\mathcal{M}}(\bm{x}) < 0 .
\end{cases}
\end{aligned}
\end{equation}
\item Creating a surrogate model $\hat{\mathcal{M}}_{\mathcal{C}}$ with the data of the classification $\mathcal{C}_{MoB}$ and evaluating the output of the metamodel with e.g. an auxiliary function
\begin{equation}
\begin{aligned}
f_{\hat{\mathcal{M}}_{\mathcal{C}}}(\bm{x}) = \begin{cases}
1,  &\text{if} \, \, \hat{\mathcal{M}}_{\mathcal{C}}(\bm{x}) \geq 0.5 \\
0, &\text{if} \, \, \hat{\mathcal{M}}_{\mathcal{C}}(\bm{x}) < 0.5 .
\end{cases}
\end{aligned}
\end{equation}
\end{enumerate}
In the following the two approaches will be compared on a modified Schwefel function given by 
\begin{equation}\label{eq::ModSchwefel}
\begin{aligned}
\mathcal{M}_{Schwefel,mod}(x) = \begin{cases}
x \sin (x),  &\text{if} \, \, x\leq 10 \\
- x \cos (x), &\text{if} \, \, x >10.
\end{cases}
\end{aligned}
\end{equation}
The function classifying the output of $\mathcal{M}_{Schwefel,mod}$ is given by 
\begin{equation}\label{eq::C_schwefelMod}
\begin{aligned}
C_{Schwefel,mod} (x) = \begin{cases}
1,  &\text{if} \, \, \mathcal{M}_{Schwefel,mod}^{1d} \geq 0 \\
0, &\text{if} \, \, \mathcal{M}_{Schwefel,mod}^{1d} < 0 .
\end{cases}
\end{aligned}
\end{equation}
$\mathcal{M}_{Schwefel,mod}$ and $C_{Schwefel,mod}$ are displayed in Figure \ref{fig::FunctionandClassifier}. Next, two metamodels are generated employing different sample sets: 
\begin{itemize}
\item a metamodel for the data function $\mathcal{M}_{Schwefel,mod}$. Here denoted by\\$\hat{\mathcal{M}}_{Schwefel,mod}$,
\item and a metamodel $\hat{M}_{C}$ for classifier $C_{Schwefel,mod}$.
\end{itemize}
\begin{figure}[h!]
\centering
\begin{subfigure}[t]{0.5\textwidth}
\includegraphics[scale=0.4]{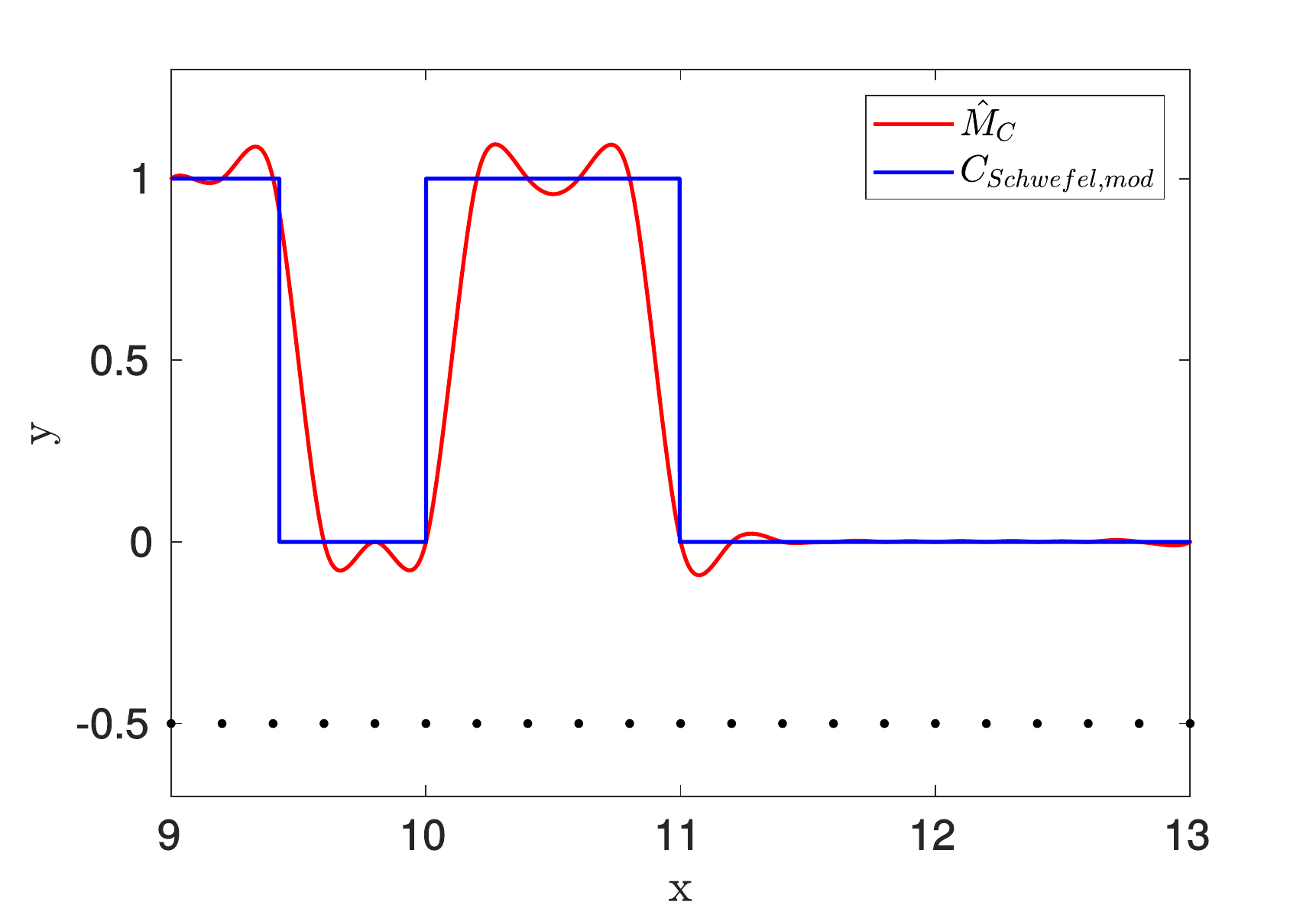} 
\subcaption{}\label{fig:hatCOp2}
\end{subfigure}%
\begin{subfigure}[t]{0.5\textwidth}
\includegraphics[scale=0.4]{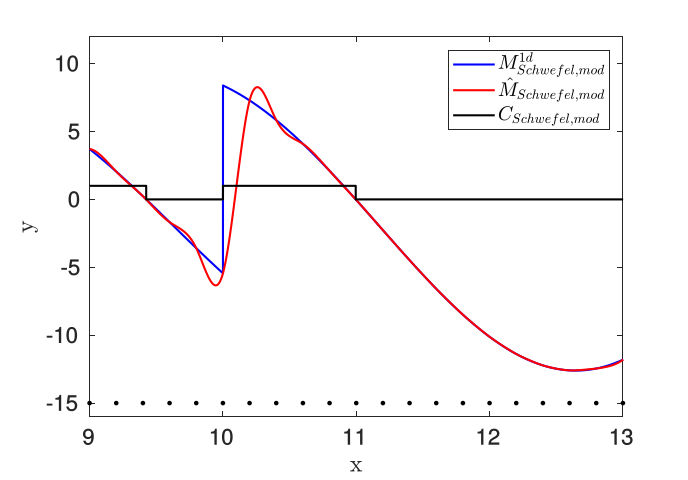} 
\subcaption{}\label{fig:hatM0p2}
\end{subfigure}
\begin{subfigure}[t]{0.5\textwidth}
\includegraphics[scale=0.4]{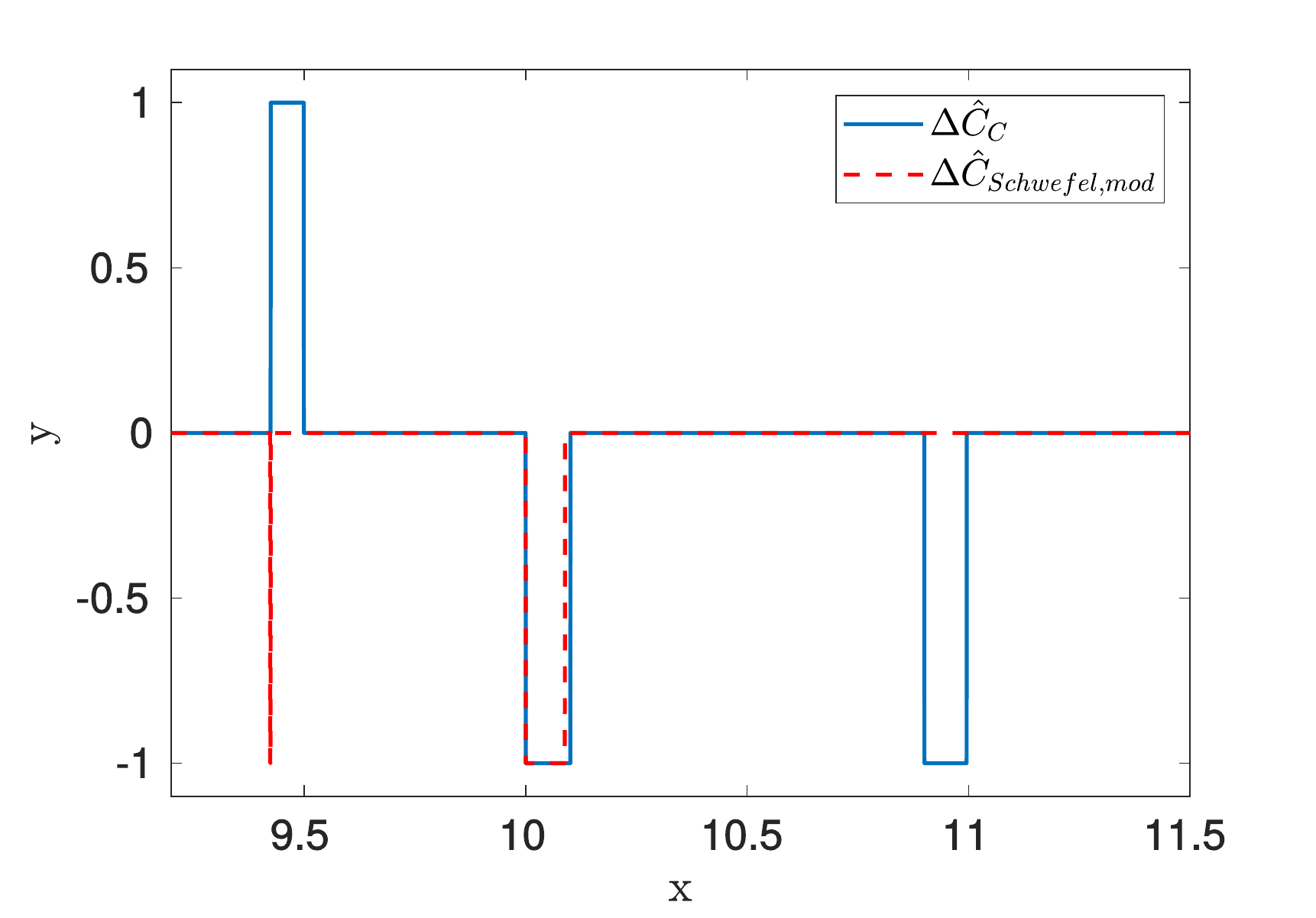} 
\subcaption{}\label{fig:Diff0p2}
\end{subfigure}%
\caption[Difference between metamodel for classifier and data samples every $0.2$ steps]{Difference between metamodel for classifier $\mathcal{C}_{Schwefel,mod}$ and metamodel for data function $\mathcal{M}_{Schwefel,mod}^{1d}$. Samples every $0.2$ steps. Black dots represent sample points. (a) Classifier and its metamodel generated from the given samples, (b) Data function and its metamodel (classifier in black), (c) Difference between the target classifier and the two metamodels.}\label{fig:0p2}
\end{figure}

In order to compare the results of the two metamodels the following auxiliary functions are defined:
\begin{itemize}
\item A  function for $\hat{\mathcal{M}}_{Schwefel,mod}$ with boolean output
\begin{equation}
\begin{aligned}
f_{ \hat{\mathcal{M}}_{Schwefel,mod}}(x) = \begin{cases}
1,  &\text{if} \, \, \hat{\mathcal{M}}_{Schwefel,mod}(x) \geq 0 \\
0, &\text{if} \, \, \hat{\mathcal{M}}_{Schwefel,mod}(x) < 0 .
\end{cases}
\end{aligned}
\end{equation}
\item A binary output function for $\hat{M}_{C}$ with
\begin{equation}
\begin{aligned}
f_{ \hat{M}_{C}}(x) = \begin{cases}
1,  &\text{if} \, \, \hat{M}_{C}(x) \geq 0.5 \\
0, &\text{if} \, \, \hat{M}_{C}(x) < 0.5 .
\end{cases}
\end{aligned}
\end{equation}
\end{itemize}
The differences between $C_{Schwefel,mod}$ and the two auxiliary functions are defined in order to compare the performance with
\begin{equation}\label{eq::Differences}
\begin{aligned}
\Delta \hat{\mathcal{C}}_{\mathcal{C}}(x)  &= f_{ \hat{M}_{C}}(x) - C_{Schwefel,mod}(x) \\
\Delta \hat{\mathcal{C}}_{Schwefel,mod}(x)  &= f_{ \hat{\mathcal{M}}_{Schwefel,mod}}(x) - C_{Schwefel,mod}(x).
\end{aligned}
\end{equation}
In order to evaluate the respective classification performance of $\hat{M}_{C}$ and\\$\mathcal{M}_{Schwefel,mod}^{1d}$ with the help of the auxiliary functions of equation (\ref{eq::Differences}) three different sets of observations over the input domain $x \in [9,13]$ are presented, \\
Equidistant samples every $0.2$ step is chosen over the domain $x \in [9,13]$ in 
Figure \ref{fig:0p2}. The metamodel for the classifier is depicted in Figure \ref{fig:hatCOp2}. $\hat{\mathcal{M}}_{Schwefel,mod}$ and the target function are displayed in Figure \ref{fig:hatM0p2}. 

\begin{figure}[h!]
\centering
\begin{subfigure}[t]{0.5\textwidth}
\includegraphics[scale=0.4]{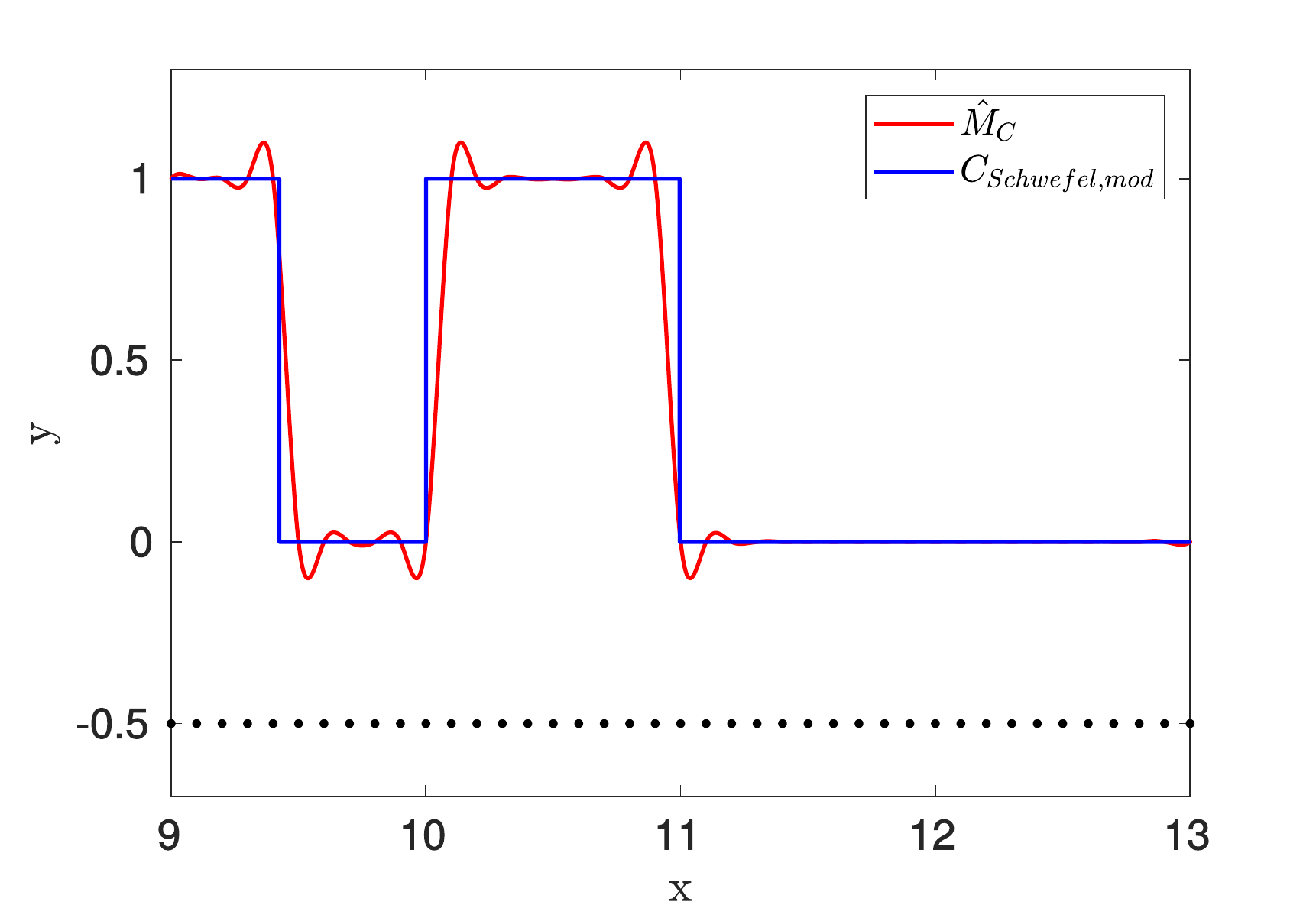} 
\subcaption{}\label{fig:hatCOp1}
\end{subfigure}%
\begin{subfigure}[t]{0.5\textwidth}
\includegraphics[scale=0.4]{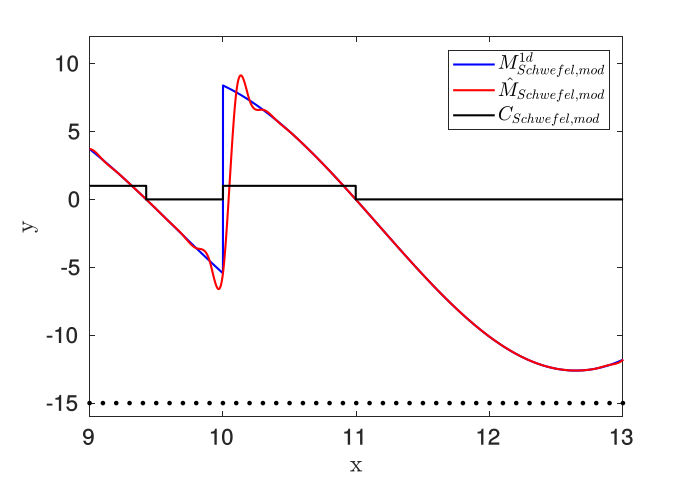} 
\subcaption{}\label{fig:hatMOp1}
\end{subfigure}
\begin{subfigure}[t]{0.5\textwidth}
\includegraphics[scale=0.4]{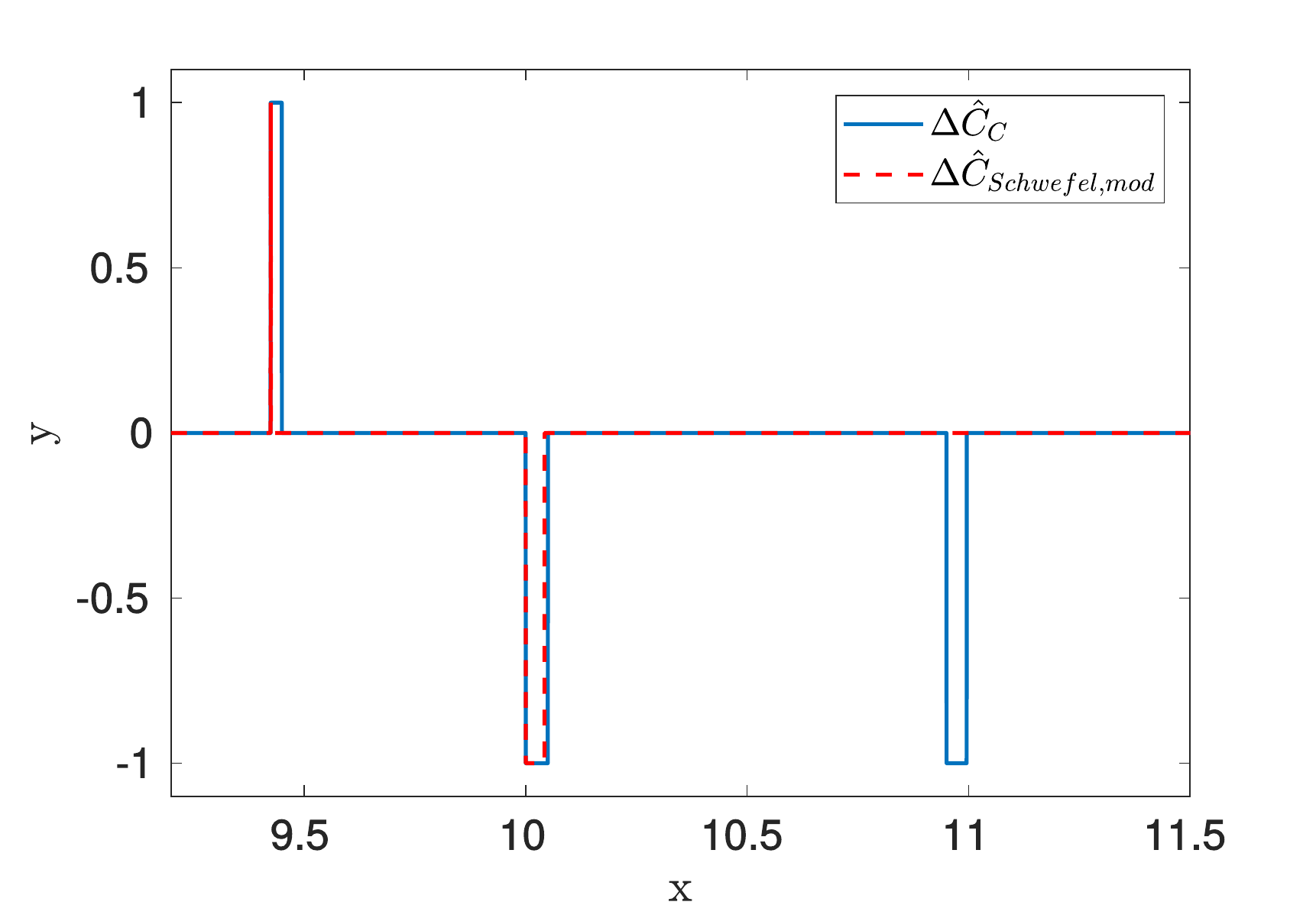} 
\subcaption{}\label{fig:DiffOp1}
\end{subfigure}%
\caption[Difference between metamodel for classifier and data samples every $0.1$ steps]{Difference between metamodel for classifier $\mathcal{C}_{Schwefel,mod}$ and metamodel for data function $\mathcal{M}_{Schwefel,mod}^{1d}$. Samples every $0.1$ steps. Black dots represent sample points. (a) Classifier and its metamodel generated from the given samples, (b) Data function and its metamodel (classifier in black), (c) Difference between the target classifier and the two metamodels. }\label{fig:0p1}
\end{figure}

\begin{figure}[h!]
\centering
\begin{subfigure}[t]{0.5\textwidth}
\includegraphics[scale=0.4]{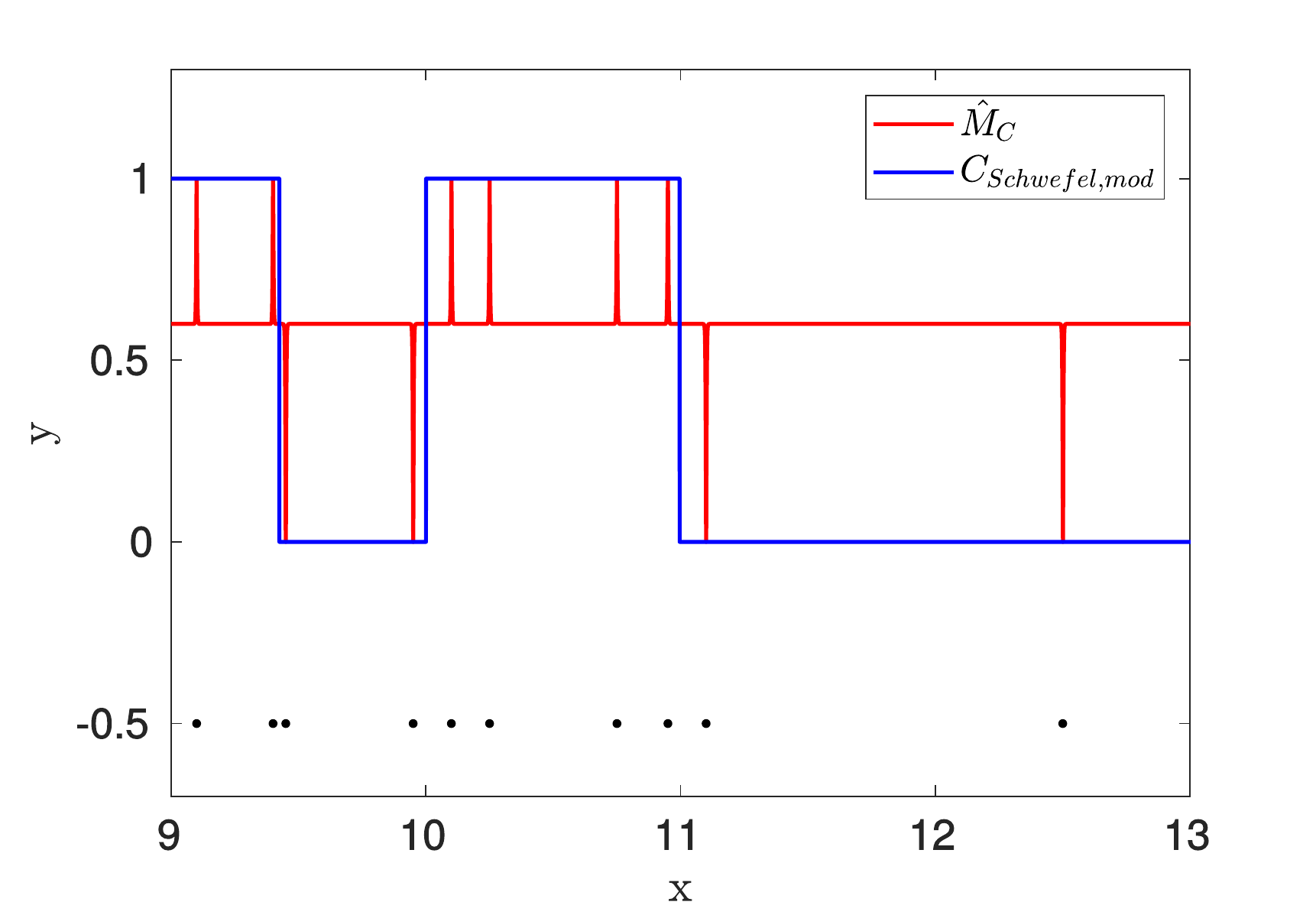} 
\subcaption{}\label{fig:hatCMy}
\end{subfigure}%
\begin{subfigure}[t]{0.5\textwidth}
\includegraphics[scale=0.4]{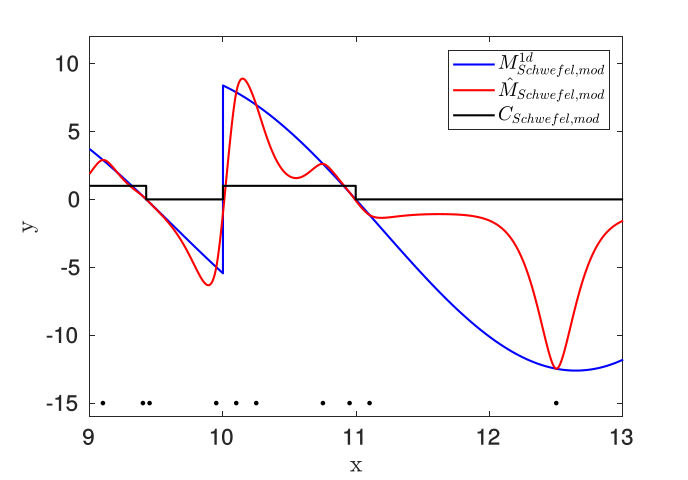} 
\subcaption{}\label{fig:hatMMy}
\end{subfigure}
\begin{subfigure}[t]{0.5\textwidth}
\includegraphics[scale=0.4]{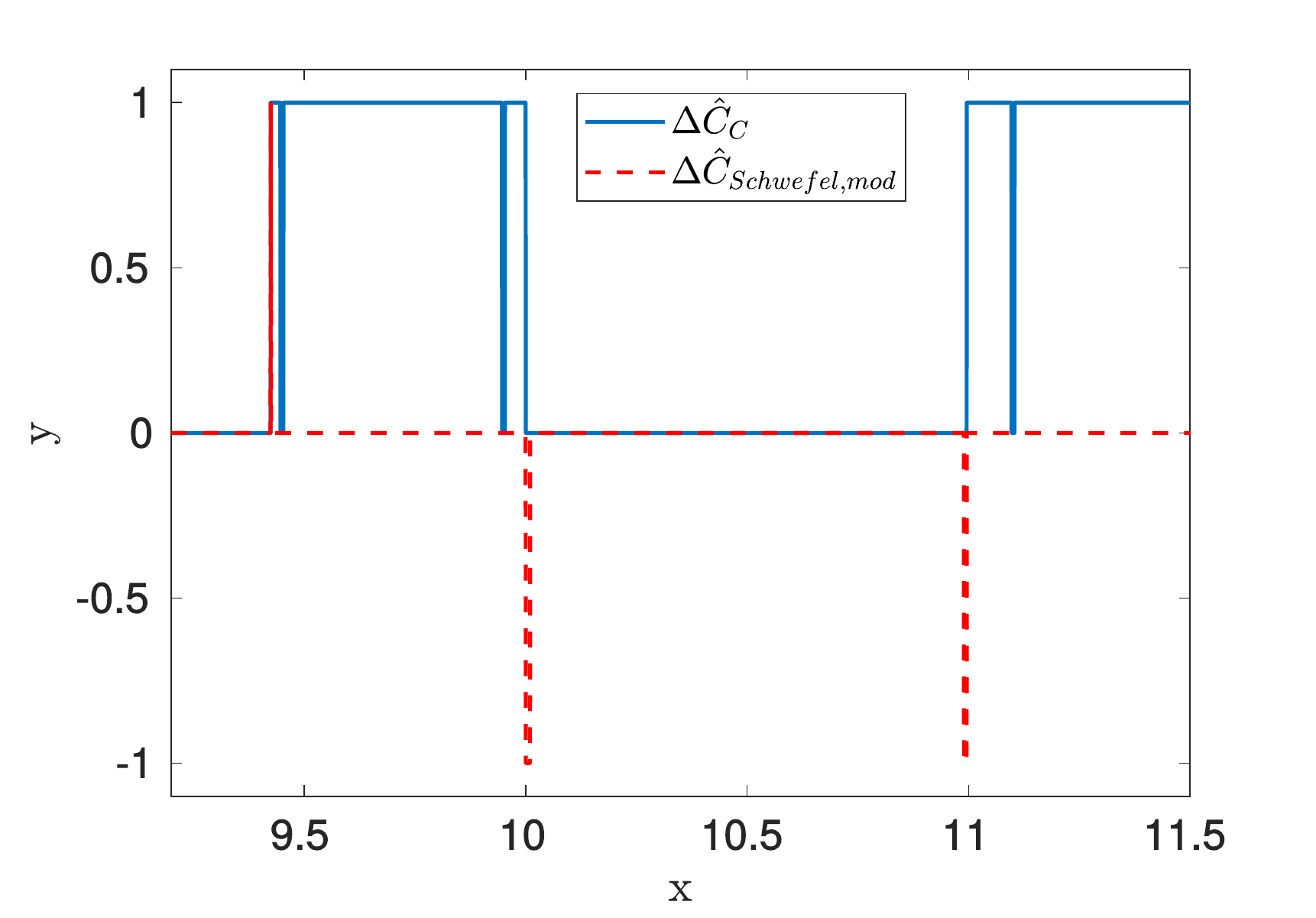} 
\subcaption{}\label{fig:DiffMy}
\end{subfigure}%
\caption[Difference between metamodel for classifier and data function for selected samples]{Difference between metamodel for classifier $\mathcal{C}_{Schwefel,mod}$ and metamodel for data function $\mathcal{M}_{Schwefel,mod}^{1d}$. Samples selected as represented by black dots. (a) Classifier and its metamodel generated from the given samples, (b) Data function and its metamodel (classifier in black), (c) Difference between the target classifier and the two metamodels.}\label{fig:My}
\end{figure}

The error between the metamodels and the target classifier as defined in equation (\ref{eq::Differences}) are illustrated in Figure \ref{eq::Differences} over the shortened domain of $[9.2, 11.5]$, which highlights the binary transition regions. It can be seen that generating a metamodel for the smooth target function instead of the classifier yields more proficient results. The same is true when refining the sample distances to include a sample every $0.1$ steps as shown in Figure \ref{fig:0p1}. Furthermore when only utilizing essential samples around the edge points as seen in Figure \ref{fig:hatMMy}, which yield very proficient results for the metamodel of $\mathcal{M}_{Schwefel,mod}$. Here, it can be noticed that $\hat{M}_{C}$ appears to exhibit numerical issues. \\
Overall, after studying the given artificial application, the generation of a surrogate model for the smooth target function instead of its classifier should be the preferred choice when aiming to create low-cost functions with Kriging for classification problems. 

\clearpage
\subsubsection{Evaluation of common adaptive sampling techniques for the metamodel generation for the LLE indicator}
\begin{figure}[h!]
\centering
\begin{subfigure}[t]{0.5\textwidth}
\includegraphics[scale=0.33]{Bilder/LLE/LLE_Example_Problem.jpg} 
\subcaption{$\mathcal{M}_{Schwfel,mod}^{1d}$ and $C_{Schwefel,mod}$}\label{fig:ModfiedPlot}
\end{subfigure}%
\begin{subfigure}[t]{0.5\textwidth}
\includegraphics[scale=0.33]{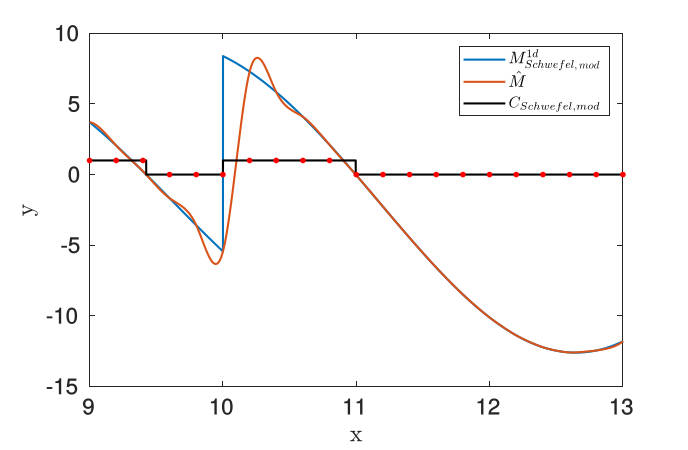} 
\subcaption{Equidistant samples every 0.2 steps.}\label{fig:fig:ModfiedPlot0p2}
\end{subfigure}
\begin{subfigure}[t]{0.5\textwidth}
\includegraphics[scale=0.33]{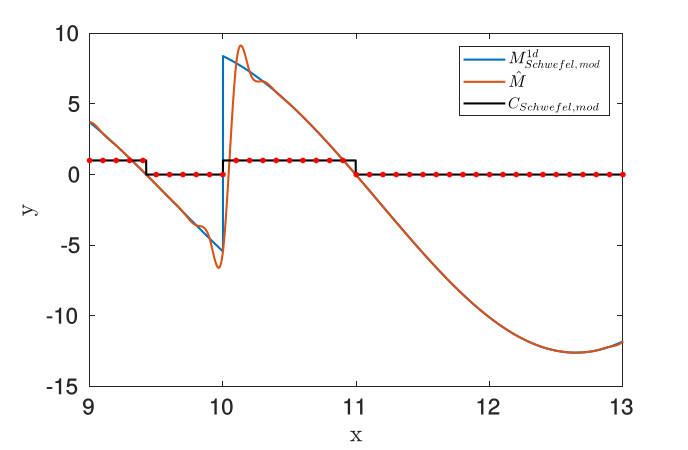} 
\subcaption{Equidistant samples every 0.1 steps.}\label{fig:fig:ModfiedPlot0p1}
\end{subfigure}%
\begin{subfigure}[t]{0.5\textwidth}
\includegraphics[scale=0.33]{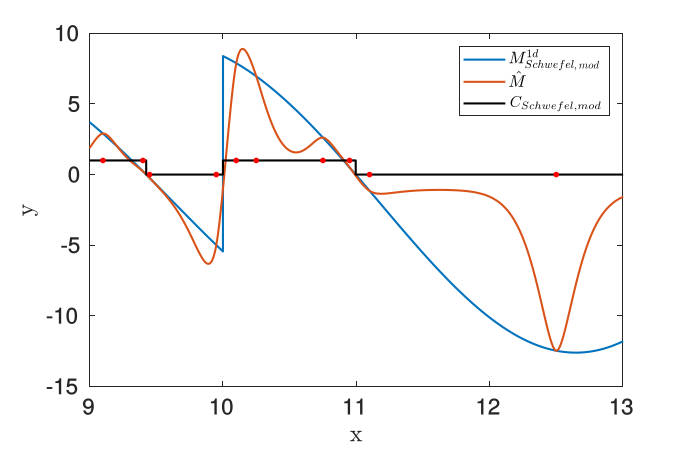} 
\subcaption{Samples located at the transitions.}\label{fig:fig:ModfiedPlotMy}
\end{subfigure}
\caption[Sample positions for classification problem]{Sample positions for classification of  $\mathcal{M}_{Schwefel,mod}^{1d}$.  }\label{fig:}
\end{figure}

The samples for a proficient Kriging metamodel $\hat{\mathcal{M}}$ for the smooth data function $\mathcal{M}_{LLE,MoB}(\bm{x})$  need to be positioned on the edges between regular (positive) and chaotic (negative) outcome. This can be seen when considering $\mathcal{M}_{Schwefel,mod}^{1d}$ of equation (\ref{eq::ModSchwefel}) with a discontinuous jump at 10 and its classifying function $C_{Schwefel,mod}$ as presented in equation (\ref{eq::C_schwefelMod}).
The function and its classification are displayed in Figure \ref{fig:ModfiedPlot} over the domain $x \in [9,13]$. The metamodel of $\mathcal{M}_{Schwefel,mod}^{1d}$ when considering $21$ equidistant samples generated every $0.2$ steps is shown in Figure \ref{fig:fig:ModfiedPlot0p2}. The samples are represented by the red dots on the classification function. The $y$-value of the points reflect the output of the classifier for the particular $x$-value. The Kriging metamodel generated with the given sample values is depicted by the red line. It can be noticed that the overall shape of the function is approximated proficiently. However the classification around the sharp transition at $x=10$ is not sufficient. The same problem can be noticed when $41$ equidistant samples every $0.1$ steps are utilized. Even though it can be seen that with respect to the last metamodel the classification is improved. Next, consider the $10$ sample points of Figure \ref{fig:fig:ModfiedPlotMy}. It can be seen that the samples are carefully placed along the transitions of the classifier value. The metamodel $\hat{\mathcal{M}}$ is not approximating $\mathcal{M}_{Schwefel,mod}^{1d}$ as proficiently as before, however the classification of the metamodel yields better results. Hence, when intentionally adding samples points at the edges between two classification values the efficiency of the metamodel can be drastically improved.
This fact yields to be problematic for the discussed adaptive sampling techniques because they focus on an exact representation of the output values and not on finding the transitions of the classifier function $\mathcal{C}_{MoB}$ and therefore exhibit a lack of ability to adaptively create samples for classification problems. 
\begin{figure}[h!]
\centering
\begin{subfigure}[t]{0.8\textwidth}
\includegraphics[scale=0.5]{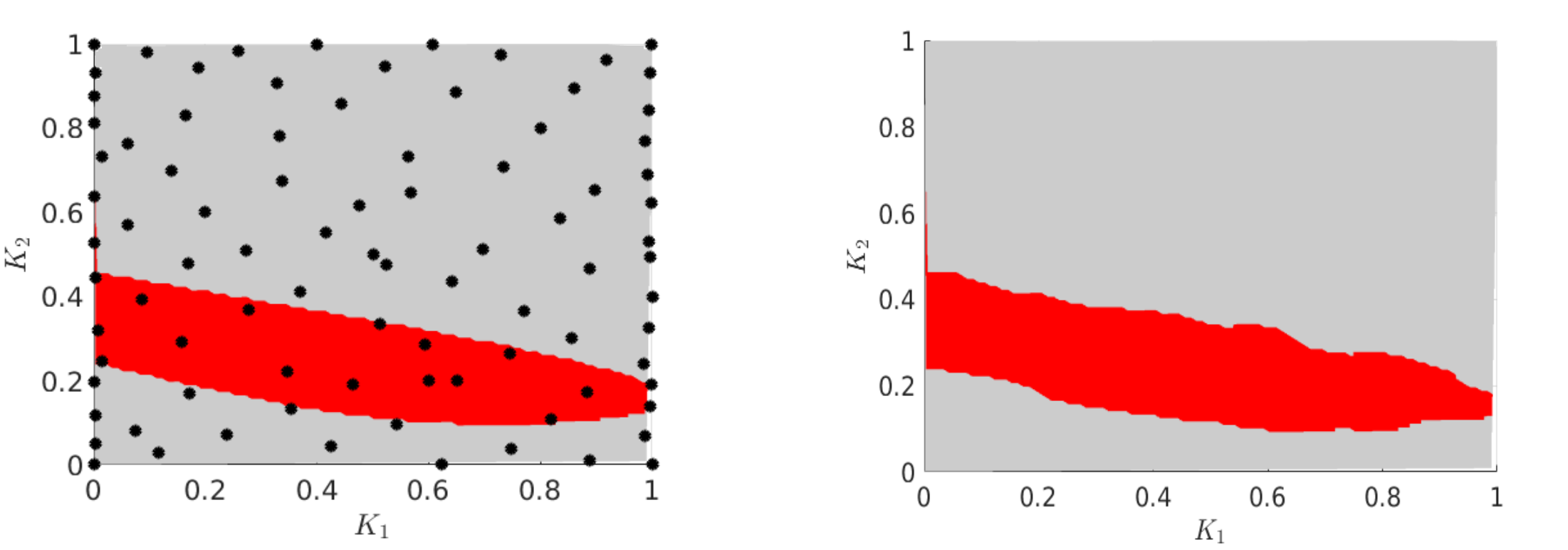} 
\subcaption{AME}\label{fig:SampleKSeta06ComAME}
\end{subfigure}
\begin{subfigure}[t]{0.8\textwidth}
\includegraphics[scale=0.5]{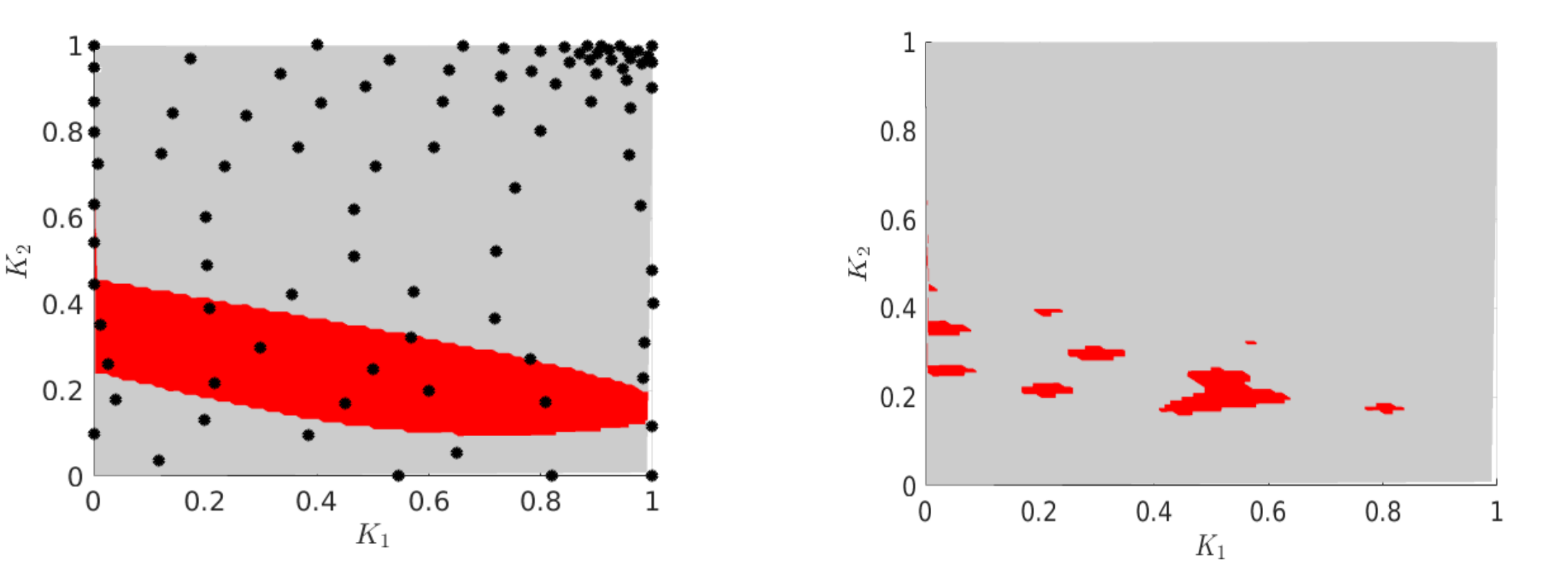} 
\subcaption{EIGF}\label{fig:SampleKSeta06ComEIGF}
\end{subfigure}
\begin{subfigure}[t]{0.8\textwidth}
\includegraphics[scale=0.5]{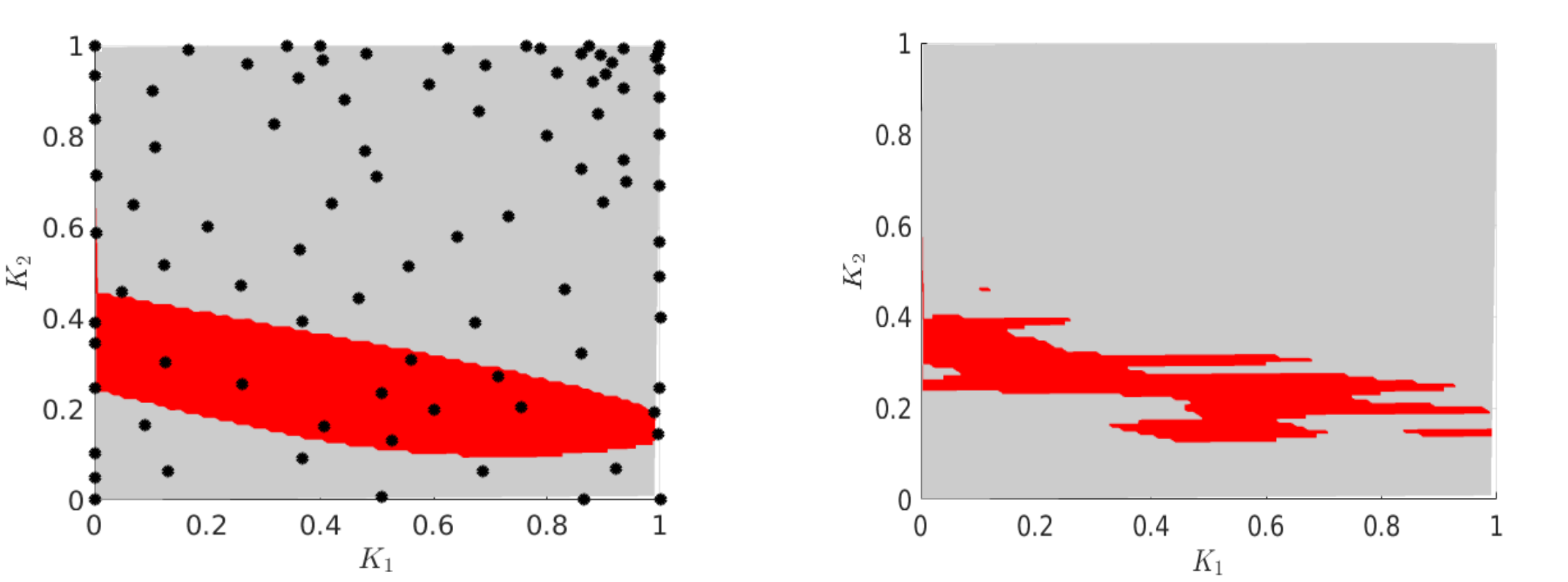} 
\subcaption{MEPE}\label{fig:SampleKSeta06ComMEPE}
\end{subfigure}
\begin{subfigure}[t]{0.8\textwidth}
\includegraphics[scale=0.5]{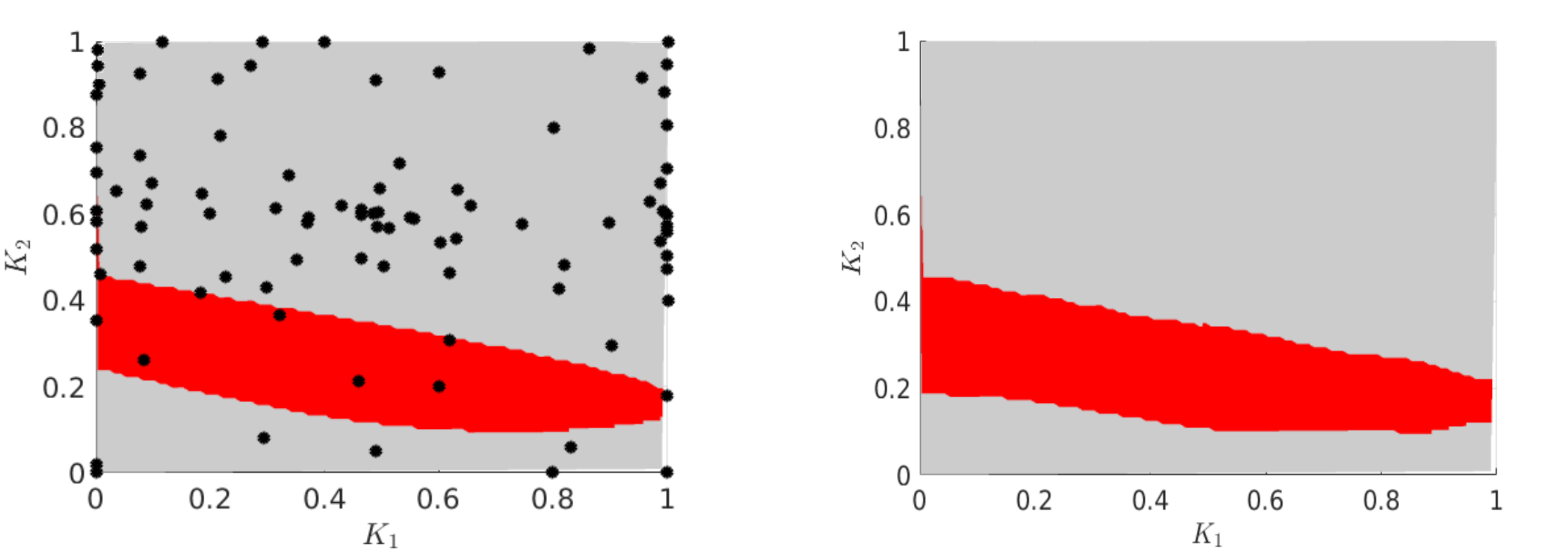} 
\subcaption{SSA}\label{fig:SampleKSeta06ComSSA}
\end{subfigure}
\caption[Sample positions for different adaptive sampling techniques for the classification of chaotic motion]{Sample positions for different adaptive sampling techniques for the classification of chaotic motion after 90 samples. $K_{1}$ is given in $\text{N}/\text{m}^{3}$, $K_{2}$ in $\text{N}/\text{m}$. }\label{fig:SampleKSeta06Com}
\end{figure}
This is demonstrated in Figure \ref{fig:SampleKSeta06Com}. Four different commonly utilized adaptive sampling techniques, representative of the previously investigated techniques of this thesis, are used to adaptively sample the problem of Figure \ref{fig:LLEExample1}, in order to create a low-cost metamodel. 
The left-hand side of each row displays 85 created samples with the respective technique after, initially, five samples were generated with TPLHD. Here, the contour of the classification is illustrated in each image. On the right-hand side the respective metamodels generated with the 90 samples (shown in the left-hand side) are displayed. It is noticeable that the majority of samples created by these techniques is in areas not around the transition edges. However even without sampling the edge regions of $\mathcal{C}_{MoB}$, AME (Figure \ref{fig:SampleKSeta06ComAME}) and SSA (Figure \ref{fig:SampleKSeta06ComSSA}) display proficient results. Nevertheless the sampling procedures are not efficient for this application. \\ \\
To the best of the authors knowledge no adaptive sampling technique presented in the literature is aimed for classification problems in context of Kriging metamodeling with focus on sampling near the transition regions.  
An innovative adaptive sampling technique aiming to fill this gap is proposed and tested in the next chapter.

\clearpage
\section{Concluding remark}
The mathematical model of a non-linear oscillator of Duffing's type with an elastoplastic friction model has been investigated. A metamodel for the sticking time was successfully created with the help of adaptive sampling. The largest Lyapunov exponent as well as the sticking time have been introduced as means to indicate the chaotic behavior of the system. A negative value of the largest Lyapunov exponent results in regular motion. Values equal or larger than zero point towards chaotic motion. Common adaptive sampling techniques have been studied for their use in the generation of a surrogate model for this indicator in order to classify the motion of the presented system system. It was found that the presented adaptive sampling techniques were not able to yield satisfying results, because of their inability to sample near transition edges.
\chapter{Novel adaptive sampling technique for surrogate classification with Kriging}\label{ch::Novel_techniques}
The proposed adaptive sampling technique is aimed for the use of binary classification with Kriging.
Mathematically this can be defined by a function\\$\mathcal{M}: \mathbb{C}^{n} \rightarrow \mathbb{R}$ and a classifier $\mathcal{C}_{\mathcal{M}}$. The classifier maps the output of $\mathcal{M}$ to the set $S = \lbrace 0,1 \rbrace$. The mapping reads
\begin{equation}
\begin{aligned}
\mathcal{C}_{\mathcal{M}}(\bm{x}) = 
\begin{cases}
1, &\text{if} \, \, \mathcal{M}(\bm{x}) \geq a \\
0, &\text{if} \, \, \mathcal{M}(\bm{x}) < a,
\end{cases}
\end{aligned}
\end{equation}
where $\bm{x} = \mathbb{C}^{n}$ and $a \in \mathbb{R}$. 
Binary classification has been of interest in many application fields including e.g. medical research \cite{tang2004granular} and chemical processes \cite{merkwirth2004ensemble}. \\
However in this thesis the focus lies on the classification of the motion of the system of equations (\ref{eq::dynamic_system}) as presented in section \ref{sec::MetaModelForLLEIndi}.
\section{Construction of the technique}
The proposed adaptive sampling technique discretizes the parametric space with Voronoi tessellation and is based on the MIPT algorithm. It was observed that LLE values of above 0, i.e. indicating chaotic motion, cluster together in the input domain. Hence, the classifier of the dynamic motion as presented in equation \ref{eq::ClassifierMotion} exhibits clearly defined edges between the two states. \\
In the next steps this phenomenon is taken as an assumption to build the innovative, hybrid MC-intersite Voronoi (MIVor) adaptive sampling technique. \\
Let the initial input data be defined by $\mathcal{X}= \lbrace \bm{x}^{(1)}, \ldots, \bm{x}^{(m)} \rbrace$ where $\bm{x}^{(i)} \in \mathbb{R}^{n}$ and the one-dimensional output data (here: e.g. the LLE) be given by $\mathcal{Y}= \lbrace y^{(1)}, \ldots, y^{(m)} \rbrace$. The output data can be classified to yield either chaotic motion ($y^{(i)}\geq 0$) or regular motion with $y^{(i)}< 0$ by the mapping $\mathcal{C}_{MoB}$ of equation (\ref{eq::ClassifierMotion}).  \\
The method is based on MIPT for exploration and discretizes the input domain with Voronoi tessellation in order to find the largest cell that indicates chaotic motion. This cell is then used to define an exploitation character. In the following the concepts of Voronoi tessellation and MIPT are shortly defined before MIVor procedure is presented. \\
\\
\textbf{MIPT} \\
MIPT as presented in \cite{crombecq2011efficient} is an efficient adaptive space-filling algorithm. It uses the Monte Carlo approach in the input domain to define a set of $nn$ points $P = \lbrace \bm{p}_{1}, \ldots, \bm{p}_{nn} \rbrace$ where $\bm{p}_{i} \in \mathbb{R}^{n}$. The points are evaluated and ranked with respect to their distance to the existing sample points. A new point is selected by the highest rank or score. A distance threshold is defined to create the space-filling character in the method with 
\begin{equation}
d_{min} = \frac{1}{nn}.
\end{equation}
MIPT over the set of Monte Carlo points $P$ then ranks each point $\bm{p}_{i} \in P$ according to
\begin{equation}
MIPT(P, \bm{p}) = \begin{cases}
0 & \text{if} \, \min_{\bm{p}_{i} \in P} \norm{\bm{p}_{i} - \bm{p}}_{-\infty} < d_{min} \\
\min_{\bm{p}_{i} \in P} \norm{\bm{p}_{i} - \bm{p}}_{2} &\text{if} \, \min_{\bm{p}_{i} \in P} \norm{\bm{p}_{i} - \bm{p}}_{-\infty} \geq d_{min}.
\end{cases}
\end{equation}
Here, $\norm{\bullet}_{2}$ describes the euclidean distance and $\norm{\bullet}_{-\infty}$ is the negative infinity norm defined by  $\norm{\bm{p}}_{-\infty} = \min (p_{i}) \forall i =1, \ldots,n $
\\
\\
\textbf{Voronoi tessellation}\label{sec::Voro} \\
To discretize the input domain the well-known Voronoi tessellation \citep{aurenhammer1991voronoi} around the existing sample points is employed. The idea is that the set of cells $\lbrace C_{1}, \ldots, C_{n} \rbrace$ tesselate the whole space vy employing the so-called dominance function
\begin{equation}
dom(\bm{x}^{(i)}, \bm{x}^{(j)} ) = \lbrace \bm{x} \in \mathbb{R}^{n} | \norm{\bm{x}- \bm{x}^{(i)}} \leq \bm{x - \bm{x}^{(j)}} \rbrace.
\end{equation}
This means, that the subset of a plane needs to be as least as close to $\bm{x}^{(i)}$ as it is to $\bm{x}^{(j)}$ in order for $\bm{x}^{(i)}$ to be dominant over $\bm{x}^{(j)} $. Utilizing this, the Voronoi cell of the point $\bm{x}^{(i)}$ defines the subset of the input space in which $\bm{x}^{(i)}$ is dominant over all other sample points in $\mathcal{X}$ with
\begin{equation}
C_{i} = \cap_{\bm{x}^{(j)} \in \mathcal{X} \setminus \bm{x}^{(i)}} dom(\bm{x}^{(i)}, \bm{x}^{(j)}).
\end{equation}
The computation of the Voronoi tessellation requires high computational effort especially in higher dimensions. However the volume of each corresponding cell can be estimated by employing the Monte Carlo method. For this the parametric space is randomly filled with points. A particular point lies in the influence zone of the sample point that it has the closest euclidean distance to. The volume of each Voronoi cell is estimated by the amount of points in each cell as explained in Box \ref{alg::EstimationVolume}. Simplified, the Voronoi cell with the largest volume is the one with the highest number of randomly sampled points in its influence domain. This is an effective way to define a volume $Vol_{i}, i=1, \ldots,n$ of the Voronoi cell for each sample point. \\
\begin{kasten}[h!]
\begin{Algorithmus}
\begin{itemize}
\itemsep1em 
\item Initial sample data given with $\mathcal{X}= \lbrace \bm{x}^{(1)}, \ldots, \bm{x}^{(m)} \rbrace$. 
\item Obtain $nn$ Monte Carlo points in the input domain and define $P = \lbrace \bm{p}_{1}, \ldots, \bm{p}_{nn} \rbrace$.
\item Define $\bm{Vol} = \lbrace Vol_{i}, \, i=1,\ldots, m \rbrace$. Initially set all $Vol_{i}=0$.
\item[] for $\bm{p} \in P$ \\
\begin{itemize}[topsep=-5px,partopsep=0px]
\item[] Find closest point $\bm{x}^{(i)}$ for $p$ in $\mathcal{X}$ with $\bm{x}^{(i)} =  \min_{\bm{x}^{\star} \in \mathcal{X}} \norm{ \bm{p} - \bm{x}^{\star}}_{2}$.
\item[] Set $Vol_{i} = Vol_{i} + 1$.
\end{itemize}
end
\item Set $Vol_{i} = \frac{Vol_{i}}{\norm{\bm{Vol}}_{1}}$ for all $i=1, \ldots m$.
\end{itemize}
\end{Algorithmus}
\captionof{kasten}{Estimation of normalized Voronoi cell volume}\label{alg::EstimationVolume}
\end{kasten}

\textbf{Mc-intersite Voronoi (MIVor)} \\
The basic idea behind the classification problems MIVor is to focus on sampling around edges between positive ($\mathcal{C}_{MoB}=0$) and negative ($\mathcal{C}_{MoB}=1$) outcomes. 
For instance in Figure \ref{fig:LLEExample1}, it has been observed that chaotic motion indicated by the red points seems to cluster in the parametric domain. \\ 
As presented in the example of Figure \ref{fig:SampleKSeta06Com} Kriging is an accurate interpolative method. The edges between areas indicating chaotic and regular motion need to be sampled sufficiently in order to create a proficient metamodel for binary classification. The exploitation component of the presented method aims to sample close to these edges. \\
The first part of the process is purely exploration-based, i.e. new
samples are added with MIPT: Then, as soon as one sample gives a chaotic outcome ($\mathcal{C}_{MoB}=1$), an exploitation contribution is added to the adaptive algorithm using a decreasing strategy as illustrated in Figure \ref{fig:Ex_ex_decreasing_strategy}. Initially an exploration rate $0 \leq r \leq 1$ is chosen by the user as well as a decreasing factor $\alpha>1.0$. 
In order to choose if the next sampling step is exploration or exploitation-based a uniformly distributed random variable $u \sim U[0,1]$ is sampled:
\begin{itemize}
\item If $u< r$ then the next sample will be found by exploration employing the presented MIPT algorithm.
\item If $u \geq r$ the algorithm aims to sample a new point close to a predicted edge between positive and negative outcome. Therefore this step represents the exploitation component. 
\end{itemize}

\begin{figure}[h!]
\centering
\begin{subfigure}[t]{0.50\textwidth}
\includegraphics[scale=0.4]{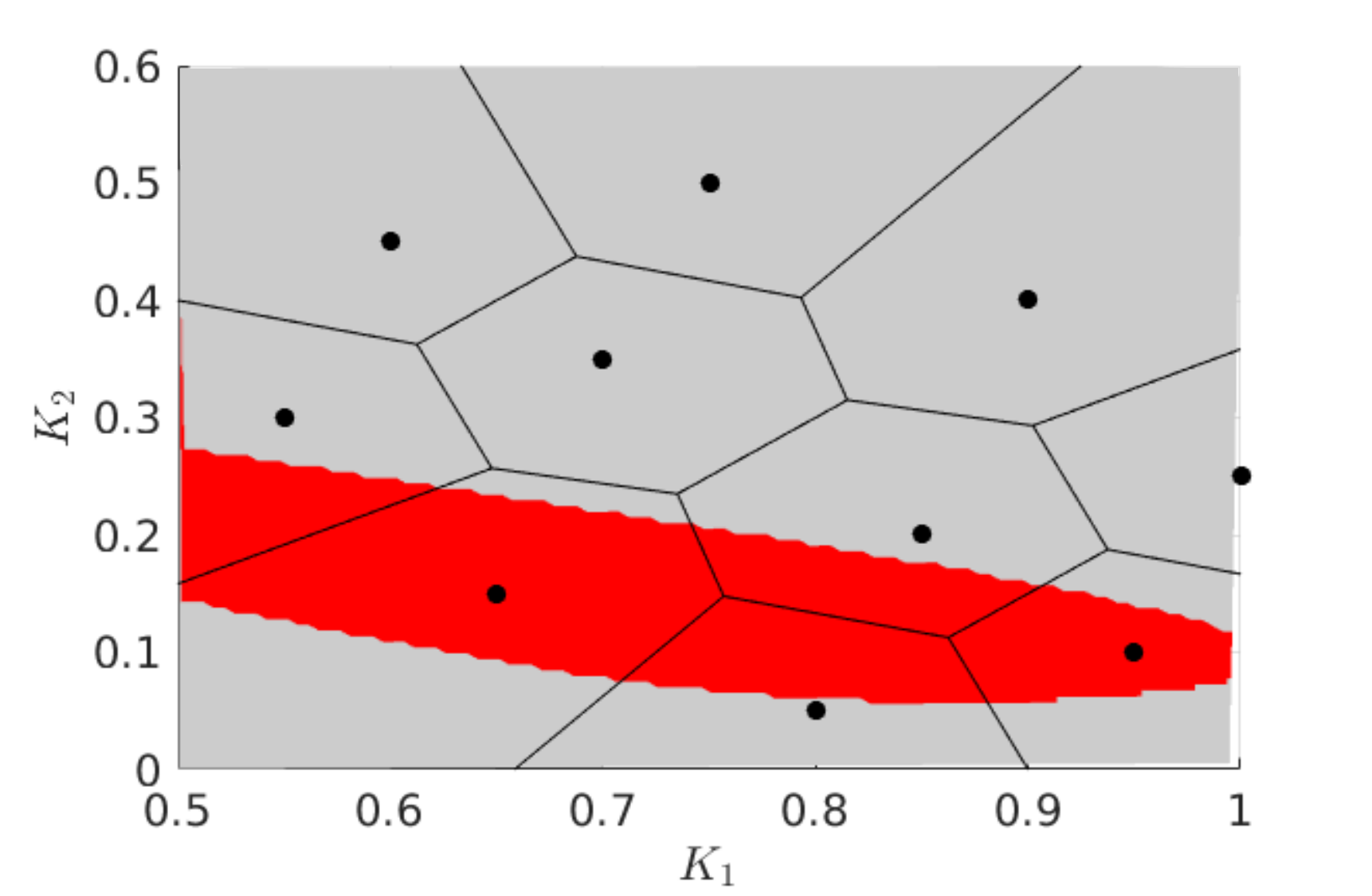} 
\subcaption{Initial samples}\label{fig:MIVOINITIAL}
\end{subfigure}%
\begin{subfigure}[t]{0.5\textwidth}
\includegraphics[scale=0.4]{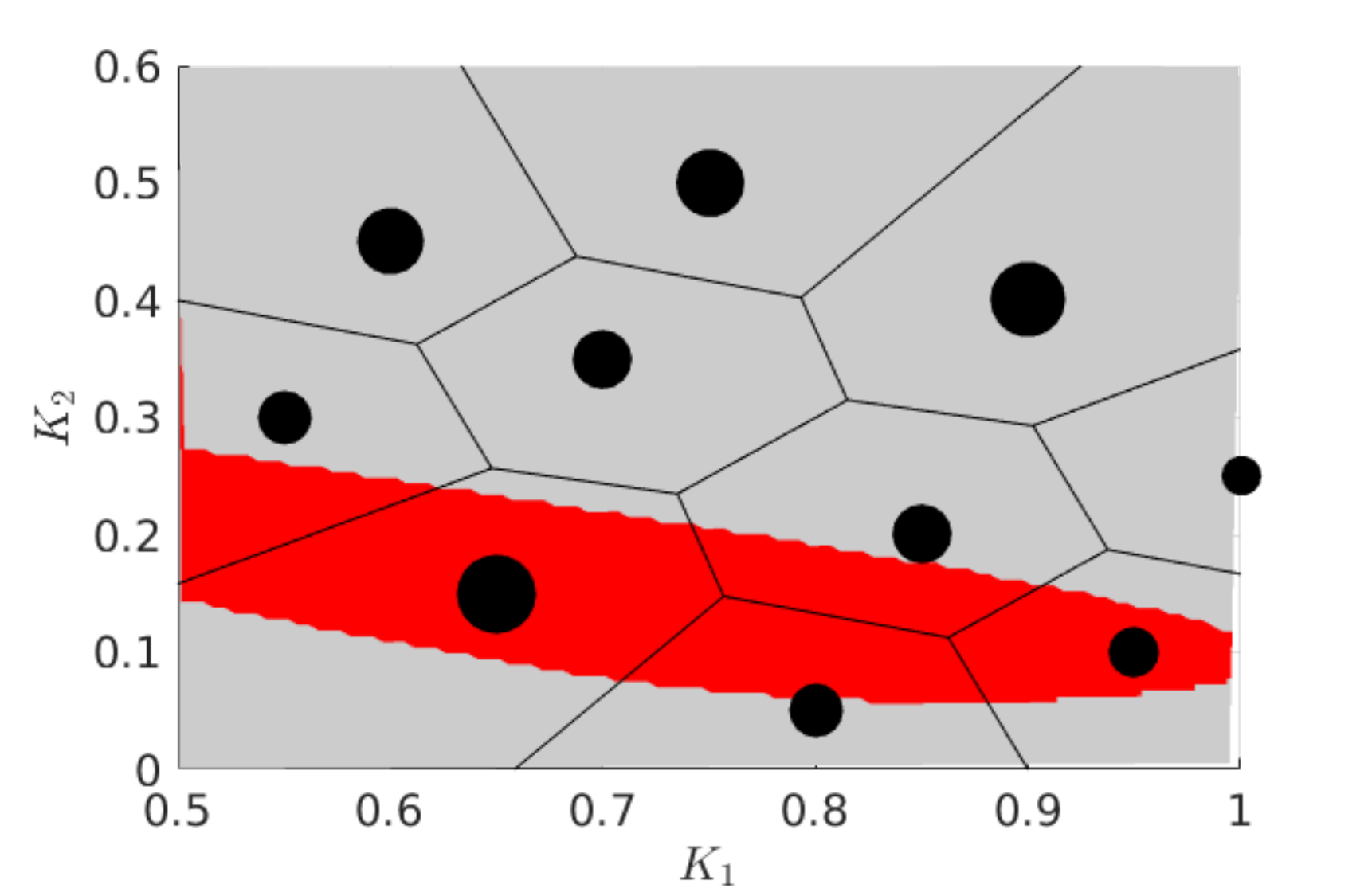} 
\subcaption{Volumes of Voronoi cells}\label{fig:MIVOVolume}
\end{subfigure}
\begin{subfigure}[t]{0.5\textwidth}
\includegraphics[scale=0.4]{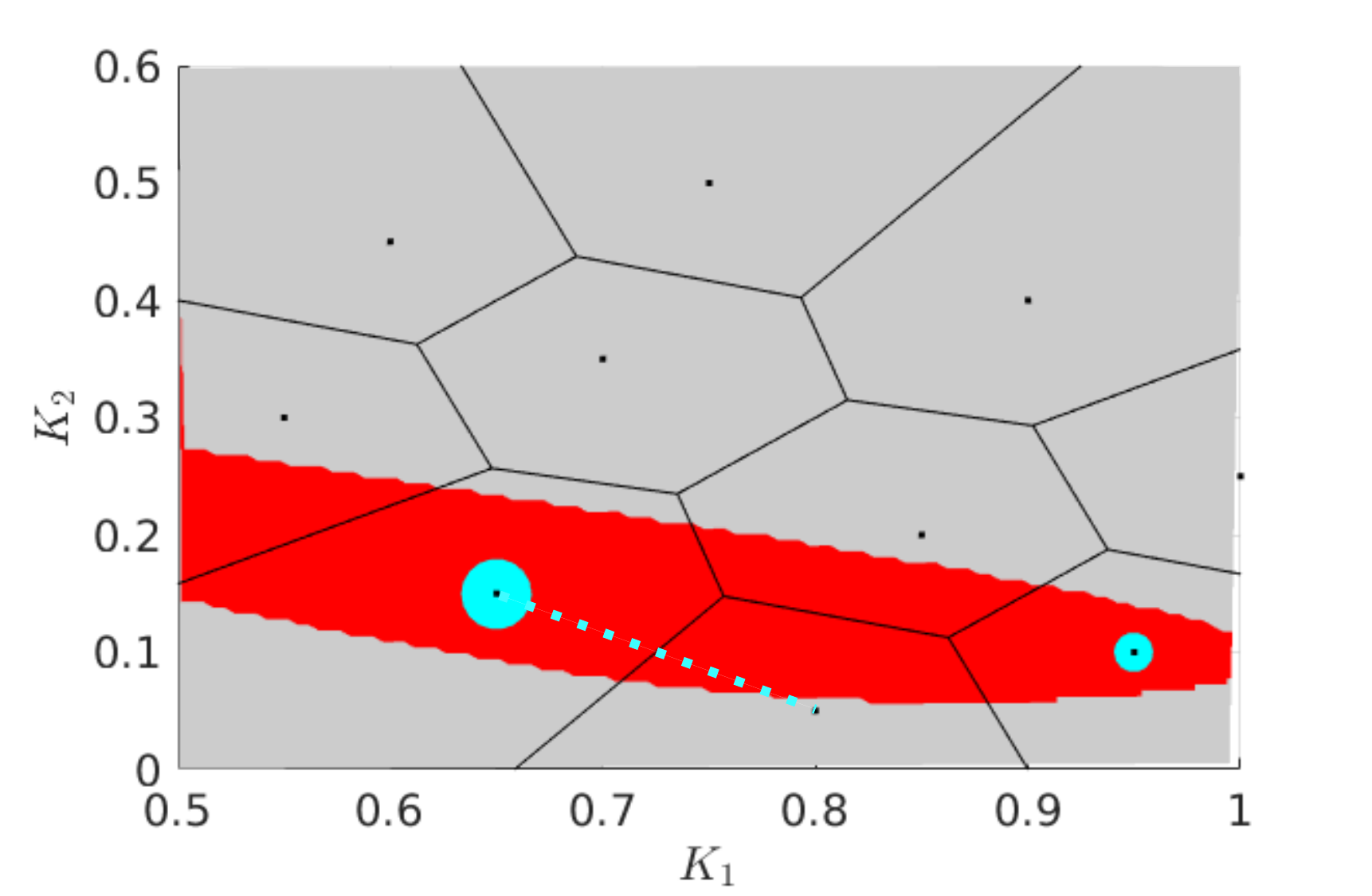} 
\subcaption{Scores}\label{fig:MIVOScore}
\end{subfigure}%
\begin{subfigure}[t]{0.5\textwidth}
\includegraphics[scale=0.4]{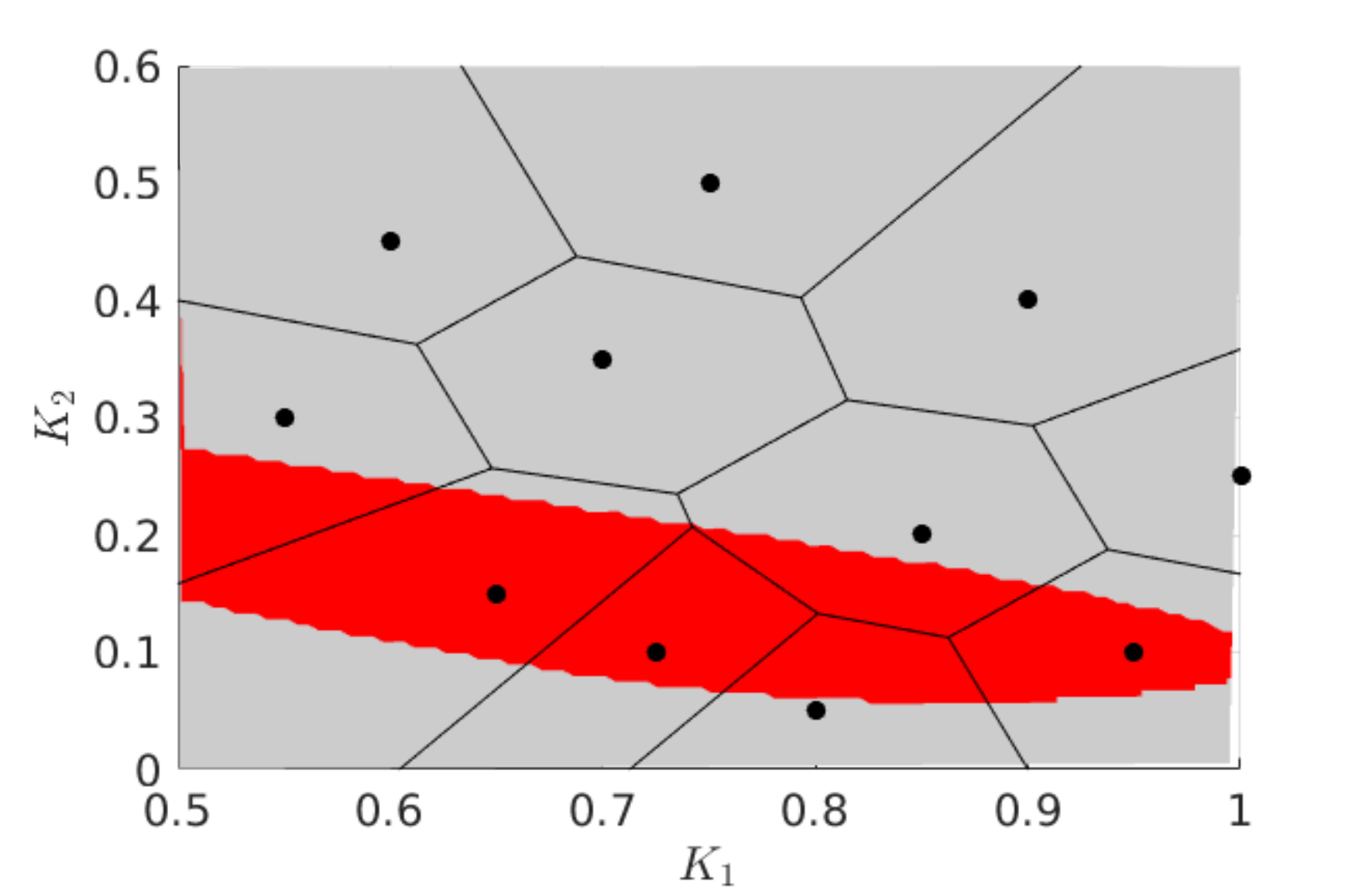} 
\subcaption{New point sampled}\label{fig:MIVONewPoint}
\end{subfigure}
\caption[Process of sampling a new point with MIVor]{Process of sampling a new point with MIVor over contour of LLE classification. (a) Initial samples and Voronoi cells, (b) Size of sample proportional to estimated volume of its Voronoi cell, (c) Size of samples proportional to their Score value (in cyan). Dotted line indicates closest point to highest scoring sample, (d) New point created in the middle of the dotted line of (c).  }\label{fig:MIVORONEStep}
\end{figure}

Let $n_{chaotic}$ be the number of samples with $\mathcal{C}_{MoB}(\bm{x}^{(i)})=1$ for $i=1, \ldots, m$. Define the set $\mathcal{X}_{chaotic}$, that contains all samples indicating chaotic motion and let $\mathcal{X}_{regular} = \mathcal{X} \setminus \mathcal{X}_{chaotic}$ be the set containing the parametric values yielding regular motion. 
Let the neighboring points of sample $\bm{x}^{(i)}$ be defined by its $2*n$ nearest points by euclidean distance of the set $\mathcal{X} \setminus \bm{x}^{(i)}$, where $n$ is the dimension of $\bm{x}^{(i)}$. \\ The exploitation component of MIVor is based the following ideas:
\begin{itemize}
\item The volumes of the Voronoi cells of samples $\bm{x} \in \mathcal{X}_{chaotic}$ are a measure for the density of samples with $\mathcal{C}_{MoB}(\bm{x}^{(i)})=1$. The larger the volume of a particular cell the more uncertain the chaotic region around this respective sample and hence, a new point needs to be created.
\item The more points in the neighborhood of a sample $\bm{x} \in \mathcal{X}_{chaotic}$ are in the set $\mathcal{X}_{regular}$ the closer $\bm{x}$ is to an edge.
\end{itemize}
MIVor combines these two ideas by ranking the points $\bm{x} \in \mathcal{X}_{chaotic}$ in a combined score of their volume and the amount of neighboring samples with a positive outcome. The samples are ranked by the multiplicative score 
\begin{equation}\label{eq::SamplePointRank}
R(\bm{x}^{(i)}) = Vol_{i} \mathcal{N}_{i}, \qquad \forall \bm{x}^{(i)} \in \mathcal{X}_{chaotic}
\end{equation}
where $\mathcal{N}_{i}$ is the number of points part of the set $\mathcal{X}_{regular}$ in the neighborhood of $\bm{x}^{(i)}$.
As described in section \ref{sec::Voro} a Monte Carlo approach is employed in order to evaluate the volumes of the Voronoi cells. \\
The sample point with the highest score  is evaluated
\begin{equation}
\bm{x}^{max} = \max_{\bm{x}^{\star} \in \mathcal{X}_{chaotic}} R(\bm{x}^{\star}).
\end{equation}
The Monte Carlo points defining the cell $C_{max}$ around this point are stored in the set $P_{max} \subset P$. \\
In order to estimate the location of the edge between two regions accurately the Monte Carlo point of the set $P_{max}$ that is closest to a sample point with positive outcome (from the set $\mathcal{X}_{regular}$) is taken as a candidate $\bm{x}^{cand}$ for the next added sample. 
\begin{equation}
\bm{x}^{cand} = \min_{\bm{p}  \in P_{max}} \norm{\bm{p} - \bm{x}^{(i)}} \, \,  \forall \bm{x}^{(i)} \in \mathcal{X}_{regular}.
\end{equation}
In order to avoid clustering of sample points and hence numerical issues a space-filling metric $S$ is introduced with 
\begin{equation}
\begin{aligned}
ds(\bm{x}^{(i)}) &= \min \left( \norm{\bm{x}^{(i)} - \bm{x}^{{j)}}} \right), && \forall \bm{x}^{(i)} \in \mathcal{X} \cap (i \neq j) \\
S &= 0.1 \max (ds(\bm{x}^{(i)}) ), && \forall \bm{x}^{(i)} \in \mathcal{X}.
\end{aligned}
\end{equation}
If the candidate point $\bm{x}^{cand}$ is closer than $S$ to an existing sample point, i.e.
\begin{equation}
\exists \, \bm{x}^{(i)} \in \mathcal{X} \, \text{with} \, \norm{\bm{x}^{(i)} - \bm{x}^{cand}}< S,
\end{equation} 
it will be rejected. The new candidate point is the Monte Carlo point of the set $P_{max}$ with the highest variance 
\begin{equation}
\bm{x}^{cand} = \max_{\bm{p} \in P_{max}} \, \sigma^{2}(\bm{p}).
\end{equation}
 If $\bm{x}^{cand}$ still violates the distance constraint, MIPT is employed.  Otherwise the candidate point is taken as the new sample point. The workflow for deciding on a new sample point is illustrated in Figure \ref{fig::flowchart_MIVOrSample}.
\begin{figure}[h!]
\tikzstyle{decision} = [rectangle, draw,  
    text width=20em, text centered, rounded corners, node distance=2cm, inner sep=0pt,minimum height=3em]
\tikzstyle{block} = [rectangle, draw, fill=blue!20, 
    text width=20em, text centered, rounded corners, minimum height=4em,node distance=4cm,]
\tikzstyle{blockA} = [rectangle, draw, fill=blue!10, 
    text width=20em, text centered, rounded corners, minimum height=3em]
\tikzstyle{line} = [draw, -latex']
\tikzstyle{cloud} = [draw, ellipse,fill=red!20, node distance=8.0cm,
    minimum height=2em]
 \centering 
\begin{tikzpicture}[node distance = 1.5cm, auto,scale=0.8, transform shape]
    \node [block] (init) {
\begin{tabular}{ l r }
Candidate point & $\bm{x}^{cand}$ \\
Space-filling metric & $S$ \\
Set of Monte Carlo points & $P_{max}$ \\
Set of samples & $\mathcal{X}$
\end{tabular}
};
    \node [decision, below of=init,node distance=3.0cm] (start) {Check if: \\
    $\exists \, \bm{x}^{(i)} \in \mathcal{X} \, \text{with} \, \norm{\bm{x}^{(i)} - \bm{x}^{cand}}< S$};
    
 \node [cloud, right of=start] (Done1) {$\bm{x}^{(m+1)} = \bm{x}^{cand}$};    
    
    \node [blockA, below of=start,node distance=2.5cm] (Evaluate) {$\bm{x}^{cand} = \max_{\bm{p} \in P_{max}} \, \sigma^{2}(\bm{p})$  };
    
   \node [decision, below of=Evaluate,node distance=2.5cm] (Check1) {Check if: \\
    $\exists \, \bm{x}^{(i)} \in \mathcal{X} \, \text{with} \, \norm{\bm{x}^{(i)} - \bm{x}^{cand}}< S$};

     \node [cloud, right of=Check1] (Done2) {$\bm{x}^{(m+1)} = \bm{x}^{cand}$};    
    
    \node [cloud, below of=Check1,node distance=2.0cm] (Update) {Find $\bm{x}^{(m+1)}$ with MIPT.};

    \path [line] (init) -- (start);
     \path [line] (start) -- node {no}(Done1);
     \path [line] (start) -- node {yes}(Evaluate);
    \path [line] (Evaluate) -- (Check1);
      \path [line] (Check1) -- node {no}(Done2);
     \path [line] (Check1) -- node {yes}(Update);

\end{tikzpicture}
\caption[Workflow of finding a new sample point with MIVor.]{Workflow of finding a new sample point $\bm{x}^{(m+1)}$ from the candidate point with MIVor.}\label{fig::flowchart_MIVOrSample}
\end{figure}
 The general procedure of the exploitation step is illustrated in Figure \ref{fig:MIVORONEStep}. Consider ten initial samples as shown in Figure \ref{fig:MIVOINITIAL} over the presented LLE classification problem of Figure \ref{fig:LLEExample1_redgray}. The Voronoi tessellation with its respective cells around each sample point is depicted with black lines. It can be seen that two initial points are inside the red area. The estimated volumes of the cells (see Box \ref{alg::EstimationVolume}) are represented proportionally by the area of the points of Figure \ref{fig:MIVOVolume}. As there are only two points with negative outcome the score only needs to be calculated twice. The respective score values of the two points with negative outcome are shown in Figure \ref{fig:MIVOScore}. Here the area around the points (in cyan) is proportional to the score value. In a next step, the neighbors of the sample point with the highest score are scanned for sample points with positive outcome.\\The closest point is highlighted by the dotted line. As seen in Figure \ref{fig:MIVONewPoint} the next sample point is chosen to be the Monte Carlo point, in the Voronoi cell of the sample with the highest score, that is closest to this neighbor.

Summarized, the procedure is defined by the following steps:
\begin{enumerate}
\item[] \textbf{Create initial metamodel $\hat{\mathcal{M}}$}.\\ With the initial design of experiments $\mathcal{D} = \lbrace \mathcal{X}, \mathcal{Y} \rbrace$ generate an initial metamodel. Define rate of exploration $r$ and reduction factor $\alpha$. 
\item[] \textbf{Check number of samples indicating chaotic motion.}\\ In the input set check for the number of samples indicating chaotic motion $n_{chaotic}$. As long as  $n_{chaotic} = 0$, sample new points with MIPT. As soon as  $n_{chaotic}>0$ go to the next step.
\item \textbf{Exploration or Exploitation.}\\
 Sample a uniform random variable $u \sim U[0,1]$, if $u <r$, find the next sample point by MIPT and reduce $r$ by factor $\alpha$ else go to step 2.  
\item \textbf{Rank the existing sample points of set $\mathcal{X}_{chaotic}$.} \\Identify $\mathcal{X}_{chaotic}$ and $\mathcal{X}_{regular}$. Evaluate the volumes of the Voronoi cells of samples in $\mathcal{X}_{chaotic}$. Rank the $n_{chaotic}$ points according to the formula of equation (\ref{eq::SamplePointRank}). Identify the highest scoring sample $\bm{x}^{max}$. Store the Monte Carlo points of $\bm{x}^{max}$ in $P_{max}$.
\item \textbf{Sample in Voronoi cell with highest score} \\Find the closest point in $\mathcal{X}_{regular}$ in the neighborhood of $\bm{x}^{max}$. Set this point as the candidate point $\bm{x}^{cand}$.
Find the new sample point by following the workflow of Figure \ref{fig::flowchart_MIVOrSample}. Go back to step 2.
\end{enumerate}

The following section validates the presented method and compares the novel approach with respect to the commonly used adaptive sampling techniques previously introduced in Table \ref{table::OverViewMethods}.
\subsection{One-dimensional LLE classification}
\begin{figure}[h!]
\centering
\begin{subfigure}[t]{0.5\textwidth}
\includegraphics[scale=0.43]{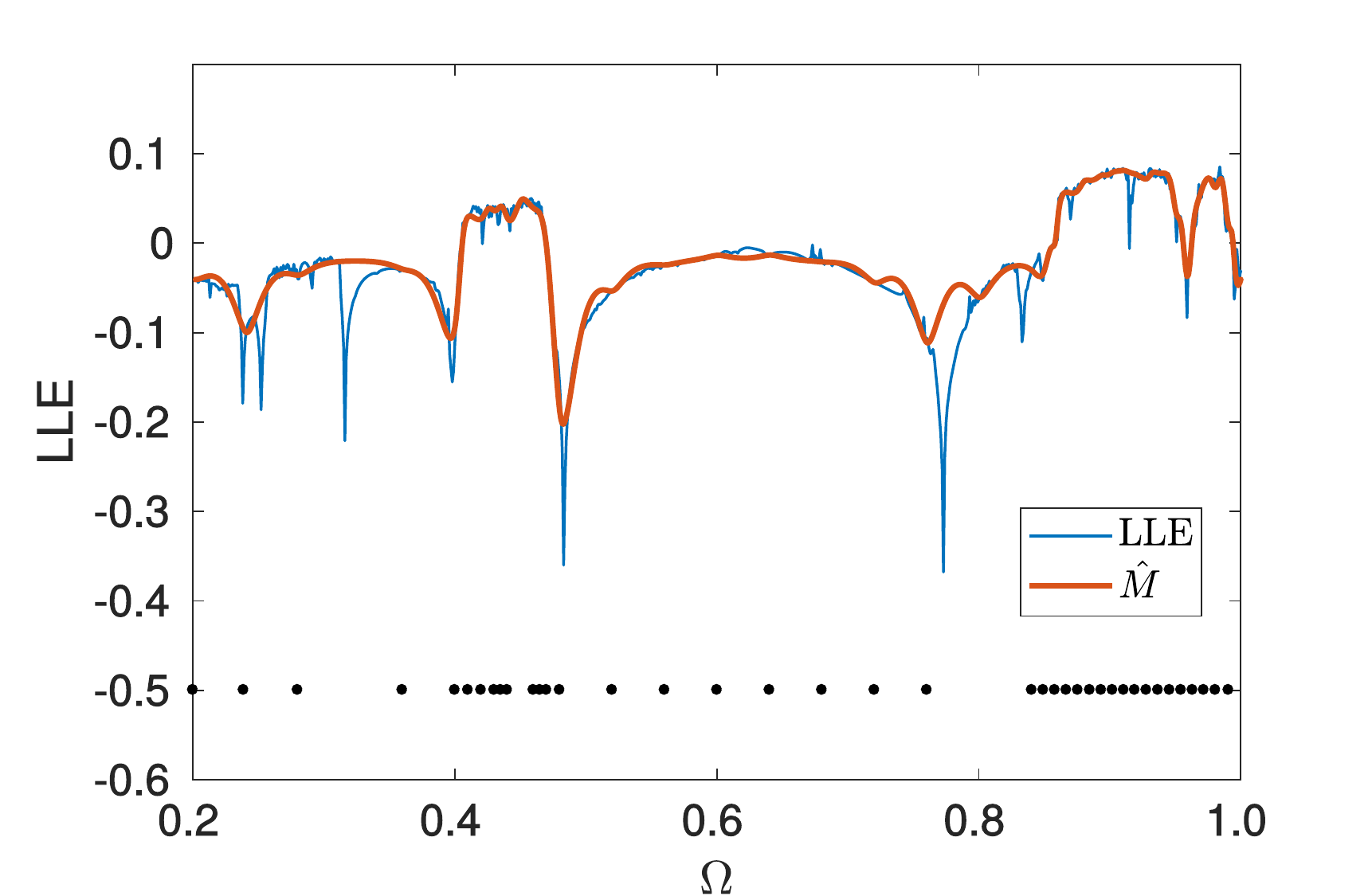} 
\subcaption{}\label{fig::LLE1d_plot}
\end{subfigure}%
\begin{subfigure}[t]{0.5\textwidth}
\includegraphics[scale=0.43]{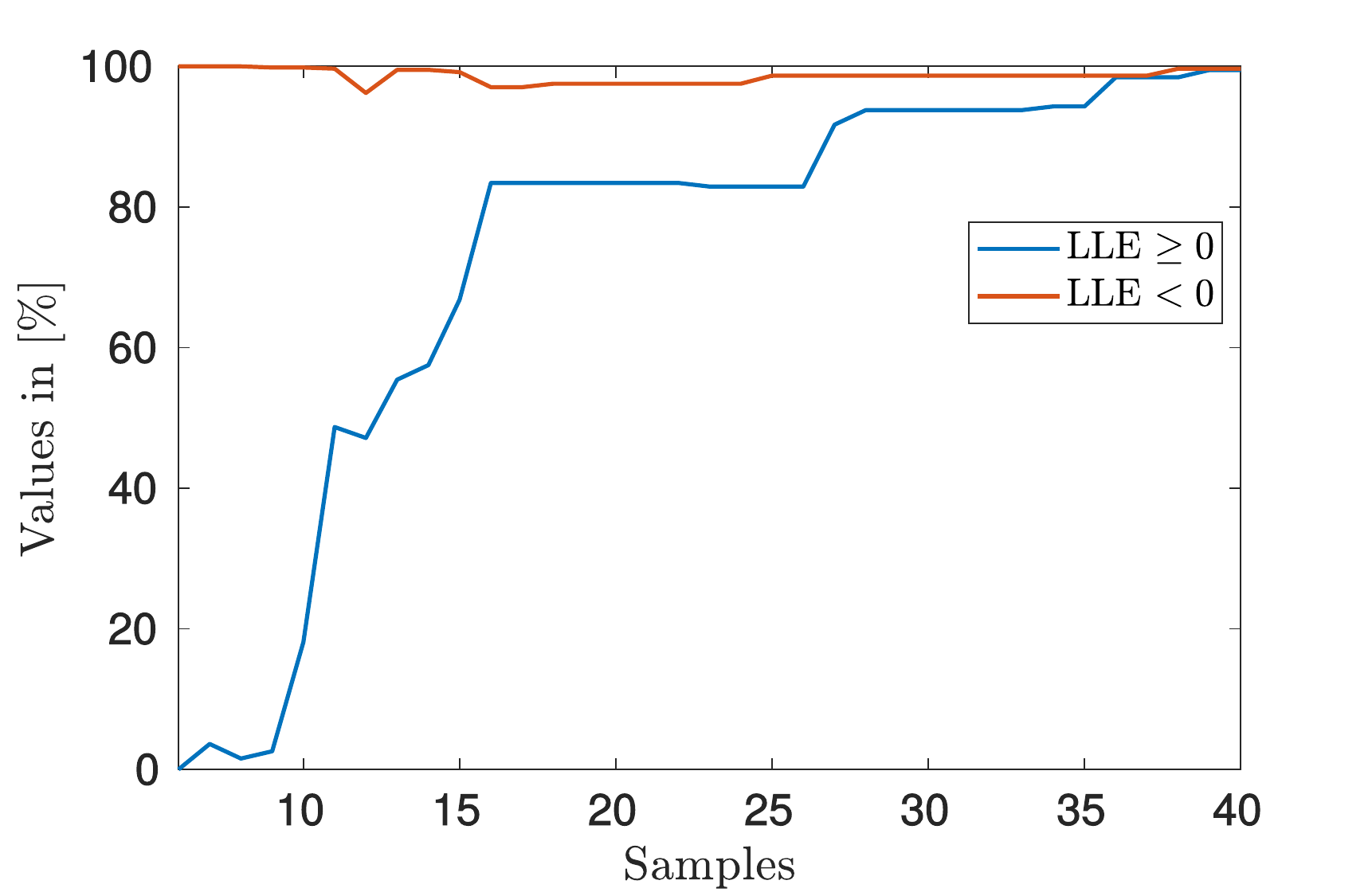} 
\subcaption{}\label{fig:LLE1D_percent}
\end{subfigure}
\caption[One-dimensional LLE example]{One-dimensional LLE example. $\Omega$ is given in $\text{rad}/\text{s}$, (a) LLE value and metamodel for classification, (b) Evolution of the perfomance of MIVor with the number of samples.}\label{fig::LLE1d}
\end{figure}

MIVor is first tested on the one-dimensional LLE values as given in Figure \ref{fig:LLEOverBifurcationMyProblem}. The input parameter values are given by 
$K_{1} =  1.0\, \text{N}/\text{m}^{3}$
and $K_{2} =  0.0 \, \text{N}/\text{m}$,
$\mu_{k} = 0.15$,
$M = 1 \, \text{kg}$,
$V_{0} = 0.1 \, \text{m}/\text{s}$,
$D = 0.0 \, \text{Ns}/\text{m}$, 
$\mu_{s} = 0.3$,
$V_{s} = 0.1 \, \text{m}/\text{s}$,
$U_{0} = 0.1 \,\text{N}$,
$N_{0} = 1.0 \, \text{N}$,
$\sigma_{0} = 100.0 \, \text{N}/\text{m} $,
$\sigma_{1} = 10.0 \,  \text{Ns}/\text{m}$ and
$\sigma_{2} = 0.1 \, \text{Ns}/\text{m}$.  As seen in Figure \ref{fig:LLEOverBifurcationMyProblem}, $\Omega$ is varies between $0.2$ and $1.0\, \text{rad}/\text{s}$.
 Five initial sample points are set with TPLHD, i.e. $[0.2,0.36,0.52, 0.68, 0.84]^{T}$. None of them yields a chaotic output. This is done to show the exploration component of the technique. \\
 5000 samples points are created inside the domain to test the performance of the metamodel. A metamodel for classification is judged by how many of the points yielding $\mathcal{C}_{MoB}=0$ on one hand and how many points yielding $\mathcal{C}_{MoB}=1$ on the other hand are predicted.
The measure is stated in percent of accurately predict points. 
 The process is stopped when both percentages yield a value of above $99\%$. This means that the probability to correctly classify a point in its chaotic or regular motion with over $99\%$. 
The evolution of the accuracy with regards to the sample size is shown in Figure \ref{fig:LLE1D_percent}. After 39 samples, i.e. 35 samples have been added to the initial samples, the procedure stops. Finally, $99.48\%$ of the 5000 random points that yield LLE$\geq 0$ are evaluated correctly with the metamodel and $99.67\%$ of the points with LLE$<0$.

The target LLE plot, the metamodel $\hat{M}$ after 39 samples and the respective sample positions (black dots) at the end of the adaptive process are shown in Figure \ref{fig::LLE1d_plot}. It can be seen that a proficient metamodel has been created. Furthermore as intended with MIVor most of the created samples lie in the area that indicates chaotic behavior.
\subsection{Two-dimensional LLE classification}
\begin{figure}[h!]
\centering
\begin{subfigure}[t]{0.5\textwidth}
\includegraphics[scale=0.43]{Bilder/LLE/KSETA06_plot2.jpg} 
\subcaption{$\mathcal{M}_{LLE,MoB}$}\label{fig:KSETA06_plot}
\end{subfigure}%
\begin{subfigure}[t]{0.5\textwidth}
\includegraphics[scale=0.43]{Bilder/LLE/KSETA06_plot_redGray2.pdf} 
\subcaption{Classification provided by $\mathcal{M}_{LLE,MoB}$}\label{fig:KSETA06_redGray}
\end{subfigure}
\begin{subfigure}[t]{0.5\textwidth}
\includegraphics[scale=0.43]{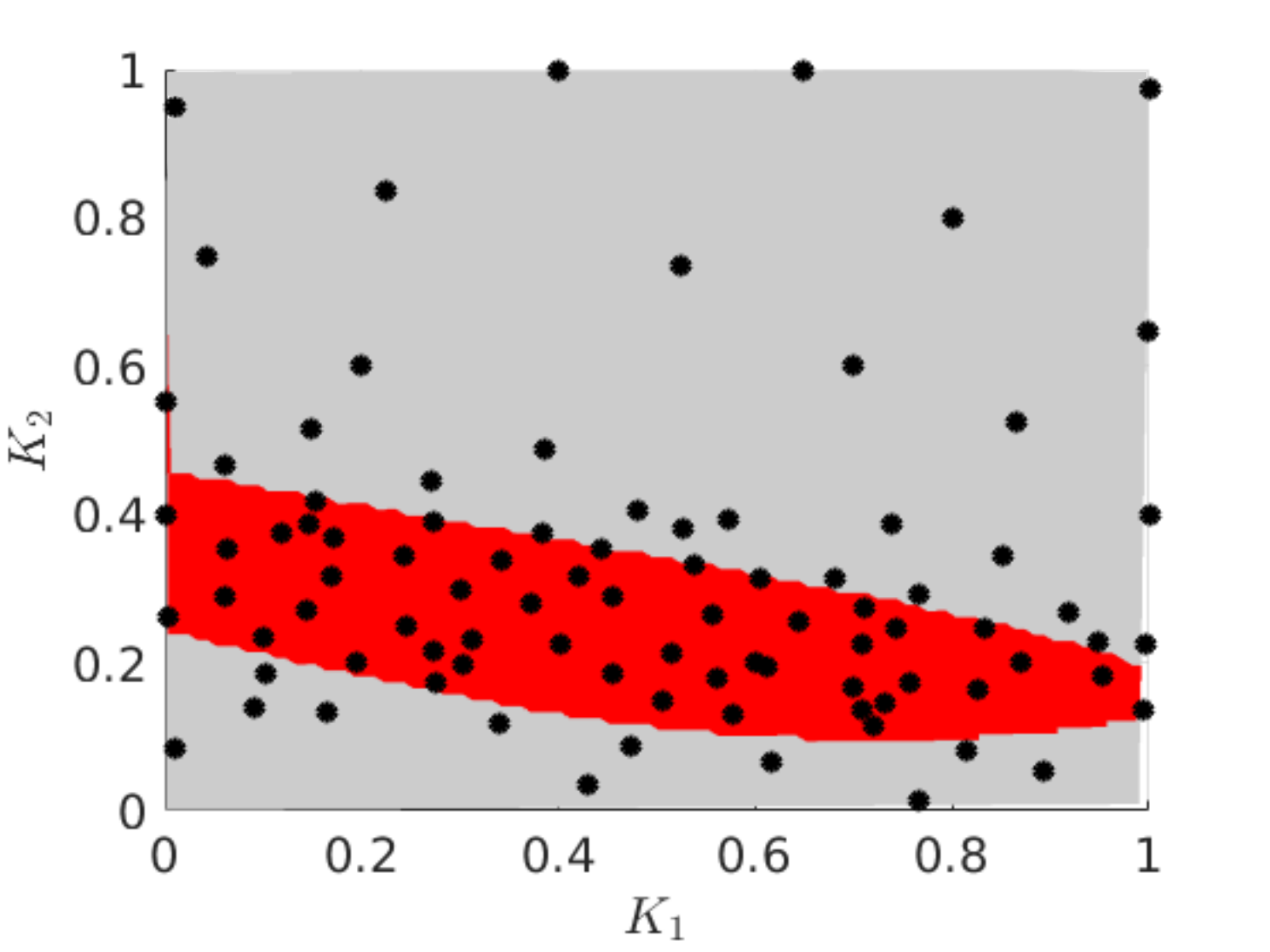} 
\subcaption{90 MIVor samples}\label{fig:KSETA06_Samples}
\end{subfigure}%
\begin{subfigure}[t]{0.5\textwidth}
\includegraphics[scale=0.43]{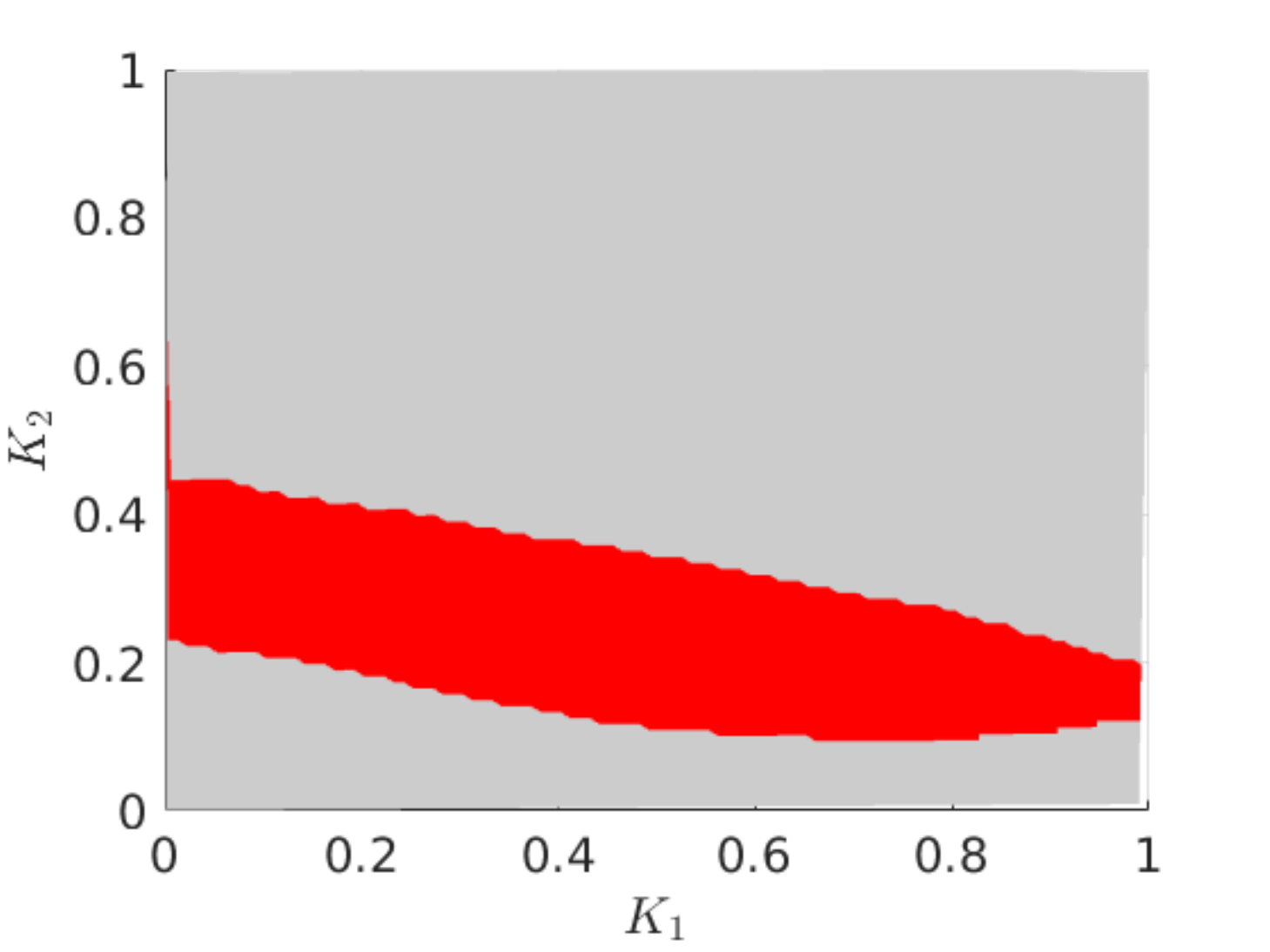} 
\subcaption{Classification provided by $\hat{\mathcal{M}}_{LLE,MoB} $ with 90 samples}\label{fig:KSETA06_Meta}
\end{subfigure}
\caption[Results and data for the 2D LLE problem case $\mathcal{P}_{1}$]{Results and data for the 2D problem case $\mathcal{P}_{1}$. $K_{1}$ is given in $\text{N}/\text{m}^{3}$, $K_{2}$ in $\text{N}/\text{m}$.LLE is unitless. }
\end{figure}

In this thesis MIVor is restricted to two-dimensional applications. Higher dimensions are out of the scope of this thesis.\\
The investigated is defined by equation (\ref{eq::dynamic_system}) with parameters 
$M = 1 \, \text{kg}$,
$V_{0} = 0.1 \, \text{m}/\text{s}$,
$D = 0.0 \, \text{Ns}/\text{m}$, 
$\mu_{s} = 0.3$,
$V_{s} = 0.1 \, \text{m}/\text{s}$,
$U_{0} = 0.1 \,\text{N}$,
$N_{0} = 1.0 \, \text{N}$,
$\sigma_{0} = 100.0 \, \text{N}/\text{m} $,
$\sigma_{1} = 10.0 \,  \text{Ns}/\text{m}$ and
$\sigma_{2} = 0.1 \, \text{Ns}/\text{m}$.  
Hereafter, four different problem settings with increasing difficulty are investigated. The methods are compared according to their ability to correctly predict if input data yields regular or chaotic motion, by considering 12000 points which are space-fillingly placed into the input domain with TPLHD. For each of these points the output LLE is calculated by the surrogate model and by the full model for comparison. \\
The following computations have been repeated 10 times to avoid performance outliers. The average of these 10 iterations has been taken for evaluating the performance. Illustrated sample positions have been  taken randomly from one of iteration set. \\ \\
\textbf{Problem $\mathcal{P}_{1}$}\label{sec::FirstProblem} \\
\\
Consider the input domain for the two spring-stiffnesses given by\\$K_{1} \in  [0.5 , 1.0]\, \text{N}/\text{m}^{3}$
and $K_{2} \in [0.0,0.6] \, \text{N}/\text{m}$ with 
$\Omega = 0.6 \, \text{rad}/\text{s}$ and
$\mu_{k} = 0.15$. The LLE over this domain is plotted in Figure \ref{fig:KSETA06_plot}. The classification problem is displayed in Figure \ref{fig:KSETA06_redGray}. It can be seen that the shape of the area belonging to LLE values above or equal $0$ is fairly simple. Values with $LLE\geq 0$ lie predominantly in the middle of the given domain and are shaped like an ellipse. However there is a small chaotic part around $(0.0,0.3)$ that is separated from the main chaotic regime. Here, common adaptive sampling techniques with an exploration component will sample in the red area. Therefore this a problem where usual space-filling techniques should be able to generate a proficient classification model for the $\mathcal{M}_{LLE,MoB}$. \\
The problem is studied until 90 samples are reached starting from 5 sample created with TPLHD. \\
The sample position of the 90 MIVor samples are displayed in  Figure \ref{fig:KSETA06_Samples}. It can be noticed that most of the generated points are located around the red area that indicates chaotic motion. The metamodel of the classified LLE values with the 90 samples is shown in Figure \ref{fig:KSETA06_Meta}. It can be seen that full model, see Figure \ref{fig:KSETA06_redGray}, can be proficiently approximated. 
\newpage
The MAE error after 90 sample points for the respective adaptive sampling techniques including MIVor as well as the initial model and a one-shot TPLHD model with 90 samples is listed in Table \ref{table::KSeta06}. Furthermore the amount of correctly correctly classified points for $LLE \geq 0$ and  $LLE < 0$ are given. It can be seen that MIVor achieves the best results out of all tested sampling techniques with over $99\%$ of points correctly identified for both cases. \\
However for this simple example the space-filling techniques (MIPT and MSD) as well as SSA and AME yield proficient results with all being able to classify more than $95\%$ of cases accurately.  Nevertheless, consistently accurate classification (similar to MIVor) is only possible with a high number of sample points for these techniques. 
\begin{table}[ht!]
\begin{center}
\resizebox{1.0\textwidth}{!}{%
\begin{tabular}{l|l c c c } \hline
& Method & MAE  & Equal or Above 0[$\%$ ] & Below 0[$\%$ ] \\ \hline\hline \\
\multirow{1}{*}{\shortstack[l]{Values after\\5 samples}} & TPLHD &0.0114  & 81.40 & 84.73 \\ \\  \hline \\
\multirow{14}{*}{\shortstack[l]{ Values after\\90 samples}} &TPLHD  &0.0024  & 94.34 & 98.25 \\
&ACE & 0.0223  & 11.07 & 0.9972 \\
&AME &   0.0024  & 96.15 & 99.63 \\
&CVD   &0.0019& 96.07 & 99.57\\
&CVVor & 0.0767  & 55.37 & 76.90 \\ 
&EI &   0.0882 & 0.04 & 100.0 \\
&EIGF   & 0.0077  & 22.56 & 100.0  \\
&MASA    & 0.0424  & 1.03 & 100.0 \\
&MEPE    &0.0023  & 88.38 & 99.33\\
&MIPT    & 0.0030  & 96.86 & 99.42\\ 
&MSD   &  0.0042  & 95.28 & 99.46  \\
&SFCVT &0.0028  & 85.00 & 99.64 \\ 
&SSA    & 0.0021  & 95.20 & 98.01 \\   \hline \\
&MIVor &0.0074 & \textbf{99.21} & \textbf{99.79}
\end{tabular}
}
\end{center}
\caption[Error measures for first 2D LLE problem case.]{Error measures for first 2D LLE classification problem case after 90 samples.}\label{table::KSeta06}
\end{table}

\begin{figure}[h!]
\centering
\includegraphics[scale=0.4]{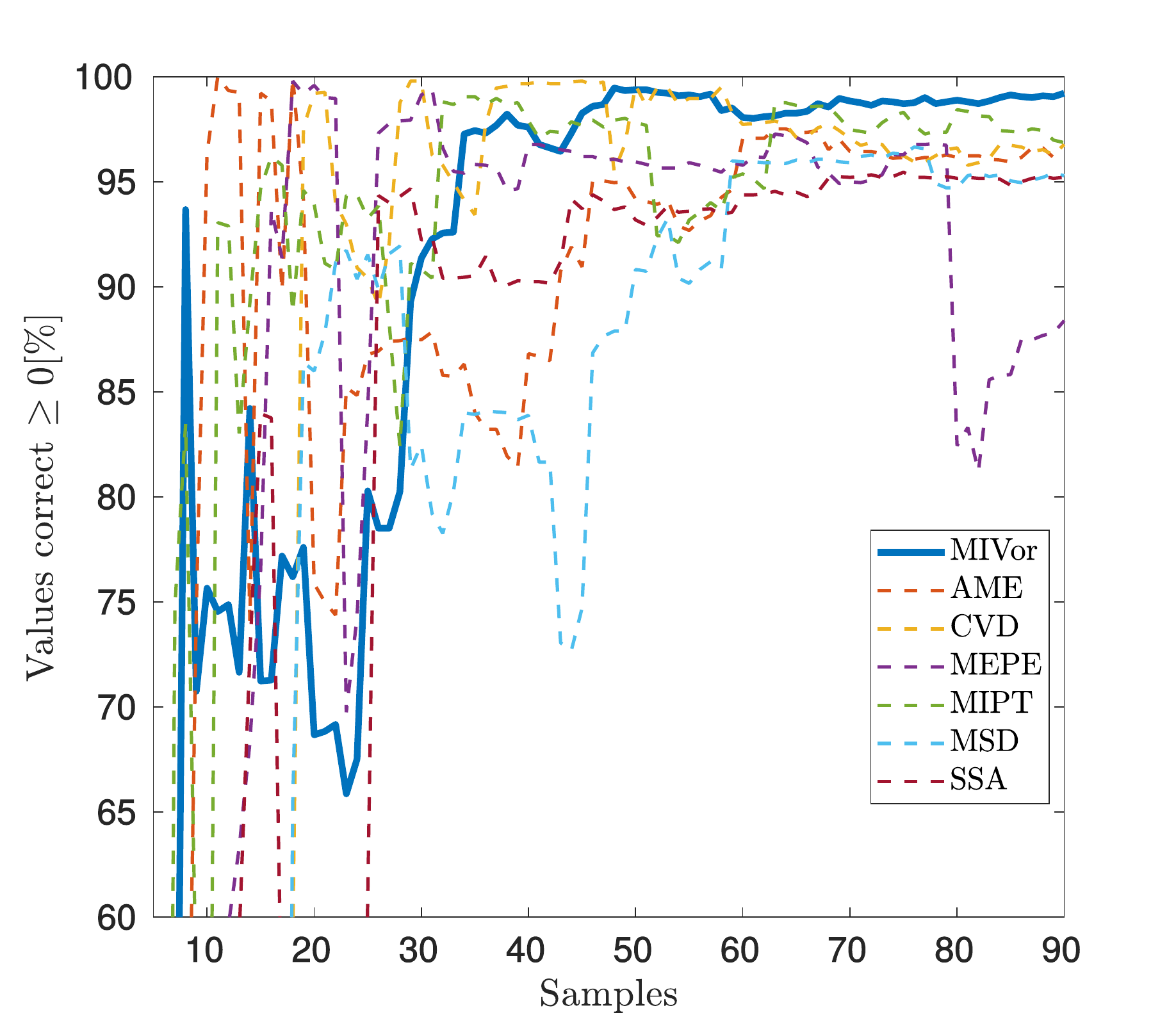}
\caption[Convergence of correctly classified points with $LLE \geq 0$ for 2D LLE case $\mathcal{P}_{1}$]{Convergence of correctly classified points with $LLE \geq 0$ for 2D LLE case $\mathcal{P}_{1}$.}\label{fig::KSEta06Above}
\end{figure}

\begin{figure}[b!]
\centering
\begin{subfigure}[t]{0.5\textwidth}
\includegraphics[scale=0.32]{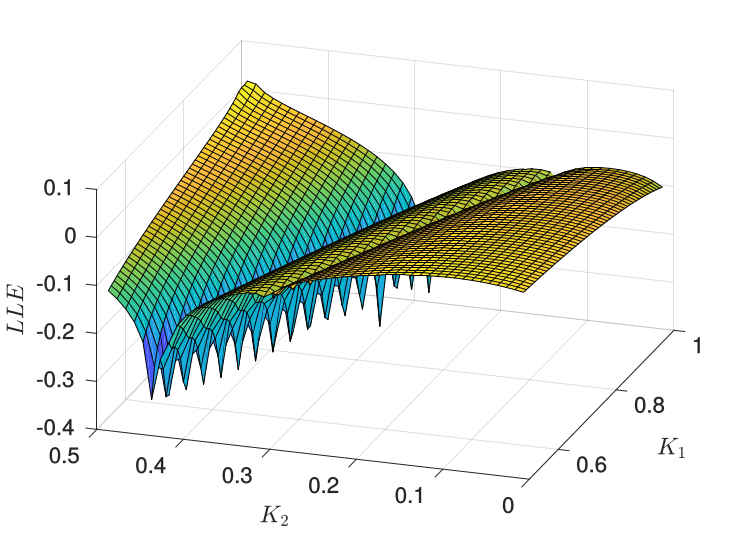} 
\subcaption{$\mathcal{M}_{LLE,MoB} $}\label{fig:KSETA06SmallPlot}
\end{subfigure}%
\begin{subfigure}[t]{0.5\textwidth}
\includegraphics[scale=0.32]{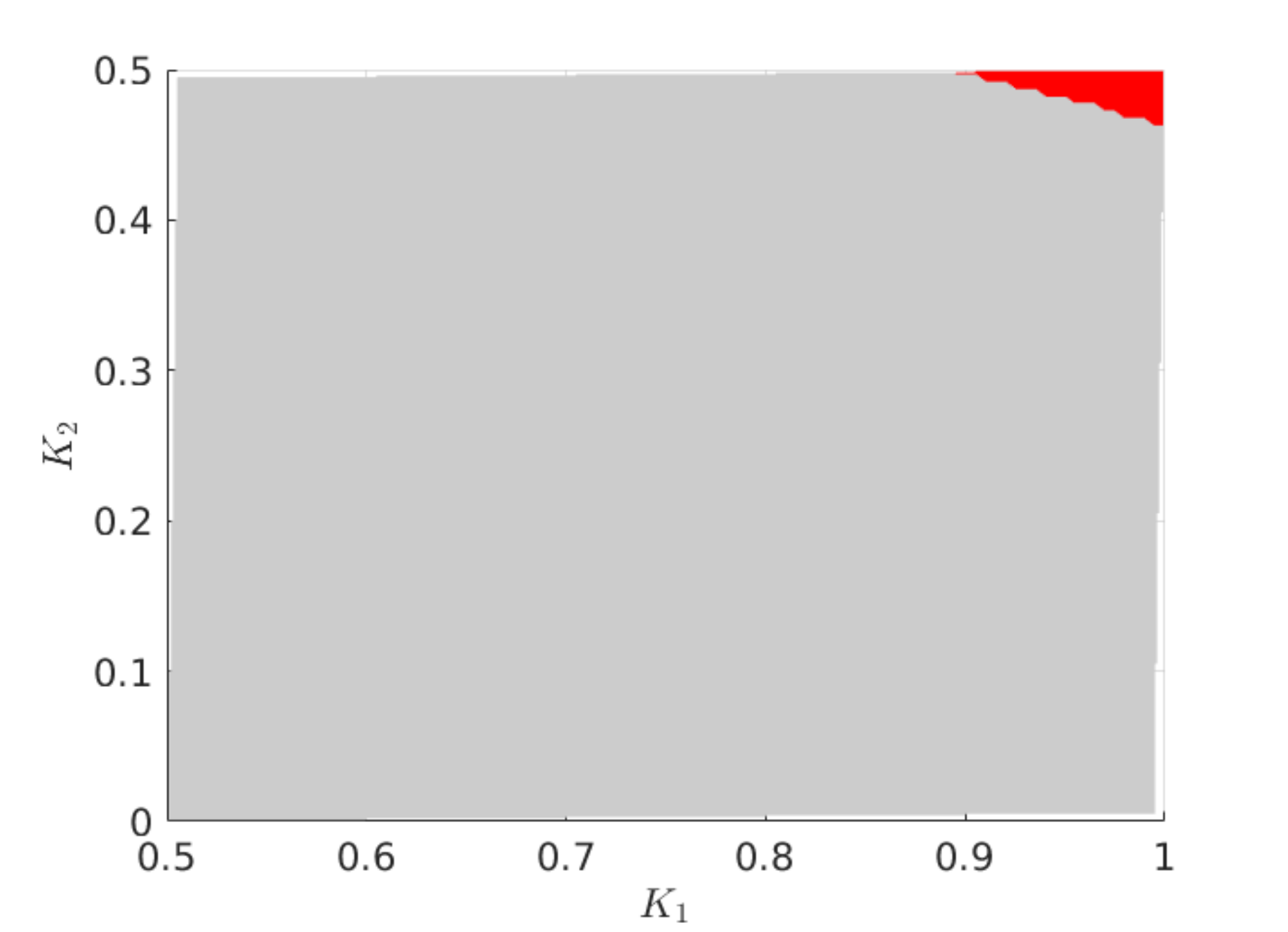} 
\subcaption{ Classification provided by $\mathcal{M}_{LLE,MoB} $}\label{fig:KSETA06SmallRedGray}
\end{subfigure}
\begin{subfigure}[t]{0.5\textwidth}
\includegraphics[scale=0.4]{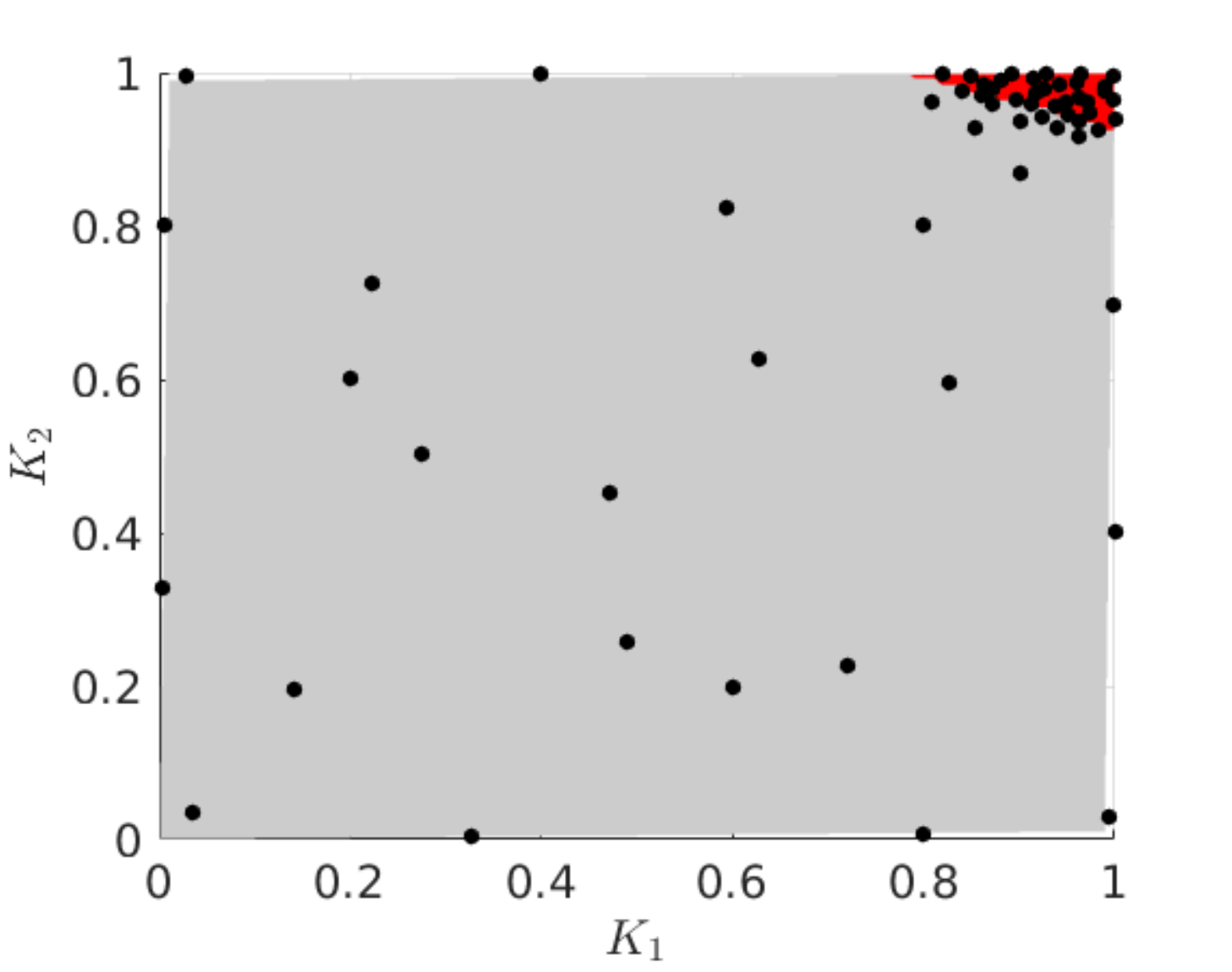} 
\subcaption{65 MIVor samples}\label{fig:KSETA06SmallSamples}
\end{subfigure}%
\begin{subfigure}[t]{0.5\textwidth}
\includegraphics[scale=0.42]{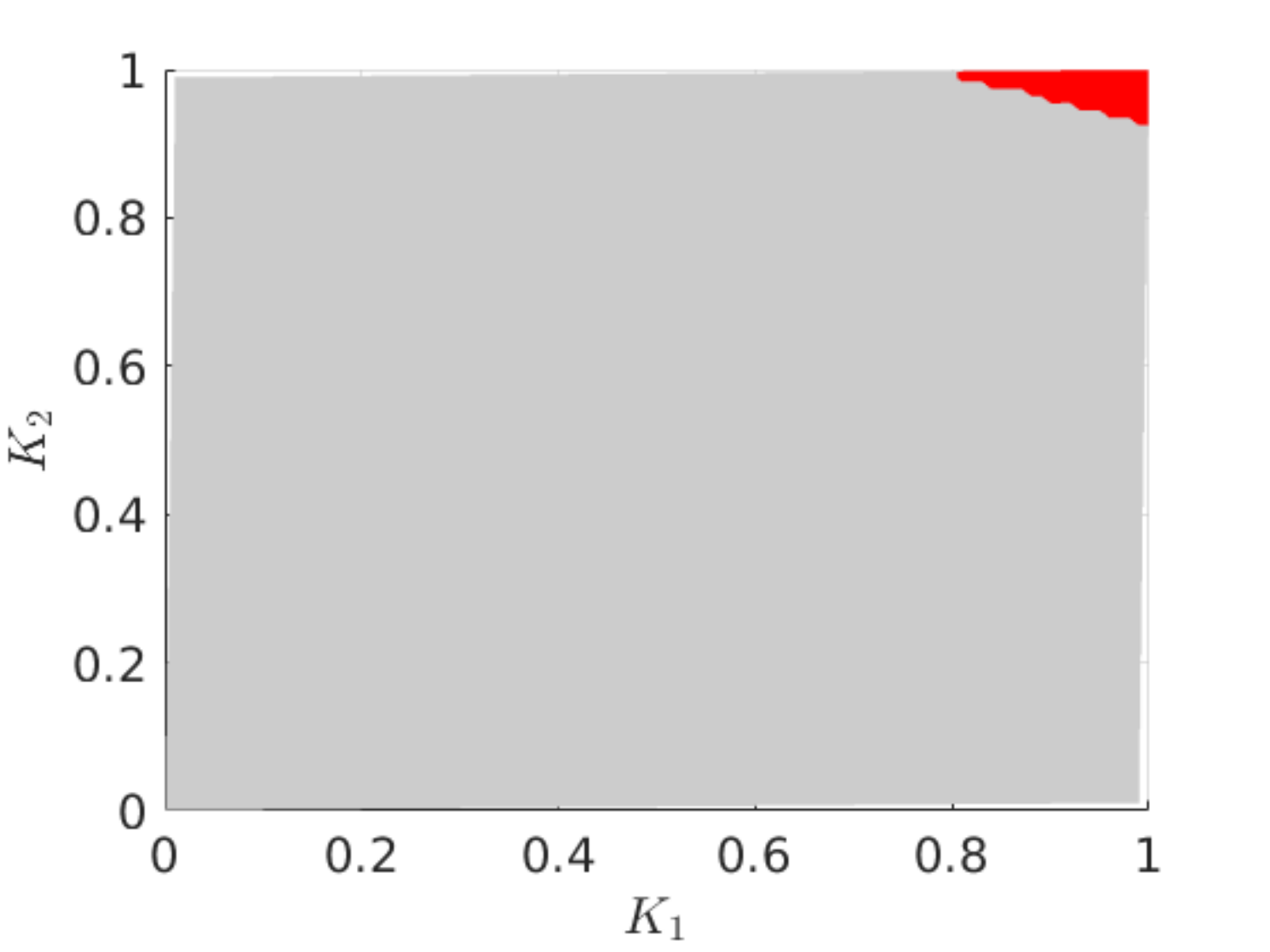}  
\subcaption{Classification provided by $\hat{\mathcal{M}}_{LLE,MoB} $ with 65 samples}\label{fig:KSETA06SmallMeta}
\end{subfigure}
\caption[Results and data for 2D LLE case $\mathcal{P}_{2}$]{Results and data second 2D problem case $\mathcal{P}_{2}$. $K_{1}$ is given in $\text{N}/\text{m}^{3}$, $K_{2}$ in $\text{N}/\text{m}$. LLE is unitless.}\label{fig:KSETA06Small}
\end{figure}

\begin{figure}[b!]
\centering
\begin{subfigure}[t]{0.5\textwidth}
\includegraphics[scale=0.4]{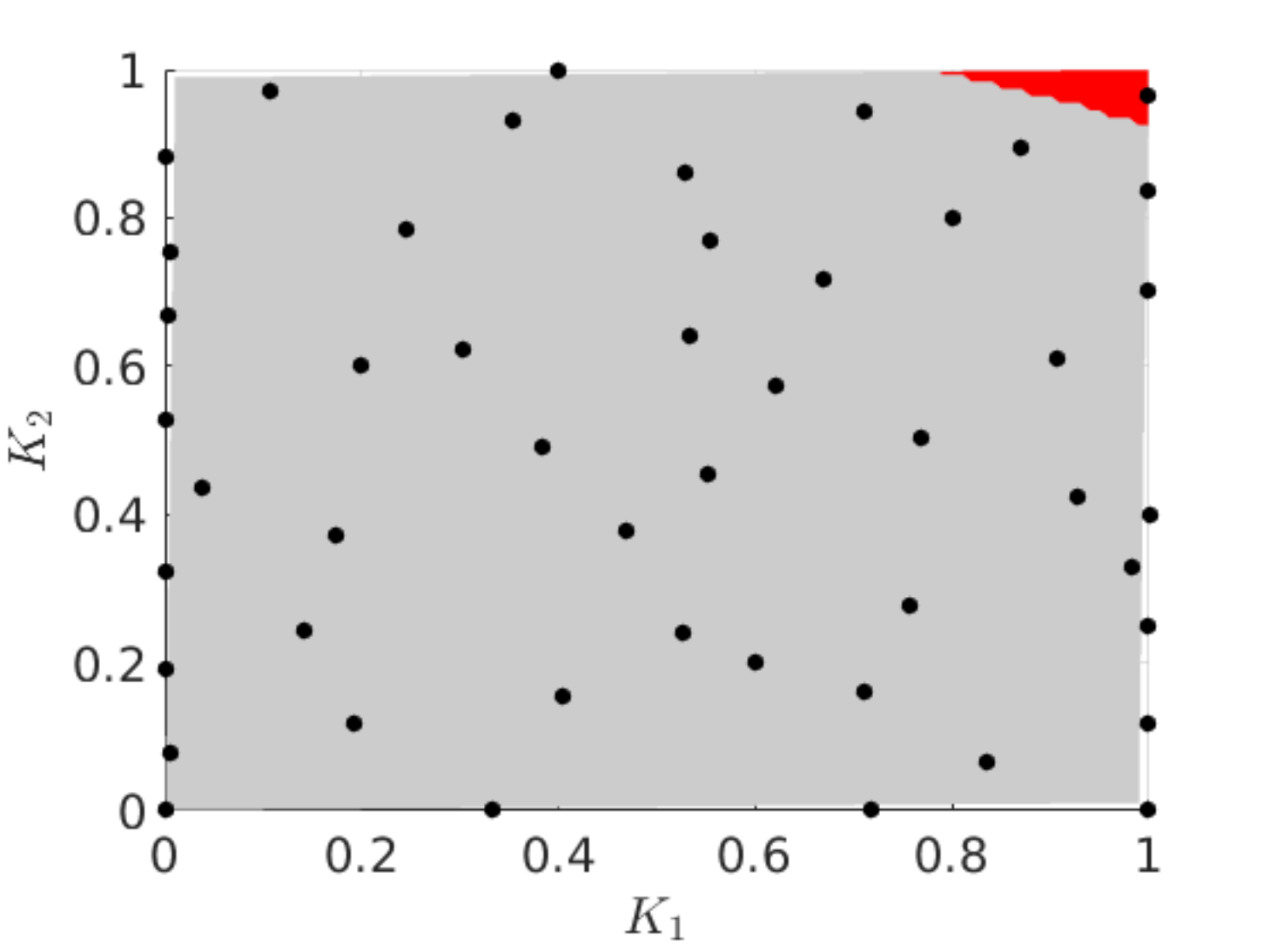}  
\subcaption{AME}\label{fig:KSEtaSmallSamplesCompareAME}
\end{subfigure}%
\begin{subfigure}[t]{0.5\textwidth}
\includegraphics[scale=0.4]{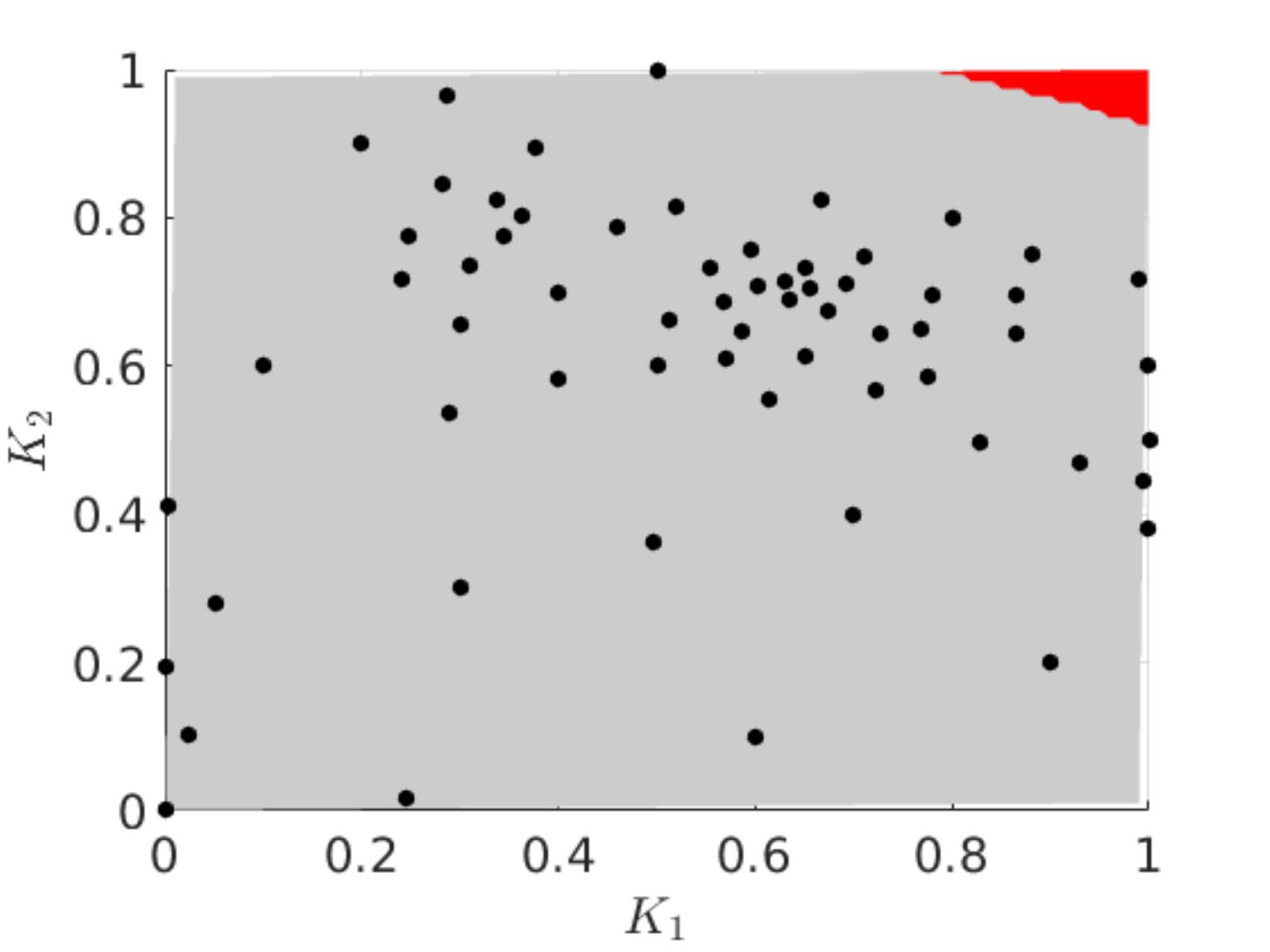} 
\subcaption{EIGF}\label{fig:KSEtaSmallSamplesCompareEIGF}
\end{subfigure}
\begin{subfigure}[t]{0.5\textwidth}
\includegraphics[scale=0.4]{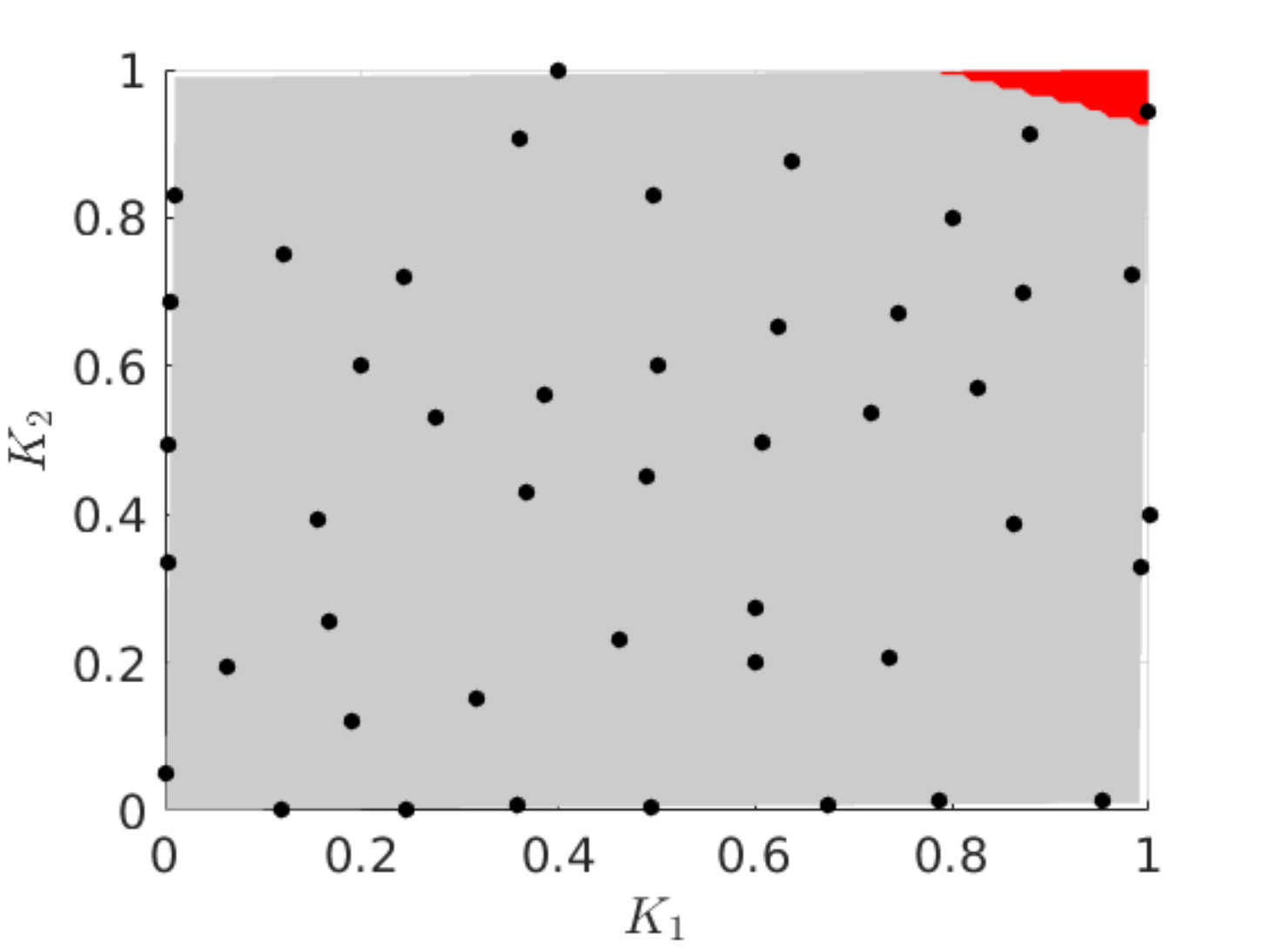}  
\subcaption{MIPT}\label{fig:KSEtaSmallSamplesCompareMIPT}
\end{subfigure}
\caption[Sample positions for AME,EIGF and MIPT for 2D LLE case $\mathcal{P}_{2}$ after 65 samples]{Sample positions for AME,EIGF and MIPT for 2D LLE case $\mathcal{P}_{2}$ after 65 samples. $K_{1}$ is given in $\text{N}/\text{m}^{3}$, $K_{2}$ in $\text{N}/\text{m}$. }\label{fig:KSEtaSmallSamplesCompare}
\end{figure}

\begin{figure}[h!]
\centering
\includegraphics[scale=0.4]{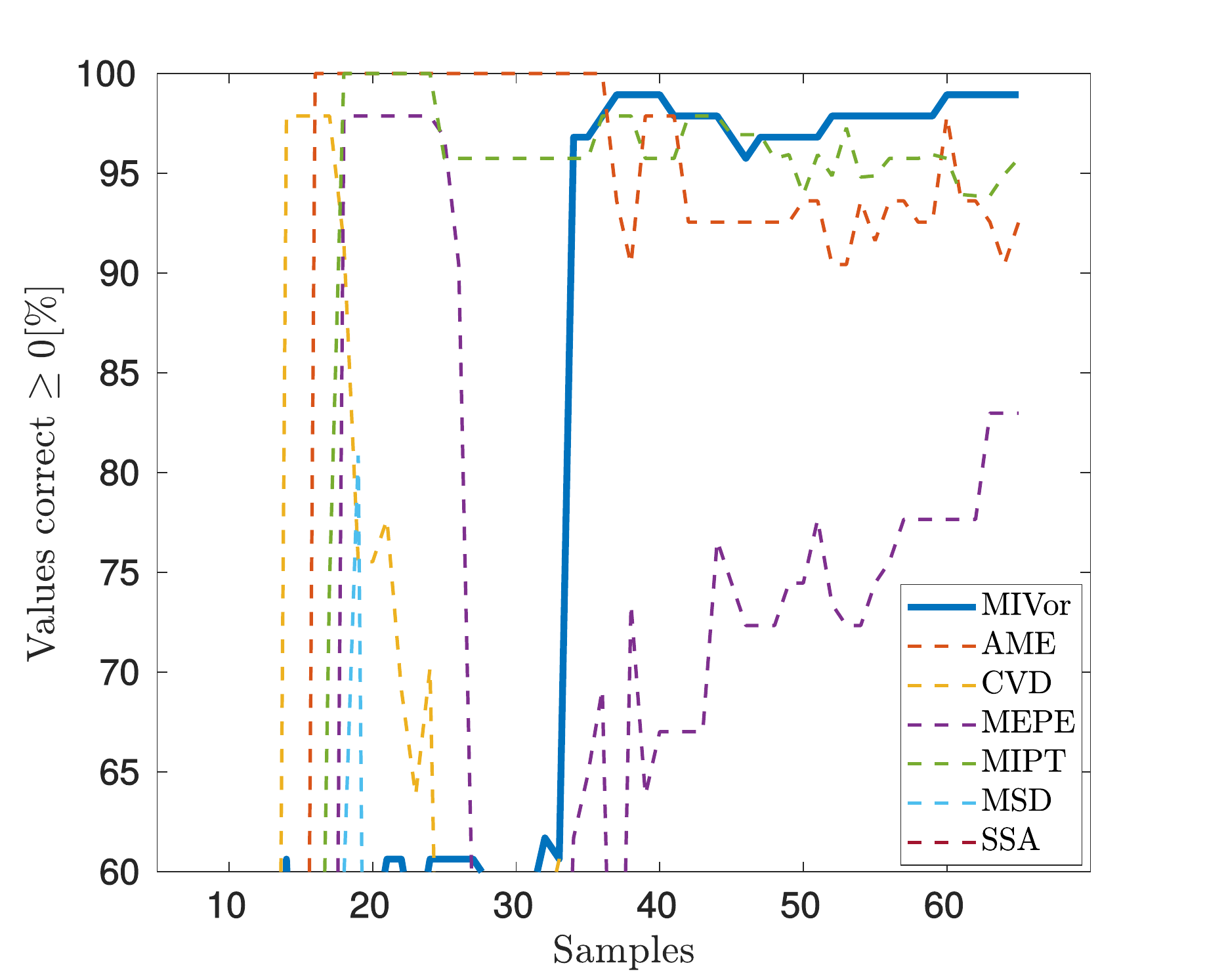}
\caption[Convergence of correctly classified points with $LLE \geq 0$ for 2D LLE case $\mathcal{P}_{2}$]{Convergence of correctly classified points with $LLE \geq 0$ for 2D LLE case $\mathcal{P}_{2}$.}\label{fig::KSeta06SmallConvergenceAbove}
\end{figure}

\begin{figure}[h!]
\centering
\includegraphics[scale=0.38]{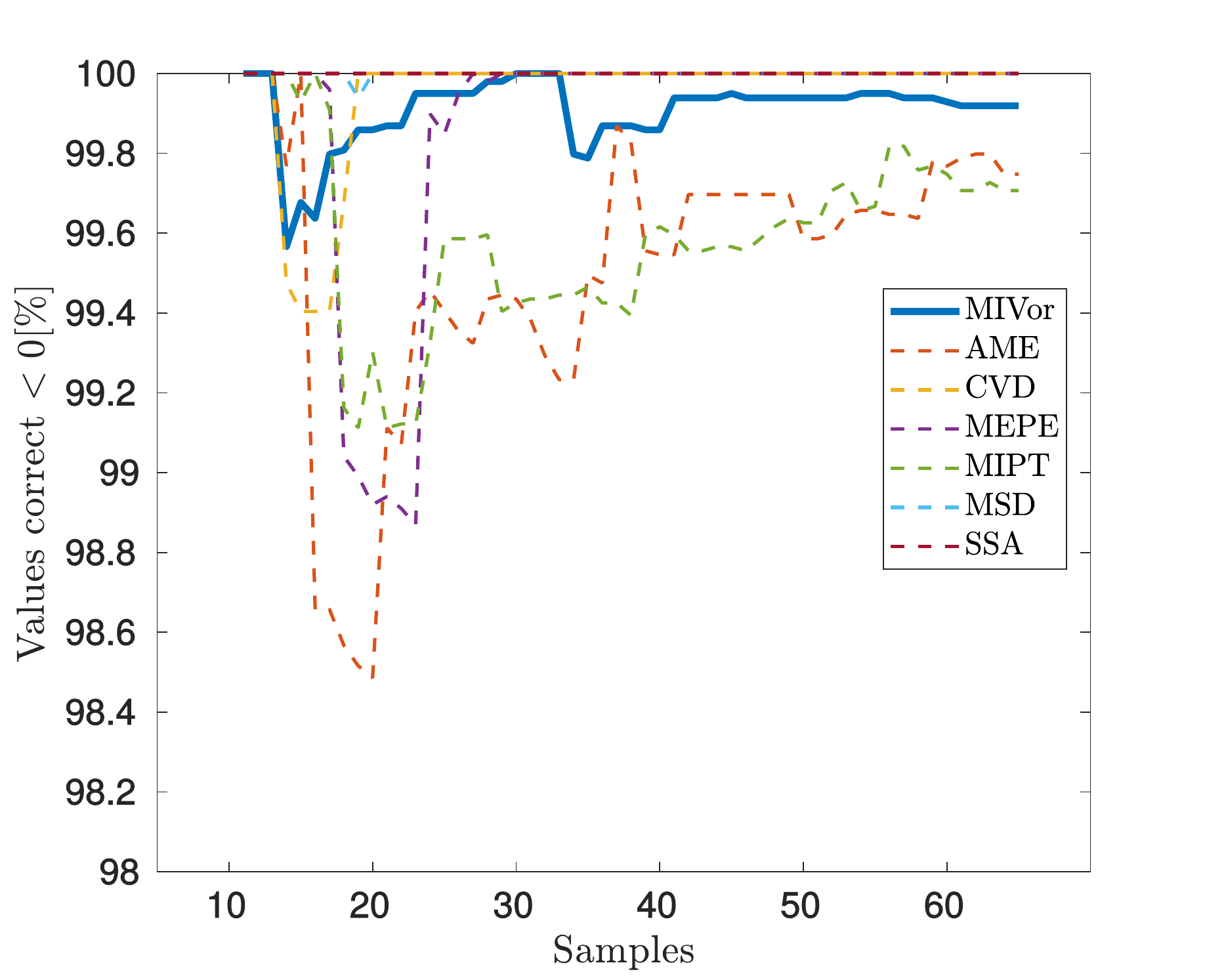}
\caption[Convergence of correctly classified points with $LLE < 0$ for 2D LLE case $\mathcal{P}_{2}$]{Convergence of correctly classified points with $LLE < 0$ for 2D LLE case $\mathcal{P}_{2}$.}\label{fig::KSeta06SmallConvergenceBelow}
\end{figure}

The percentage value of correctly classified point with $\mathcal{M}_{LLE,MoB} \geq 0$ over the sample size from 5 until 90 samples is plotted in Figure \ref{fig::KSEta06Above}. 
\clearpage
MIVor is highlighted with the thicker line. It can be noticed that after around 50 samples MIVor holds a constant value whereas the other methods fluctuate more. 
\begin{figure}[t!]
\centering
\includegraphics[scale=0.4]{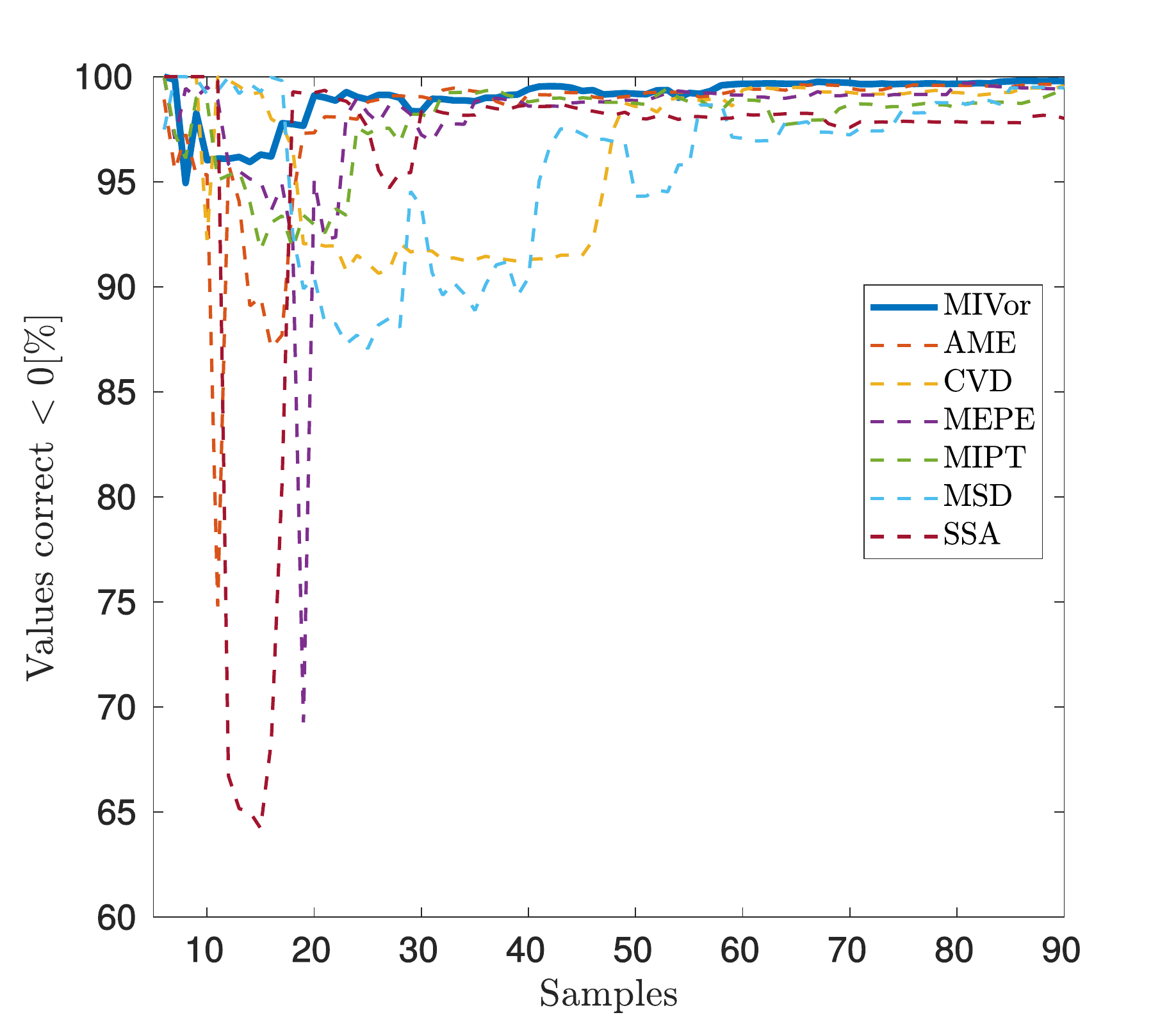}
\caption[Convergence of correctly classified points with $LLE < 0$ for 2D LLE case $\mathcal{P}_{1}$]{Convergence of correctly classified points with $LLE < 0$ for 2D LLE case $\mathcal{P}_{1}$.}\label{fig::KSEta06Below}
\end{figure}
The percentage of correctly identified points is shown in Figure \ref{fig::KSEta06Below}. MIVor yields the best results after 90 samples but has almost reached convergence performance after around 40 samples.
\clearpage
\textbf{Problem case $\mathcal{P}_{2}$} \\
\\
The proposed method is tested for a second problem test which aims at evaluating its ability of proficiently exploring the parametric domain if only a small percentage of the domain has a LLE value that indicated chaotic motion. 
Consider the input domain for the two spring-stiffnesses given by $K_{1} \in  [0.5 , 1.0]\, \text{N}/\text{m}^{3}$
and $K_{2} \in [0.0,0.5] \, \text{N}/\text{m}$ with 
$\Omega = 0.6 \, \text{rad}/\text{s}$ and
$\mu_{k} = 0.15$. Consider 10 initial sample points created with TPLHD and adaptive sampling until 65 samples.\\
The plot of $\mathcal{M}_{LLE,MoB}$, provided by the reference model, over the input domain is shown in Figure \ref{fig:KSETA06SmallPlot}. 
The indicator values are plotted in Figure \ref{fig:KSETA06SmallRedGray}.  It can be seen that only a fraction of the domain yields chaotic motion, only 112 points of the 12000 points that were spread in the domain in a space-filling manner, i.e. around $0.93 \%$. The point locations of the 55 samples created with MIVor over the normalized input space are highlighted in Figure \ref{fig:KSETA06SmallSamples}. \newpage It can be observed that they are predominantly spread in and around the red area that indicated chaos. This leads to an accurate surrogate model $\hat{\mathcal{M}}_{LLE,MoB} $ for the classification problem as shown in Figure \ref{fig:KSETA06SmallMeta}.
The sample positions of MIVor can be compared AME, EIGF and MIPT as displayed in Figure \ref{fig:KSEtaSmallSamplesCompare}. \\AME is a balance between exploration and exploitation and samples only 1 sample inside the red area, As previously seen in Figure \ref{fig:KSEtaSmallSamplesCompareEIGF}, EIGF  focuses on the regions with high absolute error and fails to sample inside the relevant chaotic domain. It can therefore not be expected to yield a good classification results. Similarly to AME, MIPT (Figure \ref{fig:KSEtaSmallSamplesCompareMIPT}) generates a single sample inside the red area. 
Even tough AME and MIPT only manage to sample one point each in the chaotic domain they represent the two best classifiers after MIVor when looking at the results given in Table \ref{table::KSEtaSmall}. 
It can be seen that MIVor fits around $99\%$ of points for both cases accurately. All other sampling techniques as well as the one-shot method perform worse.
The convergence of the percentage of $LLE \geq 0$ values for MIVor and 6 selected adaptive sampling techniques is depicted in Figure \ref{fig::KSeta06SmallConvergenceAbove}.\\ It can be seen that MIVor shows a big jump around 32 sample points. This is due to the fact that the algorithm needs to identify the small area of Figure \ref{fig:KSETA06SmallRedGray}. After the jump a constant behavior of MIVor can be noticed.
Similarly the convergence of the classification of points below 0 is plotted in Figure \ref{fig::KSeta06SmallConvergenceBelow}. As a large part of the domain is below this threshold value the metamodel prediction is accurate here for all considered methods with the values varying between $98-100\%$.
\begin{table}[h!]
\begin{center}
\resizebox{1.0\textwidth}{!}{%
\begin{tabular}{l|l c c c } \hline
& Method & MAE  & Equal or above 0[$\%$ ] & Below 0[$\%$ ] \\ \hline\hline \\
\multirow{1}{*}{\shortstack[l]{Values after\\10 samples}} & TPLHD  & 0.0498  & 0.0 & 100.0 \\ \\  \hline \\
\multirow{14}{*}{\shortstack[l]{ Values after\\65 samples}} &TPLHD &\textbf{0.0096}  & 0.0 & 100.0 \\
&ACE &  0.0346  & 0.0 & 100.0 \\
&AME &   0.0098  & 92.55 & 99.65 \\
&CVD   &0.0113  & 0.0 & 100.0 \\
&CVVor & 0.0255  & 61.70 & 99.98 \\ 
&EI &   0.0354 & 15.95 & 100.0  \\
&EIGF   & 0.0533  & 0.0 & 100.0  \\
&MASA    & 0.0380  & 0.0 & 100.0  \\
&MEPE    & 0.0116  & 82.97 & 100.0\\
&MIPT    & 0.0117   & 95.74	& 99.70 \\ 
&MSD   & 0.0178  & 0.0 & 100   \\
&SFCVT & 0.0125  & 78.72 & 99.92 \\ 
&SSA    & 0.0197  & 46.80 & 100.0 \\   \hline \\
&MIVor &0.0179  & \textbf{98.93} &	\textbf{99.91} 
\end{tabular}
}
\end{center}
\caption[Error measures for 2D LLE problem case $\mathcal{P}_{2}$]{Error measures for 2D LLE classification problem case $\mathcal{P}_{2}$ after 65 samples.}\label{table::KSEtaSmall}
\end{table}
\clearpage
\textbf{Problem case $\mathcal{P}_{3}$} \\
\\
The third problem aims at testing MIVor with regard to its ability to detect disconnected regions of the input domain associated with chaotic behavior. 
Consider the input domain for the spring-stiffness given by $K_{2} \in [0.0,0.5] \, \text{N}/\text{m}$ and the kinematic friction coefficient give in $\mu_{k} \in [0.08,0.18]$ with 
$\Omega = 0.6 \, \text{rad}/\text{s}$ and $K_{1} \in  1.0\, \text{N}/\text{m}^{3}$. 
Here 5 points sampled with TPLHD are considered initially. The adaptive sampling techniques are evaluated after adding 100 sample points and the results are average over 10 iterations. 
\begin{figure}[b!]
\centering
\begin{subfigure}[t]{0.5\textwidth}
\includegraphics[scale=0.4]{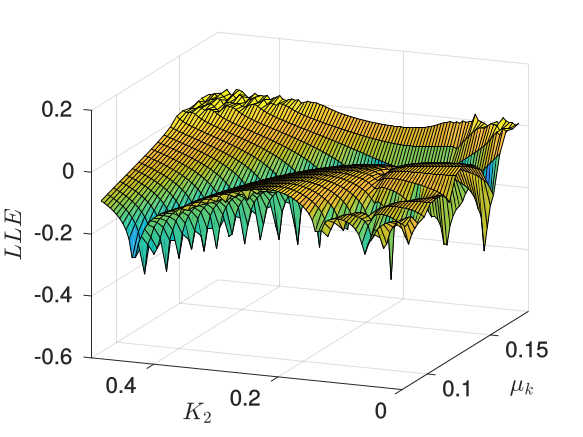} 
\subcaption{$\mathcal{M}_{LLE,MoB} $}\label{fig:muKK2Plot}
\end{subfigure}%
\begin{subfigure}[t]{0.5\textwidth}
\includegraphics[scale=0.4]{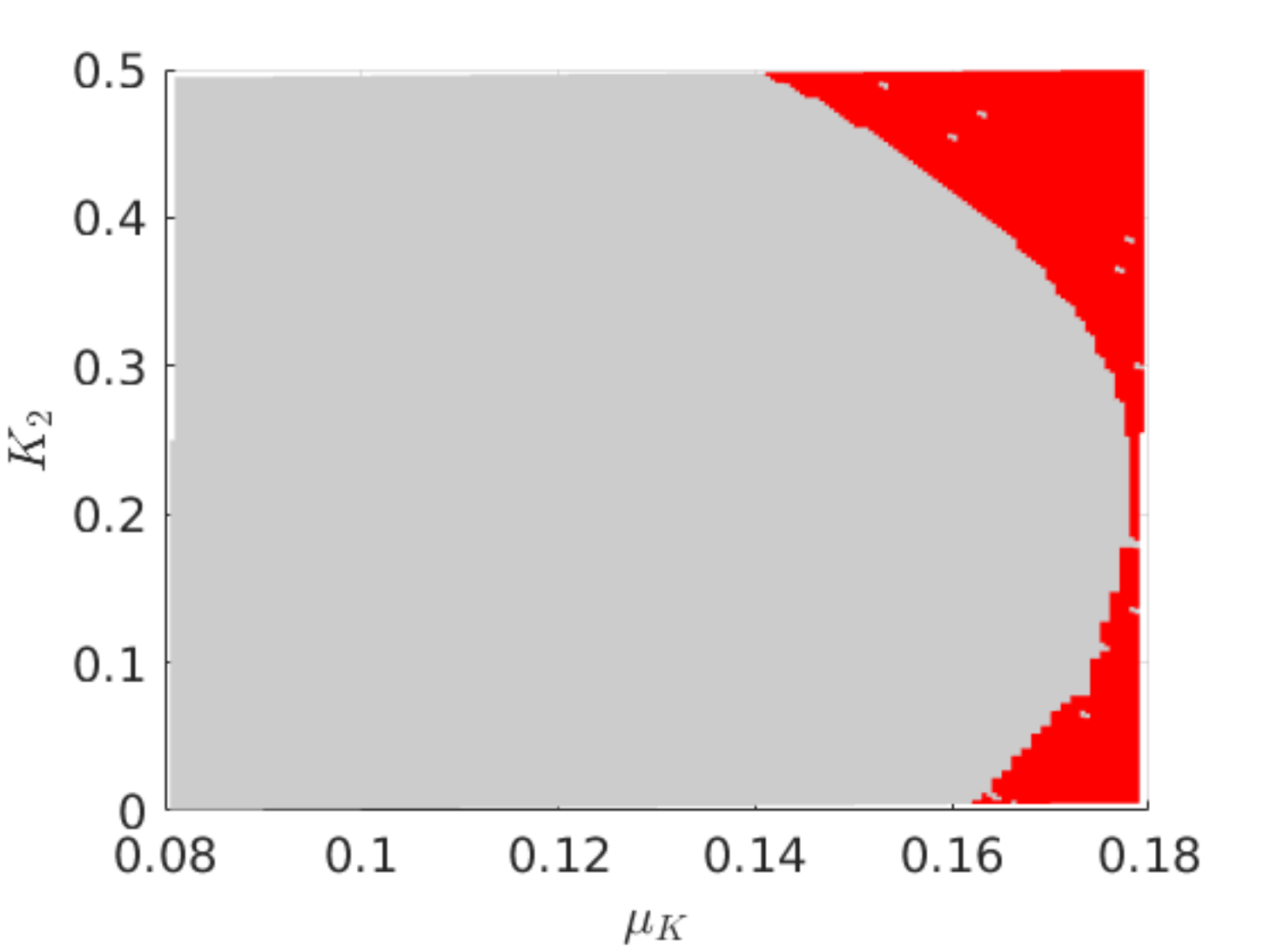} 
\subcaption{Classification provided by $\mathcal{M}_{LLE,MoB} $}\label{fig:muKK2RedGray}
\end{subfigure}
\begin{subfigure}[t]{0.5\textwidth}
\includegraphics[scale=0.45]{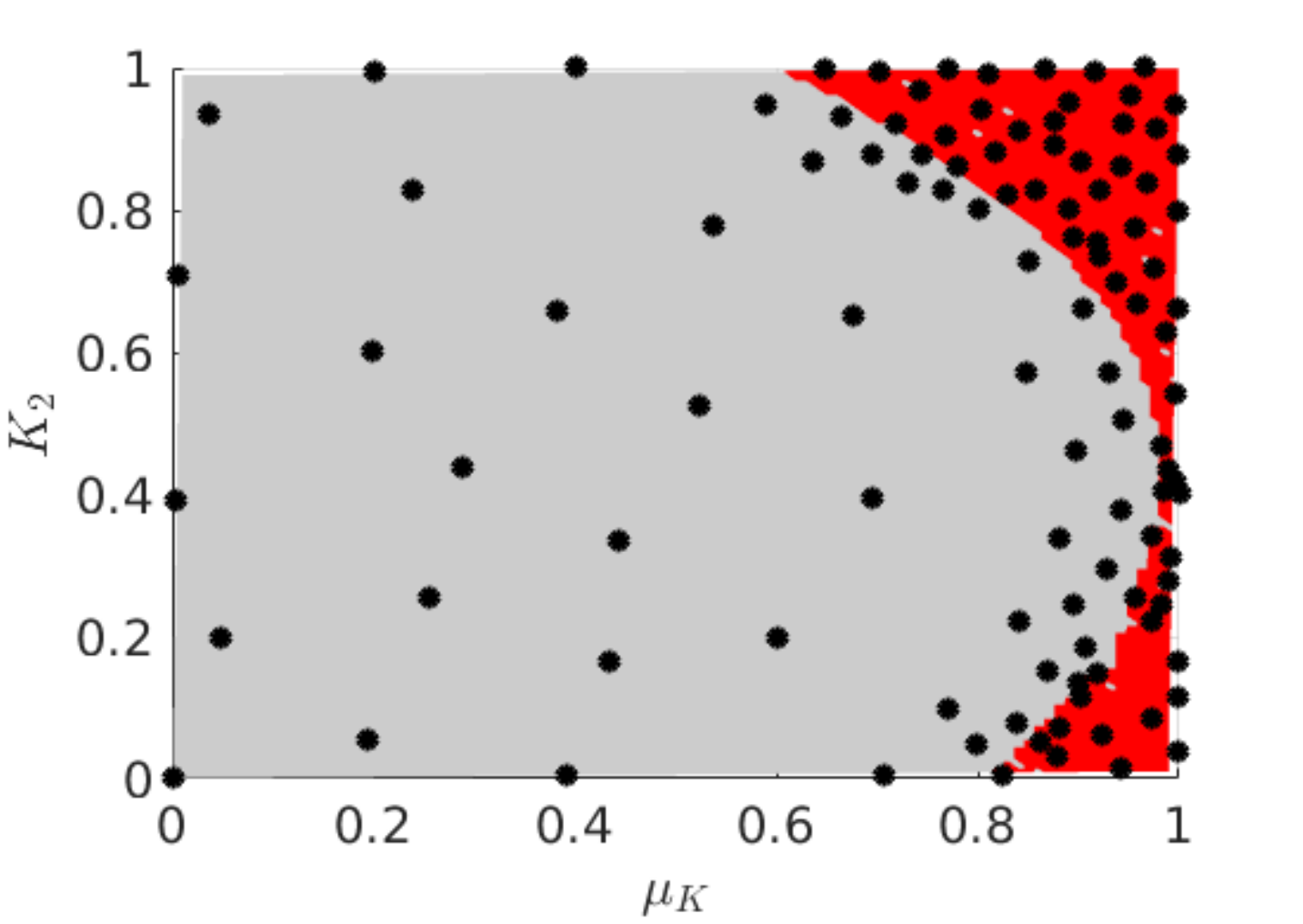} 
\subcaption{105 MIVor samples}\label{fig:muKK2Samples}
\end{subfigure}%
\begin{subfigure}[t]{0.5\textwidth}
\includegraphics[scale=0.4]{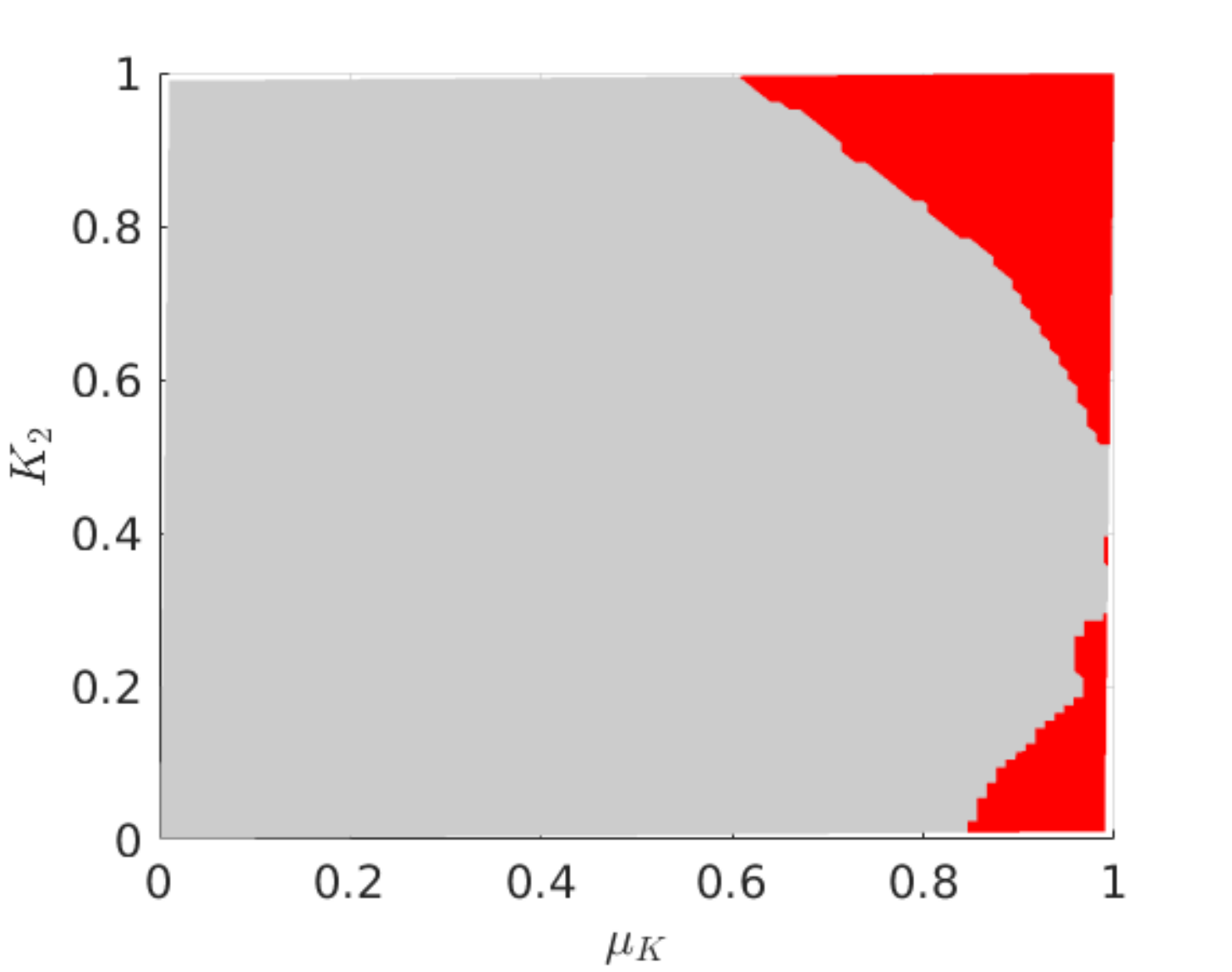} 
\subcaption{Classification provided by $\hat{\mathcal{M}}_{LLE,MoB} $ with 105 samples.}\label{fig:muKK2Meta}
\end{subfigure}
\caption[Results and data for 2D LLE case $\mathcal{P}_{3}$]{Results and data for 2D problem case $\mathcal{P}_{3}$. }\label{fig:muKK2}
\end{figure}
The LLE plot is given in Figure \ref{fig:muKK2Plot}, which yields the classification displayed in Figure \ref{fig:muKK2RedGray}. It can be seen that there are two major areas indicate chaotic motion. These two areas are almost disconnected. 
The locations of the sample points of MIVor after 105 samples are shown in Figure \ref{fig:muKK2Samples}. It can again be noticed that the method is effective in sampling around the red zones indicating chaos. Furthermore the rest of the domain is space-fillingly sampled with MIPT. 
\newpage
The resulting metamodel after 105 samples is displayed in Figure \ref{fig:muKK2Meta} and shows an accurate prediction performance.
The best performing sampling techniques of the previous two examples are compared to the results of MIVor in Table \ref{table::MUK2}. It can be seen that MIVor performs a lot better than all the other selected methods with both percentages being above $98 \%$. In this case even the techniques, e.g. AME and MIPT, that yielded accurate results in the previous problem achieve less proficient results here. 
\begin{table}[htpb]
\begin{center}
\resizebox{1.0\textwidth}{!}{%
\begin{tabular}{l|l c c c } \hline
& Method & MAE  & Above 0[$\%$ ] & Below 0[$\%$ ] \\ \hline\hline \\
\multirow{1}{*}{\shortstack[l]{Values after\\5 samples}} & TPLHD & 0.0529  & 18.75 & 99.94\\ \\  \hline \\
\multirow{8}{*}{\shortstack[l]{ Values after\\105 samples}} &TPLHD &0.0132  & 66.49 & \textbf{99.81} \\
&AME &   0.0159 & 85.59 & 99.34 \\
&CVD & 0.0142  & 75.54 & 99.80 \\ 
&MEPE    &0.0143  & 76.17 & 99.71\\
&MIPT    & 0.0146 & 85.68 & 99.55\\ 
&MSD   & 0.0191  & 83.87 & 99.69   \\
&SSA   & 0.0170  &  75.45 & 99.75 \\   \hline \\
&MIVor &0.0263 & \textbf{98.00} & 99.19
\end{tabular}
}
\end{center}
\caption[Error measures for 2D LLE problem case $\mathcal{P}_{3}$]{Error measures for 2D LLE classification problem case $\mathcal{P}_{3}$ after 105 samples.}\label{table::MUK2}
\end{table}

\begin{figure}[h!]
\centering
\includegraphics[scale=0.4]{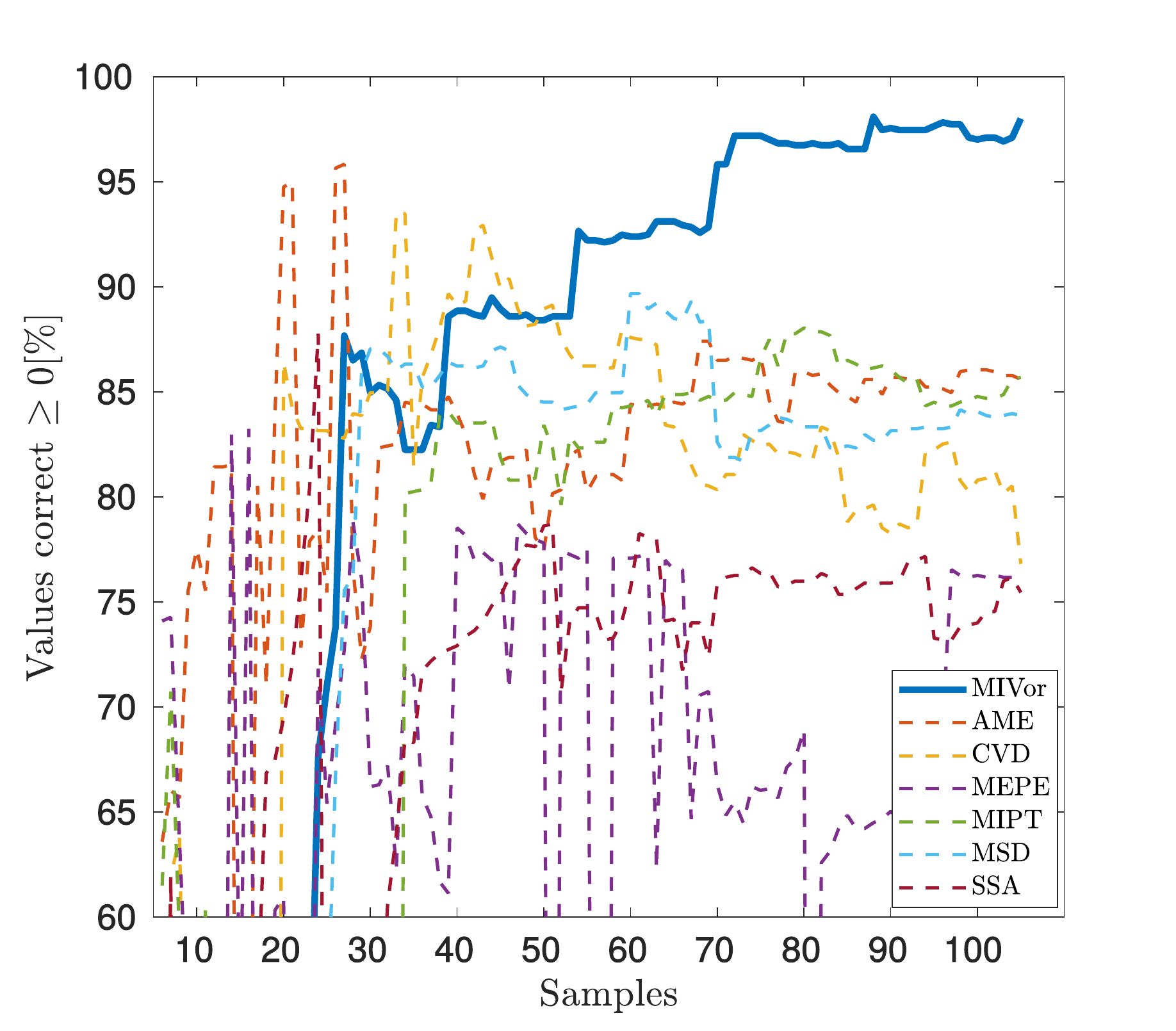}
\caption[Convergence of correctly classified points with $LLE \geq 0$ for 2D LLE problem $\mathcal{P}_{3}$]{Convergence of correctly classified points with $LLE \geq 0$ fo 2D LLE problem $\mathcal{P}_{3}$. $K_{1}$ is given in $\text{N}/\text{m}^{3}$, $K_{2}$ in $\text{N}/\text{m}$. LLE is untiless.}\label{fig::muK2above}
\end{figure}

\begin{figure}[hbtp]
\centering
\includegraphics[scale=0.4]{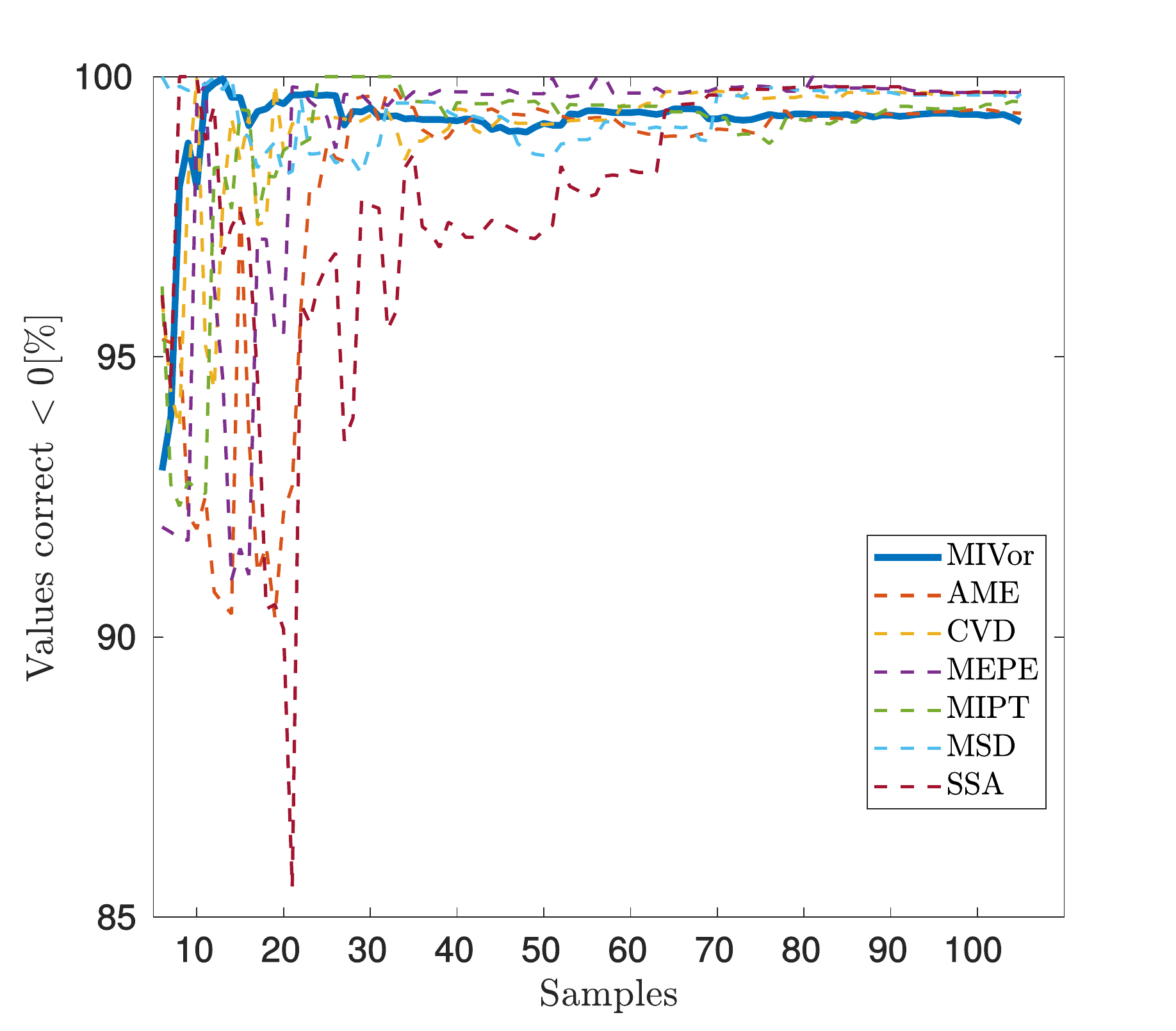}
\caption[Convergence of correctly classified points with $LLE < 0$ for 2D LLE problem case $\mathcal{P}_{3}$]{Convergence of correctly classified points with $LLE < 0$ for 2D LLE problem $\mathcal{P}_{3}$. $K_{1}$ is given in $\text{N}/\text{m}^{3}$, $K_{2}$ in $\text{N}/\text{m}$. LLE is untiless.}\label{fig::muK2Below}
\end{figure}

\begin{figure}[b!]
\centering
\begin{subfigure}[t]{0.5\textwidth}
\includegraphics[scale=0.4]{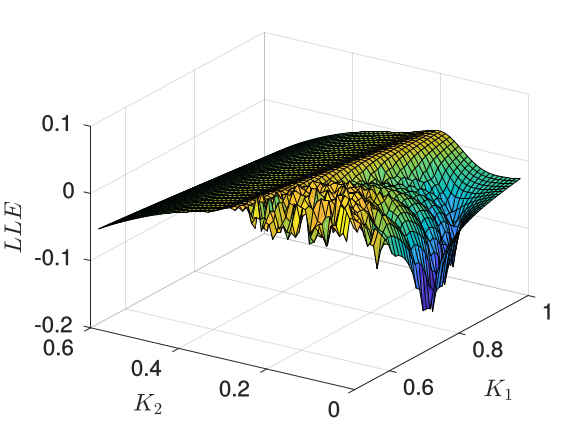} 
\subcaption{$\mathcal{M}_{LLE,MoB} $}\label{fig::Kseta07Plot}
\end{subfigure}%
\begin{subfigure}[t]{0.5\textwidth}
\includegraphics[scale=0.4]{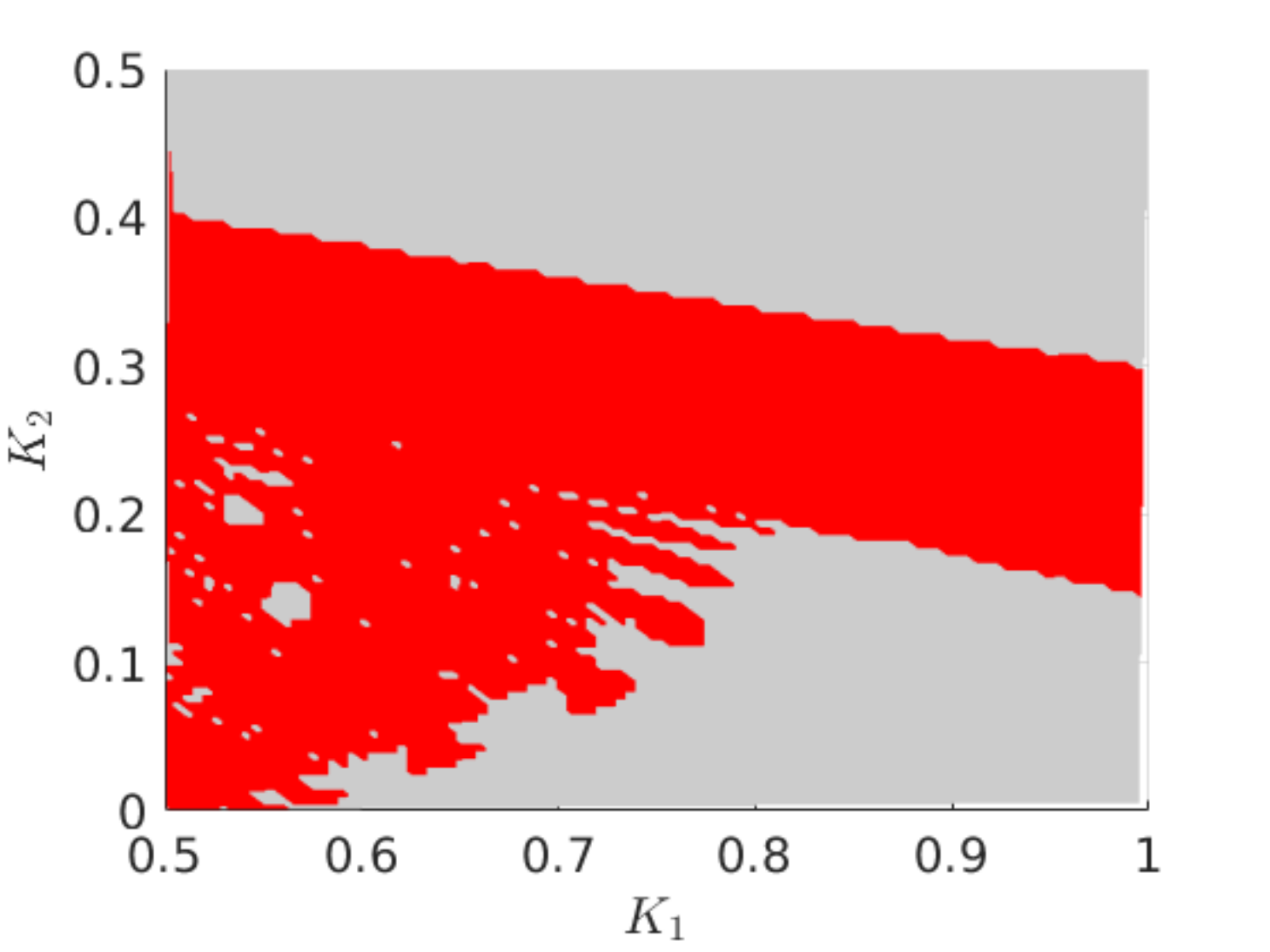} 
\subcaption{Classification provided by $\mathcal{M}_{LLE,MoB} $}\label{fig::Kseta07redGray}
\end{subfigure}
\begin{subfigure}[t]{0.5\textwidth}
\includegraphics[scale=0.4]{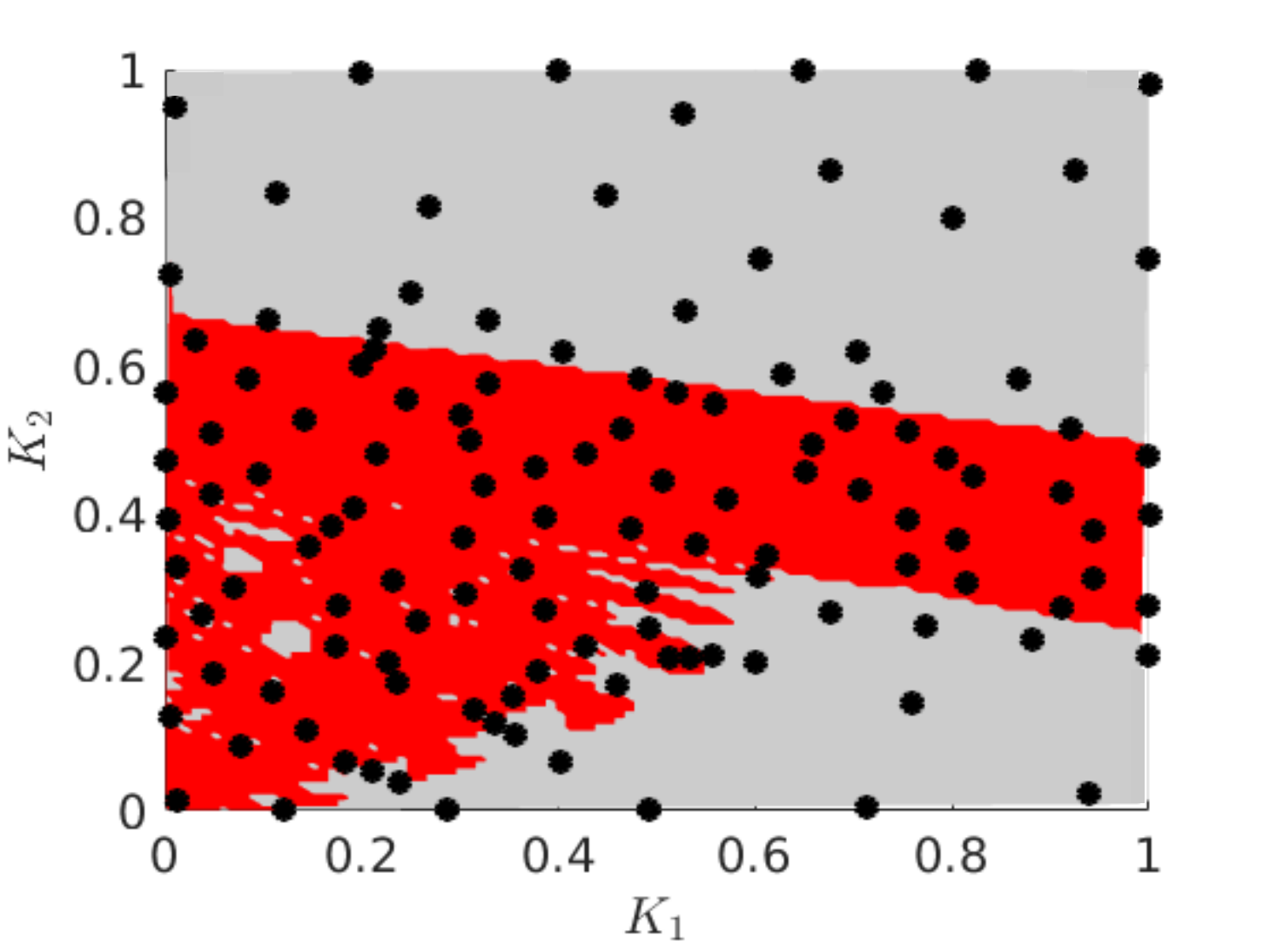} 
\subcaption{115 MIVor samples}\label{fig::Kseta07Samples}
\end{subfigure}%
\begin{subfigure}[t]{0.5\textwidth}
\includegraphics[scale=0.4]{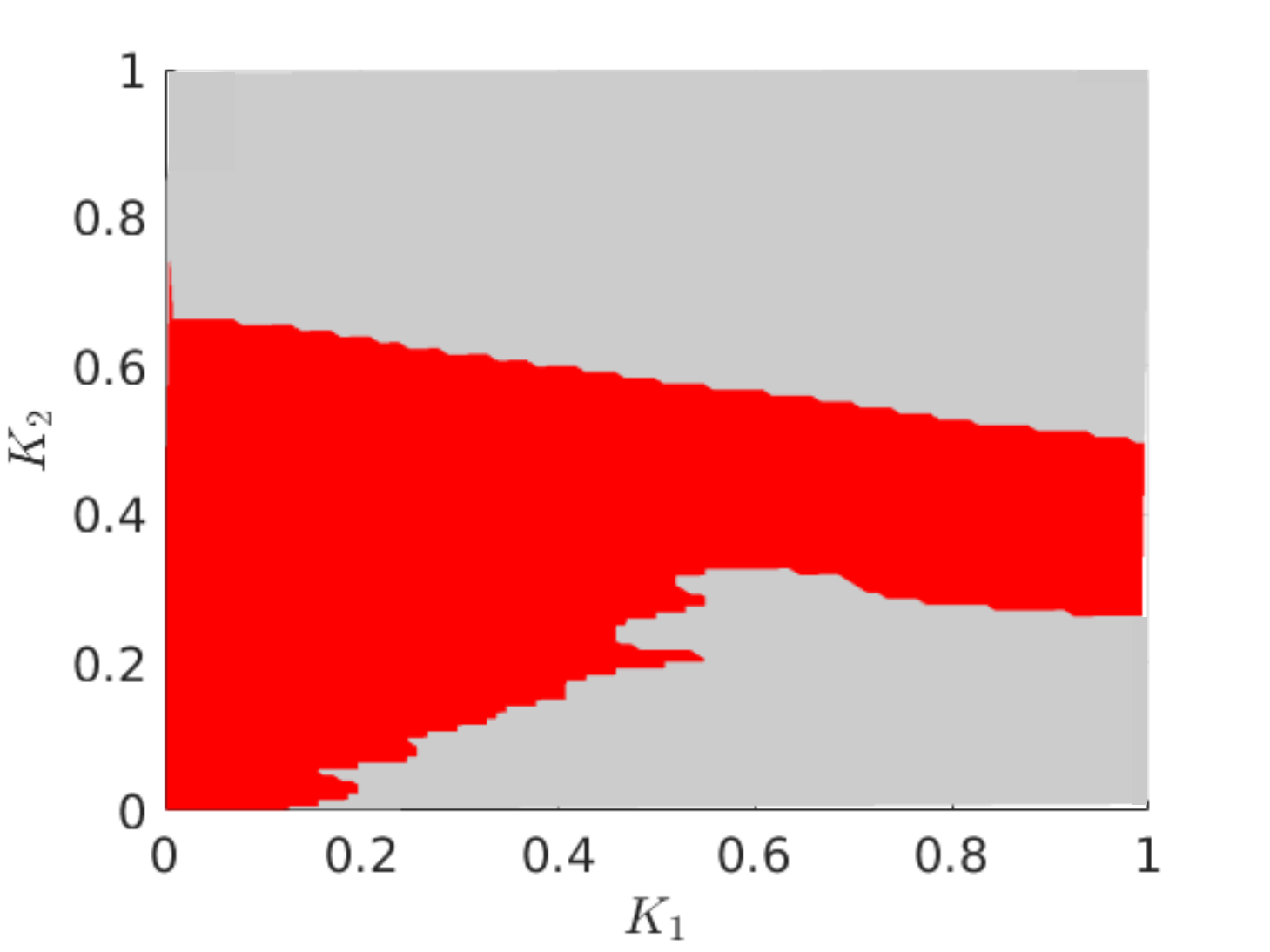} 
\subcaption{Classification provided by $\hat{\mathcal{M}}_{LLE,MoB} $ with 115 samples}\label{fig::Kseta07Meta}
\end{subfigure}
\caption[Results and data for 2D LLE case $\mathcal{P}_{4}$]{Results and data for the 2D LLE case $\mathcal{P}_{4}$. $K_{1}$ is given in $\text{N}/\text{m}^{3}$, $K_{2}$ in $\text{N}/\text{m}$. LLE is untiless.}\label{fig::Kseta07}
\end{figure}

\begin{figure}[htbp]
\centering
\includegraphics[scale=0.4]{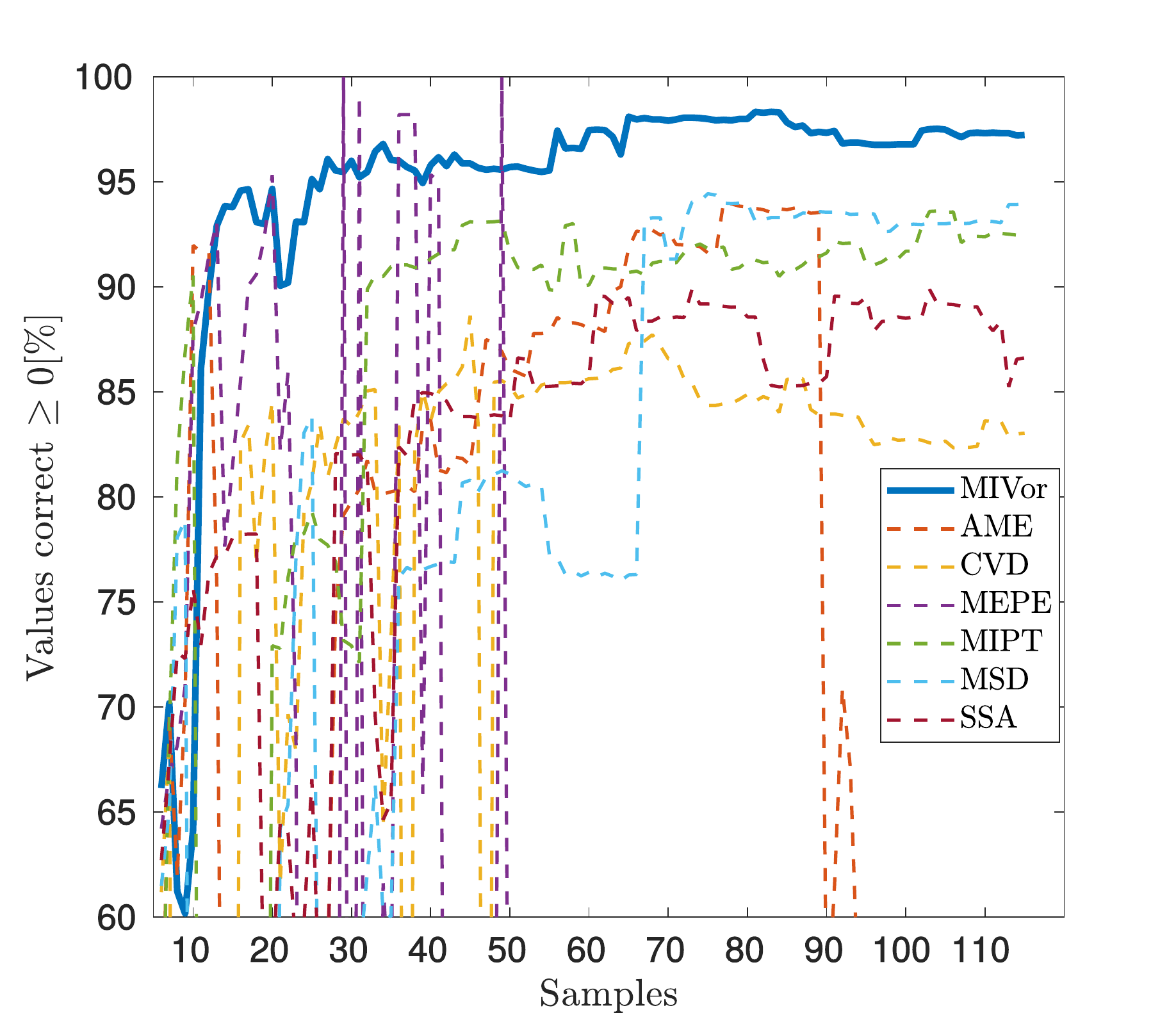}
\caption[Convergence of correctly classified points with $LLE \geq  0$ for 2D LLE case $\mathcal{P}_{4}$]{Convergence of correctly classified points with $LLE \geq  0$ for 2D LLE case $\mathcal{P}_{4}$.}\label{fig::KSeta07Above}
\end{figure}

\begin{figure}[htbp]
\centering
\includegraphics[scale=0.4]{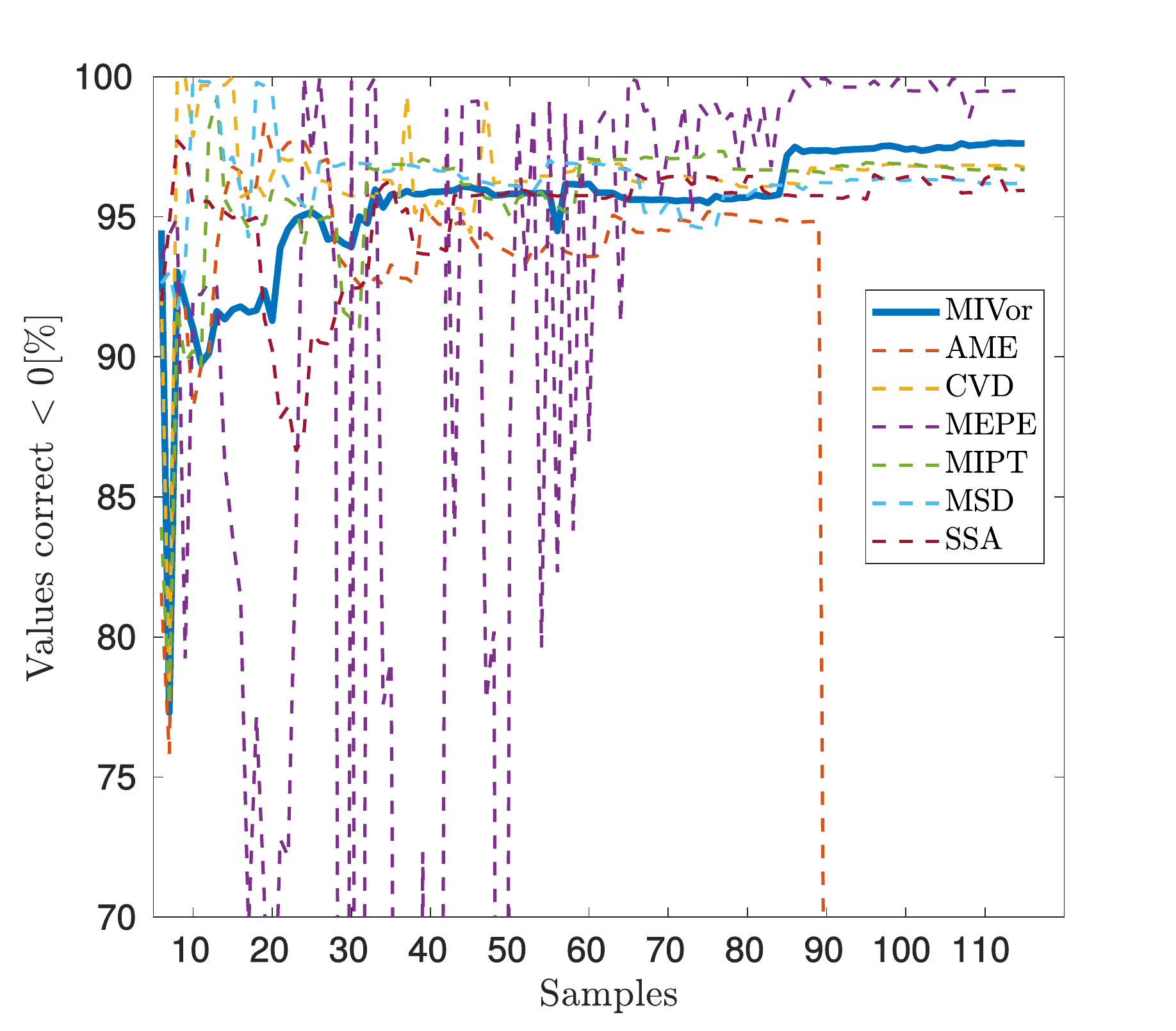}
\caption[Convergence of correctly classified points with $LLE <  0$ for 2D LLE case $\mathcal{P}_{4}$]{Convergence of correctly classified points with $LLE < 0$ for 2D LLE case $\mathcal{P}_{4}$.}\label{fig::KSeta07Below}
\end{figure}

The convergence of the percentage values is displayed in Figures \ref{fig::muK2above} and \ref{fig::muK2Below}. The performance of MIVor compared to the existing methods is clearly visible especially in Figure \ref{fig::muK2above}.
\clearpage
\textbf{Problem case $\mathcal{P}_{4}$} \\
\\
Consider the input domain for the two spring-stiffnesses given by\\$K_{1} \in  [0.5 , 1.0]\, \text{N}/\text{m}^{3}$
and $K_{2} \in [0.0,0.6] \, \text{N}/\text{m}$ with 
$\Omega = 0.7 \, \text{rad}/\text{s}$ and
$\mu_{k} = 0.15$.
Here 5 points are given initially sampled using the TPLHD technique. The adaptive sampling techniques are evaluated after reaching 115 sample points. The plot of the LLE's is shown in Figure \ref{fig::Kseta07Plot} the corresponding classification for this case is pictured in Figure \ref{fig::Kseta07redGray}. It can be noticed that the chaotic area is complex and not as smooth as the previously investigated problems. The created samples needed to generate the metamodel of Figure \ref{fig::Kseta07Meta} are shown in Figure \ref{fig::Kseta07Samples}. It can be seen that the holes inside the red domain are not detected by the surrogate model. This will reduce the percentage of correctly classified points. However an accurate prediction of these holes with Kriging would increase the number of needed samples significantly.
\newpage
Furthermore identifying a point inside the red domain as safe (with regular motion) might lead to a perceived reliability of the system's dynamics around the found area which is not the case. 
The interest of creating a surrogate model which is able to accurately reproduce these holes might thus be discussed.
The results of the MIVor metamodel after 115 samples and 6 selected adaptive sampling techniques are shown in Table \ref{table::Kseta07}. The selected techniques represent the  most proficient methods of the last problem cases. Here AME can not be used because of clustering problems. MIVor achieves the best result in comparison to the other selected methods with over $97\%$ of points classified accurately in both cases. Considering the form of the chaotic region of Figure \ref{fig::Kseta07redGray} (with holes and narrow inlets) the result must be rated highly. The space-filling methods MSD and MIPT are not able to achieve similar results.
\begin{table}[t!]
\begin{center}
\resizebox{1.0\textwidth}{!}{%
\begin{tabular}{l|l c c c } \hline
& Method & MAE  & Above 0[$\%$ ] & Below 0[$\%$ ] \\ \hline\hline \\
\multirow{1}{*}{\shortstack[l]{Values after\\5 samples}} & TPLHD & 0.0152  & 60.39 & 91.41 \\  \\  \hline \\
\multirow{8}{*}{\shortstack[l]{ Values after\\115 samples}} &TPLHD & 0.0052  & 93.53 & 96.39 \\
&AME &   -  & - & - \\
&CVD & \textbf{0.0016}  & 83.04 & 96.75\\ 
&MEPE    &0.0223  & 12.59 & 99.49\\
&MIPT    & 0.0056  & 92.46	& 96.68\\ 
&MSD   & 0.0057  & 93.95 & 96.20   \\
&SSA   & 0.0073  & 86.61 & 95.94  \\   \hline \\
&MIVor &0.0055  & \textbf{97.22} & \textbf{97.61}
\end{tabular}
}
\end{center}
\caption[Error measures for 2D LLE problem case $\mathcal{P}_{4}$]{Error measures for 2D LLE classification problem case $\mathcal{P}_{4}$ after 115 samples (methods with clustering problems are indicated by empty rows).}\label{table::Kseta07}
\end{table}
The convergence of the percentage values are displayed in Figures \ref{fig::KSeta07Above} and \ref{fig::KSeta07Below}. The validity of the MIVor with respect to the existing methods is especially noticeable in Figure \ref{fig::KSeta07Above} where the percentage value is more or less constant starting from around 55 sample points. Furthermore the average values of MIVor show less jitters in comparison with the other methods which hints at less variance over the 10 iterations.

\clearpage
\section{Conclusive evaluation of MIVor}
The four investigated problems show a range of difficulties in detecting chaotic and regular motion in the parametric space. The introduced MIVor adaptive sampling techniques appears to offer promising performances compared to established approaches. Furthermore its use is not a priori limited to this application. The technique could be applied to any problems which is defined by boolean output. 
Thus, an innovative adaptive sampling technique has been herein proposed for classification using Kriging. In future work this method should be investigated for higher dimensionality problems and other application fields.

\chapter{Multi-fidelity Kriging}\label{ch::MFKriging}
\begin{figure}[h!]
\centering
\begin{subfigure}[t]{0.5\textwidth}
\includegraphics[scale=0.5]{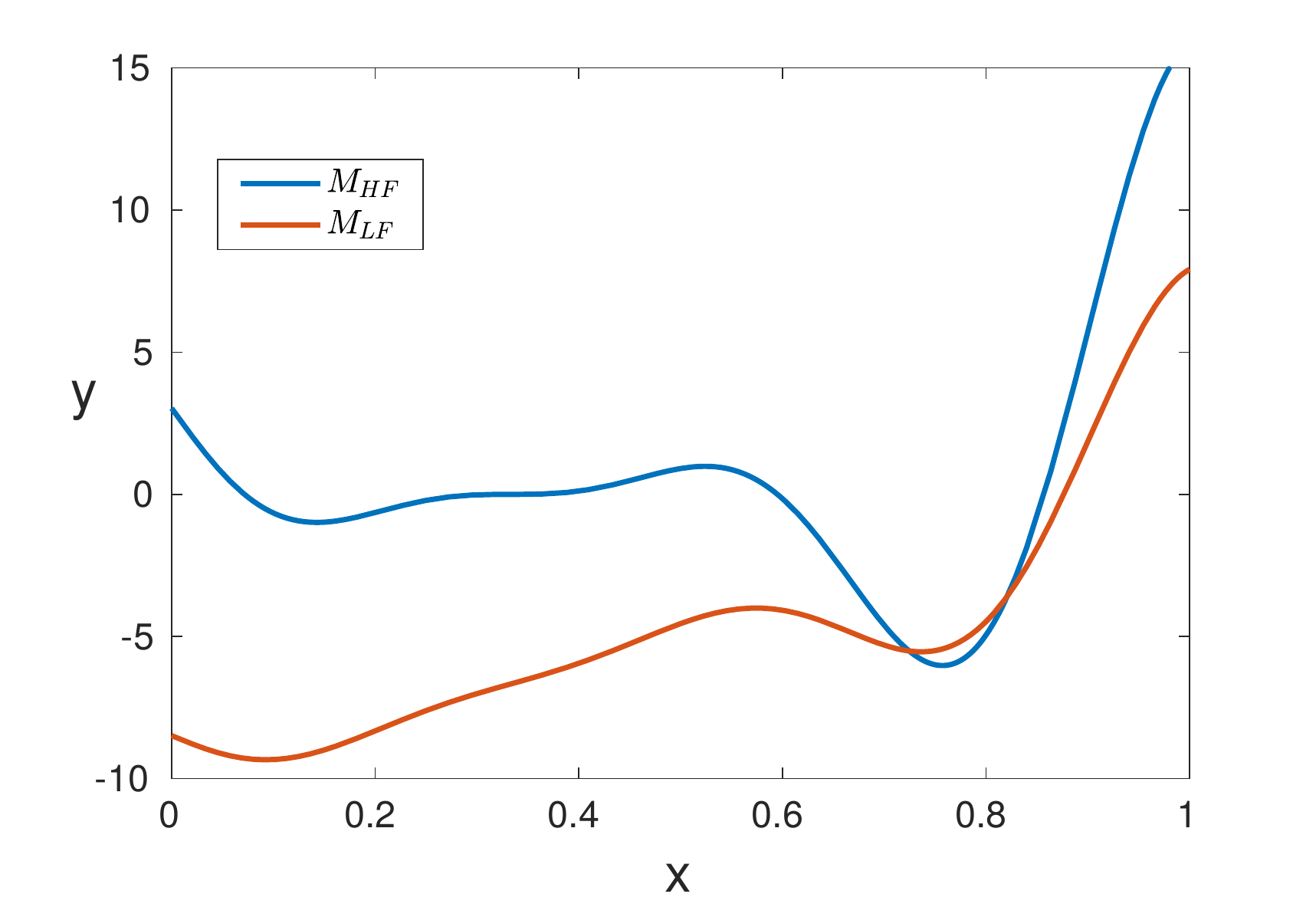} 
\subcaption{}\label{fig::HK_function}
\end{subfigure}
\begin{subfigure}[t]{0.5\textwidth}
\includegraphics[scale=0.5]{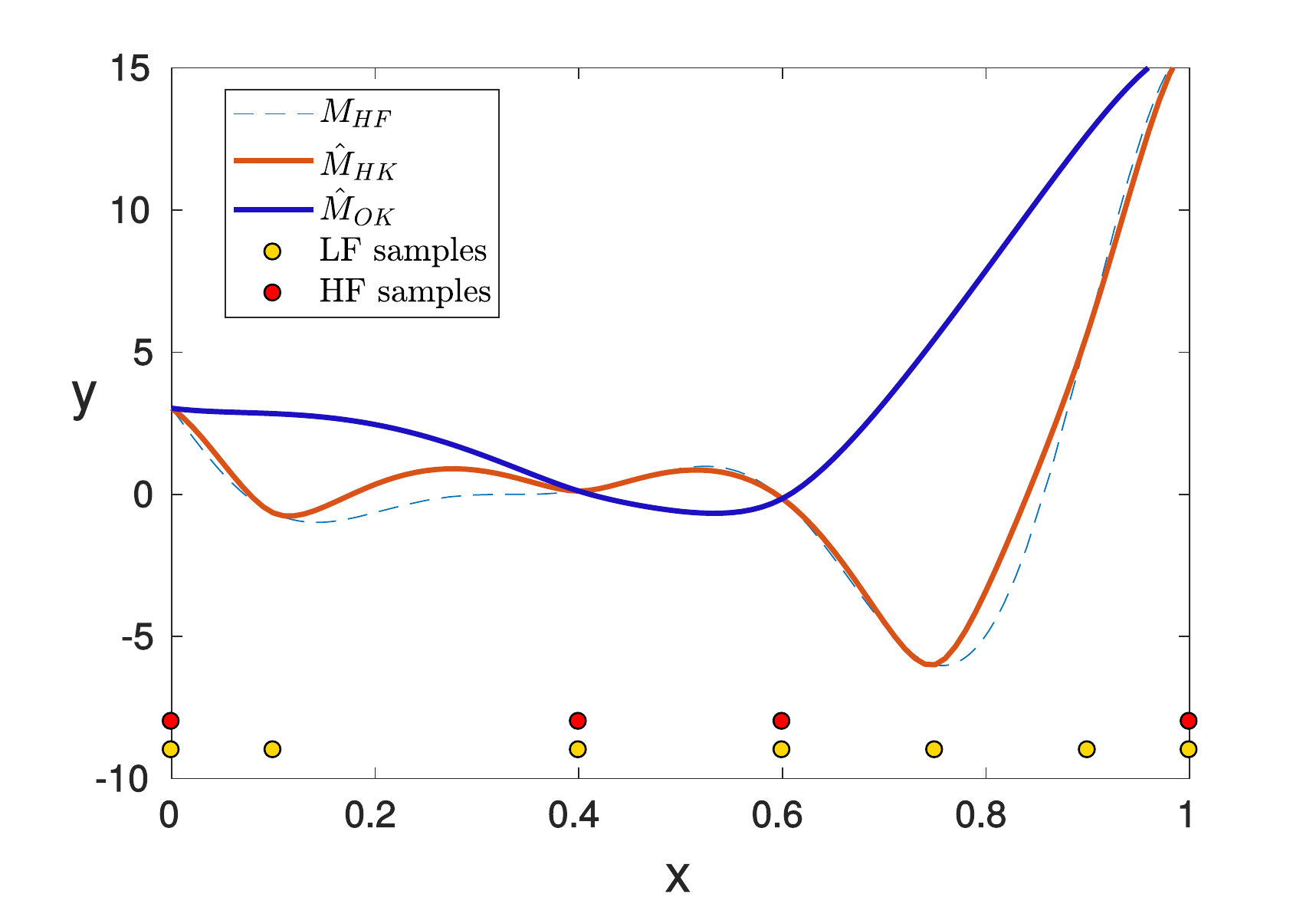} 
\subcaption{}\label{fig::HK_metamodel}
\end{subfigure}
\caption[Example of Hierarchical Kriging]{ Example of Hierarchical Kriging for $\mathcal{M}_{Test,HF}^{1d}$. (a) Example functions for Hierarchical Kriging, (b) Constructed metamodels with given sampling values. HK achieves better results compared to OK.  }\label{fig::HK_ex}
\end{figure}

Multi-fidelity modeling is used to further improve the efficiency of surrogate model construction, whereby numerical models of varying degrees of fidelity and computational expense are combined. Here, \acrfull{lf} models are employed in order to reduce the number of expensive mapping evaluations and hence yield process speedup. \acrfull{hf} models are leveraged to establish the necessary accuracy of the surrogate model. A good overview is given in \cite{peherstorfer2018survey}. \\
\cite{kennedy2000predicting} first suggested to apply this technique to the Kriging framework in context with computer experiments. The authors developed a method called Co-Kriging, in which the correlation matrix is built in a multistage process. First, a Kriging model is built for LF data. Then, a Kriging model is constructed for the difference in outputs between low- and high-fidelity model. Here, the tuning of a scaling factor is needed. Finally, the correlation matrix is built and the multi-fidelity Kriging process is fitted. However the implementation of the method was found to be complicated and less robust \citep{han2012hierarchical} and furthermore the estimation of the mean-squared error is not suitable for some adaptive sampling techniques. \\ 
In recent years multi-fidelity Kriging has been investigated and enhanced (see e.g. \cite{kuya2011multifidelity}, \cite{j2011efficient}, \cite{ghoreyshi2009accelerating}). Some considerations about the performance of multi-fidelity Kriging concerning for example the necessary correlation between LF and HF models are discussed in \cite{toal2015some}. \\
An interesting approach was introduced by \cite{han2012hierarchical}, called \acrfull{hk}, which directly takes the low-fidelity model as a trend for the construction of the high-fidelity surrogate. The respective estimation of the mean-squared error is more proficient and the correlation matrix has a lower dimensionality than in Co-Kriging. \\
Consider an expensive HF mapping
$\mathcal{M}_{HF} \, : \mathbb{X} \rightarrow \mathbb{Y}$ between an input $\bm{x} \in \mathbb{X} \subset \mathbb{R}^{n}$ and some output $\bm{y} \in \mathbb{Y} \subset \mathbb{R}$ from the 
given set of observations $\mathcal{D}_{HF} = \lbrace \left( \bm{x}^{(i)}_{HF}, \,\bm{y}^{(i)}_{HF} \right), \, i=1, \, \ldots  , \, m_{HF}  \rbrace$. The LF sampled data is denoted by $\mathcal{D}_{LF} = \lbrace \left( \bm{x}^{(i)}_{LF}, \,\bm{y}^{(i)}_{LF} \right), \, i=1, \, \ldots  , \, m_{LF}  \rbrace$ and is used to approximate the LF mapping $\mathcal{M}_{LF} \, : \mathbb{X} \rightarrow \mathbb{Y}$. Generating a metamodel for the HF mapping requires a LF surrogate model, which will assist in the prediction. \\
In HK the HF function can be written as
\begin{equation}
Y_{HF, i} (\bm{x}^{(i)}) = \mu_{HF} \, \mu_{\hat{Y}_{LF}}(\bm{x}^{(i)})  + Z(\bm{x}^{(i)}), \qquad i=1, \, \ldots, \, m
\end{equation}
where $\mu_{\hat{Y}_{LF}}$ is the LF model, which can be directly constructed by a Kriging model from the given LF sampled points. An unknown constant factor $\mu_{HF}$ yields the global trend of the HF model. As in the standard Kriging approach $Z(\bullet)$ is a stationary random process. \\
After a derivation which can be found in \cite{han2012hierarchical} the HK predictor can be written as
\begin{equation}
\mu_{\hat{Y}_{HF}} = \hat{\mu}_{HF} \mu_{\hat{Y}_{LF}}(\bm{x}^{(i)}) + \bm{r}_{0}^{T} \bm{R}^{-1} (\bm{y} - \hat{\mu}_{HF} \bm{F}),
\end{equation}
with 
\begin{equation}
\bm{F} = \left[ \mu_{\hat{Y}_{LF}}(\bm{x}^{(1)}), \,  \ldots, \,  \mu_{\hat{Y}_{LF}}(\bm{x}^{(m)}) \right]^{T}, \qquad i=1, \, \ldots, \, m_{HF}
\end{equation}
and $\hat{\mu}_{HF}$ reads
\begin{equation}
\hat{\mu}_{HF} = (\bm{F}^{T} \bm{R}^{-1} \bm{F})^{-1} \bm{F}^{-1} \bm{R}^{-1} \bm{y} \, .
\end{equation}
Finally, the mean-squared error of the HK prediction is given by
\begin{equation}
\begin{aligned}
\sigma^{2}_{\hat{Y}_{HF}} &= \sigma^{2} \lbrace 1 - \bm{r}_{0}^{T} \bm{R}^{-1} \bm{r}_{0} + \left[ \bm{r}_{0} \bm{R}^{-1} \bm{F} - \mu_{\hat{Y}_{LF}} \right] \\& \left( \bm{F}^{F} \bm{R}^{-1} \bm{F} \right)^{-1} \left[ \bm{r}_{0}^{T} \bm{R}^{-1} \bm{F} - \mu_{\hat{Y}_{LF}} \right]^{T} \rbrace \, .
\end{aligned}
\end{equation}
Similarly to standard Kriging the unknown hyperparameters can be obtained with MLE. As an example take the following high-fidelity map
\begin{equation}
\mathcal{M}_{Test,HF}^{1d}(x) = (6x - 2)^2 \,  \sin(12x-4)
\end{equation}
and the low-fidelity counterpart defined by
\begin{equation}
\mathcal{M}_{Test,LF}^{1d}(x) = 0.5  \mathcal{M}_{Test,HF}(x) + 10(x-0.5) - 5 \, \text{.}
\end{equation}
Consider the functions to be defined between $0$ and $1$ as plotted in Figure \ref{fig::HK_function}. 
For the construction of the metamodel four high-fidelity samples given at $x_{HF} = [0.0, 0.4, 0.6, 1.0]$ and six respective low-fidelity ones at\\$x_{LF} = [0.0,  0.1, 0.4,  0.6,  0.75,  0.9, 1.0]$ are used.
\newpage
 The metamodels of Hierarchical Kriging $\hat{\mathcal{M}}_{HK}$, and as a comparison Ordinary Kriging $\hat{\mathcal{M}}_{OK}$ approximated with high-fidelity samples, are depicted in Figure \ref{fig::HK_metamodel}.
With the help of the low-fidelity model Hierarchical Kriging is able to achieve a better result even though as given in Figure \ref{fig::HK_function} the low-fidelity function is a poor approximation of the high-fidelity one. An overview over adaptive sampling methods in given in the next section.

\clearpage
\section{Benchmark problems for Hierarchical Kriging}
In the following section the adaptive sampling algorithms presented in Chapter \ref{sec::adaptiveSampling} are tested and compared on benchmark problems with respect to their ability in the HK framework.
\subsection{One-dimensional Forrester function}\label{sec::Forrester}
\begin{figure}[htpb ]
\centering
\includegraphics[scale=0.5]{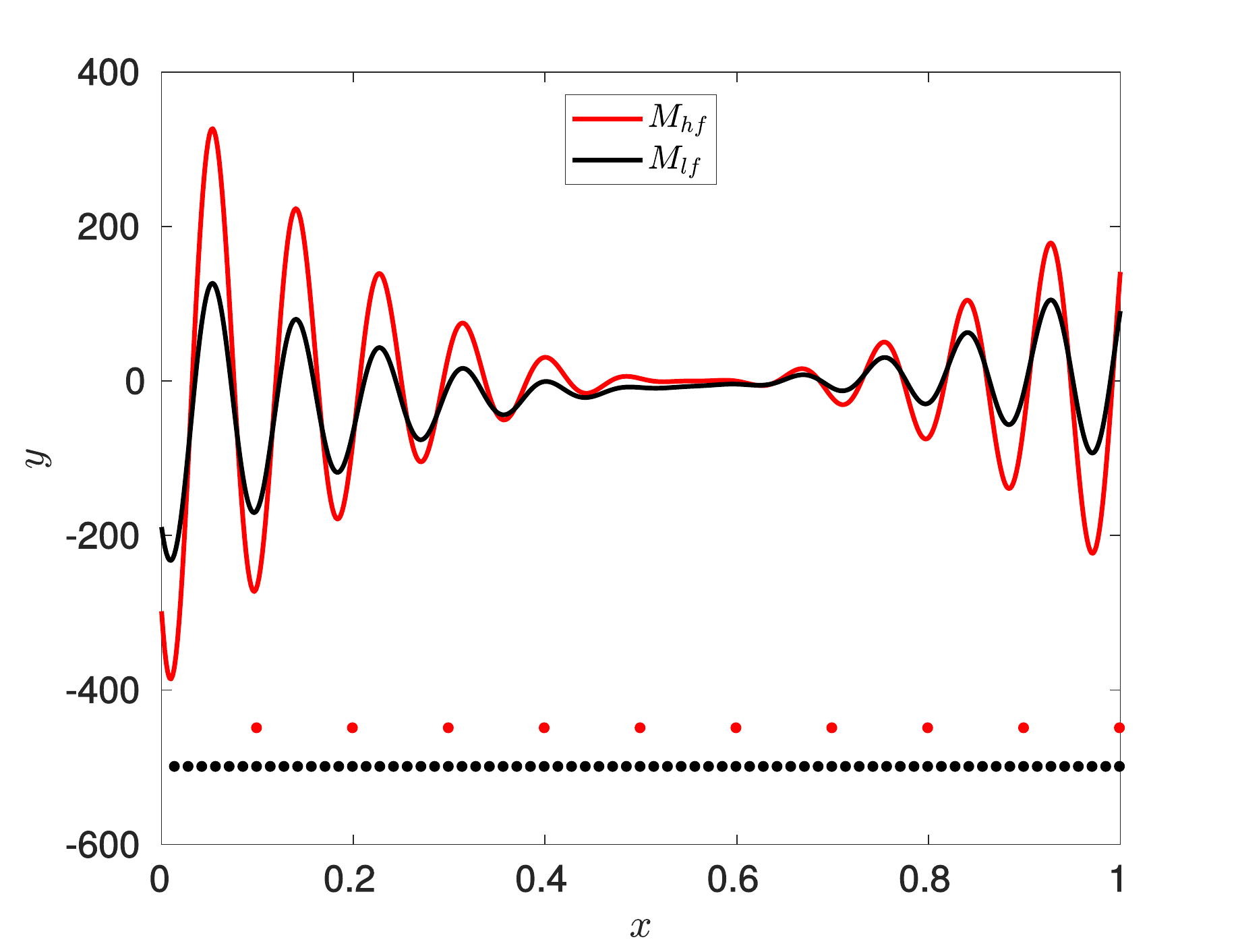} 
\caption[Initial state for metamodel generation of $\mathcal{M}_{Forrester,HF}^{1d}$]{10 HF-samples and 70 LF-samples for metamodel creation of $\mathcal{M}_{Forrester,HF}^{1d}$. Red line: HF-function. Black line: LF-function.}\label{fig:1d_Forrester}
\end{figure}

Consider the Forrester function as a multifidelity function with two levels as introduced in \cite{forrester2008c}. The high-fidelity function reads
\begin{equation}
\mathcal{M}_{Forrester,HF}^{1d}(x) =  (6 x -2)^{2} \sin(12 x - 4)
\end{equation}
and the low-fidelity is given by
\begin{equation}
\mathcal{M}_{Forrester,LF}^{1d}((x) = 0.5 \mathcal{M}_{Forrester,HF}^{1d}(xx) + 10 (x-0.5) -5 .
\end{equation}
The function is restricted to the domain $[-3,3]$ as seen in Figure \ref{fig:1d_Forrester}. 
In the following it is considered that the low-fidelity version is less expensive to compute. 

Hence, HK aims at requiring less computational effort to achieve comparable results in comparison to using the HF-model with OK. The error measures between the two fidelity levels computed from 5000 points yield: MAE: $38.7069$, RMAE:$1.8210$ and RMSE:$57.7046$. 
It is therefore obvious that the LF-version only represents a rudimentary approximation of the HF-function.
For this problem initially 10 high-fidelity and 70 low-fidelity samples are created, both of which are shown in Figure \ref{fig:1d_Forrester} with the high-fidelity points symbolized by the red dots and the low-fidelity ones in black. It can be seen that in order to create a high accuracy metamodel adaptive samples need to be created around the lower boundary and closer to the optima of the function. \\ The results when extending the high-fidelity samples with the respective adaptive sampling techniques to 20 samples are listed in Table \ref{tab::Forrester20}.
\begin{table}[h!]
\begin{center}
\resizebox{1.0\textwidth}{!}{%
\begin{tabular}{l|l c c c c} \hline
&Method & MAE & RMAE & RMSE & R$^{2}$ \\ \hline\hline \\
\multirow{1}{*}{\shortstack[l]{Errors after\\10 HF samples}} & TPLHD  & 1.2510 & 1.173 & 6.9847 & 0.9959   \\ \\  \hline \\
\multirow{14}{*}{\shortstack[l]{ Errors after\\20 HF samples}} &TPLHD &  \textbf{0.72635} & 1.1643 &  \textbf{6.8283} & 0.9961  \\
&ACE & 6.4188 & 1.2950 &	10.3576 & 0.9911   \\
&AME & 6.1958 & 1.0472 & 16.4926 & 0.9774   \\ 
&CVD & 5.2940 & 0.7767 & 10.8063 & 0.9903   \\
&CVVor & - & - & - & -  \\ 
&EI & 24.8405 & 1.4131 & 41.1386 & 0.8599  \\
&EIGF & 4.5476 & 0.7656 & 10.6914 & 0.9905   \\  
&LOLA & 1.4546 & 1.1885 & 7.0994 & 0.9958   \\
&MASA &  16.6938 & \textbf{0.6901} & 24.5789 & 0.9499  \\
&MEPE & 17.5687 & 0.9159 & 28.2937 & 0.9337   \\
&MIPT &  6.0020 & 0.9893 & 19.5133 & 0.9684   \\
&MSD &5.0636 & 1.0684 & 17.7577 & 0.9739   \\
&SFCVT &  19.3477 & 0.9434 & 29.5930 & 0.9275   \\
&SSA & 37.2579 & 0.7545 & 42.6750 & 0.8492  \\ 
\end{tabular}
}
\end{center}
\caption[Error measures for $\mathcal{M}_{Forrester,HF}^{1d}$ after 20 HF-samples.]{Error measures for $\mathcal{M}_{Forrester,HF}^{1d}$ after 20 HF samples with 70 LF samples (methods with clustering problems are indicated by empty rows).}\label{tab::Forrester20}
\end{table}

\begin{figure}[h!]
\centering
\begin{subfigure}[t]{0.7\textwidth}
\centering
\includegraphics[scale=0.4]{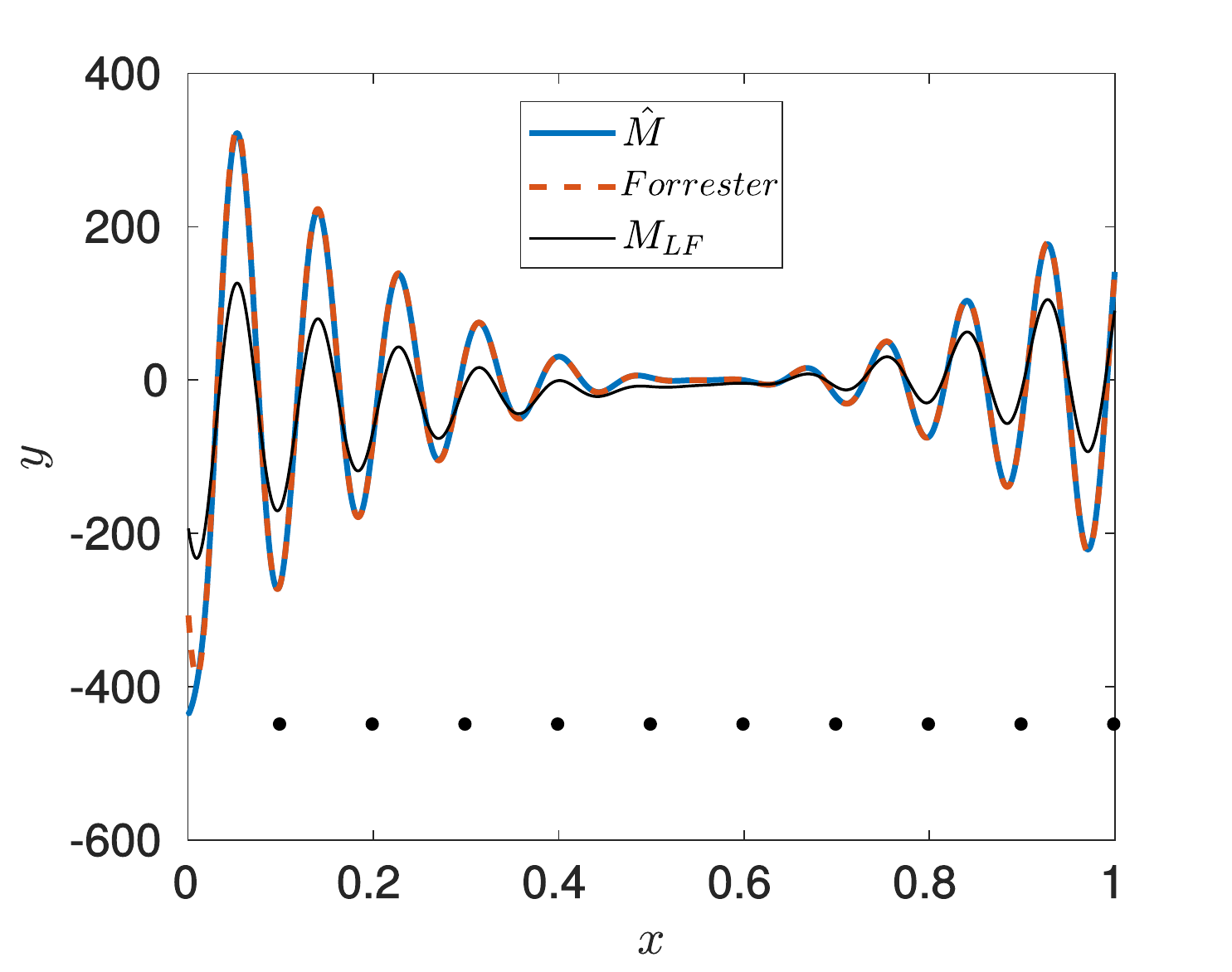} 
\subcaption{Initial metamodel with $\mathcal{M}_{Forrester,LF}^{1d}$}\label{fig:ForesterExplanation1}
\end{subfigure}
\begin{subfigure}[t]{0.7\textwidth}
\centering
\includegraphics[scale=0.4]{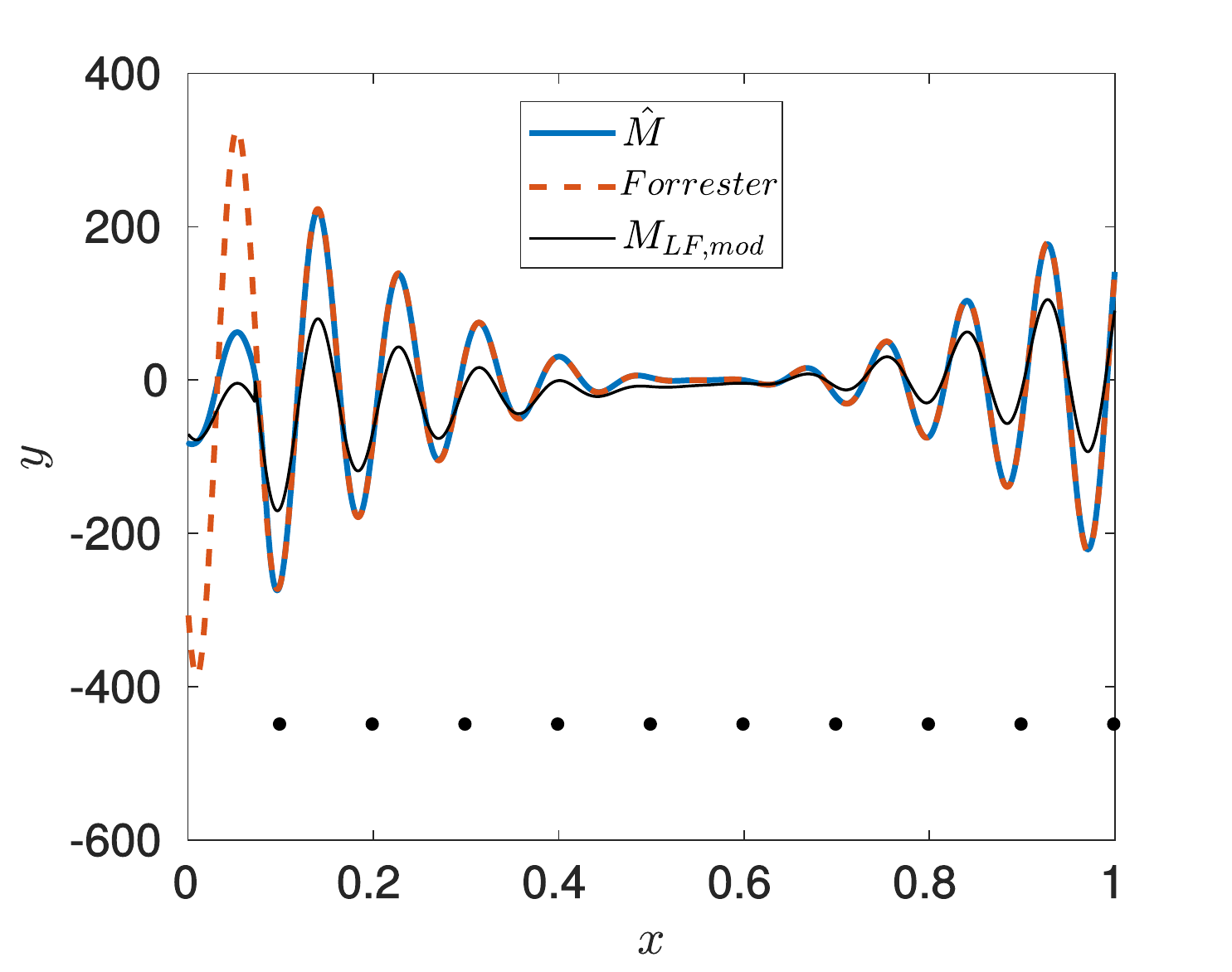} 
\subcaption{Initial metamodel with modfied $M_{LF,mod}$}\label{fig:ForesterExplanation2}
\end{subfigure}
\begin{subfigure}[t]{0.7\textwidth}
\centering
\includegraphics[scale=0.4]{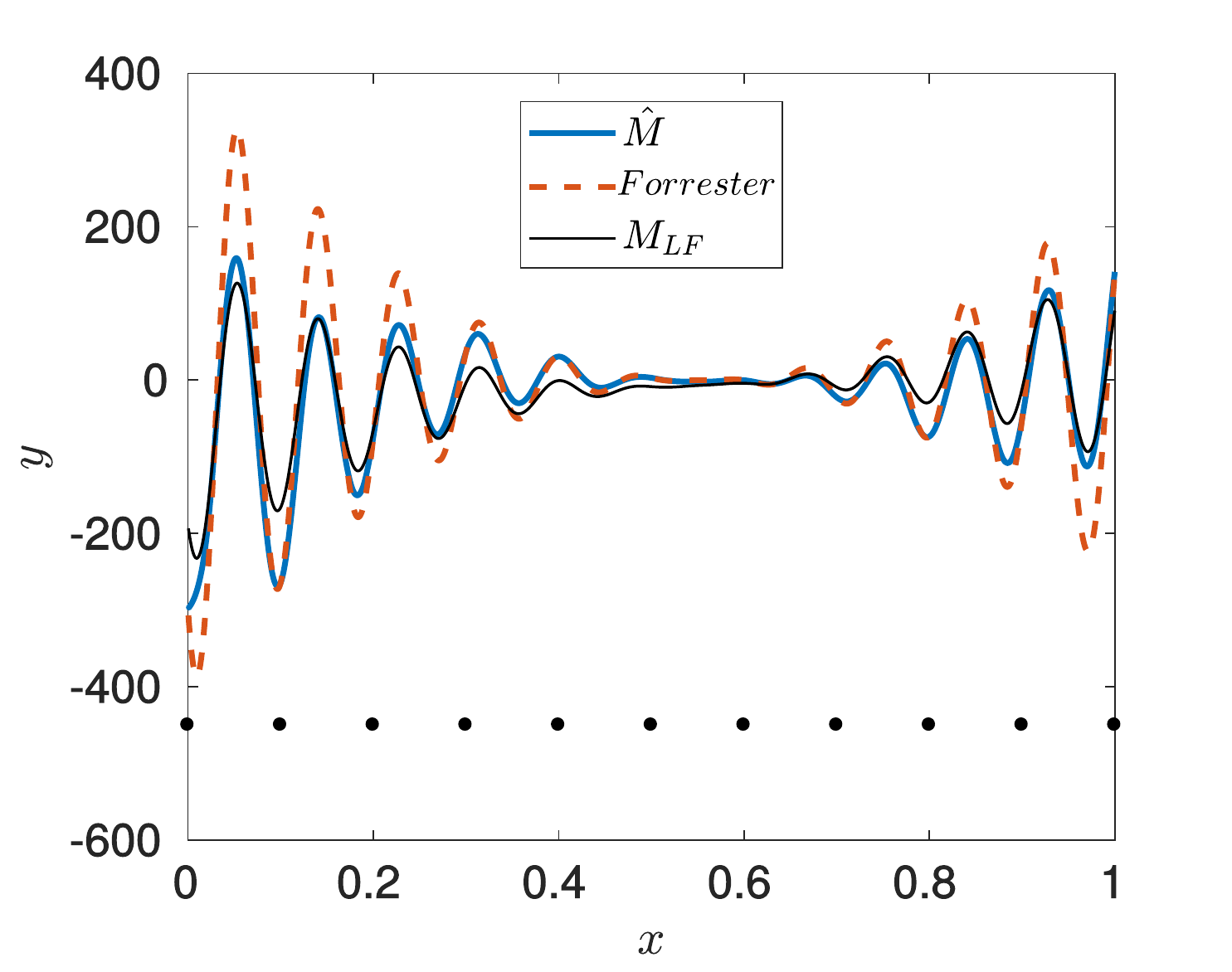} 
\subcaption{Metamodel of $\mathcal{M}_{Forrester,HF}^{1d}$ after adding one sample with EI algorithm.}\label{fig:ForesterExplanation3}
\end{subfigure}
\caption[Explanation as to why the metamodels with TPLHD obtain better error measurements for $\mathcal{M}_{Forrester,HF}^{1d}$]{Explanation as to why the metamodels with TPLHD obtain better error measurements for $\mathcal{M}_{Forrester,HF}^{1d}$. (a) Initial metamodel created with the initial 10 HF samples, (b) Metamodel created with the initial 10 HF samples and a modified low-fidelity function, (c) Metamodel after adding 1 sample with the EI algorithm.}\label{fig:ForesterExplanation}
\end{figure}

Firstly, it can be observed that the error measures get worse in comparison to the initial TPLHD sample and the metamodel created with 20 HF-TPLHD points. This can be explained by the Figures \ref{fig:ForesterExplanation}. The HK metamodel created with the initial 10 HF-samples as indicated with the black dots is shown in Figure \ref{fig:ForesterExplanation1}. It can be seen that even though there are no HF-sample points below a normalized value of around 0.1, the metamodel predicts the actual HF-function very accurately in that area. In fact even though there is no exact sample point on the global maximum the optimum is approximated without much difference. This can be explained by the property of HK to follow the shape of the low-fidelity function in unsampled areas. In fact when modifying the low-fidelity function as shown in Figure \ref{fig:ForesterExplanation2} where in the unsampled HF-area a damping factor is introduced to change the shape of the function, the quality of the metamodel decreases dramatically since the shape of $M_{LF}$ shows a decrease of function value and the HK prediction follows that. Therefore when looking at the adaptive sampling technique, samples are added in the unsampled area near 0 as shown in Figure \ref{fig:ForesterExplanation3} with EI. Here, the HK approximation looses its ties to the low-fidelity function and tries to accurately predict the value at 0. Because none of the optima have a sample point the value of the metamodel changes drastically. In the next few steps until 20 samples and beyond the adaptive sampling algorithm are required to refit the high-fidelity model. This explains the less proficient performance in comparison to general HK created with TPLHD samples. However as will be shown later HK with TPLHD has issues with reducing the error measure below a certain threshold. This is where the adaptive sampling performs exceedingly better. 
As seen in Table \ref{tab::Forrester20} the best performing adaptive sampling technique after 20 samples is LOLA. CVVor is not able to reach 20 samples because of clustering problems. 
\begin{table}[h!]
\begin{center}
\begin{tabular}{l| c c} \hline
Sampling method & RMSE HK & RMSE OK \\ \hline\hline \\
TPLHD &   6.8283 & 140.7895 (50*)  \\
ACE &  10.3576 & 203.7317 (*46)   \\
AME & 16.4926 & 14.5884 (*50)  \\ 
CVD &  10.8063 & 29.6253 (50*)   \\
CVVor & -    & - \\ 
EI &  41.1386 & 81.8168 (*26)\\
EIGF &  10.6914 & 55.6776 (*28)\\  
LOLA & 7.0994 & 98.9523 (*15)   \\
MASA &  24.5789 & 85.5948 (*21)  \\
MEPE &  28.2937 & 6.8543 (*50)  \\
MIPT &   19.5133 & 15.7892 (*50)    \\
MSD & 17.7577 & 135.4633 (*50) \\
SFCVT &  29.5930 & 74.954 (18*)   \\
SSA &  42.6750 & 41.1706 (*36)  \\ 
\end{tabular}
\end{center}
\caption[Error measures for $\mathcal{M}_{Forrester,HF}^{1d}$ comparison of OK and HK]{Comparison of error measures for $\mathcal{M}_{Forrester,HF}^{1d}$ with OK and HK. HK starting from 10 HF samples and inceasing sample size to 20. The value behind the RMSE OK version indicates at what number of samples the best RMSE value is obtainable. This caps off at 50 samples. The difference is due to clustering issues and therefore numerical problems worsening the result.}\label{tab::ForresterHKOK}
\end{table}

\begin{figure}[htpb]
\centering
\includegraphics[scale=0.45]{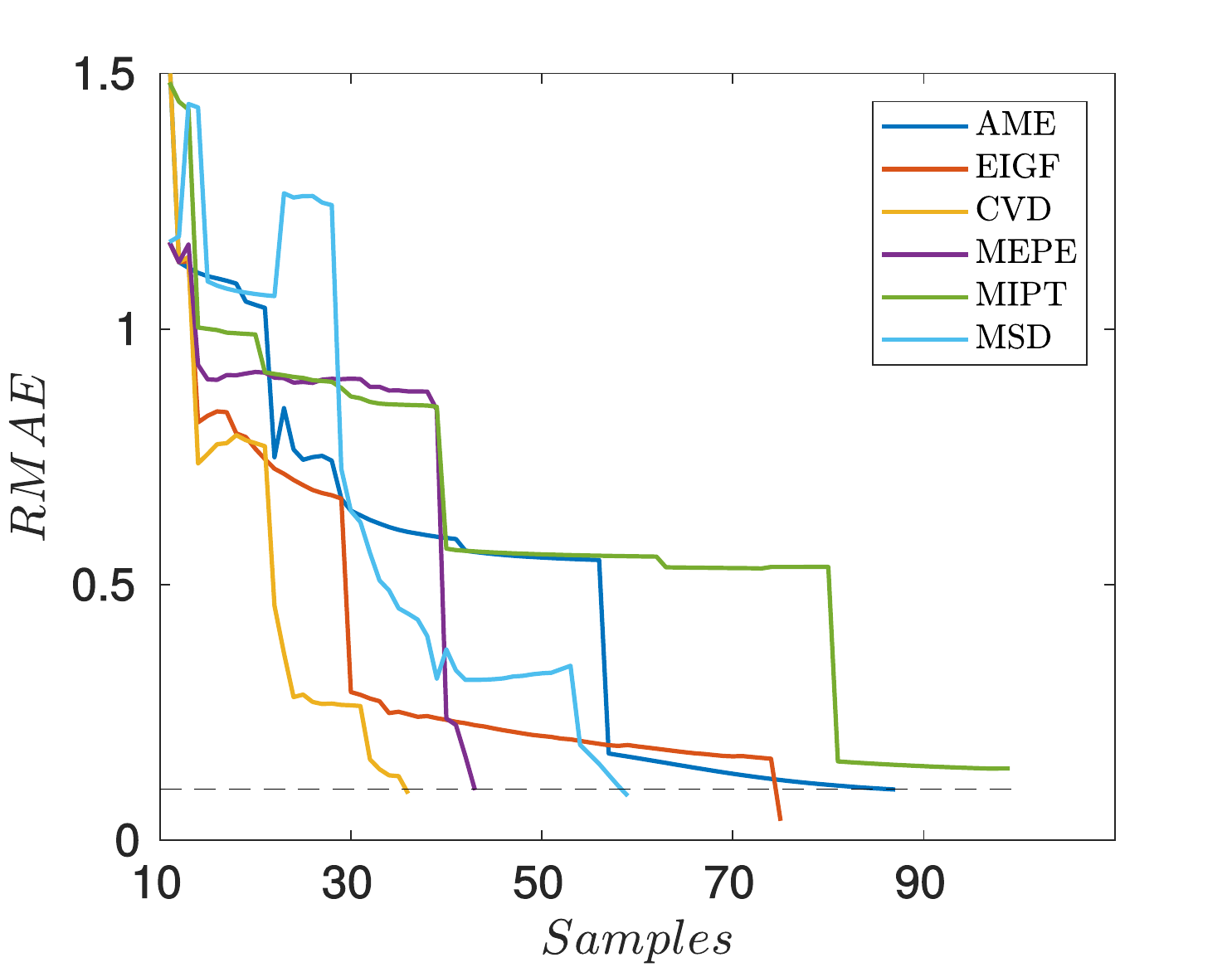}
\caption[Convergence of RMAE error for $\mathcal{M}_{Forrester,HF}^{1d}$ ]{Convergence of RMAE error for $\mathcal{M}_{Forrester,HF}^{1d}$ until threshold of 0.1 for different sampling methods.}%
\label{fig:ForesterConvergence}%
\end{figure}

\begin{figure}[htbp]
\centering
\includegraphics[scale=0.45]{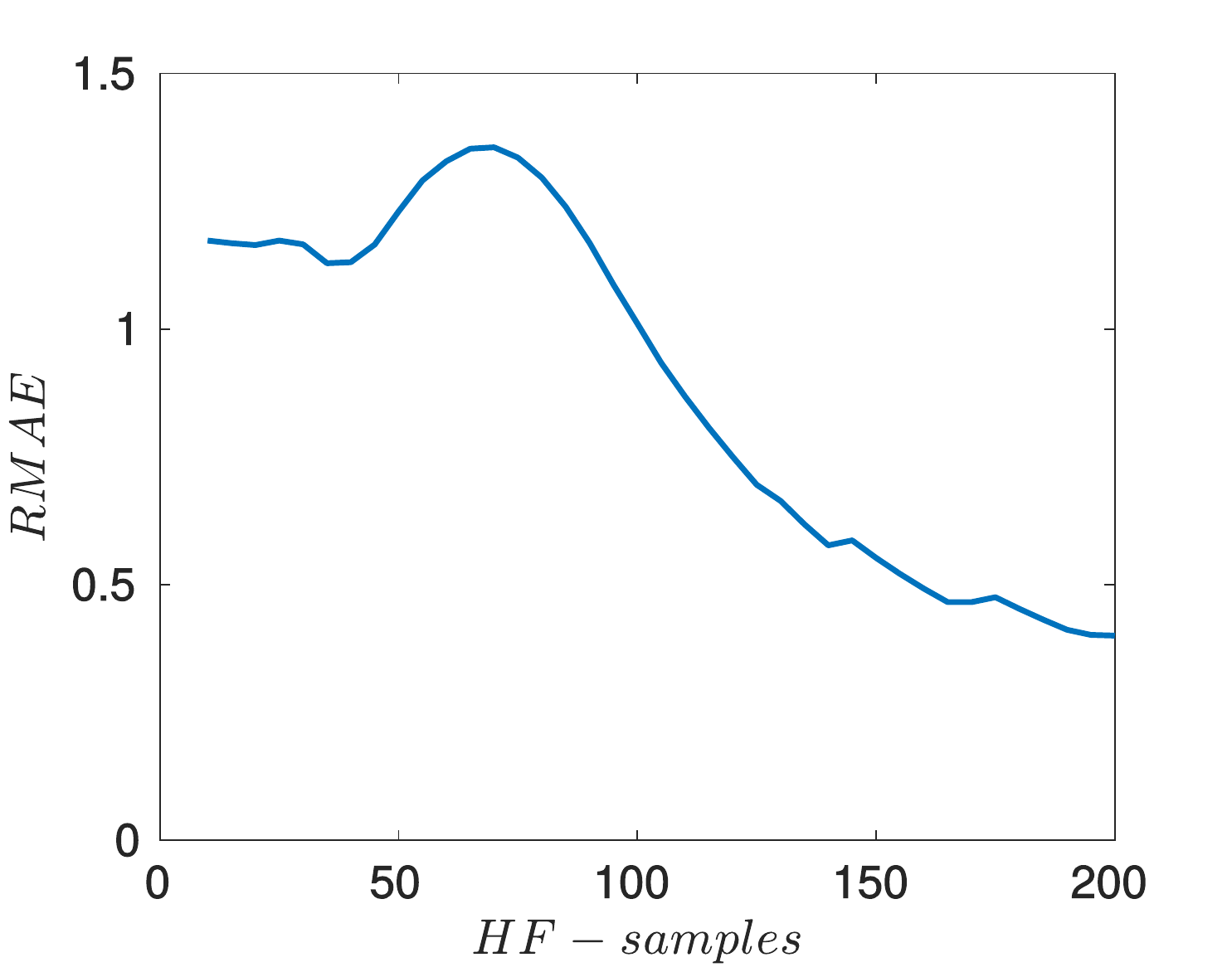}
\caption[Convergence of RMAE error for  $\mathcal{M}_{Forrester,HF}^{1d}$ utilizing HK]{Convergence of RMAE error for  $\mathcal{M}_{Forrester,HF}^{1d}$ utilizing HK generated with samples from TPLHD. }%
\label{fig:ForesterTPLHDConvergence}%
\end{figure}

In order to prove the validity of HK in comparison to OK the data listed in Table 
\ref{tab::ForresterHKOK} is studied. The RMSE values of the adaptive sampling methods created up to 20 samples as well as for TPLHD with the equivalent counterparts are evaluated in comparison with an OK model up to 50 samples.
Since some sampling techniques run into clustering problems in this 1-dimensional example for OK. The best average value with an upper limit of 50 samples is chosen, the respective sample number is indicated in brackets behind the RMSE OK values.
It can be seen that HK produces much better results for most methods except for MEPE and AME. The data can be interpreted to find out how computationally expensive the evaluation of $M_{LF}$ can be in comparison to $M_{HF}$ in order to make HK viable. Since (without taking into account the clustering issue) OK needs to evaluate 30 more high-fidelity samples. Since most of the methods used with OK have not been able to reduce the RMSE error below their HK counterparts it can be said that when using CVD for example the low-fidelity computation can be more than around 2.5 times faster than the HF-version according to this data. 
\begin{figure}[htbp]
\centering
\begin{subfigure}[t]{0.5\textwidth}
\includegraphics[scale=0.33]{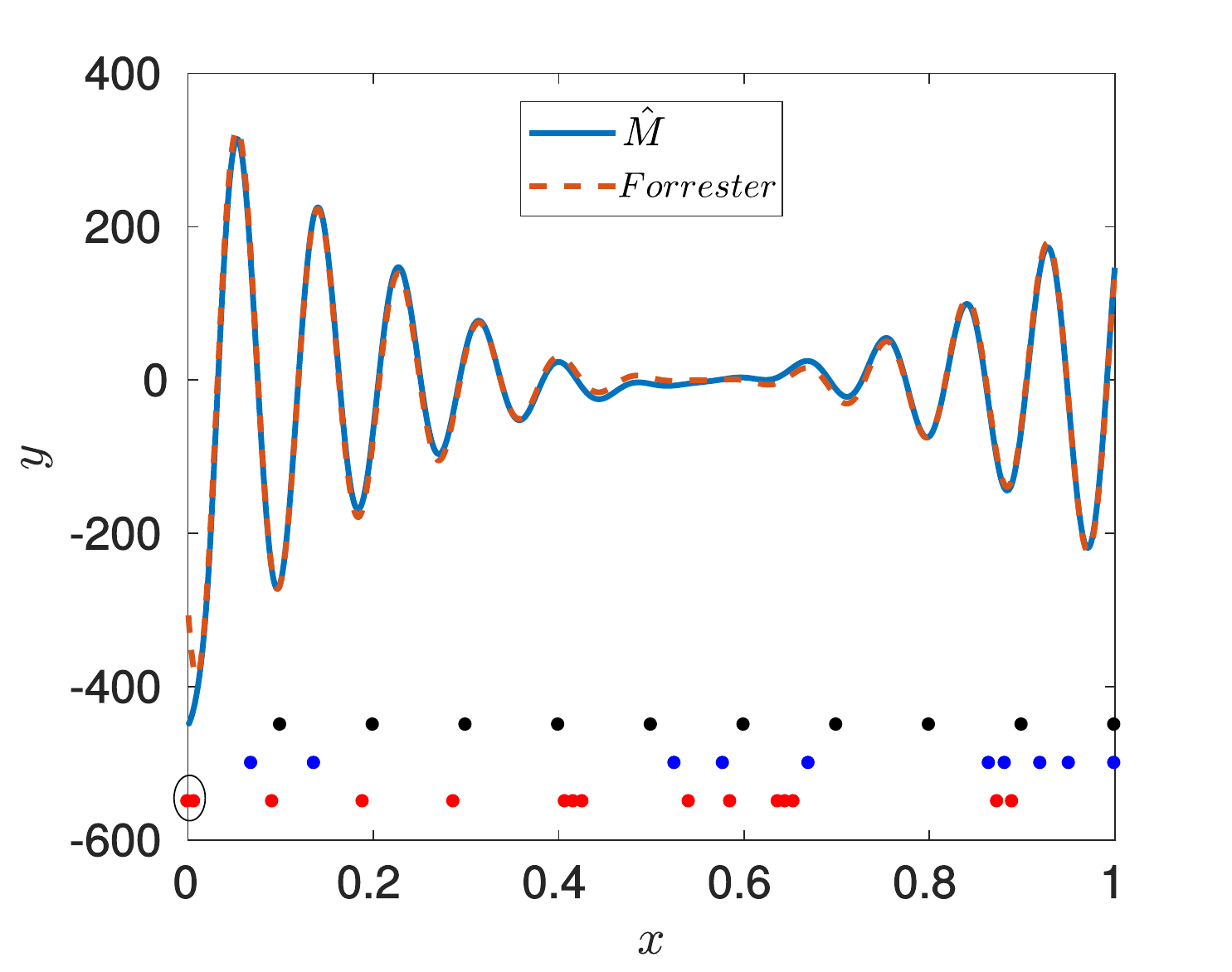}
\subcaption{ACE - 35 samples}\label{fig::ForesterNotACE}
\end{subfigure}%
\begin{subfigure}[t]{0.5\textwidth}
\includegraphics[scale=0.33]{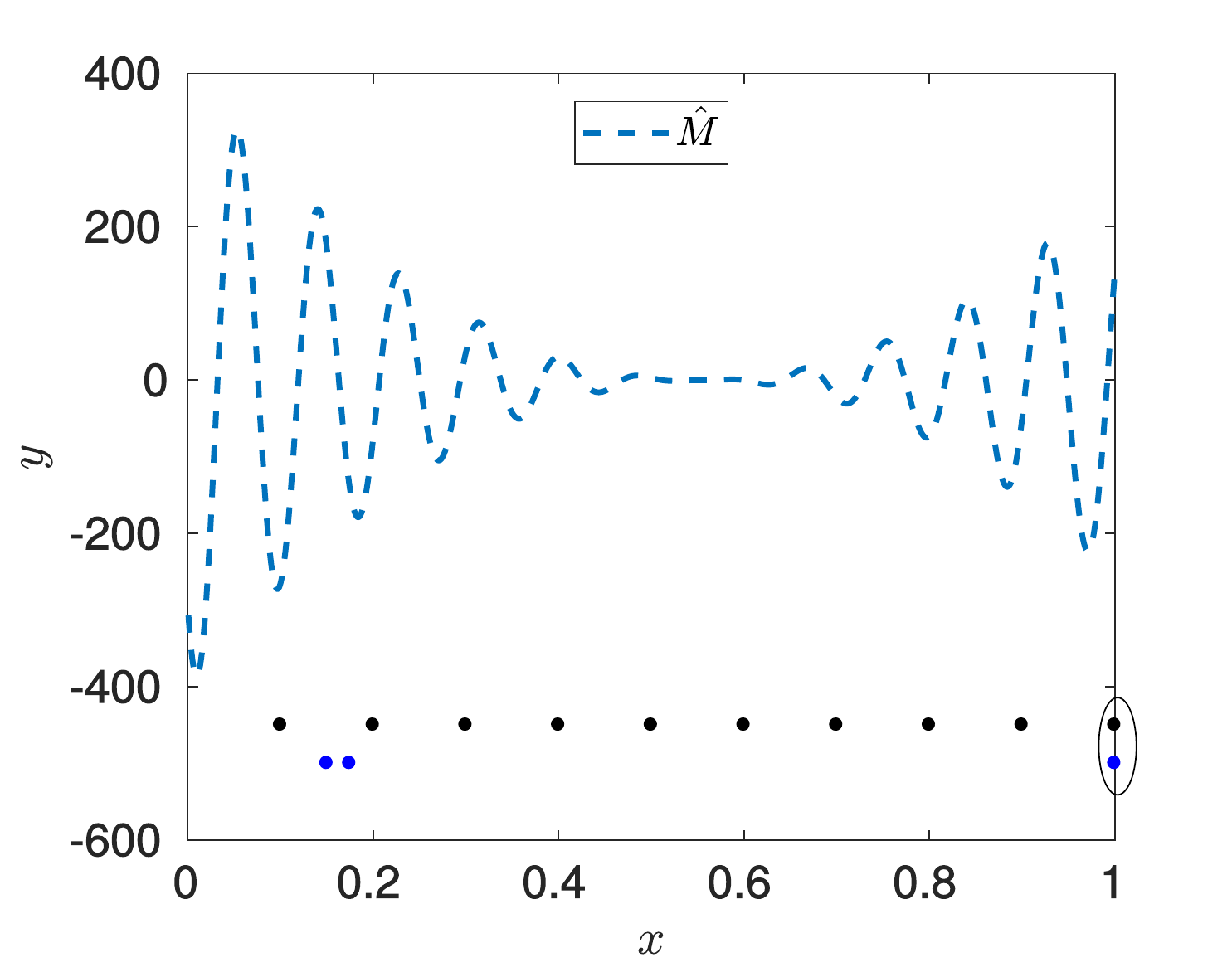}
\subcaption{CVVor - 13 samples}\label{fig::ForesterNotCVVor}
\end{subfigure}
\begin{subfigure}[t]{0.5\textwidth}
\includegraphics[scale=0.33]{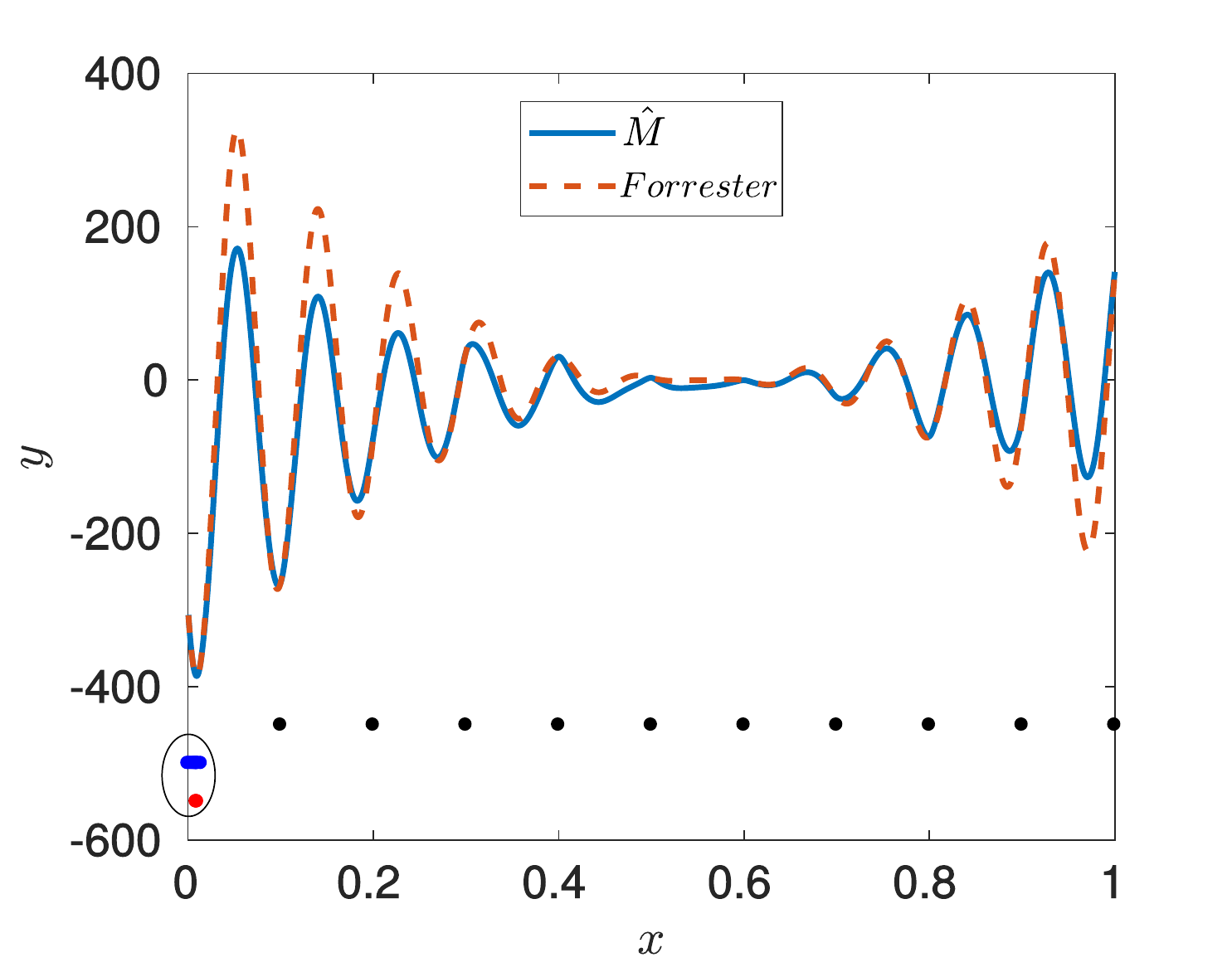}
\subcaption{EI - 25 samples}\label{fig::ForesterNotEI}
\end{subfigure}%
\begin{subfigure}[t]{0.5\textwidth}
\includegraphics[scale=0.33]{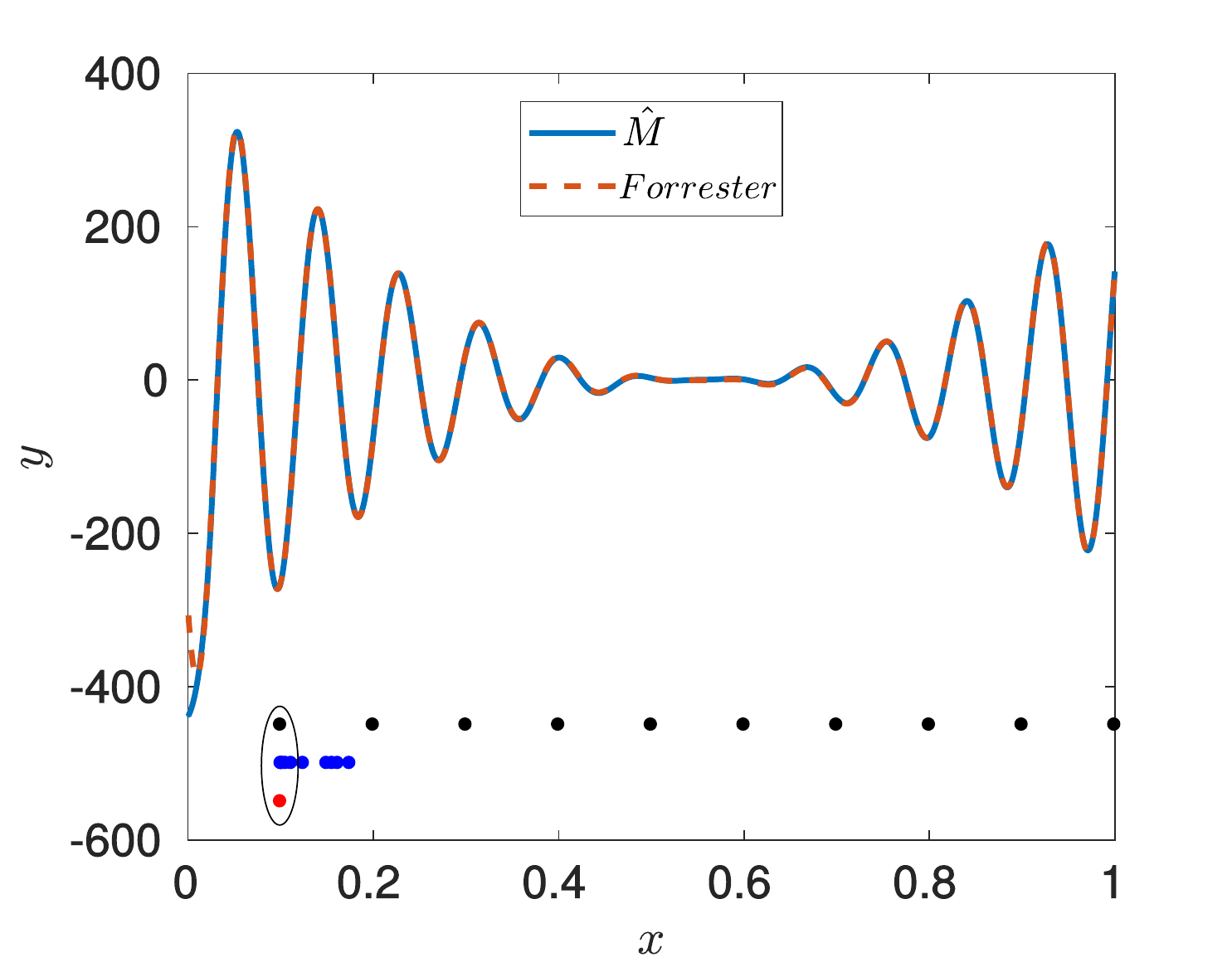}
\subcaption{LOLA - 21 samples}\label{fig::ForesterNotLOLA}
\end{subfigure}
\begin{subfigure}[t]{0.5\textwidth}
\includegraphics[scale=0.33]{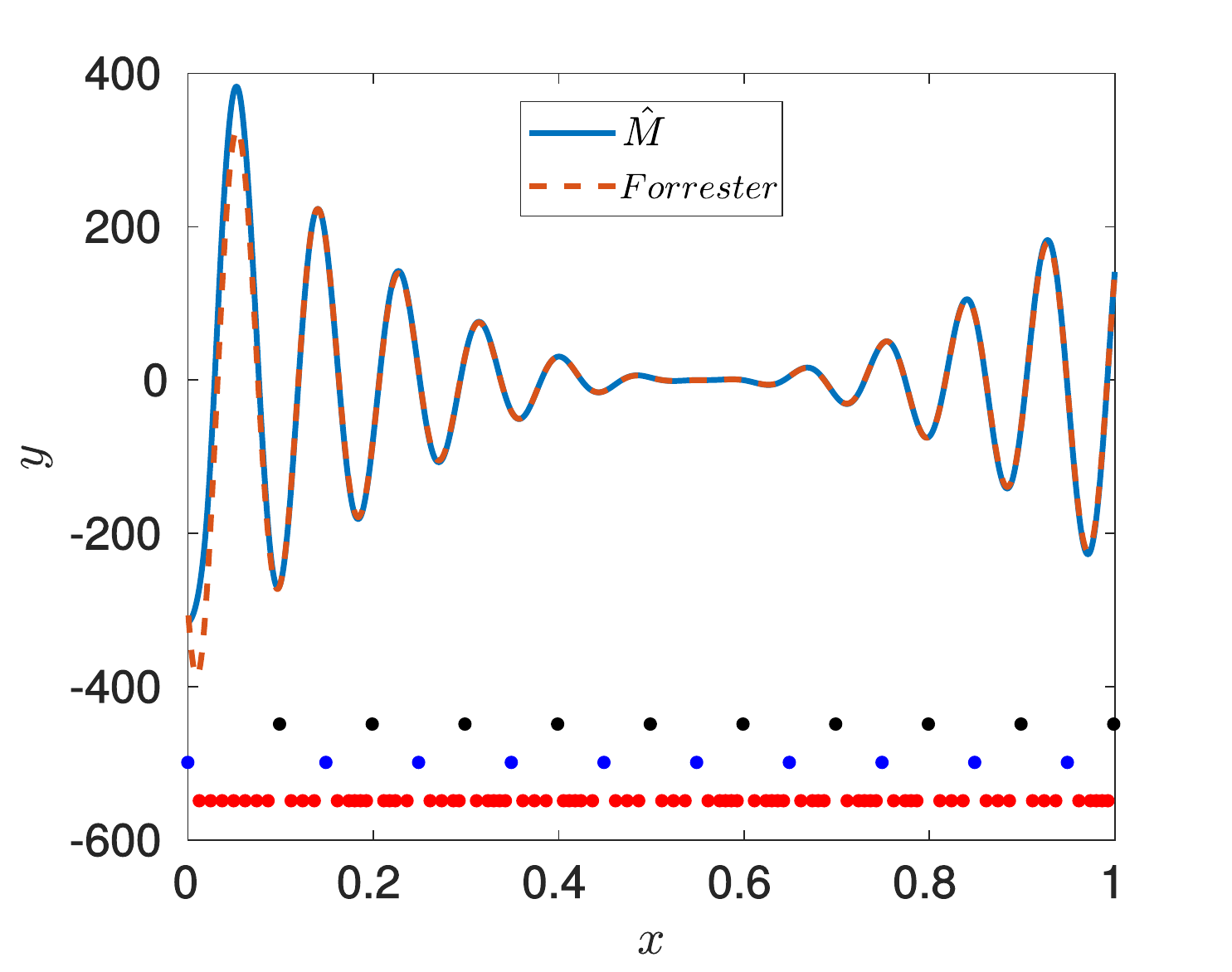}
\subcaption{MIPT - 100 samples}\label{fig::ForesterNotMIPT}
\end{subfigure}%
\begin{subfigure}[t]{0.5\textwidth}
\includegraphics[scale=0.33]{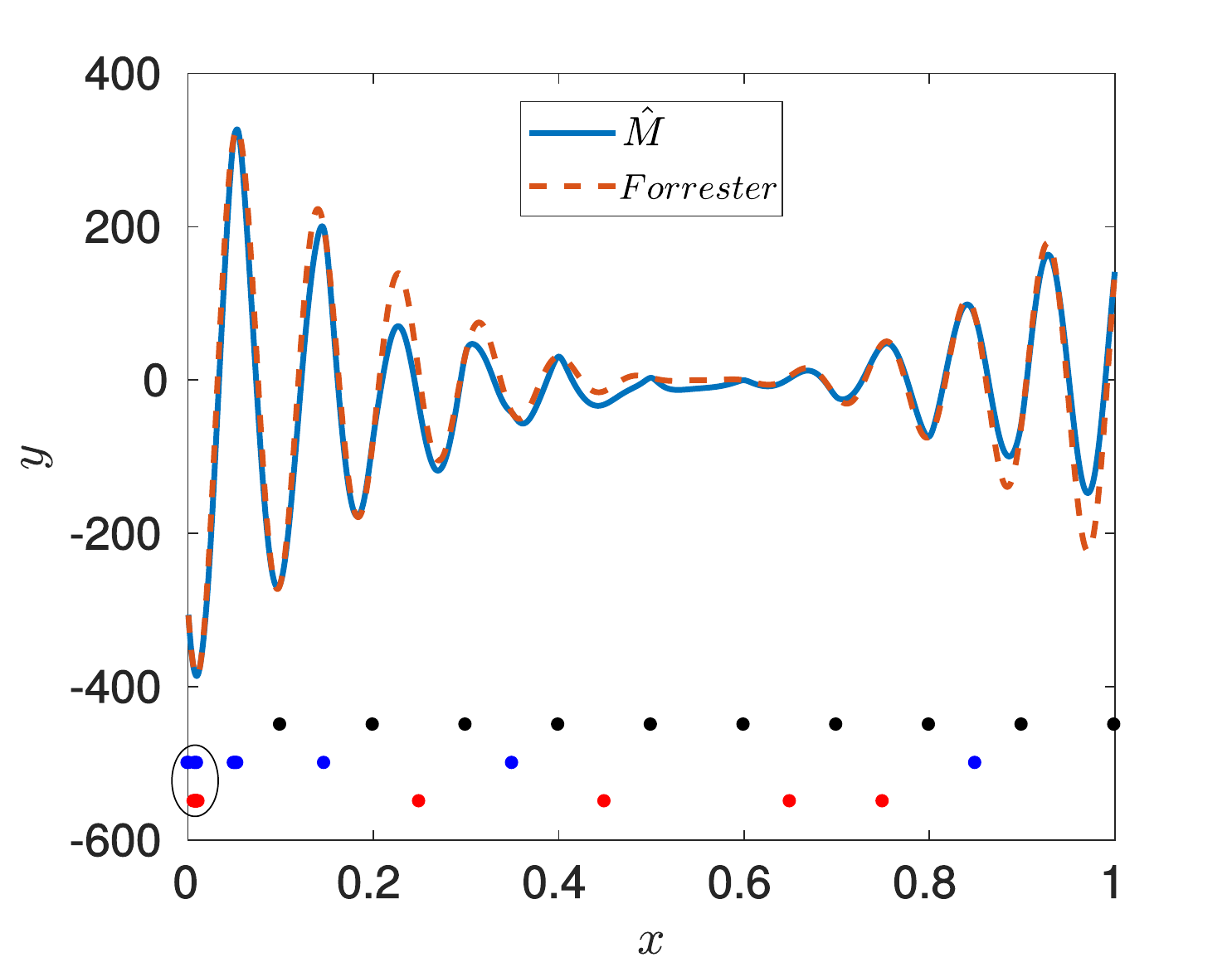}
\subcaption{MASA - 27 samples}\label{fig::ForesterNotMASA}
\end{subfigure}
\begin{subfigure}[t]{0.5\textwidth}
\includegraphics[scale=0.33]{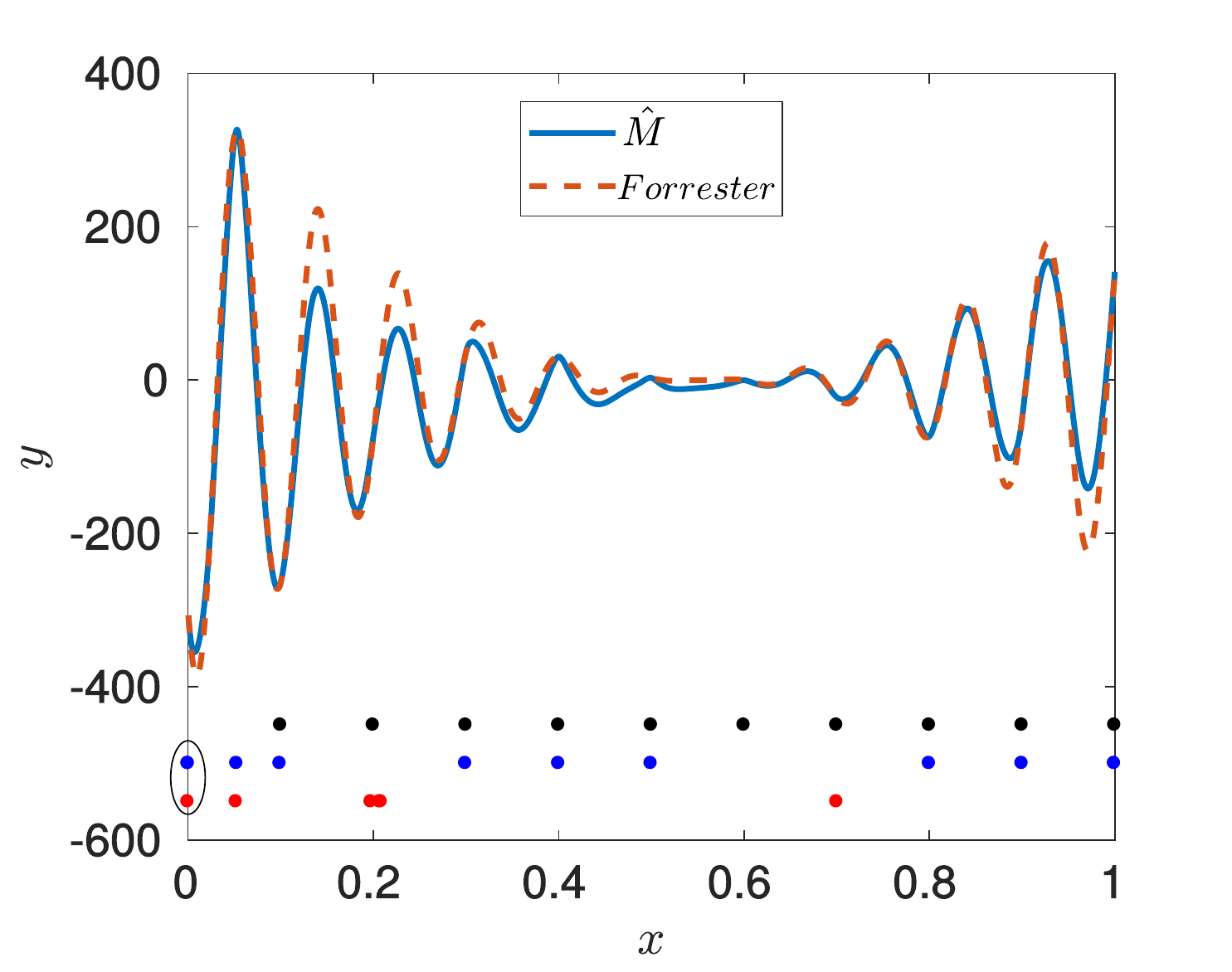}
\subcaption{SFCVT - 26 samples}\label{fig::ForesterNotSFVCT}
\end{subfigure}%
\begin{subfigure}[t]{0.5\textwidth}
\includegraphics[scale=0.33]{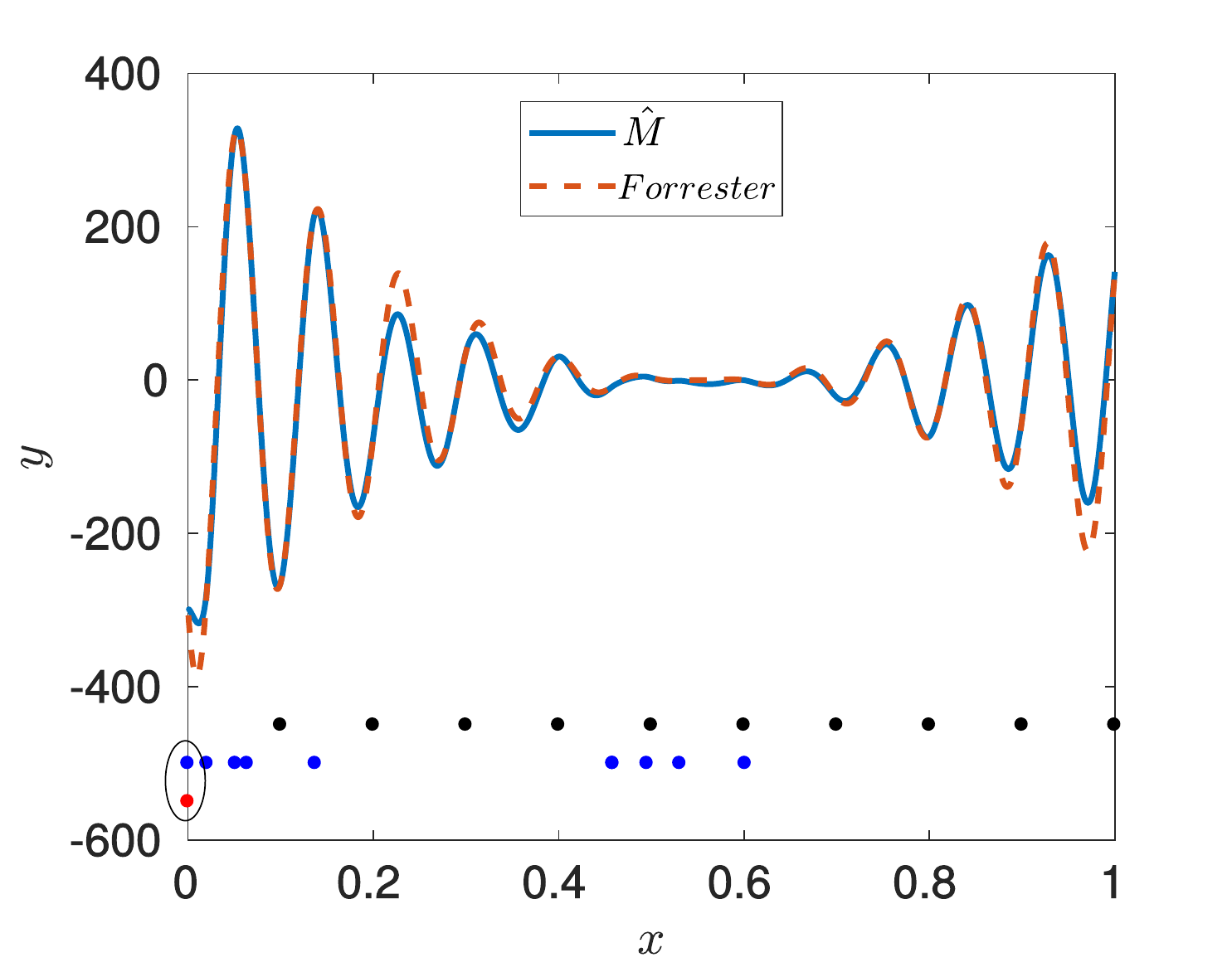}
\subcaption{SSA - 21 samples}\label{fig::ForesterNotSSA}
\end{subfigure}
\caption[Adaptive sampling techniques not able to reduce RMAE$<0.1$ for $\mathcal{M}_{Forrester,HF}^{1d}$]{Adaptive sampling techniques not able to reduce RMAE$<0.1$ for $\mathcal{M}_{Forrester,HF}^{1d}$. The cluster region is highlighted with an ellipse. MIPT has no clustering issue but needs over 100 samples. Black dots: 10 initial samples. Blue dots: 10th to 20th sample. Red dots: Supplementary samples.}\label{fig::ForesterNot}
\end{figure}
In a next step the convergence of RMAE is studied. Here a considerable low value for the complexity of the function is chosen to deliberately see which techniques show clustering problems. 8 out of the 13 investigated methods (ACE, CVVor, EI, LOLA, MASA, MIPT, SFCVT and SSA) were unable to reach the set threshold within the first 100 HF-sample points. This is mostly due to clustering issues as shown in Figure \ref{fig::ForesterNot}. Here, each subplot presents an adaptive sampling technique and shows the samples until numerical problems emerge or more than 100 samples are required without reaching the set threshold. The metamodel at 20 samples is plotted in blue and the target high-fidelity Forrester function is indicated by a dotted red-line. The black dots symbolize the initial HF samples, the blue dots represent the next 10 samples up to 20 and the rest of the points are shown in red. In order for the RMAE value to be reduced the optima need to be fitted very proficiently which means for HK that HF samples need to be created around the maxima and minima. \\
The samples for ACE are depicted in Figure \ref{fig::ForesterNotACE}. The method does not sample near zero in the first 20 points which shows that the exploitation factor of the method is not sufficient. However some of the first samples points are created close to the maxima in the lower half of the domain. In the end ACE is only able to generate 35 HF samples since the points cluster around 0. \\ As for the 1-dimensional OK function CVVor (Figure \ref{fig::ForesterNotCVVor}) produces the worst results and is the first to run into numerical problems. Here after 13 points two samples are superposed around 1 as indicated with the ellipse.  The results of the EI algorithm are displayed in Figure \ref{fig::ForesterNotEI}. As remarked in earlier OK problems EI tends to cluster samples around the minimum. This is also the case in this problem and EI is not able to continue after 25 sample points. 
\newpage
The LOLA technique (Figure \ref{fig::ForesterNotLOLA}) stops after 21 points since a clustering phenomenon appears around a point around 0.1. It indicates a high computed gradient value in that area. \\
The results for MIPT are displayed in Figure \ref{fig::ForesterNotMIPT}. It is the only method to run through to the cut off of 100 HF-samples. The points are all evenly spaced in the domain and a constant decrease of the RMAE value is visible. However the threshold is not reached until 100 points. At the 100th samples point MIPT shows an average RMAE measure of around $0.1402$. Therefore further points are needed with this method. \\ Similar to EI and ACE, MASA (Figure \ref{fig::ForesterNotMASA}) creates a cluster of points around the minimum value close to the lower boundary and the numerical process stops after 27 samples points. \\
SFCVT (Figure \ref{fig::ForesterNotSFVCT}) and SSA (Figure \ref{fig::ForesterNotSSA}) show similar clustering problems to earlier methods.
5 out of the 13 presented techniques are able to reduce the RMAE value below the target. The convergence of this value for these 6 techniques (AME, EIGF, CVD, MEPE and MSD) and the aforementioned MIPT method are shown in Figure \ref{fig:ForesterConvergence}. On average AME needs 87 samples, EIGF 75, MEPE 43, MSD 59 and CVD the lowest value of 36 samples. Since RMAE is dependent on sampling points being close to the optimum values the convergence plot as seen in Figure \ref{fig:ForesterConvergence} shows jumps when a better sample is found. As a comparison between the adaptive techniques utilizing HK and HK with TPLHD the convergence seen in Figure \ref{fig:ForesterConvergence} can be contrasted to the RMAE convergence as displayed in Figure \ref{fig:ForesterTPLHDConvergence}.

The error measure over the TPLHD sample size is shown in Figure \ref{fig:ForesterTPLHDConvergence}. Starting from 10 samples the sample size is increased by 5 samples each step. It can be seen that after a peak the RMAE value shows a rather smooth convergence until 200 sample points. However the lowest reached value after this amount of samples is around 0.4. Therefore much more points are needed to reduce the measure below the target. Here the adaptive sampling techniques are able to work a lot more efficient.

\begin{figure}[htbp]
\centering
\begin{subfigure}[t]{0.5\textwidth}
\includegraphics[scale=0.4]{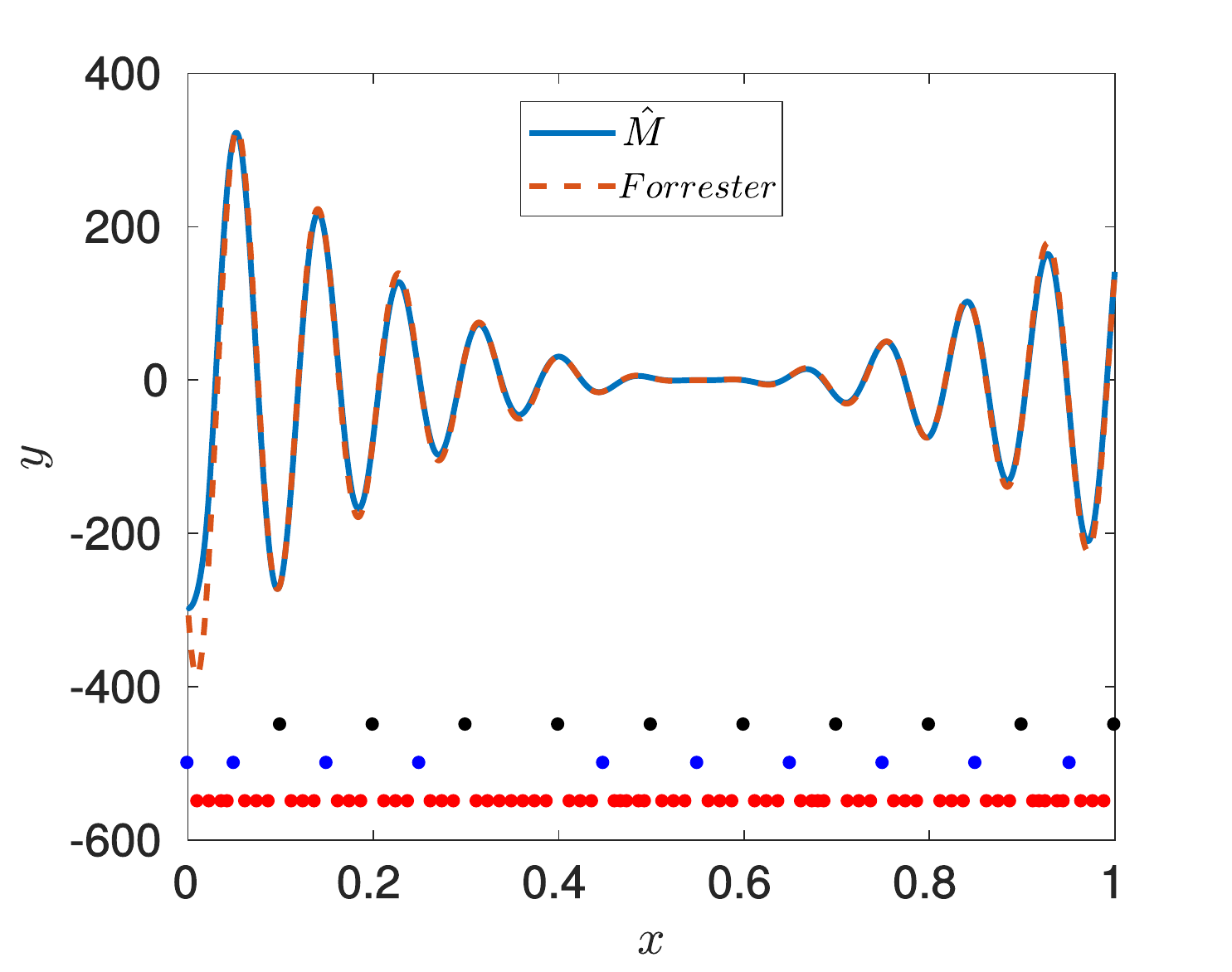}
\subcaption{AME - 87 samples}\label{fig:ForesterYesAME}
\end{subfigure}%
\begin{subfigure}[t]{0.5\textwidth}
\includegraphics[scale=0.4]{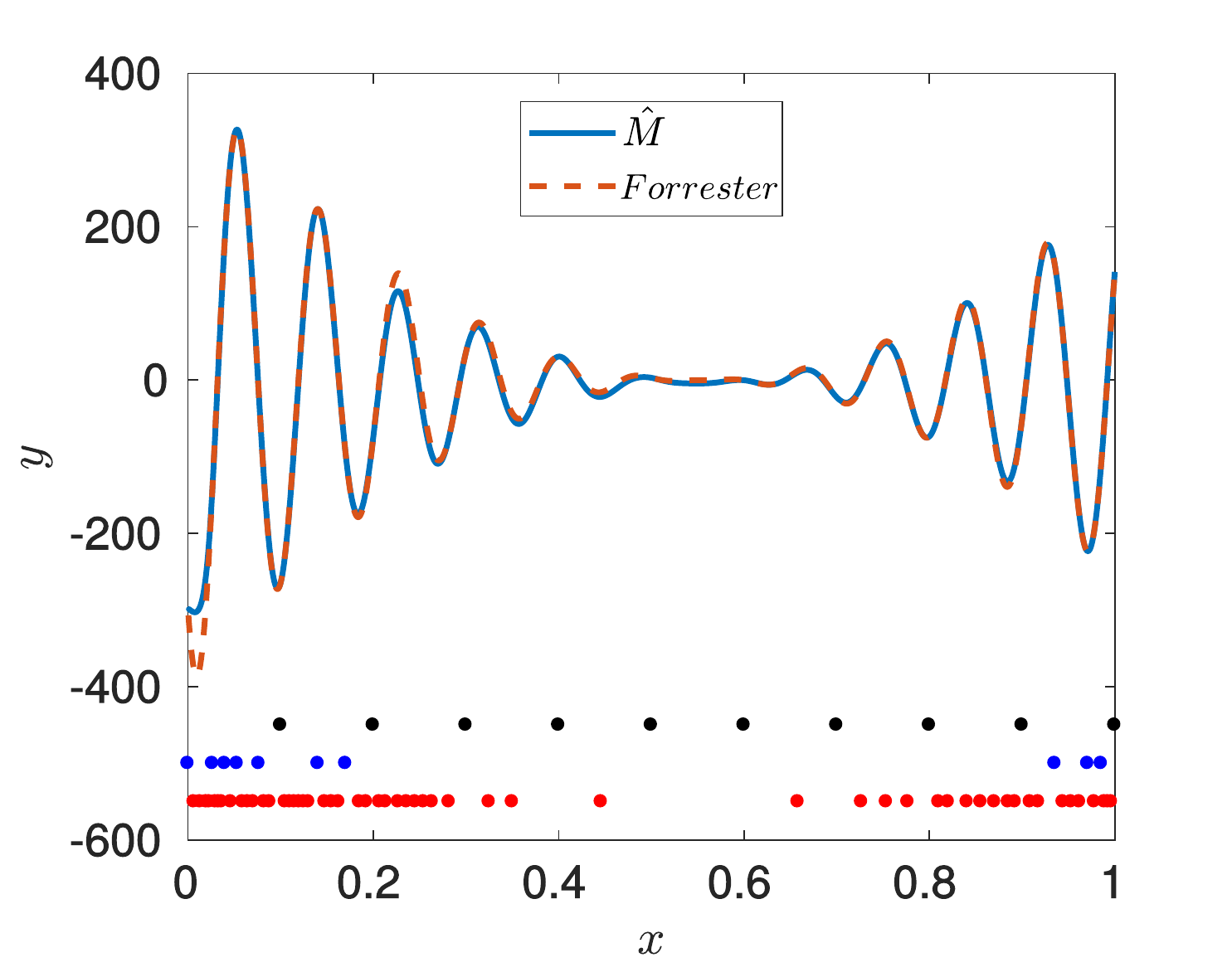}
\subcaption{EIGF - 75 samples}\label{fig:ForesterYesEIGF}
\end{subfigure}
\begin{subfigure}[t]{0.5\textwidth}
\includegraphics[scale=0.4]{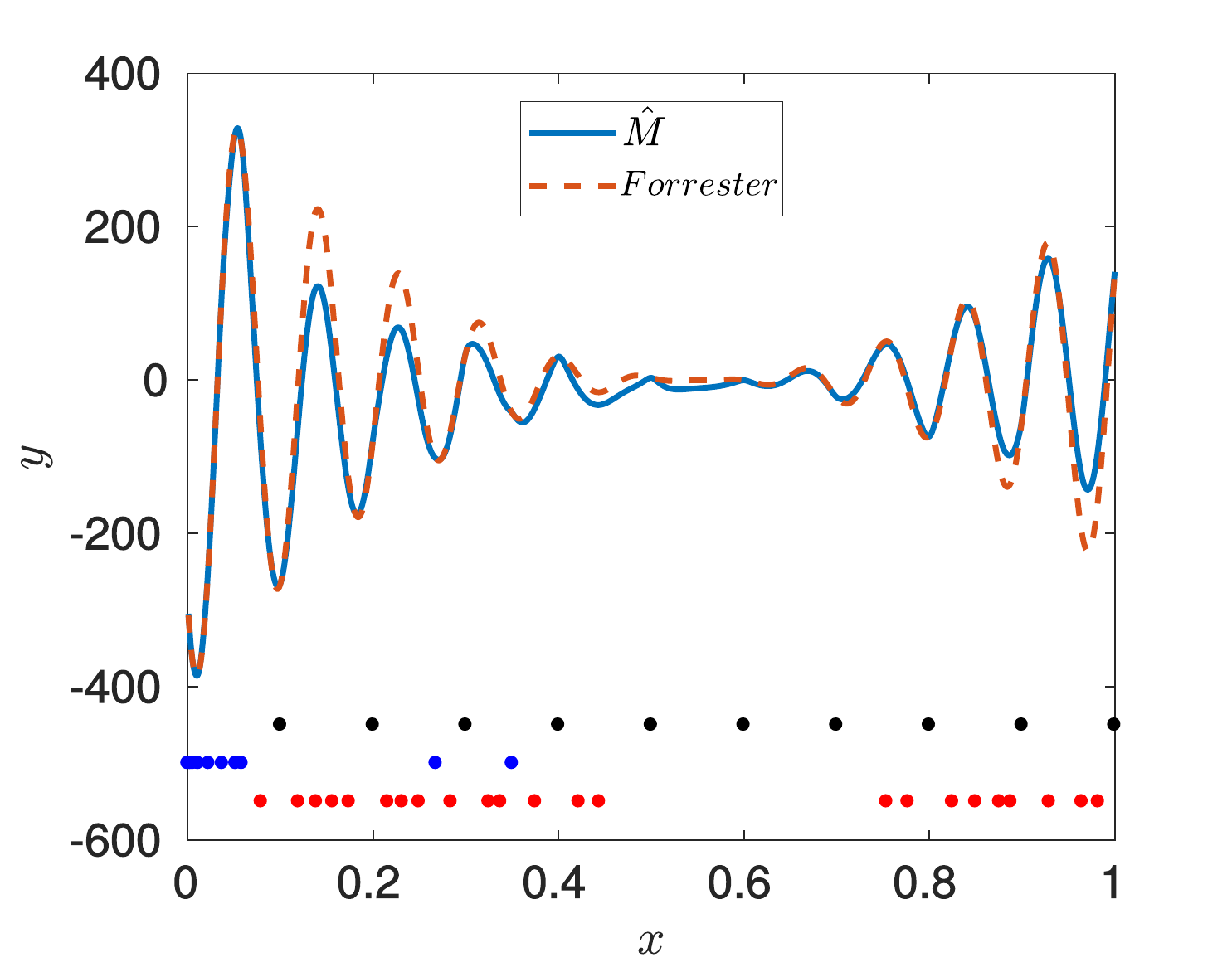}
\subcaption{MEPE - 43 samples}\label{fig:ForesterYesMEPE}
\end{subfigure}%
\begin{subfigure}[t]{0.5\textwidth}
\includegraphics[scale=0.4]{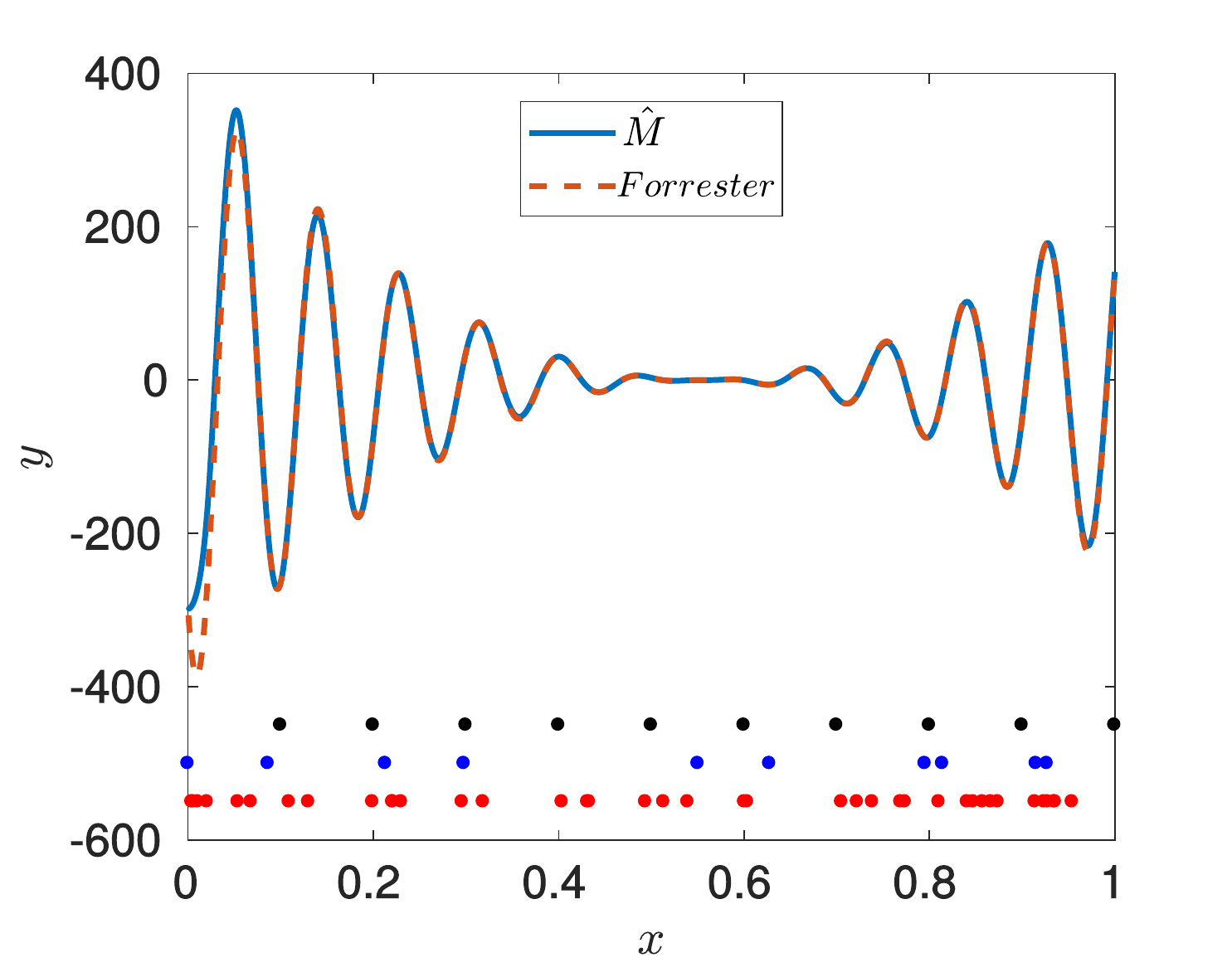}
\subcaption{MSD- 59 samples}\label{fig:ForesterYesMSD}
\end{subfigure}
\begin{subfigure}[t]{0.5\textwidth}
\includegraphics[scale=0.4]{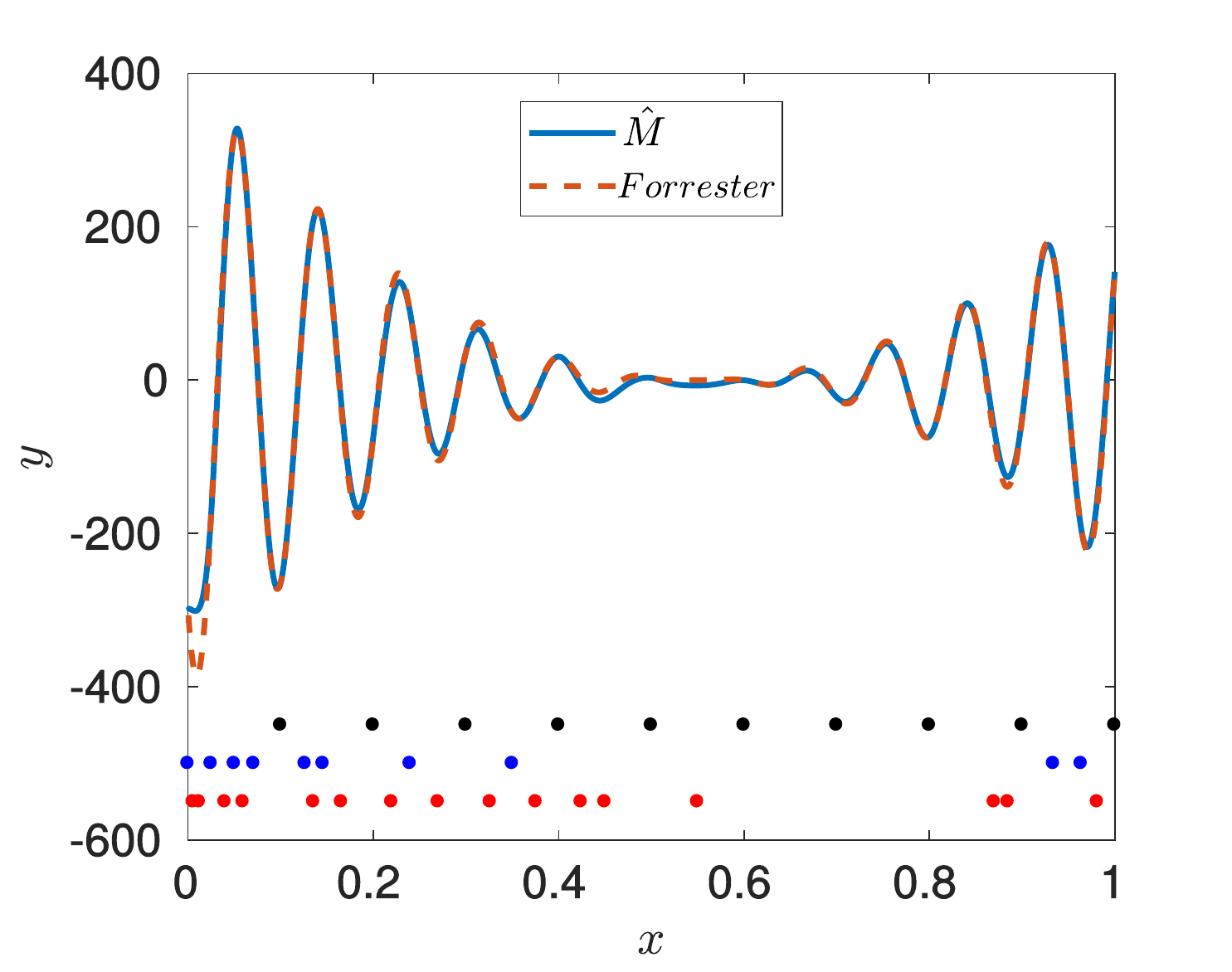}
\subcaption{CVD - 36 samples}\label{fig:ForesterYesMSE}
\end{subfigure}
\caption[Location of samples for $\mathcal{M}_{Forrester,HF}^{1d}$ reaching the error threshold of RMAE$<0.1$]{Location of samples for $\mathcal{M}_{Forrester,HF}^{1d}$ reaching the error threshold of RMAE$<0.1$. Lines show function and metamodel at 20 samples. Black dots: 10 initial samples. Blue dots: 10th to 20th sample. Red dots: Supplementary samples. }\label{fig:ForesterYes}
\end{figure}

The sample points of the techniques that converge to the error threshold are shown in Figure \ref{fig:ForesterYes}. 
Again, the black dots symbolize the initial points, the blue dots the first 10 adaptively generated samples and the red point the other points until the threshold is reached.
It can be seen that AME (Figure \ref{fig:ForesterYesAME}) focuses on exploration. All created points are equally-spaced in the domain. This way the method reduces the RMAE measure however with more points then needed. \\
The 75 HF-sample points needed for EIGF are displayed in Figure \ref{fig:ForesterYesEIGF}. The exploitation character of this method is clearly visible since the method tends to sample around high absolute values of the target function. The unknown areas around the lower boundary are proficiently sampled to reduce the error efficiently. 
The second best method for this application is MEPE (Figure \ref{fig:ForesterYesMEPE}) for which 43 samples are enough. The method shows a good balance between exploitation (sampling around the optima) and exploration (generating sample around the lower bound). Similar to EIGF no extra samples are created around the flat domain of the HF function. \\
MSD is illustrated in Figure \ref{fig:ForesterYesMSD}. MSD is a purely exploration-based method and therefore evenly spaces the created sample points. \\
As in the one-dimensional OK benchmark the most proficient method is CVD (Figure \ref{fig:ForesterYesMSE}). It shows a sufficient balance between exploration and exploitation, so it is able to sample around the lower boundary as well as the optima values.
\clearpage
\subsection{Two-dimensional Currin function}\label{sec::Currin}
\begin{figure}[h!]
\centering
\begin{subfigure}[t]{0.5\textwidth}
\begin{tikzpicture}
[
declare function={ 
fact1(\x, \y) = 1 - exp(-1/(2*\y));
fact2(\x, \y) = 2300*\x^3 + 1900*\x^2 + 2092*\x + 60;
fact3(\x, \y) = 100*\x^3 + 500*\x^2 + 4*\x + 20;
fHF(\x, \y) = fact1(\x, \y) * fact2(\x, \y)/fact3(\x, \y);  
                 },
]
\begin{axis}
[
xlabel={$x_{1}$},
ylabel={$x_{2}$},
zlabel={$\mathcal{M}_{Currin,HF}^{2d}(x_{1},x_{2})$},
view={15}{25},
zmin=0,zmax=15.0,
ztick = {0.0,5.0,10.0,15.0},
  zticklabels = {0,5,10,15},
]
\addplot3[surf,domain=0:1,domain y=0:1]
{fHF(x,y)};
\end{axis}
\end{tikzpicture}
\subcaption{High-fidelity function $\mathcal{M}_{Currin,HF}^{2d}$}\label{fig::}
\end{subfigure}
\begin{subfigure}[t]{0.5\textwidth}
\begin{tikzpicture}
[
declare function={ 
fact1(\x, \y) = 1 - exp(-1/(2*\y));
fact2(\x, \y) = 2300*\x^3 + 1900*\x^2 + 2092*\x + 60;
fact3(\x, \y) = 100*\x^3 + 500*\x^2 + 4*\x + 20;
fHF(\x, \y) = fact1(\x, \y) * fact2(\x, \y)/fact3(\x, \y);  
maxarg(\x, \y) = max(0, \y-1/20);
yh1(\x, \y)  = fHF(\x+1/20, \y+1/20);
yh2(\x, \y)  = fHF(\x+1/20, maxarg(\x, \y));
yh3(\x, \y) = fHF(\x-1/20, \y+1/20);
yh4(\x, \y)  = fHF(\x-1/20, maxarg(\x, \y));
fLF(\x, \y) = (0.4*yh1(\x, \y) +0.5* yh2(\x, \y) +0.8* yh3(\x, \y) + yh4(\x, \y)) / 4;
                 },
]
\begin{axis}
[
xlabel={$x_{1}$},
ylabel={$x_{2}$},
zlabel={$\mathcal{M}_{Currin,LF}^{2d}(x_{1},x_{2})$},
view={15}{25},
zmin=0,zmax=15.0,
ztick = {0.0,5.0,10.0,15.0},
  zticklabels = {0,5,10,15},
]
\addplot3[surf,domain=0:1,domain y=0:1]
{fLF(x,y)};
\end{axis}
\end{tikzpicture}
\subcaption{Low-fidelity function $\mathcal{M}_{Currin,LF}^{2d}$}\label{fig::}
\end{subfigure}
\caption[Multifidelity in the Currin function]{Multifidelity in the Currin function. (a) $\mathcal{M}_{Currin,HF}^{2d}$, (b) $\mathcal{M}_{Currin,LF}^{2d}$}\label{fig::CurrinHF}
\end{figure}
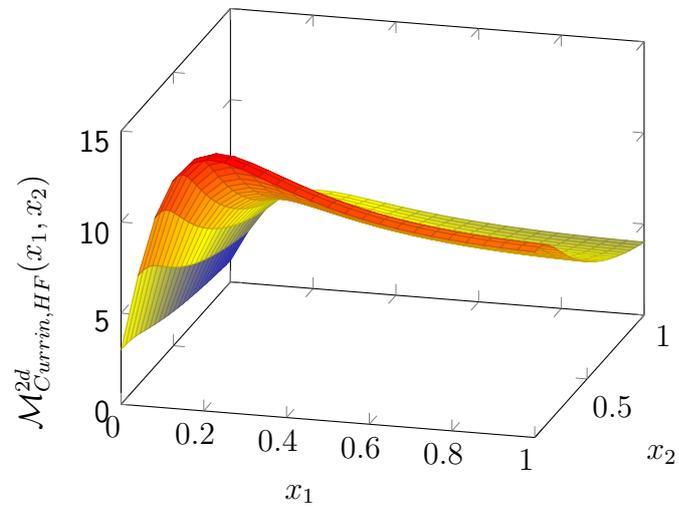
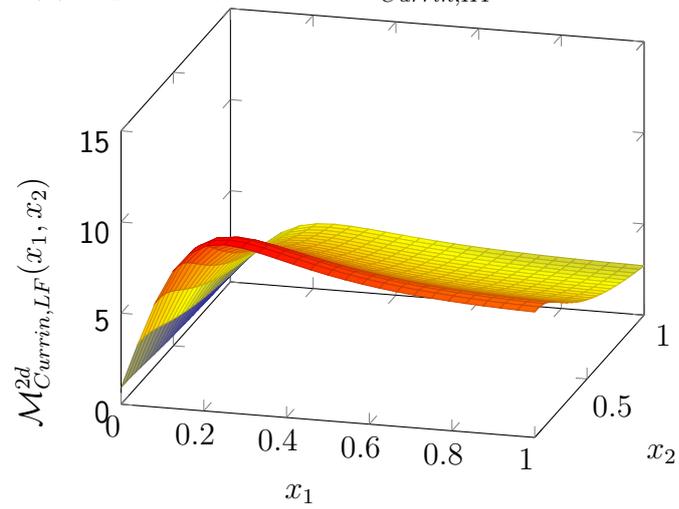

\begin{figure}
\centering
\begin{subfigure}[t]{0.5\textwidth}
\includegraphics[scale=0.4]{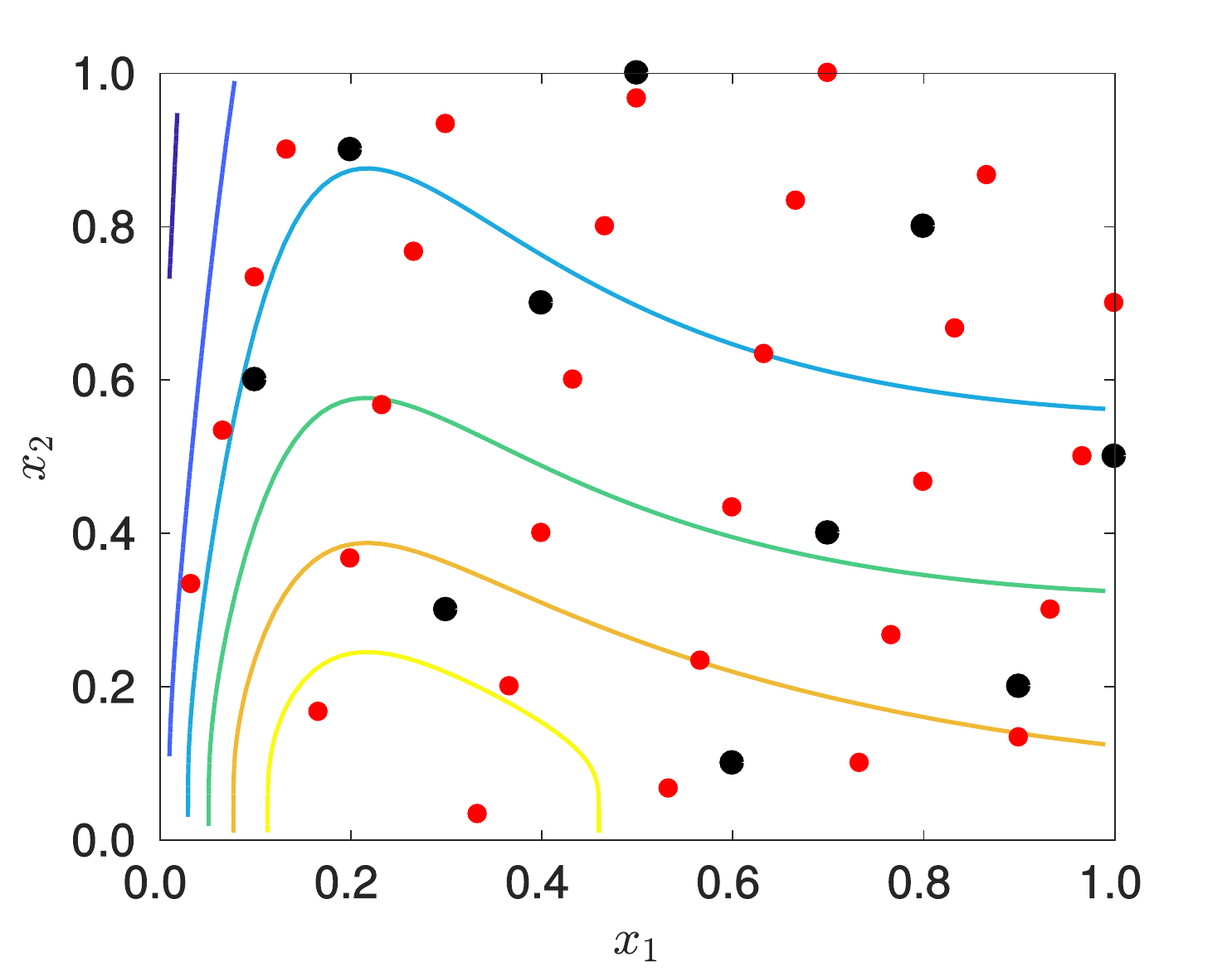} 
\subcaption{Initial samples}\label{fig:CurrinInitial}
\end{subfigure}%
\begin{subfigure}[t]{0.5\textwidth}
\includegraphics[scale=0.4]{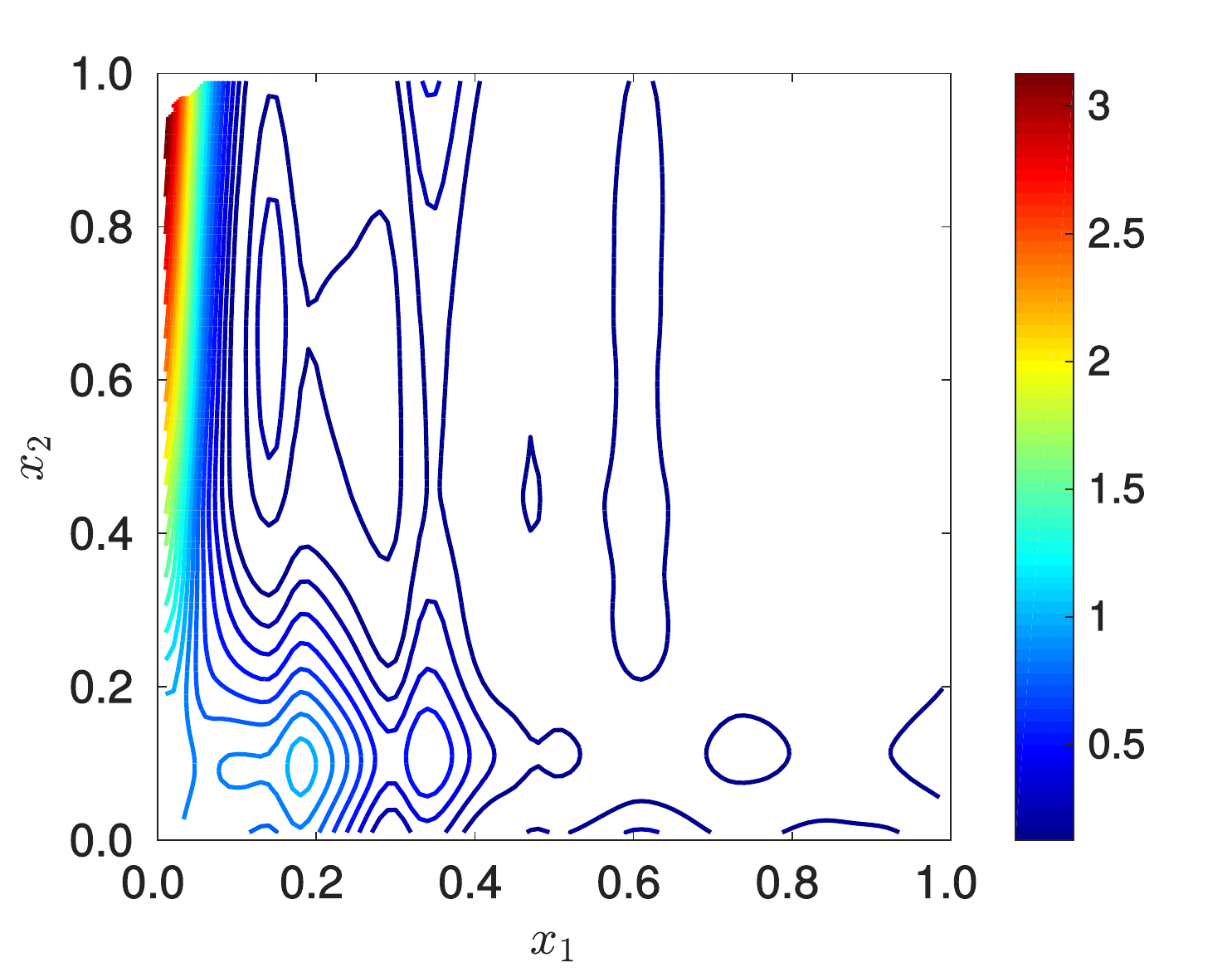} 
\subcaption{Initial absolute error}\label{fig:CurrinInitialError}
\end{subfigure}%
\caption[Initial state of the metamodel generation for $\mathcal{M}_{Currin,HF}^{2d}$.]{Initial state of the metamodel generation for $\mathcal{M}_{Currin,HF}^{2d}$. (a) Initial dataset: 10 high-fidelity (black) samples and 30 low-fidelity samples (red). Two low-fidelity samples are in the same position has high-fidelity ones. (b) Initial absolute error over the domain.}\label{fig:CurrinInitialSamples}
\end{figure}

Next consider the two-dimensional Currin function first employed by \cite{currin1988bayesian}. The high-fidelity version reads
\begin{equation}
\mathcal{M}_{Currin,HF}^{2d}(x_{1}, x_{2}) = \left[ 1 - \exp \left( -\frac{1}{2 x_{2}} \right) \right] \frac{2300 x_{1}^{3} + 1900 x_{1}^{2} + 2092 x_{1} + 60}{100 x_{1}^{3} + 500 x_{1}^{2} + 4 x_{1} + 20}.
\end{equation}
The low-fidelity version is given by
\begin{equation}
\begin{aligned}
\mathcal{M}_{Currin,LF}^{2d}(x_{1}, x_{2}) &= 0.25  \mathcal{M}_{HF}(x_{1}+0.05, x_{2}+0.05) \\&+ 0.25 \mathcal{M}_{HF} (x_{1} + 0.05, \max(0, x_{2}-0.05))\\
&+ 0.25 \mathcal{M}_{HF}(x_{1}-0.05, x_{2}+0.05) \\&+ 0.25  \mathcal{M}_{HF} (x_{1} - 0.05, \max(0, x_{2}-0.05)).
\end{aligned}
\end{equation}
The function is evaluated on the input domain $]0,1]^{2}$. $\mathcal{M}_{Currin,HF}^{2d}$ and $\mathcal{M}_{Currin,LF}^{2d}$ are displayed in Figure \ref{fig::CurrinHF}. It can be seen that the low-fidelity version is a damped version of the HF function. 
The error measures computed from 10000 points between the two fidelities yield MAE:0.1140, RMAE :0.3661 and RMSE: 0.2151. It can therefore be seen that the LF variant distorts the real HF-function somewhat.
Here it is assumed again that the LF-variant needs significantly less computation time.
All of the following computations are done starting with 10 HF samples and 30 LF samples created with TPLHD. The positions of the points are displayed in Figure \ref{fig:CurrinInitial} with an overlaid contour plot of the HF-function. The black dots represent the HF points whereas the LF samples are indicated by the red dots. It can be seen that there are no HF samples around the global optimum. Furthermore the boundary values are not covered by neither HF- nor LF samples. The adaptive sampling techniques need to generate samples in these two problematic areas in order to create proficient surrogate models. This is also visible from the absolute error of the initial metamodel over the domain as displayed in Figure \ref{fig:CurrinInitialError}. Especially the uncovered corner around $(0,1.0)$ yields a high error measure.
A comparison of the error measure of the created metamodels after 20 HF Samples is listed in Table \ref{tab::CurrinTable1}. It can be seen that out of the adaptive sampling techniques only 6 perform better (RMSE error) than if the samples are created by the one-shot TPLHD method. The most proficient method is CVD realizing the best error measures in all 4 categories. However MEPE and EIGF do not fair much worse over all. SFCVT runs into numerical problems with clustering early on in the computation.    
\begin{table}[ht!]
\begin{center}
\resizebox{1.0\textwidth}{!}{%
\begin{tabular}{l|l c c c c} \hline
&Method & MAE & RMAE & RMSE & R$^{2}$ \\ \hline\hline \\
\multirow{1}{*}{\shortstack[l]{Errors after\\10 HF samples}} & TPLHD & 0.2230 & 1.2891 & 0.4630 & 0.9696  \\ \\  \hline \\
\multirow{14}{*}{\shortstack[l]{ Errors after\\20 HF samples}} &TPLHD &  0.1280 & 1.2320 & 0.3330 & 0.9843 \\
&ACE &  0.1758 & 1.4649 & 0.3970 & 0.9775  \\
&AME & 0.1807 & 0.5519 & 0.3048 & 0.9867  \\ 
&CVD & \textbf{0.1004} &\textbf{ 0.2372} & \textbf{0.1406} & \textbf{0.9971}  \\
&CVVor &  0.1811 & 1.6343 & 0.4420 & 0.9721 \\ 
&EI &  0.2698 & 0.9870 & 0.4731 & 0.9680  \\
&EIGF & 0.1103 & 0.3359 & 0.1685 & 0.9959 \\  
&LOLA &  0.2115 & 1.3452 & 0.4775 & 0.9674  \\
&MASA &  0.1953 & 1.4147 & 0.4657 & 0.9690  \\
&MEPE & 0.1104 & 0.3209 & 0.1685 & 0.9959  \\
&MIPT &  0.1686 & 0.5419 & 0.3014 & 0.9870  \\
&MSD & 0.1486 & 0.7030 & 0.2859 & 0.9883  \\
&SFCVT &  - & -& -&  -\\
&SSA &  0.2402 & 0.9174 & 0.4882 & 0.9660 \\ 
\end{tabular}
}
\end{center}
\caption[Error measures for $\mathcal{M}_{Currin,HF}^{2d}$ after 20 HF-samples]{Error measures for $\mathcal{M}_{Currin,HF}^{2d}$ after 20 HF samples and 30 LF samples (methods with clustering problems are indicated by empty rows).}\label{tab::CurrinTable1}
\end{table}
At a next step the convergence behaviors of the adaptive sampling methods are studied. The goal is to decrease the MAE error below a threshold value of $0.1$. The average numbers of samples (over 10 iterations) needed to reach this limit are listed in Table \ref{tab::CurrinTableLimit}.
In addition to a HK method starting with 10 HF samples and 30 LF samples an OK technique is employed starting with 10 samples generated by TPLHD to compare the effectiveness of HK.
It can be seen that in addition to SFCVT, EI is not able to reach this threshold and is therefore ommited. Only focusing on HK at first, the data indicates that CVD requires the least amount of samples on average with 21 $\pm$ 1. The $\pm$-symbol illustrates the variation. CVVor needs the highest number of samples with 53. \\ \newpage When comparing these values to the average amount of samples needed for the adaptive methods in OK to reach this limit, it is noticeable that except for the SSA technique OK requires more samples to reach the accuracy target. \\
Looking closer at the CVD technique, which represents the best value for HK, when utilizing this method an algorithm with OK needs 18 HF samples and therefore 18 HF function evaluations more. Hence, HK is efficient if one LF-evaluation (with 30 initial samples) needs around half of the computational time in comparison to HK. 
\begin{table}[ht!]
\begin{center}
\begin{tabular}{l| c c} \hline
Sampling method & Number samples HK & Number Samples OK\\ \hline\hline \\
ACE &  30  $\pm$ 0 & 97 $\pm$ 2 \\
AME & 31 $\pm$ 0 & 34 $\pm$ 0 \\ 
CVD & 21 $\pm$ 1 &  39 $\pm$ 1\\
CVVor &  53 $\pm$ 2 & 75 $\pm$ 3 \\ 
EI  &  - & - \\
EIGF & 27 $\pm$ 1 &  60 $\pm$ 1 \\
LOLA &  38  $\pm$ 2 & -   \\
MASA &  49 $\pm$ 1 &   69 $\pm$ 2\\
MEPE & 27  $\pm$ 0 & 37  $\pm$ 0   \\
MIPT &  28 $\pm$ 0 & 37 $\pm$ 0 \\
MSD & 37   $\pm$ 2 & 45 $\pm$ 1  \\
SFCVT  & - &  -\\
SSA &  36  $\pm$ 2 & 34 $\pm$ 2  
\end{tabular}
\end{center}
\caption[Comparison of number of samples between OK and HK for $\mathcal{M}_{Currin,HF}^{2d}$]{Average amount of samples to get a value of MAE below $0.1$ for  $\mathcal{M}_{Currin,HF}^{2d}$. Computation until 55 H -samples. Comparison to the average amount of samples needed for OK until 100 samples (methods with clustering problems are indicated by empty rows).}\label{tab::CurrinTableLimit}
\end{table}

\begin{figure}[hbtp]
\centering
\includegraphics[scale=0.6]{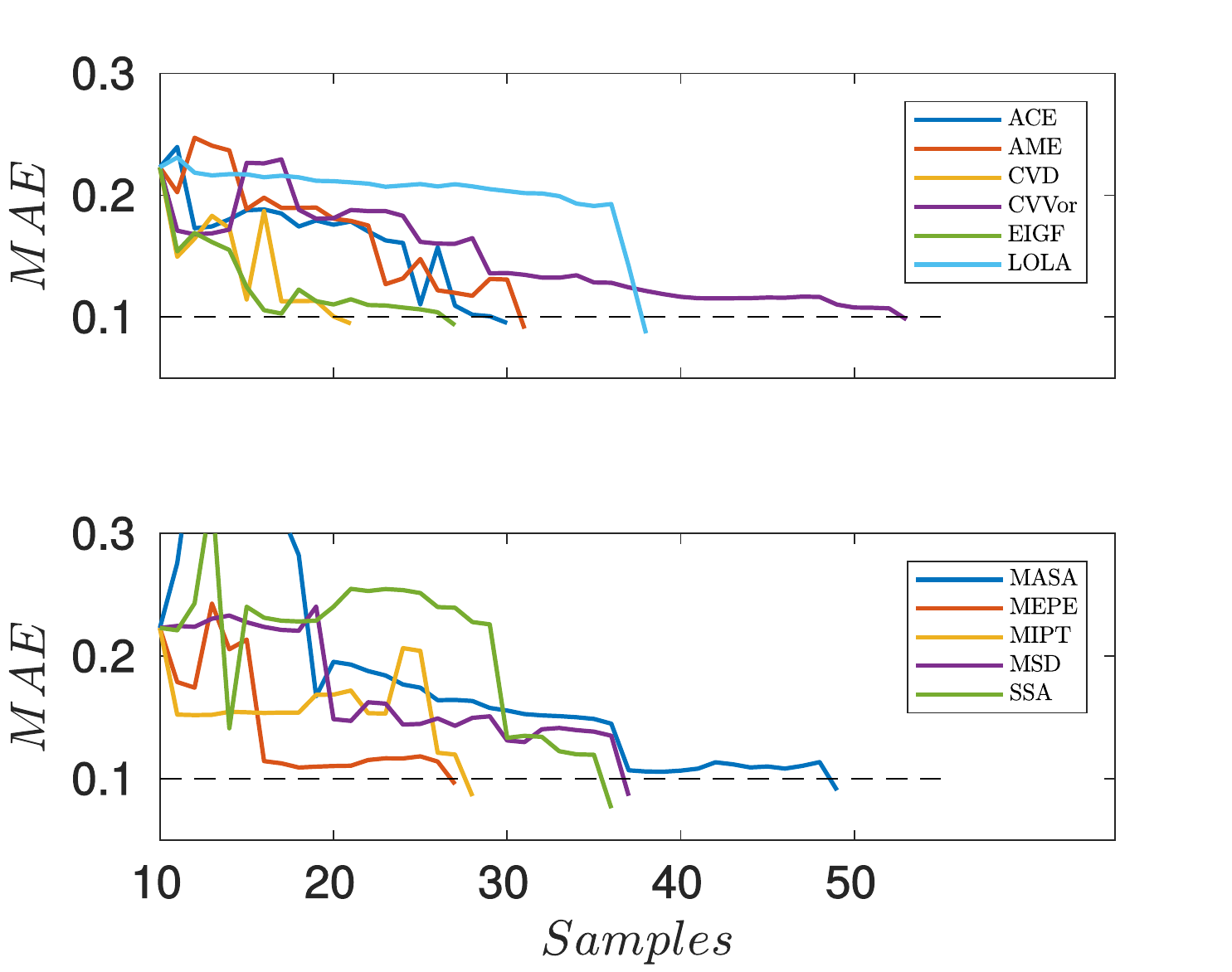}
\caption[Convergence of MAE error for $\mathcal{M}_{Currin,HF}^{2d}$]{Convergence of MAE error for $\mathcal{M}_{Currin,HF}^{2d}$. MAE threshold is illustrated with dotted-line.}\label{fig::DegreefConvergenceCurrin}
\end{figure}

The degree of convergence of the studied techniques is displayed in Figure \ref{fig::DegreefConvergenceCurrin}. Here, the MAE threshold is represented by the dotted-line.

\begin{figure}[h!]
\centering
\begin{subfigure}[t]{0.3\textwidth}
\includegraphics[scale=0.28]{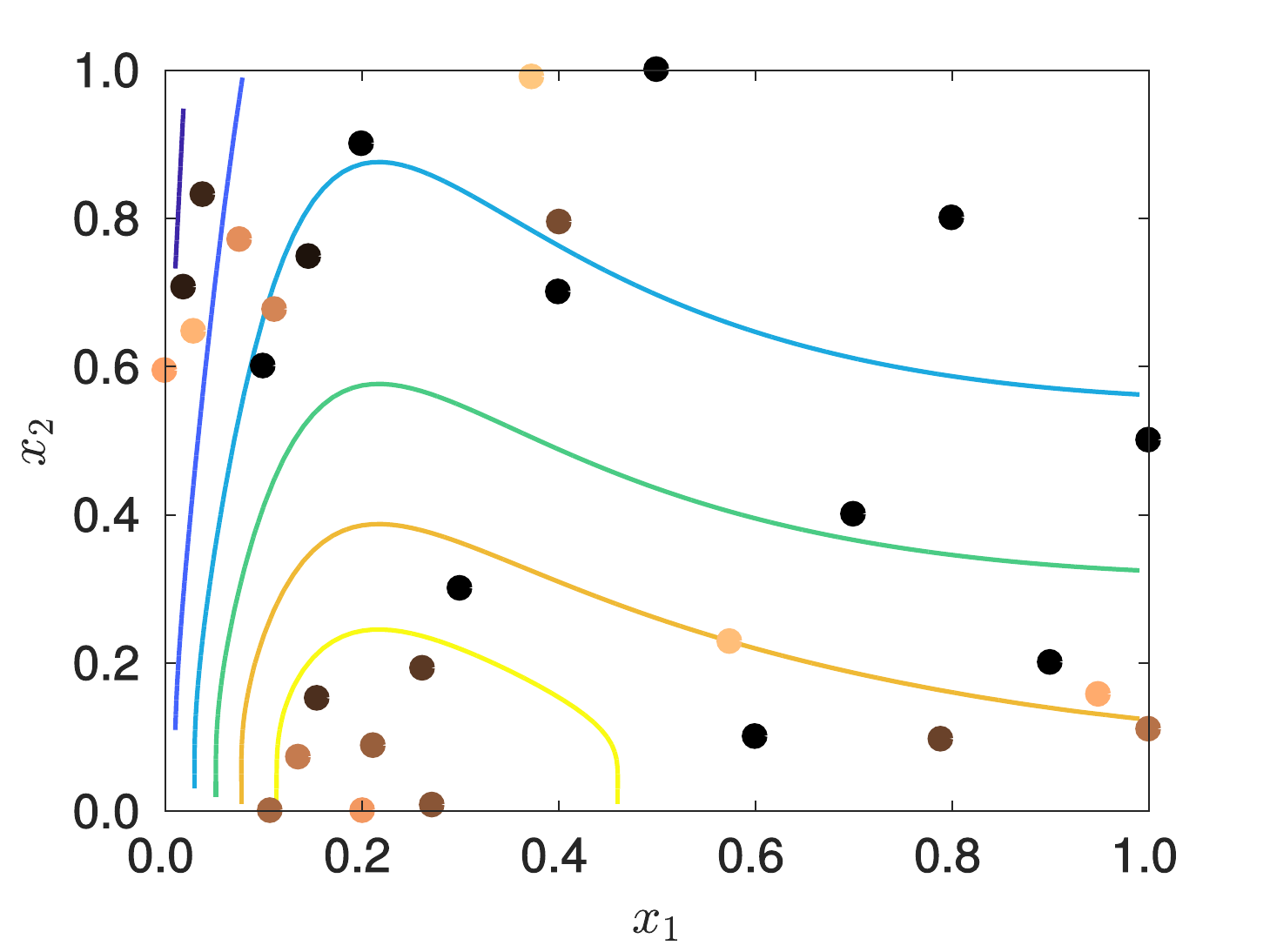}
\subcaption{ACE - 30 samples}\label{fig:CurrinYesACE}
\end{subfigure}%
\begin{subfigure}[t]{0.3\textwidth}
\includegraphics[scale=0.28]{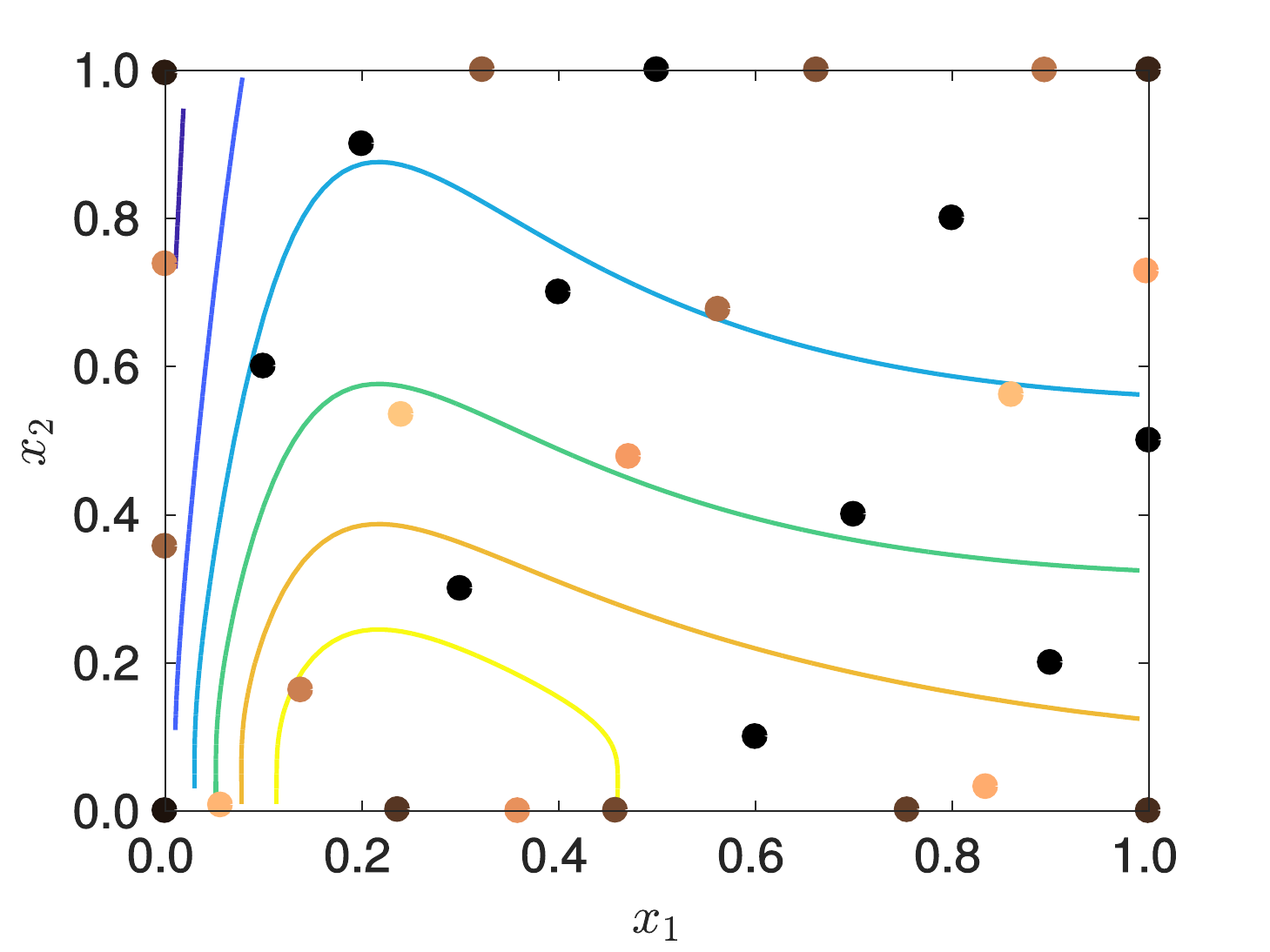}
\subcaption{AME - 31 samples}\label{fig:CurrinYesAME}
\end{subfigure}%
\begin{subfigure}[t]{0.3\textwidth}
\includegraphics[scale=0.28]{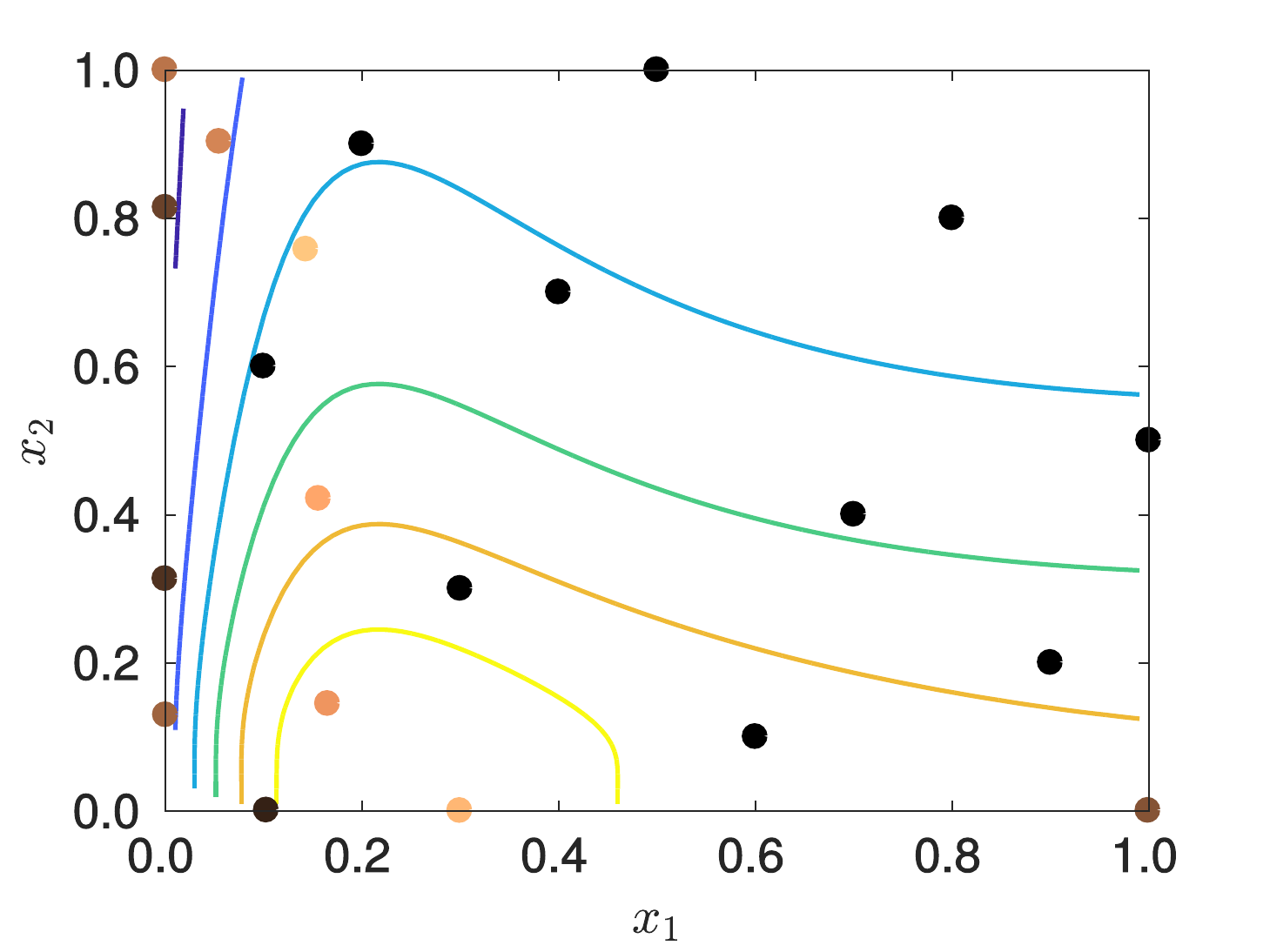}
\subcaption{CVD - 21 samples}\label{fig:CurrinYesMSE}
\end{subfigure}
\begin{subfigure}[t]{0.3\textwidth}
\includegraphics[scale=0.28]{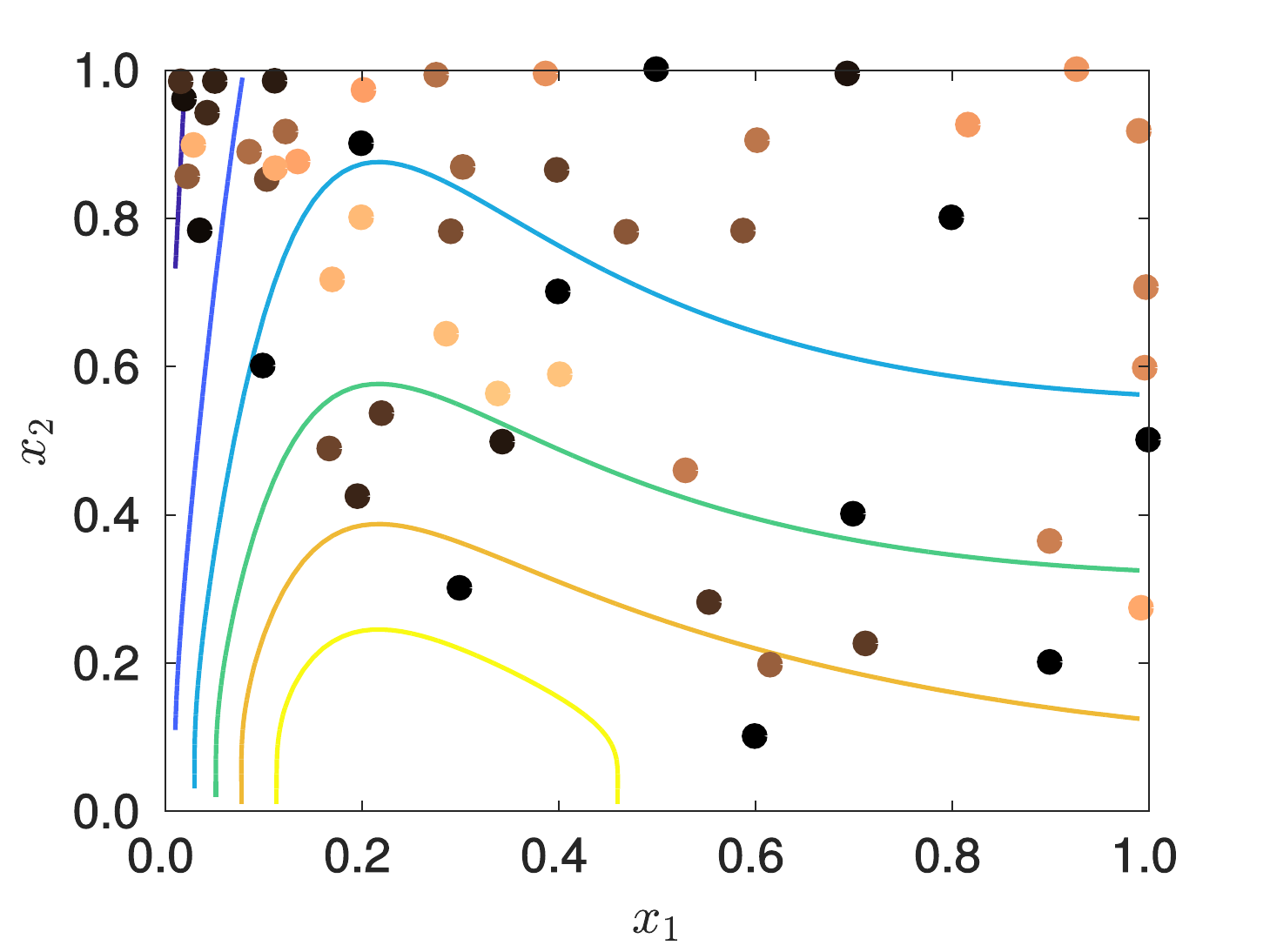}
\subcaption{CVVor - 53 samples}\label{fig:CurrinYesCVVor}
\end{subfigure}%
\begin{subfigure}[t]{0.3\textwidth}
\includegraphics[scale=0.28]{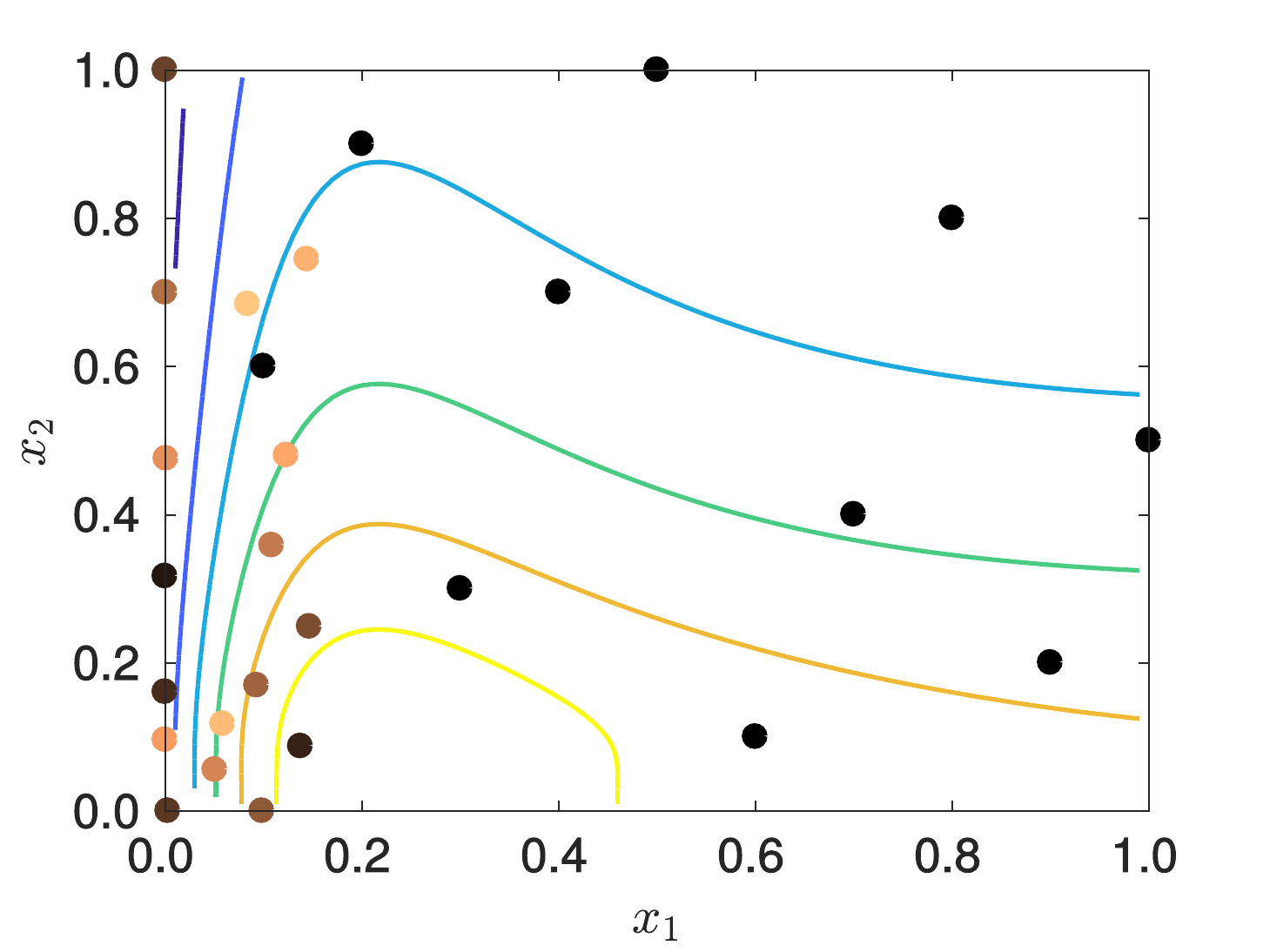}
\subcaption{EIGF - 27 samples}\label{fig:CurrinYesEIGF}
\end{subfigure}%
\begin{subfigure}[t]{0.3\textwidth}
\includegraphics[scale=0.28]{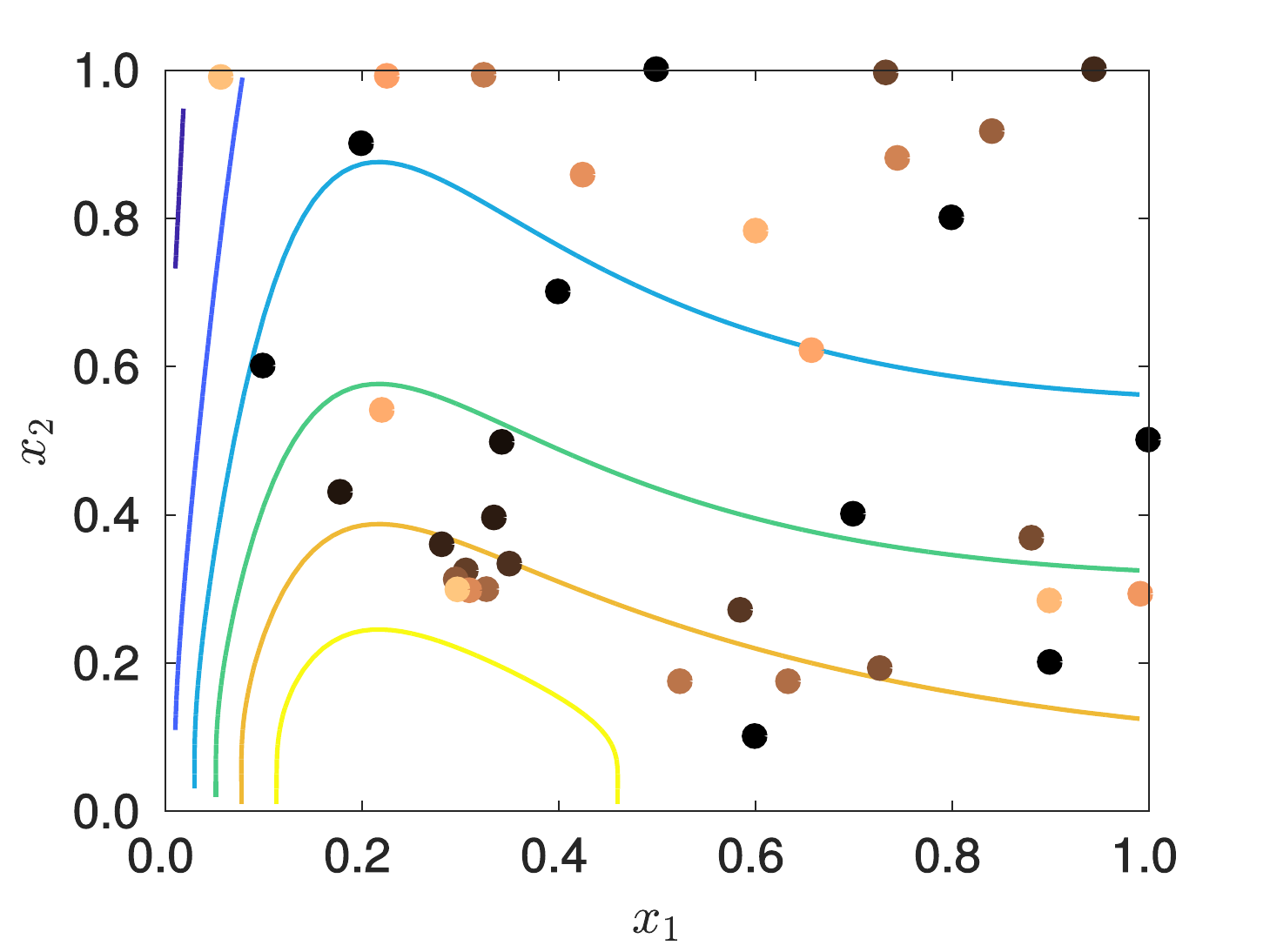}
\subcaption{LOLA - 38 samples}\label{fig:CurrinYesLOLA}
\end{subfigure}
\begin{subfigure}[t]{0.3\textwidth}
\includegraphics[scale=0.28]{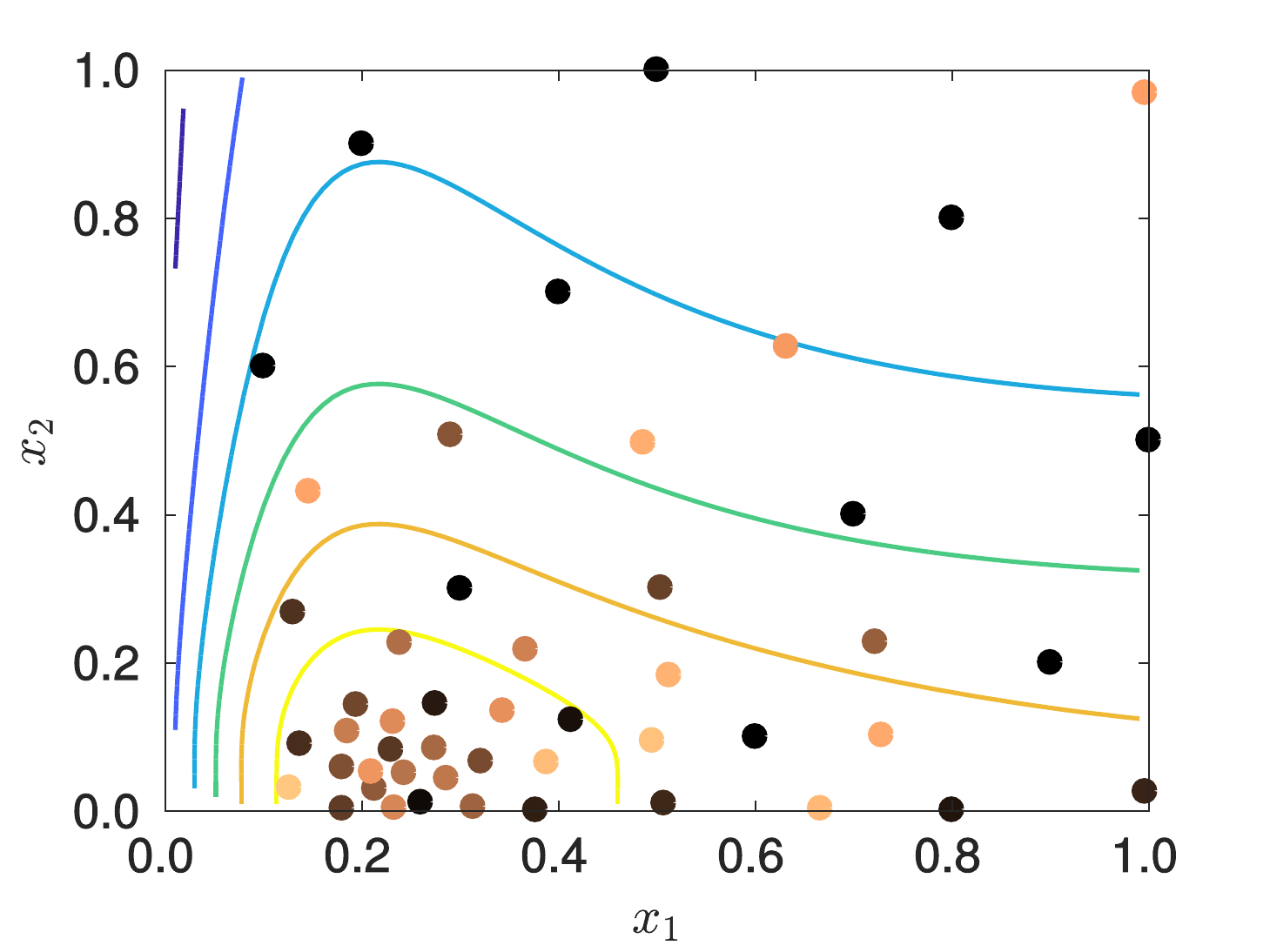}
\subcaption{MASA - 49 samples}\label{fig:CurrinYesMASA}
\end{subfigure}%
\begin{subfigure}[t]{0.3\textwidth}
\includegraphics[scale=0.28]{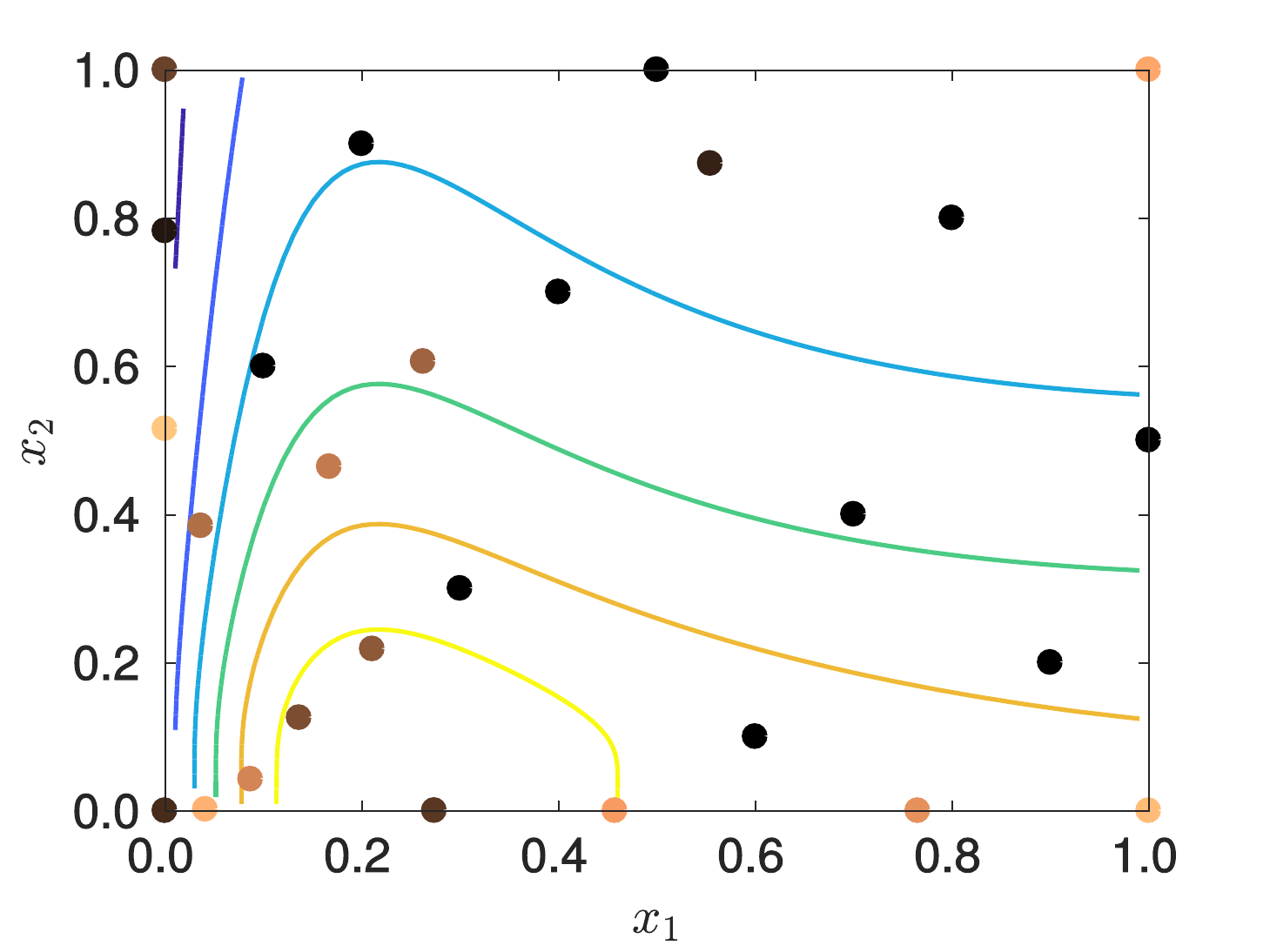}
\subcaption{MEPE - 27 samples}\label{fig:CurrinYesMEPE}
\end{subfigure}%
\begin{subfigure}[t]{0.3\textwidth}
\includegraphics[scale=0.28]{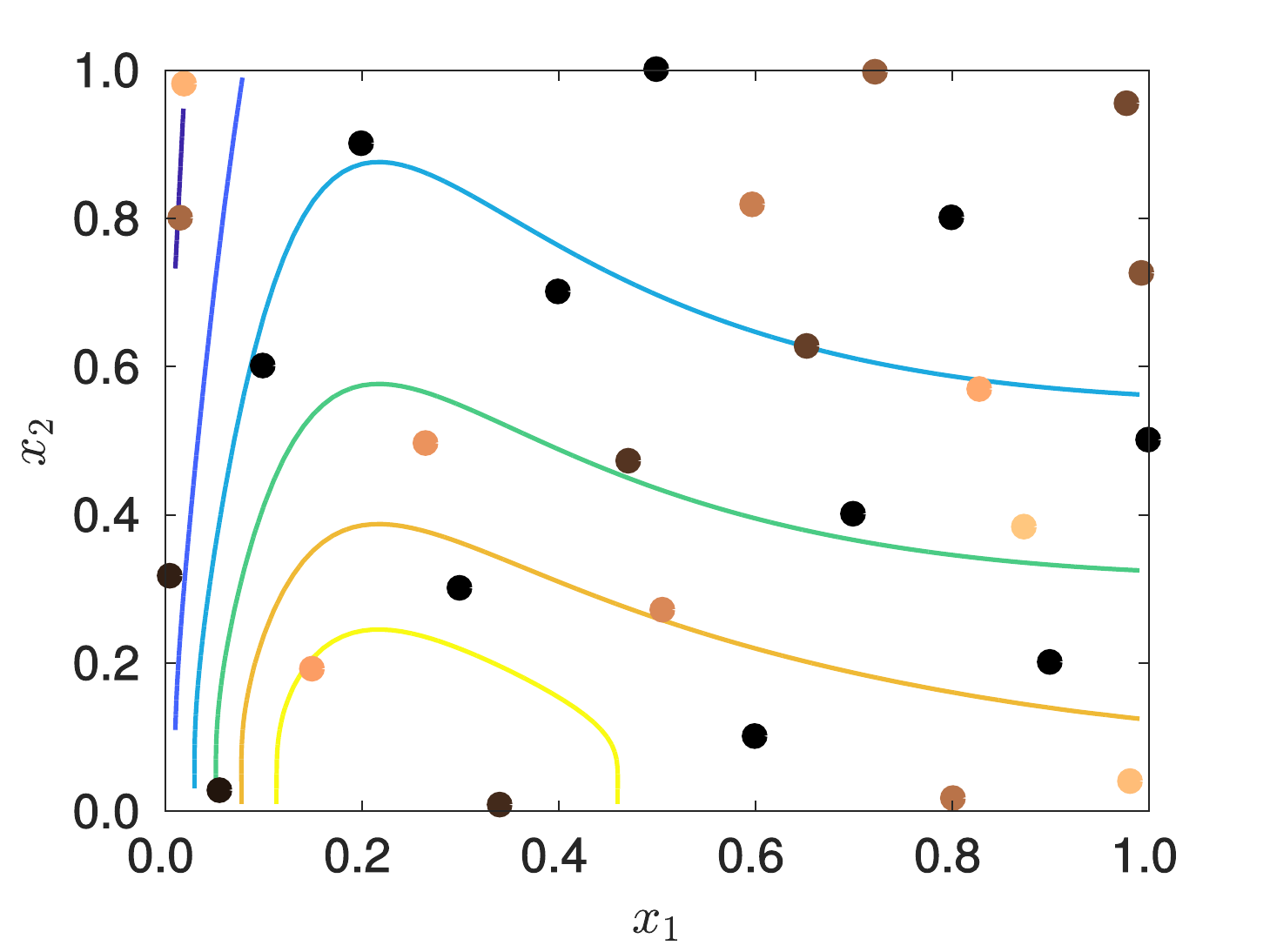}
\subcaption{MIPT - 28 samples}\label{fig:CurrinYesMIPT}
\end{subfigure}
\begin{subfigure}[t]{0.3\textwidth}
\includegraphics[scale=0.28]{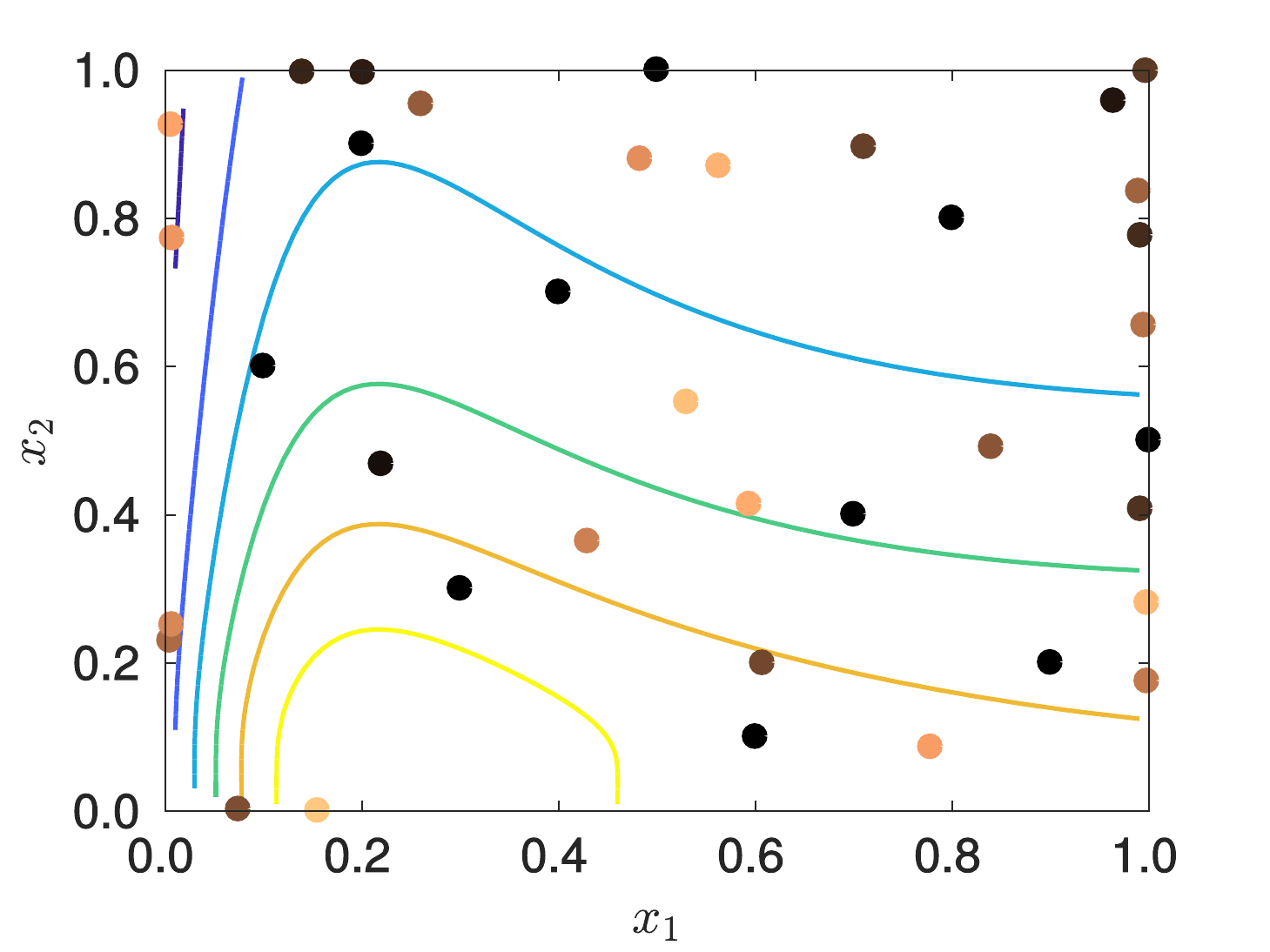}
\subcaption{MSD - 37 samples}\label{fig:CurrinYesMSD}
\end{subfigure}%
\begin{subfigure}[t]{0.3\textwidth}
\includegraphics[scale=0.28]{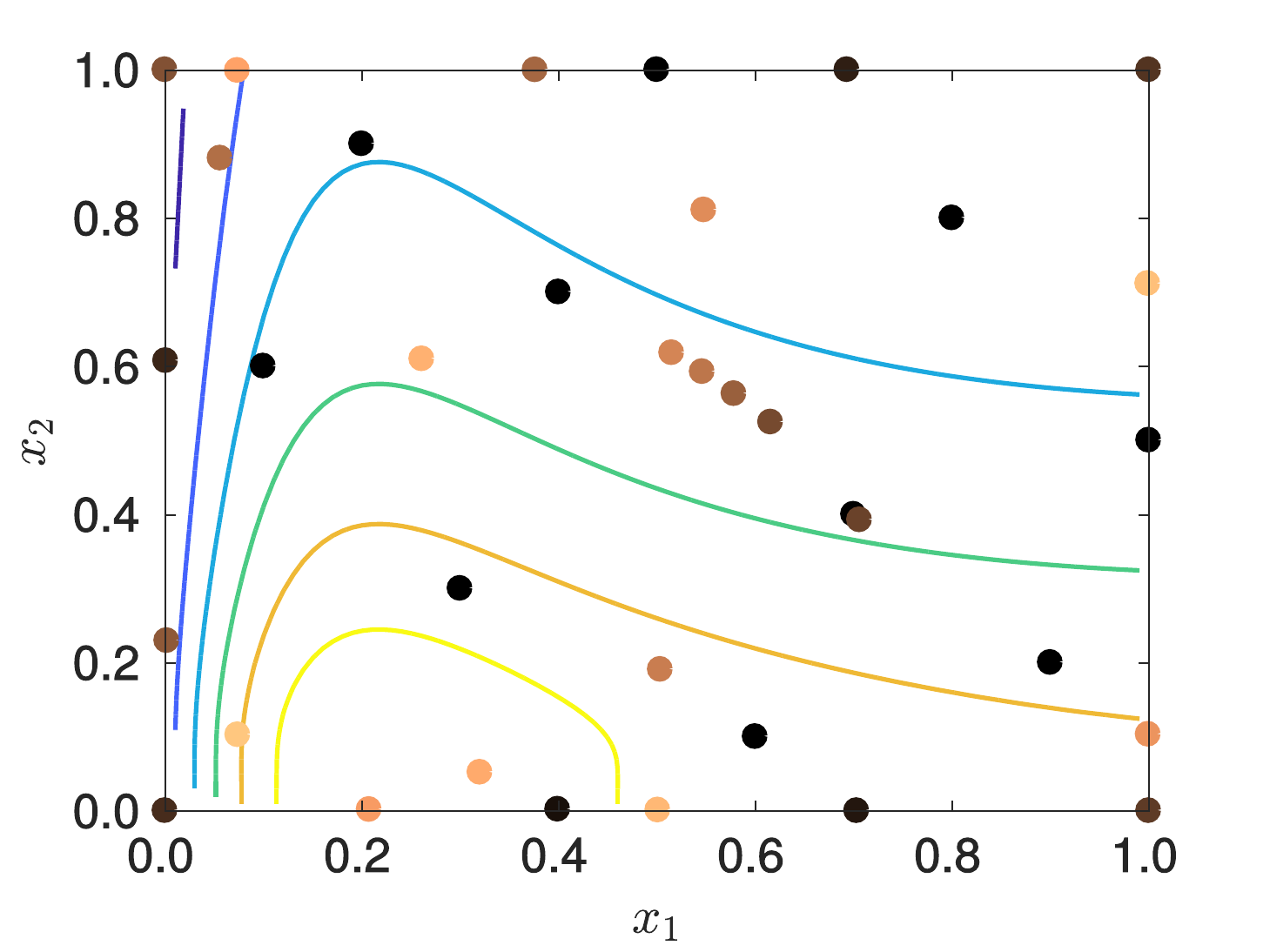}
\subcaption{SSA - 36 samples}\label{fig:CurrinYesSSA}
\end{subfigure}%
\caption[Location of samples for $\mathcal{M}_{Currin,HF}^{2d}$ of techniques reaching threshold]{Location of samples for $\mathcal{M}_{Currin,HF}^{2d}$ of techniques able to reach a MAE-value of below $0.1$ until reaching 55 HF samples.}%
\label{fig:CurrinYes}%
\end{figure}

In the next step the locations of the generated samples that are needed to reach the MAE limit are investigated. The respective plots for each technique are shown in Figure \ref{fig:CurrinYes}. Here in the background the contour of the HF Currin function is plotted. Furthermore the later a sample is added the more the color of the points in the plots tend towards light red. Hence, the initial HF-samples appear in black.
The sample locations for ACE are shown in Figure \ref{fig:CurrinYesACE}. The technique reaches the MAE limit after 30 sample points.
\newpage
It can be seen that early samples are added in the domain where initially a high absolute error was present (c.f. Figure \ref{fig:CurrinInitialError}). Later the samples also focus on the optimum area. It can be seen that ACE has a good balance between exploration and exploitation for this application in HK. \\
Figure \ref{fig:CurrinYesAME} displays the 31 samples of AME. The approach samples the domain with evenly spread points. All edges are well sampled which allows to lower the MAE error below the threshold. \\
The most proficient technique for this application is CVD (Figure \ref{fig:CurrinYesMSE}). It can be seen that the whole focus lies on the left side of the domain where the optimum and the area with the initially high MAE error was located. Here, the method shows that it balances between exploitation and exploration. \\
CVVor (Figure \ref{fig:CurrinYesCVVor}) requires 53 sample points. The algorithm faces difficulties due to its lack of exploration in this application. The area around the optimum is not sampled at all. A strong focus lies here on the area with the initially high prediction error. This leads the algorithm to be the worst performing method. 

\begin{figure}
\centering
\begin{subfigure}[t]{0.5\textwidth}
\includegraphics[scale=0.4]{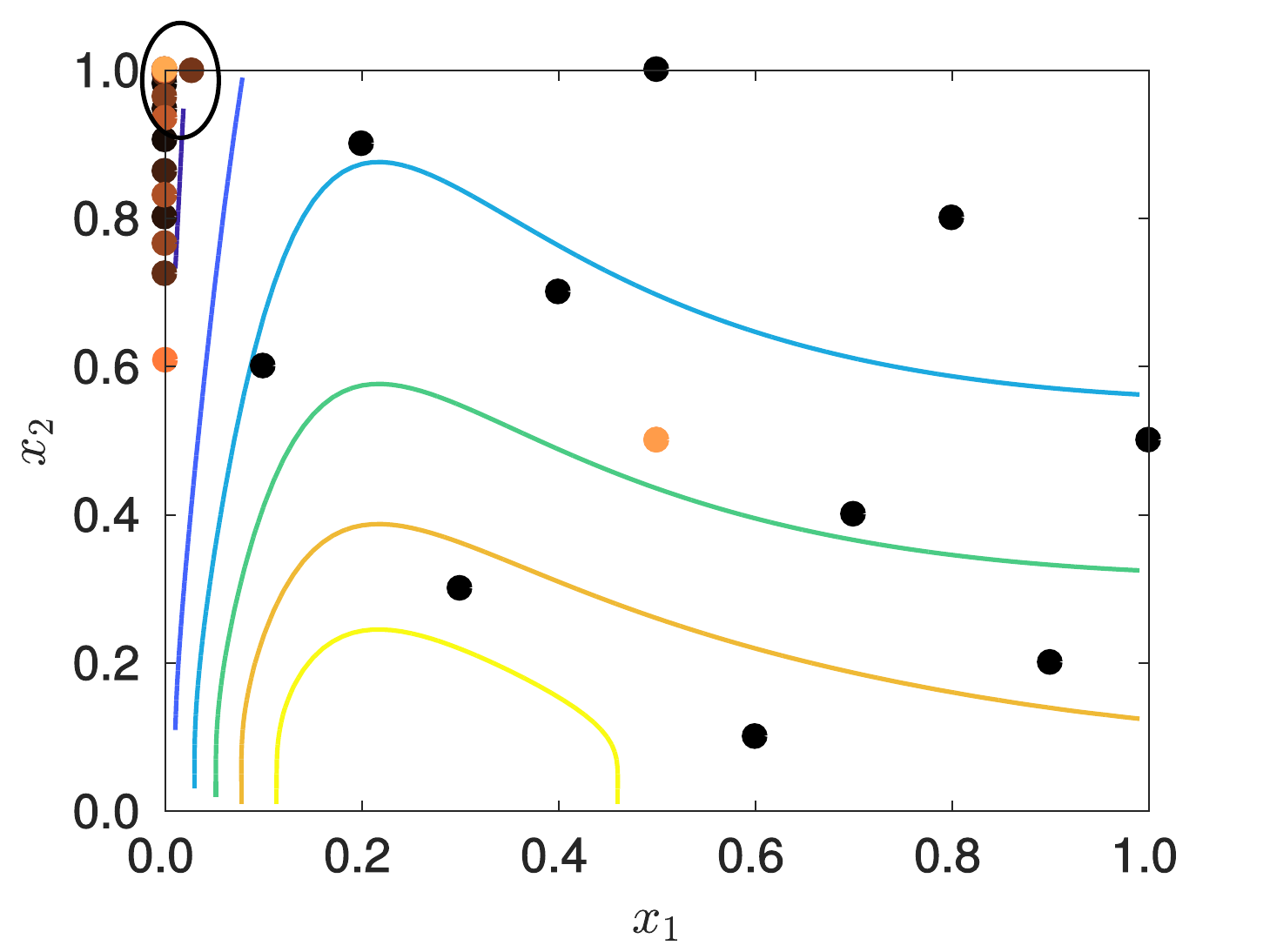}
\subcaption{EI - not able to reach limit}\label{fig:CurrinNoEI}
\end{subfigure}%
\begin{subfigure}[t]{0.5\textwidth}
\includegraphics[scale=0.4]{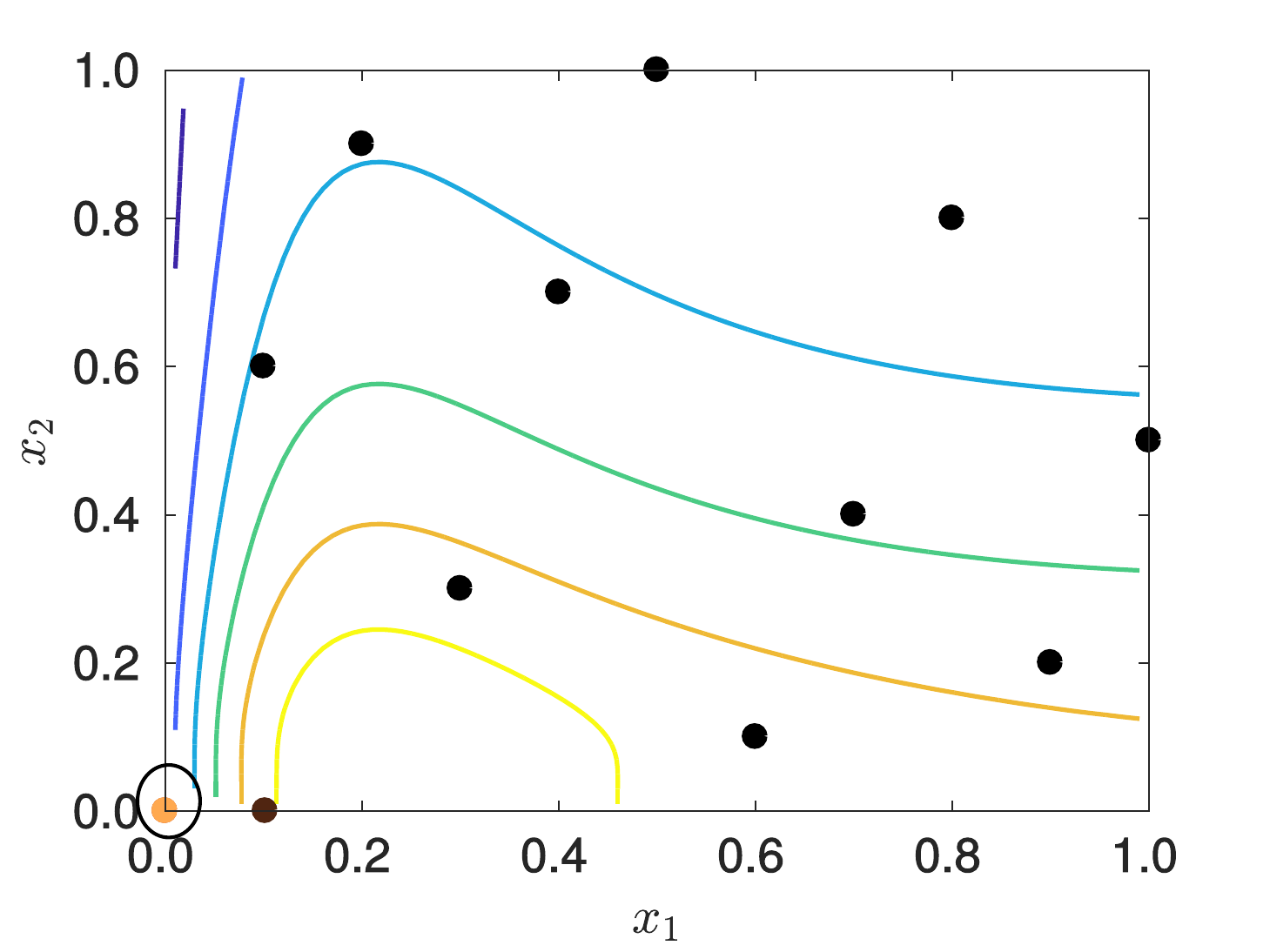}
\subcaption{SFCVT - not able to reach limit}\label{fig:CurrinNoSFVCT}
\end{subfigure}%
\caption[Location of samples for $\mathcal{M}_{Currin,HF}^{2d}$ of techniques unable to reach threshold]{Location of samples for $\mathcal{M}_{Currin,HF}^{2d}$ of techniques unable to reach error threshold because of clustering or needing more than 55 HF samples.}%
\label{fig:CurrinNo}%
\end{figure}

The EIGF scheme is displayed in Figure \ref{fig:CurrinYesEIGF}. It is visible that only the left side of the domain is sampled. This fits with the initial prediction error. The exploitation character is neglected. However it works for this application. \\
LOLA (Figure \ref{fig:CurrinYesLOLA}) manages to reduce the error below the threshold without sampling in the area where the initial metamodel had the highest prediction error. 
\newpage
Furthermore the optimum value is not covered because the algorithm bases its exploitation on the gradients of the existing sample points. Therefore the method wastes a lot of samples in unnecessary areas. \\
The samples of the MASA scheme are plotted in Figure \ref{fig:CurrinYesMASA}. It gets proven again that MASA focuses on the maximum absolute value of the target function, but without the occurrence of clustering due to a constraint.\\However the initial area of high prediction error is not sampled at all. Here, the scheme uses too many samples unnecessarily. \\
MEPE (Figure \ref{fig:CurrinYesMEPE}) needs 28 samples points. The method has a good exploration component, which provides samples at all corners of the domain. Furthermore the exploitation character adds samples in the left-hand side of the domain. \\
The exploration-based character of MIPT which evenly spaces the points is depicted in Figure \ref{fig:CurrinYesMIPT}. This procedure is very effective for the given application.\\
MSD (Figure \ref{fig:CurrinYesMSD}) covers the domain evenly but not as effective as MIPT and hence requires more sample points. \\
SSA (Figure \ref{fig:CurrinYesSSA}) shows its exploration character by sampling all corners. However the exploitation is not proficient enough to lie the main focus on the left-hand side of the domain. Hence 36 sample points are needed. \\
The two methods that were unable to reach the MAE target value are displayed in Figure \ref{fig:CurrinNo}. As indicated earlier EI (Figure \ref{fig:CurrinNoEI}) has problems with clustering around the minimum values of the target function, an issue which is also apparent here. Furthermore SFCVT suffers repeatedly from the fact that it places multiple samples on the point $(0,0)$ as highlighted within the plot.
\clearpage
\subsection{Four-dimensional Park function}
\begin{figure}[h!]
\centering
\begin{subfigure}[t]{0.5\textwidth}
\includegraphics[scale=0.35]{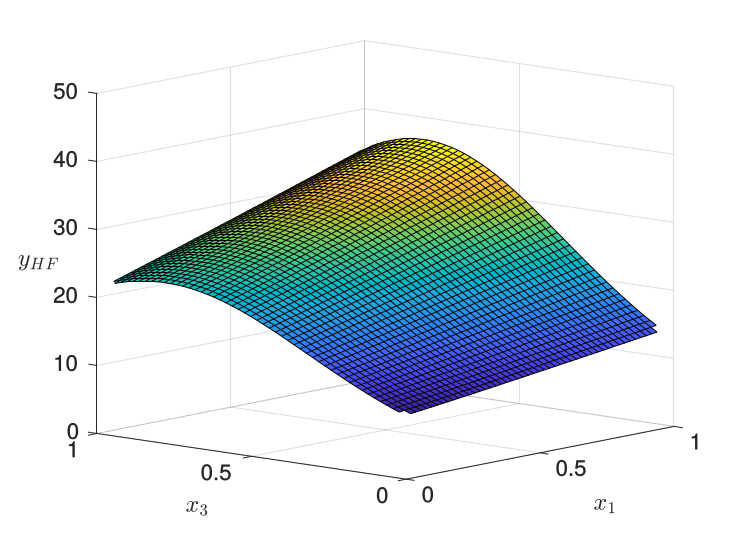} 
\subcaption{HF function}\label{fig:ParkHF}
\end{subfigure}%
\begin{subfigure}[t]{0.5\textwidth}
\includegraphics[scale=0.35]{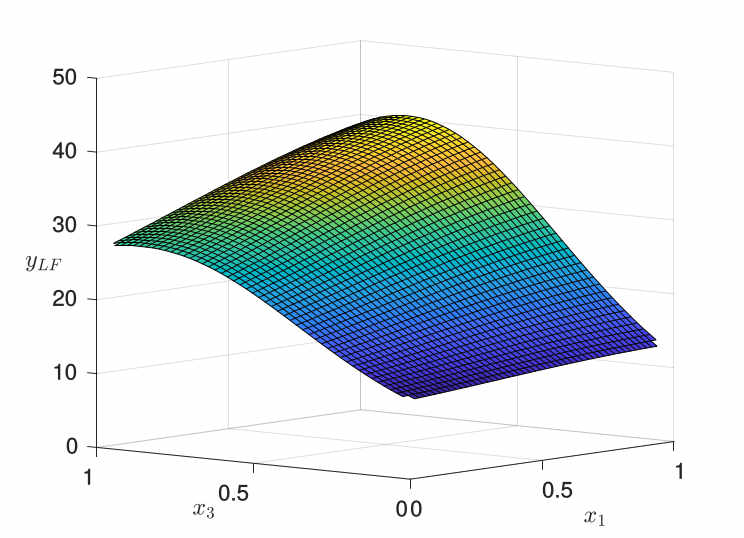} 
\subcaption{LF function}\label{fig:ParkLF}
\end{subfigure}%
\caption[Two-dimensional plots of the HF and LF functions of the Park function]{Two-dimensional plots of the HF and LF functions of the Park function. Here, $x_{1}$ and $x_{3}$ are plotted over their input domain while the other two inputs are set to unity respectively.}%
\label{fig:Park2dPlot}%
\end{figure}

Next consider the four-dimensional Park function first employed by \cite{park1991tuning} and later e.g. utilized in \cite{xiong2013sequential}. The high-fidelity version reads
\begin{equation}
\mathcal{M}_{Park,HF}^{4d}(\bm{x}) = \frac{x_{1}}{2} \left[ \sqrt{1 + (x_{2} + x_{3}^{2}) \frac{x_{4}}{x_{1}^{2}}} - 1 \right] + (x_{1} + 3 x_{4}) + \exp \left[ 1 + \sin (x_{3})\right].
\end{equation}
The low accuracy version is given by
\begin{equation}
\mathcal{M}_{Park, LF}^{4d}(\bm{x}) = \left[ 1 + \frac{\sin(x_{1})}{10} \right] \mathcal{M}_{HF}(\bm{x}) - 2 x_{1} + x_{2}^{2} + x_{3}^{2} + 0.5 .
\end{equation}
The input domain is chosen to be $x_{i} \in \,  ]0,1] \forall \, i=1,2,3,4$. To give an idea for the error measurements the spread (absolute difference between the minimum and maximum) is around 36. To visualize the complexness of the function consider $x_{2}=x_{3}=1.0$. The HF and LF functions for this scenario over the input domain are displayed in Figure \ref{fig:Park2dPlot}. 

\begin{figure}[b!]
\centering
\includegraphics[scale=0.5]{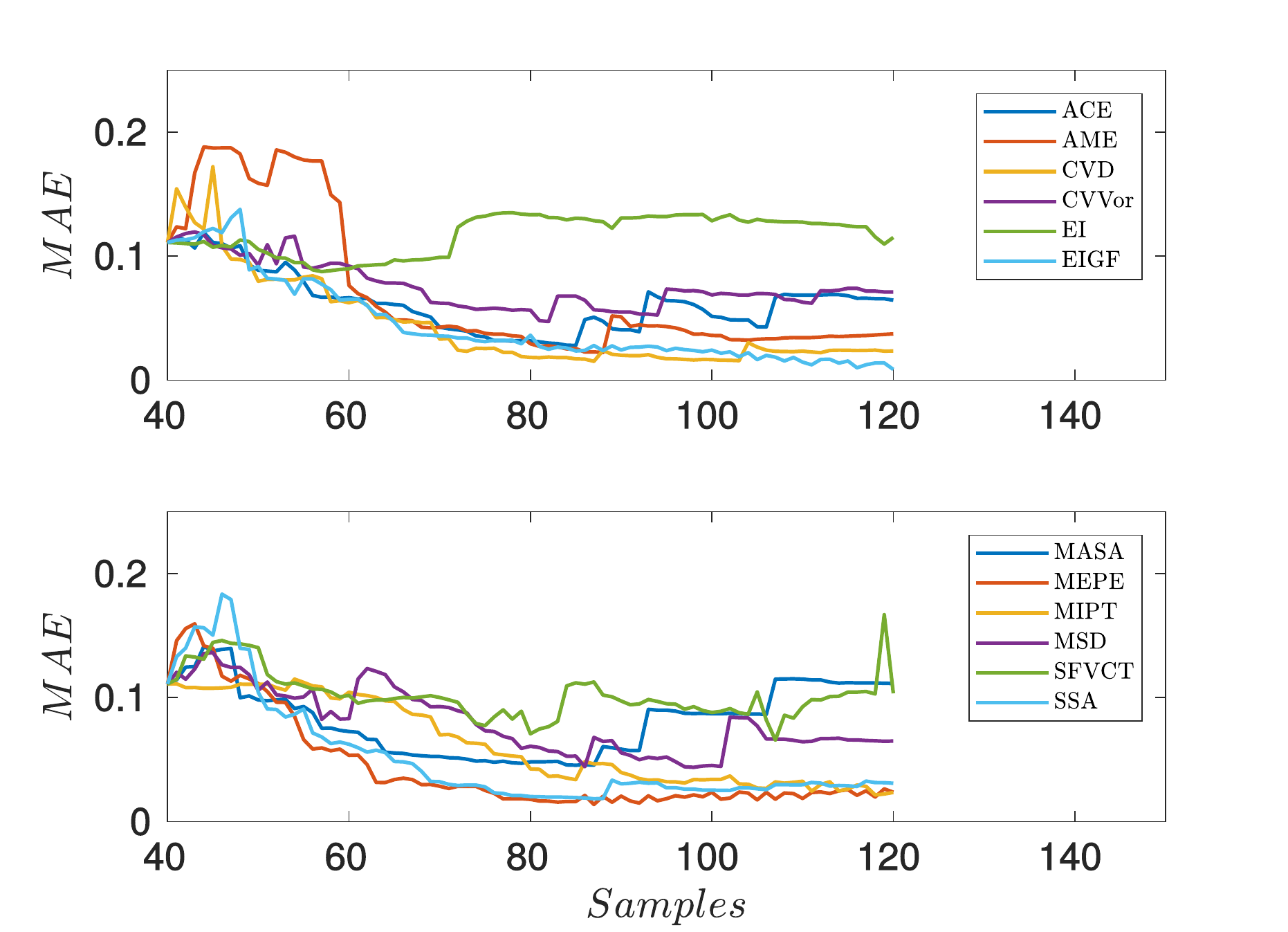} 
\caption[Convergence of MAE error for $\mathcal{M}_{Park,HF}^{4d}$]{Convergence of MAE error for $\mathcal{M}_{Park,HF}^{4d}$ for until 120 HF-samples.}\label{fig::ParkMAE}
\end{figure}

\begin{figure}[b!]
\centering
\includegraphics[scale=0.5]{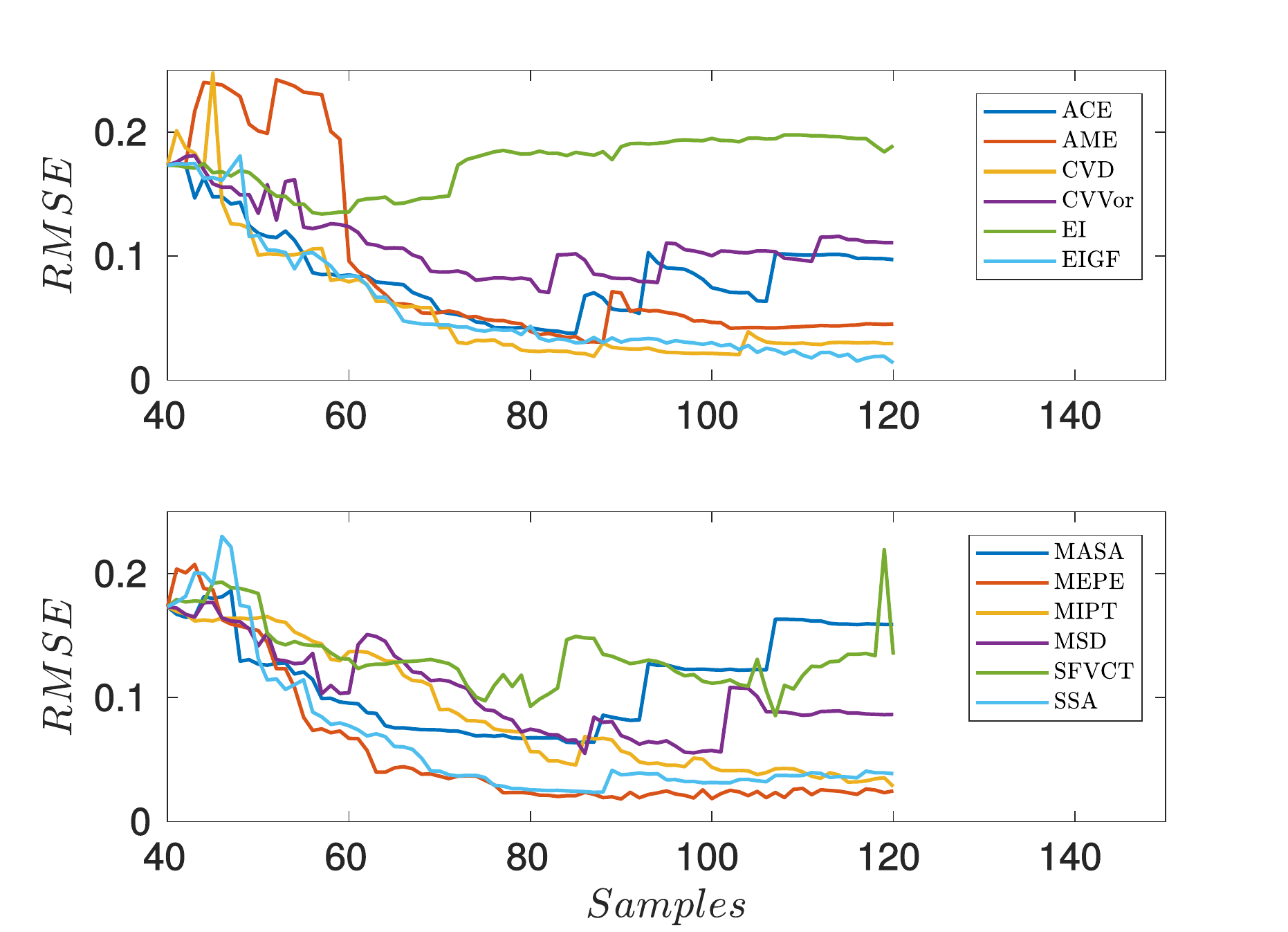} 
\caption[Convergence of RMSE error for $\mathcal{M}_{Park,HF}^{4d}$]{Convergence of RMSE error for $\mathcal{M}_{Park,HF}^{4d}$ for until 120 HF-samples.}\label{fig::ParkRMSE}
\end{figure}

The nonlinearity of the function can be seen. Furthermore the LF function is a stretched and shifted version of the higher-fidelity function. The error between the two fidelity versions yield MAE:$0.7368$, RMAE:	$0.4768$ and RMSE: $0.8947$. It can therefore be seen that the LF-function distorts the value of the HF-version.
The results of the adaptive sampling techniques and the TPLHD are compared in Table \ref{tab::Park80}. Here the initial HF-samples size is 40, which gets adaptively increased to 80. 250 LF-samples are generated with TPLHD. \newpage
It can be seen that all adaptive sampling techniques provide better results than the initial metamodel and the metamodel created with 80 HF-samples. For MAE the negative outlier is EI, which as already known tends to cluster around the global minimum due to the lack of exploration. Furthermore it can be seen that the space-filling approaches MIPT and CVVor yield the worst relative maximum absolute error value.
\begin{table}[t!]
\begin{center}
\resizebox{1.0\textwidth}{!}{%
\begin{tabular}{l|l c c c c} \hline
&Method & MAE & RMAE & RMSE & R$^{2}$ \\ \hline\hline \\
\multirow{1}{*}{\shortstack[l]{Errors after\\40 HF samples}} & TPLHD &  0.1110 & 0.3379 & 0.1733 & 0.9988   \\ \\  \hline \\
\multirow{14}{*}{\shortstack[l]{ Errors after\\80 HF samples}} &TPLHD &  0.1130 & 0.2116 & 0.1795 & 0.9988 \\
&ACE & 0.0318 & 0.0571 & 0.0422 & 0.9999  \\
AME & 0.0293 & 0.0325 & 0.0393 & 0.9999   \\
&CVVor & 0.0564 & 0.1100 & 0.0812 & 0.9997 \\ 
&EI &  0.1334 & 0.3230 & 0.1826 & 0.9987  \\
&EIGF & 0.0286 & 0.0249 & 0.0358 & 0.9999 \\
&MASA & 0.0482 & 0.0880 & 0.0676 & 0.9998   \\
&MEPE &  0.0179 & 0.0186 & \textbf{0.0229 }& 0.9999   \\
&MIPT & 0.0425 & 0.1118 & 0.0565 & 0.9998     \\ 
&MSD & 0.0608 & 0.0621 & 0.0745 & 0.9998   \\
&MSE & \textbf{0.0183} & 0.0194 & 0.0234 & 0.9999  \\
&SFCVT & 0.0708 & 0.1062 & 0.0930 & 0.9996  \\ 
&SSA &  0.0202 & \textbf{0.0192} & 0.0256 & 0.9999 
\end{tabular}
}
\end{center}
\caption[Error measures for $\mathcal{M}_{Park,HF}^{4d}$ after 80 HF samples]{Error measures for $\mathcal{M}_{Park,HF}^{2d}$ after 80 HF samples and 250 LF samples. }\label{tab::Park80}
\end{table}
The behavior of two error measure with increasing sample size is shown in Figures \ref{fig::ParkMAE} and \ref{fig::ParkRMSE}. In Figure \ref{fig::ParkMAE} the evolution of the MAE error over the sample size from 40 HF samples up to 120 samples is plotted. It can be seen that ACE, CVVor, EI, MASA, MSD and SFCVT have problems with the convergence and show flattening or in the case of SFCVT increases. This points towards numerical issues with sample points being generated too close to each other hence raising the condition number of the autocorrelation  matrix and distorting its inverse. The rest of the methods show a decline but also seem to flatten around respective values. However the MAE error at 120 samples e.g. for SSA is around 0.03. This proves the effectiveness of the adaptive scheme to create a proficient approximation for a nonlinear four-dimensional function with a spread of around 36. 
The RMSE error over the sample size is plotted in Figure \ref{fig::ParkRMSE}. It can be seen that the same behavior is observed as described for Figure \ref{fig::ParkMAE}, i.e. some adaptive schemes face convergence problems.
The final values at 120 samples are shown in Figure \ref{tab::Park}. As visible in the Figures, the best working method seems to be EIGF, which also reduces the RMAE value to the lowest value. However SSA and MEPE are not much worse. 
\begin{table}[h!]
\begin{center}
\resizebox{1.0\textwidth}{!}{%
\begin{tabular}{l|l c c c c} \hline
&Method & MAE & RMAE & RMSE & R$^{2}$ \\ \hline\hline \\
\multirow{1}{*}{\shortstack[l]{Errors after\\40 HF samples}} & TPLHD &  0.1110 & 0.3379 & 0.1733 & 0.9988   \\ \\  \hline \\
\multirow{14}{*}{\shortstack[l]{ Errors after\\120 HF samples}} &TPLHD &  0.0921 & 0.30747 & 0.1397 & 0.9992 \\
&ACE & 0.0645 & 0.2169 & 0.0972 & 0.9996  \\
&AME & 0.0373 & 0.0403 & 0.04538 & 0.9997  \\
&CVD & 0.0235 & 0.0290 & 0.0295 & 0.9999  \\
&CVVor & 0.0711 & 0.1902 & 0.1109 & 0.9995  \\ 
&EI &  0.1151 & 0.4502 & 0.1891 & 0.9986 \\
&EIGF & \textbf{0.0086 }& \textbf{0.0202} & \textbf{0.0139} & 0.9997  \\
&MASA & 0.1114 & 0.2342 & 0.1590 & 0.9990 \\
&MEPE &  0.0237 & 0.0361 & 0.0248 & 0.9998    \\
&MIPT & 0.0235 & 0.1018 & 0.0283 & 0.9998     \\ 
&MSD & 0.0650 & 0.1258 & 0.0864 & 0.9997    \\
&SFCVT & 0.1035 & 0.1802 & 0.1346 & 0.9993  \\ 
&SSA &  0.0309 & \textbf{0.0275} & 0.0386 & 0.9999 
\end{tabular}
}
\end{center}
\caption[Error measures for $\mathcal{M}_{Park,HF}^{4d}$ after 120 HF samples]{Error measures for $\mathcal{M}_{Park,HF}^{2d}$ after 120 HF samples and 250 LF samples. }\label{tab::Park}
\end{table}
\clearpage
\section{Finite-element application for Hierarchical Kriging}
\begin{figure}[hbtp]
\centering
\includegraphics[scale=0.6]{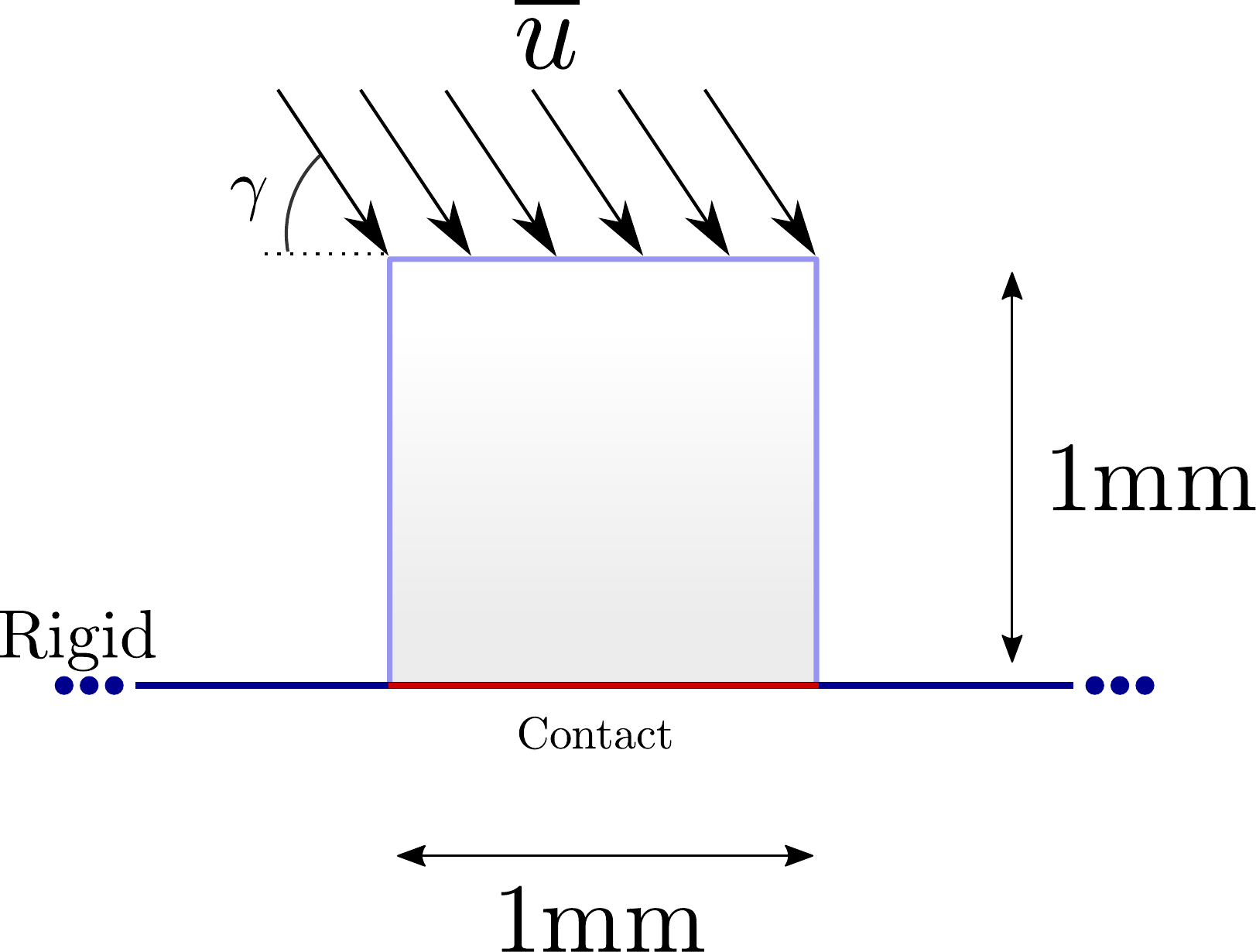}
\caption[Scheme of the two-dimensional contact problem.]{Scheme of the two-dimensional contact problem. }\label{fig::cont_application}
\end{figure}

Consider the two-dimensional contact problem shown in Figure \ref{fig::cont_application}.

\begin{figure}[hbtp]
\centering
\begin{subfigure}[t]{0.45\textwidth}
\includegraphics[scale=0.3]{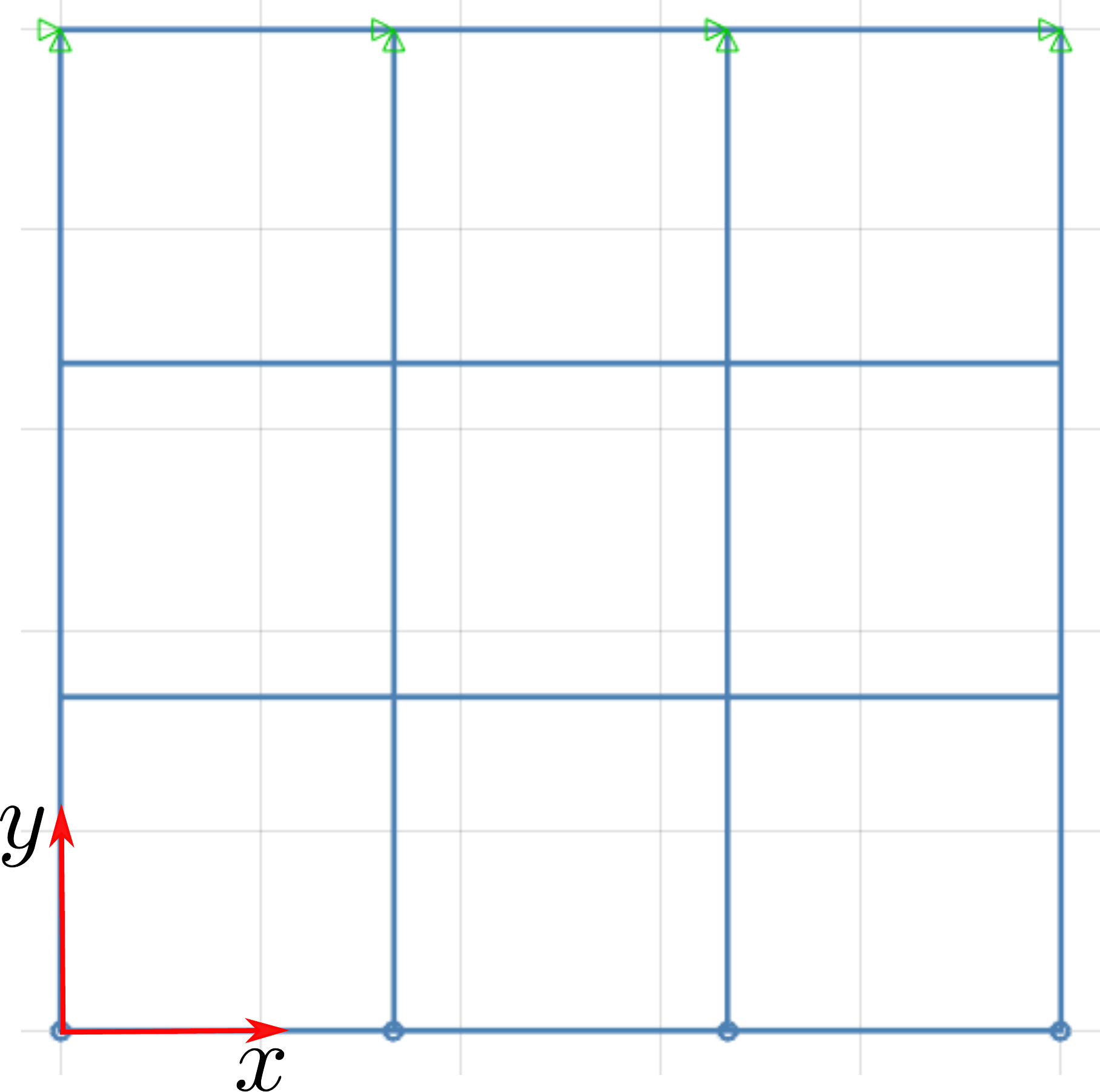} 
\subcaption{Low fidelity mesh}\label{fig::Lf_mesh}
\end{subfigure}%
\begin{subfigure}[t]{0.45\textwidth}
\includegraphics[scale=0.3]{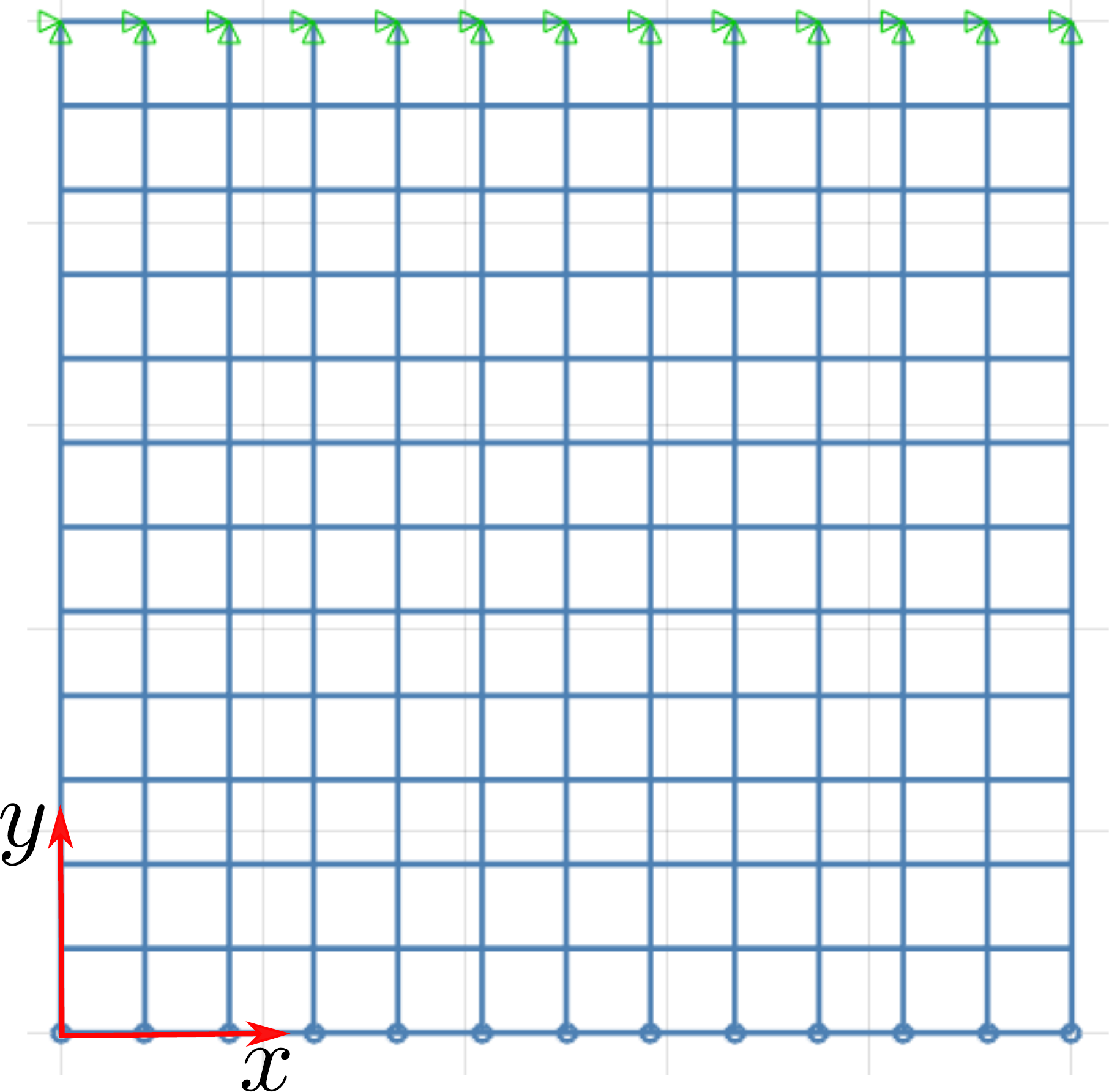} 
\subcaption{High fidelity mesh}\label{fig::hf_mesh}
\end{subfigure}
\caption[Meshes for the two-dimensional contact problem.]{Meshes for the two-dimensional contact problem. (a) Low-fidelity mesh, (b) High-fidelity- }\label{fig::d2_contactMeshes}
\end{figure}

This application is utilized to study the use of adaptive HK. An elastic-deformable
block with an area of $1 \, mm^{2}$ is
pushed onto an infinitely extended plane rigid surface by a displacement boundary condition on the top of the block. The displacement $\overline{u} = 0.3 \, mm$ is parameterized by an angle $\gamma$ in rad. The contact optimization problem is approximated by the penalty method. The mechanical problem is solved using the finite-element method. For the HK surrogate approach the high-fidelity approach discretizes the block with 144 quadrature elements considering 12 in each row and column respectively as illustrated in Figure \ref{fig::hf_mesh}. The low-fidelity approach yields a mesh of nine elements with three in each row and column as displayed in Figure \ref{fig::Lf_mesh}. 
After 10000 runs of the employed code with the same starting configurations it was found that on average the low-fidelity simulation runs 12.5 times quicker than the high-fidelity version. \\
The aim is to study the maximum von Mises stress value in x-direction over all Gauss points during the whole duration of the simulation. \\
As described in section \ref{sec::MathematicalModelOfSystem} friction force models can be distinguished into static and dynamic models. In Chapter \ref{chapt::DynamicApproach} the dynamic elastoplastic model is employed. Static models as described by
\cite{piatkowski2014dahl} are derived from the classic Coulomb model (see \cite{coulomb1785theorie}). 
When following the classification of \cite{marques2016survey}, static models can further be subdivided into models that are or are not able to capture a phenomenon called stiction.
Stiction is characterized by a higher friction force at zero relative velocity between contacting bodies \citep{morin1833new}. In this application a static model and a static model with stiction are investigated,
the classical Coulomb law (section \ref{sec::CoulombLaw}) as as well a velocity-dependent model (section \ref{label::VelDependentModel}) with stiction.  
\subsection{Static friction model: Coulomb law}\label{sec::CoulombLaw}
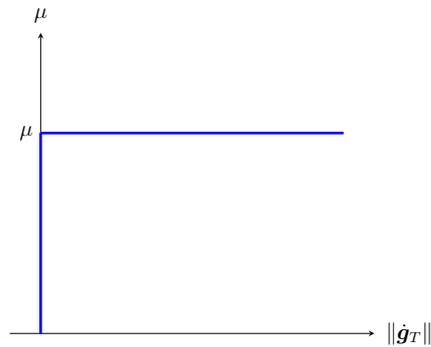
\begin{figure}[h!]
\centering
\begin{tikzpicture}[scale=0.7]
\begin{axis}
[
 axis equal,
xlabel={$\norm{\dot{\bm{g}}_{T}}$},
ylabel={$\mu$},
y label style={anchor=north},
axis lines=middle,
xtick={0},
xticklabels={},
    ytick={0},
    ymin=0,ymax=1.5,
    xmin=0,xmax=1.5,
    axis on top=false,
every axis x label/.style={
    at={(ticklabel* cs:1.01)},
    anchor=west,
},
every axis y label/.style={
    at={(ticklabel* cs:1.01)},
    anchor=south,
},
]
\addplot[domain=-1.5:0,blue,line width = 0.5mm]{-1}; 
\addplot[domain=0:1.5,blue,line width = 0.5mm]{(1}; 
\addplot [black, nodes near coords=$\mu$,every node near coord/.style={anchor=0}] coordinates {( 0, 1)};
\draw[blue,thick,line width = 0.5mm] (axis cs:0,1) -- (axis cs:0,-1);
\end{axis}

\end{tikzpicture}
\caption[Static friction law.]{Coefficient of friction for the static Coulomb friction model.}\label{fig::Coulomb}
\end{figure}

The investigated static friction model is the classical Coulomb model with friction coefficient $\mu$. The friction law takes the form
\begin{equation}\label{eq::ColoumbFriction}
\bm{t}_{T} = \mu \abs{p_{N}} \dfrac{ \dot{\bm{g}}_{T}}{\norm{\dot{\bm{g}}_{T}}} , \qquad \text{if} \, \, \norm{\bm{t}_{T}} > \mu \abs{p_{N}},
\end{equation}
where $\bm{t}_{T} $ is the tangential stress, $p_{N}$ is the contact normal pressure and $\dot{\bm{g}}_{T}$ is the relative velocity between the two contacting bodies.
The friction coefficient is constant with increasing relative velocity $\dot{\bm{g}}_{T}$ between the two contacting bodies as seen in Figure \ref{fig::Coulomb}.
\newpage
\subsubsection{Coulomb law with two input parameters}
\begin{figure}[b!]
\centering
\includegraphics[scale=0.5]{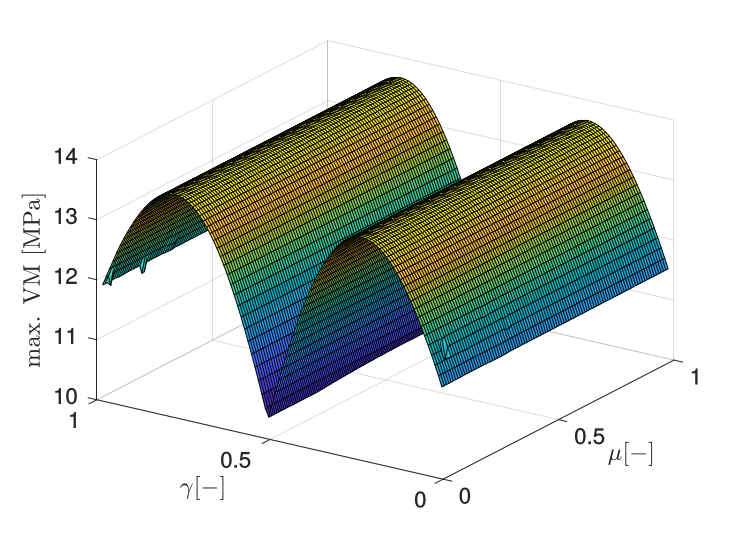} 
\caption[Shape of the output function for the two-dimensional contact problem]{Shape of the output function for the two-dimensional contact problem i.e. the maximum von Mises stress in MPa over the input domain. }\label{fig::d2_contact_result}
\end{figure}

\begin{figure}[hbtp]
\centering
\begin{subfigure}[t]{0.45\textwidth}
\includegraphics[scale=0.35]{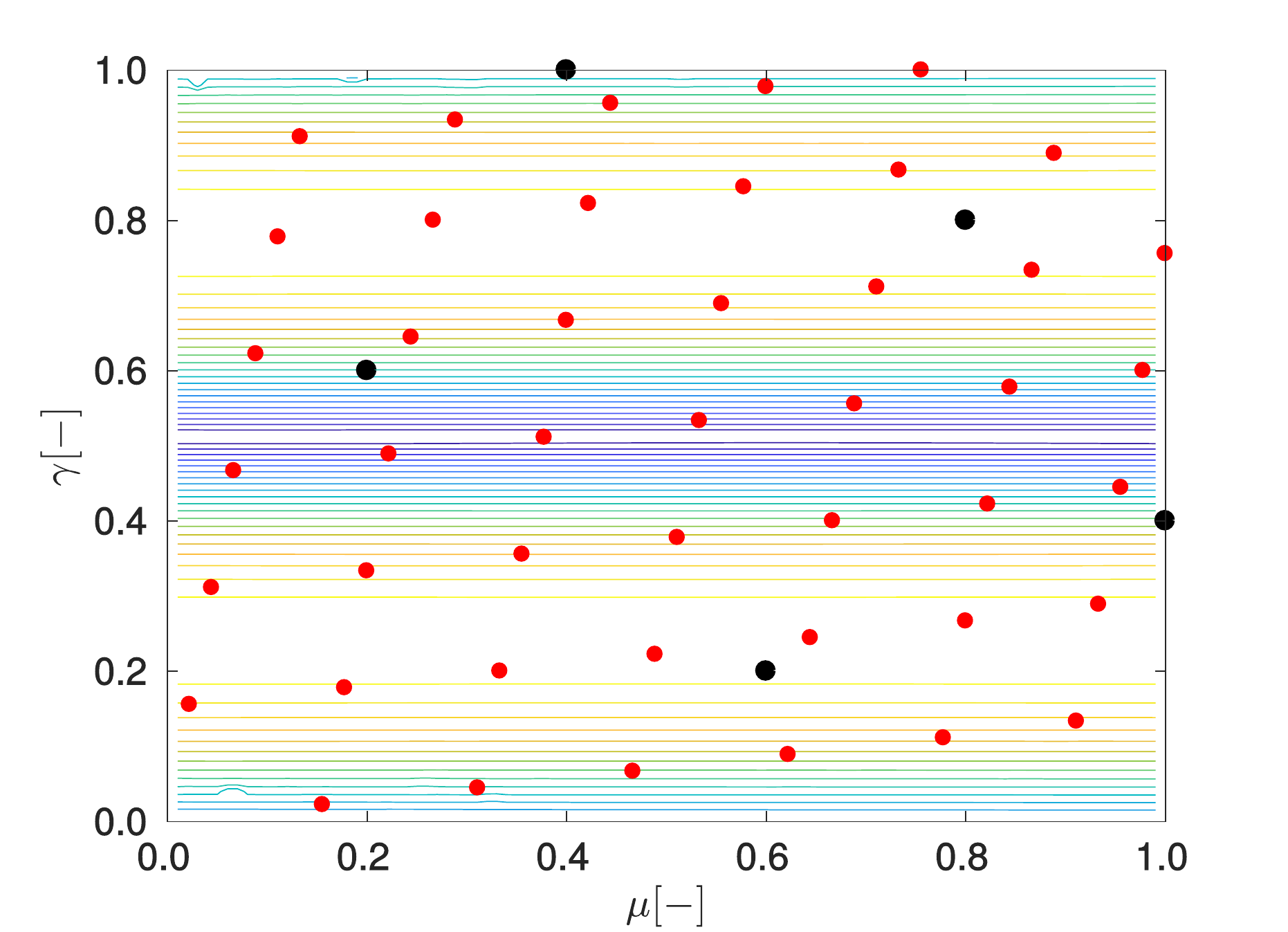} 
\subcaption{Initial samples}\label{fig::Contact2dInitialSamples}
\end{subfigure}%
\begin{subfigure}[t]{0.45\textwidth}
\includegraphics[scale=0.35]{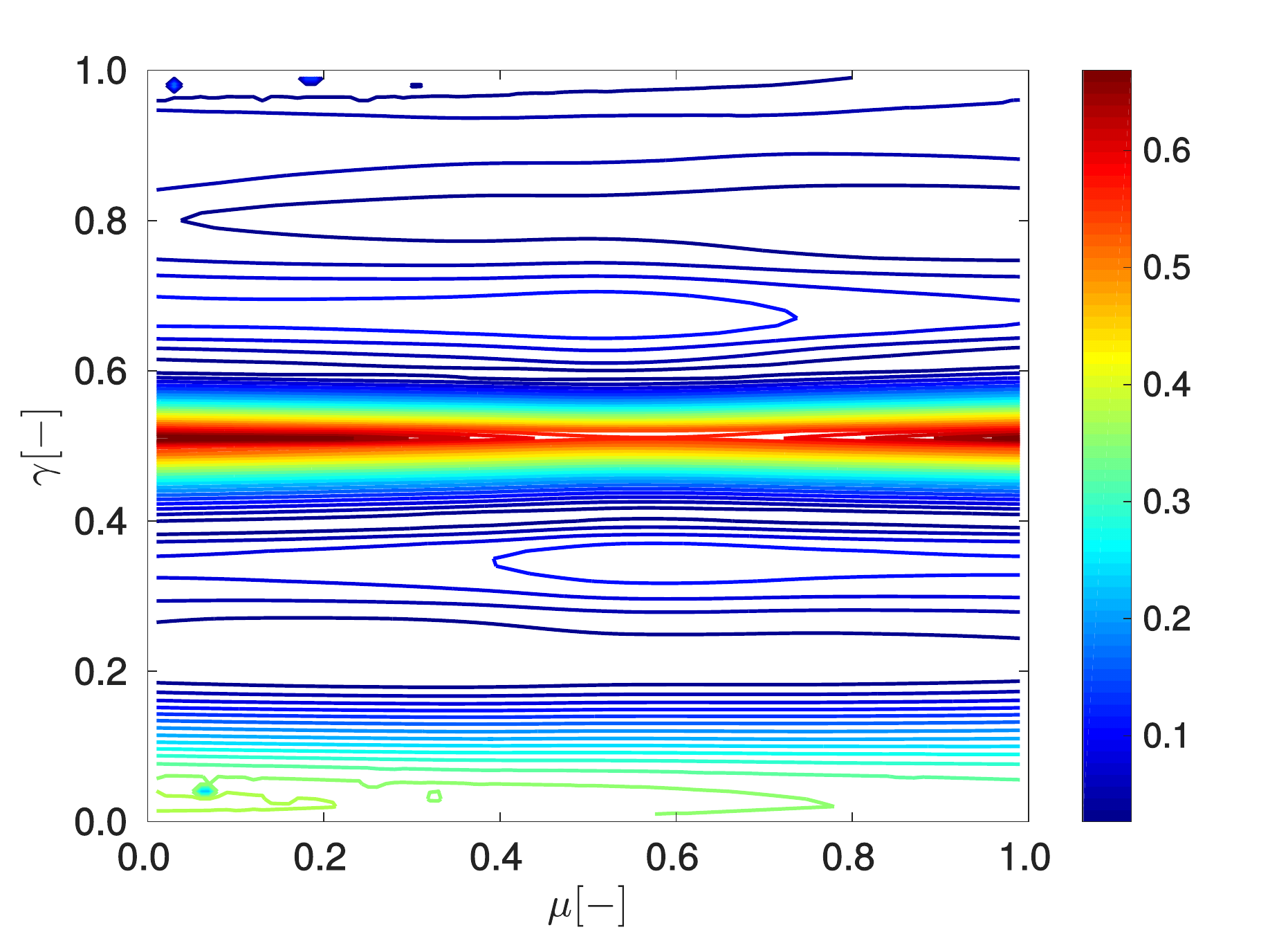} 
\subcaption{Initial absolute error}\label{fig::Contact2sInitialAB}
\end{subfigure}
\caption[Initial state for the metamodel gneration for the two-dimensional contact problem.]{Initial state for the metamodel gneration for the two-dimensional contact problem. (a) 20 Initial samples, (b) initial absolute error. }\label{fig::Contact2dInitial}
\end{figure}

In this academic example the Young's modulus of the block is given as $E = 10 \text{MPa}$ and the Poisson ration $\nu$ is set to $0.3$. 
 The parametric space for the surrogate model is given by the parameter set $\mu \in \left[ 0.3 , 0.5 \right]$ and $\gamma \in \left[ 0.7, 2.4 \right]$. The maximum von Mises stress in x-direction over this parametric space is displayed in Figure \ref{fig::d2_contact_result}.   
Initially 5 HF and 45 LF sample points are created with TPLHD as shown in Figure \ref{fig::Contact2dInitialSamples}. Therefore the computation time for the initial surrogate model corresponds to the same time required to compute 9 HF-samples. The absolute error over the input domain of this setting is displayed in Figure \ref{fig::Contact2sInitialAB}. \\It can be observed that especially the center of the domain lacks accuracy. Furthermore the left-hand side also shows increased errors. The error measures of the initial metamodel are: MAE:$0.1264$, RMAE:$ 0.7756$ and RMSE:$0.1878$.

For 10000 points the error measures between the high-fidelity and low-fidelity simulations yield the following differences: MAE: 5.9318,	RMAE: 7.3037, RMSE: 5.9698 and R$^{2}$-score: -42.9293. Therefore there is a significant difference in the output values of the two simulation configurations. \\
The error measures after 15 HF samples are listed in Table \ref{tab::FE2d15}. All the investigated techniques except EI, which shows clustering issues, improve the metamodel significantly in comparison to the initial prediction. 5 out of the 13 investigated adaptive sampling techniques yield a better MAE error than the one-shot TPLHD technique. However the RMAE measure is more proficient in 8 out of the 13 techniques. MEPE shows the best overall results with the best MAE, RMSE and R$^{2}$ measures.
\begin{table}[ht!]
\begin{center}
\resizebox{1.0\textwidth}{!}{%
\begin{tabular}{l|l c c c c} \hline
& Method & MAE & RMAE & RMSE & R$^{2}$ \\ \hline\hline \\
\multirow{1}{*}{\shortstack[l]{Errors after\\5-HF samples}} & TPLHD &  0.1264 & 0.7756 & 0.1878& 0.9565  \\\\  \hline \\
\multirow{14}{*}{\shortstack[l]{ Errors after\\15-HF samples}} &TPLHD  & 0.0355 & 0.4366 & 0.0585& 0.9957  \\
&ACE &  0.0300 & 0.1872 & 0.0410 & 0.9979 \\
&AME &  0.0243 & 0.2259 & 0.0356 & 0.9984 \\ 
&CVD & 0.0371 & \textbf{0.1809} & 0.0513 & 0.9967  \\
&CVVor & 0.0333 & 0.2305 & 0.0507 & 0.9968\\ 
&EI &  0.2789 & 0.7432 & 0.3575 & 0.8424  \\
&EIGF & 0.0453 & 0.3295 & 0.0741 & 0.9932  \\
&LOLA &  0.0390 & 0.2935 & 0.0569 & 0.9960  \\
&MASA &  0.0314 & 0.4378 & 0.0602 & 0.9955\\
&MEPE & \textbf{0.0276} & 0.2198 & \textbf{0.0348} & \textbf{0.9985} \\
&MIPT &  0.0398 & 0.4561 & 0.0681 & 0.9942  \\
&MSD &  0.0518 & 0.5951 & 0.1027 & 0.9869\\
&SFCVT &   0.0795 & 0.6299 & 0.1292 & 0.9794  \\
&SSA & 0.0372 & 0.2799 & 0.0542 & 0.9963
\end{tabular}
}
\end{center}
\caption[Error measures for FE contact problem with static friction and two-dimensional parametric space after 15 HF samples]{Error measures for FE contact problem with static friction and two-dimensional parametric space after 15 HF samples.}\label{tab::FE2d15}
\end{table}
\begin{table}[hb!]
\begin{center}
\resizebox{0.35\linewidth}{!}{%
\begin{tabular}{l| c } \hline
\shortstack[l]{Sampling\\method} & \shortstack[l]{Average number\\of Samples}\\ \hline\hline \\
ACE &  38 $\pm$ 1  \\
AME & 37 $\pm$ 0  \\ 
CVD & 61 $\pm$ 2 \\
CVVor &  - \\ 
EI & - \\
EIGF & 67 $\pm$ 0 \\
LOLA & - \\
MASA & 52 $\pm$ 3  \\
MEPE & 45 $\pm$ 0 \\
MIPT & 53 $\pm$ 1 \\
MSD & -  \\
SFCVT &  -\\
SSA &  45 $\pm$ 1  \\ 
\end{tabular}
}
\end{center}
\caption[Average amount of samples needed for FE contact problem with static friction and two-dimensional parametric space to reach error threshold]{Average amount of samples needed for the FE contact problem with static friction and two-dimensional parametric space to get MAE below $0.01$ up until 70 HF sample points (methods that do not reach the threshold are left blank).}\label{tab::FE2dLimit}
\end{table}

\begin{figure}[htbp]
\centering
\includegraphics[scale=0.4]{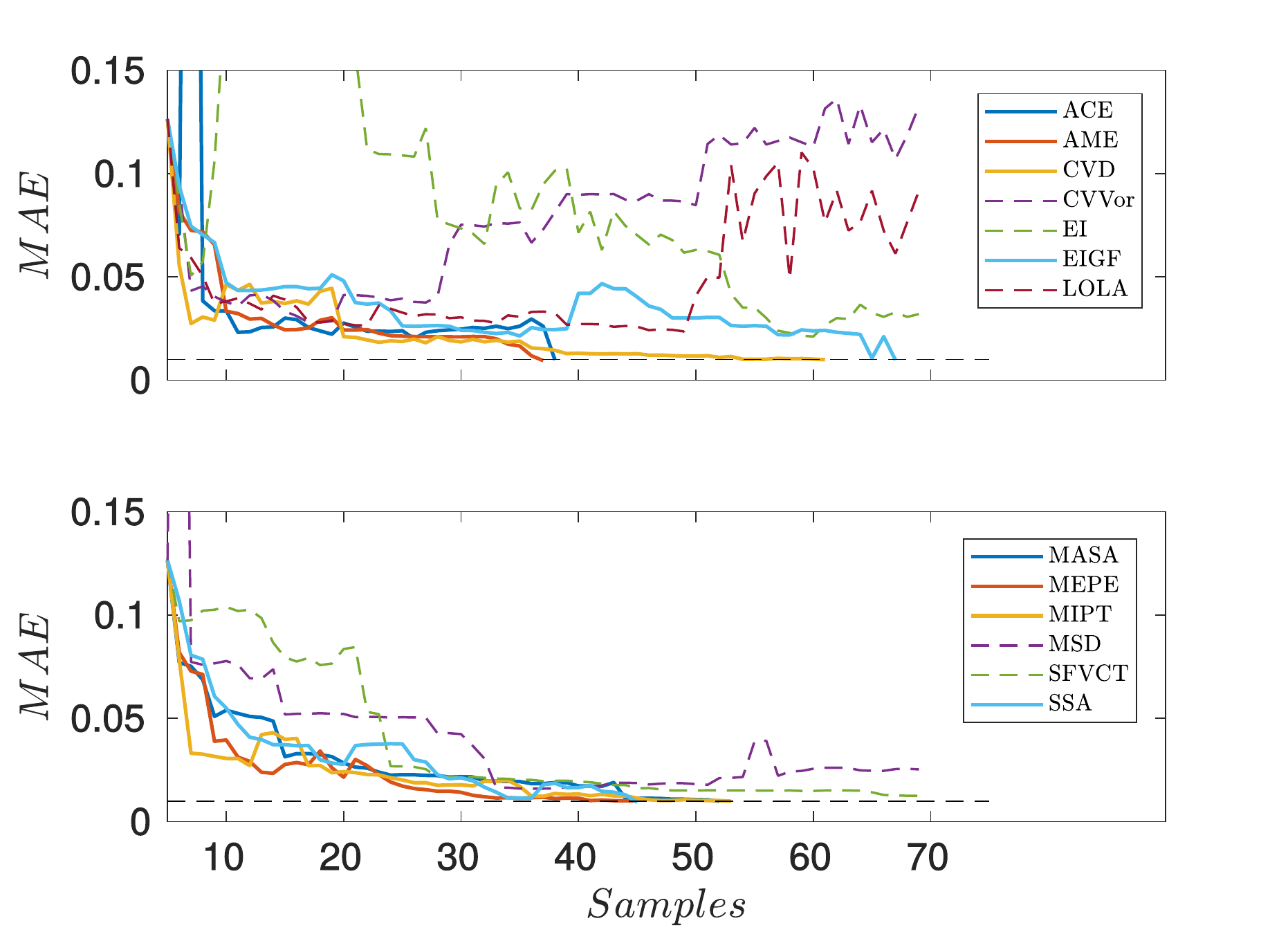}
\caption[Convergence of the MAE error for the FE contact problem with static friction and two-dimensional parametric space]{Convergence of the MAE error for the FE contact problem with static friction and two-dimensional parametric space. Dashed line indicates methods that do not reach the MAE threshold as represented by the horizontal line.}\label{fig::FE2dConvergencePlot}
\end{figure}

In a next step the convergence of the methods with respect to the MAE measure is studied. Until 70 HF samples the average number of samples for each method to reach a threshold value of $0.01$ is studied. The results are shown in Table \ref{tab::FE2dLimit}. Blank lines indicate that the respective method has not reached the limit within the 70 sample constraint. The variation number over the 10 iterations is represented by the $\pm$ values. It can be seen that 8 out of the 13 samples are able to reach the threshold. In this two-dimensional case ACE and AME perform best.
The convergence of the average MAE value for the respective methods is depicted in Figure \ref{fig::FE2dConvergencePlot}. The dashed lines indicate methods that do not reach the threshold. The horizontal line represents the MAE threshold value.

\begin{figure}[htbp]
\centering
\begin{subfigure}[t]{0.4\textwidth}
\includegraphics[scale=0.35]{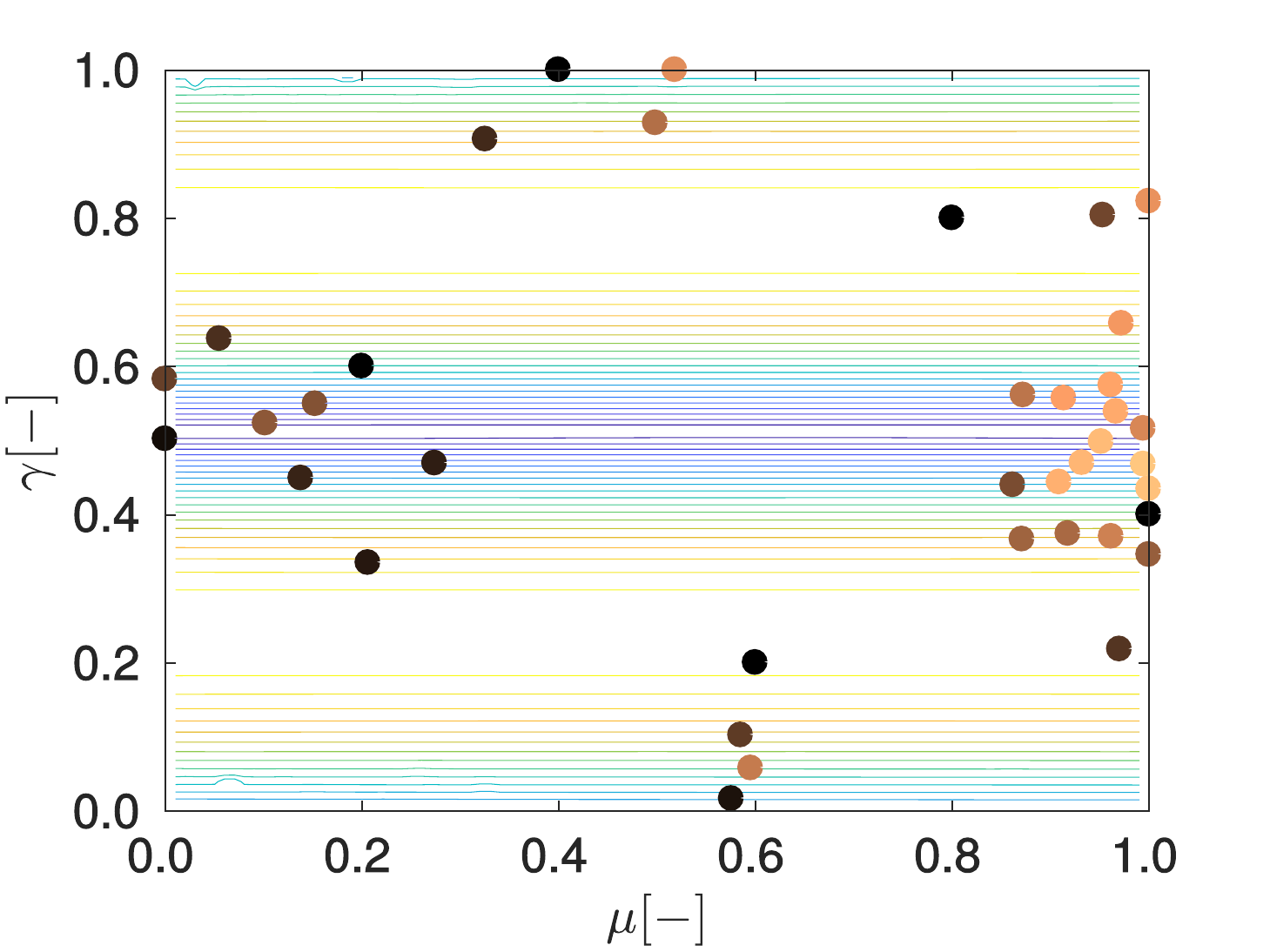}
\subcaption{ACE - 38 samples}\label{fig:Contact2dYesACE}
\end{subfigure}%
\begin{subfigure}[t]{0.4\textwidth}
\includegraphics[scale=0.35]{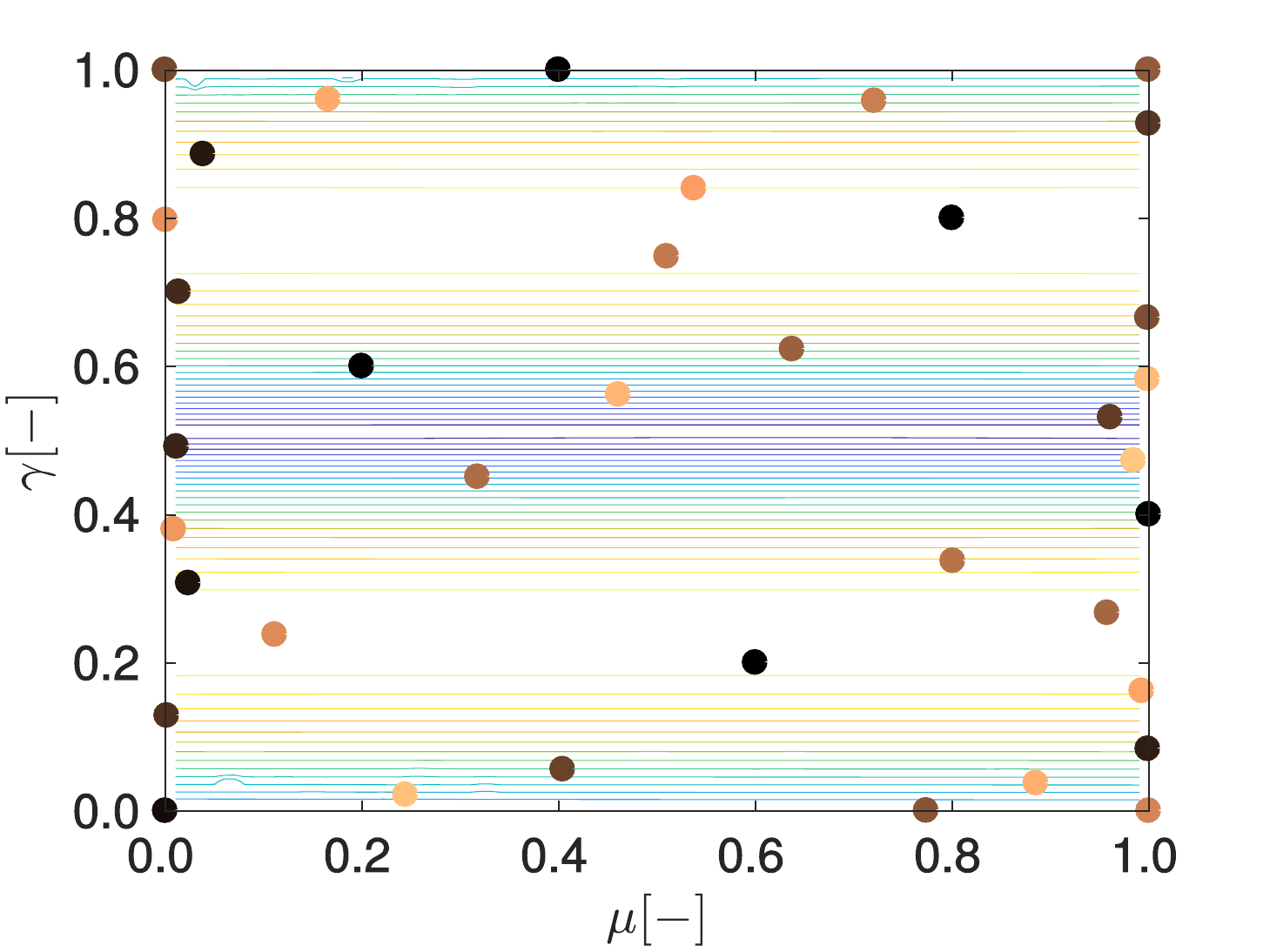}
\subcaption{AME - 37 samples}\label{fig:Contact2dYesAME}
\end{subfigure}
\begin{subfigure}[t]{0.4\textwidth}
\includegraphics[scale=0.35]{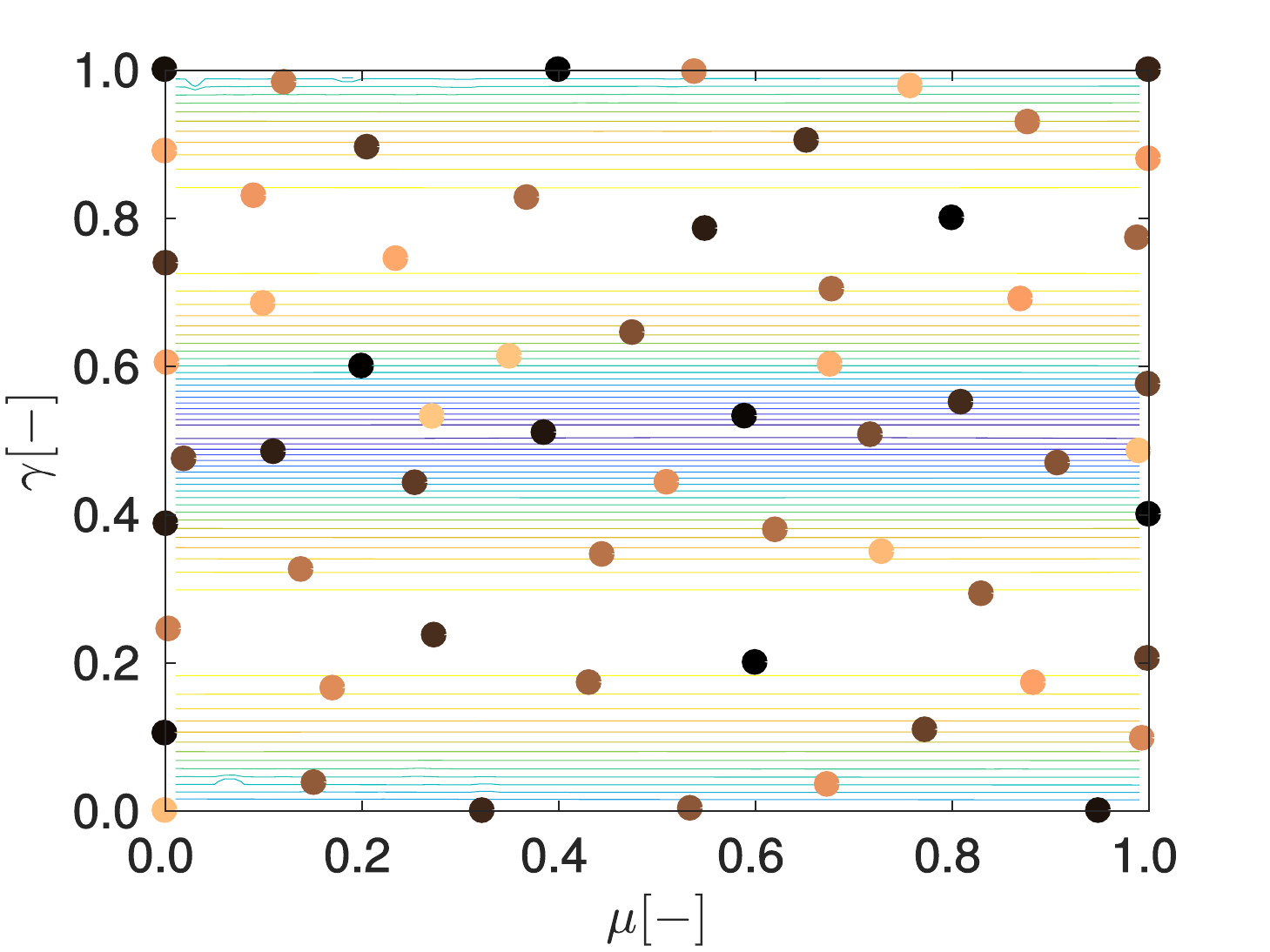}
\subcaption{CVD - 61 samples}\label{fig:Contact2dYesMSE}
\end{subfigure}%
\begin{subfigure}[t]{0.4\textwidth}
\includegraphics[scale=0.35]{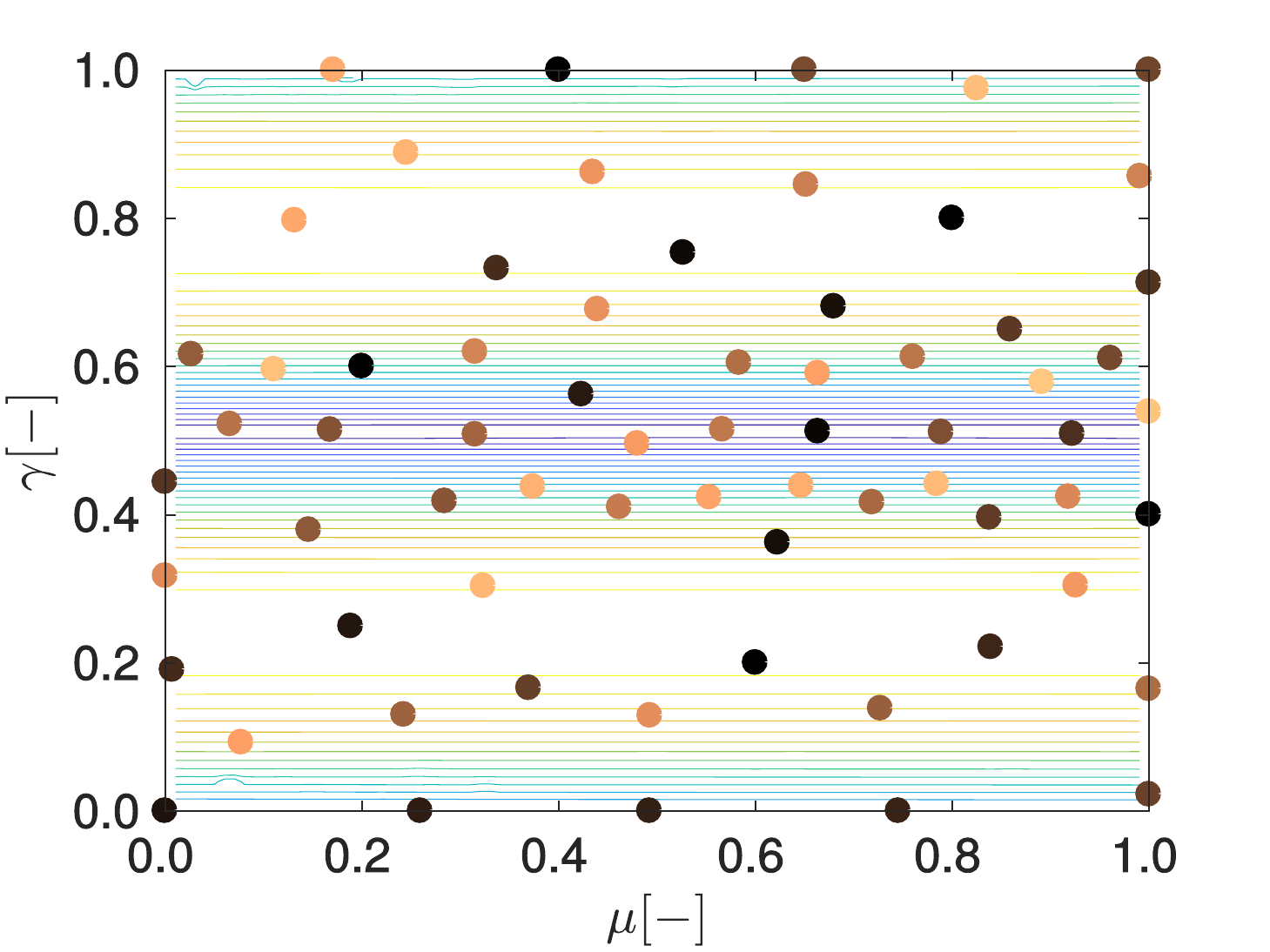}
\subcaption{EIGF - 67 samples}\label{fig:Contact2dYesEIGF}
\end{subfigure}
\begin{subfigure}[t]{0.4\textwidth}
\includegraphics[scale=0.35]{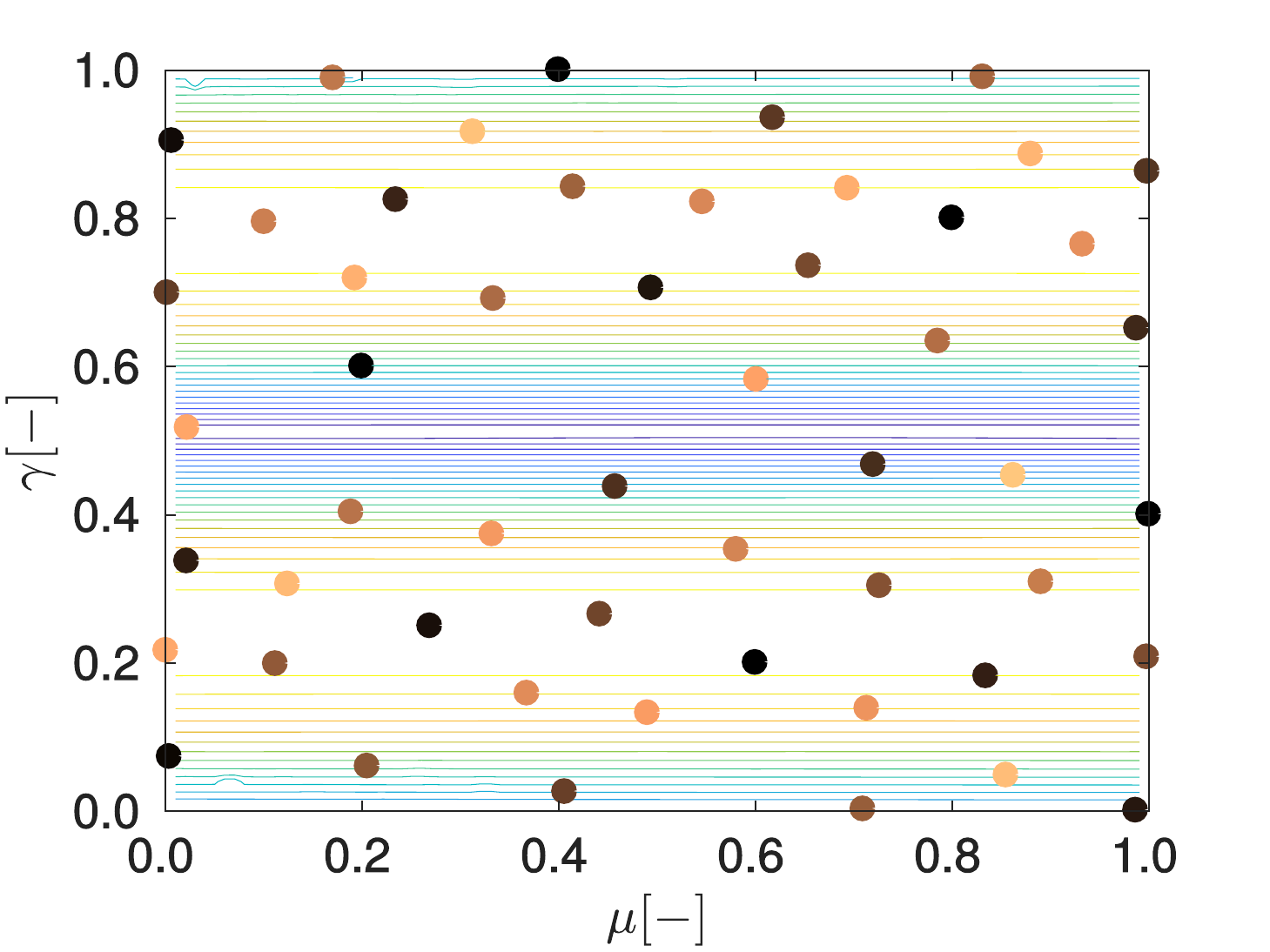}
\subcaption{MASA - 52 samples}\label{fig:Contact2dYesMASA}
\end{subfigure}%
\begin{subfigure}[t]{0.4\textwidth}
\includegraphics[scale=0.35]{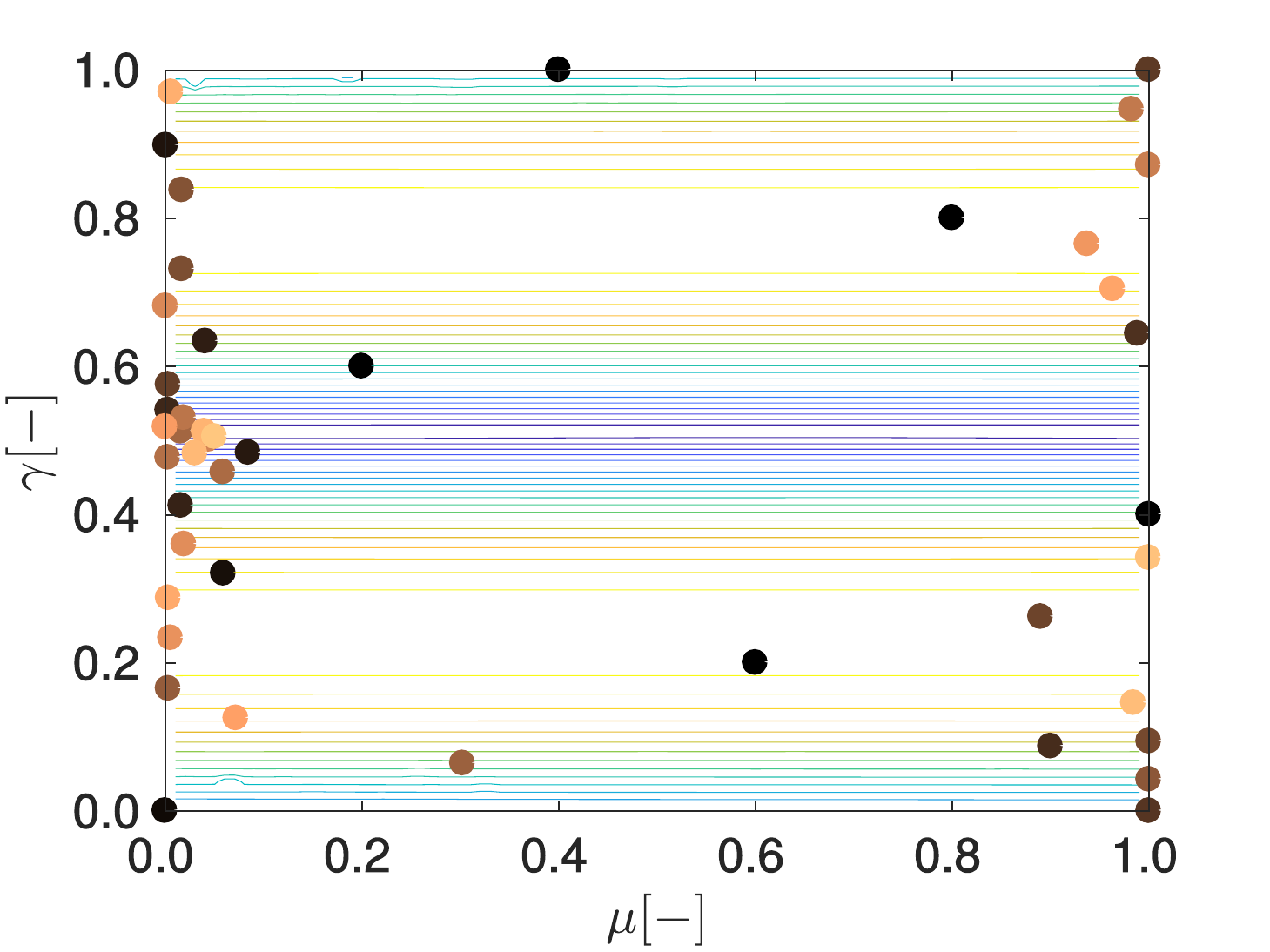}
\subcaption{MEPE - 45 samples}\label{fig:Contact2dYesMEPE}
\end{subfigure}
\begin{subfigure}[t]{0.4\textwidth}
\includegraphics[scale=0.35]{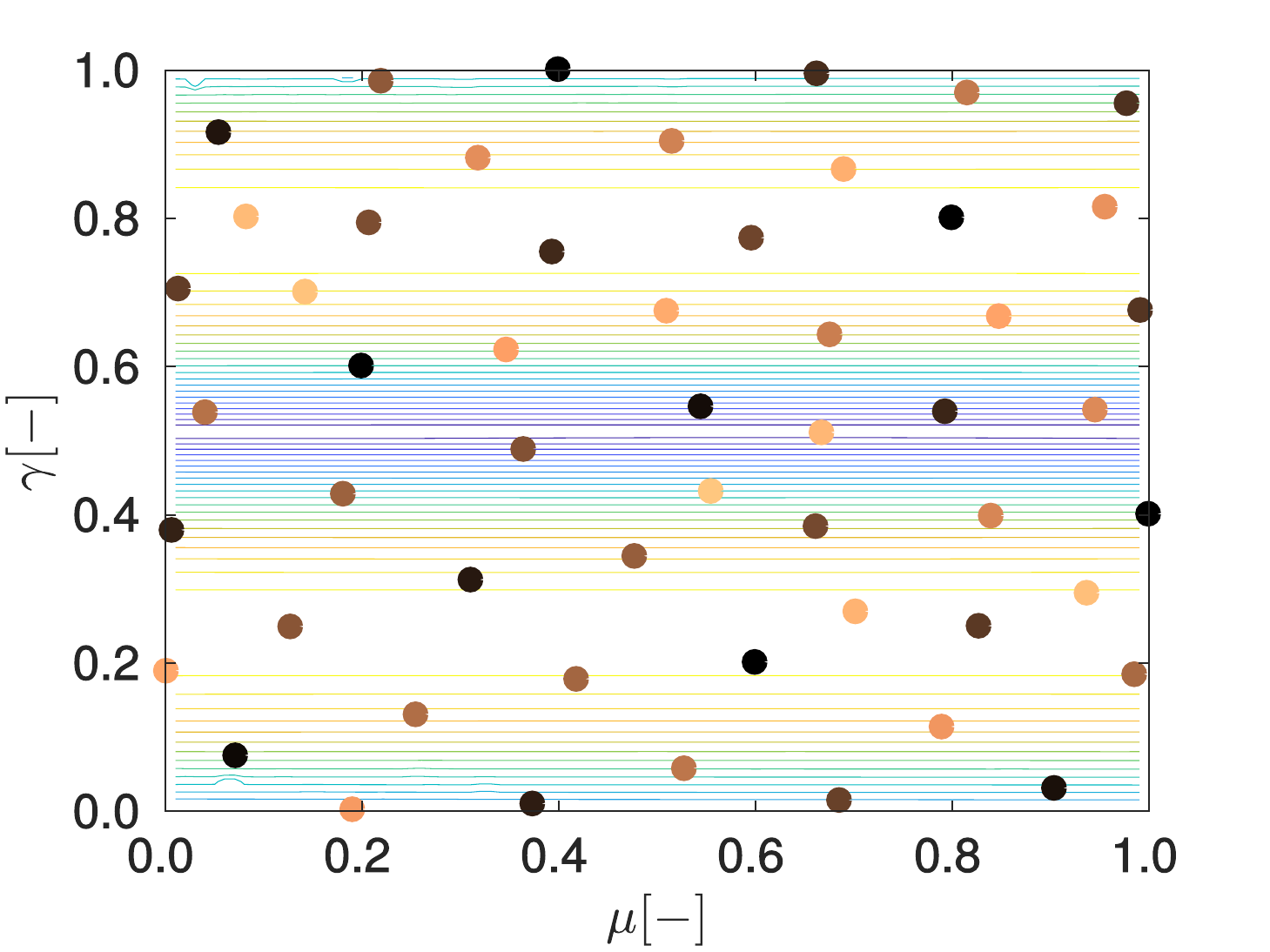}
\subcaption{MIPT - 53 samples}\label{fig:Contact2dYesMIPT}
\end{subfigure}%
\begin{subfigure}[t]{0.4\textwidth}
\includegraphics[scale=0.35]{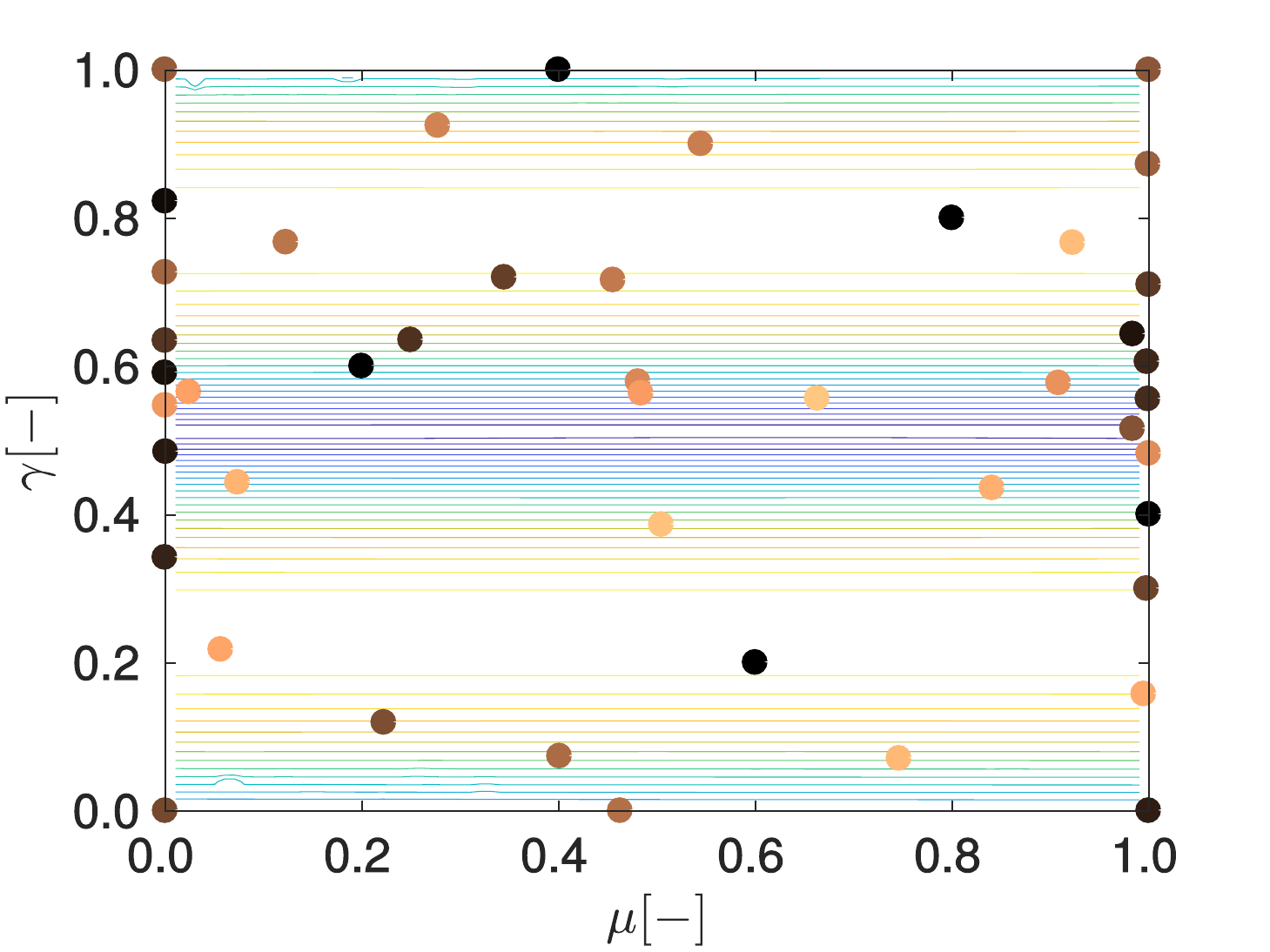}
\subcaption{SSA - 45 samples}\label{fig:Contact2dYesSSA}
\end{subfigure}%
\caption[Location of samples of respective adaptive sampling technique for the FE contact problem with static friction and two-dimensional parametric space]{Location of samples of respective adaptive sampling technique for the FE contact problem with static friction and two-dimensional parametric space to reach a MAE-value of below $0.01$.}\label{fig:Contact2dYes}%
\end{figure}

The sample positions of the methods that reach the limit are summarized in Figure \ref{fig:Contact2dYes}. The evaluation  of the sample positions is done with regards to the shape of the initial absolute error as shown in Figure \ref{fig::Contact2sInitialAB}. 
ACE requires on average 38 samples. One set of sample positions is shown in Figure \ref{fig:Contact2dYesACE}.  The exploitation component of the method can be seen. The majority of points are sampled around the initially problematic center of the domain. \\
The least amount of samples are needed with AME (Figure \ref{fig:Contact2dYesAME}). The technique samples predominantly around the edges of the domain where there are no initial HF-samples. With this approach AME is able to achieve a proficient result. 
The exploration character of CVD is noticeable in Figure \ref{fig:Contact2dYesMSE}. The points are spread evenly. As seen in previous studies, the exploitation character of CVD is the less dominant component.  
As previously noticed, EIGF (Figure \ref{fig:Contact2dYesEIGF}) focuses the majority of the points around the highest absolute value. Since the highest absolute error in this application the horizontal line of $\gamma = 0.5$ (in the normalized case) the samples are evenly spread in this area.  
The MASA technique is illustrated in Figure \ref{fig:Contact2dYesMASA}. The sample points are spread evenly in the area. This means the individual metamodels of the committee members result in rather constant prediction errors. 
\newpage
The method that balances exploitation and exploration adaptively is MEPE as shown in Figure \ref{fig:Contact2dYesMEPE}. This feature can be seen in the sample positions in this application. The edges are sampled. However the majority of the points are around the center of the domain with the highest initial absolute error. 
The sequential space-filling method MIPT is shown in Figure \ref{fig:Contact2dYesMIPT}. With 53 needed samples it is a good compromise between accuracy and required samples for symmetric application problems as seen in this application.
SSA is depicted in Figure \ref{fig:Contact2dYesSSA}. The method highlights a proficient balance between exploration and exploitation.

\subsubsection{Coulomb law with four input parameters}\label{sec::CoulombLaw4d}
As an extension to the two-dimensional case the Young's modulus and the displacement boundary are added to the input space. Hence, the
parametric space for the surrogate model is given by the parameter set $\mu \in \left[ 0.3 , 0.5 \right]$, $\gamma \in \left[ 0.7, 2.4 \right]$, $\overline{u} \in \left[ 0.28, 0.3 \right]mm$ and $E \in \left[ 9.9, 10.1 \right] \, \text{MPa}$.
The low-fidelity sample size is set to 150 which corresponds to the simulation time of 12 high-fidelity simulations. 40 initial HF samples are created with TPLHD. Here, only the proficient adaptive sampling techniques of the lower dimensional case are considered. The error measures after 80 HF-samples are shown in Table \ref{tab::d4Measures}. ACE has issues with clustering and therefore numerical problems in the majority of the 10 iterations. The values for ACE are therefore left blank. It can be noticed that CVD yields the best results. Furthermore all considered sampling techniques are able to reduce the initial error significantly and show a more proficient performance when compared to TPLHD at 80 HF-samples. When comparing these results to the two-dimensional problem of the previous section it can be observed that the added parameters do not necessarily increase the complexity (in form of e.g. nonlinearities) of the black-box function. 
\begin{table}[ht!]
\begin{center}
\resizebox{1.0\textwidth}{!}{%
\begin{tabular}{l|l c c c c} \hline
& Method & MAE & RMAE & RMSE & R$^{2}$ \\ \hline\hline \\
\multirow{1}{*}{\shortstack[l]{Errors after\\40-HF samples}} & TPLHD &  0.1660 &0.4909 & 0.2195 & 0.9748 \\\\  \hline \\
\multirow{0}{*}{\shortstack[l]{ Errors after\\80-HF samples}} &TPLHD & 0.1369 & 0.4912 & 0.1776 & 0.9835  \\
&ACE &  - & - & - & -\\
&AME &  0.0868 & 0.3854 & 0.1106 & 0.9936 \\ 
&CVD & \textbf{0.0502} & 0.4380 & \textbf{0.0716} &\textbf{ 0.9973} \\
&EIGF & 0.0969 & 0.3510 & 0.1269 & 0.9915 \\
&MASA &  0.1033 & 0.4605 & 0.1313 & 0.9896\\
&MEPE & 0.0782 & 0.4566 & 0.1071 & 0.9940 \\
&MIPT &  0.0779 & \textbf{0.3457} & 0.1011 & 0.9946 \\
&SSA & 0.0885 & 0.3760 & 0.1120 & 0.9934
\end{tabular}
}
\end{center}
\caption[Error measures for FE-contact problem with static friction and four-dimensional parametric space after 80-HF samples]{Error measures for FE-contact problem with static friction and four-dimensional parametric space after 80-HF samples (blank rows indicate numerical problems).}\label{tab::d4Measures}
\end{table}

\begin{figure}[hbtp]
\centering
\includegraphics[scale=0.5]{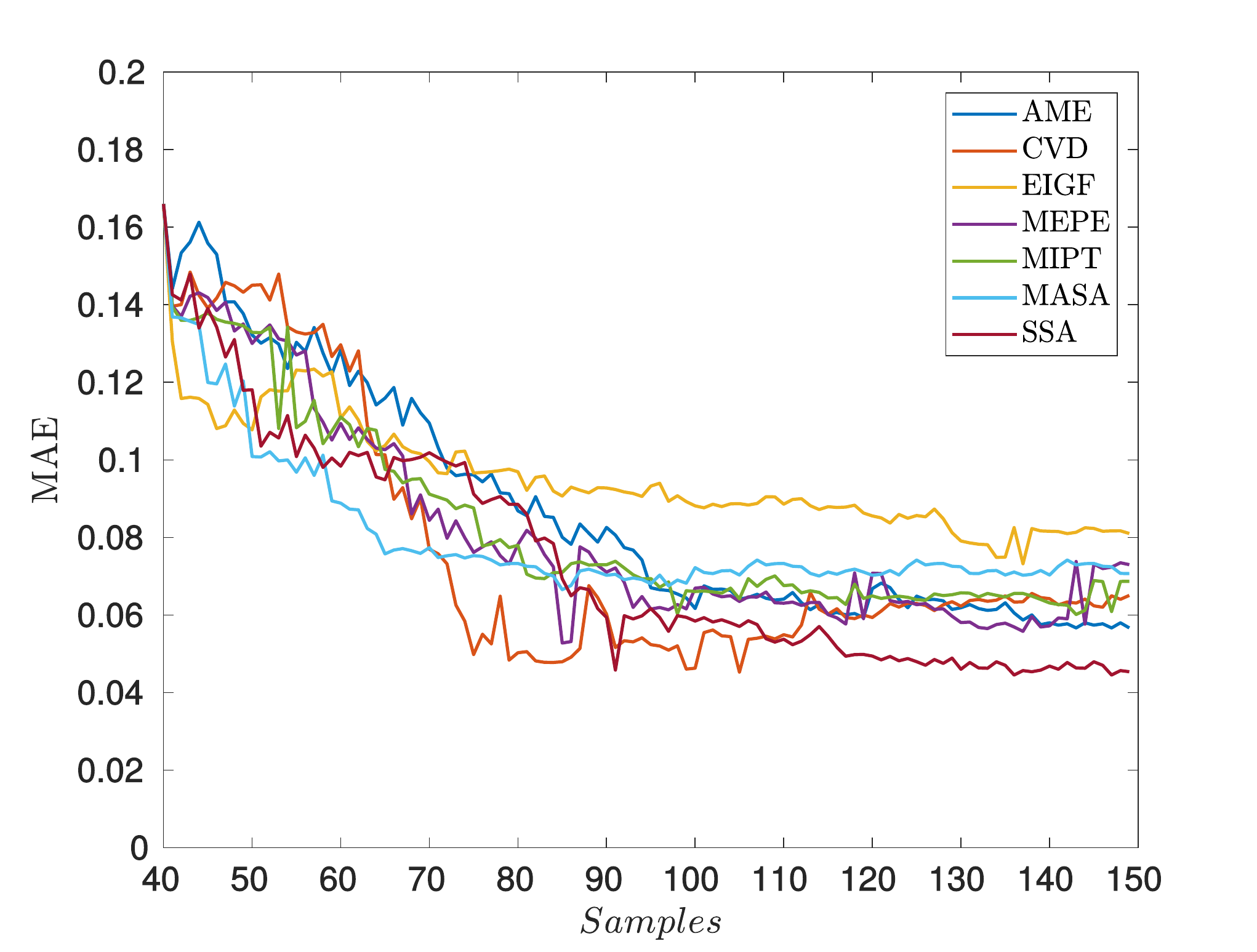}
\caption[Convergence of the MAE error for the FE-contact problem with static friction and four-dimensional parametric space]{Convergence of the MAE error for the FE-contact problem with static friction and four-dimensional parametric space. }\label{fig::ConvergenceD4COntact}
\end{figure}

The convergence of the MAE error is depicted in Figure \ref{fig::ConvergenceD4COntact}. It can be observed that all techniques show a drastic initial decrease. However after around 90 samples the measure appears to be stalling. Furthermore it can be seen that CVD, the most proficient method at 80 HF-samples (see Table \ref{tab::d4Measures}), exhibits numerical issues provoked by clustering. The MAE error value of the method increases after 100 samples. Overall SSA yields the most proficient MAE measure after 150 samples. 
\clearpage
\subsection{Velocity-dependent Coulomb's friction law}\label{label::VelDependentModel}
\begin{figure}[h!]
\centering
\begin{tikzpicture}[scale=0.7]
\begin{axis}
[
 axis equal,
xlabel={$\norm{\dot{\bm{g}}_{T}}$},
ylabel={$\mu$},
y label style={anchor=north},
axis lines=middle,
xtick={0},
xticklabels={},
    ytick={0},
    ymin=0,ymax=1.5,
    xmin=0,xmax=1.5,
    axis on top=false,
every axis x label/.style={
    at={(ticklabel* cs:1.01)},
    anchor=west,
},
every axis y label/.style={
    at={(ticklabel* cs:1.01)},
    anchor=south,
},
]
\addplot [black, nodes near coords=$\mu_{k}$,every node near coord/.style={anchor=0}] coordinates {( 0, 1.0)};
\addplot [black, nodes near coords=$\mu_{s}$,every node near coord/.style={anchor=0}] coordinates {( 0, 1.3)};
\addplot[blue,line width = 0.5mm, domain=0:1.5] { (1.0 + (1.3-1.0)*exp(-2*(abs(x)))} ;

\draw[blue,line width = 0.5mm] (axis cs:0,1.3) -- (axis cs:0,-1.3);
\draw[dashed,line width = 0.1mm] (axis cs:0,1.0) -- (axis cs:1.5,1.0);
\end{axis}

\end{tikzpicture}
\caption[Static friction law with stiction.]{Coefficient of friction for the static Coulomb friction model with stiction.}\label{fig::CoulombWithStiction}
\end{figure}
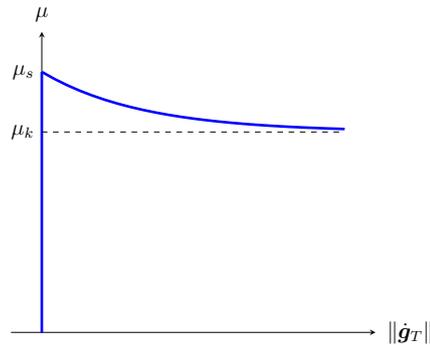

The investigated static friction law with stiction \citep{wriggers2004computational} is considered in the following. Here, the friction force is higher at standstill than when sliding occurs. Hence, the friction coefficient is now considered to be dependent on the relative velocity.  
Therefore equation (\ref{eq::ColoumbFriction}) can be rewritten to include this dependence with
\begin{equation}
\bm{t}_{T} = \mu(\dot{\bm{g}}_{T}) \abs{p_{N}} \dfrac{ \dot{\bm{g}}_{T}}{\norm{\dot{\bm{g}}_{T}}} , \qquad \text{if} \, \, \norm{\bm{t}_{T}} > \mu(\dot{\bm{g}}_{T}) \abs{p_{N}}.
\end{equation}
A heuristic friction law that incorporates $\dot{\bm{g}}_{T}$ into its expression is  
\begin{equation}
\mu(\dot{\bm{g}}_{T}) = \mu_{k} + (\mu_{s} - \mu_{k}) \exp (- c \norm{\dot{\bm{g}}_{T}}).
\end{equation}
Here three constitutive parameters, the kinetic friction $\mu_{k}$, the static friction $\mu_{s}$ and an additional parameter $c$ that governs how fast the static coefficient approaches the static one as depicted in Figure \ref{fig::CoulombWithStiction}. As a result the friction force at standstill is higher than with sliding motion between the contacting bodies.
\subsubsection{Velocity-dependent law with four input parameters}\label{sec::VelDependetLaw4d}
Consider the four-dimensional parametric space given by $\mu_{s} \in \left[ 0.2 , 0.3 \right]$, $\mu_{k} \in \left[ 0.1 , 0.15 \right]$, $c \in \left[ 1.0, 2.0 \right]$, $\gamma \in [0.7, 2.4]$. The Young's modulus is set to $E = 10 \text{MPa}$ and the Poisson ration $\nu$ to $0.3$.
The low-fidelity sample size is set to 150 which corresponds to the simulation time of 12 high-fidelity simulations. 40 initial HF-samples are created with TPLHD. The same adaptive sampling techniques as in section \ref{sec::CoulombLaw4d} are employed. The error measures after 80 HF-samples are listed in Table \ref{tab::Contactvbd4}. Similarly to section \ref{sec::CoulombLaw4d}, ACE has issues with clustering. After adding 40-HF samples it can be seen that all selected sampling techniques are able to significantly reduce the initial error. CVD is the best performing method, i.e. reducing the RMSE error by factor four when comparing to the initial error. Furthermore all methods perform more proficiently than the one-shot TPLHD method at 80 HF-samples.
\begin{table}[ht!]
\begin{center}
\resizebox{1.0\textwidth}{!}{%
\begin{tabular}{l|l c c c c} \hline
& Method & MAE & RMAE & RMSE & R$^{2}$ \\ \hline\hline \\
\multirow{1}{*}{\shortstack[l]{Errors after\\10-HF samples}} & TPLHD &  0.1427 & 0.8285 &0.2072 & 0.9473 \\\\  \hline \\
\multirow{0}{*}{\shortstack[l]{ Errors after\\20-HF samples}} &TPLHD  & 0.1266 & 0.6578 &0.1688 & 0.9650   \\
&AME &  0.0815 & 0.6289 & 0.1105 & 0.9850\\ 
&CVD & \textbf{0.0409} & 0.3444 & \textbf{0.0541} & \textbf{0.9964} \\
&EIGF & 0.1085 & 0.5927 & 0.1501 & 0.9723\\
&MASA &  0.1121 & 0.6486 & 0.1543 & 0.9710\\
&MEPE & 0.0648 & 0.4612 & 0.0855 & 0.9910 \\
&MIPT &  0.1015 & 1.0025 & 0.1390 & 0.9763\\
&SSA & 0.0731 & \textbf{0.3337} & 0.0893 & 0.9902
\end{tabular}
}
\end{center}
\caption[Error measures for FE contact problem with stiction and four-dimensional parametric space after 80 HF samples]{Error measures for FE contact problem with stiction and four-dimensional parametric space after 80 HF samples(blank rows indicate numerical problems)}\label{tab::Contactvbd4}
\end{table}

\begin{figure}[hbtp]
\centering
\includegraphics[scale=0.5]{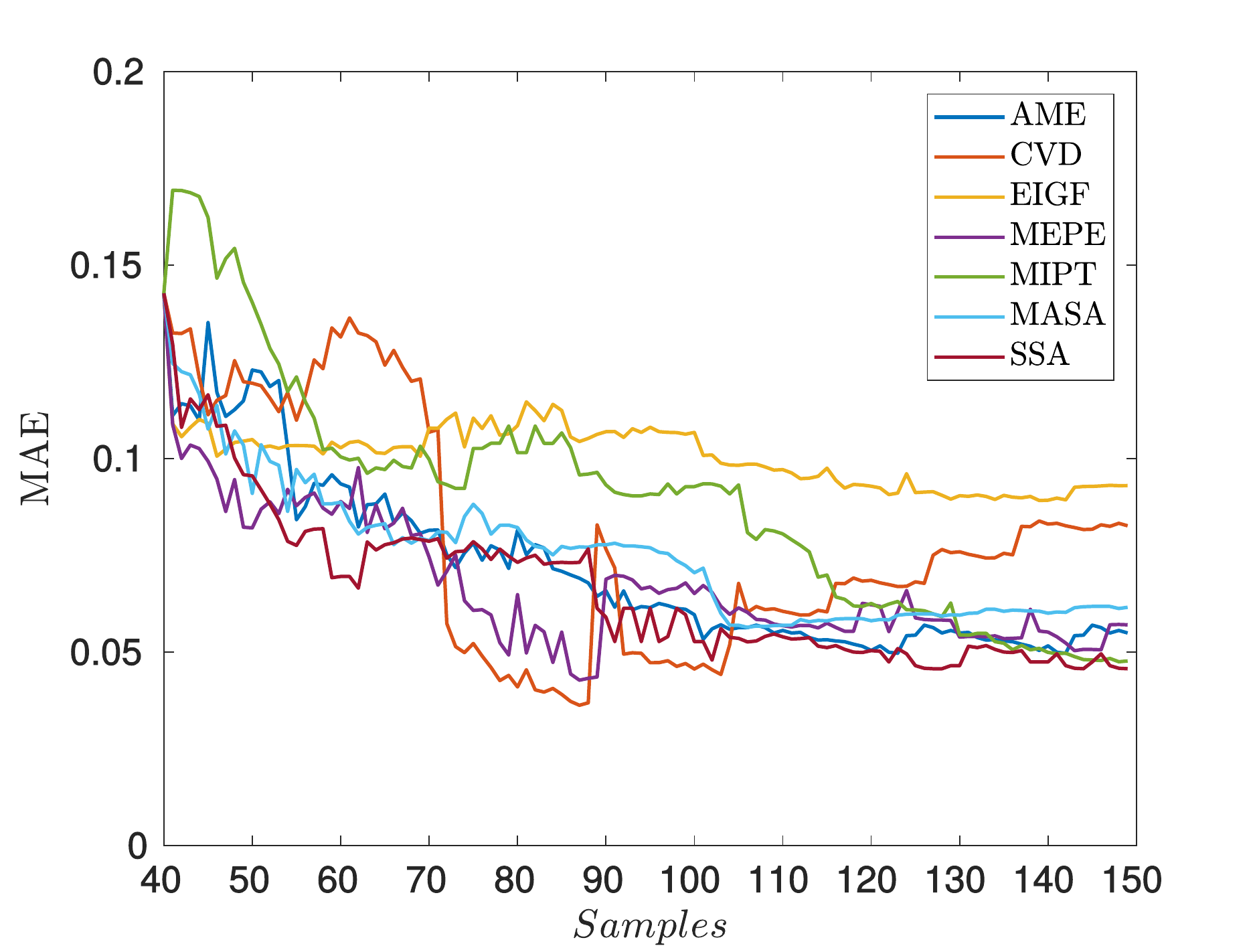}
\caption[Convergence of the MAE error for the FE-contact problem with stiction and four-dimensional parametric space]{Convergence of the MAE error for the FE-contact problem  with stiction friction and four-dimensional parametric space. }\label{fig::ConveregnceVBd4}
\end{figure}

The convergence of the MAE error is shown in Figure \ref{fig::ConveregnceVBd4}. Similarly to the results in Figure \ref{fig::ConvergenceD4COntact}, it can be seen that all methods initially drastically reduce the error measure. MIPT and SSA yield the best measure after 150 samples. 
The MAE value of CVD is increasing again due to clustering issues after 100 samples.
When comparing the two convergence behaviors of Figure \ref{fig::ConveregnceVBd4} and \ref{fig::ConvergenceD4COntact}, it can be noticed that when judging by the MAE error, adding a friction law with stiction, appears to have increased the "difficulty" of the black-box function. Since the final error values of stiction problem are higher on average. 

\subsubsection{Velocity-dependent law with six input parameters}
The four-dimensional parametric space of section \ref{sec::VelDependetLaw4d} is extended to include the Young's modulus with $E \in [9.8,10.2] \text{MPa}$ and the amplitude of the displacement boundary $\overline{u} \in [0.27,0.32]mm$. The five best performing methods of the previous investigation (CVD, EIGF, MIPT, MEPE and SSA) are considered here. \newpage
Initially, 60 HF samples and 250 LF samples are generated with TPLHD.
The error measures after 120 HF samples are shown in Table \ref{table::Contactd6}. Here no clear best performing method can be selected, since dependent on the error measure different methods perform more proficiently. Overall MEPE, SSA and MEPE show the best results. It can be seen that all methods significantly decrease the errors with respect to the initial error and that all adaptive techniques achieve more proficient error with respect to the one-shot technique. 
\begin{table}[ht!]
\begin{center}
\resizebox{1.0\textwidth}{!}{%
\begin{tabular}{l|l c c c c} \hline
& Method & MAE & RMAE & RMSE & R$^{2}$ \\ \hline\hline \\
\multirow{1}{*}{\shortstack[l]{Errors after\\60-HF samples}} & TPLHD &  0.3796 & 1.1399 & 0.4802 & 0.9077  \\\\  \hline \\
\multirow{0}{*}{\shortstack[l]{ Errors after\\120-HF samples}} &TPLHD  & 0.2962 & 1.1362 & 0.4080 & 0.9305  \\
&CVD & 0.1523 & 0.6892 & 0.1923 & 0.9789 \\
&EIGF & 0.2080 & 0.7383 & 0.2682 & 0.9699 \\
&MEPE & 0.1443 & 0.5465 & 0.1879 & 0.9852\\
&MIPT &  \textbf{0.1442} & 0.9350 & 0.2191 & 0.9799\\
&SSA & 0.1457 &\textbf{0.4716} & \textbf{0.1824} & \textbf{0.9861}
\end{tabular}
}
\end{center}
\caption[Error measures for FE contact problem with stiction and six-dimensional parametric space after 120 HF samples]{Error measures for FE contact problem with stiction and six-dimensional parametric space after 120 HF samples}\label{table::Contactd6}
\end{table}

\begin{figure}[hbtp]
\centering
\includegraphics[scale=0.5]{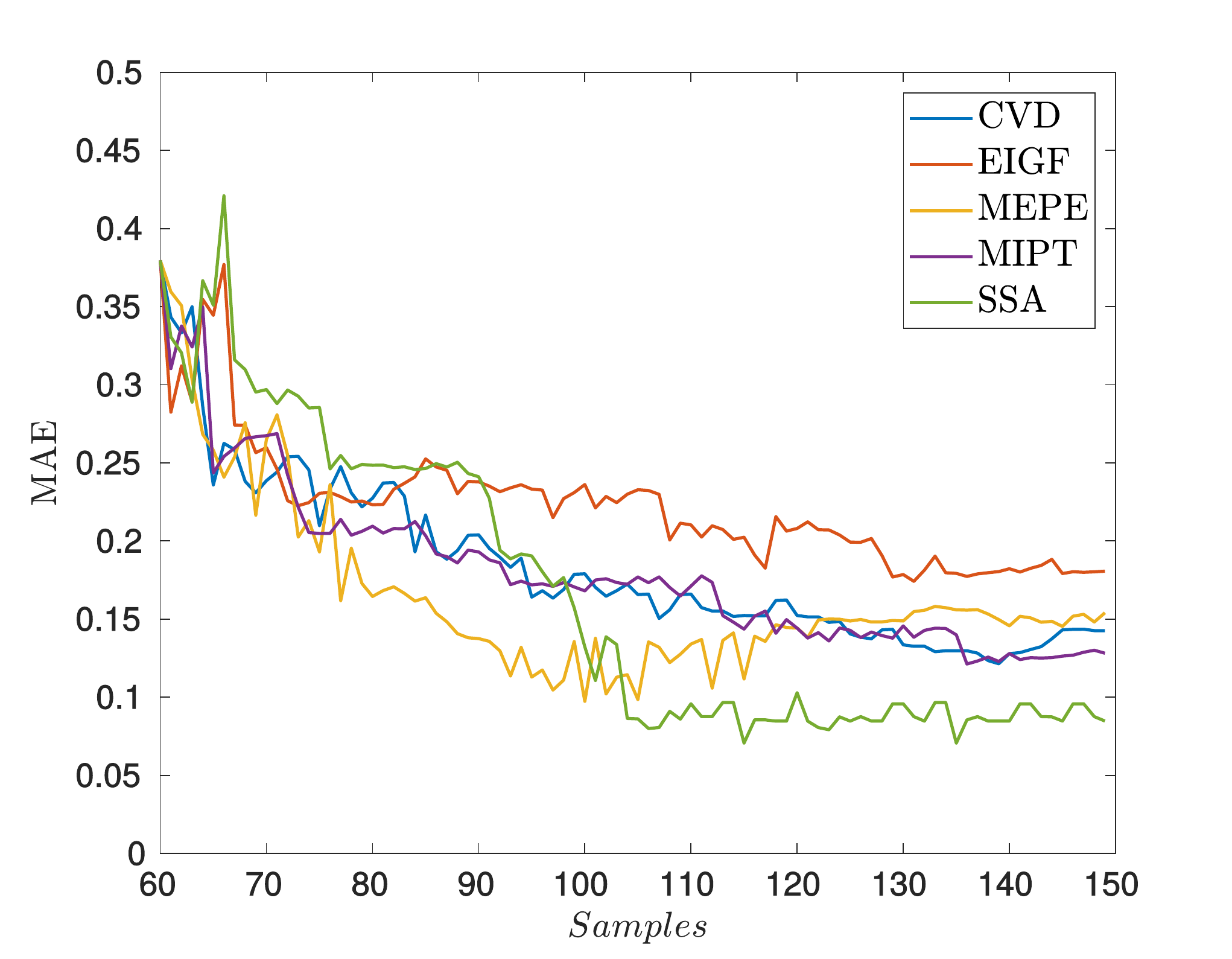}
\caption[Convergence of the MAE error for the FE-contact problem with stiction and six-dimensional parametric space]{Convergence of the MAE error for the FE-contact problem  with stiction friction and six-dimensional parametric space.}\label{fig::ConveregnceVBd6}
\end{figure}

The convergence of the MAE error of the five selected methods until 150 HF-samples is shown in Figure \ref{fig::ConveregnceVBd6}. All methods show a drastic initial decrease in error. However after around 100 samples the measure stalls. As in the previous studies SSA shows the best performance after 150 samples.

\section{Concluding remark}
Hierarchical Kriging has been herin presented as a tool to combine multi-fidelity approaches in Kriging to yield a metamodel. As a novel contribution, different adaptive sampling techniques have been compared for different benchmark problems utilizing Hierarchical Kriging. The techniques have also been successfully tested on a two-dimensional Finite-Element problem involving contact nonlinearities. Here two different friction laws have been investigated. 
\chapter{Partial-least squares Kriging}\label{ch::PLSOK}
As discussed in section \ref{sec::limitations} a limiting factor of using Kriging is the dimensionality of the inputs because of the resulting computational expenses in determining the hyperparameters with MLE in higher dimensions.  
\cite{bouhlel2016improving} suggested the use of \acrfull{pls} method in combination with Kriging to reduce the number of hyperparameters in the MLE optimization process. The author called this approach partial least squares Kriging (PLSK). Since this thesis highlights the use of OK the method will be termed \acrfull{plsok} in the following.
The idea is that the hyperparameters $\theta_{i}$ with $i=1, \ldots, n$ represent a measure of how strongly the respective inputs influence the output. The PLS technique finds a linear relationship between the inputs $\bm{X} = [\bm{x}^{(1)}, \ldots, \bm{x}^{(m)}]$ and the outputs $\bm{y} = [y^{(1)}, \ldots, y^{(m)}]^{T}$ by a projection of $\bm{X}$ onto $h$ principal components $\bm{t}^{l}$ with $l=1, \ldots h$ which is supposed to be lower than its initial dimension $m$.  Utilizing this principle \cite{bouhlel2016improving} were able to reduce the number of hyperparameters needed for Kriging and significantly speed up the MLE optimization process.
A thorough introduction to PLS is given in \cite{geladi1986partial}. The most commonly used method for calculating the principal components of a data set is the NIPALS ("nonlinear iterative partial least squares") algorithm developed by \cite{wold1973nonlinear} which is also employed in this thesis.

After employing this technique a matrix $\bm{W}_{\star} = \left[\bm{w}_{\star}^{(1)}, \ldots \bm{w}_{\star}^{(h)}  \right]$ can be defined by 
\begin{equation}
\bm{W}_{\star} = \bm{W} \left( \bm{P}^{T} \bm{W} \right)^{-1}.
\end{equation}
Here, $\bm{W}= \left[\bm{w}^{(1)}, \ldots \bm{w}^{(h)}  \right]$ contains the best directions and $\bm{P}= \left[\bm{p}^{(1)}, \ldots \bm{p}^{(h)}  \right]$ is the matrix with the coefficients  $\bm{p}^{(l)}$ that define the regression of the inputs onto the principal component $\bm{t}^{(l)}$.
The coefficients of the vectors $\bm{w}_{\star}^{l}$ measure the influence of the input on the output. These can be used to define linear transformations in the autocorrelation functions that include the PLS weights and hence reduce the dimensionality of the hyperparameters to $h$. Information for this process can be found in \cite{bouhlel2016improving}. \\
Considering the Matérn 3/2 autocorrelation matrix as used in this thesis the PLS-expanded version yields
\begin{equation}
R_{KPLS}(\bm{x},\bm{x}', \bm{\theta} ) = \prod_{l=1}^{h} \prod_{i=1}^{n} \left( 1 + \frac{\sqrt{3} \bm{m}_{i}^{(l)}}{\theta_{l}}  \right) \exp  \left( \frac{\sqrt{3} \bm{m}_{i}^{(l)}}{l_{l}}  \right) 
\end{equation}
with $\bm{m}_{i}^{(l)} = \abs{\bm{w}_{\star \,i}^{(l)} (\bm{x}^{(i)} - \bm{x}^{(i)})' }$. 
\\
In a novel approach this thesis applies the PLS model reduction of the hyperparameters to HK. In the following partial least squares hierarchical Kriging will be shortened to \acrfull{plshk}. 
A significant limitation of HK is the need of hyperparameter optimization for each fidelity level. 
Here, PLSHK enables a significant time reduction.
\section{Ten-dimensional Wong function}
\begin{figure}[h!]
\centering
\includegraphics[scale=0.5]{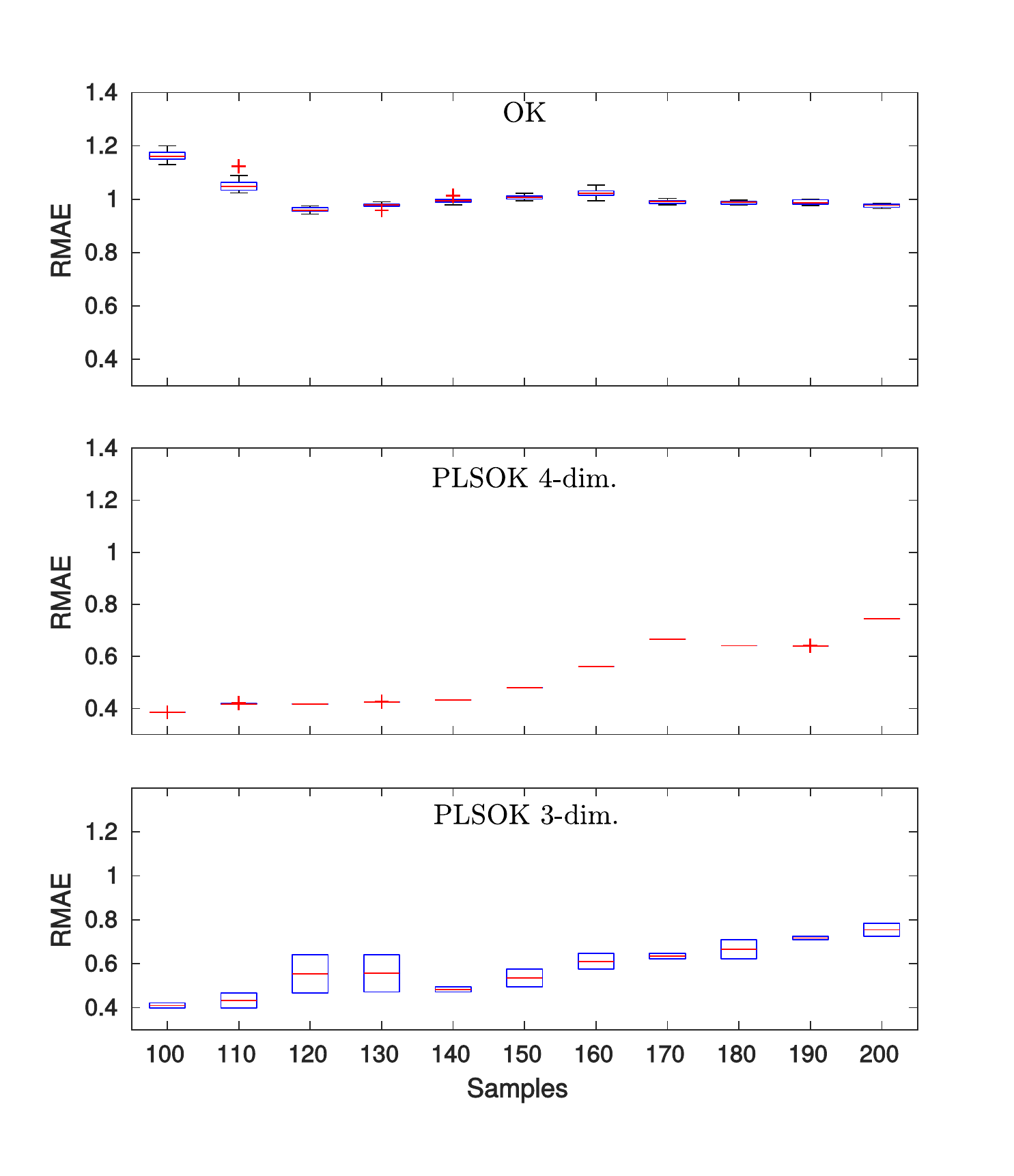} 
\caption[RMAE comparison between OK and PLSOK for $\mathcal{M}_{Wong}^{10d}$]{RMAE comparison between OK and PLSOK for $\mathcal{M}_{Wong}^{10d}$ using samples created with TPLHD and reduced dimension of the hyperparameters to 3 and 4 respectively.}\label{fig:WongRMAE}
\end{figure}

\begin{figure}[t!]
\centering
\includegraphics[scale=0.5]{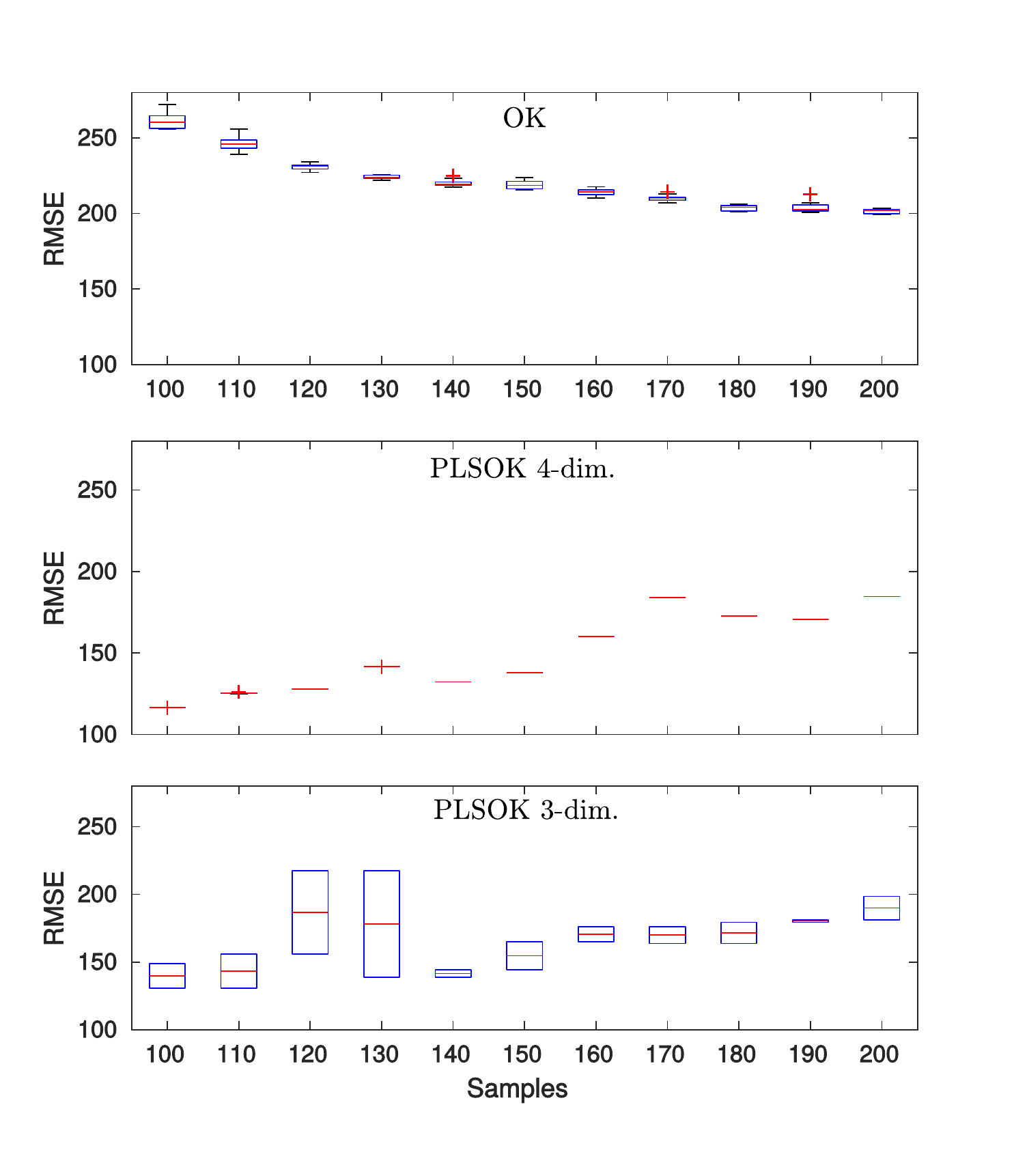} 
\caption[RMSE comparison between OK and PLSOK for $\mathcal{M}_{Wong}^{10d}$]{RMSE comparison between OK and PLSOK for $\mathcal{M}_{Wong}^{10d}$ at samples created with TPLHD and reduced dimension of the hyperparameters to 3 and 4 respectively.}\label{fig:WongRMSE}
\end{figure}

A ten-dimensional example that is used to validate adaptive sampling in combination with PLSOK is the Wong function proposed by \cite{wong1970decentralised}, later used by \cite{asaadi1973computational} and for example
\cite{michalewicz1996evolutionary}. The function reads
\begin{equation}
\begin{aligned}
\mathcal{M}_{Wong}^{10d}(\bm{x}) &= x_{1}^{2} + x_{2} + x_{1}x_{2} - 14 x_{1} - 16 x_{2} \\&+ (x_{3}-10)^{2} + (x_{4}-5)^{2} + (x_{5}-3)^{2} + 2(x_{6}-1)^{2} \\&+ 5 x_{7}^{2} + 7(x_{8}-11)^{2} + 2(x_{9}-10)^{2} + (x_{10}-7)^{2} + 45.
\end{aligned}
\end{equation}
Here, let all $x_{i}$ with $i=1, \ldots, 10$ be defined in the domain $\left[-10, 10 \right]$.
\begin{figure}[h!]
\centering
\includegraphics[scale=0.5]{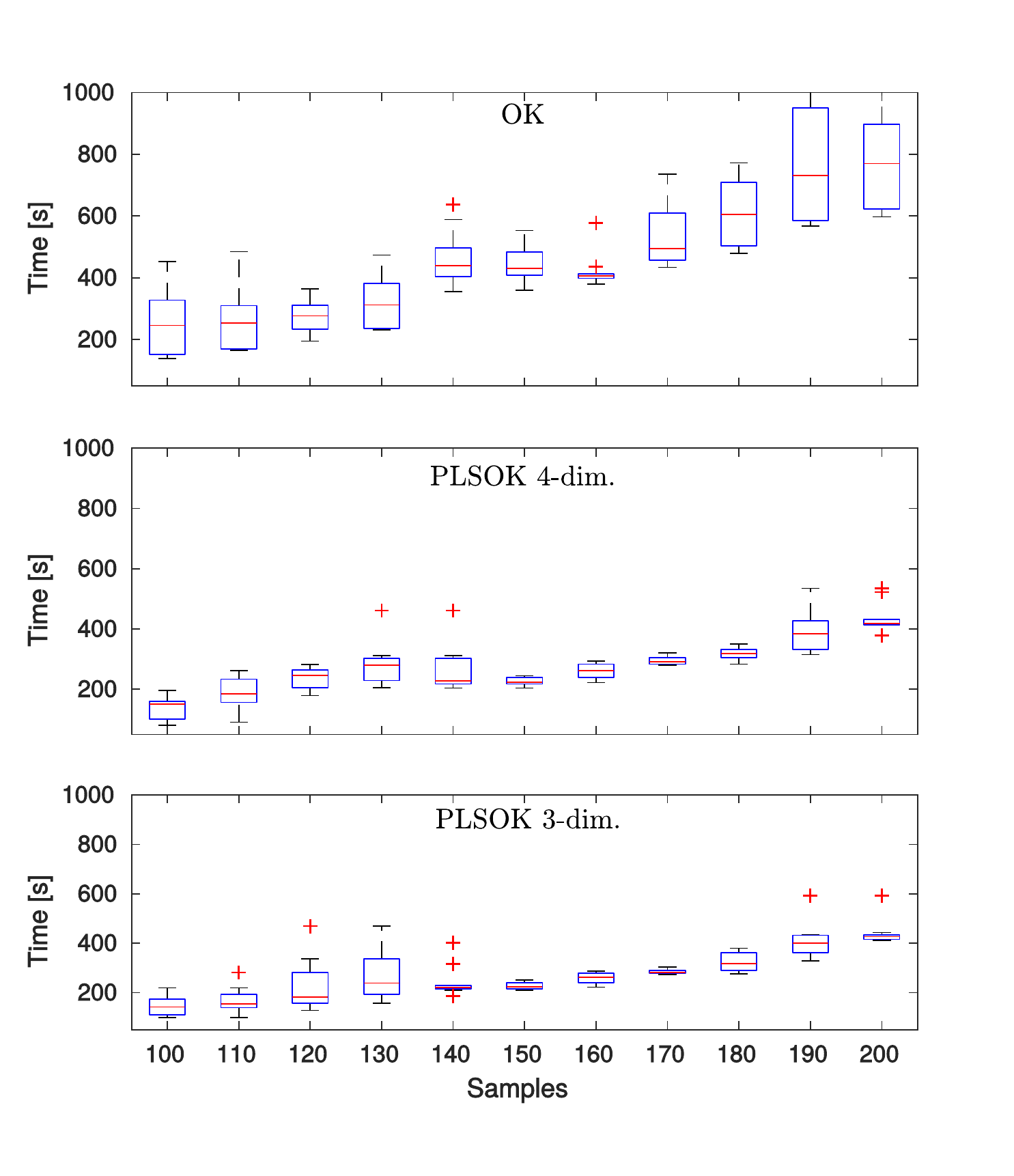} 
\caption[Time comparison between OK and PLSOK for $\mathcal{M}_{Wong}^{10d}$]{Time comparison between OK and PLSOK for $\mathcal{M}_{Wong}^{10d}$ at samples created with TPLHD and reduced dimension of the hyperparameters to 3 and 4 respectively.}\label{fig:WongTime}
\end{figure}
According to \cite{asaadi1973computational} the function has its minimum at $\mathcal{M}(\bm{x}_{min}) = 24.31 $ at $\bm{x}_{min} = (2.17, 2.36, 8.77,
5.09, 0.99, 1.43, 1.32, 9.82, 8.27, 8.37)$. The viability of the model reduction of the hyperparameters in Kriging can be seen in Figures \ref{fig:WongTime}, \ref{fig:WongRMAE} and \ref{fig:WongRMSE}. 
In each of these figures one relevant parameter is compared for Kriging models created with TPLHD. 
The parameters of interest for this study are the computation time, RMSE and RMAE. The process is as follows: every 10 steps (starting from 100) the required amount of samples are generated with TPLHD and the parameter is evaluated up until 200 samples. This procedure is evaluated 10 times in order to avoid random outliers. The parameters are shown in boxplot format over the 10 iterations. \\
The boxplot for the creation time of the metamodel for OK and PLSOK with 3 and 4 dimensions respectively is shown in  Figure \ref{fig:WongTime}. 
\newpage
The Kriging MLE for the hyperparameters is optimized with a simulated annealing algorithm and a fixed function tolerance of $10^{(-8)}$ that needs to be held over 5000 iterations. The time is shown in seconds. The computations are carried out on the cluster system at the Leibniz University of Hannover, Germany, to circumvent local computation issues. It can be seen that with increasing number of samples the computation time increases for all of the three Kriging methods. However the evaluation time of the standard OK method is on average considerably higher than the ones of PLSOK. Furthermore there seems to be a higher variance involved in the computational time for OK in comparison to the other two. This hints towards that the optimization problem for the OK model is more complex than the other two and hence there is more variation involved in the optimization algorithm until the needed function tolerance is found. This must be due to the higher dimensionality of the optimization problem for OK. When comparing the two PLSOK methods it can be seen that the lower dimensionality needs less computation time on average for all samples and the variance is also lower. However the differences are not stark. 
The results of the error indicator RMAE between the three methods are depicted in Figure \ref{fig:WongRMAE}. 
\newpage
It can be seen that the largest variation of the error measure is found for the 3-dimensional PLSOK model. OK shows less variation and the PLSOK for 4 dimensions is basically free of variance. Surprisingly it can be seen that the PLSOK models generate a metamodel with less RMAE error which indicates that the hyperparameter optimization was solved more proficiently, i.e. a better optimum was found. This hints that the used optimization scheme (Simulated annealing) is not well suited for higher dimensionality. Since the general OK model when solved correctly should be more accurate than the models with dimensionality reduction of the hyperparameters. This fact again hints towards an easier optimization problem in lower dimensionality. \\
It can be seen that whereas the error measure stays basically flat with increasing samples the PLSOK errors increase. This might be explained with the non-optimality of the TPLHD samples in higher dimensions as discussed in section \ref{page::TPLHD}. 
Furthermore it seems the model reduction works more proficient for a smaller sample size. However since the RMAE measure is still lower for PLSOK it is still a valid method. 
When comparing the resulting PLSOK measures the 4-dimensional model shows less variation and a lower error value overall and must therefore be the preferred choice here. 

The RMSE error for all three models over the sample size is shown in Figure \ref{fig:WongRMSE}. A similar effect to RMAE is visible where with increasing sample size the RMSE value for OK reduces wheres an increase is shown for PLSOK. Here, again, the PLSOK metamodels indicate a lower RMSE over all samples which hints at a better approximation of the surrogate model. The 3-dimensional version shows more variation in this parameter of interest than the other two, where the hyperparameter model reduction to 4 dimensions has almost no variation which hints at a straightforward optimization problem. \\
Considering all three of the described graphs it can be seen that PLSOK can be validated as an efficient method for Kriging metamodel generation. Due to the lower dimensionality of the optimization problem even better results are obtained. This opens PLSOK up for the use in adaptive sampling with introduced techniques for the Wong function. \\ 
The adaptive sampling is here performed with a model reduction to 4 hyperparameter dimensions. \\
For the investigation of the adaptive sampling techniques an initial sample size of 100 samples is chosen.
Because of the high-dimensionality of the problem the sampling techniques that require higher computational effort are only used to add 50 more samples from an initial 100 samples. The more time-efficient methods (EI, EIGF, MEPE, MIPT and MSD) are calculated ten times until 300 samples to yield the average error measures. \\
The errors after adding 50 samples to the initial sample size are listed in Table \ref{table::Wong150}. The values can be compared to a metamodel created with the one-shot TPLHD method. As observed in the comparison of the results for OK and PLSOK the increased TPLHD metamodel yields worse results than the initial model.
However all the adaptive sampling techniques are able to reduce the values significantly with respect to the one-shot methods. As noticed in earlier problems EI performs worst. After 150 samples EIGF reduces the errors most proficiently.
\begin{table}[h!]
\begin{center}
\resizebox{1.0\textwidth}{!}{%
\begin{tabular}{l|l c c c c} \hline
& Method & MAE & RMAE & RMSE & R$^{2}$  \\ \hline\hline \\
\multirow{1}{*}{\shortstack[l]{Errors after\\100 samples}} & TPLHD & 94.4414 & 0.3851 & 116.4710 & 0.9895 \\ \\  \hline \\
\multirow{14}{*}{\shortstack[l]{ Errors after\\150 samples}} &TPLHD &109.5913 & 0.47935 & 137.8352 & 0.9854\\
&ACE & 32.6919 & 1.1659 & 69.4532 & 0.9771 \\
&CVVor & 46.2332 & 0.2763 & 42.2743 & 0.9801   \\ 
&EI &  95.0939 & 0.3427 & 118.5459 & 0.9892 \\
&EIGF &  \textbf{26.4332} & \textbf{0.1249} & \textbf{33.6435} & \textbf{0.9991}\\
&MASA &  60.2378 & 0.2818 & 62.6726 & 0.9899  \\
&MEPE &  40.2932 & 0.2028 & 50.2783 & 0.9980\\
&MIPT &  68.8788 & 0.2740 & 83.4779 & 0.9946 \\ 
&MSD & 31.7074 & 0.1853 & 40.5203 & 0.9987 \\
&MSE & 46.7074 & 0.2839 & 56.2367 & 0.9945  \\
&SFCVT &41.2481 & 0.2148 & 51.9043 & 0.9979 \\ 
&SSA &  38.5541 & 0.1425 & 47.9274 & 0.9982
\end{tabular}
}
\end{center}
\caption[Error measures for $\mathcal{M}_{Wong}^{10d}$ after 150 samples.]{Error measures for $\mathcal{M}_{Wong}^{10d}$ after 150 samples. }\label{table::Wong150}
\end{table}

The results for the five time-efficient techniques after 300 samples are listed in Table \ref{table::Wong300}. When comparing the values to Table \ref{table::Wong150} a further significant decrease can be noticed. EI yields by far the worst result. MEPE and EIGF perform most proficiently. The space-filling techniques (MIPT and MSD) achieve similar values.
\begin{table}[h!]
\begin{center}
\resizebox{1.0\textwidth}{!}{%
\begin{tabular}{l|l c c c c} \hline
& Method & MAE & RMAE & RMSE & R$^{2}$ \\ \hline\hline \\ \\  \hline \\
\multirow{1}{*}{\shortstack[l]{Errors after\\100 samples}} & TPLHD & 94.4414 & 0.3851 & 116.4710 & 0.9895 \\
\multirow{6}{*}{\shortstack[l]{ Errors after\\30 0 samples}} &TPLHD &35.0384 &0.3356 & 41.4699 & 0.9922 \\
&EI &  62.4272 & 0.3143 & 90.7335 & 0.9883   \\
&EIGF &  16.5853 & \textbf{0.0772} & 25.2345& \textbf{0.9984}  \\
&MEPE &  \textbf{13.3693}& 0.1043 & \textbf{20.7335} & 0.9978  \\
&MIPT & 26.0037& 0.1923 & 34.8743 & 0.9934 \\ 
&MSD & 22.8202 & 0.1734 & 32.2342 & 0.9911  \\
\end{tabular}
}
\end{center}
\caption[Error measures for $\mathcal{M}_{Wong}^{10d}$ after 300 samples.]{Error measures for $\mathcal{M}_{Wong}^{10d}$ after 300 samples. }\label{table::Wong300}
\end{table}

\begin{figure}[h!]
\centering
\includegraphics[scale=0.5]{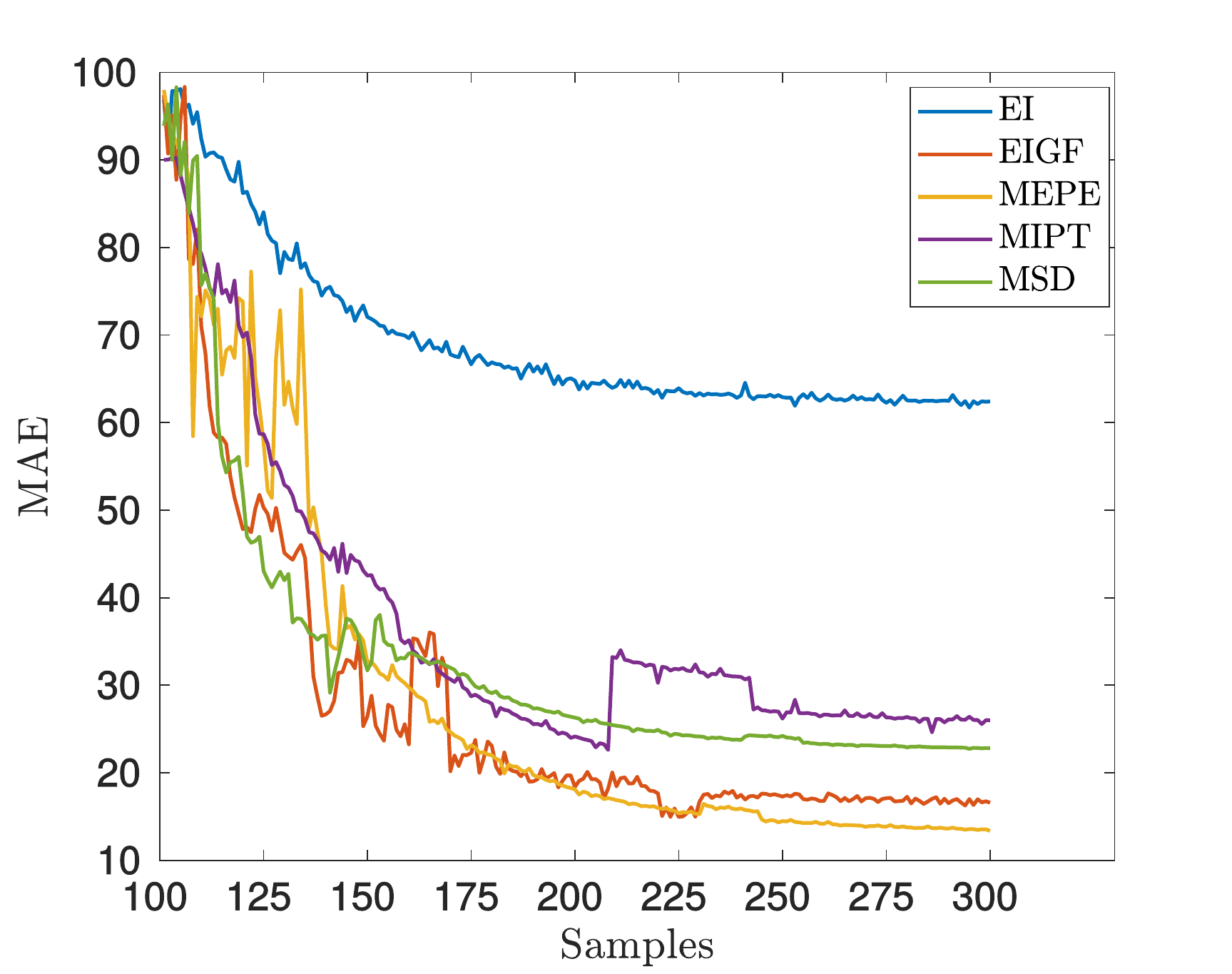}
\caption[MAE measure for$\mathcal{M}_{Wong}^{10d}$ over the sample size]{MAE measure for$\mathcal{M}_{Wong}^{10d}$ over the sample size for 5 different adaptive sampling techniques.}\label{fig::WongConvergence}
\end{figure}

Figure \ref{fig::WongConvergence} displays the convergence of the MAE error with increasing sample size for the 5 investigated techniques. All methods show a decrease in the error value. However after around 200 samples the values is stagnating.
In the following section a benchmark problem for PLSHK is investigated.
\section{Eight-dimensional Application for PLSHK: Borehole function}
As a benchmark application for PLSHK the borehole function e.g. given in \cite{harper1983sensitivity} is considered. The function is used to model water flow through a hole.
It is a usual application problem in multifidelity approaches in the literature see e.g. \cite{xiong2013sequential} or \cite{moon2012two}.
The low-fidelity version reads
\begin{equation}
\mathcal{M}_{borehole, HF}^{8d}(\bm{x}) = \dfrac{1}{\text{ln}(r/r_{w})} \dfrac{2 \pi T_{u} \left( H_{u} - H_{l} \right)}{1 + \frac{2 L T_{u}}{\text{ln}(r/r_{w}) r_{w}^{2} K_{w}} + \frac{T_{u}}{T_{l}}}
\end{equation}
and the low-accuracy model is given by
\begin{equation}
\mathcal{M}_{borehole, LF}^{8d}(\bm{x}) = \dfrac{1}{\text{ln}(r/r_{w})} \dfrac{ 5 T_{u} \left( H_{u} - H_{l} \right)}{1.5 + \frac{2 L T_{u}}{\text{ln}(r/r_{w}) r_{w}^{2} K_{w}} + \frac{T_{u}}{T_{l}}}.
\end{equation}
The difference between the two fidelity levels is captured in the following error between them calculated from 40000 points: $MAE$:$15.8715$, $RMAE$: $1.1340$, and $RMSE$: $18.3973$. Hence the LF function distorts the frame of the HF version.
The physical interpretation of the parameters as well as the input domain for the creation of the metamodel is given in Table \ref{tab::boreholeParameter}.
\begin{table}[h!]
\begin{center}
\begin{tabular}{l cr} \hline
Parameter  & Domain  \\ \hline\hline \\
$r_{w}$ - Radius of the borehole    & $\left[ 0.05, 0.15 \right] [m]$  \\
$r$ - Radius of influence  & $\left[ 100, 50000 \right] [m]$   \\ 
$T_{u}$ - Transmissivity of upper aquifer    & $\left[ 63070, 115600 \right]  [m^{2}/year] $ \\ 
$H_{u}$ - Potentiometric head of upper aquifer  & $\left[ 990, 1100 \right][m]$   \\
$T_{l}$ - Transmissivity of lower aquifer   & $\left[ 63.1,116 \right]  [m^{2}/year]$ \\
$H_{l}$ - Potentiometric head of lower aquifer  & $\left[ 700, 820 \right][m]$ \\
$L$ - Length of borehole & $\left[ 1120, 1680 \right][m]$  \\  $K_{w}$ - Hydraulic conductivity of borehole   &  $\left[ 9855, 12045 \right][m/year]$   
\end{tabular}
\end{center}
\caption[Parameter domains for borehole functions]{Parameter domains for borehole functions: $\mathcal{M}_{borehole, HF}^{8d}$ and $\mathcal{M}_{borehole, LF}^{8d}$}\label{tab::boreholeParameter}
\end{table}

Adaptive sampling for 8 dimensions as given here is a computationally expensive procedure. Furthermore the defined space of the hyperparameter optimization problem is also 8-dimensional and hence complicated to solve. As introduced PLSHK can be an option for model reduction of the hyperparameter space. In contrast to PLSOK the multifidelity version PLSHK has different levels of fidelity for which a Kriging metamodel needs to be created or at least the MLE for the hyperparameters needs to be maximized. This implies that each fidelity level can have its own hyperparameter dimensionality. This concept would require further investigation. Here, for validation of PLSHK, similar to the benchmark problem for PLSOK the parameters Time, RMSE and RMAE are compared for metamodels created for samples which were generated by TPLHD. This is done by starting with 100 samples and increasing the sample size by 10 up until 200 samples are reached.
To show the variation of the parameters 10 iterations of the computations are done and the values are then illustrated with boxplots. The comparison is done between a general HK-model and two PLSHK models which are reduced to 3 and 4 dimensions in both fidelity levels respectively. \\
\begin{figure}[h!]
\centering
\includegraphics[scale=0.6]{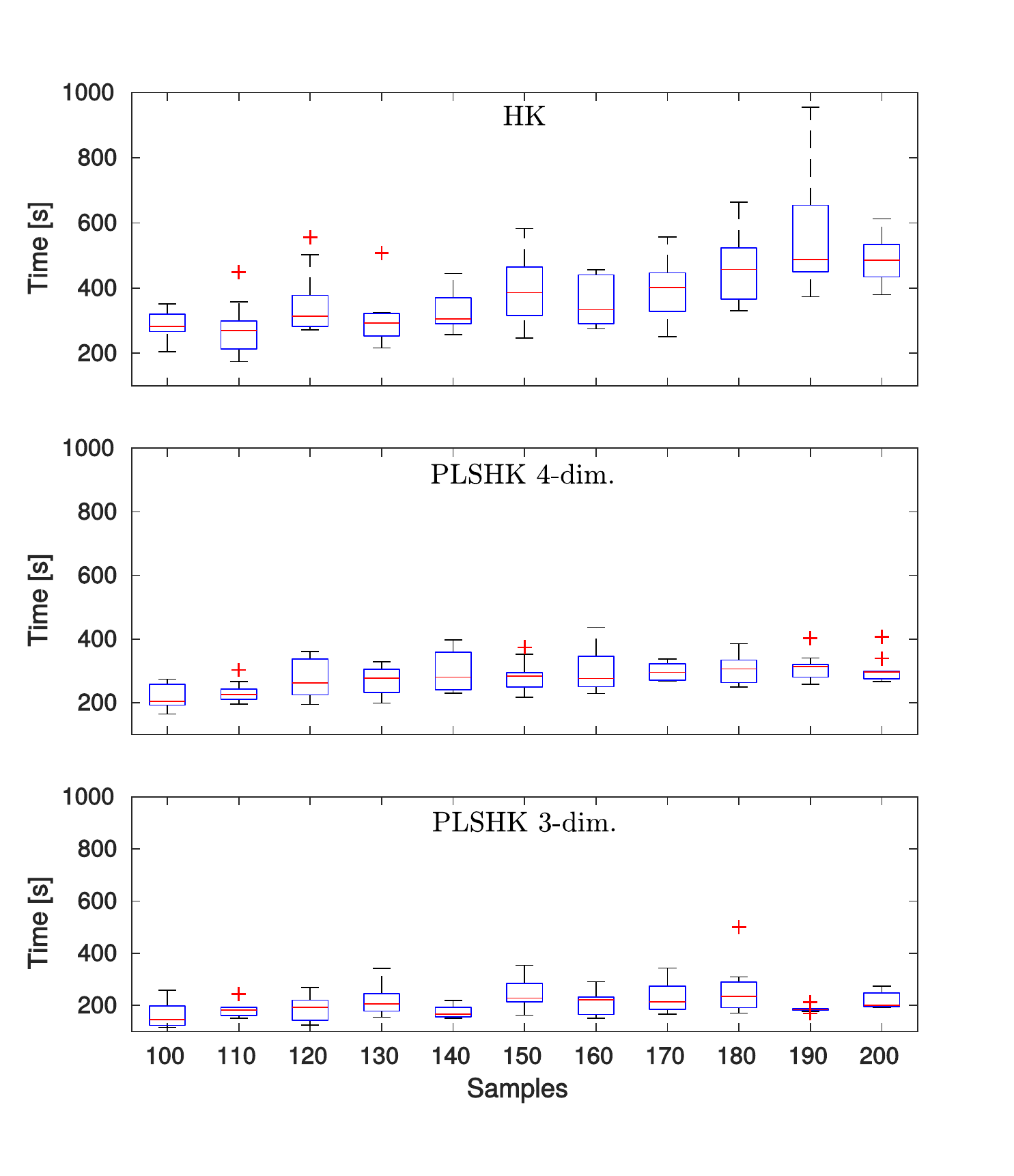}
\caption[Time comparison between HK and PLSHK for $\mathcal{M}_{borehole, HF}^{8d}$]{Time comparison between HK and PLSHK for $\mathcal{M}_{borehole, HF}^{8d}$ at samples created with TPLHD and reduced dimension of the hyperparameters to 3 and 4 in both fidelity-levels respectively. }\label{fig::PLSHKTime}
\end{figure}

\begin{figure}[h!]
\centering
\includegraphics[scale=0.6]{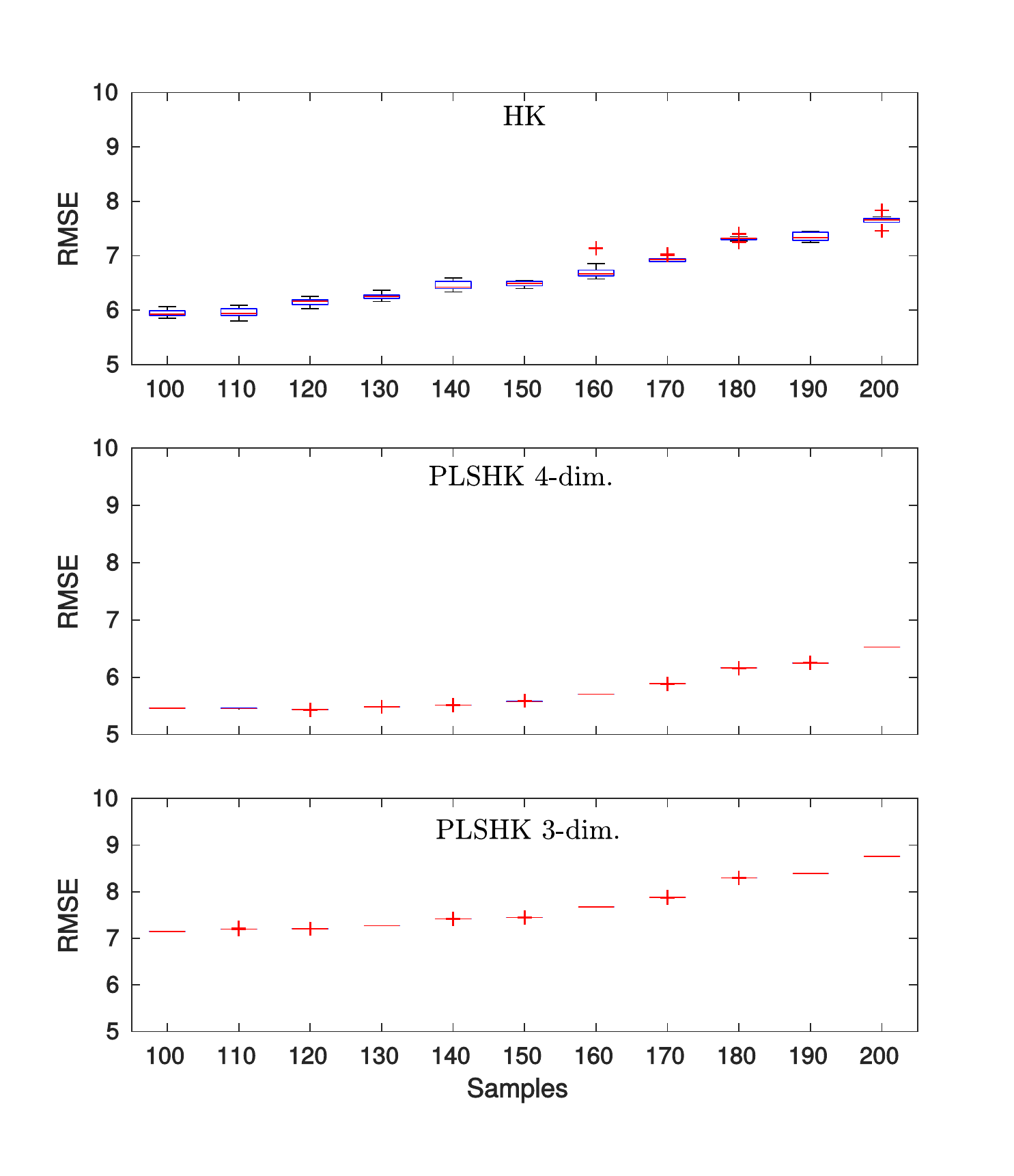}
\caption[RMSE comparison between HK and PLSHK for $\mathcal{M}_{borehole, HF}^{8d}$]{RMSE comparison between HK and PLSHK for t$\mathcal{M}_{borehole, HF}^{8d}$ at samples created with TPLHD and reduced dimension of the hyperparameters to 3 and 4 in both fidelity-levels respectively.}\label{fig::PLSHKRMSE}
\end{figure}

Furthermore the same restrictions as for PLSOK for the Wong function are placed with simulated annealing algorithm used as an approximation method with a fixed function tolerance of $10^{(-8)}$ that needs to be held over 5000 iterations. The computations are again carried out on the cluster system at the Leibniz University of Hannover, Germany, to circumvent local computation issues. \\
The time of metamodel generation of the 3 considered models for each sample size step is displayed in Figure \ref{fig::PLSHKTime}. It can be seen that HK needs the most time on average and shows a higher variation than the other two models.  Predictably, the 3-dimensional version of PLSHK needs less time than the four dimensional version. 
The RMSE errors between the 3 models are compared in Figure \ref{fig::PLSHKRMSE}. Similar to the Wong function it can be seen that with an increase in sample size the error measure actually increases which hints at either bad sample placement or problems with the optimization of the hyperparameter optimization problem with e.g. numerical issues. However is is still valuable to compare the obtained values. 

\begin{figure}[hbtp]
\centering
\includegraphics[scale=0.5]{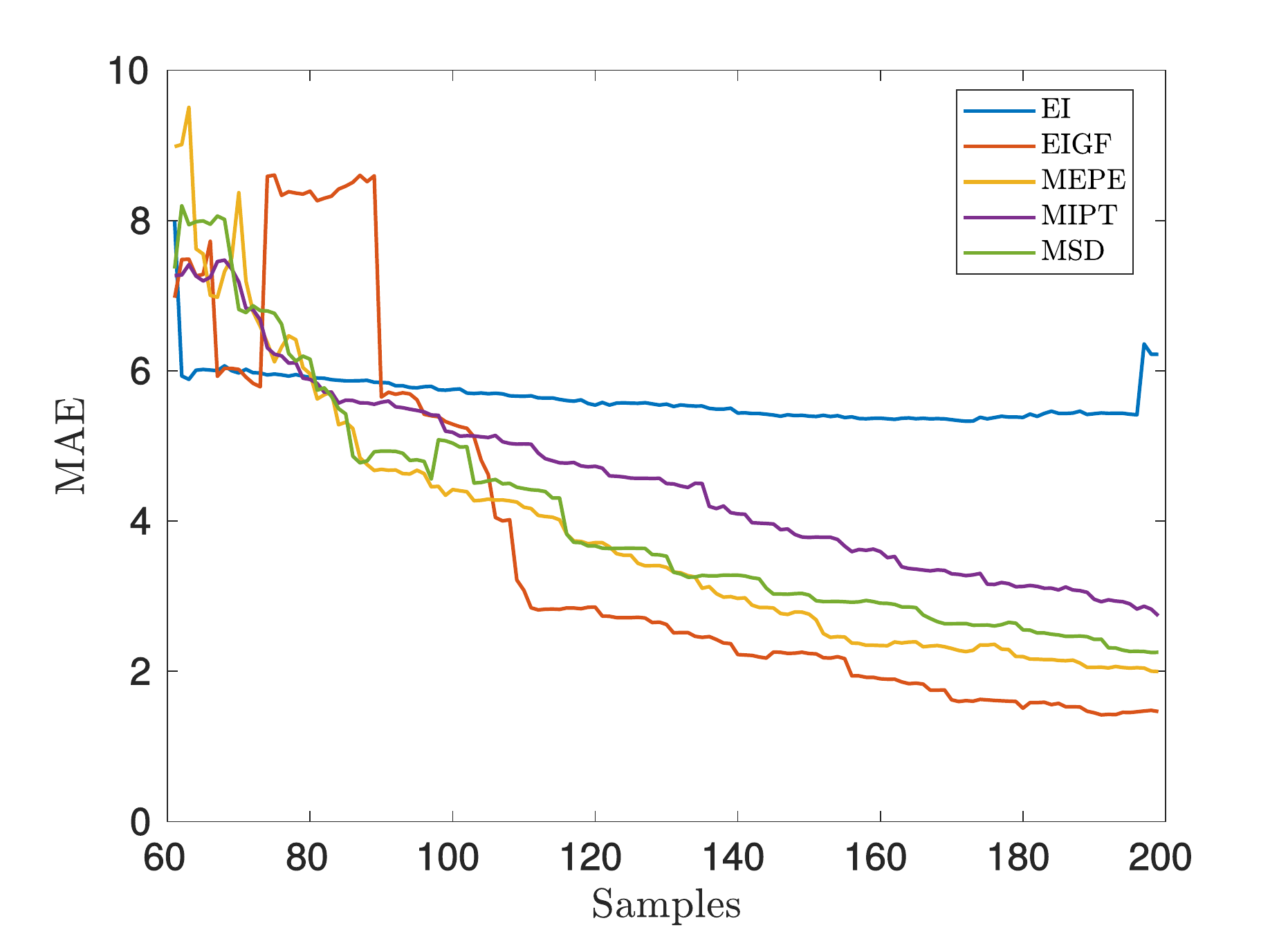}
\caption[Convergence of the MAE error for $\mathcal{M}_{borehole, HF}^{8d}$ ]{Convergence of the MAE error for $\mathcal{M}_{borehole, HF}^{8d}$ with increasing samples size for the 5 respective sampling techniques.}\label{fig::BoreholdConvergence}
\end{figure}

It can be seen that the 3-dimensional PLSHK version creates the highest measure error in comparison to the other two. The 4 dimensional model however is on average lower than the general HK and shows less variation in its values. This again hints that the used optimization tool is not proficient enough for higher (in this case 8) dimensions as the optimal solution for the hyperparameters is not found.
\begin{figure}[h!]
\centering
\includegraphics[scale=0.6]{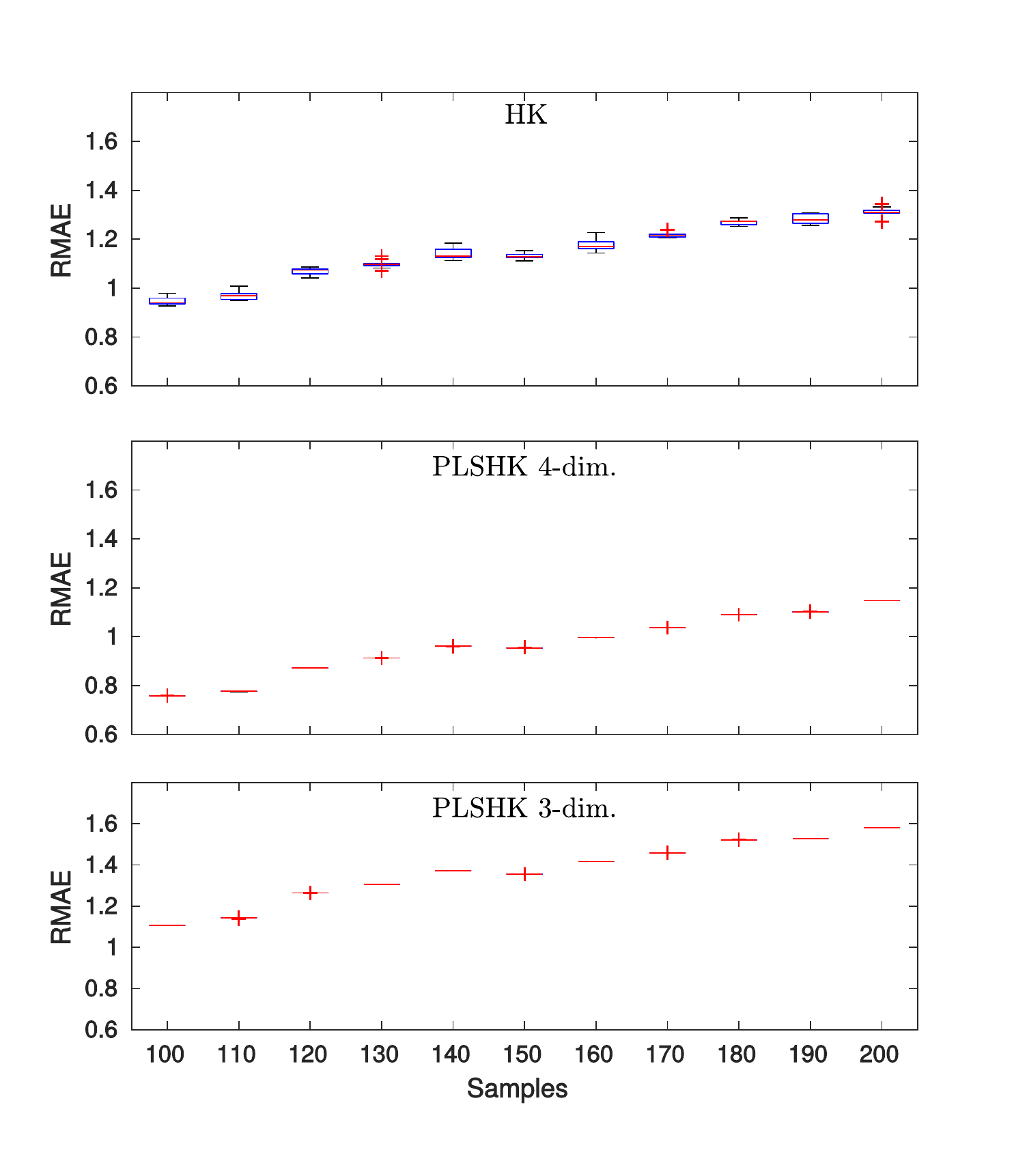}
\caption[RMAE comparison between HK and PLSHK for $\mathcal{M}_{borehole, HF}^{8d}$]{RMAE comparison between HK and PLSHK for t$\mathcal{M}_{borehole, HF}^{8d}$ at samples created with TPLHD and reduced dimension of the hyperparameters to 3 and 4 in both fidelity-levels respectively.}\label{fig::PLSHKRMAE}
\end{figure}
The RMAE measure of the 3 methods is displayed in Figure \ref{fig::PLSHKRMAE}. It shows similar results to RMSE, i.e. that the 4 dimensional PLSHK is able to obtain the best surrogate model with the least variation. 
\newpage
Therefore PLSHK is a valid method in comparison to HK for the borehole function. Furthermore the four-dimensional version needs to be the preferred choice because all studied parameter values are significantly reduced with respect to HK. 
The following adaptive sampling computations were performed with 60 HF samples and 250 LF samples initially.
Similar to the Wong function an adaptive sampling to higher sample sizes ($>150$) is only performed with 5 techniques that are most time-efficient (EI, EIGF, MEPE, MIPT and MSD). The other methods are compared after adding 60 HF-samples. Table \ref{table::Borehole120} shows a comparison of the error measure values after adding 60 points. It can be noticed that all methods significantly decrease the initial error. Judging by MAE only 2 (EI and MASA) out of the 11 methods show a worse results than the one-shot TPLHD. SSA and EIGF perform the best.
\begin{table}[h!]
\begin{center}
\resizebox{1.0\textwidth}{!}{%
\begin{tabular}{l|l c c c c} \hline
 & Method & MAE & RMAE & RMSE & R$^{2}$  \\ \hline\hline \\
\multirow{1}{*}{\shortstack[l]{Errors after\\60-HF samples}} & TPLHD  & 7.3548& 1.0219 & 9.8656 &0.95313  \\ \\  \hline \\
\multirow{10}{*}{\shortstack[l]{ Errors after\\120-HF samples}} &TPLHD &4.1312 & 0.8717 & 5.4317 & 0.9857\\
&ACE &  3.2628 & 0.5508 & 4.1053 & 0.9862\\
&CVD & 3.9450 &	0.6598 & 4.9965 & 0.9831  \\
&CVVor &  3.9901 & 1.2409 & 5.7568 & 0.9863\\
&EI &  5.5435 & 1.2943 & 7.6987 & 0.9717\\
&EIGF & 2.8550 & 0.4965 & 3.7218 & 0.9933\\
&MASA &  4.8223 & 1.1234 & 6.5625 & 0.9734 \\
&MEPE &  3.7130 & 0.5928 & 5.1018 & 0.9875\\
&MIPT &  4.7297 & 1.2173 & 6.4905 & 0.9798\\ 
&MSD & 3.6700 & 0.4447 & 4.7714 & 0.9891\\
&SFCVT &3.1776 & 0.8280 & 4.4055 & 0.9907 \\ 
&SSA &  \textbf{2.2701} & \textbf{0.3738 }& \textbf{2.9714} & \textbf{0.9957}
\end{tabular}
}
\end{center}
\caption[Error measures $\mathcal{M}_{borehole, HF}^{8d}$ after 120 HF-samples]{Error measures $\mathcal{M}_{borehole, HF}^{8d}$ after 120 HF- and 250 LF-samples. }\label{table::Borehole120}
\end{table}
\begin{table}[h!]
\begin{center}
\resizebox{1.0\textwidth}{!}{%
\begin{tabular}{l|l c c c c} \hline
& Method & MAE & RMAE & RMSE & R$^{2}$  \\ \hline\hline \\
\multirow{1}{*}{\shortstack[l]{Errors after\\60-HF samples}} & TPLHD &7.3548& 1.0219 & 9.8656 &0.95313 \\\\  \hline \\
\multirow{6}{*}{\shortstack[l]{ Errors after\\200-HF samples}} &TPLHD &4.2018 & 1.1472 & 6.5265 & 0.9794 \\
&EI &   6.2193 & 1.5798 & 9.0259 & 0.9611\\
&EIGF & \textbf{1.4656} &\textbf{0.1858 }&\textbf{1.8702} & \textbf{0.9983}\\
&MEPE &  1.9984 & 0.3717 & 2.7079 & 0.9965\\
&MIPT &  2.7403 & 0.7270 & 3.8755 & 0.9928 \\ 
&MSD & 2.2533 & 0.3537 & 3.0203 & 0.9956\\
\end{tabular}
}
\end{center}
\caption[Error measures $\mathcal{M}_{borehole, HF}^{8d}$ after 200 HF-samples]{Error measures $\mathcal{M}_{borehole, HF}^{8d}$ after 200 HF- and 260 LF-samples.}\label{table::Borehole200}
\end{table}
The results after 200 HF-samples for the 5 techniques that require the least computational effort are listed in Table \ref{table::Borehole200}. Here, all methods except EI perform better than TPLHD. EIGF yields the most proficient metamodel, by e.g. reducing the initial MAE error by factor 5.\\ It can be observed that contrary to the pure TPLHD approach the adaptive sampling methods are reducing the error with increasing sample size.
The MAE error over the added samples (from 60 to 200) is depicted in Figure \ref{fig::BoreholdConvergence} displays. The EI method shows clustering of the sample points since the error is not decreasing or even increasing. All other methods show a decreasing behavior with occasional spikes.

\section{Concluding remark}
The partial least squares method has been utilized in the generation of the Ordinary Kriging and in a novelty approach in Hierarchical Kriging in order to reduce the number of hyperparameters. It was found that the method is able to speed up the computations of Kriging in higher dimensions. Furthermore due to the simplification of the hyperparameter optimization problem the results are more proficient when using this technique. Adaptive and sequential sampling techniques have been compared utilizing this approach. Four out of five were found to be proficient when compared to one-shot techniques.

\chapter{Conclusion and outlook}\label{chapter::ConclusionandOutlook}
Concepts of the Kriging framework and adaptive sampling have been presented and studied for the use of global metamodeling in computer experiments.
An overview of the work, a conclusive evaluation of the investigated and newly introduced adaptive sampling techniques, and a perspective for future research are given in this chapter.
\paragraph*{Overview of the work}
An overview over the general concepts of global surrogate modeling, specifically Kriging, with adaptive sampling techniques and restriction to univariate output has been provided. Two different formulations of the Kriging method in Universal and Ordinary Kriging (OK) have been presented and compared. It has been observed that OK offers more consistent metamodeling results. Occurring challenges and limitations of OK have been stated and a literature review for the solutions to these problems has been performed. 
It has been found that Ordinary Kriging performs the approximation of a one-dimensional function more proficiently than other surrogate modeling techniques found in the literature. \\
A multifidelity variant of Kriging called Hierarchical Kriging (HK) has been studied. A solution for the use of OK in high-dimensional problems by reduction of the hyperparameter space has been investigated (here termed partial least squares ordinary Kriging (PLSOK)). 
In a novelty approach this thesis has also established the hyperparameter reduction for Hierarchical Kriging (PLSHK). \\
The concepts of adaptive sampling for global surrogate modeling have been discussed, in particular the role of global exploration and local exploitation for finding new samples. Common sampling techniques found in the literature have been reviewed. As a first investigation of this scale, thirteen of the methods have been implemented and compared by means of benchmark problems for OK.  Furthermore, for the first time, a majority of the methods has been utilized for surrogate model creation with HK. A two-dimensional Finite Element application including contact nonlinearities and two different friction force models has been employed for the validation of adaptive sampling for HK. 
In a novelty approach different adaptive sampling techniques have also been succesfully introduced and studied for their use in PLSOK and PLSHK. \\
In the following a conclusive evaluation of the studied sampling techniques is given.
\paragraph*{Evaluation of the investigated adaptive sampling techniques}
Thirteen sampling techniques have been compared on benchmark problems and application examples of different fidelity. It was found that for all presented problems the adaptive sampling techniques were able to significantly improve the accuracy of the metamodel when comparing it to the one-shot TPLHD approach. It could be noticed that EI, CVVor, SFCVT and SSA are prone to provoke numerical issues in form of clustering of points (see problems in sections: \ref{sec::Schwefel} or \ref{sec::SixHumb}). This is particularly the case in lower dimensions. Therefore it is not advised to employ these techniques for global metamodeling with aims to reach an accuracy threshold. \\
It was found that LOLA is restricted to lower dimensionality because of the computational efforts involved. In this thesis the limit is set to 2 dimensions. The same holds true for AME, whose use is not advised in large dimensions, i.e. higher than 5.  It was found that the methods that achieve proficient results for ordinary Kriging are also able to do so in the multi-fidelity case. However in HK, more methods including MASA and ACE appear to experience numerical issues by cumulating sample points in specific areas (see section \ref{sec::Forrester}). \\
The best results were generally found by methods relying on cross-validation or the variance to find new sample points. The chosen Query-by-committee technique MASA as well as the gradient-based method LOLA do not achieve the same accuracy of the metamodel with respect to the target function. In lower dimensionality CVD is a pertinent choice, i.e. yielding the best average outcome for the one-dimensional Schwefel function (section \ref{sec::Schwefel}) and the one-dimensional multi-fidelity Forrester function of section \ref{sec::Forrester}.  \\
In higher dimensions the space-filling methods MIPT and MSD as well as the adaptive techniques EIGF and MEPE should be the preferred choices when the time of surrogate model generation is of concern. Over all dimensionality and fidelity levels, MEPE seems to be the most consistent choice of adaptive sampling technique. It is able to consistently improve the metamodel in comparison to one-shot techniques. It combines exploration and exploitation in a switch strategy and has no issues with point clustering. A drawback is that the method is restricted to Kriging because the prediction variance needs to be known. After MEPE, EIGF is the second best choice. However the method predominantly samples around the highest absolute values of the target function, which is not ideal for global metamodeling. 
\paragraph*{Adaptive sampling for classification with Kriging}
A dynamic, nonlinear mass-on-belt oscillator of Duffing's type with an elastoplastic friction law has been studied with the goal to built a metamodel for the classification of the chaotic behavior of the system. Here, the largest Lyapunov exponent as an indicator for chaotic motion has been selected.  \\ However,
it has been observed that the adaptive sampling techniques found in the literature were not able to proficiently generate sample for classification problem with Kriging. In order to adaptively sample for this application, an innovative adaptive sampling technique termed MC-Intersite Voronoi (MIVor) for binary classification problems with Kriging has been introduced and successfully validated to classify chaotic behavior of a dynamic system. 
\paragraph*{Perspective}
Challenges remain for efficiently reducing the number of required samples for generating proficient surrogates. The majority of introduced methods found in the literature are high in computational effort, especially in higher dimensions. In general, discontinuous methods lack efficiency of continuous ones.
New techniques need to find a balance between exploration, exploitation and creating a fast optimization environment to find new samples. \\
The concept of PLSHK should be studied further in particular with regards to the number of required dimensions in each fidelity level. However it was found to be an effective method to reduce the computation time and the complexity of the MLE optimization problem. \\
Kriging should be further investigated as a modeling choice for generating surrogates for binary classification problems. Adaptive sampling for these applications should be explored.
MIVor could be applied to different binary classification problems. In addition the method should be tested on problems of higher dimensionality.
\clearpage
\appendix
\chapter{Stochastic background}
\section{Gaussian processes}\label{sec::GaussianProcess}
The fundamental mathematical object of random-variables is the probability space $[\Omega, \mathbb{F}^{(\Omega)}, P]$, where
\begin{itemize}
\item $\Omega$ is the sample space comprised of all possible elementary events $\omega_{i} \in \Omega$,
\item $\mathbb{F}^{(\Omega)}$ is a complete system of subsets of $\Omega$ covering all events,
\item $P$ assigns probabilities to the elements of $\mathbb{F}^{(\Omega)}$.
\end{itemize}
A real-random variable is formally defined as a a mapping between the probability space and an output space $\mathbb{T} \subseteq \mathbb{R}$ 
\begin{equation}\label{eq:randomVariable}
\begin{aligned}
T : \Omega &\rightarrow \mathbb{T} \\
\omega &\rightarrow T(\omega):= t .
\end{aligned}
\end{equation} 
A random variable is completely defined by its cumulative distribution function 
\begin{equation}
F_{T}(t) = P(T \leq t ).
\end{equation}
Continuous random variables can be described by a probability density function
\begin{equation}
f_{T}(t) = \frac{d F_{T}(t)}{d t}.
\end{equation}
The expectation value of the random variable $T$ reads
\begin{equation}
\mathbb{E}(T) = \int_{-\infty}^{\infty} t f_{T}(t) dt
\end{equation}
and the variance
\begin{equation}
Var(T) = \sigma^{2} = \mathbb{E} \left[ \left( T - \mathbb{E}(T)\right)^{2}\right] .
\end{equation}
The covariance between two random variables $T_{1}$ and $T_{2}$ is defined as
\begin{equation}
cov(T_{1}, T_{2}) =  \mathbb{E} \left[ \left( T_{1} - \mathbb{E}(T_{1})\right) \left( T_{2} - \mathbb{E}(T_{2})\right) \right].
\end{equation}
A random vector $\bm{T}$ is a collection of random variables
\begin{equation}
\begin{aligned}
\bm{T} : \Omega &\rightarrow \mathbb{T} \subseteq \mathbb{R}^{n} \\
\omega &\rightarrow \bm{T}(\omega):= \bm{t} .
\end{aligned}
\end{equation} 
A Gaussian random vector is a vector given by 
\begin{equation}
\bm{T} \sim \mathcal{N}(\bm{\mu}, \bm{\Sigma}).
\end{equation}
Here $\bm{\mu}$ is a vector containing the means and $\bm{\Sigma}$ is its covariance matrix. The variable is completely defined by the probability density function 
\begin{equation}
f_{\bm{T} } (\bm{t}) = \frac{1}{(2 \pi)^{n/2} \sqrt{\det \bm{\Sigma}}} \exp \left( - \frac{1}{2} (\bm{t} - \bm{\mu})^{t} \bm{\Sigma}^{-1} (\bm{t} - \bm{\mu})\right).
\end{equation}
Reusing the definition given in equation (\ref{eq:randomVariable}) a stochastic process $Z$ is a set of random variables indexed by a continuous value $\bm{s}$. It expands the random variable and reads as the mapping  
\begin{equation}
\begin{aligned}
Z(\bm{s}) : \mathbb{I} \times [\Omega, \mathbb{F}^{(\Omega)}, P] &\rightarrow \mathbb{T} \\
(s,\omega) &\rightarrow T(\bm{s},\omega):= t(\bm{s}) .
\end{aligned}
\end{equation} 
$t(\bm{s})$ is a realization of the process and $\mathbb{I} \in \mathbb{R}^{n}$ denotes an index space with $\bm{s}\in\mathbb{I}$.
A Gaussian random process is a random process in which for all $s_{1}, \ldots, s_{n}$ the random vector $\left( Z(s_{1}) , \ldots Z(s_{n}) \right)$ is Gaussian i.e. has a multivariate normal distribution. Therefore it can be uniquely defined by its mean function
\begin{equation}
\bm{\mu}(\bm{s}) = \mathbb{E}(Z(\bm{s}) )
\end{equation}
and its autocovariance function $\mathbb{C}$ that gives the covariance of the process with itself between two indexes $\bm{s}$ and $\bm{s'}$ as
\begin{equation}
\begin{aligned}
\mathbb{C}(\bm{s}, \bm{s}') &= cov(Z(\bm{s}),Z(\bm{s'})) \\&=\mathbb{E} \left[ \left( Z(\bm{s}) - \mathbb{E}(Z(\bm{s}))\right) \left( Z(\bm{s'}) - \mathbb{E}(Z(\bm{s'}))\right) \right].
\end{aligned}
\end{equation}
For a given $\omega_{0}$ the function $(s,\omega_{0}) \rightarrow T(\bm{s},\omega_{0})$ is a realization of the Gaussian process. \\
The autocovariance function is symmetric and (non-strictly) positive semi-definite. For the simplified case of a stationary Gaussian process (it is invariant by translation \citep{pavliotis2014stochastic}) the mean function is a constant
\begin{equation}
\mu(\bm{s}) = \mu_{0} \text{.}
\end{equation}
Furthermore the autocovariance function can be simplified to 
\begin{equation}
\mathbb{C}(\bm{s}, \bm{s}') = \sigma^{2} R (\bm{s} - \bm{s'}),
\end{equation}
where  $\sigma^{2}$ is the variance and the spatial autocorrelation function $R$ only depends on the euclidean distance between the input combinations $\bm{s}$ and $\bm{s'}$.
For more information on general probability theory refer to \cite{zio2007introduction}.  
\section{Least squares linear regression model}\label{seq::LeastLinearRegression}
Consider a given set of observations $\mathcal{D} = \lbrace \left( \bm{x}^{(i)}, \,y^{(i)} \right), \, i=1, \, \ldots  , \, m  \rbrace$ with an input $\bm{x} \in \mathbb{X} \subset \mathbb{R}^{n}$ and a scalar output $y \in \mathbb{Y} \subset \mathbb{R}$. 
The linear regression assumes that the relationship between the input and the output is linear
\begin{equation}\label{eq:LinearRegression}
Y_{i} = \sum_{j=1}^{m} \beta_{j} f_{j}(\bm{x}^{(i)}) + Z_{i}, \qquad \text{for} \, i=1, \ldots, m.
\end{equation}
Here, the unknown $\bm{\beta}$ is a $m$-dimensional parameter vector, $\bm{f}$ is a chosen $m$-dimensional collection of regressor functions and $\bm{Z}$ is a $m-$dimensional random vector that expresses the noise term. 
The equation can be summarized in the matrix equation
\begin{equation}
\bm{Y} = \bm{F} \bm{\beta} + \bm{Z},
\end{equation}
where $\bm{F}_{ij} = f_{j}(\bm{x}^{i})$.
For simplification and due to the lack of knowledge about the data the noise term is usually chosen to be a Gaussian random vector with zero mean and a covariance function 
\begin{equation}
cov(\bm{Z}, \bm{Z}) = \sigma^{2} \bm{R},
\end{equation}
with the variance $\sigma^{2}$ which needs to be estimated and a correlation matrix $\bm{R}$ which is known.
Since $\bm{Z}$ is a Gaussian vector the distribution of the observations $\bm{Y}$ can be set to be a multivariate normal distribution with 
\begin{equation}
\bm{Y} \sim \mathcal{N}_{m} (\bm{F} \bm{\beta} , \sigma^{2} \bm{R}).
\end{equation}
The optimal unknown parameters need to be estimated e.g. using a maximum likelihood approach which yields
\begin{equation}
\mathcal{L}(\bm{y}| \bm{\beta}, \sigma^{2}) = \frac{1}{(2 \pi \sigma^{2})^{m/2} \sqrt{\det \bm{R}}} \exp \left( - \frac{1}{2 \sigma^{2}} (\bm{y} - \bm{F} \bm{\beta})^{t} \bm{R}^{-1} (\bm{y} - \bm{F} \bm{\beta})\right).
\end{equation}
The best estimates of the parameters $(\hat{\bm{\beta}}, \hat{\sigma})$ can be found by maximizing the likelihood function
\begin{equation}
(\hat{\bm{\beta}}, \hat{\sigma}) \equiv \argmax_{(\bm{\beta}, \sigma)} \mathcal{L}(\bm{y} |  \bm{\beta}, \sigma^{2})
\end{equation}
which yields
\begin{equation}\label{eq::bestLinearWeights}
\hat{\bm{\beta}} = \left( \bm{F}^{T} \bm{R}^{-1} \bm{F}\right)^{-1} \bm{F}^{T} \bm{R}^{-1} \bm{y} \, \text{,}
\end{equation}
and
\begin{equation}
\hat{\sigma}^{2} = \frac{1}{m} \left( \bm{y} - \bm{F} \hat{\bm{\beta}} \right)^{T} \bm{R}^{-1} \left( \bm{y} - \bm{F} \bm{\beta} \right) \text{.}
\end{equation}
\section{Bayesian prediction methodology}
Consider a given set of uncertain data $\bm{Y} = \left( Y^{(i)} \right), \, i=1, \, \ldots  , \, m  \rbrace$ and an unobserved random variable $Y_{0}$. $\bm{Y}$ and $Y_{0}$ are dependent random variables. Therefore a predictor for the unobserved quantity should be dependent on the joint distribution of $\bm{Y}$ and $Y_{0}$. Hence it is assumed that a $y_{0}$ is a realization of a random vector distributed according to a joint parametric distribution $F \in \mathcal{F}$ 
 \begin{equation}
\begin{Bmatrix} 
 \bm{Y} \\
Y_{0}
\end{Bmatrix} \sim F \in \mathcal{F}\, \text{.}
\end{equation}
Bayesian prediction methodology as e.g. described in  \cite{santner2013design} tries to derive an estimator $\hat{Y}_{0}$ for the data point $y_{0}$ by utilizing this statistical dependency. This general framework allows $F$ to belong to a large class of probability distributions. In context of Kriging however it is enough to restrict $F$ to the multivariate Gaussian distribution. \\ Predictors may have a general form. However, to the best of the authors knowledge the mean-squared prediction error is usually employed as a prediction criterion. This allows to define the fundamental theorem of prediction \citep{santner2013design}.
\begin{mydef}
Suppose that $\begin{Bmatrix} 
 \bm{Y},
Y_{0}
\end{Bmatrix}$ has a joint distribution function $F$ for which the conditional mean of $Y_{0}$ given $\bm{Y}$ exists. Then
\begin{center}
$\hat{Y}_{0} = \mathbb{E} (Y_{0} | \bm{Y})$ 
\end{center}
is the best mean-squared prediction error of $Y_{0}$.
\end{mydef}
\section{Best linear unbiased predictor (BLUP)}\label{sec::BLUP}
Let the training data $\lbrace \left( \bm{x}^{(i)}, \,y^{(i)} \right), \, i=1, \, \ldots  , \, m  \rbrace$ be given of a blackbox function $\mathcal{M}$. Let $\bm{x}^{(0)}$ be an unobserved data point.
The goal is to predict a random variable
$Y_{0}= \mathcal{M}(\bm{x}^{(0)})$ based on training data which is assumed to be uncertain with $\bm{Y} = (Y_{1}, \ldots, Y_{m})^{T}$. Let $\hat{Y}_{0} = \hat{Y}_{0} (\bm{Y})$ be a predictor of the random variable $Y_{0}$ based on $\bm{Y}$.
Consider the regression model given in equation (\ref{eq::Regression_model}) 
\begin{equation}
\bm{Y}(\bm{x}) = \bm{F}(\bm{x}) \bm{\beta} + \bm{Z}(\bm{x}),
\end{equation}
where $\bm{F}$ is the regression matrix, $\bm{\beta} $ is the vector of weights and $Z(\bm{x})$ is a stationary Gaussian process with zero mean, of which the stationary autocovariance $\mathcal{C}$ is given by
\begin{equation}
\mathcal{C}(\bm{x}, \bm{x'}) = \sigma^{2} R (\bm{x} - \bm{x'}, \bm{\theta}),
\end{equation}
where $\sigma^{2}$ is the variance and the spatial autocorrelation function $R$ (or kernel) only depends on the euclidean distance between the input combinations $\bm{x}$ and $\bm{x'}$ and some hyperparameters $\bm{\theta}$ which are considered to be known. Then the joint distribution of $\hat{Y}_{0}$ and $\bm{Y}$ is given as the multivariate normal distribution of the form 
\begin{equation}
\begin{Bmatrix} 
\bm{Y} \\
Y_{0} 
\end{Bmatrix} \sim \mathcal{N}_{1+m} \left( 
\begin{Bmatrix} 
\bm{F} \bm{\beta} \\
\bm{f}_{0}^{T} \bm{\beta} 
\end{Bmatrix}, \sigma^{2} 
\begin{Bmatrix} 
\bm{r}_{0} & \bm{R} \\
\bm{1} & \bm{r}_{0}^{T}  
\end{Bmatrix} 
 \right),
\end{equation}
where $\bm{f}_{0}$ is the collection of regression functions evaluated at $\bm{x}^{(0)}$,
$\bm{r}_{0}$ describes the cross-correlations between $\bm{x}^{(0)}$ and each observation as 
\begin{equation}
\bm{r}_{0 \, i} = R(\bm{x}^{(0)} - \bm{x}^{(i)}, \bm{\theta}) \, \qquad i=1, \, \ldots \, , m.
\end{equation}
 $\bm{R}$ is the correlation matrix of the observations given by
\begin{equation}
\bm{R}_{i \, j} = R(\bm{x}^{(i)} - \bm{x}^{(j)}, \bm{\theta}) \, \qquad i, \, j\, =1, \, \ldots \, , m.
\end{equation}
Consider a class of estimators $\lbrace Y_{ul} \rbrace$ for $Y_{0}$ that are unbiased and linear. The best estimator $\mu_{\hat{Y}_{0}} = \mathbb{E} [\hat{Y}_{0}]$ in the sense of mean-squared out of this collection has the following properties:
\begin{itemize}
\item The estimator is linear in $\bm{Y}$
\begin{equation}
\hat{Y}_{0} = \sum_{i=1}^{m} a_{0\, i} Y_{i} = \bm{a}_{0}^{T} \bm{Y}.
\end{equation}
\item The estimator is unbiased
\begin{equation}
\mathbb{E} [\hat{Y}_{0} - Y_{0}] = 0.
\end{equation}
\item The estimator has the least expected mean-squared error i.e. prediction variance
\begin{equation}
\hat{Y}_{0} = \argmin_{Y_{0}^{\star} \in \lbrace Y_{ul} \rbrace} \qquad \mathbb{E} \left[ \left( Y_{0}^{\star} - Y_{0} \right)^{2} \right].
\end{equation}
\end{itemize}
Under these conditions the best estimator is (for proof see e.g. \cite{santner2013design}) 
\begin{equation}
\mu_{\hat{Y}_{0}}  = \hat{Y}_{0} = \bm{f}_{0}^{T} \hat{\bm{\beta}} + \bm{r}_{0}^{T} \bm{R}^{-1} (\bm{y}- \bm{F} \hat{\bm{\beta}})
\end{equation}
with the minimal variance
\begin{equation}
\sigma_{\hat{Y}_{0}}^{2}  = \sigma^{2} \left( 1 - \bm{r}_{0}^{T} \bm{R}^{-1} \bm{r}_{0} + \bm{u}_{0}^{T} \left( \bm{F}^{T} \bm{R}^{-1} \bm{F}\right)^{-1} \bm{u}_{0}\right) \, ,
\end{equation}
where $\hat{\bm{\beta}}$ is the best linear regression estimator for the weights see e.g. equation (\ref{eq::bestLinearWeights}) and 
and $\bm{u}_{0}$ is given by
\begin{equation}
\bm{u}_{0} = \bm{F}^{T} \bm{R}^{-1} \bm{r}_{0} - \bm{f}_{0} \text{.}
\end{equation}

\bibliographystyle{plainnat}
\addcontentsline{toc}{chapter}{Bibliography}
\bibliography{bibShort}
\end{document}